\documentclass[11pt]{report}
\usepackage[T1]{fontenc}    % use 8-bit T1 fonts
\usepackage[utf8]{inputenc}
\usepackage{fullpage}

\usepackage[en-US, useregional, showseconds=true, showzone=false]{datetime2}
\usepackage{macros/general_def}
\usepackage{macros/shortcuts}
\begin{document}
\title{On Principled Local Optimization Methods\\for Federated Learning} 

\date{May 2022}
\author{%
  Honglin Yuan \\
  Stanford University \\
  \texttt{yuanhl@alumni.stanford.edu}
  }
\maketitle

\begin{abstract}
    Federated Learning (FL), a distributed learning paradigm that scales on-device learning collaboratively, has emerged as a promising approach for decentralized AI applications. 
% FL has gained increasing popularity due to its communication efficiency, massive decentralized computations, agile personalized service, and privacy preservation. 
Local optimization methods such as Federated Averaging (\fedavg) are the most prominent methods for FL applications. 
Despite their simplicity and popularity, the theoretical understanding of local optimization methods is far from clear. 
This dissertation aims to advance the theoretical foundation of local  methods in the following three directions.

First, we establish sharp bounds for \fedavg, the most popular algorithm in Federated Learning. 
We demonstrate how \fedavg may suffer from a notion we call iterate bias, and how an additional third-order smoothness assumption may mitigate this effect and lead to better convergence rates. We explain this phenomenon from a Stochastic Differential Equation (SDE) perspective. 

Second, we propose \fedacfull (\fedac), the first principled acceleration of \fedavg, which provably improves the convergence rate and communication efficiency.  Our technique uses on a potential-based perturbed iterate analysis, a novel stability analysis of generalized accelerated SGD, and a strategic tradeoff between acceleration and stability.

Third, we study the Federated Composite Optimization problem, which extends the classic smooth setting by incorporating a shared non-smooth regularizer. We show that direct extensions of  \fedavg may suffer from the ``curse of primal averaging,'' resulting in slow convergence. As a solution, we propose a new primal-dual algorithm, Federated Dual Averaging, which overcomes the curse of primal averaging by employing a novel inter-client dual averaging procedure.
\end{abstract}

\chapter{Introduction}
The advances of machine learning, the proliferation of mobile and IoT (Internet of Things) devices, and the rapid development of communication technology led to a boom of various on-device intelligence applications, such as connected autonomous vehicles, smart homes, wearable devices, and mobile AI.
It is crucial to securely and efficiently leverage the massively distributed data to succeed in these burgeoning tasks.
Federated Learning (FL), an emerging distributed learning paradigm that scales on-device learning collaboratively, has gained increasing popularity due to its communication efficiency, massive decentralized computations, agile personalized service, and privacy preservation \cite{Hard.Rao.ea-18,Hartmann.Suh.ea-19,Hard.Partridge.ea-INTERSPEECH20}. 
% Federated Learning techniques have been successfully applied in a broad range of practical applications 

% \hynote{maybe mentino that you only focus on optimization, and mention other possible concerns, privacy, etc, and cite some papers}

% \hynote{Perhaps broader discussion here}
Federated Learning is orchestrated by a central server who oversees the clients possessing data, e.g., mobile devices or a group of organizations. 
A typical FL process involves a series of alternate training and communication rounds. 
During the training round, each client performs local training by consuming its local data.
The client models are aggregated by the orchestration server during the communication round and then broadcast to the clients.
We refer to this type of process as the \emph{local optimization method}.
For example, Federated Averaging (\fedavg, \cite{McMahan.Moore.ea-AISTATS17}), also known as Local SGD or Parallel SGD \cite{Mangasarian-95,Zinkevich.Weimer.ea-NIPS10,Coppola-14,Zhou.Cong-IJCAI18}, applies SGD on local data for local training, and occasionally aggregate the information by averaging the parameters.
% the most prominent algorithm in Federated Learning

One of the reputed advantages of the local optimization methods is the potential to improve communication efficiency since the client models may be aggregated infrequently.
The communication process is widely acknowledged as the major performance bottleneck of FL applications due to the vast number of participants, the relatively large model size, and the unreliable connection of client devices. 
Unlike the classic datacenter setting where compute nodes and backbone networks are powerful and robust, the devices in FL are usually battery-powered and wirelessly connected. 
Communication over such devices is costly, unstable, and subject to high latency. 
Hence, understanding and improving communication efficiency have been one of the primary questions since the inception of FL. 

Despite the simplicity and popularity of local optimization methods, a thorough theoretical understanding has not been established. 
% Even under the simplest assumptions the best known upper and lower bounds of \fedavg do not match, and it is not clear whether the existing analysis captures the capacity of the algorithm. 
It is not clear whether, when, and why local optimization methods may provably improve communication efficiency. 
In this dissertation, we aim to advance the theoretical foundation of local optimization methods by 
\begin{enumerate}
  \item Establishing sharp understanding of the existing FL algorithms.
  \item Improving the efficiency of FL by principled acceleration.
  \item Extending FL algorithms to more general, regularized settings.
\end{enumerate}
% The training process combines local training of a model at the clients with infrequent aggregation of the locally trained models at the central server. 
% \hynote{TM: maybe mention what are the typical metrics for success, optimization speed, communication etc?}

\section{Sharp Bounds for Federated Averaging and Continuous Perspective}
In \cref{chapter:fedavg}, we establish sharp lower bounds for homogeneous and heterogeneous $\fedavg$, the most prominent local method in Federated Learning.
By solving this open problem, we highlight the obstacles to $\fedavg$, and show how $\fedavg$ can converge faster on problems with third-order smoothness.
This chapter is based on a joint work with Margalit Glasgow and Tengyu Ma, published in AISTATS 2022 \cite{Glasgow.Yuan.ea-AISTATS22}.

% While local methods such as \fedavg are popular in practice,
% The understanding of local optimization methods such as \fedavg is one of the most important topics in distributed optimization.
% For instance, Federated Averaging applies parallelized SGD and periodically synchronizes across clients by simple averaging.
% The distributed nature poses novel challenges to Federated Learning study. 
% For instance, Federated Averaging applies parallelized SGD and periodically synchronizes across clients by simple averaging.
% The distributed and private nature of FL poses a series of novel challenges. 
% The two most distinct features of FL, when compared to classic distributed learning settings, are (1) very high cost to communicate with a client (2) heterogeneity in data amongst the clients. 
% \hynote{Perhaps missing a paragraph here.}
% Federated Averaging (\fedavg), also known as Local SGD, is one of the most popular algorithms in Federated Learning. 
% Despite its simplicity and popularity, the convergence rate of \fedavg has thus far been undetermined. 
% Even under the simplest assumptions (convex, smooth, homogeneous, and bounded covariance), the best-known upper and lower bounds do not match, and it is not clear whether the existing analysis captures the capacity of the algorithm. 
The characterization of local optimization methods such as \fedavg is one of the most important topics in distributed optimization. 
Numerous existing works have aimed to determine the convergence rate of \fedavg in various settings \cite{Stich-ICLR19,Li.Huang.ea-ICLR20,Stich.Karimireddy-arXiv19,Khaled.Mishchenko.ea-AISTATS20,Woodworth.Patel.ea-ICML20,Woodworth.Patel.ea-NeurIPS20}, though early analysis of \fedavg preceded the proposal of Federated Learning, typically under the name of Local SGD or parallel SGD \cite{Mcdonald.Mohri.ea-NIPS09,Zinkevich.Weimer.ea-NIPS10,Shamir.Srebro-Allerton14,Rosenblatt.Nadler-16,Jain.Kakade.ea-COLT18,Zhou.Cong-IJCAI18}. 
The primary focus of early literature is the special case of one-shot averaging, in which only one round of averaging (communication) is conducted at the end of the procedure. 
Despite the joint efforts, even under the simplest setting (convex, smooth, homogeneous, and bounded covariance; see \cref{asm:fo:2o}), the state-of-the-art upper bounds for \fedavg due to \cite{Khaled.Mishchenko.ea-AISTATS20} and \cite{Woodworth.Patel.ea-ICML20} do not match the state-of-the-art lower bound due to \cite{Woodworth.Patel.ea-ICML20}. 
% \hynote{Remove table?}
This gap suggests that at least one side of the analysis is not sharp. 
Therefore, a fundamental question remains:

\emph{Does the current convergence analysis of \fedavg fully capture the capacity of the algorithm?}

% In \cref{chapter:fedavg}, we resolve this question by providing a lower bound for \fedavg that matches the existing upper bound, which shows the existing \fedavg upper bound analysis is not improvable. Additionally, we establish a lower bound in a heterogeneous setting that nearly matches the existing upper bound.
Our first contribution is to answer this question definitively under the aforementioned assumptions. 
In \cref{sec:fedavg:2o:lb}, we establish a sharp lower bound for \fedavg that matches the existing upper bound (\cref{thm:fedavg:lb:homo}), showing that the existing \fedavg analysis is \emph{not} improvable.
% \hynote{at least 1 ouf of the above 3 sentences can be removed}
Moreover, we establish a stronger lower bound in the \emph{heterogeneous} setting, \cref{thm:fedavg:lb:hetero}, which suggests the best-known \emph{heterogeneous} upper bound analysis \cite{Woodworth.Patel.ea-NeurIPS20} is also (almost)\footnote{Up to a minor variation of the definition of heterogeneity measure; see \cref{rem:hetero}.} not improvable.

% The improved lower bound results seem discouraging, but fortunately o
Our proofs highlight the limitation of \fedavg, yielding these slow convergence rates. 
% \tnote{Maybe I didn't get the logic here; why ``highlighting exactly what can go wrong'' make it less discouraging?}\mg{I see. This hopefully becomes clear in later paragraphs. We could either spend 2 sentences explaining why its useful to understand the obstacles in order to model better assumptions or we could just cut the first half of the sentence?}
In \cref{sec:fedavg:2o:bias}, we show that our lower bound analysis stems from a notion we call \emph{iterate bias}, which is defined by the deviation of the expectation of the SGD trajectory from the (noiseless) gradient descent trajectory with the same initialization (see \cref{def:fedavg:bias} for details).
% \hynote{this sentence is too long. should we just formally define it?}
We show that even for convex and smooth objectives, the mean of SGD initialized at the optimum can drift away from the optimum at the rate of $\Theta(\eta^2 k^{\frac{3}{2}})$ after $k$ steps for any sufficiently small learning rate $\eta$. 
This rate is also sharp according to our matching upper and lower bounds; see \cref{thm:fedavg:2o:bias:ub,thm:fedavg:2o:bias:lb} for details.
% see \cref{thm:fedavg:2o:bias:ub,thm:fedavg:2o:bias:lb} for details \hynote{Maybe emphasize this is matching bound}.
% Under standard \cref{asm:fo:2o}, we show (via matching upper bound \cref{thm:fedavg:2o:bias:ub} and lower bound \cref{thm:fedavg:2o:bias:lb}) that the iterate bias can be as large as $\Theta(\eta^2 k^{\frac{3}{2}})$ after $k$ steps, for sufficiently small learning rate $\eta$. 
The iterate bias thus quantifies the fundamental difficulty encountered by \fedavg: 

\emph{Even with infinite number of homogeneous clients, \fedavg can drift away from the optimum even if initialized at the optimum. }

% Indeed, we show that the sharp lower bound of SGD iterate bias leads directly to our sharp lower bound of \fedavg convergence rate.

The discouraging lower bound of $\fedavg$ % under a standard smoothness assumption
does not conform well with its empirical efficiency observed in practice \cite{Lin.Stich.ea-ICLR20}. 
This motivates us to consider whether additional modeling assumptions could better explain the empirical performance of \fedavg. 
The aforementioned lower bound is attained by a special piece-wise quadratic function with a sudden curvature change, which is smooth (with bounded second-order derivatives) but has unbounded third-order derivatives. A natural assumption to exclude this corner case is the third-order smoothness, which may be representative of objectives in practice. 
For instance, loss functions used to learn many generalized linear models, such as logistic regression, often exhibit third-order smoothness \cite{Hastie.Tibshirani.ea-09}. 

With this additional third-order smoothness assumption, we show in \cref{sec:fedavg:3o:bias} that the iterate bias reduces to $\Theta(\eta^3 k^2)$ after $k$ steps, one order higher in $\eta$ than the rate under only second-order smoothness. 
This rate is sharp according to our matching upper and lower bounds; see \cref{thm:fedavg:3o:bias:ub,thm:fedavg:3o:bias:lb}.
While the proofs for bounding the iterate bias are quite technical, we show that it is easy to analyze the bias via a continuous approach. 
More specifically, by studying the stochastic differential equation (SDE) corresponding to the continuous limit of SGD, one can derive the limit of the iterate bias of generic objectives by using the Kolmogorov backward equation of the SDE; see \cref{sec:sde}.

Leveraging this intuition from the iterate bias, we prove state-of-the-art rates for \fedavg under third-order smoothness in \emph{both} convex (\cref{sec:fedavg:3o:ub}) and non-convex (\cref{sec:fedavg:ncvx}) settings. 
In non-convex settings, our convergence rate scales with $1/R^{\frac{4}{5}}$, which improves upon the best-known rate of $1/R^{\frac{2}{3}}$ \cite{Yu.Yang.ea-AAAI19} if we do not assume third-order smoothness. 
The specialty of quadratic objectives for better efficiency has been noted in various contexts \cite{Zhang.Duchi.ea-JMLR15,Jain.Kakade.ea-JMLR18,Woodworth.Patel.ea-ICML20}. 
% \cite{Woodworth.Patel.ea-ICML20} studied an acceleration of \fedavg but was limited to quadratic objectives.
Our results give a smooth interpolation of the results of \cite{Woodworth.Patel.ea-ICML20} for quadratic objectives to broader function class. 
% More generally, \cite{Dieuleveut.Patel-NeurIPS19} studied the convergence of \fedavg under bounded \nth{3}-derivative, but the bounds are still dominated by minibatch SGD baseline \cite{Woodworth.Patel.ea-ICML20}. 
% Recent work by \cite{Godichon-Baggioni.Saadane-20} studied one-shot averaging under similar assumptions. 
% \hynote{CONTEXT}

It is possible to view the iterate bias as an implicit bias of the \fedavg algorithm, which pushes the iterate towards flatter regions of the objective. 
This effect is similar to other instances of implicit bias observed for stochastic gradient descent, which has drawn connections between noise in the gradients and flat minima \cite{Hochreiter.Schmidhuber-97,Jastrzebski.Kenton.ea-ICANN18,Blanc.Gupta.ea-COLT20,Damian.Ma.ea-NeurIPS21}. 
While in many instances, implicit bias has been linked to choosing favorable optima that generalize well \cite{Neyshabur-17}, in our setting, the bias affects the convergence rate. 
The existence and effect of iterate bias have been observed in various forms in the current literature \cite{Dieuleveut.Durmus.ea-20,Charles.Konecny-20,Woodworth.Patel.ea-ICML20}, yet our work is the first to sharply characterize the rate of the bias, both in the second-order smooth case and third-order smooth case.

% \hynote{
% For non-convex objectives, a series of recent works \cite{Zhou.Cong-IJCAI18,Haddadpour.Kamani.ea-ICML19,Wang.Joshi-JMLR21,Yu.Jin-ICML19,Yu.Jin.ea-ICML19} has established various upper bounds of \fedavg in homogeneous and heterogeneous settings. 
% To the best of our knowledge, we are unaware of any lower bound for \fedavg in non-convex settings.
% }

\section{Principled Acceleration of Federated Averaging}
In \cref{chapter:fedac}, we study the acceleration of \fedavg and investigate whether it is possible to improve convergence speed and communication efficiency. 
This chapter is baed on a joint work with Tengyu Ma, published in NeurIPS 2020 \cite{Yuan.Ma-NeurIPS20}.

We propose \fedacfull (\fedac), a principled acceleration of Federated Averaging.
\fedac is the first provable acceleration of \fedavg that improves convergence speed and communication efficiency on various types of convex functions.
For example, for strongly convex and smooth functions, when using $M$ clients, the previous state-of-the-art \fedavg analysis can achieve a linear speedup in $M$ if given $\tildeo(M)$ rounds of synchronization, whereas \fedac only requires $\tildeo(M^{\frac{1}{3}})$ rounds.  
Moreover, we prove stronger guarantees for \fedac when the objectives are third-order smooth. 
Our technique is based on a potential-based perturbed iterate analysis, a novel stability analysis of generalized accelerated SGD, and a strategic tradeoff between acceleration and stability.

Our results suggest an intriguing synergy between acceleration and parallelization.
In the single-client sequential setting, the convergence is usually dominated by the term related to stochasticity,
which is generally not possible to be accelerated \cite{Nemirovski.Yudin-83}.
In distributed settings, the communication efficiency is dominated by the overhead caused by infrequent synchronization, which can be accelerated.% as we show in the convergence rates summary \cref{tab:conv:rate}. 

The main challenge for introducing acceleration to \fedavg lies in the conflict between acceleration and stability. 
Stability is essential for analyzing distributed algorithms such as \fedavg, whereas momentum applied for acceleration may amplify the instability of the algorithm. 
% Indeed, we show that standard Nesterov's accelerated gradient descent algorithm \cite{Nesterov-18} \emph{may not be initial-value stable even for smooth and strongly convex functions}, in the sense that the initial infinitesimal difference may grow exponentially fast (see \cref{thm:agd:instability}). 
% \hynote{
% We identify how stability is connected to the efficiency of federated algorithms. 
In general, stability is one important topic in machine learning and has been studied for a variety of purposes \cite{Yu.Kumbier-20}. For example, \cite{Bousquet.Elisseeff-JMLR02,Hardt.Recht.ea-ICML16} showed that algorithmic stability could be used to establish generalization bounds.
\cite{Chen.Jin.ea-arXiv18} provided stability bound of standard Accelerated Gradient Descent (AGD) for \emph{quadratic objectives}. 
To the best of our knowledge, there is no existing (positive or negative) result on the stability of AGD for general convex or strongly convex objectives. 
This work provides the first (negative) result on the stability of standard deterministic AGD, which suggests that standard AGD may not be initial-value stable even for strongly convex and smooth objectives; see \cref{thm:agd:instability}.
This evidence necessitates a more scrutinized acceleration in distributed settings, and may be of broader interest.\footnote{
We construct the counterexample for initial-value stability for simplicity and clarity. 
We conjecture that our counterexample also extends to other algorithmic stability notions (\eg, uniform stability \cite{Bousquet.Elisseeff-JMLR02}) since initial-value stability is usually milder than the others.
} 
The tradeoff technique of \fedac also provides a possible remedy to mitigate the instability issue, which may be applied to derive better generalization bounds for momentum-based methods.
% }

We empirically demonstrate the efficiency of \fedac in \cref{sec:experiment}. 
Numerical results suggest a considerable improvement of \fedac over all three baselines, namely \fedavg, (distributed) Minibatch-SGD, and (distributed) Accelerated Minibatch-SGD \cite{Dekel.Gilad-Bachrach.ea-JMLR12,Cotter.Shamir.ea-NeurIPS11}, especially in the regime of highly infrequent communication and abundant clients.

\section{Federated Composite Optimization}
In \cref{chapter:fco}, we study the \emph{Federated Composite Optimization} (FCO) problem, in which the loss function contains a non-smooth regularizer. 
Existing FL research, as in \cref{chapter:fedavg,chapter:fedac}, primarily focuses on \emph{unconstrained smooth} objectives.
However, many FL applications in practice involve non-smooth objectives. 
Such problems arise naturally in FL applications that involve sparsity, low-rank, monotonicity, or more general constraints. 
This chapter is based on a joint work with Manzil Zaheer and Sashank Reddi, published in ICML 2021 \cite{Yuan.Zaheer.ea-ICML21}.

% Existing FL research, as in \cref{chapter:fedavg,chapter:fedac}, primarily focuses on \emph{unconstrained smooth} objectives.
% However, many FL applications in practice involve non-smooth objectives. 
% Such problems arise naturally in the context of regularization (e.g., sparsity, low-rank, monotonicity, or additional constraints on the model). 

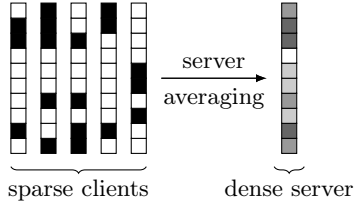
\begin{figure}%[b]
      \centering
      % !TEX root = main.tex  
\begin{tikzpicture}
  \filldraw[fill=black!0] (0.0, 0.0) rectangle (0.2, 0.2);
\filldraw[fill=black!100] (0.0, 0.2) rectangle (0.2, 0.4);
\filldraw[fill=black!0] (0.0, 0.4) rectangle (0.2, 0.6000000000000001);
\filldraw[fill=black!0] (0.0, 0.6000000000000001) rectangle (0.2, 0.8);
\filldraw[fill=black!0] (0.0, 0.8) rectangle (0.2, 1.0);
\filldraw[fill=black!0] (0.0, 1.0) rectangle (0.2, 1.2);
\filldraw[fill=black!0] (0.0, 1.2000000000000002) rectangle (0.2, 1.4000000000000001);
\filldraw[fill=black!100] (0.0, 1.4000000000000001) rectangle (0.2, 1.6);
\filldraw[fill=black!100] (0.0, 1.6) rectangle (0.2, 1.8);
\filldraw[fill=black!0] (0.0, 1.8) rectangle (0.2, 2.0);
\filldraw[fill=black!100] (0.4, 0.0) rectangle (0.6000000000000001, 0.2);
\filldraw[fill=black!0] (0.4, 0.2) rectangle (0.6000000000000001, 0.4);
\filldraw[fill=black!0] (0.4, 0.4) rectangle (0.6000000000000001, 0.6000000000000001);
\filldraw[fill=black!100] (0.4, 0.6000000000000001) rectangle (0.6000000000000001, 0.8);
\filldraw[fill=black!0] (0.4, 0.8) rectangle (0.6000000000000001, 1.0);
\filldraw[fill=black!0] (0.4, 1.0) rectangle (0.6000000000000001, 1.2);
\filldraw[fill=black!0] (0.4, 1.2000000000000002) rectangle (0.6000000000000001, 1.4000000000000001);
\filldraw[fill=black!100] (0.4, 1.4000000000000001) rectangle (0.6000000000000001, 1.6);
\filldraw[fill=black!100] (0.4, 1.6) rectangle (0.6000000000000001, 1.8);
\filldraw[fill=black!100] (0.4, 1.8) rectangle (0.6000000000000001, 2.0);
\filldraw[fill=black!100] (0.8, 0.0) rectangle (1.0, 0.2);
\filldraw[fill=black!100] (0.8, 0.2) rectangle (1.0, 0.4);
\filldraw[fill=black!0] (0.8, 0.4) rectangle (1.0, 0.6000000000000001);
\filldraw[fill=black!100] (0.8, 0.6000000000000001) rectangle (1.0, 0.8);
\filldraw[fill=black!0] (0.8, 0.8) rectangle (1.0, 1.0);
\filldraw[fill=black!0] (0.8, 1.0) rectangle (1.0, 1.2);
\filldraw[fill=black!0] (0.8, 1.2000000000000002) rectangle (1.0, 1.4000000000000001);
\filldraw[fill=black!100] (0.8, 1.4000000000000001) rectangle (1.0, 1.6);
\filldraw[fill=black!0] (0.8, 1.6) rectangle (1.0, 1.8);
\filldraw[fill=black!0] (0.8, 1.8) rectangle (1.0, 2.0);
\filldraw[fill=black!0] (1.2000000000000002, 0.0) rectangle (1.4000000000000001, 0.2);
\filldraw[fill=black!100] (1.2000000000000002, 0.2) rectangle (1.4000000000000001, 0.4);
\filldraw[fill=black!0] (1.2000000000000002, 0.4) rectangle (1.4000000000000001, 0.6000000000000001);
\filldraw[fill=black!0] (1.2000000000000002, 0.6000000000000001) rectangle (1.4000000000000001, 0.8);
\filldraw[fill=black!0] (1.2000000000000002, 0.8) rectangle (1.4000000000000001, 1.0);
\filldraw[fill=black!0] (1.2000000000000002, 1.0) rectangle (1.4000000000000001, 1.2);
\filldraw[fill=black!0] (1.2000000000000002, 1.2000000000000002) rectangle (1.4000000000000001, 1.4000000000000001);
\filldraw[fill=black!0] (1.2000000000000002, 1.4000000000000001) rectangle (1.4000000000000001, 1.6);
\filldraw[fill=black!100] (1.2000000000000002, 1.6) rectangle (1.4000000000000001, 1.8);
\filldraw[fill=black!100] (1.2000000000000002, 1.8) rectangle (1.4000000000000001, 2.0);
\filldraw[fill=black!0] (1.6, 0.0) rectangle (1.8, 0.2);
\filldraw[fill=black!0] (1.6, 0.2) rectangle (1.8, 0.4);
\filldraw[fill=black!100] (1.6, 0.4) rectangle (1.8, 0.6000000000000001);
\filldraw[fill=black!0] (1.6, 0.6000000000000001) rectangle (1.8, 0.8);
\filldraw[fill=black!100] (1.6, 0.8) rectangle (1.8, 1.0);
\filldraw[fill=black!100] (1.6, 1.0) rectangle (1.8, 1.2);
\filldraw[fill=black!0] (1.6, 1.2000000000000002) rectangle (1.8, 1.4000000000000001);
\filldraw[fill=black!0] (1.6, 1.4000000000000001) rectangle (1.8, 1.6);
\filldraw[fill=black!0] (1.6, 1.6) rectangle (1.8, 1.8);
\filldraw[fill=black!0] (1.6, 1.8) rectangle (1.8, 2.0);
\filldraw[fill=black!40.0] (3.6, 0.0) rectangle (3.8000000000000003, 0.2);
\filldraw[fill=black!60.0] (3.6, 0.2) rectangle (3.8000000000000003, 0.4);
\filldraw[fill=black!20.0] (3.6, 0.4) rectangle (3.8000000000000003, 0.6000000000000001);
\filldraw[fill=black!40.0] (3.6, 0.6000000000000001) rectangle (3.8000000000000003, 0.8);
\filldraw[fill=black!20.0] (3.6, 0.8) rectangle (3.8000000000000003, 1.0);
\filldraw[fill=black!20.0] (3.6, 1.0) rectangle (3.8000000000000003, 1.2);
\filldraw[fill=black!0.0] (3.6, 1.2000000000000002) rectangle (3.8000000000000003, 1.4000000000000001);
\filldraw[fill=black!60.0] (3.6, 1.4000000000000001) rectangle (3.8000000000000003, 1.6);
\filldraw[fill=black!60.0] (3.6, 1.6) rectangle (3.8000000000000003, 1.8);
\filldraw[fill=black!40.0] (3.6, 1.8) rectangle (3.8000000000000003, 2.0);

  \draw[decoration={brace,mirror,raise=5pt},decorate]
  (0,0) -- node[below=6pt] {\footnotesize sparse clients} (1.8,0);

  \draw[-latex] (2,1.0) -- node[above=0pt] {\footnotesize server} (3.4,1.0);
  \draw[-latex] (2,1.0) -- node[below=0pt] {\footnotesize averaging} (3.4,1.0);

  \draw[decoration={brace,mirror,raise=5pt},decorate]
  (3.5,0) -- node[below=6pt] {\footnotesize dense server} (3.9,0);
\end{tikzpicture}
    \caption{\textbf{Illustration of ``curse of primal averaging''}. While each client of \fedmid can locate a sparse solution, simply averaging them will yield a much denser solution on the server side.}
    \label{fig:curse_of_avg}
  \end{figure}

  Standard FL algorithms such as \fedavg and its variants (e.g., \cite{Li.Sahu.ea-MLSys20,Karimireddy.Kale.ea-ICML20}) are primarily tailored to \emph{smooth unconstrained} settings, and are therefore, not well-suited for FCO. 
  % Recall that in \fedavg, each client runs a local thread of SGD and periodically synchronizes with the orchestration server. %(see \cref{sec:fco:fedavg} for a formal review).
  The most straightforward extension of \fedavg towards FCO is to apply local subgradient method \cite{Shor-85} in lieu of SGD. 
  This approach is largely ineffective due to the intrinsic slow convergence of subgradient approaches \cite{Boyd.Xiao.ea-03}.
  %, which is also demonstrated in \cref{fig:haxby:simplified} (marked \fedavg ($\partial$)).
  A more natural extension of \fedavg is to replace the local SGD with proximal SGD (\cite{Parikh.Boyd-FnT14}, a.k.a. projected SGD for constrained problems), or more generally, mirror descent \cite{Duchi.Shalev-shwartz.ea-COLT10}. 
  We refer to this algorithm as \emph{\fedmidfull} (\fedmid, see \cref{alg:fedmid}). 
  The most noticeable drawback of primal-averaging methods like \fedmid is the ``curse of primal averaging,'' where the desired regularization of FCO may be rendered completely ineffective due to the server averaging step typically used in FL. For instance, consider an $\ell_1$-regularized logistic regression setting --- although each client is able to obtain a sparse solution, simply averaging the client states will inevitably yield a dense solution.
  See \cref{fig:curse_of_avg} for an illustrative example.
  
To overcome this challenge, we propose a novel primal-dual algorithm named \emph{\feddualavgfull} (\feddualavg, see \cref{alg:feddualavg}).
Unlike \fedmid (or its precursor \fedavg), the server averaging step of \feddualavg operates in the dual space instead of the primal. 
Locally, each client runs a dual averaging algorithm \cite{Nesterov-MP09} by tracking a pair of primal and dual states.
During communication, the dual states are averaged across the clients.
Thus, \feddualavg employs a novel double averaging procedure --- averaging of dual states across clients (as in \fedavg), and the averaging of gradients in dual space (as in the sequential dual averaging). 
Since both levels of averaging operate in the dual space, we can show that \feddualavg provably overcomes the curse of primal averaging.
Specifically, we prove that \feddualavg can attain significantly lower communication complexity; see \cref{sec:fco:theory}. % when deployed with a large client learning rate.  

We demonstrate the empirical performance of \fedmid and \feddualavg on various tasks in \cref{sec:fco:expr}, including $\ell_1$-regularization, nuclear-norm regularization, and various constraints in FL.

\section{Additional Related Work}

Throughout this dissertation, we mostly focus on the simplest form of each algorithm. % to keep our efforts focused.
There are many other extensions applied in practice. 
For example, instead of letting all the clients participate in computation, one may randomly draw a subset of clients to participate in every round. 
% \hynote{how generalize?}
Most of our results (e.g., all of the homogeneous results) can be directly extended to this sub-sampling variant. 
Other variants of \fedavg include letting clients run different numbers of steps per round, or averaging the client states non-uniformly.
We expect the proposed techniques can shed light on the analysis of other federated algorithms and aid the design of more efficient federated algorithms. 

Many other techniques have been studied to improve the efficiency of FL algorithms.
% Besides the $\fedavg$ framework, there are many other federated optimization algorithms that aim to improve communication efficiency \cite{Reddi.Charles.ea-ICLR21}. 
For example, researchers have studied how to compress the model updates by sparsification and quantization, which reduces the communication cost per round \cite{Alistarh.Grubic.ea-NIPS17,Wen.Xu.ea-NIPS17,Stich.Cordonnier.ea-NeurIPS18,Basu.Data.ea-NeurIPS19,Mishchenko.Gorbunov.ea-arXiv19,Reisizadeh.Mokhtari.ea-AISTATS20}. 
These compression-based approaches naturally complement our studies on efficient optimization. 
In the deep learning context, a recent array of works has studied the alternative approaches of model ensembling beyond simple averaging in parameter space \cite{Bistritz.Mann.ea-NeurIPS20,He.Annavaram.ea-NeurIPS20,Lin.Kong.ea-NeurIPS20,Chen.Chao-ICLR21,Yoon.Shin.ea-ICLR21}. 

In FL application, the dataset usually exhibits heterogeneity across clients.
That is, the client datasets do not follow the same distribution. 
People observed that data heterogeneity might cause performance degradation in practice \cite{Hsu.Qi.ea-arXiv19}. 
% In \cref{chapter:fedavg}, we provide sharp characterization on the tolerance of \fedavg to data heterogeneity. 
% However, \fedavg does not actively account for heterogeneity at the first place.
Numerous existing works have aimed to mitigate the negative effect of heterogeneity in various ways \cite{Mohri.Sivek.ea-ICML19,Liang.Shen.ea-19,Chen.Chen.ea-NeurIPS20,Deng.Kamani.ea-20,Li.Sahu.ea-20,Reisizadeh.Mokhtari.ea-AISTATS20,Wang.Liu.ea-AAAI19,Pathak.Wainwright-NeurIPS20,Zhang.Hong.ea-TSP21,Yuan.Zaheer.ea-ICML21,Acar.Zhao.ea-ICLR21,Al-Shedivat.Gillenwater.ea-ICLR21,Yuan.Morningstar.ea-ICLR22}. 
In practice, the system heterogeneity will also affect the performance of Federated Learning \cite{Smith.Chiang.ea-NIPS17,Diao.Ding.ea-ICLR21}. 

This dissertation mainly focuses on the classic FL settings in which the same model is learned from and deployed to all the clients. 
An alternative setup in FL, known as the \emph{personalized} setting, aims to learn a different (personalized) model for different clients or different groups of clients. 
Numerous recent papers have proposed Federated Learning models and algorithms to accommodate personalization, such as multi-task objectives  \cite{Smith.Chiang.ea-NIPS17,Hanzely.Hanzely.ea-NeurIPS20,T.Dinh.Tran.ea-NeurIPS20,Deng.Kamani.ea-20}, and meta-learning objectives \cite{Fallah.Mokhtari.ea-NeurIPS20,Chen.Luo.ea-19,Jiang.Konecny.ea-19}.

This dissertation is mostly concerned with the optimization perspective of Federated Learning. 
Another important research topic of FL is to enforce privacy preservation. 
Local optimization methods such as \fedavg provide a ritual layer of privacy since the training data are not directly exposed to the public. 
Nevertheless, researchers have found that local methods alone are not sufficient to guarantee privacy, since attackers may recover the private data from the model parameters sent to the server. 
Various techniques have been proposed to enhance the privacy of Federated Learning, such as secure server aggregation and integration with differential privacy \cite{Dwork-08}. 
Another related research topic is the robustness to (unintentional) failures and (intentional) attacks. 
Robustness is particularly crucial for FL due to its distributed and private nature, as the server does not have access to verify the client data. 
For example, a malfunctioning client with faulty data could feed erratic model updates to the central server and contaminate the shared model. 
Since modern ML models (especially deep network) demonstrates vulnerability to data-poisoning attacks \cite{Liu.Ma.ea-NDSS18}, attackers can backdoor the FL model by feeding poisoned data to the server via a single compromised client during training time \cite{Bhagoji.Chakraborty.ea-ICML19}. 
There are numerous works that attempt to defend against attacks in FL \cite{Pillutla.Kakade.ea-19,Xie.Koyejo.ea-ECML-PKDD19,Xie.Koyejo.ea-ICML19}.
We refer readers to \cite{Kairouz.McMahan.ea-FnT21} for more detailed discussions on these topics.
% Even worse, since the , a compromised client can send poisoned model updates directly to backdoor the training process \citep{Bagdasaryan.Veit.ea-AISTATS20}.

Analogous to FL, a related distributed setting is the \emph{decentralized consensus optimization}, also known as 
\emph{multi-agent optimization} or 
\emph{optimization over networks} in the literature \cite{Nedich-FnT15}. 
Unlike the federated settings, in decentralized consensus optimization, each client can communicate every iteration, but the communication is limited to its graphic neighborhood.
Standard algorithms for unconstrained consensus optimization include decentralized (sub)gradient methods \cite{Nedic.Ozdaglar-TACON09,Yuan.Ling.ea-SIOPT16} 
and EXTRA \cite{Shi.Ling.ea-SIOPT15,Mokhtari.Ribeiro-JMLR16}. 
For constrained or composite consensus problems, people have studied both mirror-descent type methods (with primal consensus), {e.g.}, \cite{SundharRam.Nedic.ea-JOTA10,Shi.Ling.ea-TSP15,Rabbat-CAMSAP15,Yuan.Hong.ea-Automatica18,Yuan.Hong.ea-TACON20}; and dual-averaging type methods (with dual consensus), {e.g.,} \cite{Duchi.Agarwal.ea-TACON12,Tsianos.Lawlor.ea-CDC12,Tsianos.Rabbat-ACC12,Liu.Chen.ea-TSP18}.
In particular, the distributed dual averaging \cite{Duchi.Agarwal.ea-TACON12} has gained great popularity since its dual consensus scheme elegantly handles the constraints, and overcomes the technical difficulties of primal consensus, as noted by the original paper.
We identify that while the federated settings share certain backgrounds with the decentralized consensus optimization, the motivations, techniques, challenges, and results are quite dissimilar due to the fundamental difference of communication protocol, as noted by \cite{Kairouz.McMahan.ea-FnT21}.
We refer readers to \cite{Nedich-FnT15} for a more detailed introduction to the classic decentralized consensus optimization.

\section{Notations}
Let $[n]$ denote the set $\{1, \ldots, n\}$. 
We use bold lower-case characters to denote vectors (\eg, $\x$), bold upper-case characters to denote matrices (\eg, $\A$).
We use $\langle \cdot , \cdot \rangle$ to denote the inner product.
We use  $\| \cdot\|$ to denote an arbitrary norm, and $\|\cdot\|_*$ to denote its dual norm, unless otherwise specified. 
We use $\| \cdot\|_2$ to denote the $\ell_2$ norm of a vector or the operator norm of a matrix, and $\|\cdot\|_{\A}$ to denote the vector norm induced by a positive definite matrix $\A$, namely $\|\x\|_{\A} := \sqrt{ \x^\top \A \x }$. 
For any convex function $g(\x)$, we use $g^*(\y)$ to denote its convex conjugate $g^*(\y) := \sup_{\x \in \reals^d}\{ \langle \y, \x \rangle - g(\x) \}$. 
For any distance-generating function $h$, we use $D_h(\x, \y)$ to denote the Bregman divergence, namely $D_h(\x, \y) := h(\x) - h(\y) - \langle h(\y), \x - \y \rangle$.
We use $\x^{\star}$ to denote the optimum of the objective being optimized (which should be clear from context).

For any federated algorithms, we use $M$ to denote the number of parallel clients, $R$ to denote the number of rounds, $K$ to denote the number of local steps per round. 
In general, we use superscripts to denote timesteps, italicized subscripts to denote the indices of clients. 
For federated algorithms, $\x^{(r,k)}_m$ means the state at the $k$-th local step of the $r$-th round at the $m$-th client.
% For example, $\x^{(0,0)}$ means the initialization (state at the $0$-th local step of the $0$-th round. 
We use the overline to denote the parametric averaging overall clients, \eg, $\overline{\x^{(r,k)}} := \frac{1}{M} \sum_{m=1}^M \x^{(r,k)}_m$.

% Let $B := \|\x^{(0,0)} - \x^{\star}\|$ be the Euclidean distance of the initial guess $\x^{(0,0)}$ and the optimum $\x^{\star}$.

Throughout the dissertation, we use $O, \Omega, \Theta$ notation to hide absolute constants only, whereas $\tildeo, \tilde{\Theta}$ may hides multiplicative $\polylog$ factors, which will be clarified in the formal context. 

\chapter{Sharp Bounds for Federated Averaging and Continuous Perspective}
\label{chapter:fedavg}

% \input{fedavg/fedavg_intro}
% \section{Additional Related Work}
% \label{sec:related-work}

% \input{fedavg/2o_setup}
% In many applications, the goal is to collectively minimize some training loss over the space of models.
% Federated Learning (FL) is a recently popularized distributed learning paradigm in which a massive number of clients collaboratively train a model without disclosing their private local data. 

Federated Averaging (\fedavg), also known as Local SGD, is one of the most popular algorithms in Federated Learning (FL). 
Despite its simplicity and popularity, the convergence rate of \fedavg has thus far been undetermined. 
Even under the simplest assumptions (convex, smooth, homogeneous, and bounded covariance), the best-known upper and lower bounds do not match, and it is not clear whether the existing analysis captures the capacity of the algorithm. 
In this chapter, we first resolve this question by providing a lower bound for \fedavg that matches the existing upper bound, which shows the existing \fedavg upper bound analysis is not improvable. 
Additionally, we establish a lower bound in a heterogeneous setting that nearly matches the existing upper bound. While our lower bounds show the limitations of \fedavg, under an additional assumption of third-order smoothness, we prove more optimistic state-of-the-art convergence results in both convex and non-convex settings. 
Our analysis stems from a notion we call iterate bias, which is defined by the deviation of the expectation of the SGD trajectory from the noiseless gradient descent trajectory with the same initialization. 
We prove novel sharp bounds on this quantity, and show intuitively how to analyze this quantity from a Stochastic Differential Equation (SDE) perspective.

Reflecting the goal of minimizing a loss function aggregated across a group of clients, in this chapter, we consider the following federated optimization problem:
\begin{equation}
  \min F(\x) := \frac{1}{M} \sum_{m=1}^M F_m(\x), \quad \text{where  } F_m(\x) := \expt_{\x \sim \dist_m} f(\x; \xi),
  \label{eq:fo:hetero}
\end{equation}
where each client $m \in [M]$ holds a local objective $F_m$ realized by its local data distribution $\dist_m$. 
We assume that each client can access the stochastic gradient oracle $\nabla f(\x; \xi)$ by drawing independent sample $\xi$ from $\dist_m$.
In typical machine learning settings, this objective function represents a loss function evaluated over certain data distribution. 
Federated Learning is \emph{heterogeneous} by design as $\mathcal{D}_m$ can vary across clients. 
In the special case when $\mathcal{D}_m \equiv \mathcal{D}$ for all clients $m$, the problem is called \emph{homogeneous}.
% For instance, if $\mathcal{D}$ is a distribution over data points, and $f(\x; \xi)$ is a loss function evaluated at parameter $\x$ and data point $\xi$, then we take the objective function $$F(\x) := \expt_{\xi \sim \mathcal{D}} f(\x; \xi).$$ 

% Federated Averaging (\fedavg) is one of the most popular algorithms applied in Federated Learning. 
% \footnote{In practice, \fedavg usually runs on a randomly sampled subset of heterogeneous clients for each synchronization round, whereas Local SGD or Parallel SGD usually run on a fixed set of clients. In this chapter we do not differentiate the terminology and assumed a fixed set of clients are deployed for simplicity.}
In its simplest form, \fedavg proceeds in $R$ communication rounds, where at the beginning of each round $r$, a central orchestration server sends the current state $\x^{(r,0)}$ to each of the $M$ clients. 
Each client then locally takes $K$ steps of SGD \cite{Robbins.Monro-51}, and then returns its final state to the central server. 
The central server averages these iterates to obtain the first iterate of the next round, namely $\x^{(r+1,0)}$. 
We state the \fedavg algorithm formally in pseudo code; see \cref{alg:fedavg}.
\begin{algorithm}[ht]
  \caption{\fedavgfull (\fedavg)}
  \label{alg:fedavg}
  \begin{algorithmic}[1]
  \STATE {\textbf{procedure}} \fedavg ($\x^{(0, 0)}; \eta$)
  \FOR {$r=0, \ldots, R-1$}
      \FORALL {$m \in [M]$ {\bf in parallel}}
        \STATE $\x_m^{(r,0)} \gets \x^{(r,0)}$ \COMMENT{broadcast current state}
        \FOR {$k = 0, \ldots, K-1$}
          \STATE $\xi^{(r,k)}_m \sim \mathcal{D}_m$
          \STATE $\g^{(r,k)}_m \gets \nabla f(\x^{(r,k)}_m; \xi^{(r,k)}_m)$
          \STATE $\x^{(r, k+1)}_m \gets \x^{(r, k)}_m - \eta \cdot \g_m^{(r,k)}$
          \COMMENT{client update} 
        \ENDFOR
      \ENDFOR
      $\x^{(r+1, 0)} \gets \frac{1}{M} \sum_{m=1}^M \x^{(r,K)}_m$
      \COMMENT{server averaging}
  \ENDFOR
\end{algorithmic}
\end{algorithm}

\section{Preliminaries}
\label{sec:fedavg:2o:ub}
In this section, we introduce the assumptions and review the well-known upper bounds of \fedavg obtained by \cite{Khaled.Mishchenko.ea-AISTATS20,Woodworth.Patel.ea-ICML20,Woodworth.Patel.ea-NeurIPS20}. 
We start by stating a set of assumptions that all local objectives are supposed to satisfy. 
% \begin{assumption}
%   \label{asm:fo:2o}
%   Let $f: \reals^d \times \Xi \to \reals$ be a real-valued function and $\mathcal{D}$ be a distribution supported on $\Xi$.
%   Denote $F(\x) := \expt_{\xi \sim \dist} f(\x; \xi)$. 
%   We say $(f, \dist)$ satisfies \cref{asm:fo:2o} if the following conditions are met:
%   \begin{enumerate}[(a)]
%       \item $f(\x; \xi)$ is second-order continuously differentiable w.r.t. $\x \in \reals^d$ for any $\xi$.
%       \item $F(\x)$ is convex for any $\x \in \reals^d$.
%       \item $F(\x)$ is $L$-smooth with respect to $\x \in \reals^d$. That is, for any $\x, \y \in \reals^d$, we have 
%       $$
%       \| \nabla F(\x) - \nabla F(\y) \|_2 \leq L\|\x - \y\|_2.
%       $$
%       \item For any $\x \in \reals^d$, $\expt_{\xi \sim \mathcal{D}} \|\nabla f(\x; \xi) - \nabla F(\x)\|_2^2 \leq \sigma^2.$
%   \end{enumerate}
% \end{assumption}
\begin{assumption}[Convexity, $L$-smoothness and $\sigma^2$-uniformly bounded gradient covariance]
  \label{asm:fo:2o}
  Consider the federated optimization problem \eqref{eq:fo:hetero}. 
  Assume that for any client $m \in [M]$,
  \begin{enumerate}[(a)]
    \item $F_m(\x)$ is convex for any $\x \in \reals^d$.
    \item $F_m(\x)$ is $L$-smooth. That is, for any $\x, \y \in \reals^d$, we have 
    \begin{equation*}
    \| \nabla F_m(\x) - \nabla F_m(\y) \|_2 \leq L\|\x - \y\|_2.
    \end{equation*}
    \item For any $\x \in \reals^d$, $\expt_{\xi \sim \mathcal{D}_m} \|\nabla f(\x; \xi) - \nabla F_m(\x)\|_2^2 \leq \sigma^2.$
  \end{enumerate}
\end{assumption}
In the case of only one client, it is known that \textsc{SGD} with $T$ steps can return an expected function error of order $\frac{LB^2}{T} + \frac{\sigma B}{\sqrt{T}}$, where $B$ is the bound of Euclidean distance from the initialization to the optimum.%, whereas accelerated \textsc{SGD} (also known as \textsc{AC-SA} \cite{Lan-MP12}) can attain the rate of $\frac{LB^2}{T^2} + \frac{\sigma B}{\sqrt{MT}}$.

Comparable assumptions are assumed in existing studies on \fedavg.
For example, \cite{Khaled.Mishchenko.ea-AISTATS20} assumes $f(\x; \xi)$ are convex and smooth for all $\xi$, which is more restricted.
\cite{Stich.Karimireddy-arXiv19} assumes quasi-convexity instead of convexity.
\cite{Haddadpour.Kamani.ea-NeurIPS19} assumes P-\L\ condition instead of convexity. 
In addition, the bounded covariance assumption (c) can be relaxed, for example, to $\expt_{\xi \sim \dist_m} \| \nabla f(\x; \xi) - \nabla F_m(\x)\|_2^2 \leq \sigma^2 + \tilde{\sigma}^2 \|\nabla F_m(\x)\|_2^2$ (c.f., \cite{Karimireddy.Kale.ea-ICML20}), whereas the corresponding bounds will be weaker.
Since our goal is to establish both upper and lower bounds, we will focus on the most common and representative settings as stated in \cref{asm:fo:2o}. 
Our proof technique may extend to broader settings described above.

We consider the following two settings of federated optimization. 
The first setting, known as the \emph{homogeneous} or \emph{i.i.d.} setting, assumes that all clients share the same distribution $\dist$, which we formalize as follows. 
\begin{assumption}[Homogeneous]
  \label{asm:fedavg:homo}
  Consider the federated optimization problem \eqref{eq:fo:hetero}. 
  Assume that all clients share the same distribution $\dist$, namely $\dist_m \equiv \dist$.%, and $(f, \mathcal{D})$ satisfies the local assumptions \cref{asm:fo:2o}.
\end{assumption}
While heterogeneity is commonly believed to be the major challenge in Federated Learning practice, as we will see in subsequent sections, the fundamental difficulty of local optimization methods (such as \fedavg) already arises in homogeneous settings. 
It is crucial to understand the behavior of local optimization methods under simple, homogeneous settings before advancing to more complicated, heterogeneous settings.

A less-restricted setting is to impose a bounded heterogeneity across clients, which we formalize as follows. Similar conditions have been imposed in an array of related works, c.f. \cite{Woodworth.Patel.ea-NeurIPS20}.
\begin{assumption}[Bounded Heterogeneity]
  \label{asm:fedavg:hetero}
 % $(f, \mathcal{D}_m)$ satisfies the local condition \cref{asm:fo:2o} for any $m \in [M]$, and in addition
  Consider the federated optimization problem \eqref{eq:fo:hetero}. 
  Assume that
  \begin{equation*}
    \max_{m \in [M]} \sup_{\x} \|\nabla F_m(\x) - \nabla F(\x) \|_2^2 \leq \zeta^2.
  \end{equation*}
\end{assumption}
Note that \cref{asm:fedavg:hetero} reduces to \cref{asm:fedavg:homo} if $\zeta = 0$. 

Under \cref{asm:fo:2o,asm:fedavg:hetero}, the best-known upper bound of \fedavg is due to \cite{Khaled.Mishchenko.ea-AISTATS20,Woodworth.Patel.ea-ICML20,Woodworth.Patel.ea-NeurIPS20}, which we quote below.
\begin{proposition}[label=thm:fedavg:2o:ub,restate=ThmFedAvgSecondOrderUB,name={Convergence Rate for \fedavg, adapted from \cite{Khaled.Mishchenko.ea-AISTATS20,Woodworth.Patel.ea-ICML20,Woodworth.Patel.ea-NeurIPS20}}]
  Consider the model problem \cref{eq:fo:hetero} and assume \cref{asm:fo:2o,asm:fedavg:hetero}. Consider running \fedavg with $M$ clients, $R$ rounds and $K$ steps per round, starting from $\x^{(0,0)}$. Then there exists a step-size $\eta$ such that \fedavg yields 
  \begin{equation}
      \expt \left[  \frac{1}{K R} \sum_{r=0}^{R-1} \sum_{k=1}^{K} F( \overline{\x^{(r,k)}} ) - F(\x^{\star}) \right]
      \leq
      \bigo
      \left(
      \underbrace{\frac{LB^2}{K R}}_{\text{\ding{172}}}
      + 
      \underbrace{\frac{\sigma B}{\sqrt{M K R}}}_{\text{\ding{173}}}
      +
      \underbrace{\frac{L^{\frac{1}{3}} \sigma^{\frac{2}{3}} B^{\frac{4}{3}} }{K^{\frac{1}{3}} R^{\frac{2}{3}}}}_{\text{\ding{174}}}
      + 
      \underbrace{\frac{L^{\frac{1}{3}} \zeta^{\frac{2}{3}} B^{\frac{4}{3}} }{R^{\frac{2}{3}}}}_{\text{\ding{175}}}
      \right).
      \label{eq:thm:fedavg:2o:ub}
  \end{equation}
  Particularly when \cref{asm:fedavg:homo} holds, the RHS of \cref{eq:thm:fedavg:2o:ub} becomes $\bigo(\text{\ding{172}}+\text{\ding{173}}+\text{\ding{174}})$.
\end{proposition}
\begin{remark}
  \label{rem:fedavg:2o:ub}
  \cref{thm:fedavg:2o:ub} does not include some obvious upper bounds that can be obtained by certain trivial settings.
  For example, by letting $\eta = 0$, the LHS of \cref{eq:thm:fedavg:2o:ub} can be upper bounded by $\bigo(LB^2)$ since the iterates stay at $\x^{(0,0)}$.
  In homogeneous setting (under \cref{asm:fedavg:homo}), since any single client can achieve an upper bound of $\bigo(\frac{LB^2}{KR} + \frac{\sigma B}{\sqrt{KR}})$ by applying $KR$ steps of \textsc{SGD}, \fedavg can at least attain the same bound due to convexity. 
  Hence, a comprehensive upper bound of homogeneous \fedavg can be 
  \begin{equation*}
    \bigo
    \left(
      \min \left\{ LB^2, ~~
      \frac{LB^2}{KR} + \frac{\sigma B}{\sqrt{KR}}, ~~
      \frac{LB^2}{K R} 
      + \frac{\sigma B}{\sqrt{M K R}} 
      + \frac{L^{\frac{1}{3}} \sigma^{\frac{2}{3}} B^{\frac{4}{3}} }{K^{\frac{1}{3}} R^{\frac{2}{3}}}
        \right\}
    \right).
  \end{equation*}
  \end{remark}

\subsection{Interpretation of \cref{thm:fedavg:2o:ub}}
Before we review the proof of \cref{thm:fedavg:2o:ub}, we first provide some intuitions for the convergence rates above. 
There are four terms on the RHS of \cref{eq:thm:fedavg:2o:ub}. The first two terms, namely \ding{172} and \ding{173}, are familiar from the standard SGD convergence rate. 
\begin{itemize}[leftmargin=*]
  \item The first term $\frac{LB^2}{KR}$ corresponds to the deterministic convergence, which appears even when there is no noise. 
  \item The second term $\frac{\sigma B}{\sqrt{MKR}}$ is a standard statistical noise term that applies to any algorithm which accesses $MKR$ total stochastic gradients.
  \item The third term $\frac{L^{\frac{1}{3}} \sigma^{\frac{2}{3}} B^{\frac{4}{3}}} {K^{\frac{1}{3}} R^{\frac{2}{3}}}$ depends on the variance of the noise, and arises due to the local steps applied in \fedavg. 
  This term appears even in the homogeneous setting where all clients access the same distribution.
  The previous best lower bound, due to \cite{Woodworth.Patel.ea-ICML20}, achieved in comparison the term $\frac{L^{\frac{1}{3}} \sigma^{\frac{2}{3}} B^{\frac{4}{3}}} {\boldsymbol{K^{\frac{2}{3}}} R^{\frac{2}{3}}}$, which is a factor of $K^{\frac{1}{3}}$ weaker.
  As we will see in the subsequent sections, this term \ding{174} is intrinsic and does appear in the lower bound of \fedavg.
  \item The last term $\frac{L^{\frac{1}{3}} \zeta^{\frac{2}{3}} B^{\frac{4}{3}}} {R^{\frac{2}{3}}}$ is caused by the heterogeneity of the data among the clients. As we will see in subsequent sections, this term also appears in the lower bound of \fedavg.
\end{itemize}

\subsection{Review of \fedavg Upper Bound Analysis}
In this subsection, we review the upper bound analysis of \fedavg by providing the proof of \cref{thm:fedavg:2o:ub}.\footnote{This result is well-known which we include for completeness only.
The specific exposition below is also included in \cite{Wang.Charles.ea-21} contributed by the dissertation author, mainly adapted from \cite{Woodworth.Patel.ea-NeurIPS20}.}
The upper bound analysis of \fedavg \cite{Stich-ICLR19,Stich.Karimireddy-arXiv19,Khaled.Mishchenko.ea-AISTATS20,Woodworth.Patel.ea-ICML20} typically follows the perturbed iterate analysis framework \cite{Mania.Pan.ea-SIOPT17} where the performance of \fedavg is compared with the idealized version with immediate communication.
The key idea is to control the stability of SGD so that the local iterates held by parallel clients do not differ much, even with infrequent communication.

We structure the analysis into the following two lemmas. 
The first lemma shows that the shadow trajectory, defined as $\overline{\x^{(r,k)}} := \frac{1}{M} \sum_{m=1}^M \x^{(r,k)}_m$, converges comparably to the synchronized SGD up-to a variance term. 
\begin{lemma}[label=lem:fedavg:2o:ub:1,restate=LemFedAvgSecondOrderUBFirst,name={Convergence of shadow trajectory up to variance term}]
  Under the same setting of \cref{thm:fedavg:2o:ub}, for any stepsize $\eta \leq \frac{1}{4L}$, the following inequality holds
  \begin{align*}
    & \expt \left[  \frac{1}{K R} \sum_{r=0}^{R-1} \sum_{k=1}^{K} F( \overline{\x^{(r,k)}} ) - F(\x^{\star}) 
    + \frac{1}{2 \eta KR}  \left\|\overline{\x^{(r,K)}} - \x^{\star} \right\|_2^2 \right]
    \\
    \leq &
    \underbrace{\frac{1}{2 \eta KR} \left\|\overline{\x^{(0,0)}} - \x^{\star} \right\|_2^2 + \frac{\eta \sigma^2}{M}}_{\text{synchronized SGD}}
    +
    \frac{L}{MKR} \sum_{r=0}^{R-1} \sum_{k=0}^{K-1} \sum_{m=1}^M \expt \left[  \left\| \x_m^{(r,k)} -  \overline{\x^{(r,k)}} \right\|_2^2 \right]
  \end{align*}
\end{lemma}

The second lemma shows that the intra-client variance term introduced in \cref{lem:fedavg:2o:ub:1} is indeed upper bounded by the gradient covariance bound $\sigma$ and heterogeneity bound $\zeta$.
\begin{lemma}[label=lem:fedavg:2o:ub:2,restate=LemFedAvgSecondOrderUBSecond,name={Bounded inter-client variance}]
  Under the same setting of \cref{thm:fedavg:2o:ub}, for any stepsize $\eta \leq \frac{1}{4L}$, the following inequality holds for any $r \in \{0, 1, \ldots, R-1\}$ and $k \in \{0, 1, \dots, K-1\}$.
  \begin{equation*}
      \expt \left[ \left\| \x_m^{(r,k)} -  \overline{\x^{(r,k)}} \right\|_2^2  \right] 
      \leq 
      4 K \eta^2 \sigma^2 + 18 K^2 \eta^2 \zeta^2.
  \end{equation*}
\end{lemma}

The detailed proof of \cref{lem:fedavg:2o:ub:1,lem:fedavg:2o:ub:2} is provided in \cref{sec:pf:lem:fedavg:2o:ub:1,sec:pf:lem:fedavg:2o:ub:2}, adapted from \cite{Woodworth.Patel.ea-NeurIPS20}. The upper bound \cref{thm:fedavg:2o:ub} then follows immediately once we specify the appropriate $\eta$:
\begin{proof}[Proof of \cref{thm:fedavg:2o:ub}]
  Applying \cref{lem:fedavg:2o:ub:1,lem:fedavg:2o:ub:2} gives
  \begin{align*}
      & \expt \left[  \frac{1}{K} \sum_{k=1}^{K} F( \overline{\x^{(r,k)}} ) - F(\x^{\star}) \middle| \mathcal{F}^{(r,0)} \right]
      + \frac{1}{2\eta K} \left( \expt\left[ \left\|\overline{\x^{(r,K)}} - \x^{\star} \right\|_2^2 \middle| \mathcal{F}^{(r,0)} \right]  - \left\|\overline{\x^{(r,0)}} - \x^{\star} \right\|_2^2 \right)
  \\
  \leq & \frac{\eta \sigma^2}{M}  + 4 K \eta^2 L \sigma^2 +  18 K^2 \eta^2 L \zeta^2.
  \end{align*}    
  Telescoping $r$ from $0$ to $R-1$ gives
  \begin{equation*}
      \expt \left[  \frac{1}{K R} \sum_{r=0}^{R-1} \sum_{k=1}^{K} F( \overline{\x^{(r,k)}} ) - F(\x^{\star})  \right]
      \leq \frac{B^2}{2\eta K R} 
      + \frac{\eta \sigma^2}{M} + 4 K \eta^2 L \sigma^2   +  18 K^2 \eta^2 L \zeta^2,
  \end{equation*}
  Furthermore, when the step size is chosen as
  \begin{equation*}
      \eta = \min \left\{ \frac{1}{4L}, \frac{M^{\frac{1}{2}} B}{K^{\frac{1}{2}} R^{\frac{1}{2}} \sigma}, \frac{B^{\frac{2}{3}}}{K^{\frac{2}{3}} R^{\frac{1}{3}} L^{\frac{1}{3}} \sigma^{\frac{2}{3}}}, \frac{B^{\frac{2}{3}}}{K R^{\frac{1}{3}} L^{\frac{1}{3}} \zeta^{\frac{2}{3}}} \right\},
  \end{equation*}
  we obtain the upper bound \cref{eq:thm:fedavg:2o:ub}.
\end{proof}

\section{Iterate Bias of SGD}
\label{sec:fedavg:2o:bias}
In this section, we will show why the third term of \cref{eq:thm:fedavg:2o:ub} arises in the upper bound \fedavg. 
The intuition from our lower bound comes from studying the behaviour of \fedavg when there are infinite number of homogeneous clients.
In this case, the averaged iterate $\x^{(r + 1, 0)}$ is precisely the \emph{expected} iterate after $K$ iterations of SGD starting from the last averaged iterate, $\x^{(r, 0)}$. This motivates the following definition. 
\begin{definition}[Iterate Bias of SGD]
\label{def:fedavg:bias}
Consider the stochastic approximation problem
\begin{equation}
    \min_{\x} F(\x) := \expt_{\xi \sim \dist} f(\x; \xi).
    \label{eq:fedavg:so}
\end{equation}
Let $\{\x_{\sgd}^{(k)}\}_{k=0}^{\infty}$ and $\{\z_{\gd}^{(k)}\}_{k=0}^{\infty}$ be the trajectories of SGD and GD initialized at the same point $\x$, formally
\begin{alignat}{3}
    & \x^{(k+1)}_{\sgd} \gets \x_{\sgd}^{(k)} - \eta \nabla f(\x_{\sgd}^{(k)}; \xi^{(k)}), \qquad & \x_{\sgd}^{(0)} = \x;
    \nonumber
    \\
    & \z^{(k+1)}_{\gd} \gets \z_{\gd}^{(k)} - \eta \nabla F(\z_{\gd})
    , \qquad  & \z_{\gd}^{(0)} = \x.
    \nonumber
\end{alignat}
The \textbf{iterate bias} (or in short ``bias'') from $\x$ at the $k$-th step is defined as 
\begin{equation*}
    \expt[\x_{\sgd}^{(k)}] - \z_{\gd}^{(k)},
\end{equation*}
the difference between the mean of SGD trajectory and the (deterministic) GD trajectory.
\end{definition}

One important special case of \cref{def:fedavg:bias} is the iterate bias from a stationary point  $\x^{\star}$. In this case, the gradient descent trajectory $\z_{\gd}^{(k)}$ will stay at the optimum since $\nabla F(\z_{\sgd}^{(k)}) \equiv \nabla F(\x^{\star}) = \mathbf{0}$. 
The iterate bias then reduces to $\expt [\x_{\sgd}^{(k)}] - \x^{\star}$. 
Notably, even for convex smooth objectives $f$, the expected iterate $\expt[\x_{\sgd}^{(k)}]$ may drift away from the optimum $\x^{\star}$, even if initialized at the $\x^{\star}$.
This occurs because of a difference between the gradient of the expectation of an iterate, $\nabla F(\expt[\cdot])$, and the expectation of the gradient of the iterate, $\expt [\nabla F(\cdot)]$.
% $\nabla f(\expt[\x_{\sgd}^{(k)} ])$.

\begin{figure*}[t]
  \centering
  \begin{subfigure}{0.2\textwidth}
      \centering
      \vspace{0.1cm}
      \includegraphics[width=\textwidth]{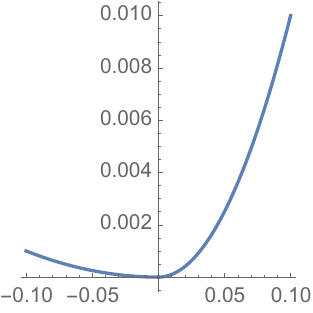}
      \vspace{0.05cm}
      \caption{Function plot}
      \label{fig:piecewise:quadratic}
  \end{subfigure}
  \hfill
  \begin{subfigure}{0.78\textwidth}
      \includegraphics[width=\textwidth]{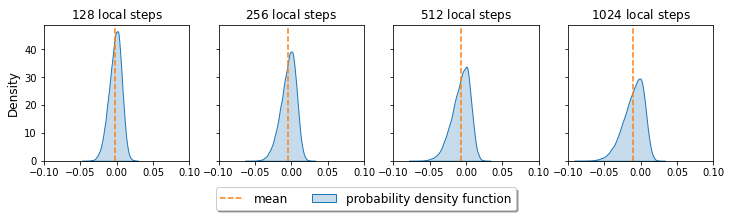}
      \caption{Probability density function after various SGD steps}
  \end{subfigure}
  \caption{\textbf{Illustration of the iterate bias of SGD.}  Consider the objective $F(x) = \begin{cases}x^2 & x \geq 0 \\
      \frac{1}{10} x^2 & x < 0\end{cases}$ as shown in (a), and $f(x; \xi) := \xi x + F(x)$ where $\xi \sim \mathcal{N}(0, 0.01)$.
      We initialize the SGD at optimum $x^{\star}=0$, and run 1024 steps of SGD with step size $10^{-2}$. We repeat this random process for 65536 times, and estimate the density function after 128, 256, 512 and 1024 steps. 
      Observe that the density function and the average gradually move to the left (away from the optimum, where the curvature is smaller). This figure explains the intrinsic difficulty for \fedavg to handle objective with drastic Hessian change.}
  \label{fig:bias}
\end{figure*}

In \cref{fig:bias}, we illustrate this phenomenon via a one-dimensional objective.\footnote{Code repository see \url{https://bit.ly/fedavg-aistats22}.} This figure, and our formal results below, illustrate that for sufficiently small step sizes, the bias increases in the number of steps $k$.
For this reason, doing more than one local step can sometimes be counterproductive. 
This phenomenon is key to the poor dependence on $K$ in the convergence rate we prove for \fedavg. 

Our goal is to characterize the iterate bias corresponding to the condition in \cref{asm:fo:2o}. 
Since \cref{asm:fo:2o} is stated for a federated optimization problem, we reformulate the local conditions of \cref{asm:fo:2o} for a proper stochastic approximation problem in the form of \cref{eq:fedavg:so}.
\begin{rassumption}{asm:fo:2o}
    % \label{asm:fedavg:2o:so}
    Consider the stochastic approximation problem \cref{eq:fedavg:so}. We say $(f, \dist)$ satisfies Assumption \ref*{asm:fo:2o}' if the following conditions are met:
    \begin{enumerate}[(a)]
      \item $F(\x)$ is convex for any $\x \in \reals^d$.
      \item $F(\x)$ is $L$-smooth with respect to $\x \in \reals^d$.
      \item For any $\x \in \reals^d$, $\expt_{\xi \sim \mathcal{D}} \|\nabla f(\x; \xi) - \nabla F(\x)\|_2^2 \leq \sigma^2.$
    \end{enumerate}
\end{rassumption}

Under Assumption \ref*{asm:fo:2o}', we can establish the following upper bound on the bias. Throughout this section, we mainly focus on the iterate bias bound in the regime of sufficiently small $\eta$ for simplicity and easy comparison. Our complete theorem in the appendix covers the case of general $\eta$ choice.
% \begin{theorem}[Upper bound of the iterate bias under \cref{asm:fo:2o}]
%     \label{thm:fedavg:2o:bias:ub}
%     Under \cref{asm:fo:2o}, there exists an absolute constant $\bar{c}$ such that for any initialization $\x$, for any $\eta \leq \frac{1}{2L}$, the iterate bias satisfies
%      \begin{equation}
%         \left \| 
%             \expt \x_{\sgd}^{(k)} - \z_{\gd}^{(k)}
%         \right\|_2
%         \leq
%         \bar{c}
%         \cdot
%         \min \{\eta^2 k^{\frac{3}{2}}  L \sigma, \eta k^{\frac{1}{2}} \sigma \}.
%     \end{equation}
% \end{theorem}
\begin{theorem}[Upper bound of the iterate bias under Assumption \ref*{asm:fo:2o}', simplified from \cref{thm:fedavg:2o:bias:ub:complete}]
    \label{thm:fedavg:2o:bias:ub}
    Consider running $\sgd$ and $\gd$ starting from some initialization $\x^{(0)}$. 
    Suppose $(f,\dist)$ satisfies Assumption \ref*{asm:fo:2o}', there exists an absolute constant $\bar{c}$ such that for any initialization $\x^{(0)}$, for any $\eta \leq \frac{1}{L}$, the iterate bias satisfies
    $\left \| 
            \expt \x_{\sgd}^{(k)} - \z_{\gd}^{(k)}
        \right\|_2
        \leq
        \bar{c}
        \cdot
        \eta^2 k^{\frac{3}{2}}  L \sigma.
    $
\end{theorem}
% \cref{thm:fedavg:2o:bias:ub} suggests that for very small $\eta$, the iterate bias scales at the order of $\eta^2$ until it passes the threshold of $\frac{1}{k L}$ in which the second term starts to dominate. \hynote{Looking for better description}.
In fact, we show in the following theorem that this upper bound of iterate bias is sharp.
\begin{theorem}[Lower bound of the iterate bias under Assumption \ref*{asm:fo:2o}', simplified from \cref{thm:fedavg:2o:bias:lb:complete}]
    \label{thm:fedavg:2o:bias:lb}
    There exists an absolute constant $\underline{c}$ such that for any $L, \sigma$,  there exists an objective $f(\x; \xi)$ and distribution $\xi \sim \dist$ satisfying Assumption \ref*{asm:fo:2o}' such that for any integer $K$, for any $\eta \leq \frac{1}{2KL}$, and integer $k \in [2, K]$,  the iterate bias from the optimum $\x^{\star}$ of $F$ is lower bounded as
    $
        \left \| 
            \expt \x_{\sgd}^{(k)} - \z_{\gd}^{(k)}
        \right\|_2
        \geq
        \underline{c} \cdot \eta^2 k^{\frac{3}{2}} L \sigma.
        % \underline{c} \cdot \min\{ \eta^2 k^{\frac{3}{2}} L \sigma,  \eta^{\frac{1}{2}} L^{-\frac{1}{2}} \sigma\}
    $
    % \tnote{it might be good to keep the lower bound having the same format as the upper bound. (make the LHS of the statement more similar in upper and lower bounds}
\end{theorem}

\cref{thm:fedavg:2o:bias:lb} shows that the SGD trajectory can indeed drift away (in expectation) from the optimum $\x^{\star}$ despite being initialized at $\x^{\star}$.
Our lower bound improves over the best-known lower bound $\Omega(\eta^2 k L \sigma)$
due to \cite{Woodworth.Patel.ea-ICML20}. 
The lower bound is attained by running SGD with Gaussian noise on the piecewise quadratic function 
$F(x) := \frac{1}{2}L \cdot \psi(x)$ where $\psi$ is a piecewise quadratic function defined as 
\begin{equation}
    \psi(x) := \begin{cases} x^2 & x \geq 0, \\
    \frac{1}{2} x^2 & x < 0.\end{cases}
    \label{eq:psi}
\end{equation}
The bias originates from the difference between $\nabla \psi(\expt[\x_{\sgd}^{(k)} ])$ and $\expt[\nabla \psi(\x_{\sgd}^{(k)})]$ due to the non-linearity of $\nabla \psi$.
     
The formal statements and proofs of \cref{thm:fedavg:2o:bias:ub,thm:fedavg:2o:bias:lb} are relegated to \cref{sec:pf:fedavg:2o:bias}. We briefly discuss the proof sketch of \cref{thm:fedavg:2o:bias:lb} in the following subsection.
\subsection{Proof Sketch of \cref{thm:fedavg:2o:bias:lb}}
Our main technique is comparing the iterates $x^{(0)}, x^{(1)}, \cdots$ from running SGD\footnote{We drop the subscript ``$\sgd$'' throughout this subsection for simplicity of notation.} on the piecewise quadratic function $\frac{1}{2} L \cdot \psi(x)$ to the iterates $\{y^{(k)}\}$ and $\{z^{(k)}\}$ obtained from running SGD on the quadratic functions 
\begin{equation*}
    f_{\ell}(x; \xi) := \frac{1}{4} L x^2 + \xi x,
    \quad
    \text{and}
    \quad
    f_{u}(x; \xi) := \frac{1}{2} L x^2 + \xi x,
\end{equation*}
respectively. We show that if $x^{(0)} = y^{(0)} = z^{(0)}$, then the iterate $x^{(k)}$ is first-order stochastically dominated by both $y^{(k)}$ and $z^{(k)}$. Fortunately, the iterates $y^{(k)}$ and $z^{(k)}$ are easy to analyze. A straightforward calculation yields the closed form solutions
\begin{equation*}
    y^{(k)} \sim \alpha_y^k y^{(0)} + \mathcal{N}(0, \sigma_y^2),
\quad
\text{and}
\quad
    z^{(k)} \sim \alpha_z^k z^{(0)} + \mathcal{N}(0, \sigma_z^2),
\end{equation*}
where 
\begin{equation}
    \label{eq:fedavg:var_defs}
\alpha_y := 1 - \eta L/2 , \quad 
\alpha_z := 1 - \eta L, \quad
\sigma_y^2 := \frac{\eta^2 \sigma^2 (1 - \alpha_y^k)}{1 - \alpha_y}, \quad
\sigma_z^2 := \frac{\eta^2 \sigma^2 (1 - \alpha_z^k)}{1 - \alpha_z}.
\end{equation}

We can then bound the expectation of $x^{(k)}$ in the following way:
\begin{equation*}
    \expt[x^{(k)}] = -\int_{c = -\infty}^0{\Pr[x^{(k)} \leq c]} +  \int_{c = 0}^\infty{\Pr[x^{(k)} \geq c]} \leq -\int_{c = -\infty}^0{\Pr[y^{(k)} \leq c]} +  \int_{c = 0}^\infty{\Pr[z^{(k)} \geq c]}.
\end{equation*}
This decomposition means that the higher variance of $y^{(k)}$ to contributes to the negative term, while the relatively lower variance of $z^{(k)}$ contributes to the positive term. 

Particularly when $x^{(0)} = y^{(0)} = z^{(0)} = 0$, we have $y^{(k)} \sim \mathcal{N}(0, \sigma_y^2)$ and $z^{(k)} \sim \mathcal{N}(0, \sigma_z^2)$. 
Plugging in the cdf of a Gaussian, we obtain 
\begin{equation*}
-\int_{c = -\infty}^0{\Pr[y^{(k)} \leq c]} +  \int_{c = 0}^\infty{\Pr[z^{(k)} \geq c]}  = -\frac{\sigma_y}{\sqrt{2\pi}} + \frac{\sigma_z}{\sqrt{2\pi}}.
\end{equation*}

Using the fact that $\eta L k \ll 1$, we can approximate 
$$\sigma_y^2 \approx \frac{\eta^2\sigma^2(\eta L k/2 + (\eta L k)^2/8)}{\eta L/2} = \eta^2\sigma^2k(1 - \eta L k/4),$$ and  $$\sigma_z^2 \approx \frac{\eta^2\sigma^2(\eta L k/2 + (\eta L k)^2/2)}{\eta L} = \eta^2\sigma^2k(1 - \eta L k/2),$$ such that 
\begin{equation*}
    \expt[x^{(k)}] \leq -\frac{\sigma_y}{\sqrt{2\pi}} + \frac{\sigma_z}{\sqrt{2\pi}} \approx \frac{\eta \sigma \sqrt{k}}{\sqrt{2\pi}}\left(\frac{\eta L k}{8}\right).
\end{equation*}

When $x^{(0)}$ is non-zero but sufficiently small, we can prove that this same negative iterate bias occurs in the expectation $\expt[x^{(k)}] - x^{(0)}$. With slightly more effort, we can show that so long as the \emph{expectation} $\expt[x^{(0)}]$ is sufficiently small, there is a negative drift in $\expt[x^{(k)}] - \expt[x^{(0)}]$. 

% Our lower bound analyzes the bias of a the iterates of a SGD on a function studied previously in \cite{Woodworth.Patel.ea-ICML20}, which we call the step function:
% $$\step(x) := \begin{cases}\frac{1}{4} L x^2 & x \geq 0 \\
%      \frac{1}{2}Lx^2 & x \leq 0\end{cases}$$
     
%  Both bounds in \cref{thm:2o:bias} are novel to the best of our knowledge. 
%     Specifically, the best known lower bound of iterate bias is $\eta^{\frac{1}{2}} \sigma$ due to \cite{Woodworth.Patel.ea-ICML20}. There is no known upper bound in existing literature.

% To understand this bound intuitively, observe that the iterate bias initially scales with $\eta^2k^{\frac{3}{2}}\sigma$. As $k$ becomes large enough, the strong convexity of the step function forces the bias to decrease, and the bias plateaus at $\eta^{\frac{1}{2}}\sigma$. 

\section{Lower Bound of \fedavg}
\label{sec:fedavg:2o:lb}
In this section, we present our lower bounds for \fedavg in both convex homogeneous and heterogeneous settings, and discuss its implications. We then show how use the lower bound on the bias of SGD from \cref{sec:fedavg:2o:bias} to establish a lower bound on the convergence of \fedavg.

% \subsection{Statement of Results}
% In our lower bounds, we consider functions meeting the following set of assumptions. The set of assumptions in \cref{asm:fo:2o} are standard in the optimization and Federated Learning literature, and these are the assumptions used to achieve the existing upper bounds for the homogeneous case given in \cref{table:complexity}. 
Our main result for the homogeneous setting is the following theorem.
\begin{theorem}[Lower bound for homogeneous \fedavg]\label{thm:fedavg:lb:homo}
% For any $K \geq 2$, $R$, $L$, $B$, $\sigma$, and $\zeta_*$, for any even positive $M$, there exists an instance of heterogeneous federated optimization problem \eqref{eq:fo:hetero}, namely an objective $f$ and distributions $\{\mathcal{D}_m\}$, each satisfying \cref{asm:fo:2o}, and together satisfying \cref{asm:hetero:star}, such that for some initialization $\x^{(0, 0)}$ with $\|\x^{(0, 0)} - \x^{\star}\|_2 < B$, the final iterate of \fedavg with any step size $\eta \in \reals_{\geq 0}$ satisfies:

For any $K \geq 2$, $R$, $M$, $L$, $B$, and $\sigma$, there exists an instance of homogeneous federated optimization problem satisfying \cref{asm:fo:2o,asm:fedavg:homo}, such that 
% after $R$ rounds of \fedavg with $K$ local steps 
% \hynote{Maybe for shortening no need to emphasize $R$ rounds?} 
% and $\mathcal{D}_m = \mathcal{D}$ for all clients $m \in [M]$, 
for some initialization $\x^{(0, 0)}$ with $\|\x^{(0, 0)} - \x^{\star}\|_2 < B$, the final state of \fedavg with any step size $\eta \in \reals_{\geq 0}$ satisfies:
\begin{equation*}
  \expt\left[F(\x^{(R, 0)})\right] - F(\x^{\star}) \geq 
  \Omega\left(
  \frac{LB^2}{KR}   + 
  \min\left\{LB^2,
  \frac{\sigma B}{\sqrt{MKR}} +
  \min\left\{\frac{\sigma B}{\sqrt{KR}}, \frac{L^\frac{1}{3} \sigma^{\frac{2}{3}} B^{\frac{4}{3}}}{K^{\frac{1}{3}} R^{\frac{2}{3}}} \right\}
  \right\}
  \right).
\end{equation*}
or rearranging
\begin{equation*}
    \expt\left[F(\x^{(R, 0)})\right] - F(\x^{\star}) \geq 
    \Omega \left(
    \min \left\{ LB^2, ~
    \frac{LB^2}{KR} + \frac{\sigma B}{\sqrt{KR}}, ~
    \frac{LB^2}{KR} + \frac{\sigma B}{\sqrt{MKR}} +  \frac{L^\frac{1}{3} \sigma^{\frac{2}{3}} B^{\frac{4}{3}}}{K^{\frac{1}{3}} R^{\frac{2}{3}}} \right\}
    \right).
\end{equation*}
\end{theorem}
This lower bound matches the best-known upper bound given by \cref{thm:fedavg:2o:ub} (see also \cref{rem:fedavg:2o:ub}). 
Our lower bound shows that under only and assumption of second order smoothness and convexity (\cref{asm:fo:2o}), \fedavg may achieve a rate as slow as ${K^{-\frac{1}{3}}R^{-\frac{2}{3}}}$. 
Prior work has pointed out that this rate can be beat by alternative algorithms that use the same (or less) communication and gradient computation. One such algorithm is \emph{minibatch SGD}, which replaces the $K$ iterations of local SGD at each client with a single iteration. 
This results in the same outcome as $R$ iterations of SGD with minibatch size $M$. 
A second such algorithm, \emph{single-client SGD} ignores all but one client, and results the same outcome as $KR$ iterations of SGD. Under \cref{asm:fo:2o}, the best of these two algorithms (minibatch SGD and single-client SGD) achieves a rate of
\begin{equation*}
    \frac{LB^2}{KR} + \frac{\sigma B}{\sqrt{MKR}} + \min\left(\frac{LB^2}{R},  \frac{\sigma B}{\sqrt{KR}}\right).
\end{equation*}
It turns out that this rate always dominates the the sharp rate we have shown for \fedavg.
% \footnote{In fact, \cite{woodworth2021min} proved that the minimax rate under \cref{asm:fo:2o} is achieved by performing either accelerated minibatch SGD,  or accelerated single-client SGD, yielding an improved rate of $\frac{LB^2}{K^2R^2} + \frac{\sigma B}{\sqrt{MKR}} + \min\left(\frac{LB^2}{R^2},  \frac{\sigma B}{\sqrt{KR}}\right)$.}
Further, when $\sigma$ and $K$ are large, this rate is dominated by $\frac{L B^2}{R}$, while the rate of \fedavg is dominated by $\frac{L^{\frac{1}{3}}\sigma^{\frac{2}{3}}B^{\frac{4}{3}}}{K^{\frac{1}{3}}R^{\frac{2}{3}}}$. In this regime, the rate of this ``naive'' algorithm may improve on the rate of \fedavg by a factor of $\left(\frac{R\sigma^2L^2}{K}\right)^{\frac{1}{3}}$.

We extend our results to the heterogeneous setting. Recall that in this setting, we allow each client $m \in [M]$ to draw $\xi$ from its own distribution $\mathcal{D}_m$. 
We prove our results under a slightly weaker notation of heterogeneity, where the heterogeneity bound is only imposed at the optimum.
\begin{assumption}[Bounded gradient heterogeneity at optimum]
    Consider the federated optimization problem \eqref{eq:fo:hetero}. Assume that
    \label{asm:hetero:star}
    \begin{equation*}
        \frac{1}{M} \sum_{m=1}^M \|\nabla F_m(\x^{\star})\|_2^2 \leq \zeta_*^2.
    \end{equation*}
\end{assumption}
\begin{remark}
\label{rem:hetero}
While the right measure of heterogeneity is a subject of significant debate in the FL community, the most popular are either a bound on gradient heterogeneity at $\x^{\star}$ (\cref{asm:hetero:star}), or a stronger assumption of uniform gradient heterogeneity (\cref{asm:fedavg:hetero}). The best-known lower bound, due to \cite{Woodworth.Patel.ea-NeurIPS20}, considers the weaker \cref{asm:hetero:star}. We remark however that the strongest upper bounds use the stronger uniform assumption (e.g., \cite{Woodworth.Patel.ea-ICML20} \footnote{While \cite{Khaled.Mishchenko.ea-AISTATS20} studies a relaxed assumption (optimum-heterogeneity like \cref{asm:hetero:star} instead of uniform-heterogeneity), these results only hold with a much smaller step-size range $\eta \lesssim \frac{1}{KL}$ (in our notation, c.f. Theorem 3, 4 and 5 in their work), instead of $\eta \lesssim \frac{1}{L}$ as in the uniform setting. Under this restricted step-size range, one cannot recover the same upper bounds as in uniform-heterogeneity by optimizing $\eta$.}).
\end{remark}

We establish the following theorem on the lower bound of heterogeneous \fedavg.
\begin{theorem}[Lower bound for heterogeneous \fedavg]\label{thm:fedavg:lb:hetero}
For any $K \geq 2$, $R$, $L$, $B$, $\sigma$, and $\zeta_*$, for any even positive $M$, there exists an instance of heterogeneous federated optimization problem \eqref{eq:fo:hetero} satisfying \cref{asm:fo:2o,asm:hetero:star}, such that for some initialization $\x^{(0, 0)}$ with $\|\x^{(0, 0)} - \x^{\star}\|_2 < B$, the final iterate of \fedavg with any step size $\eta \in \reals_{\geq 0}$ satisfies:
% for $\x^{\star} \in \argmin \frac{1}{M}\sum_m F_m(\x)$, we have
\begin{align*}
    & \expt\left[F(\x^{(R, 0)})\right] - F(\x^{\star})
    \\
    &
    \Omega\left(\frac{LB^2}{KR} + 
    \min\left\{LB^2, \frac{\sigma B}{\sqrt{MKR}} 
    + \min\left\{\frac{\sigma B}{\sqrt{KR}},  \frac{L^\frac{1}{3} \sigma^{\frac{2}{3}} B^{\frac{4}{3}}}{K^{\frac{1}{3}} R^{\frac{2}{3}}} \right\}
    + \min\left\{\frac{\zeta_*^2}{L}, \frac{L^{\frac{1}{3}}\zeta_*^{\frac{2}{3}}B^{\frac{4}{3}}}{R^{\frac{2}{3}}}\right\} \right\}\right)
    % \label{eq:fedavg:lb:hetero}
\end{align*}
\end{theorem}
% \cref{thm:fedavg:lb:homo} yields a sharp lower bound that matches the upper bound of \fedavg in the homogeneous setting, given by \cite{Woodworth.Patel.ea-ICML20}.
\cref{thm:fedavg:lb:hetero} is nearly tight, up to a difference in the definitions of heterogeneity (See \cref{rem:hetero}). We compare our result to existing lower bounds and upper bounds in \cref{table:complexity}. 

\begin{table*}[ht]
    \centering
    \caption{\textbf{Convergence Rates of \fedavg under \cref{asm:fo:2o}}. Some lower order terms as $R \rightarrow \infty$ omitted.
% Prior results for homogeneous case are due to \cite{Woodworth.Patel.ea-ICML20}. Prior results for heterogeneous case are due to \cite{Woodworth.Patel.ea-NeurIPS20}. 
$L$: smoothness, $R$: number of rounds, $K$: local iterations per round, $M$: number of clients, $\sigma$: noise, $B: \|\x^{(0, 0)} - \x^{\star}\|_2.$ The lower and upper bound use a slightly different metric of heterogeneity ($\zeta$ and $\zeta_*$), see  \cref{rem:hetero} for details. 
% $\zeta_* : \expt_m\|\nabla F_m(\x^*)\|_2^2$,  $\zeta : \max_{\x} \expt_m\|\nabla F_m(\x)\|_2^2$. 
We bold the terms where our analysis improves upon previous work. 
}\label{table:complexity}
\resizebox{\linewidth}{!}{
\begin{tabular}{@{}lll@{}} \toprule
             & Homogeneous & Heterogeneous\\ \midrule
    Previous Upper Bound &  $\frac{LB^2}{KR} + \frac{ \sigma B}{\sqrt{MKR}} + \frac{L^{\frac{1}{3}} \sigma^{\frac{2}{3}} B^{\frac{4}{3}}} {K^{\frac{1}{3}} R^{\frac{2}{3}}}$   &  $\frac{LB^2}{KR} + \frac{ \sigma B}{\sqrt{MKR}} + \frac{L^{\frac{1}{3}} \sigma^{\frac{2}{3}} B^{\frac{4}{3}}} {K^{\frac{1}{3}} R^{\frac{2}{3}}} + \frac{L^{\frac{1}{3}}\zeta^{\frac{2}{3}}B^{\frac{4}{3}}}{R^{\frac{2}{3}}}$  \vspace{0.15cm}  \\ 
                         & \cite{Khaled.Mishchenko.ea-AISTATS20} & \cite{Khaled.Mishchenko.ea-AISTATS20,Woodworth.Patel.ea-NeurIPS20} \\ \midrule
    \textbf{Our Lower Bound} &    ${\boldsymbol{\frac{LB^2}{KR}}} + \frac{ \sigma B}{\sqrt{MKR}} + \frac{L^{\frac{1}{3}} \sigma^{\frac{2}{3}} B^{\frac{4}{3}}} {{\boldsymbol{K^{\frac{1}{3}}}} R^{\frac{2}{3}}}$ 
    & $\frac{LB^2}{KR} + \frac{ \sigma B}{\sqrt{MKR}} + \frac{L^{\frac{1}{3}} \sigma^{\frac{2}{3}} B^{\frac{4}{3}}} {{\boldsymbol{K^{\frac{1}{3}}}} R^{\frac{2}{3}}} + {\boldsymbol{\frac{L^{\frac{1}{3}} \zeta^{\frac{2}{3}}_*B^{\frac{4}{3}}}{R^{\frac{2}{3}}}}}$   \vspace{0.15cm}  \\
                & \cref{thm:fedavg:lb:homo} & \cref{thm:fedavg:lb:hetero} \\ \midrule
    Previous Lower Bound &  $\frac{ \sigma B}{\sqrt{MKR}} + \frac{L^{\frac{1}{3}} \sigma^{\frac{2}{3}} B^{\frac{4}{3}}} {{\boldsymbol{K^{\frac{2}{3}}}} R^{\frac{2}{3}}} $ & $\frac{ \sigma B}{\sqrt{MKR}} +  \frac{L^{\frac{1}{3}} \sigma^{\frac{2}{3}} B^{\frac{4}{3}}} {{\boldsymbol{K^{\frac{2}{3}}}} R^{\frac{2}{3}}} + {\boldsymbol{\min\left(\frac{LB^2}{R},\frac{L^{\frac{1}{3}} \zeta^{\frac{2}{3}}_*B^{\frac{4}{3}}}{R^{\frac{2}{3}}}\right)}}$   \vspace{0.15cm}  \\
                & \cite{Woodworth.Patel.ea-ICML20} & \cite{Woodworth.Patel.ea-NeurIPS20} \\ 
    \bottomrule
\end{tabular}
}
\end{table*}

% \begin{remark}
%     Note that $\zeta_* = 0$ does not necessarily imply that the distributions are homogeneous. Hence for this theorem to imply  \cref{thm:fedavg:lb:homo}, we ensure that in the case when $\zeta_* = 0$, all of the distributions $\mathcal{D}_m$ are equal.
%     \hynote{check context}
% \end{remark}

% \subsection{Constructing Lower Bound from Iterate Bias}\label{sec:fedavg:2o:lb}
% \hynote{
% In this subsection, we theoretically establish the relationship between the iterate bias (\cref{def:fedavg:bias}) and the lower bound on the function error of \fedavg.

% Recall that in \cref{thm:fedavg:2o:bias:lb}, we proved a lower bound on bias from the optimum $x^{\star}$, which came from analyzing SGD with Gaussian noise on the the piecewise quadratic function, which we abbreviate ``$\psi(x)$'':
% \begin{equation}\label{def:pqf}
%     f(x; \xi) = \frac{1}{2} L \psi(x) + x \xi, \quad 
%     \psi(x) = 
%     \begin{cases}
%         \frac{1}{2} x^2 & x < 0 \\
%         x^2 & x \geq 0, \\
%     \end{cases}
% \end{equation}
% where $\xi \sim \mathcal{N}(0, 1)$. 
% }
\subsection{Proof of \cref{thm:fedavg:lb:homo,thm:fedavg:lb:hetero}}
In this subsection we will prove the lower bound results stated in \cref{thm:fedavg:lb:homo,thm:fedavg:lb:hetero}. We will consider the following 4-dimensional stochastic functions over $\x = (x_1, x_2, x_3, x_4)$ for our lower bound\footnote{Throughout this subsection we shall slightly abuse the notation by overloading the subscript for coordinates instead of clients. The semantics should be clear from context.}. 
\begin{equation}
    f(\x; \xi) = f^{(1)}(x_1; \xi_1) + F^{(2)}(x_2) + F^{(3)}(x_3) + f^{(4)}(x_4; \xi_2, \xi_3),
    \label{eq:fedavg:lb:f}
\end{equation}
where 
\begin{equation}
    f^{(1)}(x; \xi_1) = \frac{1}{24} L \psi(x) + x \xi_1, 
    \quad \text{where  }  \psi(x) :=     
    \begin{cases}
        \frac{1}{2} x^2 & x < 0 \\
        x^2 & x \geq 0, \\
    \end{cases} \quad \xi_1 \sim \mathcal{N}(0, \sigma^2);
    \label{eq:fedavg:lb:f1}
\end{equation}
\begin{equation}
    F^{(2)}(x)= \frac{1}{2} \mu x^2,
    \text{  where $\mu$ is a function of $\sigma, B, K, R, L, \zeta_*$ to be determined;}
    \label{eq:fedavg:lb:f2:1}
\end{equation}
\begin{equation}
    F^{(3)}(x) = \frac{1}{2} L x^2;
    \label{eq:fedavg:lb:f3}
\end{equation}
and
\begin{equation}
f^{(4)}(x; \xi_2, \xi_3) = 
\begin{cases}
    \frac{1}{8} L x^2 - x\xi_3 & \text{if }\xi_2 = 1 \\ 
    \frac{1}{16} L x^2 - x\xi_3& \text{if }\xi_2 = 2 \\ 
\end{cases}
    \label{eq:fedavg:lb:f4}
\end{equation}
The distribution of $(\xi_2, \xi_3)$ is heterogeneous across clients: For all the odd $m \in [M]$, we let $(\xi_2, \xi_3) = (1, \zeta_*)$ always, while for all the even $m \in [M]$ we let $(\xi_2, \xi_3) = (2, -\zeta_*)$. Denote $F^{(1)}(x) := \expt_{\xi_1} f^{(1)}(x; \xi_1)$ and $F^{(4)}(x) := \frac{1}{M} \sum_{m=1}^M \expt_{\xi_2, \xi_3 \sim \dist_m} f^{(4)}(x; \xi_2, \xi_3)$. 

Since the trajectory of $\x = (x_1, x_2, x_3, x_4)$ are completely decoupled by coordinates, we can analyze the four components separately. 

The role of the \textbf{first} component $f^{(1)}$ is to provide the iterate bias. We provide a sharp analysis of the bias $\expt[x^{(R, 0)}]$ on the piecewise quadratic function $\psi$.
\begin{lemma}[label=lem:fedavg:lb:f1,restate=lemsteplb]
Consider $f^{(1)}(x; \xi) = \frac{1}{24}L \psi(x) + \xi x$ for $\xi \sim \mathcal{N}(0, \sigma^2)$, as defined in \cref{eq:fedavg:lb:f1}.
Suppose we run \fedavg starting from $x^{(0,0)} = 0$ for $R$ rounds with $K$ local steps per round.
Then there exists a universal constant $c_1 > 0$ such that for any $\eta \leq \frac{2}{L}$, the following inequality holds
\begin{equation}
    \expt[x^{(R,0)}] \leq -c_1 \cdot \eta^{\frac{1}{2}} L^{-\frac{1}{2}} \sigma \min\left\{1, (\eta L K)^{\frac{1}{2}}, (\eta L K)^{\frac{3}{2}}  R\right\}.
\end{equation}
Hence there exists a universal constant $C_1$ such that 
\begin{equation}
    \expt [F^{(1)}(x^{(R,0)})] \geq F^{(1)} (\expt x^{(R,0)}) \geq C_1 \cdot \eta \sigma^2 \min\left\{1, (\eta L K), (\eta L K)^3 R^2 \right\}.
    \label{eq:fedavg:F1:lb}
\end{equation}
\end{lemma}
The second inequality \cref{eq:fedavg:F1:lb} holds due to the fact that $F^{(1)}(x) = \frac{1}{24} L \psi(x) \geq \frac{1}{48} L x^2$. We sketch the proof of the first inequality in \cref{sec:pf:lem:fedavg:lb:f1}.

The role of the \textbf{second} component $f^{(2)}$ is to provide a dimension (the $x_2$-axis) in which the objective is only slightly convex. 
Indeed, this term requires that $\eta$ is sufficiently large for convergence, which we formalize in \cref{lem:fedavg:lb:f2}. 
\begin{lemma}[label=lem:fedavg:lb:f2,restate=LemFtwo]
    Consider $F^{(2)}(x) = \frac{1}{2}\mu x^2$, as defined in \cref{eq:fedavg:lb:f2:1}.
    Suppose we run \fedavg on this deterministic function starting from some $x^{(0,0)}$ for $R$ rounds with $K$ local steps per round.
    Then there exists a universal constant $C_2$ such that for any $\eta \leq \frac{1}{\mu KR}$, the following inequality holds
    \begin{equation}
        F^{(2)}(x^{(R,0)}) \geq C_2 \cdot \mu \left( x^{(0,0)} \right)^2.
    \end{equation}
\end{lemma}
The proof of \cref{lem:fedavg:lb:f2} is relegated to \cref{sec:pf:lem:fedavg:lb:f2}.

The role of the \textbf{third} component $F^{(3)}$ is to ensure that we can limit our analysis to cases with small step size, $\eta$. Indeed, by standard arguments, if $\eta \geq \frac{2}{L}$, then \fedavg on $F^{(3)}$ will not converge.
\begin{lemma}[label=lem:fedavg:lb:f3,restate=LemFthree]
    Consider $F^{(3)}(x) = \frac{1}{2}L x^2$, as defined in \cref{eq:fedavg:lb:f3}.
    Suppose we run \fedavg on this deterministic function starting from some $x^{(0,0)}$ for $R$ rounds with $K$ local steps per round.
    Then there exists a universal constant $C_3$ such that for any $\eta \geq \frac{2}{L}$, the following inequality holds
    \begin{equation}
        F^{(3)}(x^{(R,0)}) \geq \frac{1}{2} L (x^{(0,0)})^2.
    \end{equation}
\end{lemma}

The role of the \textbf{fourth} component $f^{(4)}$ is to provide bias with heterogeneous objective. 
\begin{lemma}[label=lem:fedavg:lb:f4,restate=LemFedAvgHetero]
Consider $f^{(4)}(x; \xi_2, \xi_3)$ as defined in \cref{eq:fedavg:lb:f4}. 
Suppose we run \fedavg with even $M$ clients starting from some $x^{(0,0)}$ for $R$ rounds with $K$ local steps per round.
There exists a universal constant $c_4$ such that for $\eta \leq \frac{2}{L}$, the following inequality holds
\begin{equation}
    x^{(R, 0)} \leq - c_4 \cdot L^{-1}\zeta_* \min \{1, \eta L K, (\eta L K)^2R\}.
    \label{eq:lem:fedavg:lb:f4:1}
\end{equation} 
Hence there exists a universal constant $C_4$ such that 
\begin{equation}
    F^{(4)}(x^{(R,0)}) \geq C_4 \cdot L^{-1} \zeta_*^2 \min \{1, (\eta L K), (\eta L K)^4 R^2\}.
    \label{eq:lem:fedavg:lb:f4:2}
\end{equation}
\end{lemma}
The second inequality follows directly from \cref{eq:lem:fedavg:lb:f4:2} since $F^{(4)}(x) := \frac{1}{M} \sum_{m=1}^M F_m^{(4)}(x) = \frac{3}{32} Lx^2$. We defer the proof of the first inequality to \cref{sec:pf:lem:fedavg:lb:f4}. The functions studied in this lemma appear in the heterogeneous lower bound construction in \cite{Woodworth.Patel.ea-NeurIPS20}, but the analysis we give in this lemma is much tighter than theirs. 

Finally, we note that any first order method which uses at most $MKR$ stochastic gradients has a lower bound of $\Omega(\min\{\frac{\sigma B}{\sqrt{MKR}}, LB^2\})$ in expected function error \cite{Nemirovski.Yudin-83}. 

The proof of \cref{thm:fedavg:lb:homo,thm:fedavg:lb:hetero} then follows by summarizing the above observations.
\begin{proof}[Proof of \cref{thm:fedavg:lb:homo,thm:fedavg:lb:hetero}]    
    Since any first order method which uses at most $MKR$ stochastic gradients has a lower bound of $\Omega(\min\{\frac{\sigma B}{\sqrt{MKR}}, LB^2\})$ in expected function error, it suffices to prove the remaining three terms, namely
    \begin{equation}
        \Omega\left(\frac{LB^2}{KR}
        + \min\left\{\frac{\sigma B}{\sqrt{KR}},  \frac{L^\frac{1}{3} \sigma^{\frac{2}{3}} B^{\frac{4}{3}}}{K^{\frac{1}{3}} R^{\frac{2}{3}}}, LB^2 \right\}
        + \min\left\{\frac{\zeta_*^2}{L}, \frac{L^{\frac{1}{3}}\zeta_*^{\frac{2}{3}}B^{\frac{4}{3}}}{R^{\frac{2}{3}}}, LB^2 \right\}\right)
        \label{eq:fedavg:lb:proof:main}
    \end{equation}

    Consider running \fedavg on the four-dimensional stochastic objective defined in \cref{eq:fedavg:lb:f}, where
    \begin{equation}
        \mu := \frac{1}{2B^2}
        \max\left\{
            \min\left\{\frac{\sigma B}{\sqrt{KR}}, \frac{L^{\frac{1}{3}} \sigma^{\frac{2}{3}} B^{\frac{4}{3}}}{K^{\frac{1}{3}}R^{\frac{2}{3}}}, LB^2\right\},
            \min\left\{ \frac{\zeta_*^2}{L},   \frac{L^{\frac{1}{3}} \zeta_*^{\frac{2}{3}} B^{\frac{4}{3}}}{R^{\frac{2}{3}}}, LB^2 \right\},
            \frac{LB^2}{KR}
            \right\},
        \label{eq:fedavg:lb:f2:2}
    \end{equation}
    starting at $\x^{(0,0)} = \left(0, \frac{B}{2}, \frac{B}{2}, 0 \right)$. 
    Note that the objective satisfies the homogeneous assumption (\cref{asm:fedavg:homo}) if $\zeta_* = 0$, and the heterogeneous assumption (\cref{asm:fedavg:hetero}) for any general $\zeta_* > 0$. 
    
    By definition of $\mu$ it suffices to prove $\expt [F(\x^{(R,0)})] \geq \Omega (\mu B^2)$. We consider the following three cases:
    
    \textbf{Case 1:} $\eta > \frac{2}{L}$. In this case by \cref{lem:fedavg:lb:f3}, we have 
    \begin{equation}
        F^{(3)}(x_3^{(R,0)}) \geq \frac{1}{2} L(x_3^{(0,0)})^2 = \frac{1}{8} L B^2 \geq \frac{1}{8} \mu B^2.
        \label{eq:pf:lem:fedavg:lb:mub:1}
    \end{equation}
    
    \textbf{Case 2:} $\eta \leq \frac{1}{\mu K R}$. In this case by \cref{lem:fedavg:lb:f2}, we have 
    \begin{equation}
        F^{(2)}(x_2^{(R,0)}) \geq C_2 \cdot \mu \left( x_2^{(0,0)} \right)^2 = \frac{C_2}{4} \cdot \mu B^2.
        \label{eq:pf:lem:fedavg:lb:mub:2}
    \end{equation} 
    
    \textbf{Case 3:} $\eta \leq \frac{2}{L}$ and $\eta > \frac{1}{\mu KR}$. In this case we must have $\frac{2}{L} > \frac{1}{\mu KR}$, or by definition of $\mu$
    \begin{equation*}
        2B^2 \mu = \max\left\{
            \min\left\{\frac{\sigma B}{\sqrt{KR}}, \frac{L^{\frac{1}{3}} \sigma^{\frac{2}{3}} B^{\frac{4}{3}}}{K^{\frac{1}{3}}R^{\frac{2}{3}}}, LB^2\right\},
            \min\left\{ \frac{\zeta_*^2}{L},   \frac{L^{\frac{1}{3}} \zeta_*^{\frac{2}{3}} B^{\frac{4}{3}}}{R^{\frac{2}{3}}}, LB^2 \right\},
            \frac{LB^2}{KR}
            \right\} > \frac{LB^2}{KR}.
    \end{equation*}
    and thus
    \begin{equation}
        \max\left\{
        \min\left\{\frac{\sigma B}{\sqrt{KR}}, \frac{L^{\frac{1}{3}} \sigma^{\frac{2}{3}} B^{\frac{4}{3}}}{K^{\frac{1}{3}}R^{\frac{2}{3}}}, LB^2 \right\} ,
        \min\left\{ \frac{\zeta_*^2}{L},   \frac{L^{\frac{1}{3}} \zeta_*^{\frac{2}{3}} B^{\frac{4}{3}}}{R^{\frac{2}{3}}}, LB^2 \right\}
        \right\}
        > \frac{LB^2}{KR},
        \label{eq:fedavg:case3:0}
    \end{equation}
    Depending on which term dominates the $\max$ in \cref{eq:fedavg:case3:0}, there are two sub-cases possible:

    \textbf{Case 3.1} $\min\left\{\frac{\sigma B}{\sqrt{KR}}, \frac{L^{\frac{1}{3}} \sigma^{\frac{2}{3}} B^{\frac{4}{3}}}{K^{\frac{1}{3}}R^{\frac{2}{3}}}, LB^2\right\} \geq \min\left\{ \frac{\zeta_*^2}{L},   \frac{L^{\frac{1}{3}} \zeta_*^{\frac{2}{3}} B^{\frac{4}{3}}}{R^{\frac{2}{3}}}, LB^2 \right\}$.
    
    In this case we have $\mu = \frac{1}{2B^2}  \min\left\{\frac{\sigma B}{\sqrt{KR}}, \frac{L^{\frac{1}{3}} \sigma^{\frac{2}{3}} B^{\frac{4}{3}}}{K^{\frac{1}{3}}R^{\frac{2}{3}}},  LB^2\right\}$. 
    Since $\eta > \frac{1}{\mu KR}$ we have
    \begin{equation}
        \eta > \frac{2B^2}{KR} \cdot \max \left\{ \frac{\sqrt{KR}}{\sigma B}, \frac{K^\frac{1}{3} R^{\frac{2}{3}}}{ L^{\frac{1}{3}} \sigma^{\frac{2}{3}} B^{\frac{4}{3}}} \right\} 
        = \max \left\{  \frac{2 B }{\sigma \sqrt{KR}}, \frac{2 B^{\frac{2}{3}} }{L^{\frac{1}{3}} \sigma^{\frac{2}{3}} K^{\frac{2}{3}} R^{\frac{1}{3}}} \right\}.
        \label{eq:fedavg:case3}
    \end{equation}
    
    Meanwhile since $\eta \leq \frac{2}{L}$ we have by \cref{lem:fedavg:lb:f1}, for some constant $C_1$, for $\eta \leq \frac{2}{L}$, we have 
    \begin{align*}
        \expt [F^{(1)}(x^{(R,0)}_1)] \geq C_1 \cdot \eta \sigma^2 \min\left\{1, \eta LK, R^2(\eta L K)^{3}  \right\}.
    \end{align*}
    Since $\eta L K R \geq \eta \mu KR > 1$ we can get rid of the third term and obtain 
    \begin{equation}
        \expt [F^{(1)}(x^{(R,0)}_1)]  \geq  C_1 \min\left\{ \eta \sigma^2  , \eta^2 L K \sigma^2 \right\}.
        \label{eq:fedavg:case3:2}
    \end{equation}
    Now we plug in the lower bound of $\eta$ from \cref{eq:fedavg:case3} to \cref{eq:fedavg:case3:2}. The first term is lower bounded as 
    \begin{equation*}
        \eta \sigma^2  >  \sigma^2 \cdot \frac{2 B }{\sigma \sqrt{KR}} = \frac{2 \sigma B}{\sqrt{ KR}}.
    \end{equation*}
    The second term is lower bounded as 
    \begin{equation*}
        \eta^2 L K \sigma^2 > \frac{4 B^{\frac{4}{3}} }{L^{\frac{2}{3}} \sigma^{\frac{4}{3}} K^{\frac{4}{3}} R^{\frac{2}{3}}} \cdot LK \sigma^2 = 4 \frac{L^{\frac{1}{3}} \sigma^{\frac{2}{3}} B^{\frac{4}{3}}}{K^{\frac{1}{3}} R^{\frac{2}{3}}}.
    \end{equation*}
    Plugging the above two inequalities back to \cref{eq:fedavg:case3:2} yields
    \begin{equation}
        \expt [F^{(1)}(x^{(R,0)}_1)]  \geq  4 C_1 \min \left\{  \frac{\sigma B}{\sqrt{KR}}, \frac{L^{\frac{1}{3}} \sigma^{\frac{2}{3}} B^{\frac{4}{3}}}{K^{\frac{1}{3}} R^{\frac{2}{3}}}, LB^2  \right\} = 8 C_1 \cdot \mu B^2.
        \label{eq:pf:lem:fedavg:lb:mub:3}
    \end{equation}

    \textbf{Case 3.2:} $\min\left\{\frac{\sigma B}{\sqrt{KR}}, \frac{L^{\frac{1}{3}} \sigma^{\frac{2}{3}} B^{\frac{4}{3}}}{K^{\frac{1}{3}}R^{\frac{2}{3}}}, LB^2 \right\} \leq \min\left\{ \frac{\zeta_*^2}{L},   \frac{L^{\frac{1}{3}} \zeta_*^{\frac{2}{3}} B^{\frac{4}{3}}}{R^{\frac{2}{3}}}, LB^2  \right\}$:

    In this case we have $\mu = \frac{1}{2B^2} \min\left\{ \frac{\zeta_*^2}{L},   \frac{L^{\frac{1}{3}} \zeta_*^{\frac{2}{3}} B^{\frac{4}{3}}}{R^{\frac{2}{3}}}, LB^2  \right\}$. Since $\eta > \frac{1}{\mu KR}$ we have 
    \begin{equation}
        \eta > \frac{2B^2}{KR}  \max \left\{ \frac{L}{\zeta_*^2}, \frac{R^{\frac{2}{3}}}{L^{\frac{1}{3}} \zeta_*^{\frac{2}{3}} B^{\frac{4}{3}}} \right\}
        \label{eq:fedavg:eta:lb}
    \end{equation}

    Meanwhile since $\eta \leq \frac{2}{L}$ we have by \cref{lem:fedavg:lb:f4}, for some constant $C_4$, for $\eta \leq \frac{2}{L}$, we have 
    \begin{equation*}
        F^{(4)}(x_4^{(R,0)})  \geq C_4 \cdot  \frac{L}{\zeta_*^2} \min \left\{ 1, \eta^2 L^2 K^2, \eta^4 L^4 K^4 R^2 \right\}
    \end{equation*}
    Since $\eta L K R \geq \eta \mu K R > 1$ we can get rid of the third term and obtain
    \begin{equation*}
        F^{(4)} (x_4^{(R,0)}) \geq C_4 \cdot \frac{L}{\zeta_*^2} \min \left\{ 1, \eta^2 L^2 K^2 \right\}
    \end{equation*}
    Plugging in \cref{eq:fedavg:eta:lb} yields
    \begin{align}
        & F^{(4)} (x_4^{(R,0)}) \geq C_4 \cdot \frac{L}{\zeta_*^2} \min \left\{ 1, \frac{4 B^{\frac{4}{3}} L^2 K^2}{ L^{\frac{2}{3}} K^2 R^{\frac{2}{3}} \zeta_*^{\frac{4}{3}} } \right\} 
        = C_4 \min \left\{ \frac{\zeta_*^2}{L}, \frac{4 L^{\frac{1}{3}} \zeta_*^{\frac{2}{3}} B^{\frac{4}{3}} } {R^{\frac{2}{3}}} \right\}
        = 2C_4 \cdot \mu B^2
        \label{eq:pf:lem:fedavg:lb:mub:4}
    \end{align}
    Combining \cref{eq:pf:lem:fedavg:lb:mub:1,eq:pf:lem:fedavg:lb:mub:2,eq:pf:lem:fedavg:lb:mub:3,eq:pf:lem:fedavg:lb:mub:4}, there exists a universal constant $C$ such that for any $\eta \in \reals_{\geq 0}$, it is the case that $\expt [F(\x^{(R,0)})] \geq C \cdot \mu B^2$  This proves \cref{eq:fedavg:lb:proof:main} and therefore \cref{thm:fedavg:lb:homo,thm:fedavg:lb:hetero}.
\end{proof}
\section{The Benefit of Third-Order Smoothness}
\subsection{Mitigating Iterate Bias by Third-Order Smoothness}
\label{sec:fedavg:3o:bias}
In light of the limitations of \fedavg discussed in the previous sections, it is natural to ask if there are additional assumptions under which \fedavg may perform better. 
The aforementioned lower bound is attained by a special piece-wise quadratic function \cref{eq:psi} with a sudden curvature change, which is smooth (has bounded second-order derivatives) but has unbounded third-order derivatives. 
A natural additional assumption to exclude this corner case is third-order smoothness, stated formally in \cref{asm:fo:3o}.

% This piecewise quadratic function has an unbounded third order derivative at $0$, which causes this difference to be large whenever the distribution of $\x_{\sgd}^{(k)}$ spans both sides of $0$. This worst case construction motivates our further study of the bias under a third-order derivative bound. 
\begin{assumption}
    \label{asm:fo:3o}
    Consider the federated optimization problem \eqref{eq:fo:hetero}. In addition to \cref{asm:fo:2o}, assume that for any client $m \in [M]$,
    \begin{enumerate}[(a)]
      \item $F_m(\x)$ is $Q$-3rd-order-smooth with respect to $\x \in \reals^d$, i.e. for any $\xi$, for any $\x, \y \in \reals^d$, 
      \begin{equation*}
      \| \nabla^2 F_m(\x) - \nabla^2 F_m(\y) \|_2 \leq Q \|\x - \y\|_2.
      \end{equation*}
      \item $\nabla f(\x; \xi)$ has $\sigma^4$-bounded fourth-order central moment, i.e.,
      \begin{equation*}
          \sup_{\x \in \reals^d}\expt_{\xi \sim \mathcal{D}_m}\left[\left\|\nabla f(\x; \xi)- \nabla F_m(\x)]\right\|_2^4\right] \leq \sigma^4.
      \end{equation*}
    \end{enumerate}
\end{assumption}
% Several classes of additional assumptions have been suggested for studying the performance of \fedavg. Perhaps the most common, and the one supported from our intuition on the bias, is an assumption of third-order smoothness, stated formally in \cref{asm:fo:3o}. Previously it has been shown that under such an assumption, \fedavg may converge faster. We present several state-of-the-art bounds for \fedavg under \cref{asm:fo:3o}, including for the non-convex case.
A similar version of \cref{asm:fo:3o} was studied in \cite{Dieuleveut.Patel-NeurIPS19}. 
In fact, \cite{Dieuleveut.Patel-NeurIPS19} assumes bounded \nth{4} central moment at optimum only, which results in weaker results. 
We adopt the uniformly bounded \nth{4} central moment for consistency with \cref{asm:fo:2o}.

\cref{asm:fo:3o} is stated with respect to a federated optimization problem \eqref{eq:fo:hetero}. 
To study the iterate bias associated with \cref{asm:fo:3o}, we first reformulate the above assumption in the form of a stochastic approximation problem.
\begin{rassumption}{asm:fo:3o}
    Consider the stochastic approximation problem \eqref{eq:fedavg:so}. We say $(f, \dist)$ satisfies Assumption \ref*{asm:fo:3o}' if $(f, \dist)$ satisfy \cref{asm:fo:2o}', and the following conditions are met:
    \begin{enumerate}[(a)]
        \item $F(\x)$ is $Q$-3rd-order-smooth with respect to $\x \in \reals^d$.
        \item $\nabla f(\x; \xi)$ has $\sigma^4$-bounded fourth-order central moment.
    \end{enumerate}
\end{rassumption}
We show that under this additional assumption, the iterate bias reduces to $\bigo(\eta^3 k^2 Q \sigma^2)$, which scales on the order of $\eta^3$ (rather than $\eta^2$) as $\eta$ goes to 0.
\begin{theorem}[Simplified from \cref{thm:fedavg:3o:bias:ub:complete}]
%[Upper bound of the iterate bias under\cref{asm:fo:2o,asm:fo:3o}]
\label{thm:fedavg:3o:bias:ub}
Consider running $\sgd$ and $\gd$ starting from some initialization $\x^{(0)}$. 
Suppose $(f,\dist)$ satisfies Assumption \ref*{asm:fo:3o}', then there exists an absolute constant $\bar{c}$ such that for any initialization $\x$, for any $\eta \leq \frac{1}{2L}$, the iterate bias satisfies 
    $
        \left \| 
            \expt \x_{\sgd}^{(k)} - \z_{\gd}^{(k)}
        \right\|_2
        \leq
        \bar{c}
        \cdot
        \eta^3 k^2 Q \sigma^2.
    $
\end{theorem}
% \cref{thm:fedavg:3o:bias:ub} improves over the upper bound $\bigo (\eta^2 k^{\frac{3}{2}} L \sigma)$ provided by \cref{thm:fedavg:2o:bias:ub}, in that it 
\cref{thm:fedavg:3o:bias:ub} also reveals the dependency on the third-order smoothness $Q$. In the extreme case where $Q = 0$ ($f$ is quadratic), the iterate bias will disappear. It is worth noting that since Assumption \ref*{asm:fo:2o}' is still required in \cref{thm:fedavg:3o:bias:ub}, the original upper bound $\bigo(\eta^2 k^{\frac{3}{2}} L \sigma)$ from \cref{thm:fedavg:2o:bias:ub} still applies, and one can formulate the upper bound as the minimum of the two. 

The following lower bound shows that the upper bound in \cref{thm:fedavg:3o:bias:ub} is sharp asymptotically.
% \cref{thm:fedavg:3o:bias:ub} is also sharp. We show a matching lower bound in \cref{thm:fedavg:3o:bias:lb} that the iterate bias bound 
\begin{theorem}[Simplified from \cref{thm:fedavg:3o:bias:lb:complete}]
% [Lower bound of the iterate bias \cref{asm:fo:2o,asm:}]
    \label{thm:fedavg:3o:bias:lb}
      There exists an absolute constant $\underline{c}$ such that for any $L, \sigma, K$, for any sufficiently small $Q$ (polynomially dependent on $L, \sigma, K$), there exists an objective $f(\x; \xi)$ and distribution $\xi \sim \dist$ satisfying Assumption \ref*{asm:fo:3o}' such that for any $\eta \leq \frac{1}{2LK}$ and integer $k \in [2, K]$, the iterate bias from the optimum $\x^{\star}$ is lower bounded as
    $
        \left \| 
           \expt \x_{\sgd}^{(k)} - \z_{\gd}^{(k)}
        \right\|_2
        \geq
        \underline{c} \cdot \eta^3 k^2 Q \sigma^2.
    $
\end{theorem}
The formal statements and proofs of \cref{thm:fedavg:3o:bias:ub,thm:fedavg:3o:bias:lb} are provided in \cref{sec:pf:fedavg:3o:bias}.

% \tnote{just to make sure there is no typo---is it true that in 3 of these theorems the learning rate requirement is $\eta\le 1/(2kL)$ where in theorem 2.4 you don't have the $k$? }

\subsection{Revealing Iterate Bias via Continuous Perspective}
\label{sec:sde}
We demonstrate how the iterate bias can be analyzed from a continuous view of SGD.
As an example, we will explain how the $\Theta(\eta^3 k^2 Q \sigma^2)$ term shows up in \cref{thm:fedavg:3o:bias:ub,thm:fedavg:3o:bias:lb}. 

Consider a one-dimensional instance of SGD with Gaussian noise, where $f(x; \xi) = F(x) - \xi x$, and $\xi \sim \mathcal{N}(0, \sigma^2)$.
% \footnote{For simplicity, we assume $F$ is infinitely times differentiable throughout this subsection.}
The SGD then follows
    \begin{equation}
        x^{(k+1)}_{\sgd} = x^{(k)}_{\sgd} - \eta \nabla F(x^{(k)}_{\sgd}) + \eta \xi^{(k)}, ~~ \text{where } \xi^{(k)} \sim \mathcal{N}(0, \sigma^2).
        \label{eq:sde:sgd}
    \end{equation}
The continuous limit of \eqref{eq:sde:sgd} corresponds to the following SDE, with the scaling $t = \eta k$:
\begin{equation}
    \diff X(t) = - F'(X(t)) \diff t + \sqrt{\eta} \sigma \diff B_t,
    \label{eq:sde}
\end{equation}
where $B_t$ denotes the Brownian motion (also known as the Wiener process).\footnote{To justify the relation of \cref{eq:sde:sgd} and \cref{eq:sde}, note that \cref{eq:sde:sgd} can be viewed as a numerical discretization (Euler-Maruyama discretization \cite{Kloeden.Platen-92}) of the SDE \eqref{eq:sde} with time step-size $\eta$.}

To get a handle of the iterate bias, our goal is to study $\expt [X(t) | X(0) = x]$, the expectation of the SDE solution $X(t)$ initialized at $x$. We view this quantity as a multivariate function $u(t, x)$ of $t$ and $x$, with the objective to Taylor expand $u(t, x)$ around $u(0, x)$ in $t$:
 \begin{equation*}
     u(t, x) = u(0, x) + u_t(0,x) t + \frac{1}{2}u_{tt}(0,x) t^2 + o(t^2).
 \end{equation*}
For brevity, we use subscript notation to denote partial derivatives, e.g, $u_x$ denotes $\frac{\partial u(t, x)}{\partial x}$.
The relationship of $u(t,x)$ and the SDE \eqref{eq:sde} is established by the Kolmogorov backward equation as follows. 
% To study.
% and studying its derivatives in terms of $t$ and $x$, we can write the PDE that governs its behavior. 
\begin{claim}[Kolmogorov backward equation \cite{Oksendal-03}]
    Let $u(t,x) = \expt [X(t) | X(0) = x]$, then $u(t,x)$ satisfies the following  partial differential equation:
    \begin{equation}
        u_t = - F_x u_x + \eta \sigma^2 u_{xx},\quad \text{with $u(0,x) = x$.}
        \label{eq:pde:1d}
    \end{equation}
    % \begin{equation}
    %     \frac{\partial u(t,x)}{\partial t} = - F'(x) \frac{\partial u(t,x)}{\partial x} + \eta \sigma^2 \frac{\partial^2 u(t,x)}{\partial x^2}
    % \end{equation}
    % We will write $u_t = - F_x u_x + \eta \sigma^2 u_{xx}$  for simplicity.
\end{claim}
% \begin{claim}[Infinitesimal generator]
%     The infinitesimal generator of SDE \eqref{eq:sde} is $\mathcal{L} = -F'(x) \cdot \frac{\diff}{\diff x} + \eta \sigma^2(x) \frac{\diff^2}{\diff x^2}$.
% \end{claim}
% Goal: study the growth of $u(t,x)$ around $t = 0$.
%  Although the PDE \eqref{eq:pde:1d} does not admit a closed form solution in general, we can quantify its initial behavior by Taylor expanding $u(t, x)$ in $t$ around $t = 0$:
%  \begin{compact}\begin{equation}
%      u(t, x) = u(0, x) + u_t t + \frac{1}{2}u_t t^2 + o(t^2).
%  \end{equation}\end{compact}
%  To write this expansion,
% \mg{One line here explaining technique (converting derivatives in t to x). Then Claim 2.6 + remark on its meaning. Then derivation of derivatives. Then implication.}

Using this claim, we can compute the first two derivatives of $u(t, x)$ in $t$, as follows:
\begin{lemma}
    \label{lem:pde:utt}
    Suppose $u(t,x)$ satisfies the PDE \eqref{eq:pde:1d}, then $
        u_t(0,x) = -F_x,  u_{tt}(0,x) = F_x F_{xx} - \eta \sigma^2 F_{xxx}$.
    % \begin{enumerate}[(a)]
    %     \item $u_x(0,x) = 1$
    %     \item $u_t(0,x) = -F_x$
    %     \item $u_{xx}(0,x) = 0$
    %     \item $u_{xt}(0,x) = - F_{xx}$
    %     % \item $u_{xxx}(0,x) = 0$
    % \end{enumerate}
\end{lemma}
\begin{proof}[Proof sketch of \cref{lem:pde:utt}]
    The first equation follows from equation~\eqref{eq:pde:1d} and the fact that $u_x(0, x) \equiv 1$ and $u_{xx}(0, x) \equiv 0 $ since $u(0,x) = x$. 
    To see the second equation, we take $\partial_t$ on both sides of \eqref{eq:pde:1d}, which gives
    \begin{equation}
        u_{tt} = - F_x u_{xt} + \eta \sigma^2 u_{xxt}.
        \label{eq:proof:lem:pde:utt}
    \end{equation}
    % Taking $\partial_t$ on both sides of \cref{eq:pde:1d} yields
    % \begin{equation}
    % \end{equation}
    % (a) follows by $u(0,x) = x$.
    Since  $u_{xt} = u_{tx} = (u_t)_x$, one has (by \cref{eq:pde:1d})
    \begin{equation*}
        u_{xt} = (- F_x u_x + \eta \sigma^2 u_{xx})_x = - F_{xx} u_x + - F_x u_{xx} + \eta \sigma^2 u_{xxx}.
    \end{equation*}
    For $t = 0$ we have $u_{xt}(0, x) = -F_{xx}$ since $u_{xx}(0,x) \equiv u_{xxx}(0,x) \equiv 0$. 
    Taking another $\partial_x$ yields $u_{xxt}(0, x) = -F_{xxx}$. Plugging back to \cref{eq:proof:lem:pde:utt} yields the second equation of the \cref{lem:pde:utt}.
\end{proof}
With \cref{lem:pde:utt} we can expand $u(t,x)$ around $(0,x)$:
\begin{equation*}
    u(t,x) = x - F_x t + \frac{1}{2} \left( F_x F_{xx} - \eta \sigma^2 F_{xxx} \right) t^2 + o(t^2).
\end{equation*}
Ignoring higher order terms in $t$, the term $-\frac{1}{2}\eta \sigma^2 F_{xxx}$ reflects the difference between the noiseless GD trajectory from $x$ and $\expt[X(t)| X(0) = x]$, that is, the iterate bias. 
Converting back to the discrete trajectory (\cref{eq:sde:sgd}) via the scaling $t = \eta k$, we obtain
\begin{equation*}
    \expt[x_{\sgd}^{(k)}] - z_{\gd}^{(k)} \approx -\frac{1}{2}\eta^3k^2\sigma^2F_{xxx}(x).
\end{equation*}
When the third derivative of $F$ is bounded by $Q$, this recovers the upper bound of $O(\eta^3k^2 Q \sigma^2)$ in \cref{thm:fedavg:3o:bias:ub}. The lower bound of \cref{thm:fedavg:3o:bias:lb} follows by choosing a function with third derivative $Q$ at $x^{\star}$. 
% At $x = x^{\star}$ we have
% \begin{equation}
%     u(t, x^{\star}) = x^{\star} - \frac{1}{2} \eta \sigma^2 F_{xxx}(x^{\star}) t^2 + o(t^2).
% \end{equation}
% Converting back to the discrete trajectory (\cref{eq:sde:sgd}) via the scaling $t = \eta k$, 
% \begin{equation}
%     u(\eta k, x^\star) \approx x^{\star} - \frac{1}{2} \eta^3 k^2 \sigma^2 F_{xxx}(x^{\star}).
% \end{equation}
% When the third derivative $F_{xxx}(x^{\star}) = Q$, this recovers the $\eta^3 k^2 Q \sigma^2$ quantity claimed in \cref{thm:fedavg:3o:bias:ub,thm:fedavg:3o:bias:lb}.

While it is possible to derive these results via more-involved discrete approaches, we believe the SDE approach may be promising for understanding more general objectives and algorithms. For instance, for multi-dimensional objectives, one can apply the same techniques to derive the \emph{direction} of the iterate bias via a multi-dimensional SDE, which is difficult to derive in the discrete setting. 

\subsection{Upper Bound of \fedavg under Third-Order Smoothness}
\label{sec:fedavg:3o:ub}
In this subsection, we show that how third-order smoothness (\cref{asm:fo:3o}) can indeed improve the convergence of \fedavg.
\begin{theorem}[label=thm:fedavg:3o:ub]
  Consider the \textbf{homogeneous} federated optimization problem \cref{eq:fo:hetero} satisfying \cref{asm:fo:3o}. Consider running \fedavg with $R$ rounds and $K$ steps per round, starting from $\x^{(0,0)}$. Then there exists a step-size $\eta$ such that \fedavg yields 
  \begin{equation}
      \expt \left[ F(\hat{\x}) - F(\x^{\star}) \right]
      \leq
      \bigo
      \left(
      \underbrace{\frac{LB^2}{K R}}_{\text{\ding{172}}}
      + 
      \underbrace{\frac{\sigma B}{\sqrt{M K R}}}_{\text{\ding{173}}}
      +
      \underbrace{\frac{Q^{\frac{1}{3}} \sigma^{\frac{2}{3}} B^{\frac{5}{3}} }{K^{\frac{1}{3}} R^{\frac{2}{3}}}}_{\text{\ding{174}}}
      \right).
      \label{eq:thm:fedavg:3o:ub}
  \end{equation}
  where $\hat{\x}$ is a linear combination of $\{\x^{(r,k)}_m\}$ defined as follows.
  \begin{equation*}
    \hat{\x} := \left( \sum_{r=0}^{R-1} \sum_{k=0}^{K-1} \frac{1}{(1+\frac{1}{KR})^{rK+k+1}}  \right)^{-1} 
    \left( \frac{1}{M} \sum_{r=0}^{R-1} \sum_{k=0}^{K-1} \sum_{m=1}^M \frac{\x^{(r,k)}_m}{(1+\frac{1}{KR})^{rK + k + 1}} \right).
  \end{equation*}
\end{theorem}
Note that in \cref{thm:fedavg:3o:ub}, the overhead term \ding{174} no longer depends on the (second-order) smoothness $L$, but instead the third-order smoothness $Q$. In the extreme case when $Q = 0$ (the objective is quadratic), only \ding{172} and \ding{173} will remain in the upper bound. 
Later in \cref{chapter:fedac} we will show how this bound can be further improved by careful acceleration.

\begin{proof}[Proof of \cref{thm:fedavg:3o:ub}]
For any $r < R$ and $k < K$, we have
  \begin{align}
  & \expt \left[ \| \overline{\x^{(r,k+1)}} - \x^{\star} \|_2^2 \middle| \mathcal{F}^{(r,k)} \right]
  =
  \expt \left[  \left\| \overline{\x^{(r,k)}} - \eta \cdot \frac{1}{M} \sum_{m=1}^M \nabla f(\x_m^{(r,k)}; \xi_m^{(r,k)})  - \x^{\star} \right\|_2^2 \middle| \mathcal{F}^{(r,k)} \right]
  \nonumber \\
  \leq &  \left\|\overline{\x^{(r,k)}} - \eta \cdot \frac{1}{M} \sum_{m=1}^M \nabla F(\x_m^{(r,k)}) - \x^{\star} \right\|_2^2 + \frac{\eta^2 \sigma^2}{M}
  \tag{Bounded variance assumption}
  \\
% \end{align*}
% % Now we focus on the first term of the RHS. The following inequality holds
% \begin{align*}
  % & \left\| \overline{\x^{(r,k)}} - \eta \cdot \frac{1}{M} \sum_{m=1}^M \nabla F(\x_m^{(r,k)}) - \x^{\star} \right\|_2^2
  % \\
  = &  \left\|\left(  \overline{\x^{(r,k)}} - \eta \cdot \nabla F(\overline{\x^{(r,k)}}) - \x^{\star}  \right) + \eta \left( \nabla F(\overline{\x^{(r,k)}}) -  \frac{1}{M} \sum_{m=1}^M \nabla F(\x_m^{(r,k)}) \right) \right\|_2^2 + \frac{\eta^2 \sigma^2}{M} 
  \nonumber  \\
  \leq & \left(1 + \frac{1}{KR} \right) \left\| \overline{\x^{(r,k)}} - \eta \cdot \nabla F(\overline{\x^{(r,k)}}) - \x^{\star}  \right\|_2^2 
  \nonumber \\
  & 
  \qquad + 2 \eta^2 KR \left\| \nabla F(\overline{\x^{(r,k)}}) -  \frac{1}{M} \sum_{m=1}^M \nabla F(\x_m^{(r,k)}) \right\|_2^2
  + \frac{\eta^2 \sigma^2}{M}
  \label{eq:fedavg:3o:ub:1}
  % \\
  % \leq & \left(1 + \frac{1}{KR} \right) \left\| \overline{\x^{(r,k)}} - \eta \cdot \nabla F(\overline{\x^{(r,k)}}) - \x^{\star}  \right\|_2^2 
  % + \frac{\eta^2 \sigma^2}{M}
  % + \frac{\eta^2 Q^2 KR}{2M} \sum_{m=1}^M \left\| \x^{(r,k)}_m - \overline{\x^{(r,k)}} \right\|_2^4
\end{align}
The first term of the RHS of \cref{eq:fedavg:3o:ub:1} can be bounded by standard convex analysis as follows (for any $\eta \leq \frac{1}{2L}$):
\begin{align}
  & \left\| \overline{\x^{(r,k)}} - \eta \cdot \nabla F(\overline{\x^{(r,k)}}) - \x^{\star}  \right\|_2^2 
  \nonumber \\
  = & \| \overline{\x^{(r,k)}} - \x^{\star} \|_2^2- 2 \eta \left\langle  \overline{\x^{(r,k)}} - \x^{\star}  ,  \nabla F(\overline{\x^{(r,k)}})  \right\rangle + \eta^2 \left\| \nabla F(\overline{\x^{(r,k)}})  \right\|_2^2
  \nonumber  \\
  \leq & \| \overline{\x^{(r,k)}} - \x^{\star} \|_2^2- 2 \eta (1 - \eta L) \left(F(\overline{\x^{(r,k)}} ) - F^{\star} \right)
  \tag{By convexity and $L$-smoothness}
  \\
  \leq & \| \overline{\x^{(r,k)}} - \x^{\star} \|_2^2- \eta  \left(F(\overline{\x^{(r,k)}} ) - F^{\star} \right).
  \label{eq:fedavg:3o:ub:2}
\end{align}

To bound the second term of the RHS of \cref{eq:fedavg:3o:ub:1}, we note that by $Q$-third-order-smoothness, we have (c.f. helper \cref{helper:3rd:Lip})
\begin{equation}
  \left\| \nabla F(\overline{\x^{(r,k)}}) -  \frac{1}{M} \sum_{m=1}^M \nabla F(\x_m^{(r,k)}) \right\|_2^2 \leq \frac{Q^2}{4M} \sum_{m=1}^M \left\| \x^{(r,k)}_m - \overline{\x^{(r,k)}} \right\|_2^4.
  \label{eq:fedavg:3o:4cm}
\end{equation}
The fourth-order central moment term appeared in \cref{eq:fedavg:3o:4cm} can be upper bounded by the following lemma.
\begin{lemma}[label=fedavg:a2:stab:main,restate=LemFedAvgThirdOrderFourthCM,name=\nth{4}-order stability]%[\nth{4}-order discrepancy overhead bound for \fedavg]
  In the same settings of \cref{thm:fedavg:3o:ub}, for any $r<R$, $k<K$, and $m \in [M]$, the following inequality holds.
  \begin{align*}
    \expt \left[ \frac{1}{M} \sum_{m=1}^M \left\| \overline{\x^{(r,k)}} - \x^{(r,k)}_m \right\|_2^4 \right]
    \leq
    192 \eta^4 K^2 \sigma^4.
  \end{align*}
\end{lemma}
The proof of \cref{fedavg:a2:stab:main} is deferred to the end of this section (see \cref{sec:pf:fedavg:3o:ub}).

Now we plug in \cref{eq:fedavg:3o:ub:2,eq:fedavg:3o:4cm} and apply \cref{fedavg:a2:stab:main} to \cref{eq:fedavg:3o:ub:1}:
\begin{align*}
  \expt \left[ \| \overline{\x^{(r,k+1)}} - \x^{\star} \|_2^2 \right]
  % \\
  % \leq & \left( 1 + \frac{1}{KR} \right) \expt[\| \overline{\x^{(r,k)}} - \x^{\star} \|_2^2 ] - \eta  \left( \expt[F(\overline{\x^{(r,k)}} )] - F^{\star} \right) 
  % + 96 \eta^6 Q^2 K^3 R \sigma^4
  % + \frac{\eta^2 \sigma^2}{M}
  % \\
  \leq & \left( 1 + \frac{1}{KR} \right) \expt[\| \overline{\x^{(r,k)}} - \x^{\star} \|_2^2 ]
  \\
  & \qquad - \eta  \left[ \expt[F(\overline{\x^{(r,k)}} ) ]- F^{\star}  -  96 \eta^5 Q^2 K^3 R \sigma^4 - \frac{\eta \sigma^2}{M}  \right]
\end{align*}
% \begin{align}
%   & \frac{1}{(1+\eta \delta)^{k+1}}\expt \left[ \| \overline{\x^{(r,k+1)}} - \x^{\star} \|_2^2 \middle| \mathcal{F}^{(r,k)} \right]
%   \\
%   \leq & 
%   \frac{1}{(1+\eta \delta)^k} \| \overline{\x^{(r,k)}} - \x^{\star} \|_2^2-  \frac{\eta}{(1+\eta \delta)^{k+1}} 
%   \left[ F(\overline{\x^{(r,k)}} ) - F^{\star} - 96 \eta^4 Q^2 K^2 \sigma^4 - \frac{\eta \sigma^2}{M}\right].
% \end{align}
Recursing with respect to $k$ from 0 to $K$:
\begin{align*}
  & \frac{1}{\left( 1 + \frac{1}{KR} \right)^K}  \expt \left[ \| \overline{\x^{(r+1,0)}} - \x^{\star} \|_2^2  \right] = 
  \frac{1}{\left( 1 + \frac{1}{KR} \right)^K}  \expt \left[ \| \overline{\x^{(r,K)}} - \x^{\star} \|_2^2  \right]
  \\
  \leq &
   \expt[ \| \overline{\x^{(r,0)}} - \x^{\star} \|_2^2] - \sum_{k=0}^{K-1} \eta \left( 1 + \frac{1}{KR} \right)^{-(k+1)}
  \left[ \expt[ F(\overline{\x^{(r,k)}} )] - F^{\star} -  96 \eta^5 Q^2 K^3 R \sigma^4  - \frac{\eta \sigma^2}{M}\right].
\end{align*}
% Attaching
% \begin{align}
%   & \frac{1}{(1+\eta \delta)^{(r+1)K}} \expt \left[ \| \overline{\x^{(r+1,0)}} - \x^{\star} \|_2^2 \middle| \mathcal{F}^{(r,0)} \right]
%   \\
%   \leq &
%   \frac{1}{(1+\eta \delta)^{rK}}  \| \overline{\x^{(r,0)}} - \x^{\star} \|_2^2 - \sum_{k=0}^{K-1} \frac{\eta}{(1 + \eta \delta)^{rK + k+1}}
%   \left[ F(\overline{\x^{(r,k)}} ) - F^{\star} - 96 \eta^4 Q^2 K^2 \sigma^4 - \frac{\eta \sigma^2}{M}\right].
% \end{align}
Further recurse with respect to $r$ from $0$ to $R$ and drop the final term, one has
\begin{align*}
  % & \frac{1}{(1+\eta \delta)^{RK}} \expt \left[ \| \overline{\x^{(R,0)}} - \x^{\star} \|_2^2 \right]
  % \\
  0 \leq B^2 - \sum_{r=0}^{R-1} \sum_{k=0}^{K-1} \eta \left( 1 + \frac{1}{KR} \right)^{-(rK + k+1)}
  \left[ \expt[F(\overline{\x^{(r,k)}})] - F^{\star} -  96 \eta^5 Q^2 K^3 R \sigma^4  - \frac{\eta \sigma^2}{M}\right]
\end{align*}
where recall $B$ is defined by $\| \overline{\x^{(0,0)}} - \x^{\star} \|_2$.
% Rearranging
% \begin{equation}
%   \sum_{r=0}^{R-1} \sum_{k=0}^{K-1}  \frac{\expt[F(\overline{\x^{(r,k)}})] - F^{\star}}{(1 + \frac{1}{KR})^{rK + k+1}} 
%   \leq
%   \frac{1}{\eta}  \| \overline{\x^{(0,0)}} - \x^{\star} \|_2^2 +  \left( \sum_{r=0}^{R-1} \sum_{k=0}^{K-1}  \frac{1}{(1 + \frac{1}{KR})^{rK + k+1}}  \right) \left(  96 \eta^5 Q^2 K^3 R \sigma^4 + \frac{\eta \sigma^2}{M} \right)
% \end{equation}
Consequently by definition of $\hat{\x}$ and convexity of $F$:
\begin{align*}
  & \expt[F(\hat{\x})] - F^{\star} \leq 
  \left( \sum_{r=0}^{R-1} \sum_{k=0}^{K-1} \frac{1}{(1+\frac{1}{KR})^{rK+k+1}}  \right)^{-1}
   \left( \sum_{r=0}^{R-1} \sum_{k=0}^{K-1} \frac{ \expt[F(\overline{\x^{(r,k)}})] - F^{\star}}{{(1+\frac{1}{KR})^{rK+k+1}}}  \right)
  \\
  \leq & \frac{1}{\eta}\left( \sum_{r=0}^{R-1} \sum_{k=0}^{K-1} \frac{1}{(1+\frac{1}{KR})^{rK+k+1}}  \right)^{-1}   B^2 +   96 \eta^5 Q^2 K^3 R \sigma^4 + \frac{\eta \sigma^2}{M}.
  \\
  \leq & \frac{3}{\eta KR }  B^2 +   96 \eta^5 Q^2 K^3 R \sigma^4 + \frac{\eta \sigma^2}{M},
\end{align*}
where the last inequality is due to $(1 + \frac{1}{KR})^{rK+k+1} \leq (1 + \frac{1}{KR})^{KR} \leq \euler < 3$ for any $r < R$ and $k < K$.

Furthermore, when the step size $\eta$ is chosen as
\begin{equation*}
  \eta = \min \left\{ \frac{1}{2L}, \frac{M^{\frac{1}{2}} B}{K^{\frac{1}{2}} R^{\frac{1}{2}} \sigma}, \frac{B^{\frac{1}{3}}}{K^{\frac{2}{3}} R^{\frac{1}{3}} Q^{\frac{1}{3}} \sigma^{\frac{2}{3}}} \right\},
\end{equation*}
we obtain the upper bound as stated in \cref{thm:fedavg:3o:ub}.
\end{proof}

\subsubsection{Deferred Proof of \cref{fedavg:a2:stab:main}}
\label{sec:pf:fedavg:3o:ub}

In this subsection we prove \cref{fedavg:a2:stab:main} regarding the 4th order stability of \fedavg.

% We restate the lemma for the reader's reference.
% \LemFedAvgThirdOrderFourthCM*
% We introduce a few more notations to simplify the discussions. 
% Let $m_1, m_2 \in [M]$ be two arbitrary distinct clients. 
We first state and prove the following lemma on one-step 4th-order stability. The proof is analogous to the \nth{4}-order convergence analysis of \fedavg in \cite{Dieuleveut.Patel-NeurIPS19}.
\begin{lemma}
  \label{fedavg:a2:stab:1}
  In the same setting of \cref{fedavg:a2:stab:main}, for any $r, k$, for any $m_1, m_2 \in [M]$, the following inequality holds,
  \begin{equation*}
    \sqrt{\expt \left\| \x_{m_1}^{(r,k+1)} - \x_{m_2}^{(r,k+1)} \right\|_2^4}
    \leq
    \sqrt{\expt \left\| \x_{m_1}^{(r,k)} - \x_{m_2}^{(r,k)} \right\|_2^4} + \sqrt{192} \eta^2 \sigma^2.
  \end{equation*}
\end{lemma}

\begin{proof}[Proof of \cref{fedavg:a2:stab:1}]
  We introduce some local notations to simplify the presentation. 
  For any $(r, k)$ pair, let $\bDelta^{(r,k)} := \x_{m_1}^{(r,k)} - \x_{m_2}^{(r,k)}$, and $\bDelta_{\varepsilon}^{(r,k)} := \varepsilon_{m_1}^{(r,k)} - \varepsilon_{m_2}^{(r,k)}$ where $\varepsilon^{(r,k)}_m := \nabla f(\x_m^{(r,k)}; \xi_m^{(r,k)}) - \nabla F(\x_m^{(r,k)})$. Let $\bDelta_{\nabla}^{(r,k)} := \nabla F(\x^{(r,k)}_{m_1}) - \nabla F(\x^{(r,k)}_{m_2})$. Then
  \begin{align}
         & \expt [\|\bDelta^{(r,k+1)}\|_2^4 |\mathcal{F}^{(r,k)}] = \expt \left[ \|\bDelta^{(r,k)} - \eta (\bDelta^{(r,k)}_{\nabla} + \bDelta^{(r,k)}_{\varepsilon}) \|_2^4 |\mathcal{F}^{(r,k)} \right]
      \nonumber
    \\
    =    & \expt \left[ \left( \|\bDelta^{(r,k)}\|_2^2- 2 \eta \langle \bDelta^{(r,k)}, \bDelta^{(r,k)}_{\nabla} + \bDelta^{(r,k)}_{\varepsilon} \rangle + \eta^2 \|\bDelta^{(r,k)}_{\nabla} + \bDelta^{(r,k)}_{\varepsilon}\|_2^2\right)^2 \middle| \mathcal{F}^{(r,k)} \right]
      \nonumber
    \\
    =    & \expt \|\bDelta^{(r,k)}\|_2^4 - 4\eta \|_2\bDelta^{(r,k)}\|_2^2\langle \bDelta^{(r,k)}, \bDelta^{(r,k)}_{\nabla} \rangle
    + 4 \eta^2 \expt \left[ \langle \bDelta^{(r,k)}, \bDelta^{(r,k)}_{\nabla} + \bDelta^{(r,k)}_{\varepsilon} \rangle^2 |\mathcal{F}^{(r,k)} \right]
      \nonumber
    \\  & \quad
    + 2 \eta^2 \|\bDelta^{(r,k)}\|_2^2 \expt \left[ \| \bDelta^{(r,k)}_{\nabla} + \bDelta^{(r,k)}_{\varepsilon} \|_2^2|\mathcal{F}^{(r,k)} \right]     
    + \eta^4 \expt \left[ \| \bDelta^{(r,k)}_{\nabla} + \bDelta^{(r,k)}_{\varepsilon} \|_2^4 |\mathcal{F}^{(r,k)} \right]
      \nonumber
    \\
         & \quad - 4 \eta^3 \expt \left[ \langle \bDelta^{(r,k)}, \bDelta^{(r,k)}_{\nabla} + \bDelta^{(r,k)}_{\varepsilon} \rangle  \cdot \| \bDelta^{(r,k)}_{\nabla} + \bDelta^{(r,k)}_{\varepsilon} \|_2^2|\mathcal{F}^{(r,k)}  \right]
      \nonumber
    \\
    \leq & \expt \|\bDelta^{(r,k)}\|_2^4 - 4\eta \|_2\bDelta^{(r,k)}\|_2^2\langle \bDelta^{(r,k)}, \bDelta^{(r,k)}_{\nabla} \rangle + 6 \eta^2 \|\bDelta^{(r,k)}\|_2^2 \expt \left[ \| \bDelta^{(r,k)}_{\nabla} + \bDelta^{(r,k)}_{\varepsilon} \|_2^2|\mathcal{F}^{(r,k)} \right]
      \nonumber
    \\
         & \quad + 4 \eta^3 \|\bDelta^{(r,k)}\| \expt  \left[ \| \bDelta^{(r,k)}_{\nabla} + \bDelta^{(r,k)}_{\varepsilon} \|_2^3 |\mathcal{F}^{(r,k)} \right]
    + \eta^4 \expt \left[ \| \bDelta^{(r,k)}_{\nabla} + \bDelta^{(r,k)}_{\varepsilon} \|_2^4 |\mathcal{F}^{(r,k)} \right] \tag{Cauchy-Schwarz inequality}
    \\
    \leq & \expt \|\bDelta^{(r,k)}\|_2^4 - 4\eta \|_2\bDelta^{(r,k)}\|_2^2\langle \bDelta^{(r,k)}, \bDelta^{(r,k)}_{\nabla} \rangle
    + 8 \eta^2  \|\bDelta^{(r,k)}\|_2^2 \expt \left[ \| \bDelta^{(r,k)}_{\nabla} + \bDelta^{(r,k)}_{\varepsilon} \|_2^2|\mathcal{F}^{(r,k)} \right]
    \nonumber
    \\
    & \quad
    + 3 \eta^4 \expt \left[ \| \bDelta^{(r,k)}_{\nabla} + \bDelta^{(r,k)}_{\varepsilon} \|_2^4 |\mathcal{F}^{(r,k)} \right],
    \label{eq:fedavg:a2:stab:prop:1}
  \end{align}
  where the last inequality is due to
  \begin{align*}
    & 4 \eta^3 \|\bDelta^{(r,k)}\| \expt  \left[ \| \bDelta^{(r,k)}_{\nabla} + \bDelta^{(r,k)}_{\varepsilon} \|_2^3 |\mathcal{F}^{(r,k)} \right]
    \\
     \leq & 2 \eta^2 \|\bDelta^{(r,k)}\|_2^2 \expt  \left[ \| \bDelta^{(r,k)}_{\nabla} + \bDelta^{(r,k)}_{\varepsilon} \|_2^2|\mathcal{F}^{(r,k)} \right] +
    2 \eta^4 \expt  \left[ \| \bDelta^{(r,k)}_{\nabla} + \bDelta^{(r,k)}_{\varepsilon} \|_2^4 |\mathcal{F}^{(r,k)} \right]
  \end{align*}
  by AM-GM inequality.

  Note that by $L$-smoothness and convexity, we have the following inequality by standard convex analysis (\cf, Theorem 2.1.5 of \cite{Nesterov-18}),
  \begin{align}
    & \|\bDelta^{(r,k)}_{\nabla}\|_2^2 
    =
    \| \nabla F(\x_{m_1}^{(r,k)}) - \nabla F(\x_{m_2}^{(r,k)})\|_2^2 
    \nonumber \\
    \leq &
    L \left\langle \x_{m_1}^{(r,k)} - \x_{m_2}^{(r,k)},  \nabla F(\x_{m_1}^{(r,k)}) - \nabla F(\x_{m_2}^{(r,k)})  \right\rangle
    =
    L \langle \bDelta^{(r,k)}, \bDelta^{(r,k)}_{\nabla} \rangle.
    \label{eq:fedavg:a2:stab:prop:2}
  \end{align}
  Consequently
  \begin{align*}
    & \expt \left[ \| \bDelta^{(r,k)}_{\nabla} + \bDelta^{(r,k)}_{\varepsilon} \|_2^2|\mathcal{F}^{(r,k)} \right] = \| \bDelta^{(r,k)}_{\nabla}\|_2^2+  \expt \left[ \|\bDelta^{(r,k)}_{\varepsilon}\|_2^2|\mathcal{F}^{(r,k)} \right]
    \\
    \leq & \| \bDelta^{(r,k)}_{\nabla}\|_2^2+ 2\sigma^2
    \leq L \langle \bDelta^{(r,k)}, \bDelta^{(r,k)}_{\nabla} \rangle + 2\sigma^2.
  \end{align*}
  Similarly
  \begin{align}
         & \expt \left[ \| \bDelta^{(r,k)}_{\nabla} + \bDelta^{(r,k)}_{\varepsilon} \|_2^4 |\mathcal{F}^{(r,k)} \right]
    \leq
    8 \|\bDelta^{(r,k)}_{\nabla}\|_2^4 + 8 \expt \left[ \|\bDelta^{(r,k)}_{\varepsilon}\|_2^4 |\mathcal{F}^{(r,k)} \right]
    \tag{AM-GM inequality}
    \\
    \leq &
    8 \|\bDelta^{(r,k)}_{\nabla}\|_2^4  + 64 \sigma^4
    \tag{by helper \cref{helper:diff:4th}}
    \\
    \leq & 8L^2 \|\bDelta^{(r,k)}\|_2^2\|\bDelta^{(r,k)}_{\nabla}\|_2^2+ 64 \sigma^4
    \tag{by $L$-smoothness}
    \\
    \leq & 8L^3 \|\bDelta^{(r,k)}\|_2^2\langle \bDelta^{(r,k)}, \bDelta^{(r,k)}_{\nabla} \rangle + 64 \sigma^4.
    \tag{by \cref{eq:fedavg:a2:stab:prop:2}}
  \end{align}
  Plugging the above two bounds to \cref{eq:fedavg:a2:stab:prop:1} gives
  \begin{align}
    & \expt [\|\bDelta^{(r,k+1)}\|_2^4 |\mathcal{F}^{(r,k)}]
    \nonumber \\
    \leq &
    \|\bDelta^{(r,k)}\|_2^4 - 4\eta (1 - 2 \eta L - 6 \eta^3 L^3) \|\bDelta^{(r,k)}\|_2^2\langle \bDelta^{(r,k)}, \bDelta^{(r,k)}_{\nabla} \rangle + 16 \eta^2 \|\bDelta^{(r,k)}\|_2^2\sigma^2 + 192 \eta^4 \sigma^4.
    \label{eq:fedavg:a2:stab:prop:3}
  \end{align}
  Since $\eta L \leq \frac{1}{4}$ we have $(1 - 2 \eta L - 6 \eta^3 L^3) > 0$. By convexity $\langle \bDelta^{(r,k)}, \bDelta^{(r,k)}_{\nabla} \rangle \geq 0$. Hence the second term on the RHS of \cref{eq:fedavg:a2:stab:prop:3} is non-positive. We conclude that
  \begin{align*}
    \expt [\|\bDelta^{(r,k+1)}\|_2^4 |\mathcal{F}^{(r,k)}]
    \leq
    \|\bDelta^{(r,k)}\|_2^4 + 16 \eta^2  \sigma^2 \|\bDelta^{(r,k)}\|_2^2+ 192 \eta^4 \sigma^4.
  \end{align*}
  Taking expectation gives
  \begin{align*}
         & \expt [\|\bDelta^{(r,k+1)}\|_2^4] \leq \expt [ \|\bDelta^{(r,k)}\|_2^4 ] + 16 \eta^2 \sigma^2 \expt[\|\bDelta^{(r,k)}\|_2^2 ] + 192 \eta^4 \sigma^4
    \\
    \leq & \expt [ \|\bDelta^{(r,k)}\|_2^4 ] + 16 \eta^2 \sigma^2 \sqrt{\expt[\|\bDelta^{(r,k)}\|_2^4]} + 192 \eta^4 \sigma^4 = \left( \sqrt{\expt \|\bDelta^{(r,k)}\|_2^4} + \sqrt{192} \eta^2 \sigma^2 \right)^2.
  \end{align*}
  Taking square root on both sides completes the proof.
\end{proof}
With \cref{fedavg:a2:stab:1} at hand we are ready to prove \cref{fedavg:a2:stab:main}.
\begin{proof}[Proof of \cref{fedavg:a2:stab:main}]
  Telescoping \cref{fedavg:a2:stab:1} yields (note that $\bDelta^{(r,0)} = \zeros$)
  \begin{equation*}
    \sqrt{\expt \|\bDelta^{(r,k)}\|_2^4} \leq \sqrt{192} \eta^2 \sigma^2 k \leq \sqrt{192} \eta^2 K \sigma^2,
  \end{equation*}
  where the last inequality is because of $k \leq K$. Thus
  \begin{align*}
    \frac{1}{M} \sum_{m=1}^M \expt \left[ \left\| \overline{\x^{(r,k)}} - \x^{(r,k)}_m \right\|_2^4  \right] \leq \expt [\|\bDelta^{(r,k)}\|_2^4] \leq 192 \eta^4 K^2 \sigma^4,
  \end{align*}
  where the first ``$\leq$'' is due to Jensen's inequality.
\end{proof}

\section{\fedavg in Non-Convex Setting}
\label{sec:fedavg:ncvx}
In this section, we show that the advantage of third-order smoothness for \fedavg can be extended to non-convex settings. 

In the non-convex setting, akin to some other work in FL literature~\cite{Yu.Yang.ea-AAAI19,Reddi.Charles.ea-ICLR21}, we require an assumption bounding the norm of stochastic gradients. Note that this is stronger that \cref{asm:fo:2o} which bounds the \emph{variance} of the stochastic gradients. We remark that several other works impose weaker assumptions, though the algorithms they consider are different, or their results are weaker \cite{Stich-ICLR19, Koloskova.Loizou.ea-ICML20, Wang.Joshi-JMLR21}.
\begin{assumption}
  \label{asm:fedavg:ncvx}
  Consider the federated optimization problem \eqref{eq:fo:hetero}. 
  Assume that $f(\x; \xi)$ is second-order continuously differentiable w.r.t. $\x \in \reals^d$ for any $\xi$, and that for any client $m \in [M]$,
  \begin{enumerate}[(a)]
    \item $F_m(\x)$ is $L$-smooth. That is, for any $\x, \y \in \reals^d$, we have 
    \begin{equation*}
    \| \nabla F_m(\x) - \nabla F_m(\y) \|_2 \leq L\|\x - \y\|_2.
    \end{equation*}    
    \item For any $\x \in \reals^d$, $\expt_{\xi \sim \mathcal{D}_m} \|\nabla f(\x; \xi) - \nabla F_m(\x)\|_2^2 \leq \sigma^2.$
    \item For any $\x$ and $\xi$, it is the case that $\|\nabla f(\x; \xi)\|_2 \leq G$.
  \end{enumerate}
\end{assumption}

The best-known rate for \fedavg with non-convex objectives (under second-order smoothness alone) is due to \cite{Yu.Yang.ea-AAAI19}, which we quote as follows.
% \footnote{There are other extensions of \fedavg that can outperform this rate, e.g., \fedavg with server learning rate discussed in \cref{sec:related-work},
% since it includes mini-batch SGD as a special case.} 
Note that this rate is not explicitly given in their paper, but can be proved from their work by setting the step size $\eta$ appropriately. For completeness, we prove this rate in \cref{sec:pf:fedavg:ncvx:2o} for completeness.
\begin{theorem}[label=thm:fedavg:ncvx:2o,name={Upper bound for \fedavg with non-Convex objectives under second-order smoothness},restate=ThmFedAvgNcvxSO]
  Consider the homogeneous federated optimization problem satisfying \cref{asm:fedavg:homo,asm:fedavg:ncvx}. Then there exists a step-size $\eta$ such that \fedavg satisfies
  % \begin{compact}
  % \begin{equation}
  %   \expt\left[\left\|\nabla f(\hat{\x})\right\|_2^2 \right]\leq \bigo \left(\frac{LB}{KR} 
  %   + \frac{G \sqrt{BL}}{\sqrt{MKR}} 
  %   +  \frac{B^{\frac{2}{3}}G^{\frac{2}{3}}L^{\frac{2}{3}}}{R^{\frac{2}{3}}} 
  %   \right).
  % \end{equation}
  % \end{compact}
  % \hynote{Remove this bound (or move it to the remark)}
  \begin{equation*}
    \expt\left[\left\|\nabla F(\hat{\x})\right\|_2^2 \right]\leq
    \bigo \left(
      \frac{L \Delta}{KR} 
      + \frac{G \sqrt{L \Delta}}{\sqrt{MKR}} 
      +  \frac{L^{\frac{2}{3}} G^{\frac{2}{3}} \Delta^{\frac{2}{3}}}{R^{\frac{2}{3}}}
    \right),
  \end{equation*}
  where $\hat{\x} := \frac{1}{M}\sum_m{\x_m^{(r, k)}}$ for a uniformly random choice of $k \in \{0, 1, \ldots, K-1\}$, and $r \in \{0, 1, \ldots, R-1\}$, and $\Delta := F(\x^{(0, 0)}) - \inf_{\x} F(\x)$.
\end{theorem}

In contrast, under third-order smoothness assumption (\cref{asm:fo:3o}), we establish the following theorem:
\begin{theorem}[label=thm:fedavg:ncvx:3o,name={Upper bound for \fedavg with non-Convex objectives under third-order smoothness},restate=ThmFedAvgNcvxTO]
Consider the homogeneous federated optimization problem satisfying \cref{asm:fedavg:homo,asm:fedavg:ncvx,asm:fo:3o}. 
Then there exists a step-size $\eta$ such that \fedavg satisfies
% \begin{compact}
% \begin{equation}
%   \expt\left[\left\|\nabla f(\hat{\x})\right\|_2^2 \right]\leq \bigo \left(\frac{LB}{KR} 
%   + \frac{G \sqrt{BL}}{\sqrt{MKR}} 
%   +  \frac{B^{\frac{2}{3}}G^{\frac{2}{3}}L^{\frac{2}{3}}}{R^{\frac{2}{3}}} 
%   \right).
% \end{equation}
% \end{compact}
% \hynote{Remove this bound (or move it to the remark)}
\begin{equation*}
  \expt\left[\left\|\nabla F(\hat{\x})\right\|_2^2 \right]\leq
  \bigo \left(
  \frac{L\Delta}{KR}
  + 
  \frac{G\sqrt{L \Delta}}{\sqrt{MKR}}
  +
  \frac{Q^{\frac{2}{5}} G^{\frac{4}{5}} \Delta^{\frac{4}{5}}}{R^{\frac{4}{5}}}
  \right),
\end{equation*}
where $\hat{\x} := \frac{1}{M}\sum_m{\x_m^{(r, k)}}$ for a uniformly random choice of $k \in \{0, 1, \ldots, K-1\}$, and $r \in \{0, 1, \ldots, R-1\}$, and $\Delta := F(\x^{(0, 0)}) - \inf_{\x} F(\x)$.
\end{theorem}
The proof is relegated to \cref{sec:pf:fedavg:ncvx:3o}. 
Observe that we improve the dependence from $R^{\frac{2}{3}}$ in the third term to $R^{\frac{4}{5}}$. 
This theorem shows that the convergence rate of \fedavg improves substantially under third order smoothness. 

\chapter{Principled Acceleration of Federated Averaging}
\label{chapter:fedac}
In this chapter, we focus on the homogeneous version of the problem considered in \cref{eq:fo:hetero}, namely
\begin{equation}
    \min F(\x) := \expt_{\x \sim \dist} f(\x; \xi),
    \label{eq:fo:homo}
\end{equation}
where each client $m \in [M]$ can access the stochastic gradient oracle $\nabla f(\x; \xi)$ by drawing independent sample $\xi$ from the shared distribution $\dist$.

We propose a principled acceleration for \fedavg, namely \emph{\fedacfull} (\fedac), which provably improves convergence rate and communication efficiency. 
Our result extends the results of \cite{Woodworth.Patel.ea-ICML20} on \textsc{Local-Ac-Sa} for quadratic objectives to broader objectives.
To the best of our knowledge, this is the first provable acceleration of \fedavg (and its variants) for general or strongly convex objectives.
\fedac parallelizes a generalized version of Accelerated SGD \cite{Ghadimi.Lan-SIOPT12}, while we carefully 
balance the acceleration-stability tradeoff to accommodate distributed settings.
Under standard assumptions on homogeneity, smoothness, bounded variance, and strong convexity (see \cref{asm:fo:scvx:2o} for details), \fedac converges at rate $\tildeo ( \frac{1}{M K R} + \frac{1}{K R^4} )$.\footnote{We hide varaibles other than $K, M, R$ for simplicity. The complete bound can be found in \cref{tab:conv:rate} and the corresponding theorems.}
The bound will be dominated by $\tildeo(\frac{1}{M K R})$ for $R$ as low as $\tildeo(M^{\frac{1}{3}})$, which implies the synchronization $R$ required for linear speedup in $M$ is $\tildeo(M^{\frac{1}{3}})$.\footnote{
    ``Communication required for linear speedup'' is a simple and common measure of the communication efficiency, which can be derived from the raw convergence rate. It is defined as the minimum number of synchronization $R$, as a function of number of clients $M$ and parallel runtime $T$, required to achieve a linear speed up --- the parallel runtime of $M$ clients is equal to the $\nicefrac{1}{M}$ fraction of a sequential single client runtime.}
In comparison, the state-of-the-art \fedavg analysis \cite{Khaled.Mishchenko.ea-AISTATS20} showed that \fedavg converges at rate $\tildeo ( \frac{1}{M K R} + \frac{1}{KR^2} )$, which requires $\tildeo(M)$ synchronization for linear speedup.
For general convex objective, \fedac converges at rate $\tildeo(\frac{1}{\sqrt{M K R}} + \frac{1}{K^{\frac{1}{3}} R})$, which outperforms both state-of-the-art \fedavg $\tildeo(\frac{1}{\sqrt{M K R}} + \frac{1}{K^{\frac{1}{3}} R^{\frac{2}{3}}})$ by \cite{Woodworth.Patel.ea-ICML20} and Minibatch-SGD baseline ${\Theta}(\frac{1}{\sqrt{M K R}} + \frac{1}{R})$ \cite{Dekel.Gilad-Bachrach.ea-JMLR12}.\footnote{
    Minibatch-SGD baseline corresponds to running SGD for $R$ steps with batch size $MT/R$, which can be implemented on $M$ parallel clients with $R$ communication and each client queries $T$ gradients in total.
    }
We summarize the convergence rates in \cref{tab:conv:rate} (on the row marked A\ref{asm:fo:scvx:2o}).

\begin{table}
    \caption{\textbf{Summary of results on convergence rates.} All bounds omit multiplicative $\polylog$ factors and additive exponential decaying term (for strongly convex objective) for ease of presentation. 
    Notation: $B$: $\|\x^{(0,0)} - \x^{\star}\|_2$; $M$: number of clients; $R$: number of communication rounds; $K$: number of local steps per round;  $\mu$: strong convexity; $L$: smoothness; $Q$: \nth{3}-order-smoothness (in \cref{asm:fo:scvx:3o}).
    }
    \label{tab:conv:rate}
    \centering
    \resizebox{\linewidth}{!}{
    \begin{tabular}{llll}
        \toprule
        Assumption              & Algorithm                     & Convergence Rate $(\expt[F(\cdot)] - F^{\star} \leq \cdots $) & Reference
        \\
        \midrule
        A\ref{asm:fo:scvx:2o}($\mu > 0$) & \fedavg                       &
        exp. decay 
        $+ \frac{\sigma^2}{\mu M K R} +  \frac{L \sigma^2}{\mu^2 K R^2} $
                               & \cite{Woodworth.Patel.ea-ICML20}
        \\
                               & \fedac                        &
        exp. decay $+ \frac{\sigma^2}{\mu M K R} +  
        \min \left\{ \frac{L \sigma^2}{\mu^2 K R^3}, \frac{L^2 \sigma^2}{\mu^3 K R^4} \right\}$
                               & \textbf{\cref{fedac:a1}}
        \\
        \midrule
        A\ref{asm:fo:scvx:3o}($\mu > 0$) & \fedavg                       &
        exp. decay $+ \frac{\sigma^2}{\mu M K R} + \frac{Q^2 \sigma^4}{\mu^5 K^2 R^4}$  
                               & \textbf{\cref{fedavg:a2}}
        \\
                               & \fedac                        &
        exp. decay $+ \frac{\sigma^2}{\mu M K R} 
        + {\frac{Q^2 \sigma^4}{\mu^5 K^2 R^8}}$
                               & \textbf{\cref{fedacii:a2}}
        \\
        \midrule
        A\ref{asm:fo:scvx:2o}($\mu = 0$) & \fedavg                       &
        $\frac{LB^2}{K R} + \frac{\sigma B}{\sqrt{M K R}} + \frac{L^{\frac{1}{3}} \sigma^{\frac{2}{3}} B^{\frac{4}{3}}}{K^{\frac{1}{3}} R^{\frac{2}{3}}}$
                               & \cref{thm:fedavg:2o:ub}, adapted from \cite{Woodworth.Patel.ea-ICML20}
        \\
                               & \fedac                        &
        $\frac{LB^2}{K R^2} + \frac{\sigma B}{\sqrt{M K R}} + \min \left\{\frac{L^{\frac{1}{3}} \sigma^{\frac{2}{3}} B^{\frac{4}{3}}}{K^{\frac{1}{3}} R}, \frac{L^{\frac{1}{2}} \sigma^{\frac{1}{2}} B^{\frac{3}{2}}}{K^{\frac{1}{4}} R} \right\} $
                               & \textbf{\cref{fedaci:a1:gcvx,fedacii:a1:gcvx}}
        \\
        \midrule
        A\ref{asm:fo:scvx:3o}($\mu = 0$) & \fedavg                       &
        $\frac{LB^2}{K R} + \frac{\sigma B}{\sqrt{M K R}} + \frac{Q^{\frac{1}{3}} \sigma^{\frac{2}{3}} B^{\frac{5}{3}}}{K^{\frac{1}{3}} R^{\frac{2}{3}}}$
                               & \cref{thm:fedavg:3o:ub}
        \\
                               & \fedac                        &
        $\frac{LB^2}{K R^2} + \frac{\sigma B}{\sqrt{M K R}} + \frac{L^{\frac{1}{3}} \sigma^{\frac{2}{3}} B^{\frac{4}{3}}}{M^{\frac{1}{3}} K^{\frac{1}{3}} R} +
            \frac{Q^{\frac{1}{3}} \sigma^{\frac{2}{3}} B^{\frac{5}{3}}}{K^{\frac{1}{3}} R^{\frac{4}{3}}}$
                               & \textbf{\cref{fedacii:a2:gcvx}}
        \\
        \bottomrule
    \end{tabular}
    }
\end{table}

Analogous to \fedavg discussed in \cref{chapter:fedavg}, we also establish stronger guarantees for \fedac when objectives are \nth{3}-order-smooth, or ``close to be quadratic'' intuitively.
For strongly convex objectives, \fedac converges at rate $\tildeo (\frac{1}{M K R} + \frac{1}{K^2 R^8})$ (see \cref{fedacii:a2}).
We summarize our results in \cref{tab:conv:rate} (on the row marked A\ref{asm:fo:scvx:3o}).

\section{Preliminaries}
% In this chapter, unlike the previous one, we will focus on the convergence of \fedac 
% In this chatper
In this chapter, we will study both strongly-convex and non-strongly-convex settings.
% , which will be clarified in the context. 
To keep the dissertation focused, we will mostly analyze the algorithm under the strongly-convex setting, and then analyze the induced non-strongly-convex rate via an $\ell_2$-augmented approach.
% \hynote{TODO} 

% To incorporate the strongly-convex assumption, we formally state the following set of assumption.
We start by considering the assumption akin to \cref{asm:fo:2o} in \cref{chapter:fedavg}, with strong convexity incorporated.
% In this chapter, we will first focus on the strongly-convex version of \cref{asm:fo:2o}
\begin{assumption}[$\mu$-strong convexity, $L$-smoothness and $\sigma^2$-uniformly bounded gradient variance]
    \label{asm:fo:scvx:2o}
    Consider the homogeneous federated optimization problem \eqref{eq:fo:homo}, assume that
    \begin{enumerate}[(a)]
        \item $F$ is $\mu$-strongly convex. That is, for any $\x, \y \in \reals^d$, we have
        \begin{equation*}
            F(\y) \geq F(\x) + \langle \nabla F(\x), \y - \x  \rangle + \frac{1}{2} \mu \|\y - \x\|_2^2.
        \end{equation*}
        In addition, assume $F$ attains a finite optimum $\x^{\star} \in \reals^d$.
        % (We will study both the strongly convex case $(\mu > 0)$ and the general convex case $(\mu = 0)$, which will be clarified in the context.)
        \item $F$ is $L$-smooth. That is, for any $\x, \y \in \reals^d$, we have
        \begin{equation*}
            F(\y) \leq F(\x) + \langle \nabla F(\x), \y - \x  \rangle + \frac{1}{2} L \|\y-\x\|_2^2
        \end{equation*}
        \item $\nabla f(\x; \xi)$ has $\sigma^2$-bounded variance. That is, 
        \begin{equation*}
        \sup_{\x \in \reals^d}  \expt_{\xi \in \mathcal{D}} \| \nabla f(\x; \xi) - \nabla F(\x)\|_2^2  \leq \sigma^2.
        \end{equation*}
    \end{enumerate}
\end{assumption}

% We conduct our analysis on \fedac in two settings with two sets of assumptions.
% The following \cref{asm:fo:scvx:2o} consists of a set of standard assumptions: convexity, smoothness and bounded variance.

The following \cref{asm:fo:scvx:3o}, akin to \cref{asm:fo:3o}, consists of an additional set of assumptions: \nth{3} order smoothness and bounded $\nth{4}$ central moment.
\begin{assumption}\label{asm:fo:scvx:3o} 
    Consider the homogeneous federated optimization problem \eqref{eq:fo:homo}.    
    In addition to \cref{asm:fo:scvx:2o}, assume that
    \begin{enumerate}
        \item[(a)] $F$ is $Q$-\nth{3}-order-smooth. That is, for any $\x, \y \in \reals^d$, we have 
        \begin{equation*}
            F(\y) \leq F(\x) + \langle \nabla F(\x), \y - \x\rangle +  \frac{1}{2} \langle \nabla^2 F(\x) (\y - \x), (\y-\x) \rangle + \frac{1}{6} Q \|\y-\x\|_2^3.
        \end{equation*}
        \item[(b)] $\nabla f(\x; \xi)$ has $\sigma^4$-bounded \nth{4} central moment. That is, 
        \begin{equation*}
            \sup_{\x \in \reals^d} \expt_{\xi \in \mathcal{D}} \| \nabla f(\x; \xi) - \nabla F(\x)\|_2^4 \leq \sigma^4
        \end{equation*}
    \end{enumerate}
\end{assumption}

% \subsection{Notations}
% Let $\x^{\star}$ be the optimum of $F$ and denote $F^{\star}:=F(\x^{\star})$. 
% Let $B := \|\x^{(0,0)} - \x^{\star}\|_2$.
% be the Euclidean distance of the initial guess $\x^{(0,0)}$ and the optimum $\x^{\star}$.
% For both \fedac and \fedavg, we use $M$ to denote the number of parallel clients, $R$ to denote the number of rounds, $K$ to denote the number of local steps per round, and $T = KR$ to denote the parallel runtime. 
% We use the subscript to denote timestep, italicized superscript to denote the index of client and 
% We use unitalicized superscript ``md'' or ``ag'' to denote modifier of iterates in \fedac (see definition in \cref{alg:fedac}).
% \hynote{FIX}
% !TEX root = main.tex
\section{Algorithm: Federated Accelerated Stochastic Gradient Descent (\fedac)} 
We formally introduce our algorithm \fedac in \cref{alg:fedac}. \fedac parallelizes a generalized version of Accelerated SGD by \cite{Ghadimi.Lan-SIOPT12}. 
As in \fedavg, \fedac proceeds in $R$ communication rounds.
At the beginning of each round $r$, a central orchestration server sends the current state $\x^{(r,0)}$, and the current ``aggregated'' state $\x^{(r,0)}_{\mathrm{ag}}$ to each of the $M$ clients. 
Each client then locally takes $K$ steps of (generalized) accelerated SGD, as described in Line 7 -- 10. 
Here $\x^{(r,k)}_{\mathrm{ag}, m}$ aggregates the past iterates, $\x^{(r,k)}_{\mathrm{md},m}$ is the auxiliary sequence of ``middle points'' on which the gradients are queried, and $\x^{(r,k)}_m$ is the main sequence of iterates.
After $K$ local steps, the central orchestration server collects and averages \emph{both} the main states $\x^{(r,K)}_{m}$ and the aggregated states $\x^{(r,K)}_{\mathrm{ag}, m}$ to obtain the first iterate pair of the next round, namely $\x^{(r+1,0)}, \x^{(r+1,0)}_{\mathrm{ag}}$.  
% In \fedac, each client $m \in [M]$ maintains three intertwined sequences $\{\x^{(r,k)}_m, \x^{(r,k)}_{\mathrm{ag}, m}, \x^{(r,k)}_{\mathrm{md},m}\}$, where the superscript $(r,k)$ stands for the $k$-th local step of the $r$-th round, and the subscript $m$ stands for the $m$-th client.
% At each local step $k < K$, the next itearate $\x^{(r,k+1)}_{\mathrm{ag}, m}$ and $\x^{(r,k+1)}_{m}$. 
% After $K$ local steps, the central orcheastration server will collect, average, and broadcast the average states.

\begin{algorithm}
    \caption{\fedacfull (\fedac)}
    \label{alg:fedac}
    \begin{algorithmic}[1]
    \STATE{\textbf{procedure}} \fedac ($\x^{(0,0)}; \alpha, \beta, \eta, \gamma$) 
      \COMMENT{See \cref{fedaci,fedacii} for hyperparameter choices}
    \STATE Initialize $\x^{(r,0)}_{\mathrm{ag}} \gets \x^{(0,0)}$
    \FOR{$r = 0, \ldots, R-1$}
      \FORALL {$m \in [M]$ {\bf in parallel}}
        \STATE $\x^{(r,0)}_m \gets \x^{(r,0)}$;~~ $\x^{(r,0)}_{\mathrm{ag}, m} \gets \x^{(r,0)}_{\mathrm{ag}} $ \COMMENT{broadcast current iterate}
        \FOR {$k = 0, \ldots, K-1$}
          \STATE $\x^{(r,k)}_{\mathrm{md}, m} \gets \beta^{-1} \x^{(r,k)}_m + (1 - \beta^{-1}) \x^{(r,k)}_{\mathrm{ag}, m}$
          \COMMENT{Compute $\x^{(r,k)}_{\mathrm{md}, m}$ by coupling}
          \STATE $\g^{(r,k)}_m \gets \nabla f(\x^{(r,k)}_{\mathrm{md}, m}; \xi^{(r,k)}_m)$ 
          \COMMENT{Query gradient at $\x^{(r,k)}_{\mathrm{md}, m}$}
          \STATE $\x^{(r, k+1)}_{\mathrm{ag}, m} \gets \x^{(r,k)}_{\mathrm{md}, m} - \eta \cdot \g^{(r,k)}_m$
          \STATE $\x^{(r, k+1)}_m \gets (1 - \alpha^{-1}) \x^{(r,k)}_m + \alpha^{-1} \x^{(r,k)}_{\mathrm{md}, m} - \gamma \cdot \g^{(r,k)}_m$
        \ENDFOR
      \ENDFOR
      $\x^{(r+1, 0)} \gets \frac{1}{M} \sum_{m=1}^M \x^{(r,K)}_m$;~~$\x^{(r+1, 0)}_{\mathrm{ag}} \gets \frac{1}{M} \sum_{m=1}^M \x^{(r,K)}_{\mathrm{ag}, m}$
      \COMMENT{server averaging}
    \ENDFOR
    \end{algorithmic}
\end{algorithm}
\paragraph{Hyperparameter choice.}
We note that the particular version of Accelerated SGD in \fedac is more flexible than the most standard Nesterov's version \cite{Nesterov-18}, as it has four hyperparameters instead of two. 
Our analysis suggests that this flexibility seems crucial for principled acceleration in the distributed setting to allow for acceleration-stability trade-off. 

However, we note that our theoretical analysis gives a very concrete choice of hyperparameter $\alpha, \beta$, and $\gamma$ in terms of $\eta$.
For $\mu$-strongly-convex objectives, we introduce the following two sets of hyperparameter choices, which are referred to as \fedaci and \fedacii, respectively.
As we will see in the \cref{sec:thm:a1}, under \cref{asm:fo:scvx:2o},
\fedaci has a better dependency on condition number $\nicefrac{L}{\mu}$, whereas \fedacii has better communication efficiency.
\begin{alignat}{5}
     & \text{\fedaci}: \quad
     &                        & \eta \in \left(0, \frac{1}{L}\right], \quad &  & \gamma = \max \left\{ \sqrt{\frac{\eta}{\mu K}} , \eta\right\}, \quad &  & \alpha  = \frac{1}{\gamma \mu},  \quad                &  & \beta = \alpha + 1;
    \label{fedaci}
    \\
     & \text{\fedacii}: \quad
     &                        & \eta \in \left(0, \frac{1}{L}\right], \quad &  & \gamma = \max \left\{ \sqrt{\frac{\eta}{\mu K}}, \eta \right\}, \quad &  & \alpha  = \frac{3}{2 \gamma \mu} - \frac{1}{2}, \quad &  & \beta = \frac{2 \alpha^2 - 1}{\alpha - 1}.
    \label{fedacii}
\end{alignat}

Therefore, practically, if the strong convexity estimate $\mu$ is given (which is often taken to be the $\ell_2$ regularization strength), the only hyperparameter to be tuned is $\eta$, whose optimal value depends on the problem parameters.

\section{Theoretical Results and Discussions}
% Now we present main theorems of \fedac for strongly convex objectives under \cref{asm:fo:scvx:2o} or \ref{asm:fo:scvx:3o}.
\subsection{Convergence of \fedac under \cref{asm:fo:scvx:2o}}
\label{sec:thm:a1}

We first introduce the convergence theorem on \fedac under \cref{asm:fo:scvx:2o}. \fedaci and \fedacii lead to slightly different convergence rates. 
\begin{theorem}[Convergence of \fedac]
    \label{fedac:a1}
    Let $F$ be $\mu > 0$-strongly convex, and assume \cref{asm:fo:scvx:2o}.
    \begin{enumerate}
        \item [(a)] (Full version see \cref{fedaci:full}) For $\eta = \min \{ \frac{1}{L}, \tilde{\Theta} (  \frac{1}{\mu K R^2}) \}$, \fedaci yields
              \begin{equation}
                  \expt \left[ F( \x^{(R,0)}_{\mathrm{ag}})  \right]  - F^{\star}  \leq
                  \exp \left( \min \left\{ -\frac{\mu K R}{L}, - \sqrt{\frac{\mu K R^2}{L}} \right\} \right)
                  L B^2 +  \tildeo\left(\frac{\sigma^2}{\mu M K R} + \frac{L \sigma^2} {\mu^2 K R^3}\right).
                  \label{eq:fedac:a1:1}
              \end{equation}
        \item [(b)] (Full version see \cref{fedacii:a1:full}) For $\eta = \min \{ \frac{1}{L}, \tilde{\Theta} (  \frac{1}{\mu K R^2} ) \}$, \fedacii yields
              \begin{equation}
                  \expt \left[  F( \x^{(R,0)}_{\mathrm{ag}})  \right]  - F^{\star} \leq
                  \exp \left( \min \left\{ -\frac{\mu K R}{3L}, - \sqrt{\frac{\mu K R^2}{9L}}  \right\} \right)
                  L B^2 +
                  \tildeo \left( \frac{\sigma^2}{\mu M K R} +  \frac{L^2 \sigma^2}{\mu^3 K R^4} \right).
                  \label{eq:fedac:a1:2}
              \end{equation}
    \end{enumerate}
\end{theorem}

In comparison, the state-of-the-art \fedavg analysis \cite{Khaled.Mishchenko.ea-AISTATS20,Woodworth.Patel.ea-ICML20} reveals the following result.\footnote{\cref{fedavg:a1} can be (easily) adapted from the Theorem 2 of \cite{Woodworth.Patel.ea-ICML20} which analyzes a decaying learning rate with convergence rate $\mathcal{O}\left( \frac{L^2B^2}{\mu T^2}  + \frac{\sigma^2}{\mu M K R} \right) +  \tildeo \left( \frac{L \sigma^2}{\mu^2 K R^2} \right)$. This bound has no $\log$ factor attached to $\frac{\sigma^2}{\mu M K R}$ term but worse (polynomial) dependency on initial state $B$ than \cref{fedavg:a1}. We present \cref{fedavg:a1} for consistency and the ease of comparison.}
\begin{proposition}[Convergence of \fedavg under \cref{asm:fo:scvx:2o}, adapted from \cite{Woodworth.Patel.ea-ICML20}]
    \label{fedavg:a1}
    In the settings of \cref{fedac:a1}, for $\eta = \min \{ \frac{1}{L}, \tilde{\Theta} ( \frac{1}{\mu K R}  ) \}$, for appropriate non-negative $\{\rho^{(r,k)}\}$ with $\sum_{r=0}^{R-1} \sum_{k=0}^{K-1} \rho^{(r,k)} = 1$, $\fedavg$ yields
        \begin{equation}
            \expt \left[ F \left( \sum_{r=0}^{R-1} \sum_{k=0}^{K-1} \rho^{(r,k)} \overline{\x^{(r,k)}} \right) \right] - F^{\star} 
            \leq \exp \left( - \frac{\mu K R}{L} \right) L B^2 +
            \tildeo \left(\frac{\sigma^2}{\mu M K R} + \frac{L \sigma^2}{\mu^2 K R^2} \right).
            \label{eq:fedavg:main}
        \end{equation}
\end{proposition}
The bound for \fedaci \eqref{eq:fedac:a1:1} \textbf{asymptotically universally outperforms} \fedavg \eqref{eq:fedavg:main}.
The first term in \eqref{eq:fedac:a1:1} corresponds to the deterministic convergence, which is better than the one for \textsc{FedAvg}. 
The second term corresponds to the stochasticity of the problem which is not improvable. 
The third term corresponds to the overhead of infrequent communication, which is also better than \textsc{FedAvg} due to acceleration.
On the other hand, \fedacii has better communication efficiency since the third term of \eqref{eq:fedac:a1:2} decays at rate $R^{-4}$.

\subsection{Convergence of \fedac under \cref{asm:fo:scvx:3o} --- Faster when Close to be Quadratic}
Similar to the situation of \cref{sec:fedavg:3o:ub}, we can establish stronger guarantees for \fedacii \eqref{fedacii} under \cref{asm:fo:scvx:3o}.
\begin{theorem}[Simplified version of \cref{fedacii:a2:full}]
    \label{fedacii:a2}
    Let $F$ be $\mu > 0$-strongly convex, and assume \cref{asm:fo:scvx:3o}, then 
    for $R \geq \sqrt{\frac{L}{\mu}}$,\footnote{The assumption $R \geq \sqrt{{L}/{\mu}}$ is removed in the full version (\cref{fedacii:a2:full}).} 
    for $\eta = \min \{ \frac{1}{L}, \tilde{\Theta} (  \frac{1}{\mu K R^2} ) \}$, \fedacii yields
        \begin{equation}
            \expt \left[  F( \x^{(R,0)}_{\mathrm{ag}}) \right] - F^{\star}  
            \leq  \exp \left( \min \left\{ -\frac{\mu K R}{3L}, - \sqrt{\frac{\mu K R^2}{9L}} \right\} \right)
            2L B^2 + \tildeo\left(\frac{\sigma^2}{\mu M K R} 
            + \frac{Q^2 \sigma^4}{\mu^5 K^2 R^8}\right).
            \label{eq:fedacii:a2}
        \end{equation}
\end{theorem}
In comparison, we also establish and prove the convergence rate of \fedavg under \cref{asm:fo:scvx:3o}. 
\begin{theorem}[Simplified version of \cref{fedavg:a2:full}]
    \label{fedavg:a2}
    In the settings of \cref{fedacii:a2}, for $\eta = \min \left\{ \frac{1}{4L}, \tilde{\Theta}\left( \frac{1}{\mu K R} \right)\right\}$, for appropriate non-negative $\{\rho^{(r,k)}\}$ with $\sum_{r=0}^{R-1} \sum_{k=0}^{K-1} \rho^{(r,k)} = 1$, \fedavg yields
    \begin{equation}
        \expt \left[ F \left( \sum_{r=0}^{R-1} \sum_{k=0}^{K-1} \rho^{(r,k)} \overline{\x^{(r,k)}} \right) \right] - F^{\star}  \leq \exp \left( - \frac{\mu K R}{8L} \right) 4L B^2 + \tildeo \left( \frac{\sigma^2}{\mu M K R} + \frac{Q^2 \sigma^4}{\mu^5 K^2 R^4} \right).
        \label{eq:fedavg:a2} 
    \end{equation}
\end{theorem}
Our results give a smooth interpolation of the results of \cite{Woodworth.Patel.ea-ICML20} for quadratic objectives to broader function class --- the third term regarding infrequent communication overhead will vanish when the objective is quadratic since $Q = 0$.
The bound of \fedac \eqref{eq:fedacii:a2} outperforms the bound of \fedavg \eqref{eq:fedavg:a2} as long as $R \geq \sqrt{L/\mu}$ holds. 
% Particularly in the case of $T \geq M$, our analysis suggests that only $\tildeo(1)$ synchronization are required for linear speedup in $M$.
We summarize our results in \cref{tab:conv:rate}.

\subsection{Convergence for General Convex Objectives}
We also study the convergence of \fedac for general convex objectives ($\mu = 0$). 
The idea is to apply \fedac to $\ell_2$-augmented objective $\tilde{F}_{\lambda}(\x) := F(\x) + \frac{\lambda}{2}\|\x - \x^{(0,0)}\|_2^2 $ as a $\lambda$-strongly-convex and $(L + \lambda)$-smooth objective for appropriate $\lambda$, which is similar to the technique of \cite{Woodworth.Patel.ea-ICML20}. 
This augmented technique allows us to reuse most of the analysis for strongly-convex objectives. 
We conjecture that it is possible to construct direct versions of \fedac for general convex objectives that attain the same rates, which we defer for the future work.
% We summarize the synchronization bounds in \cref{tab:sync:bound} and the convergence rates in \cref{tab:conv:rate}. 
We defer the statement of formal theorems to \cref{sec:gcvx}.

\section{Proof of \cref{fedac:a1}(a)}
\label{sec:proof:sketch:1}
In this section we will prove \fedaci. 
We start by providing a complete, non-asymptotic version of \cref{fedac:a1}(a) on the convergence of \fedaci under \cref{asm:fo:scvx:2o}, and then provide the detailed proof.
%, which expands the proof sketch in \cref{sec:proof:sketch:1}. \hynote{fix}
Recall that \fedaci is defined as the \fedac (\cref{alg:fedac}) with the following hyperparameters choice
\begin{equation}
  \eta \in \left( 0, \frac{1}{L} \right], \quad \gamma = \max \left\{ \sqrt{\frac{\eta}{\mu K}} , \eta\right\}, \quad   \alpha  = \frac{1}{\gamma \mu},  \quad \beta = \alpha + 1
  \tag{\fedaci}.
\end{equation}
Recall $\overline{\x^{(r,k)}}$ is defined as $\frac{1}{M} \sum_{m=1}^M \x_m^{(r,k)}$.
Formally, we use $\mathcal{F}^{(r,k)}$ to denote the $\sigma$-algebra generated by $\{\x^{(\rho,\kappa)}_m, \x^{(\rho,\kappa)}_{\mathrm{ag},m}\}$ for $\rho < r$ or $\rho = r$ but $\kappa \leq k$. Since \fedac is Markovian, conditioning on $\mathcal{F}^{(r,k)}$ is equivalent to conditioning on $\{\x_m^{(r,k)}, \x_{\mathrm{ag}, m}^{(r,k)}\}_{m \in [M]}$.

We keep track of the convergence progress via the \emph{decentralized} potential 
\begin{equation}
  \Psi^{(r,k)} := \frac{1}{M} \sum_{m=1}^M F( {\x_{\mathrm{ag},m}^{(r,k)}})  - F^{\star} + \frac{1}{2}\mu \|\overline{\x^{(r,k)}} - \x^{\star}\|_2^2 .
  \label{eq:fedaci:potential}
\end{equation}
$\Psi^{(r,k)}$ is adapted from the common potential for acceleration analysis \cite{Bansal.Gupta-19}.

Now we introduce the main theorem on the convergence of \fedaci. Throughout this paper we do not optimize the $\polylog$ factors or the constants. We conjecture that certain $\polylog$ factors can be improved or removed via averaging techniques such as \cite{Lacoste-Julien.Schmidt.ea-arXiv12,Stich-arXiv19}.
% In fact, our bound \cref{fedaci:full} in terms of $K$ also holds for irregular synchronization setting as long as the maximum local steps is bounded by $K$.
% \footnote{Note that we state our full \cref{fedaci:full} in terms of the synchronization gap $K$ instead of the synchronization round $R$ as in the simplified \cref{fedac:a1}(a). This two quantities are trivially related as $T = KR$.}
\begin{theorem}[Convergence of \fedaci, complete version of \cref{fedac:a1}(a)]
  \label{fedaci:full}
  Let $F$ be $\mu>0$-strongly convex, and assume \cref{asm:fo:scvx:2o}, then for
  \begin{equation*}
    \eta = \min \left\{ \frac{1}{L},  \frac{1}{\mu K R^2} \log^2 \left( \euler + \min \left\{ \frac{\mu M K R \Psi^{(0,0)}}{\sigma^2}, \frac{\mu^2 KR^3 \Psi^{(0,0)}}{L \sigma^2} \right\} \right) \right\},
  \end{equation*}
  \fedaci yields
  \begin{align*}
    \expt [\Psi^{(R,0)}]
    \leq &
    \min \left\{ \exp \left( - \frac{\mu K R}{L} \right), \exp \left( - \frac{\mu^{\frac{1}{2}} K^{\frac{1}{2}} R}{L^{\frac{1}{2}}} \right)\right\}\Psi^{(0,0)}
    \\
    & + \frac{2\sigma^2}{\mu M K R} \log^2 \left(  \euler + \frac{\mu M K R \Psi^{(0,0)}}{\sigma^2} \right) 
    + \frac{400 L \sigma^2}{\mu^2 K R^3} \log^4 \left( \euler + \frac{\mu^2 K R^3 \Psi^{(0,0)}}{L \sigma^2} \right),
   \end{align*}
  where $\Psi$ is the decentralized potential defined in \cref{eq:fedaci:potential}.
\end{theorem}
\begin{remark}
  The simplified version in \cref{fedac:a1}(a) can be obtained by bounding the potential $\Psi^{(0,0)}$ with $LB^2$.
\end{remark}

\subsection{Proof Overview}
Our proof framework consists of the following four steps.
\paragraph{Step 1: potential-based perturbed iterate analysis.} 
The first step is to study the difference between \fedac and its fully synchronized idealization. 
To this end, we extend the perturbed iterate analysis \cite{Mania.Pan.ea-SIOPT17} to potential-based setting to analyze accelerated convergence.

To explicate the hyperparameter dependency, we state these lemmas for general $\gamma \in \left[\eta, \sqrt{ \frac{\eta}{\mu}} \right]$, which has one more degree of freedom than \fedaci where $\gamma = \max \left\{ \sqrt{\frac{\eta}{\mu K}}, \eta\right\}$ is fixed.
\begin{lemma}[label=fedaci:conv:main,restate=FedAcIConvMain,name={Potential-based perturbed iterate analysis for \textsc{FedAcI}}]
  Let $F$ be $\mu>0$-strongly convex, and assume \cref{asm:fo:scvx:2o}, then for $\alpha = \frac{1}{\gamma \mu}$, $\beta = \alpha + 1$, $\gamma \in \left[\eta, \sqrt{ \frac{\eta}{\mu}} \right]$, $\eta \in \left(0, \frac{1}{L}\right]$, \fedac yields
  \begin{align}
     & \expt \left[\Psi^{(R,0)} \right]          \leq \exp \left( - \gamma \mu KR \right)  \Psi^{(0,0)} + \frac{\eta^2 L \sigma^2}{2\gamma \mu} + \frac{\gamma\sigma^2}{2M} \tag{\text{Convergence when fully synchronized}}
    \\
     & + \underbrace{L \cdot \max_{\substack{0 \leq r < R \\ 0 \leq k < K}}  \expt
    \left[\frac{1}{M} \sum_{m=1}^M \left\| \overline{\x^{(r,k)}_{\mathrm{md}}} - \x^{(r,k)}_{\mathrm{md},m}  \right\|_2
    \left\|  \frac{1}{1+\gamma\mu}(\overline{\x^{(r,k)}} - \x_m^{(r,k)}) + \frac{\gamma \mu}{1+\gamma\mu} (\overline{\x^{(r,k)}_{\mathrm{ag}}} - \x_{\mathrm{ag},m}^{(r,k)})  \right\|_2  \right]}_{\text{Discrepancy overhead}},
    \label{eq:fedaci:conv:main}
  \end{align}
  where $\Psi$ is the decentralized potential defined in \cref{eq:fedaci:potential}.
\end{lemma}
We refer to the last term of \eqref{eq:fedaci:conv:main} as ``discrepancy overhead'' since it characterizes the dissimilarities among clients due to infrequent synchronization. The proof of \cref{fedaci:conv:main} is deferred to \cref{sec:fedaci:conv:main}.

\paragraph{Step 2: bounding discrepancy overhead.}
The second step is to bound the discrepancy overhead in \eqref{eq:fedaci:conv:main} via stability analysis.
Before we look into \fedac, let us first review the intuition for \fedavg. 
There are two forces governing the growth of discrepancy of \fedavg, namely the (negative) gradient and stochasticity. 
Thanks to the convexity, the gradient only makes the discrepancy lower.
The stochasticity incurs $\mathcal{O}(\eta^2 \sigma^2)$ variance per step, so the discrepancy 
$\expt [\frac{1}{M} \sum_{m=1}^M \|\overline{\x^{(r,k)}} - \x^{(r,k)}_m\|_2^2  ]$ 
grows at rate $\mathcal{O}(\eta^2 K \sigma^2)$ linear in $K$. 
The detailed proof can be found in \cite{Khaled.Mishchenko.ea-AISTATS20,Woodworth.Patel.ea-ICML20}.

For \fedac, the discrepancy analysis is subtler since acceleration and stability are at odds --- the momentum may amplify the discrepancy accumulated from previous steps.
Indeed, we establish the following \cref{thm:agd:instability}, which shows that the \emph{standard deterministic} Accelerated GD (\agd) may \emph{not} be initial-value stable even for strongly convex and smooth objectives, in the sense that initial infinitesimal difference may grow exponentially fast. 
We defer the formal setup and the proof of \cref{thm:agd:instability} to the next section (\cref{sec:instability}).
\begin{theorem}[name={Initial-value instability of deterministic standard \agd},label=thm:agd:instability,restate=instability]
  For any $L, \mu > 0$ such that $\nicefrac{L}{\mu} \geq 25$, and for any $K \geq 1$, there exists a 1D objective $F$ that is $L$-smooth and $\mu$-strongly-convex, and an $\varepsilon_0 > 0$, such that for any positive $\varepsilon < \varepsilon_0$, there exists $w^{(0)}, u^{(0)}, w^{(0)}_{\mathrm{ag}}, u^{(0)}_{\mathrm{ag}}$ such that $|w^{(0)} - u^{(0)}| \leq \varepsilon$, $|w^{(0)}_{\mathrm{ag}} - u^{(0)}_{\mathrm{ag}}| \leq \varepsilon$, but
  the sequence $\{w^{(t)}_{\mathrm{ag}}, w^{(t)}_{\mathrm{md}}, w^{(t)}\}_{t=0}^{3K}$ output by $\agd(w^{(0)}_{\mathrm{ag}}, w^{(0)}, L, \mu)$ and sequence
  $\{u^{(t)}_{\mathrm{ag}}, u^{(t)}_{\mathrm{md}}, u^{(t)}\}_{t=0}^{3K}$ output by $\agd(u^{(0)}_{\mathrm{ag}}, u^{(0)}, L, \mu)$ satisfies
  \begin{equation*}
      |w^{(3K)} - u^{(3K)}| \geq \frac{1}{2} \varepsilon (1.02)^K, 
      \qquad
      |w^{(3K)}_{\mathrm{ag}} - u^{(3K)}_{\mathrm{ag}}| \geq \varepsilon (1.02)^K.
  \end{equation*}
\end{theorem} 
\begin{remark}
  It is worth mentioning that the instability theorem \textbf{does not contradicts the convergence} of \agd \cite{Nesterov-18}. The convergence of \agd suggests that $w^{(t)}_{\mathrm{ag}}$, $w^{(t)}$, $u^{(t)}_{\mathrm{ag}}$, and $u^{(t)}$ will all converge to the same point $\x^{\star}$ as $t \to \infty$, which implies $\lim_{t \to \infty} \|w^{(t)}_{\mathrm{ag}} - u^{(t)}_{\mathrm{ag}}\| = \|w^{(t)} - u^{(t)}\| = 0$. However, the convergence theorem does not imply the stability with respect to the initialization --- it does not exclude the possibility that the difference between two instances (possibly with very close initialization) first expand and only shrink until they both approach $\x^{\star}$. Our \cref{thm:agd:instability} suggests this possibility: for any finite steps, no matter how small the (positive) initial difference is, it is possible that the difference will grow exponentially fast. 
This is fundamentally different from the Gradient Descent (for convex objectives), for which the difference between two instances does not expand for standard choice of learning rate $\eta = \frac{1}{L}$ (where $L$ is the smoothness). 
\end{remark}

Fortunately, we can show that the discrepancy can grow at a slower exponential rate via less aggressive acceleration, see \cref{fedaci:stab:main}. As we will discuss shortly, we adjust $\gamma$ according to $K$ to restrain the growth of discrepancy within the linear regime. The proof of \cref{fedaci:stab:main} is deferred to \cref{sec:fedaci:stab:main}.
\begin{lemma}[name={Discrepancy overhead bound},label=fedaci:stab:main,restate=FedAcIStabMain]
  In the same setting of \cref{fedaci:conv:main}, the following inequality holds
  \begin{align}
         & \expt
    \left[ \frac{1}{M} \sum_{m=1}^M
    \left\| \overline{\x^{(r,k)}_{\mathrm{md}}} - \x^{(r,k)}_{\mathrm{md},m} \right\|_2
    \left\|  \frac{1}{1+\gamma\mu}(\overline{\x^{(r,k)}} - \x_m^{(r,k)}) + \frac{\gamma \mu}{1+\gamma\mu} (\overline{\x^{(r,k)}_{\mathrm{ag}}} - \x_{\mathrm{ag},m}^{(r,k)})  \right\|_2  \right]
    \nonumber \\
    \leq &
    \begin{cases}
      7 \eta \gamma K \sigma^2 \left(1 + \frac{2\gamma^2\mu}{\eta}\right)^{2K}
       & \text{if~} \gamma \in \left(\eta, \sqrt{\frac{\eta}{\mu}} \right],
      \\
      7 \eta^2 K \sigma^2
       &
      \text{if~} \gamma = \eta.
    \end{cases}
    \label{eq:fedaci:stab:main}
  \end{align}
\end{lemma}
The proof of \cref{fedaci:stab:main} is deferred to \cref{sec:fedaci:stab:main}.

\paragraph{Step 3: trading-off acceleration and discrepancy.} 
Combining \cref{fedaci:conv:main,fedaci:stab:main} gives 
  \begin{equation}
    \label{fedaci:combine:sketch}
    \expt \left[\Psi^{(R,0)} \right]
    \leq \underbrace{\exp \left( - \gamma \mu K R \right)  \Psi^{(0,0)}}_{\mathrm{(I)}}
    + \frac{\eta^2 L \sigma^2}{2\gamma \mu}
    + \frac{\gamma\sigma^2}{2M}
    +
    \underbrace{
      \begin{cases}
        7 \eta \gamma L K \sigma^2 \left(1 + \frac{2\gamma^2\mu}{\eta}\right)^{2K}
         & \text{if~} \gamma \in (\eta, \sqrt{ \frac{\eta}{\mu}} ],
        \\
        7 \eta^2 L K \sigma^2
         &
        \text{if~} \gamma = \eta.
      \end{cases}
    }_{\mathrm{(II)}}
  \end{equation}
The value of $\gamma \in [\eta, \sqrt{\eta/\mu}]$ controls the magnitude of acceleration in (I) and discrepancy growth in (II).
The upper bound choice $\sqrt{\eta/\mu}$ gives full acceleration in (I) but makes (II) grow exponentially in $K$.
On the other hand, the lower bound choice $\eta$ makes (II) linear in $K$ but loses all acceleration.
We wish to attain as much acceleration in (I) as possible while keeping the discrepancy (II) grow moderately. \textbf{Our balanced solution} is to pick $\gamma = \max \{ \sqrt{{\eta}/{(\mu K)}}, \eta\}$. One can verify that the discrepancy grows (at most) linearly in $K$.
Substituting this choice of $\gamma$ to \cref{fedaci:combine:sketch} leads to the following lemma.
\begin{lemma}[name={Convergence of \fedaci for general $\eta$},label=fedaci:general:eta]
    Let $F$ be $\mu > 0$-strongly convex, and assume \cref{asm:fo:scvx:2o}, then for any $\eta \in \left(0, \frac{1}{L}\right]$, \fedaci yields
    \begin{align}
      \expt[\Psi^{(R,0)}]
      \leq & 
      \underbrace{\exp \left(  - \max \left\{ \eta \mu, \sqrt{\frac{\eta \mu}{K}}\right\}KR \right) \Psi^{(0,0)}}_{\text{Monotonically decreasing } \varphi_{\downarrow}(\eta)}
      \nonumber \\
      & \quad + 
      \underbrace{\frac{\eta^{\frac{1}{2}} \sigma^2}{2 \mu^{\frac{1}{2}} M K^{\frac{1}{2}} }
      + \frac{\eta \sigma^2}{2M}
      + \frac{390 \eta^{\frac{3}{2}} L K^{\frac{1}{2}} \sigma^2} {\mu^{\frac{1}{2}}}
      + 7 \eta^2 L K \sigma^2}_{\text{Monotonically increasing} \varphi_{\uparrow}(\eta)}.
      \label{eq:fedaci:general:eta}
    \end{align}
\end{lemma}
The proof of \cref{fedaci:general:eta} is deferred to \cref{sec:fedaci:step3}.

\paragraph{Step 4: finding $\eta$ to optimize the RHS of \cref{eq:fedaci:general:eta}.}
It remains to show that \eqref{eq:fedaci:general:eta} gives the desired bound with our choice of $\eta = \min \{ \frac{1}{L}, \tilde{\Theta} (\frac{1}{\mu K R^2} )\}$.
The increasing $\varphi_{\uparrow}(\eta)$ in \eqref{eq:fedaci:general:eta} is bounded by
\(
\tildeo ( \frac{\sigma^2}{\mu M K R} +  \frac{L \sigma^2}{\mu^2 K R^3} ).
\)
The decreasing term $\varphi_{\downarrow}(\eta)$ in \eqref{eq:fedaci:general:eta} is bounded by $\varphi_{\downarrow}({\frac{1}{L}}) + \varphi_{\downarrow} (\tilde{\Theta} (  \frac{1}{\mu K R^2} ) )$, where $    \varphi_{\downarrow} ({\frac{1}{L}} )
  = \exp ( \min \{ - \frac{\mu K R}{L}, - \frac{\mu^{\frac{1}{2}} K^{\frac{1}{2}} R}{L^{\frac{1}{2}}}\} )$,
and
$\varphi_{\downarrow} (\tilde{\Theta} (  \frac{1}{\mu K R^2} ) ) 
\leq
\exp \left( - \mu^{\frac{1}{2}} K^{\frac{1}{2}} R \cdot \sqrt{\tilde{\Theta} \left(  \frac{1}{\mu K R^2} \right)} \right)$ 
can be controlled by the bound of $\varphi_{\uparrow}(\eta)$ provided $\tilde{\Theta}$ has appropriate $\polylog$ factors.
Plugging the bounds to \eqref{eq:fedaci:general:eta} completes the proof of \cref{fedac:a1}(a).
We defer the details to \cref{sec:fedaci:step4}.

\subsection{Details of Step 1: Proof of \cref{fedaci:conv:main}}
\label{sec:fedaci:conv:main}
In this section we will prove \cref{fedaci:conv:main}.
We start by the one-step analysis of the decentralized potential $\Psi^{(r,k)}$ defined in \cref{eq:fedaci:potential}. The following two propositions establish the one-step analysis of the two quantities in $\Psi^{(r,k)}$, namely $\|\overline{\x^{(r,k)}} - \x^{\star}\|_2^2 $ and $\frac{1}{M} \sum_{m=1}^M F(\x_{\mathrm{ag},m}^{(r,k)}) - F^{\star}$. 
We only require minimal hyperparameter assumptions, namely $\alpha \geq 1, \beta \geq 1, \eta \leq \frac{1}{L}$, for these two propositions.
We will then show how the choice of $\alpha, \beta$ is determined towards the proof of \cref{fedaci:conv:main} in order to couple the two quantities into potential $\Psi^{(r,k)}$.
\begin{proposition}
  \label{fedaci:conv:1}
  Let $F$ be $\mu>0$-strongly convex, and assume \cref{asm:fo:scvx:2o}, then for \fedac with hyperparameters assumptions $\alpha \geq 1$, $\beta \geq 1$, $\eta \leq \frac{1}{L}$, the following inequality holds
  \begin{align*}
         & \expt [ \|\overline{\x^{(r,k+1)}} - \x^{\star}\|_2^2  |\mathcal{F}^{(r,k)}]
    \\
    \leq & (1 - \alpha^{-1}) \| \overline{\x^{(r,k)}}  - \x^{\star}\|_2^2   + \alpha^{-1} \| \overline{\x^{(r,k)}_{\mathrm{md}}} - \x^{\star}\|_2^2  + \gamma^2 \left\|  \frac{1}{M} \sum_{m=1}^M \nabla F (\x_{\mathrm{md},m}^{(r,k)}) \right\|_2^2+ \frac{1}{M}\gamma^2 \sigma^2
    \\
         & - 2 \gamma \cdot \frac{1}{M} \sum_{m=1}^M \left\langle \nabla F (\x^{(r,k)}_{\mathrm{md},m}), (1 - \alpha^{-1}(1 - \beta^{-1})) {\x_m^{(r,k)}} + \alpha^{-1} (1 - \beta^{-1}) {\x_{\mathrm{ag},m}^{(r,k)}} - \x^{\star} \right\rangle
    \\
         & + 2 \gamma L \frac{1}{M} \sum_{m=1}^M
         \left(
    \left\| \overline{\x^{(r,k)}_{\mathrm{md}}} - \x^{(r,k)}_{\mathrm{md},m} \right\|_2 \cdot \right.
    \\
        & \qquad \qquad
    \left. \left\| (1-\alpha^{-1}(1-\beta^{-1})) (\overline{\x^{(r,k)}} - \x_m^{(r,k)}) + \alpha^{-1} (1 - \beta^ {-1}) (\overline{\x^{(r,k)}_{\mathrm{ag}}} - \x^{(r,k)}_{\mathrm{ag}, m})   \right\|_2 \right).
  \end{align*}
\end{proposition}
\begin{proposition}
  \label{fedaci:conv:2}
  In the same setting of \cref{fedaci:conv:1}, the following inequality holds
  \begin{align*}
         & \expt \left[ \frac{1}{M} \sum_{m=1}^M F( {\x^{(r,k+1)}_{\mathrm{ag}, m}}) - F^{\star} \middle| \mathcal{F}^{(r,k)} \right]
    \\
    \leq & (1 - \alpha^{-1}) \left( \frac{1}{M} \sum_{m=1}^M F( {\x_{\mathrm{ag},m}^{(r,k)}}) - F^{\star} \right)
    - \frac{1}{2} \eta \left\| \frac{1}{M} \sum_{m=1}^M  \nabla F (\x^{(r,k)}_{\mathrm{md},m})  \right\|_2^2 
    +  \frac{1}{2} \eta^2 L \sigma^2
    \\
         & + \alpha^{-1} \frac{1}{M} \sum_{m=1}^M \left\langle \nabla F(\x^{(r,k)}_{\mathrm{md},m}), \alpha \beta^{-1} \x_m^{(r,k)} + (1 - \alpha \beta^{-1}) \x_{\mathrm{ag},m}^{(r,k)} - \x^{\star}\right\rangle
    - \frac{1}{2} \mu \alpha^{-1} \| \overline{\x^{(r,k)}_{\mathrm{md}}} - \x^{\star}\|_2^2 .
  \end{align*}
\end{proposition}
We defer the proofs of \cref{fedaci:conv:1,fedaci:conv:2} to \cref{sec:proof:fedaci:conv:1,sec:proof:fedaci:conv:2}, respectively.

With \cref{fedaci:conv:1,fedaci:conv:2} at hand, we are ready to prove \cref{fedaci:conv:main}.
\begin{proof}[Proof of \cref{fedaci:conv:main}]
  Applying \cref{fedaci:conv:1} with the specified $\alpha = \frac{1}{\gamma \mu}, \beta = \alpha + 1$ yields (for any $r, k$)
  \begin{align}
         & \expt [ \|\overline{\x^{(r,k+1)}} - \x^{\star}\|_2^2  |\mathcal{F}^{(r,k)}]
    \nonumber \\
    \leq & (1 - \gamma \mu) \| \overline{\x^{(r,k)}}  - \x^{\star}\|_2^2   + \gamma\mu \| \overline{\x^{(r,k)}_{\mathrm{md}}} - \x^{\star}\|_2^2  + \gamma^2 \left\|  \frac{1}{M} \sum_{m=1}^M \nabla F (\x_{\mathrm{md},m}^{(r,k)}) \right\|_2^2+ \frac{1}{M}\gamma^2 \sigma^2
    \nonumber \\
         & - 2 \gamma \cdot \frac{1}{M} \sum_{m=1}^M \left\langle \nabla F (\x^{(r,k)}_{\mathrm{md},m}), \frac{1}{1 + \gamma\mu} {\x_m^{(r,k)}} + \frac{\gamma \mu}{1 + \gamma\mu} {\x_{\mathrm{ag},m}^{(r,k)}} - \x^{\star} \right\rangle
    \nonumber \\
         & + 2 \gamma L \cdot \frac{1}{M} \sum_{m=1}^M \left\| \overline{\x^{(r,k)}_{\mathrm{md}}} - \x^{(r,k)}_{\mathrm{md},m}  \right\|_2
    \left\|  \frac{1}{1+\gamma\mu}(\overline{\x^{(r,k)}} - \x_m^{(r,k)}) + \frac{\gamma \mu}{1+\gamma\mu} (\overline{\x^{(r,k)}_{\mathrm{ag}}} - \x_{\mathrm{ag},m}^{(r,k)})  \right\|_2.
    \label{eq:fedaci:conv:main:1}
  \end{align}
  Applying \cref{fedaci:conv:2} with the specified $\alpha = \frac{1}{\gamma \mu}, \beta = \alpha + 1$ yields (for any $r, k$)
  \begin{align}
         & \expt \left[ \frac{1}{M} \sum_{m=1}^M F( {\x^{(r,k+1)}_{\mathrm{ag}, m}}) - F^{\star} \middle| \mathcal{F}^{(r,k)} \right]
    \nonumber \\
    \leq & (1 - \gamma \mu) \left( \frac{1}{M} \sum_{m=1}^M F( {\x_{\mathrm{ag},m}^{(r,k)}}) - F^{\star} \right)
    - \frac{1}{2} \eta \left\| \frac{1}{M} \sum_{m=1}^M  \nabla F (\x^{(r,k)}_{\mathrm{md},m})  \right\|_2^2 
    +  \frac{1}{2} \eta^2 L \sigma^2
    \nonumber \\
         & + \gamma \mu \cdot \frac{1}{M} \sum_{m=1}^M \left\langle \nabla F(\x^{(r,k)}_{\mathrm{md},m}), \frac{1}{1+\gamma\mu} \x^{(r,k)}_{m} + \frac{\gamma \mu}{1 + \gamma \mu} \x^{(r,k)}_{\mathrm{ag}, m} - \x^{\star} \right\rangle
    - \frac{1}{2} \gamma \mu^2 \| \overline{\x^{(r,k)}_{\mathrm{md}}} - \x^{\star}\|_2^2 .
    \label{eq:fedaci:conv:main:2}
  \end{align}
  Adding \cref{eq:fedaci:conv:main:2} with $\frac{1}{2}\mu$ times of \cref{eq:fedaci:conv:main:1} yields
  \begin{align*}
     & \expt [\Psi^{(r,k+1)} |\mathcal{F}^{(r,k)}] \leq (1 - \gamma \mu) \Psi^{(r,k)} + \frac{1}{2} \left( \eta^2 L + \frac{1}{M} \gamma^2 \mu \right) \sigma^2 
     \\
     & \quad +  \frac{1}{2}\left( \gamma^2 \mu - \eta \right) \left\| \frac{1}{M} \sum_{m=1}^M  \nabla F (\x^{(r,k)}_{\mathrm{md},m})  \right\|_2^2 
    \\
     & \quad + \gamma \mu L \cdot \frac{1}{M} \sum_{m=1}^M \left\| \overline{\x^{(r,k)}_{\mathrm{md}}} - \x^{(r,k)}_{\mathrm{md},m}  \right\|_2
    \left\|  \frac{1}{1+\gamma\mu}(\overline{\x^{(r,k)}} - \x_m^{(r,k)}) + \frac{\gamma \mu}{1+\gamma\mu} (\overline{\x^{(r,k)}_{\mathrm{ag}}} - \x_{\mathrm{ag},m}^{(r,k)})  \right\|_2.
  \end{align*}
  Since $\gamma^2 \mu \leq \eta$, the coefficient of $\left\| \frac{1}{M} \sum_{m=1}^M  \nabla F (\x^{(r,k)}_{\mathrm{md},m})  \right\|_2^2 $ is non-positive. Thus
  \begin{align*}
     & \expt [\Psi^{(r,k+1)} |\mathcal{F}^{(r,k)}] \leq (1 - \gamma \mu) \Psi^{(r,k)} + \frac{1}{2} \left( \eta^2 L + \frac{1}{M} \gamma^2 \mu \right) \sigma^2
    \\
     & \quad + \gamma \mu L \cdot \frac{1}{M} \sum_{m=1}^M \left\| \overline{\x^{(r,k)}_{\mathrm{md}}} - \x^{(r,k)}_{\mathrm{md},m}  \right\|_2
    \left\|  \frac{1}{1+\gamma\mu}(\overline{\x^{(r,k)}} - \x_m^{(r,k)}) + \frac{\gamma \mu}{1+\gamma\mu} (\overline{\x^{(r,k)}_{\mathrm{ag}}} - \x_{\mathrm{ag},m}^{(r,k)})  \right\|_2.
  \end{align*}
  Telescoping the above inequality up to the $R$-th round yields
  \begin{align*}
         & \expt \left[\Psi^{(R,0)} \right] \leq  \left( 1 - \gamma \mu \right)^{KR} \Psi^{(0,0)} +
    \left( \sum_{t=0}^{KR-1} \left( 1 - \gamma \mu \right)^t \right) \cdot \frac{1}{2} \left( \eta^2 L + \frac{1}{M} \gamma^2 \mu \right) \sigma^2
    \\
         & +
    \gamma \mu L  \cdot \sum_{r=0}^{R-1}\sum_{k=0}^{K-1} \left\{   \left( 1 - \gamma\mu \right)^{T-(rK+k)-1} \cdot \right.
    \\
        & \qquad \left. \expt
    \left[ \frac{1}{M} \sum_{m=1}^M \left\| \overline{\x^{(r,k)}_{\mathrm{md}}} - \x^{(r,k)}_{\mathrm{md},m}  \right\|_2
    \left\|  \frac{1}{1+\gamma\mu}(\overline{\x^{(r,k)}} - \x_m^{(r,k)}) + \frac{\gamma \mu}{1+\gamma\mu} (\overline{\x^{(r,k)}_{\mathrm{ag}}} - \x_{\mathrm{ag},m}^{(r,k)})  \right\|_2  \right] \right\}
    \\
    \leq & \exp \left( - \gamma \mu KR \right) \Psi^{(0,0)} + \frac{\eta^2 L \sigma^2}{2\gamma \mu} + \frac{\gamma\sigma^2}{2M}
    \\
         & + L \cdot \max_{\substack{0 \leq r < R \\ 0 \leq k < K}}  \expt
    \left[\frac{1}{M} \sum_{m=1}^M \left\| \overline{\x^{(r,k)}_{\mathrm{md}}} - \x^{(r,k)}_{\mathrm{md},m}  \right\|_2 \cdot \right.
    \\
    & \left. \qquad \qquad \qquad \qquad
    \left\|  \frac{1}{1+\gamma\mu}(\overline{\x^{(r,k)}} - \x_m^{(r,k)}) + \frac{\gamma \mu}{1+\gamma\mu} (\overline{\x^{(r,k)}_{\mathrm{ag}}} - \x_{\mathrm{ag},m}^{(r,k)})  \right\|_2  \right],
  \end{align*}
  where in the last inequality we used the fact that $\sum_{t=0}^{\infty} \left( 1 - \gamma \mu \right)^t  \leq \frac{1}{\gamma \mu}$.
\end{proof}

\subsubsection{Proof of \cref{fedaci:conv:1}}
\label{sec:proof:fedaci:conv:1}
\begin{proof}[Proof of \cref{fedaci:conv:1}]
  By definition of the \fedac procedure (\cref{alg:fedac}), for any $m \in [M]$,
  \begin{equation*}
    \x^{(r,k+1)}_m = (1 - \alpha^{-1})\x_m^{(r,k)} + \alpha^{-1} \x_{\mathrm{md},m}^{(r,k)} - \gamma \cdot \nabla f(\x_{\mathrm{md},m}^{(r,k)};\xi^{(r,k)}_m).
  \end{equation*}
  Taking average over $m = 1, \ldots, M$ gives
  \begin{equation*}
    \overline{\x^{(r,k+1)}} - \x^{\star} = (1 - \alpha^{-1}) \overline{\x^{(r,k)}} + \alpha^{-1} \overline{\x^{(r,k)}_{\mathrm{md}}} - \gamma \cdot \frac{1}{M} \sum_{m=1}^M \nabla f(\x_{\mathrm{md},m}^{(r,k)}; \xi^{(r,k)}_m) - \x^{\star}.
  \end{equation*}
  Taking conditional expectation gives
  \begin{align}
    & \expt [ \|\overline{\x^{(r,k+1)}} - \x^{\star}\|_2^2  |\mathcal{F}^{(r,k)}]
    \nonumber \\
    =    & \left\| (1 - \alpha^{-1}) \overline{\x^{(r,k)}} + \alpha^{-1} \overline{\x^{(r,k)}_{\mathrm{md}}} - \gamma \cdot \frac{1}{M} \sum_{m=1}^M \nabla F (\x_{\mathrm{md},m}^{(r,k)})- \x^{\star} \right\|_2^2 
    \nonumber \\
         & \quad +  \expt \left[  \left\| \frac{1}{M} \sum_{m=1}^M  \left(  \nabla f(\x_{\mathrm{md},m}^{(r,k)}; \xi^{(r,k)}_m) - \nabla F(\x^{(r,k)}_{\mathrm{md}, m}) \right) \right\|_2^2 \middle| \mathcal{F}^{(r,k)} \right]
    \tag{independence}
    \\
    \leq & \left\| (1 - \alpha^{-1}) \overline{\x^{(r,k)}} + \alpha^{-1} \overline{\x^{(r,k)}_{\mathrm{md}}} - \gamma \cdot \frac{1}{M} \sum_{m=1}^M \nabla F (\x_{\mathrm{md},m}^{(r,k)})- \x^{\star} \right\|_2^2 
    + \frac{1}{M}\gamma^2 \sigma^2,
    \label{eq:fedaci:conv:1:0}
  \end{align}
  where the last inequality of \cref{eq:fedaci:conv:1:0} is due to the bounded variance assumption (\cref{asm:fo:scvx:2o}(c)) and independence.  Expanding the squared norm term of \cref{eq:fedaci:conv:1:0} and applying Jensen's inequality,
  \begin{align}
    & \left\| (1 - \alpha^{-1}) \overline{\x^{(r,k)}} + \alpha^{-1} \overline{\x^{(r,k)}_{\mathrm{md}}} - \gamma \cdot \frac{1}{M} \sum_{m=1}^M \nabla F (\x_{\mathrm{md},m}^{(r,k)})- \x^{\star} \right\|_2^2 
    \nonumber \\
    =    & \left\| (1 - \alpha^{-1}) \overline{\x^{(r,k)}} + \alpha^{-1} \overline{\x^{(r,k)}_{\mathrm{md}}} - \x^{\star} \right\|_2^2+ \gamma^2 \left\|  \frac{1}{M} \sum_{m=1}^M \nabla F (\x_{\mathrm{md},m}^{(r,k)}) \right\|_2^2 
    \nonumber \\
         & - 2 \gamma \cdot \frac{1}{M} \sum_{m=1}^M \left\langle \nabla F (\x_{\mathrm{md},m}^{(r,k)}), (1 - \alpha^{-1}) \overline{\x^{(r,k)}} + \alpha^{-1} \overline{\x^{(r,k)}_{\mathrm{md}}} - \x^{\star} \right\rangle
    \tag{expansion of squared norm}
    \\
    \leq & (1 - \alpha^{-1}) \| \overline{\x^{(r,k)}}  - \x^{\star}\|_2^2  + \alpha^{-1} \|
    \overline{\x^{(r,k)}_{\mathrm{md}}} - \x^{\star}\|_2^2   + \gamma^2 \left\|  \frac{1}{M} \sum_{m=1}^M \nabla F (\x_{\mathrm{md},m}^{(r,k)}) \right\|_2^2 
    \nonumber \\
         & - 2 \gamma \cdot \frac{1}{M} \sum_{m=1}^M  \left\langle\nabla F (\x_{\mathrm{md},m}^{(r,k)}), (1 - \alpha^{-1}) \overline{\x^{(r,k)}} + \alpha^{-1} \overline{\x^{(r,k)}_{\mathrm{md}}} - \x^{\star} \right\rangle.
    \label{eq:fedaci:conv:1:1}
  \end{align}
  It remains to analyze the inner product term of \cref{eq:fedaci:conv:1:1}. Note that
  \begin{small}
  \begin{align}
         & -\frac{1}{M} \sum_{m=1}^M \left\langle  \nabla F (\x^{(r,k)}_{\mathrm{md},m}), (1 - \alpha^{-1}) \overline{\x^{(r,k)}} + \alpha^{-1} \overline{\x^{(r,k)}_{\mathrm{md}}} - \x^{\star} \right\rangle
    \nonumber \\
    =    & -\frac{1}{M} \sum_{m=1}^M \left\langle \nabla F (\x^{(r,k)}_{\mathrm{md},m}) , (1-\alpha^{-1}(1-\beta^{-1})) \overline{\x^{(r,k)}} + \alpha^{-1} (1 - \beta^ {-1}) \overline{\x^{(r,k)}_{\mathrm{ag}}} - \x^{\star}\right\rangle
    \tag{definition of $\overline{\x^{(r,k)}_{\mathrm{md}}}$}
    \\
    =    & -\frac{1}{M} \sum_{m=1}^M \left\langle \nabla F (\x^{(r,k)}_{\mathrm{md},m}), (1-\alpha^{-1}(1-\beta^{-1})) (\overline{\x^{(r,k)}} - \x_m^{(r,k)}) + \alpha^{-1} (1 - \beta^ {-1}) (\overline{\x^{(r,k)}_{\mathrm{ag}}} - \x^{(r,k)}_{\mathrm{ag}, m})\right\rangle
    \nonumber \\
         & -\frac{1}{M} \sum_{m=1}^M \left\langle \nabla F (\x^{(r,k)}_{\mathrm{md},m}), (1 - \alpha^{-1}(1 - \beta^{-1})) {\x_m^{(r,k)}} + \alpha^{-1} (1 - \beta^{-1}) {\x_{\mathrm{ag},m}^{(r,k)}} - \x^{\star} \right\rangle
    \nonumber \\
    =    & \frac{1}{M} \sum_{m=1}^M \left\langle \nabla F(\overline{\x^{(r,k)}_{\mathrm{md}}}) - \nabla F (\x^{(r,k)}_{\mathrm{md},m}), (1-\alpha^{-1}(1-\beta^{-1})) (\overline{\x^{(r,k)}} - \x_m^{(r,k)}) + \alpha^{-1} (1 - \beta^ {-1}) (\overline{\x^{(r,k)}_{\mathrm{ag}}} - \x^{(r,k)}_{\mathrm{ag}, m})  \right\rangle
    \nonumber \\
         & - \frac{1}{M} \sum_{m=1}^M \left\langle \nabla F (\x^{(r,k)}_{\mathrm{md},m}), (1 - \alpha^{-1}(1 - \beta^{-1})) {\x_m^{(r,k)}} + \alpha^{-1} (1 - \beta^{-1}) {\x_{\mathrm{ag},m}^{(r,k)}} - \x^{\star} \right\rangle
    \nonumber \\
    \leq & L \cdot \frac{1}{M} \sum_{m=1}^M
    \left\| \overline{\x^{(r,k)}_{\mathrm{md}}} - \x^{(r,k)}_{\mathrm{md},m} \right\|_2
    \left\| (1-\alpha^{-1}(1-\beta^{-1})) (\overline{\x^{(r,k)}} - \x_m^{(r,k)}) + \alpha^{-1} (1 - \beta^ {-1}) (\overline{\x^{(r,k)}_{\mathrm{ag}}} - \x^{(r,k)}_{\mathrm{ag}, m})   \right\|
    \nonumber \\
         & - \frac{1}{M} \sum_{m=1}^M \left\langle \nabla F (\x^{(r,k)}_{\mathrm{md},m}), (1 - \alpha^{-1}(1 - \beta^{-1})) {\x_m^{(r,k)}} + \alpha^{-1} (1 - \beta^{-1}) {\x_{\mathrm{ag},m}^{(r,k)}} - \x^{\star} \right\rangle,
    \label{eq:fedaci:conv:1:2}
  \end{align}
  \end{small}
  where the last equality is due to the $L$-smoothness (\cref{asm:fo:scvx:2o}(b)).
  Combining \cref{eq:fedaci:conv:1:0,eq:fedaci:conv:1:1,eq:fedaci:conv:1:2} completes the proof of \cref{fedaci:conv:1}.
\end{proof}

\subsubsection{Proof of \cref{fedaci:conv:2}}
\label{sec:proof:fedaci:conv:2}
Before stating the proof of \cref{fedaci:conv:2}, we first introduce and prove the following claim for a single client $m \in [M]$.
\begin{claim}
  \label{fedaci:conv:2:claim}
  Under the same assumptions of \cref{fedaci:conv:2}, for any $m \in [M]$, the following inequality holds
  \begin{align*}
    & \expt \left[ F(\x^{(r,k+1)}_{\mathrm{ag}, m}) - F^{\star} |\mathcal{F}^{(r,k)} \right]
    \leq  ~
    (1 - \alpha^{-1}) \left( F( {\x_{\mathrm{ag},m}^{(r,k)}}) - F^{\star} \right)
    - \frac{1}{2} \eta \left\| \nabla F (\x^{(r,k)}_{\mathrm{md},m})  \right\|_2^2+  \frac{1}{2} \eta^2 L \sigma^2
    \\
    & 
    - \frac{1}{2} \mu \alpha^{-1} \|\x^{(r,k)}_{\mathrm{md},m} - \x^{\star}\|_2^2 
        + \alpha^{-1} \left\langle \nabla F(\x^{(r,k)}_{\mathrm{md},m}) ,  \alpha \beta^{-1} \x_m^{(r,k)} + (1 - \alpha \beta^{-1}) \x_{\mathrm{ag},m}^{(r,k)} - \x^{\star} \right\rangle.  
  \end{align*}
\end{claim}
\begin{proof}[Proof of \cref{fedaci:conv:2:claim}]
  \sloppy By definition of \fedac (\cref{alg:fedac}), $\x^{(r,k+1)}_{\mathrm{ag}, m} = {\x^{(r,k)}_{\mathrm{md},m}} - \eta \cdot \nabla f(\x_{\mathrm{md},m}^{(r,k)}; \xi^{(r,k)}_m)$. Thus, by $L$-smoothness (\cref{asm:fo:scvx:2o}(b)),
  \begin{equation*}
    F( \x^{(r,k+1)}_{\mathrm{ag}, m})
    \leq
    F( {\x^{(r,k)}_{\mathrm{md},m}}) - \eta \left\langle \nabla F({\x^{(r,k)}_{\mathrm{md},m}}), \nabla f(\x_{\mathrm{md},m}^{(r,k)}; \xi^{(r,k)}_m) \right\rangle + \frac{1}{2}  \eta^2 L \left\| \nabla f(\x_{\mathrm{md},m}^{(r,k)}; \xi^{(r,k)}_m)  \right\|_2^2 .
  \end{equation*}
  Taking conditional expectation gives
  \begin{align*}
    \expt \left[ F( \x^{(r,k+1)}_{\mathrm{ag}, m} ) |\mathcal{F}^{(r,k)} \right]
     & \leq
    F( {\x^{(r,k)}_{\mathrm{md},m}}) - \eta \left\| \nabla F (\x^{(r,k)}_{\mathrm{md},m})  \right\|_2^2+ \frac{1}{2} \eta^2 L \left\| \nabla F (\x^{(r,k)}_{\mathrm{md},m})  \right\|_2^2+ \frac{1}{2} \eta^2 L \sigma^2
    \\
     & = F( {\x^{(r,k)}_{\mathrm{md},m}}) - \eta \left( 1 - \frac{1}{2} \eta L \right) \left\| \nabla F (\x^{(r,k)}_{\mathrm{md},m})  \right\|_2^2+ \frac{1}{2} \eta^2 L \sigma^2.
  \end{align*}
  Since $\eta \leq \frac{1}{L}$ we have $1 - \frac{1}{2} \eta L \geq \frac{1}{2}$. Thus
  \begin{equation}
    \expt \left[ F( { \x^{(r,k+1)}_{\mathrm{ag}, m} }) \middle| \mathcal{F}^{(r,k)} \right]
    \leq
    F( {\x^{(r,k)}_{\mathrm{md},m}}) - \frac{1}{2} \eta \left\| \nabla F (\x^{(r,k)}_{\mathrm{md},m})  \right\|_2^2+ \frac{1}{2} \eta^2 L \sigma^2.
    \label{eq:fedaci:conv:2:1}
  \end{equation}
  Now we connect $F(\x_{\mathrm{md},m}^{(r,k)})$ with $F(\x_{\mathrm{ag},m}^{(r,k)})$ as follows.
  \begin{align}
         & F(\x_{\mathrm{md},m}^{(r,k)}) - F^{\star}
    \nonumber \\
    =    & (1 - \alpha^{-1}) \left( F(\x^{(r,k)}_{\mathrm{ag}, m}) - F^{\star} \right)
    + \alpha^{-1} \left( F(\x^{(r,k)}_{\mathrm{md},m}) - F^{\star} \right)
    + (1 - \alpha^{-1}) \left( F(\x^{(r,k)}_{\mathrm{md},m}) - F(\x_{\mathrm{ag},m}^{(r,k)}) \right)
    \nonumber \\
    \leq & (1 - \alpha^{-1}) \left( F(\x^{(r,k)}_{\mathrm{ag}, m}) - F^{\star} \right) - \frac{1}{2} \mu \alpha^{-1} \|\x^{(r,k)}_{\mathrm{md},m} - \x^{\star}\|_2^2 
    + \alpha^{-1} \left\langle \nabla F(\x^{(r,k)}_{\mathrm{md},m}), \x^{(r,k)}_{\mathrm{md},m} - \x^{\star}  \right\rangle
    \nonumber \\
         & \quad + (1 - \alpha^{-1}) \left\langle \nabla F(\x^{(r,k)}_{\mathrm{md},m}),  \x^{(r,k)}_{\mathrm{md},m} - \x^{(r,k)}_{\mathrm{ag}, m}  \right\rangle 
    \tag{$\mu$-strong-convexity}
    \\
    =    & (1 - \alpha^{-1}) \left( F(\x^{(r,k)}_{\mathrm{ag}, m}) - F^{\star} \right)  - \frac{1}{2}\mu  \alpha^{-1} \|\x^{(r,k)}_{\mathrm{md},m} - \x^{\star}\|_2^2 
    \nonumber \\
         & \quad + \alpha^{-1} \left\langle \nabla F(\x^{(r,k)}_{\mathrm{md},m}) ,  \alpha \beta^{-1} \x_m^{(r,k)} + (1 - \alpha \beta^{-1}) \x_{\mathrm{ag},m}^{(r,k)} - \x^{\star} \right\rangle,
    \label{eq:fedaci:conv:2:2}
  \end{align}
  where the last equality is due to the definition of $\x^{(r,k)}_{\mathrm{md},m}$. Plugging \cref{eq:fedaci:conv:2:2} to \cref{eq:fedaci:conv:2:1} completes the proof of \cref{fedaci:conv:2:claim}.
\end{proof}
Now we complete the proof of \cref{fedaci:conv:2} by assembling the bound for all clients in \cref{fedaci:conv:2:claim}.
\begin{proof}[Proof of \cref{fedaci:conv:2}]
  Average the bounds of \cref{fedaci:conv:2:claim} for $m = 1,\ldots,M$, which gives
  \begin{small}
  \begin{align*}
         & \expt \left[ \frac{1}{M} \sum_{m=1}^M F( {\x^{(r,k+1)}_{\mathrm{ag}, m}}) - F^{\star} \middle|\mathcal{F}^{(r,k)} \right]
    \\
    \leq & (1 - \alpha^{-1}) \left( \frac{1}{M} \sum_{m=1}^M F( {\x_{\mathrm{ag},m}^{(r,k)}}) - F^{\star} \right)
    - \frac{1}{2} \eta  \cdot \frac{1}{M} \sum_{m=1}^M   \left\|\nabla F (\x^{(r,k)}_{\mathrm{md},m})  \right\|_2^2 
    +  \frac{1}{2} \eta^2 L \sigma^2
    \\
         & + \alpha^{-1} \frac{1}{M} \sum_{m=1}^M \left\langle \nabla F(\x^{(r,k)}_{\mathrm{md},m}), \alpha \beta^{-1} \x_m^{(r,k)} + (1 - \alpha \beta^{-1}) \x_{\mathrm{ag},m}^{(r,k)} - \x^{\star} \right\rangle
    - \frac{1}{2} \mu \alpha^{-1} \frac{1}{M} \sum_{m=1}^M   \| {\x_{\mathrm{md},m}^{(r,k)}} - \x^{\star}\|_2^2 
    \\
    \leq & (1 - \alpha^{-1}) \left( \frac{1}{M} \sum_{m=1}^M F( {\x_{\mathrm{ag},m}^{(r,k)}}) - F^{\star} \right)
    - \frac{1}{2} \eta \left\| \frac{1}{M} \sum_{m=1}^M  \nabla F (\x^{(r,k)}_{\mathrm{md},m})  \right\|_2^2 
    +  \frac{1}{2} \eta^2 L \sigma^2
    \\
         & + \alpha^{-1} \frac{1}{M} \sum_{m=1}^M \left\langle \nabla F(\x^{(r,k)}_{\mathrm{md},m}), \alpha \beta^{-1} \x_m^{(r,k)} + (1 - \alpha \beta^{-1}) \x_{\mathrm{ag},m}^{(r,k)} - \x^{\star}\right\rangle
      - \frac{1}{2} \mu \alpha^{-1} \| \overline{\x^{(r,k)}_{\mathrm{md}}} - \x^{\star}\|_2^2 ,
  \end{align*}
  \end{small}
  where the last inequality is due to Jensen's inequality on the convex function $\|\cdot\|_2^2 $.
\end{proof}

\subsection{Details of Step 2: Proof of \cref{fedaci:stab:main}}
\label{sec:fedaci:stab:main}
In this subsection we prove \cref{fedaci:stab:main} regarding the growth of discrepancy overhead introduced in \cref{fedaci:conv:main}.

We first introduce a few more notations to simplify the discussions throughout this subsection. Let $m_1, m_2 \in [M]$ be two arbitrary distinct clients. For any timestep $(r, k)$, denote $\bDelta^{(r,k)} := \x^{(r,k)}_{m_1} - \x^{(r,k)}_{m_2}$,  $\bDelta^{(r,k)}_{\mathrm{ag}} := \x^{(r,k)}_{\mathrm{ag}, m_1} - \x^{(r,k)}_{\mathrm{ag}, m2}$ and $\bDelta^{(r,k)}_{\mathrm{md}} := \x^{(r,k)}_{\mathrm{md}, m_1} - \x^{(r,k)}_{\mathrm{md}, m_2}$ be the corresponding vector differences. Let $\bDelta^{(r,k)}_{\varepsilon} = \varepsilon^{(r,k)}_{m_1} - \varepsilon^{(r,k)}_{m_2}$, where $\varepsilon^{(r,k)}_m := \nabla f(\x_{\mathrm{md},m}^{(r,k)}; \xi^{(r,k)}_m) - \nabla F(\x^{(r,k)}_{\mathrm{md},m})$ be the noise of the stochastic gradient oracle of the $m$-th client evaluated at $\x^{(r,k)}_{\mathrm{md},m}$.

The proof of \cref{fedaci:stab:main} is based on the following propositions.

The following \cref{fedaci:stab:1} studies the growth of $\begin{bmatrix} \bDelta^{(r,k)}_{\mathrm{ag}} \\ \bDelta^{(r,k)} \end{bmatrix}$ at each step. The proof of \cref{fedaci:stab:1} is deferred to \cref{sec:fedaci:stab:1}.
\begin{proposition}
  \label{fedaci:stab:1}
  In the same setting of \cref{fedaci:stab:main},
  there exists a matrix $\H^{(r,k)}$ such that $\mu \I \preceq \H^{(r,k)} \preceq L \I$ satisfying
  \begin{equation*}
    \begin{bmatrix}
      \bDelta^{(r,k+1)}_{\mathrm{ag}} \\ \bDelta^{(r,k+1)}
    \end{bmatrix}
    =
    \A (\mu, \gamma, \eta, \H^{(r,k)})
    \begin{bmatrix} \bDelta^{(r,k)}_{\mathrm{ag}} \\ \bDelta^{(r,k)} \end{bmatrix}
    -
    \begin{bmatrix} \eta \I \\ \gamma \I \end{bmatrix}
    \bDelta^{(r,k)}_{\varepsilon},
  \end{equation*}
  where $\A(\mu, \gamma, \eta, \H)$ is a matrix-valued function defined as
  \begin{equation}
    \A(\mu, \gamma, \eta, \H) = \frac{1}{1 + \gamma\mu}
    \begin{bmatrix}
      \I - \eta \H            & \gamma \mu (\I - \eta \H)
      \\
      - \gamma (\H - \mu \I) & \I - \gamma^2 \mu \H
      \\
    \end{bmatrix}.
    \label{eq:fedaci:A:def}
  \end{equation}
\end{proposition}
Let us pause for a moment and discuss the intuition of the next steps of our plan. Our goal is to bound the product of several $\A(\mu, \gamma, \eta, \H_i)$ where the $\H_i$ matrix may be different. The natural idea is to bound the uniform norm bound of $\A$ for some norm $\| \cdot \|_2$: $\sup_{\mu \I \preceq \H \preceq L \I} \|\A\|_2$. It is worth noticing that the matrix operator norm will not give the desired bound --- $\sup_{\mu \I \preceq \H \preceq L \I} \|\A\|_2$ is not sufficiently small for our purpose. Our approach is to leverage the ``transformed'' norm \cite{Golub.VanLoan-13} $\|\A\|_{\X} := \| \X^{-1} \A \X \|_2$ for certain non-singular $\X$ and analyze the uniform norm bound for $\sup_{\mu \I \preceq \H \preceq L \I} \| \X^{-1} \A \X \|_2$. 

Formally, the following \cref{fedaci:stab:bound} studies the uniform norm bound of $\A$ under the proposed transformation $\X$. The proof of \cref{fedaci:stab:bound} is deferred to \cref{sec:fedaci:stab:bound}.
\begin{proposition}[Uniform norm bound of $\A$ under transformation $\X$]
  \label{fedaci:stab:bound}
  Let $\A(\mu, \gamma, \eta, \H)$ be defined in \cref{eq:fedaci:A:def}.
  and assume $\mu > 0$, $\gamma \in [\eta, \sqrt{\frac{\eta}{\mu}}]$, $\eta \in (0,\frac{1}{L}]$.
  Then the following uniform norm bound holds
  \begin{equation*}
    \sup_{\mu \I \preceq \H \preceq L \I}
    \left\| \X(\gamma, \eta)^{-1} \A(\mu, \gamma, \eta, \H) \X(\gamma, \eta) \right\|_2 \leq
    \begin{cases}
      1 + \frac{2\gamma^2 \mu}{\eta} & \text{if~} \gamma \in \left(\eta, \sqrt{\frac{\eta}{\mu}}\right], \\
      1                              & \text{if~} \gamma =  \eta,
    \end{cases}
  \end{equation*}
  where $\X (\gamma, \eta)$ is a matrix-valued function defined as
  \begin{equation}
    \X(\gamma, \eta) :=
    \begin{bmatrix}
      \frac{\eta}{\gamma} \I & \zeros
      \\
      \I                     & \I
    \end{bmatrix}.
    \label{eq:fedaci:X:def}
  \end{equation}
\end{proposition}

\cref{fedaci:stab:1,fedaci:stab:bound} suggest the one step growth of $ \left\| \X(\gamma, \eta)^{-1} \begin{bmatrix}
    \bDelta^{(r,k)}_{\mathrm{ag}}
    \\
    \bDelta^{(r,k)}
  \end{bmatrix}  \right\|_2^2$ as follows.
\begin{proposition}
  \label{fedaci:stab:2}
  In the same setting of \cref{fedaci:stab:main}, the following inequality holds (for all possible $(r, k)$)
  \begin{align*}
    & \expt \left[ \left\| \X(\gamma, \eta)^{-1} \begin{bmatrix}
        \bDelta^{(r,k+1)}_{\mathrm{ag}}
        \\
        \bDelta^{(r,k+1)}
      \end{bmatrix}  \right\|_2^2\middle| \mathcal{F}^{(r,k)} \right]
    \\ 
    \leq &
    2\gamma^2 \sigma^2 +
    \left\| \X(\gamma, \eta)^{-1} \begin{bmatrix}
      \bDelta^{(r,k)}_{\mathrm{ag}}
      \\
      \bDelta^{(r,k)}
    \end{bmatrix}  \right\|_2^2 
    \cdot
    \begin{cases}
      \left(1 + \frac{2\gamma^2 \mu}{\eta} \right)^2 & \text{if~} \gamma \in \left(\eta, \sqrt{\frac{\eta}{\mu}}\right], \\
      1                                              & \text{if~} \gamma =  \eta,
    \end{cases}
  \end{align*}
  where $\X$ is the matrix-valued function defined in \cref{eq:fedaci:X:def}.
\end{proposition}
The proof of \cref{fedaci:stab:2} is deferred to \cref{sec:fedaci:stab:2}.

The following \cref{fedaci:stab:3} relates the discrepancy overhead we wish to bound for \cref{fedaci:stab:main} with the quantity analyzed in \cref{fedaci:stab:2}.
The proof of \cref{fedaci:stab:3} is deferred to \cref{sec:fedaci:stab:3}.
\begin{proposition}
  \label{fedaci:stab:3}
  In the same setting of \cref{fedaci:stab:main}, the following inequality holds (for all $r, k$)
    \begin{align*}
      & \frac{1}{M} \sum_{m=1}^M
      \left\| \overline{\x^{(r,k)}_{\mathrm{md}}} - \x^{(r,k)}_{\mathrm{md},m}  \right\|_2
      \left\|  \frac{1}{1+\gamma\mu}(\overline{\x^{(r,k)}} - \x_m^{(r,k)}) + \frac{\gamma \mu}{1+\gamma\mu} (\overline{\x^{(r,k)}_{\mathrm{ag}}} - \x_{\mathrm{ag},m}^{(r,k)})  \right\|_2
      \\
      \leq & \frac{\sqrt{10} \eta}{\gamma}
      \left\| \X(\gamma, \eta)^{-1} \begin{bmatrix} \bDelta^{(r,k)}_{\mathrm{ag}} \\ \bDelta^{(r,k)} \end{bmatrix} \right\|_2^2 ,
    \end{align*}
  where $\X$ is the matrix-valued function defined in \cref{eq:fedaci:X:def}.
\end{proposition}

We are ready to finish the proof of \cref{fedaci:stab:main}.
\begin{proof}[Proof of \cref{fedaci:stab:main}]
  Recursively apply \cref{fedaci:stab:2} from $(r,0)$-th step to the $(r,k)$-th step (note that $\bDelta^{(r,0)}_{\mathrm{ag}} = \zeros$)
  \begin{align*}
    \expt \left[ \left\| \X(\gamma, \eta)^{-1} \begin{bmatrix}
      \bDelta^{(r,k)}_{\mathrm{ag}}
     \\
      \bDelta^{(r,k)}
    \end{bmatrix}  \right\|_2^2\middle| \mathcal{F}^{(r,0)} \right]
  & \leq
  2 \gamma^2 \sigma^2 k
  \cdot
  \begin{cases}
    \left( 1 + \frac{2\gamma^2 \mu}{\eta} \right)^{2k} & \text{if~} \gamma \in \left(\eta, \sqrt{\frac{\eta}{\mu}} \right], \\
    1                              & \text{if~} \gamma =  \eta.
  \end{cases}
  \end{align*}
  By \cref{fedaci:stab:3} we have
  \begin{align*}
    & \frac{1}{M} \sum_{m=1}^M
    \expt \left[ \left\| \overline{\x^{(r,k)}_{\mathrm{md}}} - \x^{(r,k)}_{\mathrm{md},m}  \right\|_2
    \left\|  \frac{1}{1+\gamma\mu}(\overline{\x^{(r,k)}} - \x_m^{(r,k)}) + \frac{\gamma \mu}{1+\gamma\mu} (\overline{\x^{(r,k)}_{\mathrm{ag}}} - \x_{\mathrm{ag},m}^{(r,k)})  \right\|_2 \middle| \mathcal{F}^{(r,0)} \right]
    \\
    \leq & \frac{\sqrt{10} \eta}{\gamma} \expt \left[ \left\| \X(\gamma, \eta)^{-1} \begin{bmatrix}
      \bDelta^{(r,k)}_{\mathrm{ag}}
     \\
      \bDelta^{(r,k)}
    \end{bmatrix}  \right\|_2^2\middle| \mathcal{F}^{(r,0)} \right]
    \leq 
    \begin{cases}
      7 \eta \gamma K \sigma^2 \left(1 + \frac{2\gamma^2\mu}{\eta}\right)^{2K}
       & \text{if~} \gamma \in \left(\eta, \sqrt{\frac{\eta}{\mu}}\right],
      \\
      7 \eta^2 K \sigma^2
       &
      \text{if~} \gamma = \eta,
    \end{cases}
   \end{align*}
  where in the last inequality we used the estimate that $2\sqrt{10} < 7$ and the fact that $k < K$. 
\end{proof}

\subsubsection{Proof of \cref{fedaci:stab:1}}
\label{sec:fedaci:stab:1}
In this section we will prove \cref{fedaci:stab:1}.
Let us first state and prove a more general version of \cref{fedaci:stab:1} regarding \fedac with general hyperparameter assumptions $\alpha \geq 1$, $\beta \geq 1$ .
\begin{claim}
  \label{fedac:general:stab}
  Assume \cref{asm:fo:scvx:2o} and assume $F$ to be $\mu > 0$-strongly convex. 
  For any $r, k$, there exists a matrix $\H^{(r,k)}$ such that $\mu \I \preceq \H^{(r,k)} \preceq L\I$ satisfying
  \begin{align*}
    & \begin{bmatrix} \bDelta^{(r,k+1)}_{\mathrm{ag}} \\ \bDelta^{(r,k+1)} \end{bmatrix}
    \\
    = &
    \begin{bmatrix}
      (1 - \beta^{-1}) (\I - \eta \H^{(r,k)})
       &
      \beta^{-1} (\I - \eta \H^{(r,k)})
      \\
      (1 - \beta^{-1}) (\alpha^{-1} - \gamma \H^{(r,k)})
       &
      \beta^{-1} (\alpha^{-1} \I - \gamma \H^{(r,k)}) + (1 - \alpha^{-1}) \I
    \end{bmatrix}
    \begin{bmatrix} \bDelta^{(r,k)}_{\mathrm{ag}} \\ \bDelta^{(r,k)} \end{bmatrix}
    -
    \begin{bmatrix} \eta \I \\  \gamma \I \end{bmatrix} \bDelta^{(r,k)}_{\varepsilon}.
  \end{align*}
\end{claim}
\begin{proof}[Proof of \cref{fedac:general:stab}]
  First note that \fedac can be written as the following two-state recursions.
  \begin{align*}
    \x^{(r,k+1)}_{\mathrm{ag}, m} & =  (1 - \beta^{-1}) \x^{(r,k)}_{\mathrm{ag}, m} + \beta^{-1} \x_m^{(r,k)}  - \eta \cdot \nabla F ( \x^{(r,k)}_{\mathrm{md},m} ) - \eta \varepsilon^{(r,k)}_m;
    \\
    \x^{(r,k+1)}_m                & = \alpha^{-1} \x^{(r,k)}_{\mathrm{md},m} + (1 - \alpha^{-1}) \x_m^{(r,k)} - \gamma \cdot \nabla F(\x^{(r,k)}_{\mathrm{md},m}) - \gamma \varepsilon^{(r,k)}_m
    \\
                             & =  \alpha^{-1} (1 - \beta^{-1}) \x^{(r,k)}_{\mathrm{ag}, m} + (1 - \alpha^{-1} + \alpha^{-1} \beta^{-1}) \x_m^{(r,k)} - \gamma \cdot \nabla F(\x^{(r,k)}_{\mathrm{md},m}) - \gamma \varepsilon^{(r,k)}_m.
  \end{align*}
  Taking difference gives
  \begin{align*}
    \bDelta^{(r,k+1)}_{\mathrm{ag}} & = (1 - \beta^{-1}) \bDelta^{(r,k)}_{\mathrm{ag}}  + \beta^{-1} \bDelta^{(r,k)}
    - \eta \left( \nabla F ( \x^{(r,k)}_{\mathrm{md}, m_1} )  -  \nabla F ( \x^{(r,k)}_{\mathrm{md}, m_2} ) \right)
    - \eta \bDelta^{(r,k)}_{\varepsilon};
    \\
    \bDelta^{(r,k+1)}               & = \alpha^{-1} (1 - \beta^{-1}) \bDelta^{(r,k)}_{\mathrm{ag}} + (1 - \alpha^{-1} + \alpha^{-1} \beta^{-1}) \bDelta^{(r,k)} 
    \\
    & \qquad \qquad - \gamma \left( \nabla F ( \x^{(r,k)}_{\mathrm{md}, m_1} )  -  \nabla F ( \x^{(r,k)}_{\mathrm{md}, m_2} ) \right)
    - \gamma  \bDelta^{(r,k)}_{\varepsilon}.
  \end{align*}
  By mean-value theorem, there exists a symmetric positive-definite matrix $\H^{(r,k)}$ such that $\mu \I \preceq \H^{(r,k)} \preceq L \I$ satisfying
  \begin{equation*}
    \nabla F(\x^{(r,k)}_{\mathrm{md}, {m_1}}) -  \nabla F(\x^{(r,k)}_{\mathrm{md}, {m_2}})
    =
    \H^{(r,k)} \bDelta^{(r,k)}_{\mathrm{md}}
    =
    \H^{(r,k)} \left( (1 - \beta^{-1}) \bDelta^{(r,k)}_{\mathrm{ag}} +  \beta^{-1} \bDelta^{(r,k)} \right).
  \end{equation*}
  Thus
  \begin{align*}
    \bDelta^{(r,k+1)}_{\mathrm{ag}} & = (1 - \beta^{-1}) \bDelta^{(r,k)}_{\mathrm{ag}}  + \beta^{-1} \bDelta^{(r,k)}
    - \eta \H^{(r,k)} \left( (1 - \beta^{-1}) \bDelta^{(r,k)}_{\mathrm{ag}} +  \beta^{-1} \bDelta^{(r,k)} \right)
    - \eta \bDelta^{(r,k)}_{\varepsilon}
    \\
    \bDelta^{(r,k+1)}               & = \alpha^{-1} (1 - \beta^{-1}) \bDelta^{(r,k)}_{\mathrm{ag}} + (1 - \alpha^{-1} + \alpha^{-1} \beta^{-1}) \bDelta^{(r,k)}
    \\
    & \qquad \qquad - \gamma \H^{(r,k)} \left( (1 - \beta^{-1}) \bDelta^{(r,k)}_{\mathrm{ag}} +  \beta^{-1} \bDelta^{(r,k)} \right)
    - \gamma \bDelta^{(r,k)}_{\varepsilon}
  \end{align*}
  Rearranging into matrix form completes the proof of  \cref{fedac:general:stab}.
\end{proof}
\cref{fedaci:stab:1} is a special case of \cref{fedac:general:stab}.
\begin{proof}[Proof of \cref{fedaci:stab:1}]
  The proof follows instantly by applying \cref{fedac:general:stab} with particular choice $\alpha = \frac{1}{\gamma \mu}$ and $\beta = \alpha + 1 = \frac{1 + \gamma \mu}{\gamma \mu}$.
\end{proof}

\subsubsection{Proof of \cref{fedaci:stab:bound}: uniform norm bound}
\label{sec:fedaci:stab:bound}
\begin{proof}[Proof of \cref{fedaci:stab:bound}]
  Define another matrix-valued function $\B$ as
  \begin{equation*}
    \B(\mu, \gamma, \eta, \H) := \X(\gamma, \eta)^{-1} \A(\mu, \gamma, \eta, \H) \X(\gamma, \eta).
  \end{equation*}
  Since $ \X(\gamma, \eta)^{-1} =
  \begin{bmatrix}
    \frac{\gamma}{\eta} \I  & \zeros
    \\
    -\frac{\gamma}{\eta} \I & \I
  \end{bmatrix}$ we can compute that
  \begin{equation*}
    \B(\mu, \gamma, \eta, \H) = \frac{1}{(1 + \gamma\mu)\eta}
    \begin{bmatrix}
      (\eta + \gamma^2 \mu) (\I - \eta \H) & \gamma^2 \mu (\I - \eta \H)
      \\
      - \mu (\gamma^2 - \eta^2) \I        & (\eta - \gamma^2 \mu) \I
    \end{bmatrix}.
  \end{equation*}
  \sloppy Define the four blocks of $\B(\mu, \gamma, \eta, \H)$ as $\B_{11}(\mu, \gamma, \eta, \H)$, $\B_{12}(\mu, \gamma, \eta, \H)$, $\B_{21}(\mu, \gamma, \eta)$, $\B_{22}(\mu, \gamma, \eta)$ (note that the lower two blocks do not involve $H$), \ie,
  \begin{align*}
     & \B_{11}(\mu, \gamma, \eta, \H) = \frac{\eta + \gamma^2 \mu}{(1 + \gamma \mu)\eta}(\I - \eta \H),
     &
     & \B_{12}(\mu, \gamma, \eta, \H)  = \frac{\gamma^2 \mu}{(1 + \gamma \mu) \eta}(\I - \eta \H),
    \\
     & \B_{21}(\mu, \gamma, \eta)    = -\frac{\mu (\gamma^2 - \eta^2)}{(1 + \gamma \mu) \eta} \I,
     &
     & \B_{22}(\mu, \gamma, \eta)    = \frac{\eta - \gamma^2 \mu}{(1 + \gamma \mu) \eta} \I.
  \end{align*}
  \paragraph{Case I: $\eta < \gamma \leq \sqrt{\frac{\eta}{\mu}}$.} In this case we have
  \begin{align}
    \|\B_{11}(\mu, \gamma, \eta, \H)\|_2 &
    \leq \frac{\eta + \gamma^2 \mu}{(1 + \gamma\mu)\eta}(1 - \eta \mu)
    \leq \frac{\eta + \gamma^2 \mu}{\eta}
    = 1 + \frac{\gamma^2 \mu}{\eta},
    \tag{since $\eta \mu \leq 1$}
    \\
    \|\B_{12}(\mu, \gamma, \eta, \H)\|_2 &
    \leq \frac{\gamma^2 \mu}{(1 + \gamma\mu)\eta}(1 - \eta \mu)
    \leq \frac{\gamma^2 \mu}{\eta},
    \tag{since $\eta \mu \leq 1$}
    \\
    \|\B_{21}(\mu, \gamma, \eta)\|_2    &
    = \frac{\mu (\gamma^2 - \eta^2)}{(1 + \gamma \mu) \eta}
    \leq \frac{\gamma^2 \mu}{\eta},
    \tag{since $\eta < \gamma \leq \sqrt{\frac{\eta}{\mu}}$}
    \\
    \|\B_{22}(\mu, \gamma, \eta)\|_2    & = \frac{\eta - \gamma^2\mu}{(1 + \gamma \mu)\eta} \leq \frac{1}{1 + \gamma \mu}\leq 1.
    \tag{since $\gamma \leq \sqrt{\frac{\eta}{\mu}}$}
  \end{align}
  The operator norm of $\B$ can be bounded via its blocks via helper \cref{helper:blocknorm} as
  \begin{align}
         & \|\B(\mu, \gamma, \eta, \H)\|_2
    \nonumber \\
    \leq & \max \left\{\| \B_{11}(\mu, \gamma, \eta, \H) \|_2, \| \B_{22}(\mu, \gamma, \eta) \|)_2 \right\}
    +
    \max \left\{\| \B_{12}(\mu, \gamma, \eta, \H) \|_2, \| \B_{21}(\mu, \gamma, \eta) \|)_2 \right\} \tag{by \cref{helper:blocknorm}}
    \\
    \leq & \max \left\{ 1 + \frac{\gamma^2 \mu}{\eta}, 1 \right\} + \max \left\{ \frac{\gamma^2 \mu}{\eta}, \frac{\gamma^2 \mu}{\eta} \right\} = 1 + \frac{2\gamma^2\mu}{\eta}.
    \nonumber 
  \end{align}
  \paragraph{Case II: $\gamma = \eta$.} In this case we have
  \begin{align*}
    \|\B_{11}(\mu, \gamma, \eta, \H)\|_2 &
    \leq \frac{\eta + \eta^2 \mu}{(1 + \eta\mu)\eta}(1 - \eta \mu)
    = 1 - \eta \mu,
    \\
    \|\B_{12}(\mu, \gamma, \eta, \H)\|_2 &
    \leq \frac{\eta^2 \mu}{(1 + \eta\mu)\eta}(1 - \eta \mu)
    = \frac{(1 - \eta \mu)\eta \mu}{1 + \eta \mu},
    \\
    \|\B_{21}(\mu, \gamma, \eta)\|_2    &
    = 0,
    \\
    \|\B_{22}(\mu, \gamma, \eta)\|_2    & = \frac{\eta - \eta^2\mu}{(1 + \eta \mu)\eta}
    = \frac{1 - \eta \mu}{1 + \eta \mu}.
  \end{align*}
   Similarly, the operator norm of block matrix $\B$ can be bounded via its blocks via helper \cref{helper:blocknorm} as
  \begin{align}
         & \B(\mu, \gamma, \eta, \H)
    \nonumber \\
    \leq & \max \left\{\| \B_{11}(\mu, \gamma, \eta, \H) \|_2, \| \B_{22}(\mu, \gamma, \eta) \|)_2 \right\}
    +
    \max \left\{\| \B_{12}(\mu, \gamma, \eta, \H) \|_2, \| \B_{21}(\mu, \gamma, \eta) \|)_2 \right\} \tag{by \cref{helper:blocknorm}}
    \\
    \leq & \max \left\{ 1 - \eta \mu, \frac{1 - \eta \mu}{1 + \eta \mu} \right\} + \frac{\eta \mu ( 1- \eta \mu)}{1 + \eta \mu} =  1 - \eta \mu + \frac{\eta \mu ( 1- \eta \mu)}{1 + \eta \mu}
    =
    \frac{1 + \eta \mu - 2 \eta^2 \mu^2}{1 + \eta \mu} \leq 1.
    \nonumber 
  \end{align}
  Summarizing the above two cases completes the proof of \cref{fedaci:stab:bound}.
\end{proof}

\subsubsection{Proof of \cref{fedaci:stab:2}}
\label{sec:fedaci:stab:2}
In this section we apply \cref{fedaci:stab:1,fedaci:stab:bound} to establish \cref{fedaci:stab:2}.
\begin{proof}[Proof of \cref{fedaci:stab:2}]
  Multiplying $\X(\gamma, \eta)^{-1}$ to the left on both sides of \cref{fedaci:stab:1} gives
  \begin{align*}
    \X(\gamma, \eta)^{-1} \begin{bmatrix}
      \bDelta^{(r,k+1)}_{\mathrm{ag}}
      \\
      \bDelta^{(r,k+1)}
    \end{bmatrix}
     & =
    \X(\gamma, \eta)^{-1}  \A(\mu, \gamma, \eta, \H)
    \begin{bmatrix}
      \bDelta^{(r,k)}_{\mathrm{ag}}
      \\
      \bDelta^{(r,k)}
    \end{bmatrix}
    -
    \X(\gamma, \eta)^{-1}  \begin{bmatrix} \eta \I \\ \gamma \I \end{bmatrix} \bDelta^{(r,k)}_{\varepsilon}
    \\
     & =
    \X(\gamma, \eta)^{-1} \A(\mu, \gamma, \eta, \H^{(r,k)}) \X(\gamma, \eta)^{-1} \left( \X(\gamma, \eta) \begin{bmatrix}
        \bDelta^{(r,k)}_{\mathrm{ag}}
        \\
        \bDelta^{(r,k)}
      \end{bmatrix}  \right)
    -
    \begin{bmatrix}   \gamma \I \\ \zeros \end{bmatrix} \bDelta^{(r,k)}_{\varepsilon},
  \end{align*}
  where the last equality is due to
  \begin{equation*}
    \X(\gamma, \eta)^{-1} =
    \begin{bmatrix}
      \frac{\gamma}{\eta} \I  & \zeros
      \\
      -\frac{\gamma}{\eta} \I & \I
    \end{bmatrix},\qquad
    \X(\gamma, \eta)^{-1}
    \begin{bmatrix} \eta \I \\ \gamma \I \end{bmatrix}
    =
    \begin{bmatrix}
      \gamma \I \\ \zeros
    \end{bmatrix}.
  \end{equation*}
  Taking conditional expectation,
  \begin{align}
         & \expt \left[ \left\| \X(\gamma, \eta)^{-1} \begin{bmatrix}
        \bDelta^{(r,k+1)}_{\mathrm{ag}}
        \\
        \bDelta^{(r,k+1)}
      \end{bmatrix}  \right\|_2^2\middle| \mathcal{F}^{(r,k)} \right]
    \nonumber  \\
    =    & \left\| \X^{-1} \A \X
    \left( \X^{-1}
    \begin{bmatrix}
      \bDelta^{(r,k)}_{\mathrm{ag}} \\ \bDelta^{(r,k)}
    \end{bmatrix} \right) \right\|_2^2 
    +
    \expt \left[\left\| \begin{bmatrix}
        \gamma \I \\ \zeros
      \end{bmatrix}
      \bDelta^{(r,k)}_{\varepsilon} \right\|_2^2\middle| \mathcal{F}^{(r,k)}  \right]
    \tag{independence}
    \\
    \leq & \| \X^{-1} \A \X \|_2^2 
    \left\|  \X^{-1}
    \begin{bmatrix}
      \bDelta^{(r,k)}_{\mathrm{ag}} \\ \bDelta^{(r,k)}
    \end{bmatrix}\right\|_2^2 
    + 2\gamma^2 \sigma^2
    \tag{bounded variance, sub-multiplicativity}
    \\
    \leq & 2\gamma^2 \sigma^2 +
    \left\| \X(\gamma, \eta)^{-1} \begin{bmatrix}
      \bDelta^{(r,k)}_{\mathrm{ag}}
      \\
      \bDelta^{(r,k)}
    \end{bmatrix}  \right\|_2^2 
    \cdot
    \begin{cases}
      \left(1 + \frac{2\gamma^2 \mu}{\eta} \right)^2 & \text{if~} \gamma \in \left(\eta, \sqrt{\frac{\eta}{\mu}}\right], \\
      1                                              & \text{if~} \gamma =  \eta.
    \end{cases}
    \tag{by \cref{fedaci:stab:bound}}
  \end{align}
\end{proof}

\subsubsection{Proof of \cref{fedaci:stab:3}}
\label{sec:fedaci:stab:3}
In this section we will prove \cref{fedaci:stab:3} in three steps via the following three claims. For all the three claims $\X$ stands for the matrix-valued functions defined in \cref{eq:fedaci:X:def}.
\begin{claim}
  \label{fedaci:stab:3:1}
  In the same setting of  \cref{fedaci:stab:3}, 
  \begin{align*}
        &  \frac{1}{M} \sum_{m=1}^M
         \left\| \overline{\x^{(r,k)}_{\mathrm{md}}} - \x^{(r,k)}_{\mathrm{md},m}  \right\|_2
         \left\|  \frac{1}{1+\gamma\mu}(\overline{\x^{(r,k)}} - \x_m^{(r,k)}) + \frac{\gamma \mu}{1+\gamma\mu} (\overline{\x^{(r,k)}_{\mathrm{ag}}} - \x_{\mathrm{ag},m}^{(r,k)})  \right\|_2
    \\
    \leq &  \left\| \begin{bmatrix} \frac{1}{1 + \gamma \mu} \I \\ \frac{\gamma \mu}{1 + \gamma \mu} \I \end{bmatrix}^\top \X(\gamma, \eta) \right\|_2 \cdot
    \left\| \begin{bmatrix} \frac{\gamma \mu}{1 + \gamma \mu} \I \\ \frac{1}{1 + \gamma \mu} \I \end{bmatrix}^\top \X(\gamma, \eta) \right\|_2 \cdot
    \left\| \X(\gamma, \eta)^{-1}  \begin{bmatrix} \bDelta^{(r,k)}_{\mathrm{ag}} \\ \bDelta^{(r,k)} \end{bmatrix}
    \right\|_2^2 .
  \end{align*}
\end{claim}
\begin{claim}
  \label{fedaci:stab:left1}
  Assume $\mu > 0$, $\gamma \in [\eta, \sqrt{\frac{\eta}{\mu}}]$, $\eta \in (0,\frac{1}{L}]$, then
  \(
    \left\| \X(\gamma, \eta)^\top
  \begin{bmatrix} \frac{1}{1 + \gamma \mu} \I \\ \frac{\gamma \mu}{1 + \gamma \mu} \I \end{bmatrix} \right\|_2 \leq \frac{\sqrt{5} \eta}{\gamma}.
  \)
\end{claim}
\begin{claim}
  \label{fedaci:stab:left2}
  Assume $\mu > 0$, $\gamma \in [\eta, \sqrt{\frac{\eta}{\mu}}]$, $\eta \in (0,\frac{1}{L}]$, then
  \(
    \left\| \X(\gamma, \eta)^\top
    \begin{bmatrix} \frac{\gamma \mu}{1 + \gamma \mu} \I \\ \frac{1}{1 + \gamma \mu} \I \end{bmatrix}
    \right\|_2 \leq \sqrt{2}.
  \)
\end{claim}
\cref{fedaci:stab:3} follows immediately once we have \cref{fedaci:stab:3:1,fedaci:stab:left1,fedaci:stab:left2}.
\begin{proof}[Proof of \cref{fedaci:stab:3}]
  Follows trivially with \cref{fedaci:stab:3:1,fedaci:stab:left1,fedaci:stab:left2}.
  \begin{align*}
    & \frac{1}{M} \sum_{m=1}^M
    \left\| \overline{\x^{(r,k)}_{\mathrm{md}}} - \x^{(r,k)}_{\mathrm{md},m}  \right\|_2
    \left\|  \frac{1}{1+\gamma\mu}(\overline{\x^{(r,k)}} - \x_m^{(r,k)}) + \frac{\gamma \mu}{1+\gamma\mu} (\overline{\x^{(r,k)}_{\mathrm{ag}}} - \x_{\mathrm{ag},m}^{(r,k)})  \right\|_2
    \\
    \leq & 
    \frac{\sqrt{10} \eta}{\gamma}
    \left\| \X(\gamma, \eta)^{-1} \begin{bmatrix} \bDelta^{(r,k)}_{\mathrm{ag}} \\ \bDelta^{(r,k)} \end{bmatrix} \right\|_2^2 .
  \end{align*}
\end{proof}
Now we finish the proof of the three claims.
\begin{proof}[Proof of \cref{fedaci:stab:3:1}]
  Note that 
  \begin{align}
         & \frac{1}{M} \sum_{m=1}^M  \left\| \overline{\x^{(r,k)}_\mathrm{md}} - \x^{(r,k)}_{\mathrm{md}, m} \right\|_2^2 
          \leq \|\bDelta^{(r,k)}_{\mathrm{md}}\|_2^2 
    \tag{convexity of $\|\cdot\|_2^2 $}
    \\
    =    & \left\|
    \begin{bmatrix} (1 - \beta^{-1}) \I \\ \beta^{-1} \I \end{bmatrix}^\top
    \begin{bmatrix} \bDelta^{(r,k)}_{\mathrm{ag}} \\ \bDelta^{(r,k)} \end{bmatrix}
    \right\|_2^2 
        = 
      \left\|
    \begin{bmatrix} \frac{1}{1 + \gamma \mu} \I \\ \frac{\gamma \mu}{1 + \gamma \mu} \I \end{bmatrix}^\top
    \begin{bmatrix} \bDelta^{(r,k)}_{\mathrm{ag}} \\ \bDelta^{(r,k)} \end{bmatrix}
    \right\|_2^2 
        \tag{definition of ``md''}
    \\
    \leq & \left\| \begin{bmatrix} \frac{1}{1 + \gamma \mu} \I \\ \frac{\gamma \mu}{1 + \gamma \mu} \I \end{bmatrix}^\top \X(\gamma, \eta) \right\|_2^2 
    \left\|  \X(\gamma, \eta)^{-1}
    \begin{bmatrix} \bDelta^{(r,k)}_{\mathrm{ag}} \\ \bDelta^{(r,k)} \end{bmatrix}
    \right\|_2^2,
    \tag{sub-multiplicativity}
  \end{align}
  and similarly
  \begin{align}
         & \frac{1}{M} \sum_{m=1}^M \left\|  \frac{1}{1+\gamma\mu}(\overline{\x^{(r,k)}} - \x_m^{(r,k)}) + \frac{\gamma \mu}{1+\gamma\mu} (\overline{\x^{(r,k)}_{\mathrm{ag}}} - \x_{\mathrm{ag},m}^{(r,k)})  \right\|_2^2 
    \nonumber \\
    \leq & \left\|
      \begin{bmatrix} \frac{\gamma \mu}{1 + \gamma \mu} \I \\ \frac{1}{1 + \gamma \mu} \I \end{bmatrix}^\top
      \begin{bmatrix} \bDelta^{(r,k)}_{\mathrm{ag}} \\ \bDelta^{(r,k)} \end{bmatrix}
      \right\|_2^2
    \tag{convexity of $\|\cdot\|_2^2 $}
    \\
    \leq & \left\| 
    \begin{bmatrix} \frac{\gamma \mu}{1 + \gamma \mu} \I \\ \frac{1}{1 + \gamma \mu} \I \end{bmatrix}^\top \X(\gamma, \eta) \right\|_2^2 
      \left\| \X(\gamma, \eta)^{-1}
      \begin{bmatrix} \bDelta^{(r,k)}_{\mathrm{ag}} \\ \bDelta^{(r,k)} \end{bmatrix}
      \right\|_2^2.
      \tag{sub-multiplicativity}
  \end{align}
  Thus, by Cauchy-Schwarz inequality,
  \begin{small}
  \begin{align}
         & \frac{1}{M} \sum_{m=1}^M
         \left\| \overline{\x^{(r,k)}_{\mathrm{md}}} - \x^{(r,k)}_{\mathrm{md},m}  \right\|_2
         \left\|  \frac{1}{1+\gamma\mu}(\overline{\x^{(r,k)}} - \x_m^{(r,k)}) + \frac{\gamma \mu}{1+\gamma\mu} (\overline{\x^{(r,k)}_{\mathrm{ag}}} - \x_{\mathrm{ag},m}^{(r,k)})  \right\|_2
    \nonumber \\
    \leq &
    \sqrt{
    \left(\frac{1}{M} \sum_{m=1}^M  \left\| \overline{\x^{(r,k)}_\mathrm{md}} - \x^{(r,k)}_{\mathrm{md}, m} \right\|_2^2\right)
    \left(\frac{1}{M} \sum_{m=1}^M \left\|  \frac{1}{1+\gamma\mu}(\overline{\x^{(r,k)}} - \x_m^{(r,k)}) + \frac{\gamma \mu}{1+\gamma\mu} (\overline{\x^{(r,k)}_{\mathrm{ag}}} - \x_{\mathrm{ag},m}^{(r,k)})  \right\|_2^2\right)
    }
    \tag{Cauchy-Schwarz}
    \\
    \leq & \left\| \begin{bmatrix} \frac{1}{1 + \gamma \mu} \I \\ \frac{\gamma \mu}{1 + \gamma \mu} \I \end{bmatrix}^\top \X(\gamma, \eta) \right\|_2 \cdot
    \left\| \begin{bmatrix} \frac{\gamma \mu}{1 + \gamma \mu} \I \\ \frac{1}{1 + \gamma \mu} \I \end{bmatrix}^\top \X(\gamma, \eta) \right\|_2 \cdot
    \left\| \X(\gamma, \eta)^{-1}  \begin{bmatrix} \bDelta^{(r,k)}_{\mathrm{ag}} \\ \bDelta^{(r,k)} \end{bmatrix}
    \right\|_2^2 ,
    \nonumber 
  \end{align}
  \end{small}
  completing the proof of \cref{fedaci:stab:3:1}.
\end{proof}

\begin{proof}[Proof of \cref{fedaci:stab:left1}]
  Direct calculation shows that
  \begin{equation*}
    \X(\gamma, \eta)^\top
    \begin{bmatrix} \frac{1}{1 + \gamma\mu}  \I \\ \frac{\gamma \mu}{1 + \gamma \mu} \I \end{bmatrix}
    =
    \begin{bmatrix}
      \frac{\eta}{\gamma} \I & \I
      \\
      0                     & \I
    \end{bmatrix}
    \begin{bmatrix} \frac{1}{1 + \gamma\mu}  \I \\ \frac{\gamma \mu}{1 + \gamma \mu} \I \end{bmatrix}
    =
    \frac{1}{1 + \gamma \mu}
    \begin{bmatrix} (\frac{\eta}{\gamma} + \gamma \mu)  \I \\ \gamma \mu \I \end{bmatrix}.
  \end{equation*}
  Since
  \begin{equation}
    \left\| \begin{bmatrix} (\frac{\eta}{\gamma} + \gamma \mu)  \I \\ \gamma \mu \I \end{bmatrix} \right\|_2
    =
    \sqrt{\left(\frac{\eta}{\gamma} + \gamma \mu\right)^2  + (\gamma \mu)^2}
    \leq
    \sqrt{\left(\frac{2\eta}{\gamma} \right)^2  + \left(\frac{\eta}{\gamma} \right)^2}
    = \frac{\sqrt{5}\eta}{\gamma}.
    \tag{since $\gamma \mu \leq \frac{\eta}{\gamma}$}
  \end{equation}
  We conclude that
  \begin{equation*}
    \left\| \X(\gamma, \eta)^\top
  \begin{bmatrix} \frac{1}{1 + \gamma\mu}  \I \\ \frac{\gamma \mu}{1 + \gamma \mu} \I \end{bmatrix}
    \right\|_2
    \leq \frac{1}{1 + \gamma \mu} \cdot \frac{\sqrt{5}\eta}{\gamma} \leq \frac{\sqrt{5} \eta}{\gamma}.
  \end{equation*}
\end{proof}

\begin{proof}[Proof of \cref{fedaci:stab:left2}]
  Direct calculation shows that
  \begin{equation*}
    \X(\gamma, \eta)^\top
    \begin{bmatrix} \frac{\gamma \mu}{1 + \gamma \mu} \I \\ \frac{1}{1 + \gamma \mu} \I \end{bmatrix}
    =
    \begin{bmatrix}
      \frac{\eta}{\gamma} \I & \I
      \\
      0                     & \I
    \end{bmatrix}
    \begin{bmatrix} \frac{\gamma \mu}{1 + \gamma \mu} \I \\ \frac{1}{1 + \gamma \mu} \I \end{bmatrix}
    =
    \begin{bmatrix} \frac{1 + \eta \mu}{1 + \gamma \mu} \I \\  \frac{1}{1 + \gamma \mu} \I \end{bmatrix},
  \end{equation*}
  and
  \begin{equation}
    \left\| \begin{bmatrix} \frac{1 + \eta \mu}{1 + \gamma \mu} \I \\  \frac{1}{1 + \gamma \mu} \I \end{bmatrix} \right\|_2
    =
    \sqrt{ \left( \frac{1 + \eta \mu}{1 + \gamma \mu} \right)^2 + \left( \frac{1}{1 + \gamma \mu} \right)^2 }
    \leq \sqrt{2}
    \tag{since $\eta \leq  \gamma$},
  \end{equation}
  completing the proof of \cref{fedaci:stab:left2}.
\end{proof}

\subsection{Details of Step 3: Proof of \cref{fedaci:general:eta}}
\label{sec:fedaci:step3}
\begin{proof}[Proof of \cref{fedaci:general:eta}]
  It is direct to verify that $\gamma = \max \left\{\eta, \sqrt{\frac{\eta}{\mu K}} \right\} \in \left[\eta, \sqrt{\frac{\eta}{\mu}}\right]$ so both \cref{fedaci:conv:main,fedaci:stab:main} are applicable.
  Applying \cref{fedaci:conv:main} yields
    \begin{align}
      & \expt[\Psi^{(R,0)}] \leq \exp\left( - \max \left\{ \eta \mu, \sqrt{\frac{\eta \mu}{K}}\right\} KR \right) \Psi^{(0,0)}
      \nonumber \\
      & +
      \min\left\{ \frac{\eta L \sigma^2}{2 \mu}, \frac{\eta^{\frac{3}{2}} L K^{\frac{1}{2}} \sigma^2}{2 \mu^{\frac{1}{2}}} \right\}
      + \max\left\{ \frac{\eta \sigma^2}{2M}, \frac{\eta^{\frac{1}{2}} \sigma^2}{2 \mu^{\frac{1}{2}} M K^{\frac{1}{2}} }          \right\}
      \nonumber \\
                        & + L \cdot \max_{\substack{0 \leq r < R \\ 0 \leq k < K}}  \expt
      \left[\frac{1}{M} \sum_{m=1}^M \left\| \overline{\x^{(r,k)}_{\mathrm{md}}} - \x^{(r,k)}_{\mathrm{md},m}  \right\|_2
      \left\|  \frac{1}{1+\gamma\mu}(\overline{\x^{(r,k)}} - \x_m^{(r,k)}) + \frac{\gamma \mu}{1+\gamma\mu} (\overline{\x^{(r,k)}_{\mathrm{ag}}} - \x_{\mathrm{ag},m}^{(r,k)})  \right\|_2  \right].
      \label{eq:fedaci:proof:1}
    \end{align}
  We bound $\max\left\{ \frac{\eta \sigma^2}{2M}, \frac{\eta^{\frac{1}{2}} \sigma^2}{2 \mu^{\frac{1}{2}} M K^{\frac{1}{2}} } \right\}$ by $\frac{\eta \sigma^2}{2M}+ \frac{\eta^{\frac{1}{2}} \sigma^2}{2 \mu^{\frac{1}{2}} M K^{\frac{1}{2}} } $, and bound $    \min\left\{ \frac{\eta L \sigma^2}{2 \mu}, \frac{\eta^{\frac{3}{2}} L K^{\frac{1}{2}} \sigma^2}{2 \mu^{\frac{1}{2}}} \right\}$ by    $\frac{\eta^{\frac{3}{2}} L K^{\frac{1}{2}} \sigma^2}{2 \mu^{\frac{1}{2}}}$, which gives
  \begin{equation}
    \min\left\{ \frac{\eta L \sigma^2}{2 \mu}, \frac{\eta^{\frac{3}{2}} L K^{\frac{1}{2}} \sigma^2}{2 \mu^{\frac{1}{2}}} \right\}
    + \max\left\{ \frac{\eta \sigma^2}{2M}, \frac{\eta^{\frac{1}{2}} \sigma^2}{2 \mu^{\frac{1}{2}} M K^{\frac{1}{2}} }          \right\}
    \leq
    \frac{\eta^{\frac{3}{2}} L K^{\frac{1}{2}} \sigma^2}{2 \mu^{\frac{1}{2}}}  +  \frac{\eta \sigma^2}{2M} + \frac{\eta^{\frac{1}{2}} \sigma^2}{2 \mu^{\frac{1}{2}} M K^{\frac{1}{2}} }.
    \label{eq:fedaci:proof:1.1}
  \end{equation}
  Applying \cref{fedaci:stab:main} with $\gamma = \max\left\{ \eta, \sqrt{\frac{\eta}{\mu K}}\right\}$ gives
  \begin{align}
         & \expt
    \left[ \frac{1}{M} \sum_{m=1}^M
    \left\| \overline{\x^{(r,k)}_{\mathrm{md}}} - \x^{(r,k)}_{\mathrm{md},m} \right\|_2
    \left\|  \frac{1}{1+\gamma\mu}(\overline{\x^{(r,k)}} - \x_m^{(r,k)}) + \frac{\gamma \mu}{1+\gamma\mu} (\overline{\x^{(r,k)}_{\mathrm{ag}}} - \x_{\mathrm{ag},m}^{(r,k)})  \right\|_2  \right]
    \nonumber \\
    \leq &
    \begin{cases}
      7  \eta \sqrt{\frac{\eta}{\mu K}} K \sigma^2 \left(1 + \frac{2}{K}\right)^{2K}
       & \text{if~} \gamma = \sqrt{\frac{\eta}{\mu K}}
      \\
      7  \eta^2 K \sigma^2
       &
      \text{if~} \gamma = \eta
    \end{cases}
    \nonumber \\
    \leq & \frac{7  \euler^4 \eta^{\frac{3}{2}} K^{\frac{1}{2}} \sigma^2} {\mu^{\frac{1}{2}}} + 7 \eta^2 K \sigma^2.
    \label{eq:fedaci:proof:2}
  \end{align}
  Combining \cref{eq:fedaci:proof:1,eq:fedaci:proof:1.1,eq:fedaci:proof:2} yields
    \begin{equation*}
      \expt[\Psi^{(r,k)}]
      \leq \exp \left(  - \max \left\{ \eta \mu, \sqrt{\frac{\eta \mu}{K}}\right\} KR \right) \Psi^{(0,0)}
      + 
      \frac{\eta^{\frac{1}{2}} \sigma^2}{2 \mu^{\frac{1}{2}} M K^{\frac{1}{2}} }
      + \frac{\eta \sigma^2}{2M}
      + \frac{(7 \euler^4 + \frac{1}{2}) \eta^{\frac{3}{2}} L K^{\frac{1}{2}} \sigma^2} {\mu^{\frac{1}{2}}}
      + 7 \eta^2 L K \sigma^2.
    \end{equation*}
  The lemma then follows by leveraging the estimate $7 \euler^4 + \frac{1}{2} < 390$ for the coefficient of $\frac{\eta^{\frac{3}{2}} L K^{\frac{1}{2}} \sigma^2} {\mu^{\frac{1}{2}}}$.
\end{proof}

\subsection{Details of Step 4: Finishing the Proof of \cref{fedaci:full}}
\label{sec:fedaci:step4}
The main \cref{fedaci:full} then follows by plugging an appropriate $\eta$ to \cref{fedaci:general:eta}.
\begin{proof}[Proof of \cref{fedaci:full}]
  To simplify the notation, we denote the decreasing term in \cref{eq:fedaci:general:eta} as $\varphi_{\downarrow}(\eta)$ and the increasing term as $\varphi_{\uparrow}(\eta)$, namely
  \begin{align*}
    \varphi_{\downarrow}(\eta) & := \exp \left(  - \max \left\{ \eta \mu, \sqrt{\frac{\eta \mu}{K}}\right\} KR \right) \Psi^{(0,0)},
    \\
    \varphi_{\uparrow}(\eta) & :=  \frac{\eta^{\frac{1}{2}} \sigma^2}{2 \mu^{\frac{1}{2}} M K^{\frac{1}{2}} }
    + \frac{\eta \sigma^2}{2M}
    + \frac{390 \eta^{\frac{3}{2}} L K^{\frac{1}{2}} \sigma^2} {\mu^{\frac{1}{2}}}
    + 7 \eta^2 L K \sigma^2.
  \end{align*}
  Now let 
  \begin{equation*}
    \eta_0 := \frac{1}{\mu KR^2} \log^2 \left( \euler + \min \left\{ \frac{\mu M KR \Psi^{(0,0)}}{\sigma^2}, \frac{\mu^2 KR^3 \Psi^{(0,0)}}{L \sigma^2} \right\} \right),
  \end{equation*}
  and then $ \eta = \min \left\{ \frac{1}{L}, \eta_0  \right\}$. 
  Therefore, the decreasing term $\varphi_{\downarrow}(\eta)$  is upper bounded by $\varphi_{\downarrow}(\frac{1}{L}) + \varphi_{\downarrow}(\eta_0)$, where
  \begin{equation}
    \varphi_{\downarrow} \left(\frac{1}{L} \right)
    =
    \min \left\{ \exp \left( - \frac{\mu KR}{L} \right), \exp \left( - \frac{\mu^{\frac{1}{2}} K^{\frac{1}{2}} R}{L^{\frac{1}{2}} } \right)\right\}\Psi^{(0,0)},
    \label{eq:fedaci:proof:4}
  \end{equation}
  and
  \begin{align}
    \varphi_{\downarrow}(\eta_0) & \leq \exp \left(  - \sqrt{\eta_0 \mu KR^2} \right) \Psi^{(0,0)}
    \nonumber \\
    & =
    \left( \euler + \min \left\{ \frac{\mu M K R \Psi^{(0,0)}}{\sigma^2}, \frac{\mu^2 KR^3 \Psi^{(0,0)}}{L \sigma^2} \right\} \right)^{-1} \Psi^{(0,0)}
    \leq
    \frac{\sigma^2}{\mu M K R} + \frac{L \sigma^2}{ \mu^2 K R^3}.
    \label{eq:fedaci:proof:5}
  \end{align}
  On the other hand
  \begin{align}
    \varphi_{\uparrow}(\eta)
    \leq
    \varphi_{\uparrow}(\eta_0)
    \leq
    & 
    \frac{\sigma^2}{2\mu M K R} \log \left(  \euler + \frac{\mu M K R \Psi^{(0,0)}}{\sigma^2} \right) 
    +
    \frac{\sigma^2}{2 \mu M K R^2} \log^2 \left(  \euler + \frac{\mu M K R \Psi^{(0,0)}}{\sigma^2} \right) 
    \nonumber \\
    & + \frac{390  L \sigma^2}{\mu^2 K R^3} \log^3 \left( \euler + \frac{\mu^2 K R^3 \Psi^{(0,0)}}{L \sigma^2} \right)
    + \frac{7 L \sigma^2}{\mu^2 K R^4} \log^4 \left( \euler + \frac{\mu^2 K R^3 \Psi^{(0,0)}}{L \sigma^2} \right)
    \nonumber  \\
    \leq & \frac{\sigma^2}{\mu M K R} \log^2 \left(  \euler + \frac{\mu M K R \Psi^{(0,0)}}{\sigma^2} \right) 
    + \frac{397 L \sigma^2}{\mu^2 K R^3} \log^4 \left( \euler + \frac{\mu^2 K R^3 \Psi^{(0,0)}}{L \sigma^2} \right).
    \label{eq:fedaci:proof:6}
  \end{align}
  Combining \cref{fedaci:general:eta,eq:fedaci:proof:4,eq:fedaci:proof:5,eq:fedaci:proof:6} gives
  \begin{align*}
    & \expt [\Psi^{(r,k)}]
    \leq 
    \varphi_{\downarrow} \left( \frac{1}{L} \right) + \varphi_{\downarrow}(\eta_0) + \varphi_{\uparrow}(\eta) 
    \\
    \leq & 
    \min \left\{ \exp \left( - \frac{\mu K R}{L} \right), \exp \left( - \frac{\mu^{\frac{1}{2}} K^{\frac{1}{2}} R}{L^{\frac{1}{2}}} \right)\right\}\Psi^{(0,0)}
    + \frac{2\sigma^2}{\mu M K R} \log^2 \left(  \euler + \frac{\mu M K R \Psi^{(0,0)}}{\sigma^2} \right) 
    \\
    & + \frac{400 L \sigma^2}{\mu^2 K R^3} \log^4 \left( \euler + \frac{\mu^2 K R^3 \Psi^{(0,0)}}{L \sigma^2} \right),
  \end{align*}
  completing the proof of main \cref{fedaci:full}.
\end{proof}

% \section{Proof sketch}
% In this section we sketch the proof for two of our main results, namely \cref{fedac:a1}(a) and \ref{fedacii:a2}.
% We focus on the proof of \cref{fedac:a1}(a) to outline our proof framework, and then illustrate the difference in the proof of \cref{fedacii:a2}.
\section{Proof Sketch of \cref{fedacii:a2}}
\label{sec:proof:sketch:2}
In this section, we outline the proof of \cref{fedacii:a2} by contrasting the differences from the proof in \cref{sec:proof:sketch:1}.
The first difference is that for \fedacii we study an alternative \emph{centralized potential}
\begin{equation}
  \Phi^{(r,k)} := F(\overline{\x^{(r,k)}_{\mathrm{ag}}}) - F^{\star} + \frac{1}{6}\mu \| \overline{\x^{(r,k)}} - \x^{\star}\|_2^2 
  \label{eq:centralied:potential}
\end{equation}
which leads to an alternative version of \cref{fedaci:conv:main} as follows.

\begin{lemma}[name={Potential-based perturbed iterate analysis for \fedacii},label=fedacii:conv:main,restate=FedAcIIConvMain]
  Let $F$ be $\mu>0$-strongly convex, and assume \cref{asm:fo:scvx:2o}, then for $\alpha = \frac{3}{2 \gamma \mu} - \frac{1}{2}$, $\beta =\frac{2 \alpha^2 - 1}{\alpha - 1}$,
  $\gamma \in \left[\eta, \sqrt{ \frac{\eta}{\mu}}\right]$, $\eta \in (0, \frac{1}{L}]$, \fedac yields
  \begin{align*}
    & \expt [\Phi^{(R,0)}] \\
    \leq & \exp \left( - \frac{1}{3}\gamma \mu K R \right) \Phi^{(0,0)}
    + \frac{3\eta^2 L \sigma^2 }{2 \gamma \mu M}
    + \frac{\gamma \sigma^2}{2 M}
    + \frac{3}{\mu} \max_{\substack{0 \leq r < R \\ 0 \leq k < K}}   \expt \left[ \left\|  \nabla F(\overline{\x^{(r,k)}_{\mathrm{md}}}) -  \frac{1}{M} \sum_{m=1}^M \nabla F (\x^{(r,k)}_{\mathrm{md}, m})  \right\|_2^2\right],
  \end{align*}
where $\Phi^{(r,k)}$ is the decentralized potential defined in \cref{eq:centralied:potential}.
\end{lemma}
  % \begin{equation}
  %   \expt [\Phi^{(R,0)}] \leq \exp \left( - \frac{\gamma \mu K R}{3} \right) \Phi^{(0,0)}
  %   + \frac{3\eta^2 L \sigma^2 }{2 \gamma \mu M}
  %   + \frac{\gamma \sigma^2}{2 M}
  %   + \frac{3}{\mu} \max_{\substack{0 \leq r < R \\ 0 \leq k < K}}   \expt \left\| \frac{1}{M} \sum_{m=1}^M \nabla F (\x^{(r,k)}_{\mathrm{md}, m}) - \nabla F(\overline{\x^{(r,k)}_{\mathrm{md}}})    \right\|_2^2.
  %   \label{eq:fedacii:conv:sketch}
  % \end{equation}
The second difference is that the particular discrepancy in \cref{fedacii:conv:main} can be bounded via \nth{3}-order smoothness $Q$ since
\begin{align}
  & \left\|  \frac{1}{M} \sum_{m=1}^M \nabla F(\x_m) - \nabla F (\overline{\x})    \right\|_2^2= \left\|  \frac{1}{M} \sum_{m=1}^M \left( \nabla F(\x_m) - \nabla F (\overline{\x}) - \nabla^2 F (\overline{\x})(\x_m - \overline{\x}) \right)   \right\|_2^2 
\\
\leq & \frac{1}{M} \sum_{m=1}^M\left\|   \nabla F(\x_m) - \nabla F (\overline{\x}) - \nabla^2 F (\overline{\x})(\x_m - \overline{\x}) \right\|_2^2  \leq \frac{Q^2}{4M}  \sum_{m=1}^M\left\| \x_m - \overline{\x} \right\|_2^4.
\end{align}
This results in the following lemma.
\begin{lemma}[name={Discrepancy overhead bounds},label=fedacii:stab:main,restate=FedAcIIStabMain]
  Let $F$ be $\mu>0$-strongly convex, and assume \cref{asm:fo:scvx:3o}, then for the same hyperparameter choice as in \cref{fedacii:conv:main}, \fedac satisfies (for all $r, k$)
  \begin{align*}
     & \expt \left[ \left\|  \nabla F(\overline{\x^{(r,k)}_{\mathrm{md}}}) -  \frac{1}{M} \sum_{m=1}^M \nabla F (\x^{(r,k)}_{\mathrm{md}, m})  \right\|_2^2\right]
    \leq  \begin{cases}
      44 \eta^4 Q^2 K^2 \sigma^4 \left(1 + \frac{\gamma^2\mu}{\eta}\right)^{4K}
       & \text{if~} \gamma \in \left(\eta, \sqrt{\frac{\eta}{\mu}} \right],
      \\
      44 \eta^4 Q^2 K^2 \sigma^4
       &
      \text{if~} \gamma = \eta.
    \end{cases}
  \end{align*}
\end{lemma}
We relegate the remaining proof details and formal statement to \cref{sec:fedacii}.

\section{The Challenge: Instability of Standard Accelerated Gradient Descent}
\label{sec:instability}
In this section, we revisit the difficulty of \fedac caused by the instability of the \agd as discussed in the previous subsection.
We will show that standard accelerated gradient descent \cite{Nesterov-18} may not be initial-value stable even for strongly convex and smooth objectives in the sense that the initial infinitesimal difference may grow exponentially fast. 
This provides evidence on the necessity of acceleration-stability tradeoff.

We formally define the standard deterministic AGD in \cref{algo:agd} for $L$-smooth and $\mu$-strongly-convex objective $F$ \cite{Nesterov-18}.
\begin{algorithm}
    \caption{Nesterov's Accelerated Gradient Descent Method (\agd)}
    \begin{algorithmic}[1]
        \label{algo:agd}
        \STATE{\textbf{procedure}} \agd ($\x^{(0)}, \x^{(0)}_{\mathrm{ag}}; L, \mu$)
        \STATE $\kappa \gets L/\mu$
        \FOR{$t = 0, \ldots, T-1$}
        \STATE $\x^{(t)}_{\mathrm{md}} \gets \frac{1}{\sqrt{\kappa} + 1} \x^{(t)} + \frac{\sqrt{\kappa}}{\sqrt{\kappa} + 1} \x^{(t)}_{\mathrm{ag}} $
        \STATE $\x^{(t+1)}_{\mathrm{ag}} \gets \x^{(t)}_{\mathrm{md}} - \frac{1}{L} \cdot \nabla F(\x^{(t)}_{\mathrm{md}})$
        \STATE $\x^{(t+1)} \gets \left( 1 - \frac{1}{\sqrt{\kappa}} \right) \x^{(t)} + \frac{1}{\sqrt{\kappa}} \x^{(t)}_{\mathrm{md}} - \sqrt{\frac{1}{L \mu}} \nabla F(\x^{(t)}_{\mathrm{md}})$
        \ENDFOR
    \end{algorithmic}
\end{algorithm}

We restate the instability theorem below for the reader's reference.
\instability*

We first introduce the supporting lemmas for \cref{thm:agd:instability}. \cref{instability:1} shows the existence of an objective $F$ and a trajectory of \agd on $F$ such that $F''(w^{(t)}_{\mathrm{md}})=L$ (including also the neighborhood) once every three steps and $F''(w^{(t)}_{\mathrm{md}})=\mu$ otherwise. 
% The proof of \cref{instability:1} is deferred to \cref{sec:instability:1}.
\begin{lemma}
    \label{instability:1}
    For any $L > \mu > 0$, and for any $K \geq 1$, there exists a 1D objective $F$ that is $L$-smooth and $\mu$-strongly convex, a neighborhood bound $\delta > 0$, and initial points $w^{(0)}$ and $w^{(0)}_{\mathrm{ag}}$ such that the sequence $\{w^{(t)}_{\mathrm{ag}}, w^{(t)}_{\mathrm{md}}, w^{(t)}\}_{t=0}^{3K-1}$ output by $\agd(w^{(0)}_{\mathrm{ag}}, w^{(0)}, L, \mu)$ satisfies for any $t = 0, \ldots, 3K-1$,
    \begin{align*}
        \text{ if }  t \mathrm{~mod~} 3 \neq 1, \text{ then } F''(w) \equiv \mu \text{, for all } w \in [w^{(t)}_{\mathrm{md}}-\delta, w^{(t)}_{\mathrm{md}} + \delta],
        \\
        \text{ if }  t \mathrm{~mod~} 3 = 1, \text{ then } F''(w) \equiv L \text{, for all } w \in [w^{(t)}_{\mathrm{md}}-\delta, w^{(t)}_{\mathrm{md}} + \delta].
    \end{align*}
\end{lemma}

% \subsection{Proof of \cref{instability:1}}
% \label{sec:instability:1}
% In this section we prove \cref{instability:1} on the existence of objective $F$ and the trajectory with specific curvature at certain intervals. 
The high-level rationale is that \cref{instability:1} only specifies local curvatures of $F$, and therefore we can modify an objective at certain local points to make \cref{instability:1} satisfied.
Here we provide a constructive approach by incrementally updating $F$. 

We inductively prove the following claim.
    \begin{claim}
        \label{instability:induction}
        For any $k = 0, \ldots, K$, there exists a function $H_k$ valued in $[\mu, L]$, 
        a neighborhood bound $\delta_k > 0$,  and a pair of initial points $(w^{(0)}_{\mathrm{ag}}, w^{(0)})$, such that for objective
        $F_k(w) := \int_0^{w} \int_0^{y} H_k(x) \diff x \diff y$,  the sequence output by \agd($w^{(0)}_{\mathrm{ag}}, w^{(0)}, L, \mu$) on $F_k$ satisfies $|w^{(t_1)}_{\mathrm{md}} - w^{(t_2)}_{\mathrm{md}}| \geq 2\delta_k$ if $t_1 \neq t_2$, and  for any $t = 0, \ldots, 3K - 1$,
        \begin{align}
            & \text{ if }  t \mathrm{~mod~} 3 \neq 1 \text{ or } t \geq 3k, \text{ then } F''(w) \equiv H_k(w) \equiv \mu \text{ for all } w \in [w^{(t)}_{\mathrm{md}}-\delta_k, w^{(t)}_{\mathrm{md}} + \delta_k];
            \label{eq:curvature:1}
            \\
            & \text{ if }  t \mathrm{~mod~} 3 = 1 \text{ and } t < 3k, \text{ then } F''(w) \equiv H_k(w) \equiv L \text{ for all } w \in [w^{(t)}_{\mathrm{md}}-\delta_k, w^{(t)}_{\mathrm{md}} + \delta_k].
            \label{eq:curvature:2}
        \end{align}
    \end{claim}
    To simplify the notation, we refer to \cref{eq:curvature:1,eq:curvature:2} as ``curvature conditions'' and denote $\mathcal{U}(x;r) := \{y: |y - x| < r\}$, and $\bar{\mathcal{U}}(x;r) := \{y: |y - x| \leq r\}$.
    \begin{proof}[Inductive proof of \cref{instability:induction}]
        For $k = 0$, we can put $H_0(w) \equiv \mu$ (then $F_k(w) = \frac{1}{2}\mu w^2$) and select any arbitrary initial points $(w^{(0)}_{\mathrm{ag}}, w^{(0)})$ as long as $w^{(t_1)}_{\mathrm{md}} \neq w^{(t_2)}_{\mathrm{md}}$ for $t_1 \neq t_2$, which is trivially possible.
        
        Suppose \cref{instability:induction} holds for $k$,     
        now we construct $H_{k+1}$ and $\delta_{k+1}$.  
        Let $\{{w}^{(t)}_{\mathrm{ag}, k}, {w}^{(t)}_{\mathrm{md}, k}, {w}^{(t)}_{k}\}_{t=0}^{3K-1}$ be the trajectory output by \agd ($w^{(0)}_{\mathrm{ag}}, w^{(0)}, L, \mu$) on $F_k$.
        For some positive $\varepsilon_{k} < \frac{1}{2}\delta_k$ to be determined, consider
        \begin{equation*}
            \tilde{H}_{k+1}(w) = H_{k}(w) + (L - \mu) \mathbf{1} \left[ w \in \bar{\mathcal{U}}(w^{(3k+1)}_{\mathrm{md}, k}; \varepsilon_k) \right],
            \quad
            \tilde{F}_{k+1}(w) = \int_0^w \int_0^y \tilde{H}_{k+1}(x) \diff x \diff y.
        \end{equation*}
        Let $\{\tilde{w}^{(t)}_{\mathrm{ag}, k+1}, \tilde{w}^{(t)}_{\mathrm{md}, k+1}, \tilde{w}^{(t)}_{k+1}\}_{t=0}^{3K-1}$
        be  the trajectory output by \agd($w^{(0)}_{\mathrm{ag}}, w^{(0)}, L, \mu$) on $\tilde{F}_{k+1}$. 
        Since the trajectory is continuous with respect to $\varepsilon_k$, 
        there exists a $\bar{\varepsilon} < \frac{1}{2} \delta_k$ such that for any $\varepsilon_k < \bar{\varepsilon}$ (which we assume from now on),  it is the case that $|\tilde{w}^{(t)}_{\mathrm{md}, k+1} - w^{(t)}_{\mathrm{md}, k}| \leq \frac{1}{2}\delta_k$ for all $t \leq 3k+1$.
        Then let
        \begin{equation*}
            {H}_{k+1}(w) = H_{k}(w) + (L - \mu) \mathbf{1} \left[ w \in  \bar{\mathcal{U}}(\tilde{w}^{(3k+1)}_{\mathrm{md}, k+1}; \varepsilon_k) \right],
            \quad
            {F}_{k+1}(w) = \int_0^w \int_0^y {H}_{k+1}(x) \diff x \diff y.
        \end{equation*}   
        and let $\{{w}^{(t)}_{\mathrm{ag}, k+1}, {w}^{(t)}_{\mathrm{md}, k+1}, {w}^{(t)}_{k+1}\}_{t=0}^{3K-1}$
        be  the trajectory output by \agd($w^{(0)}_{\mathrm{ag}}, w^{(0)}, L, \mu$) on ${F}_{k+1}$. 

        Consequently,
        \begin{enumerate}
            \item[(a)] By construction of $H_{k+1}$ and $\tilde{H}_{k+1}$, we have $H_{k+1}(w) = \tilde{H}_{k+1}(w) = H_k(w)$ and $\nabla F_{k+1}(w) = \nabla \tilde{F}_{k+1}(w)$ for all $w \notin \bar{U}(w^{(3k+1)}_{\mathrm{md}, k}; \delta_k)$. 
            \item[(b)] Since $\tilde{w}^{(t)}_{\mathrm{md}, k+1} \notin \bar{U}(w^{(3k+1)}_{\mathrm{md}, k}; \delta_k)$, by (a), we can inductively show that $\tilde{w}^{(t)}_{\mathrm{md}, k+1} = w^{(t)}_{\mathrm{md}, k+1}$ for $t < 3k+1$, namely the trajectories for $F_{k+1}$ and $\tilde{F}_{k+1}$ are identical up to timestep $t < 3k+1$. 
            \item[(c)] Since $|\tilde{w}^{(t)}_{\mathrm{md}, k+1} - w^{(t)}_{\mathrm{md}, k}| \leq \frac{1}{2}\delta_k$, by (b), we further have $|w^{(t)}_{\mathrm{md}, k+1} - w^{(t)}_{\mathrm{md}, k}| \leq \frac{1}{2} \delta_k$ for $t < 3k + 1$. 
            Thus, by (a), the curvature conditions will be satisfied for $w^{(t)}_{\mathrm{md}, k+1}$ and $H_{k+1}$ up to $t < 3k+1$ and any neighborhood bound $\delta_{k+1} < \frac{1}{2} \delta_k$ since $H_{k+1} \equiv H_k$ for $w \notin \bar{U}(w^{(3k+1)}_{\mathrm{md}, k}; \delta_k)$. 
            \item[(d)] By (b), we have $w^{(3k+1)}_{\mathrm{md}, k+1} = \tilde{w}^{(3k+1)}_{\mathrm{md}, k+1}$ since all previous gradients evaluated are identical for $F_{k+1}$ and $\tilde{F}_{k+1}$. Thus, by construction of $H_{k+1}$ the curvature conditions hold for $w^{(3k+1)}_{\mathrm{md}, k+1}$ and $H_{k+1}$.
            \item[(e)] Similarly, for sufficiently small $\varepsilon_k$, we have $|w^{(t)}_{\mathrm{md}, k+1} - w^{(t)}_{\mathrm{md}, k}| \leq \frac{1}{2} \delta_k$ for $t > 3k+1$, and the curvature conditions also hold for $t > 3k+1$.
        \end{enumerate}
        Summarizing (c), (d), and (e) completes the induction.
    \end{proof}
\begin{proof}[Proof of \cref{instability:1}]
    Follows by applying \cref{instability:induction}.
\end{proof}

The following \cref{instability:2} analyzes the growth of the difference of two instances of \agd. The proof is very similar to the analysis of \fedac.
\begin{lemma}
    \label{instability:2}
    Let $F$ be a $L$-smooth and $\mu>0$-strongly convex 1D function.
    Let $(w^{(t+1)}_{\mathrm{ag}}, w^{(t+1)})$, $(u^{(t+1)}_{\mathrm{ag}}, u^{(t+1)})$ be generated by applying one step of \agd on $F$ with hyperparameter $(L, \mu)$ from $(w^{(t)}_{\mathrm{ag}}, w^{(t)})$ and $(u^{(t)}_{\mathrm{ag}}, u^{(t)})$, respectively. Then there exists a $z^{(t)}$ within the interval between $w^{(t)}_{\mathrm{md}}$ and $u^{(t)}_{\mathrm{md}}$, such that
    \begin{equation*}
        \begin{bmatrix}
            w^{(t+1)}_{\mathrm{ag}} - u^{(t+1)}_{\mathrm{ag}}
            \\
            w^{(t+1)} - u^{(t+1)}
        \end{bmatrix}
        =
        \begin{bmatrix}
            \frac{\sqrt{\kappa}}{\sqrt{\kappa} + 1} \left(1- \frac{1}{L} F''(z^{(t)}) \right) 
            &
            \frac{1}{\sqrt{\kappa} + 1} \left(1- \frac{1}{L} F''(z^{(t)}) \right) 
            \\
            \frac{1}{\sqrt{\kappa} + 1} \left(1- \frac{1}{\mu} F''(z^{(t)}) \right) 
            &
            \frac{\sqrt{\kappa}}{\sqrt{\kappa} + 1} \left(1- \frac{1}{L} F''(z^{(t)}) \right) 
        \end{bmatrix}
        \begin{bmatrix}
            w^{(t)}_{\mathrm{ag}} - u^{(t)}_{\mathrm{ag}}
            \\
            w^{(t)} - u^{(t)}
        \end{bmatrix}.
    \end{equation*}
\end{lemma}
\begin{proof}[Proof of \cref{instability:2}]
    This is a special case of \cref{fedac:general:stab} with no noise. 
\end{proof}

With \cref{instability:1,instability:2} at hand we are ready to prove \cref{thm:agd:instability}. The proof follows by constructing an auxiliary trajectory for around the one given by \cref{instability:1}.
\begin{proof}[Proof of \cref{thm:agd:instability}]
    First apply  \cref{instability:1}. 
    Let $F$ be the objective, $(w^{(0)}_{\mathrm{ag}}, w^{(0)})$ be the initial point and $\delta$ be the neighborhood bound given by  \cref{instability:1}.
    Since $\{w^{(t)}_{\mathrm{ag}}, w^{(t)}_{\mathrm{md}}, w^{(t)}\}_{t=0}^{3K-1}$ is a continuous function with respect to the initial point $(w^{(0)}_{\mathrm{ag}}, w^{(0)})$, there exists a $\varepsilon_0$ such that for any $(v^{(0)}_{\mathrm{ag}}, v^{(0)})$ such that $|v^{(0)}_{\mathrm{ag}} - w^{(0)}_{\mathrm{ag}}| \leq \varepsilon_0$ and $|v^{(0)} - w^{(0)}| \leq \varepsilon_0$, trajectory $\{v^{(t)}_{\mathrm{ag}}, v^{(t)}_{\mathrm{md}}, v^{(t)}\}_{t=0}^{3K}$ output by \agd $(v^{(0)}_{\mathrm{ag}}, v^{(0)}, L, \mu)$ satisfies $\max_{0 \leq t < 3K} |v^{(t)}_{\mathrm{md}} - w^{(t)}_{\mathrm{md}}| \leq \delta$. 
    
    Thus, by \cref{instability:2}, for any $t = 0, \ldots, 3K -1$,
    \begin{alignat*}{2}
        \begin{bmatrix}
            w^{(t+1)}_{\mathrm{ag}} - v^{(t+1)}_{\mathrm{ag}}
            \\
            w^{(t+1)} - v^{(t+1)}
        \end{bmatrix}
        & =
        \begin{bmatrix}
            1 - \frac{1}{\sqrt{\kappa}}
            &
            \frac{1}{\kappa}(\sqrt{\kappa} - 1) 
            \\
            0
            &
            1 - \frac{1}{\sqrt{\kappa}}
        \end{bmatrix}
        \begin{bmatrix}
            w^{(t)}_{\mathrm{ag}} - v^{(t)}_{\mathrm{ag}}
            \\
            w^{(t)} - v^{(t)}
        \end{bmatrix},
        \quad
        &&
        \text{if } t \mathrm{~mod~} 3 \neq 1;
        \\
        \begin{bmatrix}
            w^{(t+1)}_{\mathrm{ag}} - v^{(t+1)}_{\mathrm{ag}}
            \\
            w^{(t+1)} - v^{(t+1)}
        \end{bmatrix}
        & =
        \begin{bmatrix}
            0
            &
            0
            \\
            1 - \sqrt{\kappa}
            &
            0
        \end{bmatrix}
        \begin{bmatrix}
            w^{(t)}_{\mathrm{ag}} - v^{(t)}_{\mathrm{ag}}
            \\
            w^{(t)} - v^{(t)}
        \end{bmatrix},
        \quad
        &&
        \text{if } t \mathrm{~mod~} 3 = 1.
    \end{alignat*}
    Hence for any $k = 0, \ldots, K-1$,
    \begin{align*}
        \begin{bmatrix}
            w^{(3k+3)}_{\mathrm{ag}} - v^{(3k+3)}_{\mathrm{ag}}
            \\
            w^{(3k+3)} - v^{(3k+3)}
        \end{bmatrix}
        & =
        -
        \begin{bmatrix}
            \frac{1}{\kappa^{\frac{3}{2}}}(\sqrt{\kappa} - 1)^3 
            & 
            \frac{1}{\kappa^2} (\sqrt{\kappa} - 1)^3 
            \\
            \frac{1}{\kappa}  (\sqrt{\kappa} - 1)^3 
            &
            \frac{1}{\kappa^\frac{3}{2}} (\sqrt{\kappa} - 1)^3 
        \end{bmatrix}
        \begin{bmatrix}
            w^{(3k)}_{\mathrm{ag}} - v^{(3k)}_{\mathrm{ag}}
            \\
            w^{(3k)} - v^{(3k)}
        \end{bmatrix}
        \\
        & = 
        - 2 \left( 1 - \frac{1}{\sqrt{\kappa}} \right)^3
        \begin{bmatrix}
            \frac{1}{2} & \frac{1}{2\sqrt{\kappa}}
            \\
            \frac{1}{2}\sqrt{\kappa} & \frac{1}{2}
        \end{bmatrix}
        \begin{bmatrix}
            w^{(3k)}_{\mathrm{ag}} - v^{(3k)}_{\mathrm{ag}}
            \\
            w^{(3k)} - v^{(3k)}
        \end{bmatrix}.
    \end{align*}
    Note that
    \(
        \begin{bmatrix}
            \frac{1}{2} & \frac{1}{2\sqrt{\kappa}}
            \\
            \frac{1}{2}\sqrt{\kappa} & \frac{1}{2}
        \end{bmatrix}
    \)
    is idempotent, \ie,
    \(
        \begin{bmatrix}
            \frac{1}{2} & \frac{1}{2\sqrt{\kappa}}
            \\
            \frac{1}{2}\sqrt{\kappa} & \frac{1}{2}
        \end{bmatrix}^{K}
        =
        \begin{bmatrix}
            \frac{1}{2} & \frac{1}{2\sqrt{\kappa}}
            \\
            \frac{1}{2}\sqrt{\kappa} & \frac{1}{2}
        \end{bmatrix}.
    \)
    Thus
    \begin{equation*}
        \begin{bmatrix}
            w^{(3K)}_{\mathrm{ag}} - v^{(3K)}_{\mathrm{ag}}
            \\
            w^{(3K)} - v^{(3K)}
        \end{bmatrix}
        =
        \left( -2 \left( 1 - \frac{1}{\sqrt{\kappa}} \right)^3 \right)^K
        \begin{bmatrix}
            \frac{1}{2} & \frac{1}{2\sqrt{\kappa}}
            \\
            \frac{1}{2}\sqrt{\kappa} & \frac{1}{2}
        \end{bmatrix}
        \begin{bmatrix}
            w^{(0)}_{\mathrm{ag}} - v^{(0)}_{\mathrm{ag}}
            \\
            w^{(0)} - v^{(0)}
        \end{bmatrix}.
    \end{equation*}
    Thus for any given $\varepsilon \leq \varepsilon_0$, put $u^{(0)}_{\mathrm{ag}} = w^{(0)}_{\mathrm{ag}} - \varepsilon$, and $u^{(0)} = w^{(0)} - \varepsilon$, we have
    \begin{align*}
        \begin{bmatrix}
            w^{(3K)}_{\mathrm{ag}} - u^{(3K)}_{\mathrm{ag}}
            \\
            w^{(3K)} - u^{(3K)}
        \end{bmatrix}
        =
        \frac{1}{2} \varepsilon \left( -2 \left( 1 - \frac{1}{\sqrt{\kappa}} \right)^3 \right)^K
        \begin{bmatrix}
            1 + \frac{1}{\sqrt{\kappa}}
            \\
            \sqrt{\kappa} + 1
        \end{bmatrix}.
    \end{align*}
    For $\kappa \geq 25$ we have $ \left|2 \left( 1 - \frac{1}{\sqrt{\kappa}} \right)^3 \right| > 1.02 $. 
    Therefore,
    \begin{equation*}
        | w^{(3K)}_{\mathrm{ag}} - u^{(3K)}_{\mathrm{ag}}| 
        \geq
        \frac{1}{2} (1.02)^K \cdot \varepsilon,
        \quad
        | w^{(3K)} - u^{(3K)}| 
        \geq
        (1.02)^K \cdot \varepsilon,
    \end{equation*}
    completing the proof.
\end{proof}
% As a sanity check, the proof framework above for instability does not apply to the convergence of \agd. For instability, we only need to locally change the curvature to ``separate'' two instances. This trick does not break the convergence proof where the progress depends on the global curvature. We refer readers to \cite{Lessard.Recht.ea-SIOPT16} for the relative discussion.

% !TEX root = main.tex
\section{Numerical Experiments}
\label{sec:experiment}
In this section, we validate our theory and demonstrate the efficiency of \fedac via numerical experiments.
The source code is available at \url{https://bit.ly/fedac-neurips20}.
\subsection{General Setup}
\paragraph{Baselines.}
The performance of \fedac is tested against three baselines: \fedavg (a.k.a., Local SGD), (distributed) Minibatch-SGD (\mbsgd), and (distributed) Minibatch-Accelerated-SGD (\mbacsgd) \cite{Dekel.Gilad-Bachrach.ea-JMLR12,Cotter.Shamir.ea-NeurIPS11}.
We fix the product $KR$ to be 4096, and test variant levels of synchronization interval $K$ and parallel clients $M$.
\mbsgd and \mbacsgd baselines correspond to running SGD or accelerated SGD for $\nicefrac{T}{K}$ steps with batch size $MK$. 
The comparison is fair since all algorithms can be parallelized to $M$ clients with $\nicefrac{T}{K}$ rounds of communication where each client queries $T$ gradients in total. 
We simulate the parallelization with a \texttt{NumPy} program on a local CPU cluster. 
We start from the same random initialization for all algorithms under all settings. 

\paragraph{Datasets.}
The algorithms are tested on $\ell_2$-regularized logistic regression on the following two binary classification datasets from \texttt{LibSVM} \cite{Chang.Lin-11}. The preprocessing information and the download links can be found at \url{https://www.csie.ntu.edu.tw/~cjlin/libsvmtools/datasets/binary.html}. 
\begin{enumerate}[leftmargin=*]
    \item The  ``adult'' \texttt{a9a} dataset with 123 features and 32,561 training samples from the UCI Machine Learning Repository \cite{Dua.Graff-17}.
    \item The \texttt{epsilon} dataset with 2,000 features and 400,000 training samples from the PASCAL Challenge 2008 \cite{Sonnenburg.Franc.ea-08}.
\end{enumerate}

\paragraph{Evaluation.}
For all algorithms and all settings, we evaluate the suboptimality (regularized population loss) every $512$ parallel timesteps (gradient queries). 
We compute the suboptimality by comparing with a pre-computed optimum $F^{\star}$.
We record the best suboptimality attained over the evaluations.

\paragraph{Hyperparameter Choice.}
For all four algorithms, we tune the ``learning-rate'' hyperparameter $\eta$ only and record the best suboptimality attained. For \mbacsgd, the rest of hyperparameters are determined by the strong-convexity estimate $\mu$ which is taken to be the $\ell_2$-regularization strength $\lambda$. 
For \fedac, the default choice of hyperparameters $(\gamma, \alpha, \beta)$ is \fedaci \cref{fedaci}, where the strong-convexity estimate $\mu$ is also taken to be the $\ell_2$-regularization strength $\lambda$. \fedacii is qualitatively similar to \fedaci empirically so we show \fedaci only.

% The regularization strength is set as $10^{-3}$. 
% The hyperparameters $(\gamma, \alpha, \beta)$ of \fedac follows \fedaci 
% where strong-convexity $\mu$ is chosen as regularization strength $10^{-3}$.
% We test the settings of $M = 2^2, \ldots, 2^{13}$ clients and $K = 2^0, \ldots, 2^8$ synchronization interval.
% For all four algorithms, we tune the learning-rate $\eta$ \emph{only} from the same set of levels within $[10^{-3}, 10]$.
% We choose $\eta$ based on the best suboptimality (regularized population loss).
% We claim that the best $\eta$ lies in the range $[10^{-3}, 10]$ for all algorithms under all settings. 
% We defer the rest of setup details to \cref{additional:expr}.

\subsection{Experiments on Dataset \texttt{a9a}}
\begin{figure}[ht]
    \centering
    \includegraphics[width=\textwidth]{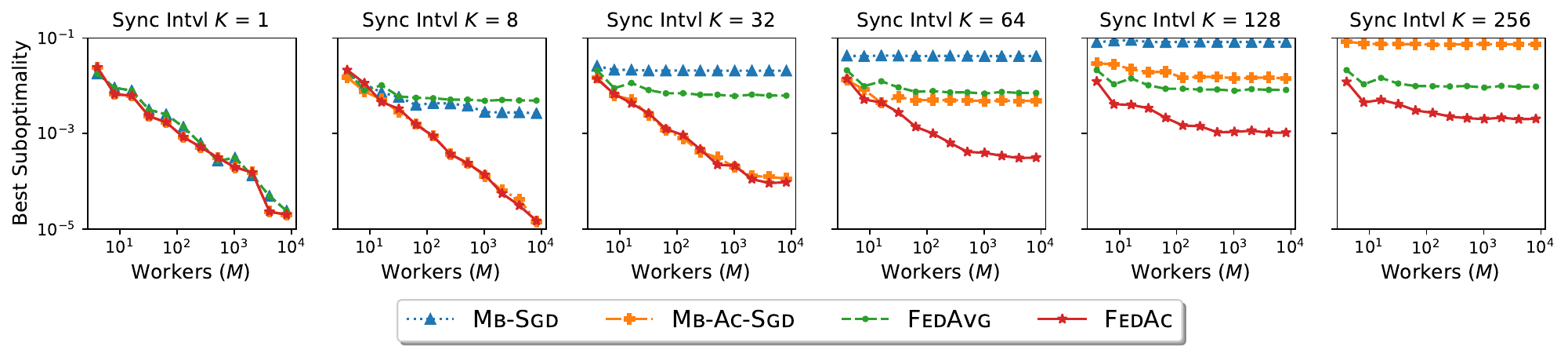}
    \caption{\textbf{Observed linear speedup with respect to the number of clients $M$ under various synchronization intervals $K$.} 
    Our \fedac is tested against three baselines \fedavg, \mbsgd, and \mbacsgd.
    While all four algorithms attain linear speedup for the fully synchronized ($K=1$) setting, \fedavg and \mbsgd lose linear speedup for $K$ as low as 8. 
    \mbacsgd is comparably better than the other two baselines but still deteriorates significantly for $K \geq 64$.
    \fedac is most robust to infrequent synchronization and outperforms the baselines by a margin for $K \geq 64$.
    }
    \label{fig:a9a:1e-3:M}
\end{figure}

We first test on the \texttt{a9a} dataset with  $\ell_2$-regularization strength $10^{-3}$.
We test the setting of $K = 2^0, \ldots, 2^8$ and $M = 2^2, \ldots, 2^{13}$.
For all algorithms, we tune $\eta$ from the same sets: \{0.001, 0.002, 0.005, 0.01, 0.02, 0.05, 0.1, 0.2, 0.5, 1, 2, 5, 10\}. We claim that the best $\eta$ lies in $[0.001, 10]$ for all algorithms for all settings.\footnote{We search for this range to guarantee that the optimal $\eta$ lies in this range for all algorithms and all settings. One could save effort in tuning if only one algorithm were implemented.}
In \cref{fig:a9a:1e-3:M}, we compare the algorithms by measuring the effect of linear speedup under variant $K$.

\begin{figure}[ht]
    \centering
    \includegraphics[width=\textwidth]{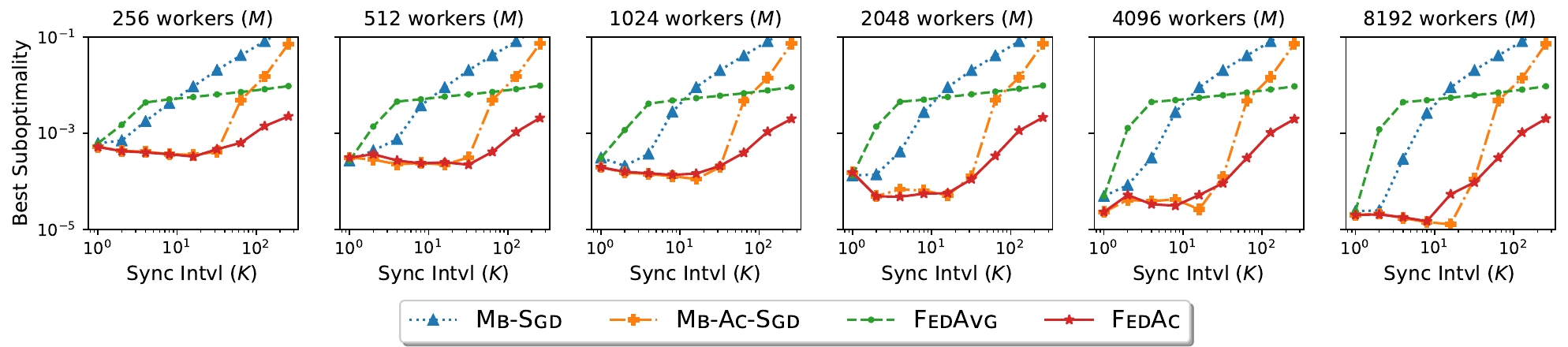}
    \caption{
        \textbf{\fedac versus baselines on the dependency of synchronization interval $K$ under various clients $M$.}
        For all tested $M$, \fedavg and \mbsgd start to deteriorate once $K$ passes $2$;
        \mbacsgd is more robust to moderate $K$ than  \fedavg and \mbsgd but sharply deteriorate once it passes a threshold at around $K=32$. 
        This is because \mbacsgd does not have enough gradient steps for convergence when the communication is too sparse. 
        In comparison, \fedac is more robust to infrequent communication.
        Dataset: \texttt{a9a}, $\ell_2$-regularization strength: $10^{-3}$.}
    \label{fig:a9a:1e-3:K}
\end{figure}

To better understand the dependency on synchronization intervals $K$, we plot the following \cref{fig:a9a:1e-3:K}. 
The results suggest that \fedac is more robust to infrequent synchronization and thus more communication-efficient. 
For example, when using 8192 clients, \fedac requries only 32 rounds of communication to attain $10^{-3}$ suboptimality, whereas \mbacsgd, \mbsgd and \fedavg require 128, 1024, 4096 rounds, respectively.

\begin{figure}[ht]
    \centering
    \includegraphics[width=\textwidth]{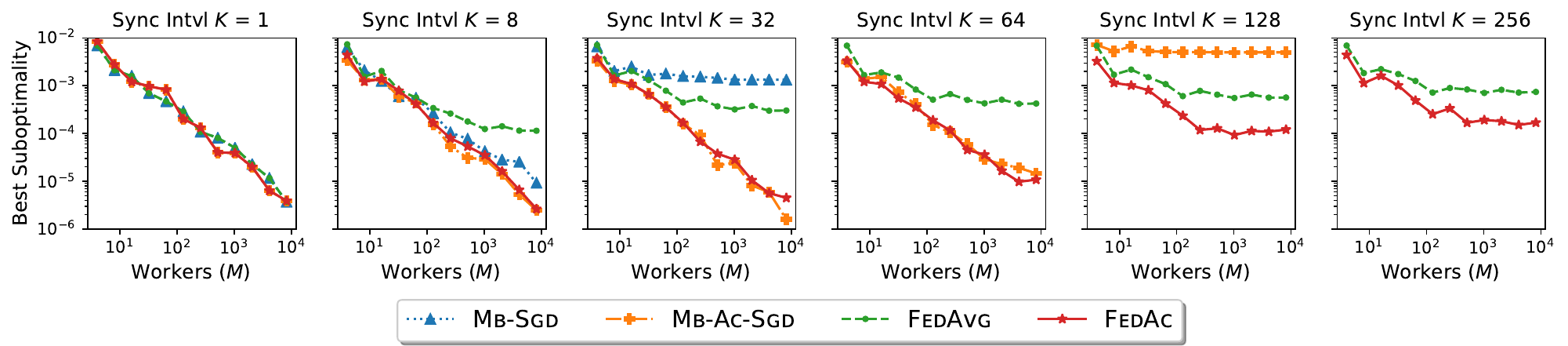}
    \caption{
        \textbf{\fedac versus baselines on the observed linear speedup w.r.t $M$ under various synchronization interval $K$.}
        The results are qualitatively similar to \cref{fig:a9a:1e-3:M}. 
        Dataset: \texttt{a9a}, $\ell_2$-regularization strength: $10^{-2}$.}
    \label{fig:a9a:1e-2:M}
\end{figure}

\begin{figure}[ht]
    \centering
    \includegraphics[width=\textwidth]{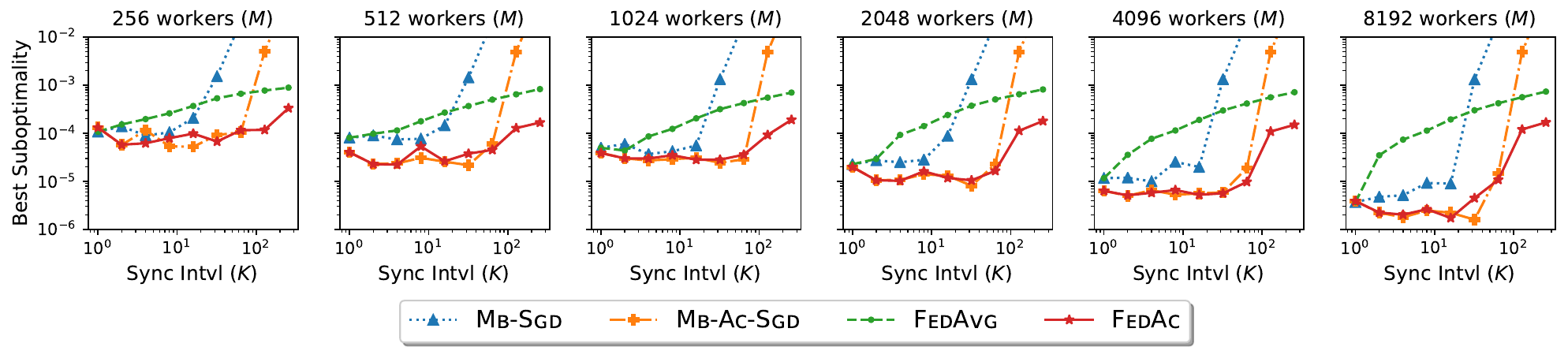}
    \caption{
        \textbf{\fedac versus baselines on the dependency of synchronization interval $K$ under various clients $M$.}
        The results are qualitatively similar to \cref{fig:a9a:1e-3:K}.
        Dataset: \texttt{a9a}, $\ell_2$-regularization strength: $10^{-2}$.}
    \label{fig:a9a:1e-2:K}
\end{figure}

We repeat the experiments with an alternative choice of $\lambda = 10^{-2}$. This problem is relatively ``easier'' in terms of optimization since the condition number $\nicefrac{L}{\mu}$ is lower. We test the same levels of $M$, $K$ and tune the $\eta$ from the same set as above. The results are shown in \cref{fig:a9a:1e-2:M,fig:a9a:1e-2:K}. The results are qualitatively similar to the $\lambda = 10^{-3}$ case. 
For $K \leq 64$, the performance of \fedac and \mbacsgd are similar, which both outperform the other two baselines \fedavg and \mbsgd. For $K \geq 128$, the \mbacsgd drastically worsen because the gradient steps are too few, and \fedac outperforms the other baselines by a margin.

\subsection{Vanilla \fedac Versus (Stable) \fedaci}
\begin{figure}[ht]
    \centering
    \includegraphics[width=\textwidth]{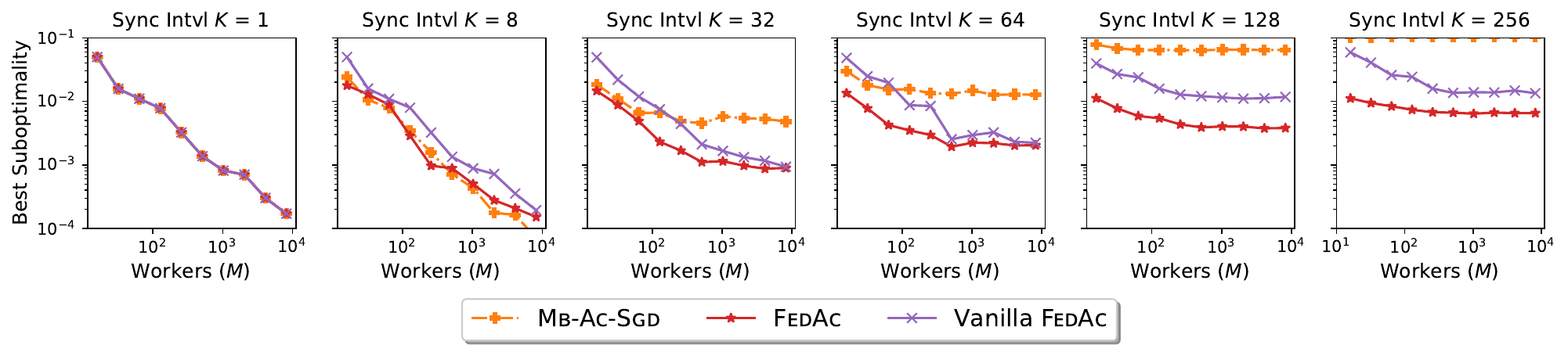}
    \caption{
        \textbf{Vanilla \fedac versus (stable) \fedaci and baseline \mbacsgd on the observed linear speedup w.r.t. $M$ under various synchronization intervals $K$.}
        Observe that Vanilla \fedac is indeed less robust to infrequent synchronization and thus worse than the \fedaci. (dataset: \textsc{a9a}, $\lambda=10^{-4}$)}
    \label{fig:instability}
\end{figure}

In the next experiments, we provide an empirical example to show that the direct parallelization of standard accelerated SGD may indeed suffer from instability. 
This complements our \cref{thm:agd:instability}) on the initial-value instability of standard AGD. 
Recall that \fedaci \cref{fedaci} and \fedacii \cref{fedacii} adopt an acceleration-stability tradeoff technique that takes $\gamma = \max \{ \sqrt{\frac{\eta}{\mu K}} , \eta\}$. Formally, we denote the following direct acceleration of \fedac without such tradeoff as ``vanilla \fedac'':
$
    \eta \in (0, \frac{1}{L} ], \gamma =  \sqrt{\frac{\eta}{\mu}} , \alpha  = \frac{1}{\gamma \mu},   \beta = \alpha + 1.
$
In \cref{fig:instability}, we compare the vanilla \fedac with the (stable) \fedaci and the baseline \mbacsgd.

\subsection{Experiments on Dataset \texttt{epsilon}}
\begin{figure}[ht]
    \centering
    \includegraphics[width=\textwidth]{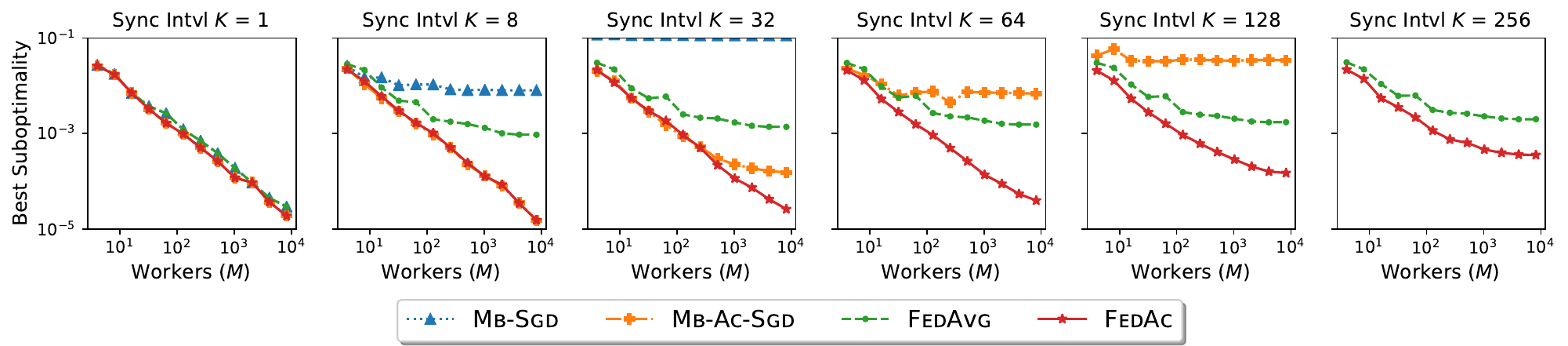}
    \caption{
        \textbf{\fedac versus baselines on the observed linear speedup w.r.t $M$ under various synchronization interval $K$.}
        The results are qualitatively similar to \cref{fig:a9a:1e-3:M}. 
        Dataset: \texttt{epsilon}, $\ell_2$-regularization strength: $10^{-4}$.}
    \label{fig:epsilon:1e-4:M}
\end{figure}

\begin{figure}[ht]
    \centering
    \includegraphics[width=\textwidth]{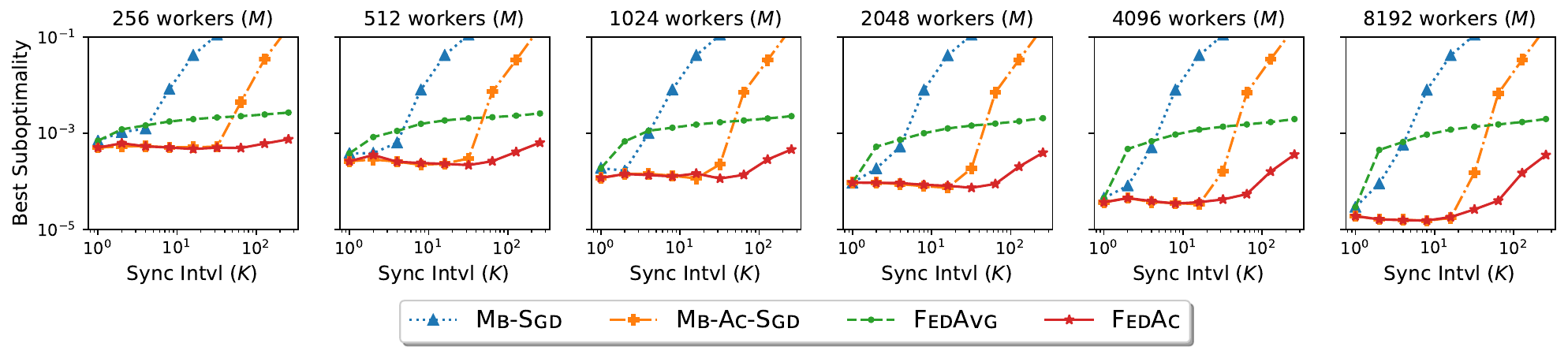}
    \caption{
        \textbf{\fedac versus baselines on the dependency of synchronization interval $K$ under various clients $M$.}
        The results are qualitatively similar to \cref{fig:a9a:1e-3:K}.
        Dataset: \texttt{epsilon}, $\ell_2$-regularization strength: $10^{-4}$.}
    \label{fig:epsilon:1e-4:K}
\end{figure}

In this section we repeat the experiments above on the larger \texttt{epsilon} dataset with $\ell_2$-regularization $\lambda$ taken to be $10^{-4}$. 
 $\eta$ is tuned from $\{0.005, 0.01, 0.02, 0.05, 0.1, 0.2, 0.5, 1, 2, 5, 10, 20, 50\}$.
The optimal $\eta$ lies in the corresponding range for all algorithm under all tested settings. The results are shown in \cref{fig:epsilon:1e-4:M,fig:epsilon:1e-4:K}. 
The results are qualitatively similar to the previous experiments on \texttt{a9a} dataset.
\fedac is more communication-efficient than the baselines. 
For example, when using 2048 clients, \fedac requires only 64 rounds of communication (synchronization) to attain $10^{-4}$ suboptimality, whereas \mbacsgd, \mbsgd and \fedavg require 256, 4096 and 4096 rounds of communication, respectively.

\chapter{Federated Composite Optimization}
\label{chapter:fco}
% !TEX root = main.tex  
% In previous chapters, we mostly focused on the smooth federated optimization problem \eqref{eq:fo:hetero}.

In this chapter, we propose to study the  \emph{Federated Composite Optimization} (FCO) problem. 
As in the previous chapters, the losses are distributed to $M$ clients. 
In addition, we assume all the clients share the same, possibly non-smooth, non-finite regularizer $\psi$. Formally, (FCO) is of the following form
\begin{equation}
\min_{w \in \reals^d} \Phi(\x) := F(\x) + \psi(\x) := \frac{1}{M} \sum_{m=1}^M F_m(\x) + \psi(\x), 
\label{FCO}
\end{equation}
where $F_m(\x) := \expt_{\xi \sim \mathcal{D}_m} f(w; \xi)$ is the loss at the $m$-th client, assuming $\mathcal{D}_m$ is its local data distribution. We assume that each client $m$ can access $\nabla f(\x; \xi)$ by drawing independent samples $\xi$ from its local distribution $\mathcal{D}_m$. 
Common examples of $\psi(\x)$ include $\ell_1$-regularizer or more broadly $\ell_p$-regularizer, nuclear-norm regularizer (for matrix variable), total variation (semi-)norm, etc. 
The (FCO) reduces to the standard federated optimization problem if $\psi \equiv 0$. 
The (FCO) also covers the constrained federated optimization if one takes $\psi$ to be the following constraint characteristics
$\chi_{\cstr}(\x) := %$ if $w \in \mathcal{C}$ or $+\infty$ otherwise.
  \begin{cases} 
      0 & \text{if $w \in \mathcal{C}$}, \\
      +\infty & \text{if $w \notin \mathcal{C}$}.
  \end{cases}
 $

\begin{figure}
    \centering
    \centerline{\includegraphics[width=\columnwidth]{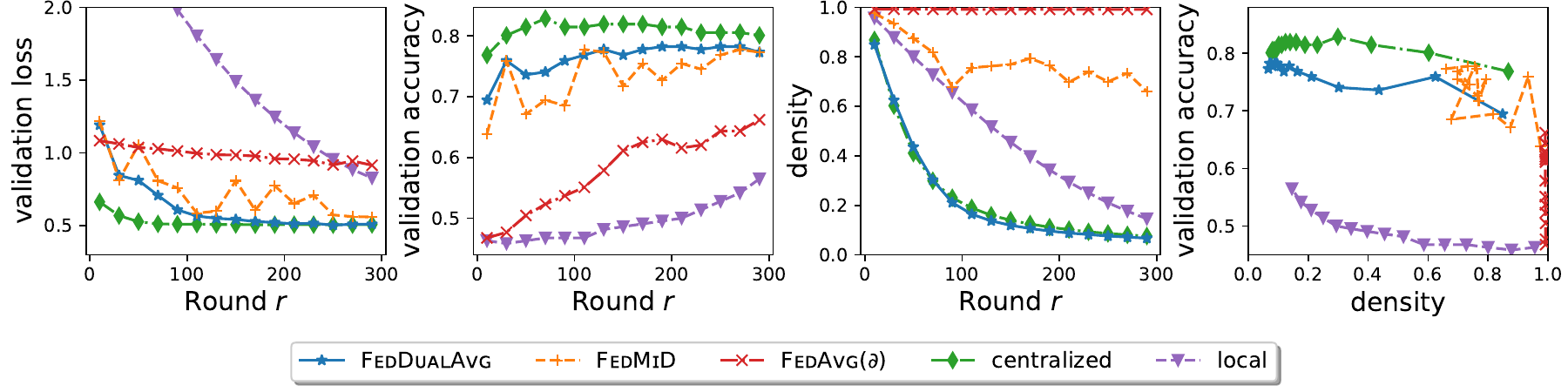}}
    \caption{
    \textbf{Results on sparse ($\ell_1$-regularized) logistic regression for a federated fMRI dataset based on  \cite{Haxby-01}.}
    \texttt{centralized} corresponds to training on the centralized dataset gathered from \textbf{all} the training clients.
    \texttt{local} corresponds to training on the local data from only \textbf{one} training client without communication. 
    \fedavg($\partial$) corresponds to running \fedavg algorithms with subgradient in lieu of SGD to handle the non-smooth $\ell_1$-regularizer. 
    \fedmid is another straightforward extension of \fedavg running local proximal gradient method (see \cref{sec:fedmid} for details). 
    We show that using our proposed algorithm \feddualavg, one can 1) achieve performance comparable to the \texttt{centralized} baseline without the need to gather client data, and 2) significantly outperforms the \texttt{local} baseline on the isolated data and the \fedavg baseline. 
    See \cref{sec:fco:expr:fmri} for details.}
    \label{fig:haxby:simplified}
  \end{figure}

For instance, consider the problem of cross-silo biomedical federated learning application, where medical organizations collaboratively aim to learn a global model on their patients' data without sharing. 
 In such applications, sparsity constraints are of paramount importance due to the nature of the problem as it involves only a few data samples (e.g., patients) but with very high dimensions (e.g., fMRI scans). 
For the purpose of illustration, in \cref{fig:haxby:simplified}, we present results on a federated sparse ($\ell_1$-regularized) logistic regression task for an fMRI dataset \cite{Haxby-01}.
As shown, using a federated approach that can handle non-smooth objectives enables us to find a highly accurate sparse solution without sharing client data.
  
% !TEX root = main.tex  
% \section{Additional Related Work}
% \label{sec:related_works}
% % Shortly after the initial preprint release of the present work, \cite{Tong.Liang.ea-20} proposed a related federated $\ell_0$-constrained problem (which does not belong to FCO due to the non-convexity of $\ell_0$), and two algorithms to solve (similar to \fedmid(\textsc{-OSP}) but with hard-thresholding instead). 
% % As in most hard-thresholding work, the convergence is weaker since it depends on the sparsity level $\tau$ (worsens as $\tau$ gets tighter).

% !TEX root = main.tex
\section{Preliminaries}
\label{sec:prelim}
In this section, we review the necessary background for composite optimization and federated learning. 
A detailed technical exposition of these topics is relegated to \cref{sec:background}.

\subsection{Composite Optimization}
\label{sec:co}
Composite optimization covers a variety of statistical inference, machine learning, signal processing problems.
Standard (non-distributed) composite optimization is defined as
\begin{equation}
    \min_{\x \in \reals^d} \quad \expt_{\xi \sim \mathcal{D}}f(\x; \xi) + \psi(\x),
    \label{eq:CO}
\end{equation}
where $\psi$ is a non-smooth, possibly non-finite regularizer.

\paragraph{Proximal Gradient Method.}
A natural extension of SGD for (CO) is the following \emph{proximal gradient method} (PGM):
\begin{align}
    \x^{(t+1)} \gets & \prox_{\eta \psi} \left(\x^{(t)} - \eta \nabla f(\x^{(t)};\xi^{(t)})  \right)
    \nonumber \\
    = & \argmin_{\x} \left( \eta \langle \nabla f(\x^{(t)};\xi^{(t)}) , \x \rangle + \frac{1}{2} \|\x - \x^{(t)}\|_2^2   + \eta \psi(\x)  \right). 
    \label{eq:pgm}
\end{align}
The sub-problem \cref{eq:pgm} can be motivated by optimizing a quadratic upper bound of $f$ together with the original $\psi$. 
This problem can often be efficiently solved by virtue of the special structure of $\psi$. 
For instance, one can verify that PGM reduces to projected gradient descent if $\psi$ is a constraint characteristic $ {\chi}_{\mathcal{C}}$, soft thresholding if $\psi(\x) = \lambda \|w\|_1$, or weight decay if $\psi(\x) := \lambda \|w\|_2^2$. 

\paragraph{Mirror Descent / Bregman-PGM.}
PGM can be generalized to the Bregman-PGM if one replaces the Euclidean proximity term by the general Bregman divergence, namely
\begin{equation}
    \x^{(t+1)} \gets \argmin_{\x} \left( \eta  \left\langle \nabla f(\x^{(t)}; \xi^{(t)}), \x \right \rangle + \eta \psi(\x) + D_{h}(\x, \x^{(t)})  \right),
     \label{eq:bpgm}
\end{equation}
where $h$ is a strongly convex distance-generating function, $D_h(\x, \y) = h(\x) - h(\y) - \langle \nabla h(\y), \x - \y \rangle$ is the Bregman divergence which reduces to Euclidean distance if one takes $h(\x) = \frac{1}{2} \|\x\|_2^2$. 
We will still refer to this step as a proximal step for ease of reference.
This general formulation \eqref{eq:bpgm} enables an equivalent primal-dual interpretation:
\begin{equation}
    \x^{(t+1)} \gets \nabla (h + \eta \psi)^*(\nabla h(\x^{(t)}) - \nabla f(\x^{(t)};\xi^{(t)})).
    \label{eq:md:oneline}
\end{equation}
A common interpretation of \eqref{eq:md:oneline} is to decompose it into the following three sub-steps \cite{Nemirovski.Yudin-83}: 
\begin{enumerate}[(a),leftmargin=*]
    \item Apply $\nabla h$ to carry $\x^{(t)}$ to a dual state (denoted as $\y^{(t)}$)
    \item Update $\y^{(t)}$ to $\y^{(t+1)}$ with the gradient queried at $\x^{(t)}$.
    \item Map $\y^{(t+1)}$ back to primal via $\nabla (h+ \eta\psi)^*$
\end{enumerate}
This formulation is known as the \emph{composite objective mirror descent} (\textsc{Comid}, \cite{Duchi.Shalev-shwartz.ea-COLT10}), or simply \emph{mirror descent} in the literature \cite{Flammarion.Bach-COLT17}. 
% We will refer to this step as ``proximal'' step or ``mirror descent'' step interchangeably hereinafter.

\paragraph{Dual Averaging.} 
An alternative approach for (CO) is the following \emph{dual averaging} algorithm \cite{Nesterov-MP09}:
\begin{equation}
    \y^{(t+1)} \gets \y^{(t)} - \eta \nabla f \left(\nabla (h + \eta t \psi)^*(\y^{(t)}); \xi^{(t)} \right).
    \label{eq:da}
\end{equation}
Similarly, we can decompose \eqref{eq:da} into two sub-steps:
\begin{enumerate}[(a),leftmargin=*]
    \item Apply $\nabla (h + \eta t \psi)^*$ to map dual state $\y^{(t)}$ to primal $\x^{(t)}$. 
        Note that this sub-step can be reformulated into
        \begin{equation}
            \x^{(t)} = \argmin_{\x} \left( \left\langle - \y^{(t)}, \x \right\rangle + \eta t \psi(\x) + h(\x)\right),
        \end{equation}
        which allows for efficient computation for many $\psi$, as in PGM.
    \item Update $\y^{(t)}$ to $\y^{(t+1)}$ with the gradient queried at $\x^{(t)}$.
\end{enumerate}
Dual averaging is also known as the \emph{``lazy'' mirror descent} algorithm \cite{Bubeck-15} since it skips the forward mapping $(\nabla h)$ step. 
Theoretically, mirror descent and dual averaging often share the similar convergence rates for sequential \eqref{eq:CO} (e.g., for smooth convex $f$, c.f. \cite{Flammarion.Bach-COLT17}).

\begin{remark}
    There are other algorithms that are popular for certain types of \eqref{eq:CO} problems. 
    For example, \emph{Frank-Wolfe} method \cite{Frank.Wolfe-56,Jaggi-ICML13} solves constrained optimization with a linear optimization oracle.
    Smoothing method \cite{Nesterov-MP05} can also handle non-smoothness in objectives but is in general less efficient than specialized CO algorithms such as dual averaging (c.f., \cite{Nesterov-18}). 
    In this work, we mostly focus on Mirror Descent and Dual Averaging algorithms since they only employ simple proximal oracles such as projection and soft-thresholding.
% FW does not directly extend to general additive composite other than hard constraints.
    % We refer readers to \cref{sec:literature:co} for additional related work in composite optimization.
\end{remark}

Composite optimization has been a classic problem in convex optimization, which covers a variety of statistical inference, machine learning, signal processing problems. 
Mirror Descent (MD, a generalization of proximal gradient method) and Dual Averaging (DA, a.k.a. lazy mirror descent) are two representative algorithms for convex composite optimization.
The \emph{Mirror Descent} (MD) method was originally introduced by \cite{Nemirovski.Yudin-83} for the constrained case and reinterpreted by  \cite{Beck.Teboulle-ORL03}. 
MD was generalized to the composite case by \cite{Duchi.Shalev-shwartz.ea-COLT10} under the name of \textsc{Comid}, though numerous preceding work had studied the special case of \textsc{Comid} under a variety of names such as gradient mapping \cite{Nesterov-MP13}, forward-backward splitting method (FOBOS,\cite{Duchi.Singer-JMLR09}), iterative shrinkage and thresholding (ISTA, \cite{Daubechies.Defrise.ea-04}), and truncated gradient \cite{Langford.Li.ea-JMLR09}.
The \emph{Dual Averaging} (DA) method was introduced by \cite{Nesterov-MP09} for the constrained case, which is also known as \emph{Lazy Mirror Descent} in the literature \cite{Bubeck-15}. The DA method was generalized to the composite (regularized) case by \cite{Xiao-JMLR10,Dekel.Gilad-Bachrach.ea-JMLR12} under the name of Regularized Dual Averaging, and extended by recent works \cite{Flammarion.Bach-COLT17,Lu.Freund.ea-SIOPT18} to account for non-Euclidean geometry induced by an arbitrary distance-generating function $h$. 
DA also has its roots in online learning \cite{Zinkevich-ICML03}, and is related to the follow-the-regularized-leader (FTRL) algorithms \cite{McMahan-AISTATS11}. 
Other variants of MD or DA (such as delayed / skipped proximal step) have been investigated to mitigate the expensive proximal oracles \cite{Mahdavi.Yang.ea-NeurIPS12,Yang.Lin.ea-ICML17}. 
% Composite optimization has also been studied for non-convex objective \cite{Attouch.Bolte.ea-MP13,Chouzenoux.Pesquet.ea-JOTA14,Bredies.Lorenz.ea-JOTA15,Li.Pong-SIOPT15}. 
% These works are typically limited to special cases due to the hardness of non-convex composite optimization, which is in sharp constrast to smooth non-convex settings.
% Unlike smooth unconstrained cases in which simple algorithms like SGD can readily work, the non-convex composite optimization are particularly challenging. 
% \cite{Agarwal.Duchi-NIPS11,Feyzmahdavian.Aytekin.e}
% In addition to MD and DA, there are other algorithms that are popular for certain types of composite optimization problems. For example, Frank-Wolfe method \cite{Frank.Wolfe-56,Jaggi-ICML13} solves constrained optimization with a linear optimization oracle, which is different from the proximal oracle applied by MD and DA. 
We refer readers to \cite{Flammarion.Bach-COLT17,Diakonikolas.Orecchia-SIOPT19} for more detailed discussions on the recent advances of MD and DA.

\subsection{Federated Averaging}
\label{sec:fco:fedavg}
Federated Averaging (\textsc{FedAvg}, \cite{McMahan.Moore.ea-AISTATS17}) is the \emph{de facto} standard algorithm for Federated Learning with unconstrained smooth objectives (namely $\psi \equiv 0$ for (FCO)). 
In this chapter, we follow the exposition of \cite{Reddi.Charles.ea-ICLR21} which splits the client learning rate and server learning rate, offering more flexibility (see \cref{alg:fedavg:slr}).
In this generalized setting, \fedavg involves a series of \emph{rounds} in which each round consists of a client update phase and server update phase. 
We still denote the total number of rounds as $R$. 
At the beginning of each round $r$, a subset of clients $\mathcal{S}^{(r)}$ are sampled from the client pools of size $M$. 
The server state is then broadcast to the sampled client as the client initialization.
During the client update phase, each sampled client runs local SGD for $K$ steps with client learning rate $\eta_{\client}$ with their own data. 
We still use $\x^{(r,k)}_m$ to denote the $m$-th client state at the $k$-th local step of the $r$-th round. 
During the server update phase, the server averages the updates of the sampled clients and treats it as a pseudo-anti-gradient $\bDelta^{(r)}$ (Line 9). 
The server then takes a server update step to update its server state with server learning rate $\eta_{\server}$ and the pseudo-anti-gradient $\bDelta^{(r)}$ (Line 10). 
Note that \cref{alg:fedavg:slr} reduces to the classic setting (\cref{alg:fedavg}) if $\eta_{\client} = \eta$, $\eta_{\server} = 1$, and $\mathcal{S}^{(r)} \equiv [M]$.

% !TEX root = main.tex  
\begin{algorithm}
  \caption{\fedavgfull (\fedavg)}
  \label{alg:fedavg:slr}
  \begin{algorithmic}[1]
  \STATE {\textbf{procedure}} \fedavg ($\x^{(0,0)}, \eta_{\client}, \eta_{\server}$)
  \FOR {$r=0, \ldots, R-1$}
    \STATE sample a subset of clients $\mathcal{S}^{(r)} \subseteq [M]$
      \FORALL {$m \in \mathcal{S}^{(r)}$ {\bf in parallel}}
        \STATE $\x^{(r,0)}_m \gets \x^{(r,0)}$ \COMMENT{broadcast client initialization}
        \FOR {$k = 0, \ldots, K-1$}
          \STATE $\g^{(r,k)}_m \gets \nabla f(\x^{(r,k)}_m; \xi^{(r,k)}_m)$ \COMMENT{query gradient}
          \STATE $\x^{(r,k+1)}_m \gets \x^{(r,k)}_m - \eta_{\client} \cdot \g^{(r,k)}_m$ 
          \COMMENT{client update} 
        \ENDFOR
      \ENDFOR
  \STATE $\bDelta^{(r)} = \frac{1}{|\mathcal{S}^{(r)}|} \sum_{m \in \mathcal{S}^{(r)}} (\x^{(r, K)}_m - \x^{(r, 0)}_m)$ 
  \STATE $\x^{(r+1, 0)} \gets \x^{(r,0)} + \eta_{\server} \cdot \bDelta^{(r)}$ \COMMENT{server update}
  \ENDFOR
\end{algorithmic}
\end{algorithm}

% !TEX root = main.tex  
\section{Proposed Algorithms: \fedmid and \feddualavg}
\label{sec:alg}
In this section, we explore the possible solutions to approach (FCO).
As mentioned earlier, existing FL algorithms such as \fedavg do not apply to (FCO) directly. Although it is possible to apply \fedavg to non-smooth settings by using subgradient in place of the gradient, such an approach is usually ineffective owing to the intrinsic slow convergence of subgradient methods \cite{Boyd.Xiao.ea-03}. 

\subsection{Federated Mirror Descent (\fedmid)}
\label{sec:fedmid}
A more natural extension of \fedavg towards (FCO) is to replace the local SGD steps in \fedavg with local proximal gradient (mirror descent) steps \eqref{eq:md:oneline}.
% The first proposal is to replace the SGD steps in \fedavg with the proximal (mirror descent) steps to handle the non-smooth composite $\psi$.
The resulting algorithm, which we refer to as \emph{\fedmidfull} (\fedmid)\footnote{Despite sharing the same term ``prox'', \fedmid is fundamentally different from \fedprox \cite{Li.Sahu.ea-MLSys20}. 
The proximal step in \fedprox was to regularize the client drift caused by heterogeneity, whereas the proximal step in this work is to overcome the non-smoothness of $\psi$. 
The problems approached by the two methods are also different -- \fedprox still solves an unconstrained smooth problem, whereas ours concerns with approaches (FCO).
}, is outlined in \cref{alg:fedmid}. 
% !TEX root = main.tex  
\begin{algorithm}
  \caption{\fedmidfull (\fedmid)}
  \label{alg:fedmid}
  \begin{algorithmic}[1]
  \STATE {\textbf{procedure}} \fedmid ($\x^{(0,0)}, \eta_{\client}, \eta_{\server}$)
  \FOR {$r=0, \ldots, R-1$}
    \STATE sample a subset of clients $\mathcal{S}^{(r)} \subseteq [M]$
      \FORALL {$m \in \mathcal{S}^{(r)}$ {\bf in parallel}}
        \STATE $\x^{(r,0)}_m \gets \x^{(r,0)}$ \COMMENT{broadcast \emph{primal} initialization}
        \FOR {$k = 0, \ldots, K-1$}
          \STATE $\g^{(r,k)}_m \gets \nabla f(\x^{(r,k)}_m; \xi^{(r,k)}_m)$ \COMMENT{query gradient}
          \STATE $\x^{(r,k+1)}_m \gets \nabla (h + \eta_{\client} \psi)^*(\nabla h(\x^{(r,k)}_m) - \eta_{\client} \cdot \g^{(r,k)}_m)$
            \hfill  \COMMENT{client update}
        \ENDFOR
      \ENDFOR
  \STATE $\bDelta_r = \frac{1}{|\mathcal{S}^{(r)}|} \sum_{m \in \mathcal{S}^{(r)}} (\x^{(r, K)}_m - \x^{(r, 0)}_m)$ 
  \STATE $\x^{(r+1,0)} \gets \nabla (h + \eta_{\server}\eta_{\client}K \psi)^*(\nabla h(\x^{(r,0)}) + \eta_{\server} \cdot \bDelta^{(r)})$ 
          \COMMENT{server update} 
  \ENDFOR
\end{algorithmic}
\end{algorithm}

Specifically, we make two changes compared to \fedavg: 
\begin{itemize}[leftmargin=*]%,partopsep=0pt,parsep=0pt]
    \item The client local SGD steps in \fedavg are replaced with proximal gradient steps (Line 8).
    \item The server update step is replaced with another proximal step (Line 10). 
    % The coefficient of $\psi$ is taken to be $\eta_{\server}\eta_{\client}K$ to keep consistency with the magnitude of $\bDelta^{(r)}$.
\end{itemize}
As a sanity check, for constrained (FCO) with $\psi = \chi_{\mathcal{C}}$, if one takes server learning rate $\eta_{\server}=1$ and Euclidean distance $h(\x) = \frac{1}{2}\|\x\|_2^2$, \fedmid will simply reduce to the following parallel projected SGD with periodic averaging:
\begin{enumerate}[(a),leftmargin=*]
    \item Each sampled client runs $K$ steps of projected SGD following $\x^{(r,k+1)}_m \gets \mathbf{Proj}_{\mathcal{C}}(\x^{(r,k)}_m - \eta_{\client} \g^{(r,k)}_m)$.
    \item After $K$ local steps, the server simply average the client states following $\x^{(r+1,0)} \gets \frac{1}{|\mathcal{S}^{(r)}|} \sum_{m \in \mathcal{S}^{(r)}} \x^{(r,K)}_m$.
\end{enumerate}

\subsection{Limitation of \fedmid: Curse of Primal Averaging}
\label{sec:curse}
Despite its simplicity, \fedmid exhibits a major limitation, which we refer to as ``curse of primal averaging'': the server averaging step in \fedmid may severely impede the optimization progress.
To understand this phenomenon, let us consider the following two illustrative examples:
\begin{itemize}[leftmargin=*]
    \item {\bf Constrained problem}: Suppose the optimum of the aforementioned constrained problem resides on a non-flat boundary $\mathcal{C}$. Even when each client is able to obtain a local solution \emph{on} the boundary, the average of them will almost surely be \emph{off} the boundary (and hence away from the optimum) due to the curvature. 
    \item {\bf Federated $\ell_1$-regularized logistic regression problem}: Suppose each client obtains a local \emph{sparse} solution, simply averaging them across clients will invariably yield a non-sparse solution.
\end{itemize}
As we will see theoretically (\cref{sec:fco:theory}) and empirically (\cref{sec:fco:expr}), the ``curse of primal averaging'' indeed hampers the performance of \fedmid.
% Note that \fedmid requires proximal steps at both the client and the server. To reduce the computation cost at the client-side, we also consider a variant where the proximal step is only performed at the server (see \cref{sec:fco:expr} for details). As we shall see later, this variant of \fedmid performs well in certain settings. 

\subsection{Federated Dual Averaging (\feddualavg)}
\label{sec:feddualavg}
Before we look into the solution of the curse of primal averaging, let us briefly investigate the cause of this effect.
Recall that in standard smooth FL settings, server averaging step is helpful because it implicitly pools the stochastic gradients and thereby reduces the variance \cite{Stich-ICLR19}. 
In \fedmid, however, the server averaging operates on the post-proximal \textbf{primal} states, but the gradient is updated in the \textbf{dual} space (recall the primal-dual interpretation of mirror descent in \cref{sec:co}). 
% Since averaging and proximal are in general not commutable, t
This primal/dual mismatch creates an obstacle for primal averaging to benefit from the pooling of stochastic gradients in dual space. 
% This observation motivates the importance of aligning the 
This thought experiment suggests the importance of aligning the gradient update and server averaging.

Building upon this intuition, we propose a novel primal-dual algorithm, named \emph{\feddualavgfull} (\feddualavg, \cref{alg:feddualavg}), which provably addresses the curse of primal averaging. 
The major novelty of \feddualavg, in comparison with \fedmid or its precursor \fedavg, is to operate the server averaging in the dual space instead of the primal. 
This facilitates the server to aggregate the gradient information since the gradients are also accumulated in the dual space.

Formally, each client maintains a pair of primal and dual states $(\x^{(r,k)}_m, \y^{(r,k)}_m)$. 
At the beginning of each client update round, the client dual state is initialized with the server dual state. 
During the client update stage, each client runs  dual averaging steps following \eqref{eq:da} to update its primal and dual state. 
The coefficient of $\psi$, namely $\tilde{\eta}^{(r,k)}$, is to balance the contribution from $F$ and $\psi$.
At the end of each client update phase, the \emph{dual updates} (instead of primal updates) are returned to the server. 
The server then averages the dual updates of the sampled clients and updates the server dual state.
We observe that the averaging in \feddualavg is two-fold: (1) averaging of gradients in dual space within a client and (2) averaging of dual states across clients at the server. As we shall see shortly in our theoretical analysis, this novel ``double'' averaging of \feddualavg in the non-smooth case enables lower communication complexity and faster convergence of \feddualavg under realistic assumptions.

% !TEX root = main.tex  
\begin{algorithm}
  \caption{\feddualavgfull (\feddualavg)}
  \label{alg:feddualavg}
  \begin{algorithmic}[1]
  \STATE {\textbf{procedure}} \feddualavg ($\x^{(0,0)}, \eta_{\client}, \eta_{\server}$)
  \STATE $\y^{(0,0)} \gets \nabla h(\x^{(0,0)})$ \COMMENT{server dual initialization}
  \FOR {$r=0, \ldots, R-1$}
    \STATE sample a subset of clients $\mathcal{S}^{(r)} \subseteq [M]$
      \FORALL {$m \in \mathcal{S}^{(r)}$ {\bf in parallel}}
        \STATE $\y^{(r,0)}_m \gets \y^{(r,0)}$ \COMMENT{broadcast \emph{dual} initialization}
        \FOR {$k = 0, \ldots, K-1$}
          \STATE $\tilde{\eta}^{(r,k)} \gets \eta_{\server} \eta_{\client} r K + \eta_{\client} k$ 
          \STATE $\x^{(r,k)}_m \gets \nabla (h + \tilde{\eta}^{(r,k)} \psi)^*(\y^{(r,k)}_m)$
            \COMMENT{retrieve primal}
          \STATE $\g^{(r,k)}_m \gets \nabla f (\x^{(r,k)}_m; \xi^{(r,k)}_m) $
            \COMMENT{query gradient}
          \STATE $\y^{(r,k+1)}_m \gets \y^{(r,k)}_m - \eta_{\client} \g^{(r,k)}_m$
            \COMMENT{client \emph{dual} update}
          % \STATE \COMMENT{client update\tikzmark{fedmid:end}} 
        \ENDFOR
      \ENDFOR
  \STATE $\bDelta^{(r)} = \frac{1}{|\mathcal{S}^{(r)}|} \sum_{m \in \mathcal{S}^{(r)}} (\y^{(r, K)}_m - \y^{(r, 0)}_m)$
  \STATE $\y^{(r+1,0)} \gets \y^{(r,0)} + \eta_{\server} \bDelta^{(r)}$
            \COMMENT{server \emph{dual} update}
  \STATE $\x^{(r+1,0)} \gets \nabla (h + \eta_{\server} \eta_{\client} (r+1) K \psi)^* (\y^{(r+1,0)})$
            \COMMENT{(optional) retrieve server primal state}
  \ENDFOR
\end{algorithmic}
\end{algorithm}

% !TEX root = main.tex  
\section{Theoretical Results and Discussions}
\label{sec:fco:theory}
In this section, we demonstrate the theoretical results of \fedmid and \feddualavg. 
We assume the following assumptions throughout the paper. 
The convex analysis definitions in \cref{a1} are reviewed in \cref{sec:background}.
\begin{assumption} Let $\|\cdot\|$ be a norm and $\|\cdot\|_*$ be its dual. Consider the Federated Composite Optimization problem \eqref{FCO}. Assume that
  \label{a1}
  \begin{enumerate}[(a),leftmargin=*]
    \item $\psi: \reals^d \to \reals \cup \{+ \infty\}$ is a closed convex function with closed $\dom \psi$. 
    Assume that $\Phi(\x) = F(\x) + \psi(\x)$ attains a finite optimum at $\x^{\star} \in \dom \psi$.
    \item  $h: \reals^d \to \reals \cup \{+\infty\}$ is a Legendre function that is 1-strongly-convex w.r.t. $\|\cdot\|$. 
    Assume $\dom h \supset \dom \psi$.
    \item
    $f(\cdot; \xi): \reals^{d} \to \reals$ is a closed convex function that is differentiable on $\dom \psi$ for any fixed $\xi$.
    In addition, $f(\cdot; \xi)$ is $L$-smooth w.r.t. $\|\cdot\|$ on $\dom \psi$, % for any fixed $\xi$.
      namely for any $\x, \y \in \dom \psi$, 
      \begin{equation*}
        f(\y;\xi) \leq f(\x;\xi) + \left\langle \nabla f(\x; \xi),  \y - \x \right\rangle + \frac{1}{2} L \|\y - \x\|^2.
      \end{equation*}
    \item $\nabla f$ has $\sigma^2$-bounded variance over $\mathcal{D}_m$ under $\|\cdot\|_*$  within $\dom \psi$, namely for any $\x \in \dom \psi$,
      \begin{equation*}
          \expt_{\xi \sim \mathcal{D}_m} \left\| \nabla f(\x; \xi) - \nabla F_m(\x) \right\|_*^2 \leq \sigma^2, \text{ for any $m \in [M]$}
      \end{equation*}
    \item Assume that all the $M$ clients participate in the client updates for every round, namely $\mathcal{S}^{(r)} = [M]$.
  \end{enumerate}
 
\end{assumption}
\cref{a1}(a) \& (b) are fairly standard for composite optimization analysis (c.f. \cite{Flammarion.Bach-COLT17}). \cref{a1}(c) \& (d) are standard assumptions in stochastic federated optimization as in \cref{asm:fo:2o,asm:fedavg:hetero}. (e) is assumed to simplify the exposition of the theoretical results. All results presented can be easily generalized to the partial participation case.

\begin{remark}
  This work focuses on convex settings because the non-convex composite optimization (either $F$ or $\psi$ non-convex) is noticeably challenging and under-developed \textbf{even for non-distributed settings}.  
This is in sharp contrast to non-convex smooth optimization for which simple algorithms such as SGD can readily work.
Existing literature on non-convex CO (e.g., \cite{Attouch.Bolte.ea-MP13,
Chouzenoux.Pesquet.ea-JOTA14,
Li.Pong-SIOPT15,
Bredies.Lorenz.ea-JOTA15}) typically relies on non-trivial additional assumptions (such as K-Ł conditions) and sophisticated algorithms.
Hence, it is beyond the scope of this work to study non-convex FCO. \footnote{However, we conjecture that for simple non-convex settings (e.g., optimize non-convex $f$ on a convex set, as tested in \cref{sec:emnist}), it is possible to show the convergence and obtain similar advantageous results for \textsc{FedDualAvg}.}
\end{remark}

\subsection{\fedmid and \feddualavg: Small Client Learning Rate Regime}
\label{subsec:small:client:lr}
We first show that both \fedmid and \feddualavg are (asymptotically) at least as good as stochastic mini-batch algorithms with $R$ iterations and batch-size $MK$ when client learning rate $\eta_{\client}$ is sufficiently small. 

\begin{theorem}[Simplified from \cref{small_lr}]
  \label{thm:0}
  Assuming \cref{a1}, then for sufficiently small client learning rate $\eta_{\client}$, and server learning rate $\eta_{\server} = \Theta (\min \{\frac{1} {\eta_{\client} K L}, \frac{B M^{\frac{1}{2}} }{ \eta_{\client} K^{\frac{1}{2}} R^{\frac{1}{2}} \sigma} \} )$, both \feddualavg and \fedmid can output $\hat{\x}$ such that
  \begin{equation}
    \expt \left[ \Phi (\hat{\x}) \right] - \Phi(\x^{\star})  
    \lesssim
    \frac{L B^2}{R} 
    +
    \frac{\sigma B}{\sqrt{MKR}},
    \label{eq:thm:0}
  \end{equation}
  where $B := \sqrt{D_h(\x^{\star}, \x^{(0,0)})}$.
\end{theorem}
The intuition is that when $\eta_{\client}$ is small, the client update will not drift too far away from its initialization of the round. Due to space constraints, the proof is relegated to \cref{sec:small_lr}. 

\subsection{\feddualavg with a Larger Client Learning Rate: Usefulness of Local Step}
\label{sec:feddualavg-benefit}
In this subsection, we show that \feddualavg may attain stronger results with a larger client learning rate. 
In addition to possible faster convergence, \cref{thm:1:simplified,thm:2} also indicate that \feddualavg allows for much broader searching scope of efficient learning rates configurations, which is of key importance for practical purpose.
% We establish faster convergence rates in two important machine learning settings: (i) loss functions with bounded gradients and (ii) ridge regression (or more general quadratic functions)
% Specifically, we focus on the setting when the server learning rate $\eta_{\server} = 1$. 

\paragraph{Bounded Gradient.}
We first consider the setting with bounded gradient.
Unlike unconstrained, the gradient bound may be particularly useful when the constraint is finite.
\begin{theorem}[Simplified from \cref{thm:1}]
  \label{thm:1:simplified}
  Assuming \cref{a1} and $\sup_{\x \in \dom \psi} \|\nabla f(\x; \xi)\|_* \leq G$, then for \feddualavg with $\eta_{\server} = 1$ and  $\eta_{\client} \leq \frac{1}{4L}$, 
  considering
    \begin{equation}
      \hat{\x} := \frac{1}{KR} \sum_{r=0}^{R-1} \sum_{k=1}^{K} 
      \left[ \nabla \left( h + \tilde{\eta}^{(r,k)} \psi \right)^* \left( \frac{1}{M} \sum_{m=1}^M \y^{(r,k)}_m \right) \right],
      \label{eq:x_hat}
    \end{equation}
  the following inequality holds
  \begin{equation}
    \expt \left[ \Phi \left( \hat{\x} \right) \right] - \Phi(\x^{\star})  
    \lesssim
    \frac{B^2}{\eta_{\client} KR}  + \frac{\eta_{\client} \sigma^2}{M} +  \eta_{\client}^2 L K^2 G^2,
  \end{equation}
  where $B := \sqrt{D_h(\x^{\star},\x^{(0,0)})}$. 
  Moreover, there exists $\eta_{\client}$ such that
  \begin{equation}
    \expt \left[ \Phi (\hat{\x}) \right] - \Phi(\x^{\star})  
    \lesssim
    \frac{L B^2}{KR} 
    +
    \frac{\sigma B}{\sqrt{MKR}}
    +
    \frac{L^{\frac{1}{3}} B^{\frac{4}{3}} G^{\frac{2}{3}}}{R^{\frac{2}{3}}}.
    \label{eq:thm:1:simplified:2}
  \end{equation}
\end{theorem}
We refer the reader to \cref{sec:proof:thm:1} for complete proof details of \cref{thm:1:simplified}.
\begin{remark}
The result in \cref{thm:1:simplified} not only matches the rate by \cite{Stich-ICLR19} for smooth, unconstrained \fedavg but also allows for a general non-smooth composite $\psi$, general Bregman divergence induced by $h$, and arbitrary norm $\|\cdot\|$. 
Compared with the small learning rate result \cref{thm:0}, the first term in \cref{eq:thm:1:simplified:2} is improved from $\frac{L B^2}{R}$ to $\frac{L B^2}{KR}$, whereas the third term incurs an additional loss regarding infrequent communication. 
One can verify that the bound \cref{eq:thm:1:simplified:2} is better than \cref{eq:thm:0} if $R \lesssim \frac{L^2 B^2}{G^2}$. 
Therefore, the larger client learning rate may be preferred when the communication is not too infrequent.
\end{remark}

\paragraph{Bounded Heterogeneity.}
Next, we consider the settings with bounded heterogeneity. 
For simplicity, we focus on the case when the loss $F$ is quadratic, as shown in \cref{a3}.
We will discuss other options to relax the quadratic assumption in \cref{sec:proof_sketch}.

\begin{assumption}[Bounded heterogeneity, quadratic]\;
  \label{a3}
  \begin{enumerate}[(a),leftmargin=*]
      \item The heterogeneity of $\nabla F_m$ is bounded, namely
      \begin{equation}
        \sup_{\x \in \dom \psi} \|\nabla F_m(\x) - \nabla F(\x) \|_* \leq  \zeta^2,
        \text{ for any $m \in [M]$}
      \end{equation}
    \item $F(\x) := \frac{1}{2} \x^{\top} \A \x + \c^{\top} \x$ for some $\A \succ 0$.
    \item Assume \cref{a1} is satisfied in which the norm $\|\cdot\|$ is taken to be the $\frac{\A}{\|\A\|_2}$-norm, namely $\|\x\| = \sqrt{\frac{\x^{\top} \A \x}{\|\A\|_2}}$.
  \end{enumerate}
\end{assumption}
\begin{remark}
\cref{a3}(a) is a straightforward extension of bounded heterogeneity \cref{asm:fedavg:hetero}.
Note that \cref{a3} only assumes the objective $F$ to be  quadratic. We do not impose any stronger assumptions on either the composite function $\psi$ or the distance-generating function $h$. 
Therefore, this result still applies to a broad class of problems such as \textsc{Lasso}.
\end{remark}
The following results hold under \cref{a3}. 
\begin{theorem}
  \label{thm:2}
  Assuming \cref{a3}, then for any initialization $\x^{(0,0)} \in \dom \psi$, for unit server learning rate $\eta_{\server} = 1$ and any client learning rate $\eta_{\client} \leq \frac{1}{4L}$, \feddualavg yields
  \begin{equation}
    \expt [\Phi(\hat{\x})] - \Phi(\x^{\star}) 
    \lesssim
    \frac{B^2}{\eta_{\client} KR }
    + \frac{\eta_{\client}  \sigma^2}{M}
    + 7 \eta_{\client}^2 L K \sigma^2
    + 14 \eta_{\client}^2 L K^2  \zeta^2,
    \label{eq:thm:2}
  \end{equation}
  where $\hat{\x}$ is the same as defined in \cref{eq:x_hat}, and $B := \sqrt{D_h(\x^{\star},\x^{(0,0)})}$.

  Particularly for 
  \begin{equation*}
    \eta_{\client} = \min \left\{ \frac{1}{4L}, 
    \frac{M^{\frac{1}{2}} B}{\sigma K^{\frac{1}{2}} R^{\frac{1}{2}}} ,
    \frac{B^{\frac{2}{3}}}{L^{\frac{1}{3}} K^{\frac{2}{3}} R^{\frac{1}{3}} \sigma^{\frac{2}{3}}}, 
    \frac{B^{\frac{2}{3}}}{L^{\frac{1}{3}} K R^{\frac{1}{3}} \zeta^{\frac{2}{3}}}  \right\},
  \end{equation*}  
  we have
  \begin{equation*}
    \expt \left[  \Phi \left(  \hat{\x} \right) - \Phi(\x^{\star}) \right] 
  \leq
  \frac{4L B^2}{KR} 
    +
    \frac{2\sigma B}{M^{\frac{1}{2}} K^{\frac{1}{2}} R^{\frac{1}{2}}}
    +
    \frac{8L^{\frac{1}{3}} B^{\frac{4}{3}} \sigma^{\frac{2}{3}}}{K^{\frac{1}{3}} R^{\frac{2}{3}}}
    +
    \frac{15L^{\frac{1}{3}} B^{\frac{4}{3}} \zeta^{\frac{2}{3}}}{R^{\frac{2}{3}}}.
  \end{equation*}
\end{theorem}

\begin{remark}
    The result in \cref{thm:2} asymptotically matches the best-known convergence rate for smooth, unconstrained \fedavg, namely \cref{thm:fedavg:2o:ub}, while our results allow for general composite $\psi$ and non-Euclidean distance. 
    Compared with \cref{thm:1:simplified}, the overhead in \cref{eq:thm:2} involves variance $\sigma$ and heterogeneity $\zeta$ but no longer depends on $G$. 
    The bound \cref{eq:thm:2} could significantly outperform the previous ones when the variance $\sigma$ and heterogeneity $\zeta$ are relatively mild.
\end{remark}
% !TEX root = main.tex  
\section{Proof of \cref{thm:2}}
\label{sec:proof_sketch}
In this section, we demonstrate our proof framework by providing the proof for \cref{thm:2}. The proofs of other theorems are relegated to the appendix.

\subsection{Proof Overview}
% which provides stronger guarantee for \feddualavg with larger client learning rate $\eta_{\client}$ and unit server learning rate $\eta_{\server}=1$.
\paragraph{Step 1: Convergence of Dual Shadow Sequence.}
We start by characterizing the convergence of the dual shadow sequence $\overline{\y^{(r,k)}} := \frac{1}{M} \sum_{m=1}^M \y^{(r,k)}_m$. 
The key observation for \feddualavg when $\eta_{s} = 1$ is the following relation
\begin{equation}
  \overline{\y^{(r,k+1)}} = \overline{\y^{(r,k)}} - \eta_{\client} \cdot \frac{1}{M} \sum_{m=1}^M \nabla f(\x^{(r,k)}_m; \xi^{(r,k)}_m).
  \label{eq:shadow}
\end{equation}
This suggests that the shadow sequence $\overline{\y^{(r,k)}}$ almost executes a dual averaging update \eqref{eq:da}, but with some perturbed gradient $\frac{1}{M} \sum_{m=1}^M \nabla f(\x^{(r,k)}_m; \xi^{(r,k)}_m)$. 
To this end, we extend the perturbed iterate analysis framework \cite{Mania.Pan.ea-SIOPT17} to the dual space.
Theoretically we show the following \cref{pia:general}, with proof relegated to \cref{sec:pia:general}.

\begin{lemma}[name={Convergence of dual shadow sequence of \feddualavg},label=pia:general,restate=PiaGeneral]
  Assume \cref{a1}, 
  then for any initialization $\x^{(0,0)} \in \dom \psi$, 
  for $\eta_{\server} = 1$, 
  for any $\eta_{\client} \leq \frac{1}{4L}$, \feddualavg yields
  \begin{align}
    & \expt \left[  \Phi \left(  \frac{1}{KR} \sum_{r=0}^{R-1} \sum_{k=1}^K \widehat{\x^{(r,k)}} \right) - \Phi(\x^{\star}) \right] 
    \nonumber \\
    \leq &
    \underbrace{\frac{1}{\eta_{\client} KR } D_h(\x^{\star}, \x^{(0,0)}) 
    + \frac{\eta_{\client}  \sigma^2}{M}}_{\substack{\text{\footnotesize Rate if synchronized} \\ 
    \text{\footnotesize every iteration}}}
    + \underbrace{\frac{L}{M KR }   \left[ \sum_{r=0}^{R-1}  \sum_{k=0}^{K-1} \sum_{m=1}^M   \expt\left \| \overline{\y^{(r,k)}} - \y^{(r,k)}_m \right\|_*^2 \right]}_{\text{\footnotesize  Discrepancy overhead}},
    \label{eq:pia}
  \end{align}
  where
  \begin{equation}
    \widehat{\x^{(r,k)}} :=  \nabla \left( h + \tilde{\eta}^{(r,k)} \psi \right)^* \left( \overline{\y^{(r,k)}}\right)
  \end{equation}
\end{lemma}
\cref{pia:general} decomposes the convergence of \feddualavg into two parts: the first part $\frac{1}{\eta_{\client} KR} D_{h}(\x^{\star}, \x^{(0,0)}) + \frac{\eta_{\client} \sigma^2}{2M} + \frac{L}{MKR}$ corresponds to the convergence rate if all clients were synchronized every iteration. 
The second part $\frac{L}{MKR} \sum_{r=0}^{R-1} \sum_{k=0}^{K-1} \sum_{m=1}^M \expt \| \y^{(r,k)}_m - \overline{\y^{(r,k)}} \|_*^2$ corresponds to the overhead for not synchronizing every step, which we call ``discrepancy overhead''.
\cref{pia:general} can serve as a general interface towards the convergence of \feddualavg as it only assumes the blanket \cref{a1}. 
We defer the proof of \cref{pia:general} to \cref{sec:pia:general}. 

% This rate matches the (centralized) minibatch stochastic dual averaging with batch-size $M$ per iteration and $KR$ iterations in total.
% The last term corresponds to the overhead for not synchronizing every step, which we call ``discrepancy overhead''.
% We refer to this term as ``discrepancy overhead'' since it represents the difference among clients. 
\begin{remark}
Note that the relation \eqref{eq:shadow} is not satisfied by \fedmid due to the incommutability of the proximal operator and the the averaging operator, which thereby breaks \cref{pia:general}. 
Intuitively, this means \fedmid fails to pool the gradients properly (up to a high-order error) in the absence of communication. 
\feddualavg overcomes the incommutability issue because all the gradients are accumulated and averaged  in the dual space, whereas the proximal step only operates at the interface from dual to primal.
% problem because the inter-client averaging is performed in the dual space. 
% Algebraically, the incommutability issue is resolved 
This key difference explains the ``curse of primal averaging'' from the theoretical perspective.
% because of the curse of primal averaging.
\end{remark}

\paragraph{Step 2: Bounding Discrepancy Overhead via Stability Analysis.}
The next step is to bound the discrepancy term introduced in \cref{eq:pia}. 
Intuitively, this term characterizes the \emph{stability} of \feddualavg, in the sense that how far away a single client can deviate from the average (in dual space) if there is no synchronization for $k$ steps. 

However, unlike the smooth convex unconstrained settings in which the stability of SGD is known to be well-behaved \cite{Hardt.Recht.ea-ICML16}, the stability analysis for composite optimization is challenging and absent from the literature. 
We identify that the main challenge originates from the asymmetry of the Bregman divergence. 
In this work, we provide a set of simple conditions, namely \cref{a3}, such that the stability of \feddualavg is well-behaved. 

\begin{lemma}[Dual stability of \feddualavg under \cref{a3}]
  Under the same settings of \cref{thm:2}, the following inequality holds for any $k \in \left\{ 0, 1, \ldots, K\right\}$ and $r \in \left\{ 0, 1, \ldots, R\right\}$,
  \label{quad:stability}
  \begin{equation*}
    \frac{1}{M} \sum_{m=1}^M \expt \left[ \left\| \overline{\y^{(r,k)}} - \y^{(r,k)}_{m} \right\|^2_* \right] 
    \leq
    7 \eta_{\client}^2 K \sigma^2 
    + 14 \eta_{\client}^2 K^2  \zeta^2.
  \end{equation*}
\end{lemma}
The proof of \cref{quad:stability} is deferred to \cref{sec:quad:stability}.

\paragraph{Step 3: Deciding $\eta_{\client}$.}
The final step is to plug in the bound in step 2 back to step 1,
and find appropriate $\eta_{\client}$ to optimize such upper bound. 
For example, combining the results of \cref{pia:general,quad:stability} immediately gives \cref{eq:thm:2} in \cref{thm:2}, namely,
\begin{align}
    \expt [\Phi(\hat{\x})] - \Phi(\x^{\star}) 
    \lesssim 
    \underbrace{\frac{B^2}{\eta_{\client} KR }}_{\substack{\text{Decreasing} \\ \varphi_{\downarrow}(\eta_{\client})}}
    + 
    \underbrace{\frac{\eta_{\client}  \sigma^2}{M}
    + \eta_{\client}^2 L K \sigma^2 
    + \eta_{\client}^2 L K^2 \zeta^2}_{\text{Increasing } \varphi_{\uparrow}(\eta_{\client})},
    \label{eq:thm:2:3}
  \end{align}
\sloppy We claim that \cref{eq:thm:2:3} can be obtained by setting $\eta_{\client} = \min \left\{ \frac{1}{4L}, 
\frac{M^{\frac{1}{2}} B}{\sigma K^{\frac{1}{2}} R^{\frac{1}{2}}} ,
\frac{B^{\frac{2}{3}}}{L^{\frac{1}{3}} K^{\frac{2}{3}} R^{\frac{1}{3}} \sigma^{\frac{2}{3}}}, 
\frac{B^{\frac{2}{3}}}{L^{\frac{1}{3}} K R^{\frac{1}{3}} \zeta^{\frac{2}{3}}}  \right\}$. 
In fact, the decreasing term $\phi_{\downarrow}(\eta_{\client})$ is upper bounded by $\phi_{\downarrow}\left( \frac{1}{4L} \right) 
+ \phi_{\downarrow}\left( \frac{M^{\frac{1}{2}} B}{\sigma K^{\frac{1}{2}} R^{\frac{1}{2}}}  \right) 
+ \phi_{\downarrow}\left( \frac{B^{\frac{2}{3}}}{L^{\frac{1}{3}} K^{\frac{2}{3}} R^{\frac{1}{3}} \sigma^{\frac{2}{3}}} \right) 
+ \phi_{\downarrow}\left( \frac{B^{\frac{2}{3}}}{L^{\frac{1}{3}} K R^{\frac{1}{3}} \zeta^{\frac{2}{3}}}  \right)$, which is upper bounded by the RHS of \cref{eq:thm:2}. Similarly, one can show that the $\varphi_{\uparrow}(\eta_{\client})$ is also upper bounded by the RHS of \cref{eq:thm:2} for the same choice of $\eta_{\client}$. This concludes the proof of \cref{thm:2}, and \cref{thm:1:simplified} can be obtained through the same argument. We defer the details to \cref{sec:feddualavg:step3}.

\subsection{Details of Step 1: Proof of \cref{pia:general}}
\label{sec:pia:general}
In this subsection, we prove \cref{pia:general}. We start by showing the following \cref{one:step:analysis} regarding the one step improvement of the shadow sequence $\overline{\y^{(r,k)}}$. 
\begin{proposition}[One step analysis of \feddualavg]
  \label{one:step:analysis}
  Under the same assumptions of \cref{pia:general}, the following inequality holds
  \begin{align*}
    \expt \left[ \tilde{D}_{h_{r,k+1}}(\x^{\star}, \overline{\y^{(r,k+1)}}) \middle| \mathcal{F}^{(r,k)} \right] 
    \leq & \tilde{D}_{h_{r,k}} (\x^{\star}, \overline{\y^{(r,k)}}) - \eta_{\client} \expt \left[  \Phi(\widehat{\x^{(r,k+1)}}) - \Phi(\x^{\star}) \middle| \mathcal{F}^{(r,k)} \right] 
    \\ 
    & 
    + \eta_{\client} L \cdot \expt \left[ \frac{1}{M}  \sum_{m=1}^M \left \| \overline{\y^{(r,k)}} - \y^{(r,k)}_m \right\|_*^2 \middle| \mathcal{F}^{(r,k)} \right] 
    + \frac{\eta_{\client}^2 \sigma^2}{M},
  \end{align*}
  where $\tilde{D}$ is the generalized Bregman divergence defined in \cref{def:generalized_bregman}.
\end{proposition}
The proof of \cref{one:step:analysis} relies on the following two claims regarding the deterministic analysis of \feddualavg. We defer the proof of \cref{one:step:analysis:claim:1,one:step:analysis:claim:2} to \cref{sec:one:step:analysis:claim:1,sec:one:step:analysis:claim:2}, respectively.
\begin{claim}
  \label{one:step:analysis:claim:1}
  Under the same assumptions of \cref{pia:general}, the following inequality holds
  \begin{align}
    & \tilde{D}_{h_{r,k+1}}(\x^{\star}, \overline{\y^{(r,k+1)}}) 
    \nonumber \\
    = &
    \tilde{D}_{h_{r,k}}(\x^{\star}, \overline{\y^{(r,k)}}) - \tilde{D}_{h_{r,k}}(\widehat{\x^{(r,k+1)}}, \overline{\y^{(r,k)}}) 
    \nonumber \\
    & - \eta_{\client} (\psi(\widehat{\x^{(r,k+1)}}) - \psi(\x^{\star}))) - \eta_{\client} \left\langle \frac{1}{M} \sum_{m=1}^{M} \nabla f(\x^{(r,k)}_m; \xi^{(r,k)}_m), \widehat{\x^{(r,k+1)}} - \x^{\star} \right\rangle.
    \label{eq:one:step:analysis:claim:1}
  \end{align}
\end{claim}
\begin{claim}
  \label{one:step:analysis:claim:2}
  Under the same assumptions of \cref{pia:general}, it is the case that 
  \begin{align}
     & F(\widehat{\x^{(r,k+1)}}) - F(\x^{\star}) \leq \left\langle \frac{1}{M} \sum_{m=1}^M \nabla f(\x^{(r,k)}_m; \xi^{(r,k)}_m), \widehat{\x^{(r,k+1)}} - \x^{\star} \right\rangle  
     \nonumber  \\
        & \qquad \qquad + \left\langle \frac{1}{M} \sum_{m=1}^M \left( \nabla F_m(\x^{(r,k)}_m) - \nabla f(\x^{(r,k)}_m; \xi^{(r,k)}_m) \right), \widehat{\x^{(r,k+1)}} - \x^{\star} \right\rangle  
      \nonumber \\
        & \qquad \qquad + L \| \widehat{\x^{(r,k+1)}} - \widehat{\x^{(r,k)}} \|^2  
         +  \frac{L}{M} \sum_{m=1}^M \left \| \overline{\y^{(r,k)}} - \y^{(r,k)}_m \right\|_*^2.
         \label{eq:one:step:analysis:claim:2}
  \end{align}
\end{claim}
With \cref{one:step:analysis:claim:1,one:step:analysis:claim:2} at hand, we are ready to prove the one step analysis \cref{one:step:analysis}.
\begin{proof}[Proof of \cref{one:step:analysis}]
  Applying \cref{one:step:analysis:claim:1,one:step:analysis:claim:2} yields 
  (summating \cref{eq:one:step:analysis:claim:1} with $\eta_c$ times of \cref{eq:one:step:analysis:claim:2}),
  \begin{align}
    \tilde{D}_{h_{r,k+1}}(\x^{\star}, \overline{\y^{(r,k+1)}})
    \leq &
    \tilde{D}_{h_{r,k}}(\x^{\star}, \overline{\y^{(r,k)}}) - \tilde{D}_{h_{r,k}}(\widehat{\x^{(r,k+1)}}, \overline{\y^{(r,k)}})
    + \eta_{\client} L \| \widehat{\x^{(r,k+1)}} - \widehat{\x^{(r,k)}} \|^2 
    \nonumber \\
    & + \eta_{\client}  \left\langle \frac{1}{M} \sum_{m=1}^M \left( \nabla F_m(\x^{(r,k)}_m) - \nabla f(\x^{(r,k)}_m; \xi^{(r,k)}_m) \right), \widehat{\x^{(r,k+1)}} - \x^{\star} \right\rangle 
    \nonumber \\
    & - \eta_{\client} \left( \Phi(\widehat{\x^{(r,k+1)}}) - \Phi(\x^{\star}) \right) 
      + \eta_{\client} L \cdot \frac{1}{M} \sum_{m=1}^M \left \| \overline{\y^{(r,k)}} - \y^{(r,k)}_m \right\|_*^2.
      \label{eq:proof:one:step:analysis:1}
  \end{align}
  Note that 
  \begin{equation*}
    \tilde{D}_{h_{r,k}}(\widehat{\x^{(r,k+1)}}, \overline{\y^{(r,k)}}) \geq D_h(\widehat{\x^{(r,k+1)}}, \nabla h_{r,k}^* ( \overline{\y^{(r,k)}})) = D_h(\widehat{\x^{(r,k+1)}},  \widehat{\x^{(r,k)}}) \geq \frac{1}{2}\| \widehat{\x^{(r,k+1)}} -  \widehat{\x^{(r,k)}}\|^2,
  \end{equation*}
  and 
  \begin{equation*}
    \eta_{\client} L \| \widehat{\x^{(r,k+1)}} - \widehat{\x^{(r,k)}} \|^2 
    \leq
    \frac{1}{4} \| \widehat{\x^{(r,k+1)}} - \widehat{\x^{(r,k)}} \|^2,
  \end{equation*}
  since $\eta_{\client} \leq \frac{1}{4L}$ by assumption. Therefore,
  \begin{equation}
    - \tilde{D}_{h_{r,k}}(\widehat{\x^{(r,k+1)}}, \overline{\y^{(r,k)}}) 
    +
    \eta_{\client} L \| \widehat{\x^{(r,k+1)}} - \widehat{\x^{(r,k)}} \|^2 
    \leq
    -\frac{1}{4}  \| \widehat{\x^{(r,k+1)}} - \widehat{\x^{(r,k)}} \|^2.
    \label{eq:proof:one:step:analysis:2}
  \end{equation}
  Plugging \cref{eq:proof:one:step:analysis:2} to \cref{eq:proof:one:step:analysis:1} gives
  \begin{align}
    \tilde{D}_{h_{r,k+1}}(\x^{\star}, \overline{\y^{(r,k+1)}})
    \leq &
    \tilde{D}_{h_{r,k}}(\x^{\star}, \overline{\y^{(r,k)}}) - \frac{1}{4} \| \widehat{\x^{(r,k+1)}} - \widehat{\x^{(r,k)}} \|^2 
    - \eta_{\client} \left( \Phi(\widehat{\x^{(r,k+1)}}) - \Phi(\x^{\star}) \right) 
    \nonumber \\
    & + \eta_{\client}  \left\langle \frac{1}{M} \sum_{m=1}^M \left( \nabla F_m(\x^{(r,k)}_m) - \nabla f(\x^{(r,k)}_m; \xi^{(r,k)}_m) \right), \widehat{\x^{(r,k+1)}} - \x^{\star} \right\rangle 
    \nonumber \\
    & + \eta_{\client} L \cdot \frac{1}{M} \sum_{m=1}^M \left \| \overline{\y^{(r,k)}} - \y^{(r,k)}_m \right\|_*^2.
      \label{eq:proof:one:step:analysis:3}
  \end{align}

  Now we take conditional expectation. Note that
  \begin{align}
        &   \expt \left[ \left \langle \frac{1}{M} \sum_{m=1}^M \nabla F_m(\x^{(r,k)}_m) - \nabla f(\x^{(r,k)}_m; \xi^{(r,k)}_m), \widehat{\x^{(r,k+1)}} - \x^{\star} \right\rangle  \middle| \mathcal{F}^{(r,k)} \right]
      \nonumber \\
      = &  \expt \left[ \left \langle \frac{1}{M}  \sum_{m=1}^M \nabla F_m(\x^{(r,k)}_m) - \nabla f(\x^{(r,k)}_m; \xi^{(r,k)}_m), \widehat{\x^{(r,k+1)}} - \widehat{\x^{(r,k)}}  \right\rangle  \middle| \mathcal{F}^{(r,k)} \right]
      \tag{since $\expt_{\xi^{(r,k)}_m \sim \mathcal{D}_m} [\nabla f(\x^{(r,k)}_m; \xi^{(r,k)}_m)] = \nabla F_m(\x^{(r,k)}_m)$ }
      \\
      \leq & \expt \left[ \left\| \frac{1}{M}  \sum_{m=1}^M \nabla F_m(\x^{(r,k)}_m) - \nabla f(\x^{(r,k)}_m; \xi^{(r,k)}_m) \right\|_* \middle| \mathcal{F}^{(r,k)} \right]
      \cdot \expt \left[  \left\| \widehat{\x^{(r,k+1)}} - \widehat{\x^{(r,k)}}   \right\| \middle| \mathcal{F}^{(r,k)} \right]
      \tag{by definition of dual norm $\|\cdot\|_*$}
      \\
      \leq & \frac{\sigma}{\sqrt{M}}  \expt \left[ \left\| \widehat{\x^{(r,k+1)}} - \widehat{\x^{(r,k)}}   \right\| \middle| \mathcal{F}^{(r,k)} \right].
      \tag{by bounded variance assumption and independence}
  \end{align}
  Plugging the above inequality to \cref{eq:proof:one:step:analysis:3} gives
  \begin{align}
      & \expt \left[ \tilde{D}_{h_{r,k+1}}(\x^{\star}, \overline{\y^{(r,k+1)}}) \middle| \mathcal{F}^{(r,k)} \right] 
     \nonumber \\
     \leq & 
     \tilde{D}_{h_{r,k}} (\x^{\star}, \overline{\y^{(r,k)}}) - \eta_{\client} \expt \left[  \Phi(\widehat{\x^{(r,k+1)}}) - \Phi(\x^{\star}) \middle| \mathcal{F}^{(r,k)} \right] 
     \nonumber \\
     & + \eta_{\client} L \cdot \expt \left[ \frac{1}{M}  \sum_{m=1}^M \left \| \overline{\y^{(r,k)}} - \y^{(r,k)}_m \right\|_*^2 \middle| \mathcal{F}^{(r,k)} \right]
     \nonumber \\ 
     & + \frac{\eta_{\client} \sigma}{\sqrt{M}}   \expt \left[ \left\| \widehat{\x^{(r,k+1)}} - \widehat{\x^{(r,k)}}   \right\| \middle| \mathcal{F}^{(r,k)} \right] - \frac{1}{4} \expt \left[ \| \widehat{\x^{(r,k+1)}} - \widehat{\x^{(r,k)}} \|^2 \middle| \mathcal{F}^{(r,k)} \right]
     \nonumber \\
     \leq & \tilde{D}_{h_{r,k}} (\x^{\star}, \overline{\y^{(r,k)}}) - \eta_{\client} \expt \left[  \Phi(\widehat{\x^{(r,k+1)}}) - \Phi(\x^{\star}) \middle| \mathcal{F}^{(r,k)} \right] 
     \nonumber \\
     & + \eta_{\client} L \cdot \expt \left[ \frac{1}{M}  \sum_{m=1}^M  \left \| \overline{\y^{(r,k)}} - \y^{(r,k)}_m \right\|_*^2 \middle| \mathcal{F}^{(r,k)} \right] + \frac{\eta_{\client}^2 \sigma^2}{M},
     \tag{by quadratic maximum}
  \end{align}
  completing the proof of \cref{one:step:analysis}.
\end{proof}

The \cref{pia:general} then follows by telescoping the one step analysis \cref{one:step:analysis}.
\begin{proof}[Proof of \cref{pia:general}]
  Let us first telescope \cref{one:step:analysis} within the same round $r$, from $k=0$ to $K$, which gives
  \begin{align*}
    \expt \left[ \tilde{D}_{h_{r,K}}(\x^{\star}, \overline{\y^{(r,K)}}) \middle| \mathcal{F}^{(r,0)} \right] 
    \leq & \tilde{D}_{h_{r,0}} (\x^{\star}, \overline{\y^{(r,0)}}) - \eta_{\client} \sum_{k=1}^{K} \expt \left[  \Phi(\widehat{\x^{(r,k)}}) - \Phi(\x^{\star}) \middle| \mathcal{F}^{(r,0)} \right] 
    \\
    &   + \eta_{\client} L \cdot \expt \left[ \frac{1}{M} \sum_{k=0}^{K-1} \sum_{m=1}^M  \left \| \overline{\y^{(r,k)}} - \y^{(r,k)}_m \right\|_*^2\middle| \mathcal{F}^{(r,0)} \right] 
    + \frac{\eta_{\client}^2 K \sigma^2}{M}.
  \end{align*}
  Since server learning rate $\eta_{\server}=1$ we have $\overline{\y^{(r,K)}} = \overline{\y^{(r+1,0)}}$. 
  Therefore, we can telescope the round from $r = 0$ to $R$, which gives
  \begin{align*}
    \expt \left[ \tilde{D}_{h_{R,0}}(\x^{\star}, \overline{\y^{(R,0)}}) \right] 
    \leq &
    \tilde{D}_{h_{0,0}}(\x^{\star}, \overline{\y^{(0,0)}}) - \eta_{\client} \sum_{r=0}^{R-1} \sum_{k=1}^K \expt \left[  \Phi(\widehat{\x^{(r,k)}}) - \Phi(\x^{\star}) \right] \\
    & + \eta_c L \cdot \expt \left[ \frac{1}{M}\sum_{r=0}^{R-1}  \sum_{k=0}^{K-1} \sum_{m=1}^M  \left \| \overline{\y^{(r,k)}} - \y^{(r,k)}_m \right\|_*^2 \right] 
    + \frac{\eta_{\client}^2 K  R \sigma^2}{M}.
  \end{align*}
  Dividing both sides by $\eta_{\client} \cdot KR$ and rearranging 
  \begin{align*}
    \frac{1}{KR }\sum_{r=0}^{R-1} \sum_{k=1}^K \expt \left[  \Phi(\widehat{\x^{(r,k)}}) - \Phi(\x^{\star}) \right] 
    \leq &
    \frac{1}{\eta_{\client} KR }\left( \tilde{D}_{h_{0,0}}(\x^{\star}, \overline{\y^{(0,0)}}) - \expt \left[ \tilde{D}_{h_{R,0}}(\x^{\star}, \overline{\y^{(R,0)}}) \right]  \right)
    \\
    & + L \cdot \expt \left[ \frac{1}{M KR }\sum_{r=0}^{R-1}  \sum_{k=0}^{K-1} \sum_{m=1}^M  \left \| \overline{\y^{(r,k)}} - \y^{(r,k)}_m \right\|_*^2 \right] 
    + \frac{\eta_{\client}  \sigma^2}{M}.
  \end{align*}
  Applying Jensen's inequality on the LHS and dropping the negative term on the RHS yield
  \begin{align}
    & \expt \left[  \Phi \left(  \frac{1}{KR} \sum_{r=0}^{R-1} \sum_{k=1}^K \widehat{\x^{(r,k)}} \right) - \Phi(\x^{\star}) \right] 
    \nonumber \\
    \leq &
    \frac{1}{\eta_{\client} KR } \tilde{D}_{h_{0,0}}(\x^{\star}, \overline{\y^{(0,0)}}) 
    + \frac{L}{M KR }   \left[ \sum_{r=0}^{R-1}  \sum_{k=0}^{K-1} \sum_{m=1}^M   \expt\left \| \overline{\y^{(r,k)}} - \y^{(r,k)}_m \right\|_*^2 \right] 
    + \frac{\eta_{\client}  \sigma^2}{M}.
    \label{eq:pia:1}
  \end{align}

  Since $\overline{\y^{(0,0)}} = \nabla h (\x^{(0,0)})$ and $\x^{(0,0)} \in \dom \psi$, we have $\nabla h_{0,0}^*(\nabla h(\x^{(0,0)})) = \x^{(0,0)}$ by \cref{prop:legendre} since $h$ is assumed to be of Legendre type. Consequently
  \begin{align}
    \tilde{D}_{h_{0,0}}(\x^{\star}, \overline{\y^{(0,0)}}) & = h(\x^{\star}) - h(\nabla h_{0,0}^*(\nabla h(\x^{(0,0)}))) - \left\langle \y^{(0,0)}, \x^{\star} - \nabla h_{0,0}^*(\nabla h(\x^{(0,0)})) \right\rangle
    \nonumber \\
    & = h(\x^{\star}) - h(\x^{(0,0)}) - \left\langle \nabla h(\x^{(0,0)}), \x^{\star} - \x^{(0,0)} \right\rangle = D_h(\x^{\star},\x^{(0,0)}).
    \label{eq:pia:2}
  \end{align}
  Plugging \cref{eq:pia:2} back to  \cref{eq:pia:1} completes the proof of \cref{pia:general}.
\end{proof} 

\subsubsection{Deferred Proof of  \cref{one:step:analysis:claim:1}}
\label{sec:one:step:analysis:claim:1}
\begin{proof}[Proof of \cref{one:step:analysis:claim:1}]
  By definition of \feddualavg procedure, for all $m \in [M]$, $k \in \{0, 1, \ldots, K-1\}$, we have
  \begin{equation*}
    \y^{(r,k+1)}_m = \y^{(r,k)}_m - \eta_{\client} \nabla f(\x^{(r,k)}_m; \xi^{(r,k)}_m).
  \end{equation*}
  Taking average over $m \in [M]$ gives (recall the overline denotes the average over clients)
  \begin{equation}
    \overline{\y^{(r,k+1)}} = \overline{\y^{(r,k)}} - \eta_{\client} \cdot \frac{1}{M} \sum_{m=1}^{M} \nabla f(\x^{(r,k)}_m; \xi^{(r,k)}_m).
    \label{eq:one:step:analysis:1}
  \end{equation}
  
  Now we study generalized Bregman divergence $\tilde{D}_{h,k+1}(\x^{\star}, \overline{\y^{(r,k+1)}})$ for any arbitrary pre-fixed $w \in \dom h_{r,k}$ 
  \begin{align}
        & \tilde{D}_{h_{r,k+1}}(\x^{\star}, \overline{\y^{(r,k+1)}})
      \nonumber \\
      = &
      h_{r,k+1}(\x^{\star}) - h_{r,k+1} \left( \nabla h_{r,k+1}^*(\overline{\y^{(r,k+1)}}) \right) - \left\langle \overline{\y^{(r,k+1)}}, \x^{\star} - \nabla h_{r,k+1}^*(\overline{\y^{(r,k+1)}}) \right\rangle \tag{By definition of $\tilde{D}$}
      \\
      = &
      h_{r,k+1}(\x^{\star}) - h_{r,k+1} \left( \widehat{\x^{(r,k+1)}} \right) - \left\langle \overline{\y^{(r,k+1)}}, \x^{\star} - \widehat{\x^{(r,k+1)}} \right\rangle
      \tag{By definition of $\widehat{\x^{(r,k+1)}}$}
      \\
      = &
      h_{r,k+1}(\x^{\star}) - h_{r,k+1} \left( \widehat{\x^{(r,k+1)}} \right) - \left\langle \overline{\y^{(r,k)}} - \eta_{\client} \cdot \frac{1}{M} \sum_{m=1}^{M} \nabla f(\x^{(r,k)}_m; \xi^{(r,k)}_m), \x^{\star} - \widehat{\x^{(r,k+1)}} \right\rangle
      \tag{By \cref{eq:one:step:analysis:1}}
      \\
      = & (h_{r,k}(\x^{\star}) + \eta_{\client} \psi(\x^{\star})) - (h_{r,k}(\widehat{\x^{(r,k+1)}}) + \eta_{\client} \psi(\widehat{\x^{(r,k+1)}}))
      \nonumber \\
      & - \left\langle \overline{\y^{(r,k)}} - \eta_{\client} \cdot \frac{1}{M} \sum_{m=1}^{M} \nabla f(\x^{(r,k)}_m; \xi^{(r,k)}_m), \x^{\star} - \widehat{\x^{(r,k+1)}} \right\rangle
      \tag{Since $h_{r,k+1} = h_{r,k} + \eta_{\client}\psi$ by definition of $h_{r,k+1}$} 
      \\
      = & \left[ h_{r,k}(\x^{\star}) - h_{r,k}(\widehat{\x^{(r,k)}}) - \left\langle \overline{\y^{(r,k)}}, \x^{\star} - \widehat{\x^{(r,k)}} \right\rangle \right] 
      \nonumber \\
      & - 
      \left[ h_{r,k}(\widehat{\x^{(r,k+1)}}) - h_{r,k}(\widehat{\x^{(r,k)}}) - \left\langle \overline{\y^{(r,k)}}, \widehat{\x^{(r,k+1)}} - \widehat{\x^{(r,k)}}  \right\rangle \right]
      \nonumber \\
      & - \eta_{\client} \left( \psi(\widehat{\x^{(r,k+1)}}) - \psi(\x^{\star}) \right) - \eta_{\client} \left\langle \frac{1}{M} \sum_{m=1}^{M} \nabla f(\x^{(r,k)}_m; \xi^{(r,k)}_m), \widehat{\x^{(r,k+1)}} - \x^{\star} \right\rangle
      \tag{Rearranging}
      \\
      = & \tilde{D}_{h_{r,k}}(\x^{\star}, \overline{\y^{(r,k)}}) - \tilde{D}_{h_{r,k}}(\widehat{\x^{(r,k+1)}}, \overline{\y^{(r,k)}}) 
      - \eta_{\client} (\psi(\widehat{\x^{(r,k+1)}}) - \psi(\x^{\star}))) 
      \nonumber \\
      & - \eta_{\client} \left\langle \frac{1}{M} \sum_{m=1}^{M} \nabla f(\x^{(r,k)}_m; \xi^{(r,k)}_m), \widehat{\x^{(r,k+1)}} - \x^{\star} \right\rangle,
      \nonumber
  \end{align} 
  where the last equality is by definition of $\tilde{D}$.
  \end{proof}

\subsubsection{Deferred Proof of  \cref{one:step:analysis:claim:2}}
\label{sec:one:step:analysis:claim:2}
\begin{proof}[Proof of \cref{one:step:analysis:claim:2}]
  By smoothness and convexity of $F_m$, we know
    \begin{align}
      F_m(\widehat{\x^{(r,k+1)}}) & \leq F_m(\x^{(r,k)}_m) + \left\langle \nabla F_m(\x^{(r,k)}_m),  \widehat{\x^{(r,k+1)}} - \x^{(r,k)}_m \right\rangle + \frac{L}{2} \| \widehat{\x^{(r,k+1)}} - \x^{(r,k)}_m \|^2 
      \tag{smoothness}
      \\
      & \leq F_m(\x^{\star}) + \left\langle \nabla F_m(\x^{(r,k)}_m),  \widehat{\x^{(r,k+1)}} - \x\right\rangle + \frac{L}{2} \| \widehat{\x^{(r,k+1)}} - \x^{(r,k)}_m \|^2.
      \tag{convexity}
  \end{align}
  Taking summation over $m$ gives
  \begin{align}
      &     F(\widehat{\x^{(r,k+1)}}) - F(\x^{\star})  = \frac{1}{M} \sum_{m=1}^M \left( F_m(\widehat{\x^{(r,k+1)}}) - F_m(\x^{\star}) \right)  
      \nonumber \\
      \leq & \left\langle \frac{1}{M} \sum_{m=1}^M \nabla F_m(\x^{(r,k)}_m), \widehat{\x^{(r,k+1)}} - \x^{\star} \right\rangle  + \frac{L}{2M} \sum_{m=1}^M  \| \widehat{\x^{(r,k+1)}} - \x^{(r,k)}_m \|^2
      \nonumber \\
      = & \left\langle \frac{1}{M} \sum_{m=1}^M \nabla f(\x^{(r,k)}_m; \xi^{(r,k)}_m), \widehat{\x^{(r,k+1)}} - \x^{\star} \right\rangle  
      + \frac{L}{2M} \sum_{m=1}^M  \| \widehat{\x^{(r,k+1)}} - \x^{(r,k)}_m \|^2 
      \nonumber \\ 
      & + \left\langle \frac{1}{M} \sum_{m=1}^M \left( \nabla F_m(\x^{(r,k)}_m) - \nabla f(\x^{(r,k)}_m; \xi^{(r,k)}_m) \right), \widehat{\x^{(r,k+1)}} - \x^{\star} \right\rangle  
      \nonumber \\
      \leq & \left\langle \frac{1}{M} \sum_{m=1}^M \nabla f(\x^{(r,k)}_m; \xi^{(r,k)}_m), \widehat{\x^{(r,k+1)}} - \x^{\star} \right\rangle 
      + L \| \widehat{\x^{(r,k+1)}} - \widehat{\x^{(r,k)}} \|^2  +  \frac{L}{M} \sum_{m=1}^M  \| \widehat{\x^{(r,k)}} - \x^{(r,k)}_m \|^2 
      \nonumber \\
          & + \left\langle \frac{1}{M} \sum_{m=1}^M \left( \nabla F_m(\x^{(r,k)}_m) - \nabla f(\x^{(r,k)}_m; \xi^{(r,k)}_m) \right), \widehat{\x^{(r,k+1)}} - \x^{\star} \right\rangle,
      \label{eq:one_step_analysis:claim:2:1}
  \end{align}
  where in the last inequality we applied the triangle inequality (for an arbitrary norm $\|\cdot\|$):
  \begin{align*}
    \| \widehat{\x^{(r,k+1)}} - \x^{(r,k)}_m \|^2 
    \leq &
    \left(  \| \widehat{\x^{(r,k+1)}} - \widehat{\x^{(r,k)}}\| + \|\widehat{\x^{(r,k)}} - \x^{(r,k)}_m \| \right)^2 
    \\
    \leq &
    2 \| \widehat{\x^{(r,k+1)}} - \widehat{\x^{(r,k)}}\|^2 
    + 2\|\widehat{\x^{(r,k)}} - \x^{(r,k)}_m \|^2.
  \end{align*}

  Since $\psi$ is convex and $h$ is $1$-strongly-convex according to \cref{a1}, we know that $h_{r,k} = h + \eta_{\client} (rK + k) \psi$ is also $1$-strongly-convex. 
  Therefore, $h_{r,k}^*$ is $1$-smooth by \cref{strongly:convex:conjugate}.
  Consequently,
  \begin{equation}
       \left\| \x^{(r,k)}_m - \widehat{\x^{(r,k)}} \right\|^2 = 
       \left\| \nabla h_{r,k}^*(\y^{(r,k)}_m) - \nabla h_{r,k}^*(\overline{\y^{(r,k)}}) \right\|^2 
      \leq  \left\| \y^{(r,k)}_m - \overline{\y^{(r,k)}} \right\|_*^2,
      \label{eq:one_step_analysis:claim:2:2}
  \end{equation}
  where the first equality is by definition of $\x^{(r,k)}_m$ and $\widehat{\x^{(r,k)}}$ and the second inequality is by $1$-smoothness. Plugging \cref{eq:one_step_analysis:claim:2:2} back to \cref{eq:one_step_analysis:claim:2:1} completes the proof of \cref{one:step:analysis:claim:2}.
\end{proof}

\subsection{Details of Step 2: Proof of \cref{quad:stability}}
\label{sec:quad:stability}
In this subsection, we prove \cref{quad:stability} on the stability of \feddualavg for quadratic $F$. We first state and prove the following \cref{quad:one_step_stab} on the one-step analysis of stability. 
\begin{proposition}
  In the same settings of \cref{thm:2}, let $m_1, m_2 \in [M]$ be two arbitrary clients. Then the following inequality holds
  \label{quad:one_step_stab}
  \begin{align*}
    & \expt \left[ \left\|\y^{(r,k+1)}_{m_1} - \y^{(r,k+1)}_{m_2} \right\|_{\A^{-1}}^2  \middle| \mathcal{F}^{(r,k)} \right]
    \\
    \leq &
    \left( 1 + \frac{1}{K} \right)  
    \left\| \y^{(r,k)}_{m_1} - \y^{(r,k)}_{m_2} \right\|_{\A^{-1}}^2 
    + 2 \left( 1 + \frac{1}{K} \right)   \eta_{\client}^2 \sigma^2 \|\A\|_2^{-1}
    + 4 ( 1+ K) \eta_{\client}^2 \zeta^2 \|\A\|_2^{-1}.
  \end{align*}
\end{proposition}
The proof of \cref{quad:one_step_stab} relies on the following three claims.
To simplify the exposition, we introduce two more notations for this subsection. For any $r, k, m$, let
\begin{equation*}
  \bvarepsilon^{(r,k)}_{m} :=  \nabla f(\x^{(r,k)}_m; \xi^{(r,k)}_{m}) - \nabla F_m(\x^{(r,k)}_m),
  \qquad
  \bdelta^{(r,k)}_m := \nabla F_m(\x^{(r,k)}_m) - \nabla  F(\x^{(r,k)}_m).
\end{equation*}

The following claim upper bounds the growth of $\left\|\y^{(r,k+1)}_{m_1} - \y^{(r,k+1)}_{m_2} \right\|_{\A^{-1}}^2$. The proof of \cref{quad:one_step_stab:claim:1} is deferred to \cref{sec:proof:quad:one_step_stab:claim:1}.
\begin{claim}
  \label{quad:one_step_stab:claim:1}
  In the same settings of \cref{quad:one_step_stab}, the following inequality holds
  \begin{align}
    & \left\|\y^{(r,k+1)}_{m_1} - \y^{(r,k+1)}_{m_2} \right\|_{\A^{-1}}^2 
    \leq
    \left( 1 + K \right) \eta_{\client}^2 \left\| \bdelta^{(r,k)}_{m_1} - \bdelta^{(r,k)}_{m_2}  \right\|_{\A^{-1}}^2
    \nonumber \\ 
    & \quad + \left( 1 + \frac{1}{K} \right) 
    \left\| \y^{(r,k)}_{m_1} - \y^{(r,k)}_{m_2} - \eta_{\client} \cdot \A \left( \x^{(r,k)}_{m_1} - \x^{(r,k)}_{m_2} \right) - \eta_{\client} \left( \bvarepsilon^{(r,k)}_{m_1} - \bvarepsilon^{(r,k)}_{m_2} \right) \right\|_{\A^{-1}}^2.
    \label{eq:quad:one_step_stab:claim:1}
  \end{align}
\end{claim}

The next claim upper bounds the growth of the first term in \cref{eq:quad:one_step_stab:claim:1} in conditional expectation.
We extend the stability technique in \cite{Flammarion.Bach-COLT17} to bound this term. 
The proof of \cref{quad:one_step_stab:claim:2} is deferred to \cref{sec:proof:quad:one_step_stab:claim:2}.
\begin{claim}
  \label{quad:one_step_stab:claim:2}
  In the same settings of \cref{quad:one_step_stab}, the following inequality holds
  \begin{align*}
    & \expt \left[ \left\| \y^{(r,k)}_{m_1} - \y^{(r,k)}_{m_2} - \eta_{\client} \A \left( \x^{(r,k)}_{m_1} - \x^{(r,k)}_{m_2} \right) - \eta_{\client} \left( \bvarepsilon^{(r,k)}_{m_1} - \bvarepsilon^{(r,k)}_{m_2} \right) \right\|_{\A^{-1}}^2 \middle| \mathcal{F}^{(r,k)} \right]
    \\
    \leq &
    \left\| \y^{(r,k)}_{m_1} - \y^{(r,k)}_{m_2} \right\|_{\A^{-1}}^2 + 2 \eta_{\client}^2 \sigma^2 \|\A\|_2^{-1}.
  \end{align*}
\end{claim}

The third claim upper bounds the growth of the second term in \cref{eq:quad:one_step_stab:claim:1} under conditional expectation. This is a result of the bounded heterogeneity assumption (\cref{a3}(c)). 
The proof of \cref{quad:one_step_stab:claim:3} is deferred to \cref{sec:proof:quad:one_step_stab:claim:3}.
\begin{claim}
  \label{quad:one_step_stab:claim:3}
  In the same settings of \cref{quad:one_step_stab}, the following inequality holds
  \begin{equation*}
    \expt \left[ \left\| \bdelta^{(r,k)}_{m_1} - \bdelta^{(r,k)}_{m_2}  \right\|_{\A^{-1}}^2  \middle| \mathcal{F}^{(r,k)} \right]
    \leq
    4 \|\A\|_2^{-1} \zeta^2.
  \end{equation*}
\end{claim}
The proof of the above claims as well as the main lemma require the following helper claim which we also state here. The proof is also deferred to \cref{sec:proof:quad:one_step_stab:claim:3}.
\begin{claim}
  \label{quad:one_step_stab:claim:4}
  In the same settings of \cref{quad:one_step_stab}, the dual norm $\|\cdot\|_*$ corresponds to the $\|\A\|_2 \cdot {\A^{-1}}$-norm, namely $    \|\y\|_* = \sqrt{ \|\A\|_2 \cdot \y^{\top} \A^{-1} \y}$. 
\end{claim}

The proof of \cref{quad:one_step_stab} is immediate once we have \cref{quad:one_step_stab:claim:1,quad:one_step_stab:claim:2,quad:one_step_stab:claim:3}.
\begin{proof}[Proof of \cref{quad:one_step_stab}]
  By  \cref{quad:one_step_stab:claim:1,quad:one_step_stab:claim:2,quad:one_step_stab:claim:3},
  \begin{align}
    & \expt \left[ \left\|\y^{(r,k+1)}_{m_1} - \y^{(r,k+1)}_{m_2} \right\|_{\A^{-1}}^2 \middle| \mathcal{F}^{(r,k)} \right]
    \nonumber \\
    \leq
    & \left( 1 + \frac{1}{K} \right) 
    \expt \left[ \left\| \y^{(r,k)}_{m_1} - \y^{(r,k)}_{m_2} - \eta_{\client} \A \left( \x^{(r,k)}_{m_1} - \x^{(r,k)}_{m_2} \right) - \eta_{\client} \left( \bvarepsilon^{(r,k)}_{m_1} - \bvarepsilon^{(r,k)}_{m_2} \right) \right\|_{\A^{-1}}^2 \middle| \mathcal{F}^{(r,k)} \right]
    \nonumber \\
    & \qquad
    + 
    \left( 1 + K \right) \eta_{\client}^2 \expt \left[ \left\| \bdelta^{(r,k)}_{m_1} - \bdelta^{(r,k)}_{m_2}  \right\|_{\A^{-1}}^2 \middle| \mathcal{F}^{(r,k)} \right]
    \tag{by \cref{quad:one_step_stab:claim:1}}
    \\
    \leq &  \left( 1 + \frac{1}{K} \right)  
    \left\| \y^{(r,k)}_{m_1} - \y^{(r,k)}_{m_2} \right\|_{\A^{-1}}^2 
    + 2 \left( 1 + \frac{1}{K} \right)   \eta^2 \sigma^2 \|\A\|_2^{-1}
    + 4 ( 1+ K) \eta_{\client}^2 \zeta^2 \|\A\|_2^{-1},
    \tag{by \cref{quad:one_step_stab:claim:2,quad:one_step_stab:claim:3}}
  \end{align}
  completing the proof of \cref{quad:one_step_stab}.
\end{proof}

The main \cref{quad:stability} then follows by telescoping \cref{quad:one_step_stab}.
\begin{proof}[Proof of \cref{quad:stability}]
Let $m_1, m_2$ be two arbitrary clients. 
Telescoping  \cref{quad:one_step_stab} from $\mathcal{F}^{(r,0)}$ to $\mathcal{F}^{(r,k)}$ gives
\begin{align}
         & \expt \left[ \left\| \y^{(r,k)}_{m_1} - \y^{(r,k)}_{m_2} \right\|_{\A^{-1}}^2 \right]
    \nonumber \\
    \leq & 
    \frac{\left( 1 + \frac{1}{K} \right)^k - 1}{\frac{1}{K}}
     \left( 2 \left( 1 + \frac{1}{K} \right)   \eta_{\client}^2 \sigma^2 \|\A\|_2^{-1}
    + 4 ( 1+ K) \eta_{\client}^2 \zeta^2 \|\A\|_2^{-1} \right) 
    \tag{telescoping \cref{quad:one_step_stab}}
    \\
    \leq & 
    (\euler - 1) K
     \left( 2 \left( 1 + \frac{1}{K} \right)   \eta_{\client}^2 \sigma^2 \|\A\|_2^{-1}
    + 4 ( 1+ K) \eta_{\client}^2 \zeta^2 \|\A\|_2^{-1} \right) 
    \tag{since $(1 + \frac{1}{K})^k \leq (1 + \frac{1}{K})^K < \euler$}
    \\
    \leq & (\euler - 1) K \left( 4 \eta_{\client}^2 \sigma^2 \|\A\|_2^{-1} + 8K \eta_{\client}^2 \zeta^2 \|\A\|_2^{-1} \right)
    \tag{since $1 + \frac{1}{K} \leq 2$ and $1 + K \leq 2K$}
    \\
    \leq & 7 \eta_{\client}^2  K \sigma^2 \|\A\|_2^{-1} + 14 \eta_{\client}^2 K^2  \zeta^2 \|\A\|_2^{-1}
    \tag{since $4 (\euler - 1) < 7$ and $8 (\euler - 1) < 14$ }
\end{align}

By convexity of $\|\cdot\|^2_{\A^{-1}}$ and \cref{quad:one_step_stab} one has
\begin{align*}
  & \frac{1}{M} \sum_{m=1}^M \expt \left[ \left\| \overline{\y^{(r,k)}} - \y^{(r,k)}_{m} \right\|_{\A^{-1}}^2 \right] 
  \leq 
  \expt \left[ \left\| \y^{(r,k)}_{m_1} - \y^{(r,k)}_{m_2} \right\|_{\A^{-1}}^2 \right]
  \\
  \leq & 
  7 \eta_{\client}^2 K \sigma^2 \|\A\|_2^{-1} + 14 \eta_{\client}^2 K^2  \zeta^2 \|\A\|_2^{-1}.
\end{align*}
Finally, we switch back to $\|\cdot\|_*$ norm following \cref{quad:one_step_stab:claim:4} 
\begin{equation*}
  \frac{1}{M} \sum_{m=1}^M \expt \left[ \left\| \overline{\y^{(r,k)}} - \y^{(r,k)}_{m} \right\|_*^2 \right] 
  \leq
  7 \eta_{\client}^2 K \sigma^2  + 14 \eta_{\client}^2 K^2  \zeta^2,
\end{equation*}
completing the proof of \cref{quad:stability}.
\end{proof}

\subsubsection{Deferred Proof of  \cref{quad:one_step_stab:claim:1}}
\label{sec:proof:quad:one_step_stab:claim:1}
\begin{proof}[Proof of \cref{quad:one_step_stab:claim:1}]
  By definition of \feddualavg procedure one has
  \begin{align}
    & \y^{(r,k+1)}_{m} = \y^{(r,k)}_m - \eta_{\client} \nabla f(\x^{(r,k)}_m; \xi^{(r,k)}_m) 
    \nonumber \\
    = & \y^{(r,k)}_m - \eta_{\client} \nabla F(\x^{(r,k)}_m) 
    + 
    \eta_{\client} \left( \nabla F_m(\x^{(r,k)}_m) - \nabla F(\x^{(r,k)}_m) \right) 
    \nonumber \\
    & +
    \eta_{\client} \left( \nabla f(\x^{(r,k)}_m;\xi^{(r,k)}_{m}) - \nabla F_m(\x^{(r,k)}_m)  \right) 
    \nonumber \\
    = &  \y^{(r,k)}_m - \eta_{\client} \nabla F(\x^{(r,k)}_m) - \eta_{\client} \bvarepsilon^{(r,k)}_m - \eta_{\client} \bdelta^{(r,k)}_m,
    \label{eq:quad:one_step_stab:1}
  \end{align}
  where the last equality is by definition of $\bvarepsilon^{(r,k)}_m$ and $\bdelta^{(r,k)}_m$.
  Therefore,
  \begin{align}
    &  \left\|\y^{(r,k+1)}_{m_1} - \y^{(r,k+1)}_{m_2} \right\|_{\A^{-1}}^2 
    \nonumber \\
    = & \left\| \y^{(r,k)}_{m_1} - \y^{(r,k)}_{m_2} - \eta_{\client} \A \left( \x^{(r,k)}_{m_1} - \x^{(r,k)}_{m_2} \right) - \eta_{\client} \left( \bvarepsilon^{(r,k)}_{m_1} - \bvarepsilon^{(r,k)}_{m_2} \right) 
    - \eta_{\client} \left( \bdelta^{(r,k)}_{m_1} - \bdelta^{(r,k)}_{m_2} \right) 
    \right\|_{\A^{-1}}^2 
    \tag{by \cref{eq:quad:one_step_stab:1}}
    \\
    = & \left\| \y^{(r,k)}_{m_1} - \y^{(r,k)}_{m_2} - \eta_{\client} \A \left( \x^{(r,k)}_{m_1} - \x^{(r,k)}_{m_2} \right) - \eta_{\client} \left( \bvarepsilon^{(r,k)}_{m_1} - \bvarepsilon^{(r,k)}_{m_2} \right) \right\|_{\A^{-1}}^2 
    + \eta_{\client}^2 \left\| \bdelta^{(r,k)}_{m_1} - \bdelta^{(r,k)}_{m_2}  \right\|_{\A^{-1}}^2 
    \nonumber \\
    & ~
    + 
    2 \left\langle \y^{(r,k)}_{m_1} - \y^{(r,k)}_{m_2} - \eta_{\client} \A \left( \x^{(r,k)}_{m_1} - \x^{(r,k)}_{m_2} \right) - \eta_{\client} \left( \bvarepsilon^{(r,k)}_{m_1} - \bvarepsilon^{(r,k)}_{m_2}  \right), \eta_{\client} \A^{-1} \left(  \bdelta^{(r,k)}_{m_1} - \bdelta^{(r,k)}_{m_2}  \right) \right\rangle.
    % \tag{expansion of $\|\cdot\|_{\A^{-1}}^2$}
    \label{eq:quad:one_step_stab:2}
  \end{align}
  By Cauchy-Schwartz inequality and AM-GM inequality one has (for any $\gamma > 0$)
  \begin{align}
    & \left\langle \y^{(r,k)}_{m_1} - \y^{(r,k)}_{m_2} - \eta_{\client} \A \left( \x^{(r,k)}_{m_1} - \x^{(r,k)}_{m_2} \right) - \eta_{\client} \left( \bvarepsilon^{(r,k)}_{m_1} - \bvarepsilon^{(r,k)}_{m_2}  \right), \eta_{\client} \A^{-1} \left(  \bdelta^{(r,k)}_{m_1} - \bdelta^{(r,k)}_{m_2}  \right) \right\rangle 
    \nonumber \\
    \leq & \left\| \y^{(r,k)}_{m_1} - \y^{(r,k)}_{m_2} - \eta_{\client} \A \left( \x^{(r,k)}_{m_1} - \x^{(r,k)}_{m_2} \right) - \eta_{\client} \left( \bvarepsilon^{(r,k)}_{m_1} - \bvarepsilon^{(r,k)}_{m_2} \right) \right\|_{\A^{-1}}
    \left\| \eta_{\client}  \left( \bdelta^{(r,k)}_{m_1} - \bdelta^{(r,k)}_{m_2} \right)  \right\|_{\A^{-1}}
    \tag{Cauchy-Schwarz inequality}
    \\
    % \leq & \frac{1}{2\gamma} \left\| \y^{(r,k)}_{m_1} - \y^{(r,k)}_{m_2} - \eta_{\client} \A \left( \x^{(r,k)}_{m_1} - \x^{(r,k)}_{m_2} \right) - \eta_{\client} \left( \bvarepsilon^{(r,k)}_{m_1} - \bvarepsilon^{(r,k)}_{m_2} \right) \right\|^2_{\A^{-1}} 
    % +
    % \frac{1}{2}  \gamma  \left\| \eta_{\client}  \left( \bdelta^{(r,k)}_{m_1} - \bdelta^{(r,k)}_{m_2} \right)  \right\|_{\A^{-1}}^2.
    % \tag{AM-GM inequality}
    % \\
    \leq & \frac{1}{2\gamma} \left\| \y^{(r,k)}_{m_1} - \y^{(r,k)}_{m_2} - \eta_{\client} \A \left( \x^{(r,k)}_{m_1} - \x^{(r,k)}_{m_2} \right) - \eta_{\client} \left( \bvarepsilon^{(r,k)}_{m_1} - \bvarepsilon^{(r,k)}_{m_2} \right) \right\|^2_{\A^{-1}} 
    \nonumber \\
    & +
    \frac{1}{2}  \gamma \eta_{\client}^2 \left\|  \left( \bdelta^{(r,k)}_{m_1} - \bdelta^{(r,k)}_{m_2} \right)  \right\|_{\A^{-1}}^2.
    \label{eq:quad:one_step_stab:3}
  \end{align}
  Plugging \cref{eq:quad:one_step_stab:3}
     to \cref{eq:quad:one_step_stab:2} with $\gamma = K$ gives
  \begin{align*}
    &  \left\|\y^{(r,k+1)}_{m_1} - \y^{(r,k+1)}_{m_2} \right\|_{\A^{-1}}^2 \leq  \left( 1 + K \right) \eta_{\client}^2 
    {\left\| \bdelta^{(r,k)}_{m_1} - \bdelta^{(r,k)}_{m_2}  \right\|_{\A^{-1}}^2}
    \\
    & \quad + \left( 1 + \frac{1}{K} \right) 
    {\left\| \y^{(r,k)}_{m_1} - \y^{(r,k)}_{m_2} - \eta_{\client} \A \left( \x^{(r,k)}_{m_1} - \x^{(r,k)}_{m_2} \right) - \eta_{\client} \left( \bvarepsilon^{(r,k)}_{m_1} - \bvarepsilon^{(r,k)}_{m_2} \right) \right\|_{\A^{-1}}^2},
  \end{align*}
  completing the proof of \cref{quad:one_step_stab:claim:1}.
\end{proof}

\subsubsection{Deferred Proof of  \cref{quad:one_step_stab:claim:2}}
\label{sec:proof:quad:one_step_stab:claim:2}
The proof technique of this claim is similar to \cite[Lemma 8]{Flammarion.Bach-COLT17} which we adapt to fit into our settings.
\begin{proof}[Proof of \cref{quad:one_step_stab:claim:2}]
  Let us first expand the $\|\cdot\|_{\A^{-1}}^2$:
\begin{align*}
   &  \left\| \y^{(r,k)}_{m_1} - \y^{(r,k)}_{m_2} - \eta_{\client} \A \left( \x^{(r,k)}_{m_1} - \x^{(r,k)}_{m_2} \right) - \eta_{\client} \left( \bvarepsilon^{(r,k)}_{m_1} - \bvarepsilon^{(r,k)}_{m_2} \right) \right\|_{\A^{-1}}^2  
   \\
  = & \left\| \y^{(r,k)}_{m_1} - \y^{(r,k)}_{m_2} \right\|_{\A^{-1}}^2 
  + \left\| \eta \A \left( \x^{(r,k)}_{m_1} - \x^{(r,k)}_{m_2} \right)  \right\|_{\A^{-1}}^2
  + \left\| \eta  \left( \bvarepsilon^{(r,k)}_{m_1} - \bvarepsilon^{(r,k)}_{m_2} \right)  \right\|_{\A^{-1}}^2 
  \\
  & \quad 
  + 2 \left\langle \eta \left( \x^{(r,k)}_{m_1} - \x^{(r,k)}_{m_2} \right),   \eta \left( \bvarepsilon^{(r,k)}_{m_1} - \bvarepsilon^{(r,k)}_{m_2} \right)  \right\rangle
  \\
  & \quad 
  + 2 \left\langle  \y^{(r,k)}_{m_1} - \y^{(r,k)}_{m_2}, - \eta \left( \x^{(r,k)}_{m_1} - \x^{(r,k)}_{m_2} \right)   \right\rangle 
  + 2 \left\langle \y^{(r,k)}_{m_1} - \y^{(r,k)}_{m_2}, - \eta \A^{-1} \left( \bvarepsilon^{(r,k)}_{m_1} - \bvarepsilon^{(r,k)}_{m_2} \right)  \right\rangle.
\end{align*}
Now we take conditional expectation. 
Note that by bounded variance assumption one has
\begin{align*}
  & \expt \left[ \left\| \eta_{\client}  \left( \bvarepsilon^{(r,k)}_{m_1} - \bvarepsilon^{(r,k)}_{m_2} \right)  \right\|_{\A^{-1}}^2 \middle| \mathcal{F}^{(r,k)} \right] 
  \\
  = &
  \|\A\|_2^{-1} \cdot \expt \left[ \left\| \eta_{\client}  \left( \bvarepsilon^{(r,k)}_{m_1} - \bvarepsilon^{(r,k)}_{m_2} \right)  \right\|_*^2 \middle| \mathcal{F}^{(r,k)} \right]
  \leq 
  2 \eta_{\client}^2 \sigma^2  \|\A\|_2^{-1},
\end{align*}
where in the first equality we applied \cref{quad:one_step_stab:claim:4}.

By unbiased and independence assumptions
\begin{equation*}
  \expt \left[ \bvarepsilon^{(r,k)}_{m_1} - \bvarepsilon^{(r,k)}_{m_2} \middle| \mathcal{F}^{(r,k)} \right] = 0.
\end{equation*}
Thus
\begin{align}
  & \expt \left[ \left\| \y^{(r,k)}_{m_1} - \y^{(r,k)}_{m_2} - \eta_{\client} \A \left( \x^{(r,k)}_{m_1} - \x^{(r,k)}_{m_2} \right) - \eta_{\client} \left( \bvarepsilon^{(r,k)}_{m_1} - \bvarepsilon^{(r,k)}_{m_2} \right) \right\|_{\A^{-1}}^2  \middle | \mathcal{F}^{(r,k)} \right]
  \nonumber \\
  \leq & 
  \left\| \y^{(r,k)}_{m_1} - \y^{(r,k)}_{m_2} \right\|_{\A^{-1}}^2   + 2 \eta_{\client}^2 \sigma^2 \|\A\|_2^{-1}
  \nonumber \\
  & \qquad
  \underbrace{+  \eta_{\client}^2 \left\|  \A \left( \x^{(r,k)}_{m_1} - \x^{(r,k)}_{m_2} \right)  \right\|_{\A^{-1}}^2}_{\text{(I)}}
  \underbrace{- 2  \eta_{\client} \left\langle  \y^{(r,k)}_{m_1} - \y^{(r,k)}_{m_2},  \x^{(r,k)}_{m_1} - \x^{(r,k)}_{m_2} \right\rangle}_{\text{(II)}} 
  \label{eq:quad:1}
\end{align}

Now we analyze (I), (II) in \cref{eq:quad:1}. First note that
\begin{align}
  & 
  \text{(I)} = \eta_{\client}^2 \left\|  \A \left( \x^{(r,k)}_{m_1} - \x^{(r,k)}_{m_2} \right)  \right\|_{\A^{-1}}^2
  \nonumber \\
  = & \eta_{\client}^2 \left\langle \x^{(r,k)}_{m_1} - \x^{(r,k)}_{m_2} , \A \left( \x^{(r,k)}_{m_1} - \x^{(r,k)}_{m_2} \right)  \right\rangle 
  \tag{by definition of $\|\cdot\|_{\A^{-1}}^2$}
  \\
  = & \eta_{\client} \left\langle \x^{(r,k)}_{m_1} - \x^{(r,k)}_{m_2} , \eta_{\client} \left( \nabla F(\x^{(r,k)}_{m_1}) - \nabla F(\x^{(r,k)}_{m_2}) \right) \right\rangle
  \tag{since $F$ is quadratic}
  \\
  = & \eta_{\client} \left\langle \x^{(r,k)}_{m_1} - \x^{(r,k)}_{m_2} ,  \nabla (\eta_{\client} F - 2 h) (\x^{(r,k)}_{m_1}) - \nabla (\eta_{\client} F - 2h)(\x^{(r,k)}_{m_2})  \right\rangle
  \nonumber \\
  & + 2 \eta_{\client} \left\langle \x^{(r,k)}_{m_1} - \x^{(r,k)}_{m_2} ,  \nabla h (\x^{(r,k)}_{m_1}) - \nabla h(\x^{(r,k)}_{m_2})  \right\rangle
  \nonumber 
\end{align}
By $L$-smoothness of $F_m$ (\cref{a1}(c)) we know that $F := \frac{1}{M} \sum_{m=1}^M F_m$ is also $L$-smooth. Thus $\eta_{\client} F$ is $\frac{1}{4}$-smooth since $\eta_{\client} \leq \frac{1}{4L}$. Thus $\eta_{\client} F - 2h$ is concave since $h$ is $1$-strongly convex, which implies
\begin{equation*}
   \left\langle \x^{(r,k)}_{m_1} - \x^{(r,k)}_{m_2} ,  \nabla (\eta_{\client} F - 2 h) (\x^{(r,k)}_{m_1}) - \nabla (\eta_{\client} F - 2h)(\x^{(r,k)}_{m_2})  \right\rangle \leq 0.
\end{equation*}
We obtain
\begin{equation}
  \text{(I)} \leq 2 \eta_{\client} \left\langle \x^{(r,k)}_{m_1} - \x^{(r,k)}_{m_2} ,  \nabla h (\x^{(r,k)}_{m_1}) - \nabla h(\x^{(r,k)}_{m_2})  \right\rangle.
  \label{eq:quad:2}
\end{equation}

Now we study (I)+(II) in \cref{eq:quad:1}:
\begin{align}
  & \text{(I) + (II)} =  
   \eta_{\client}^2 \left\|  \A \left( \x^{(r,k)}_{m_1} - \x^{(r,k)}_{m_2} \right)  \right\|_{\A^{-1}}^2
   - 2  \eta_{\client} \left\langle  \y^{(r,k)}_{m_1} - \y^{(r,k)}_{m_2},  \x^{(r,k)}_{m_1} - \x^{(r,k)}_{m_2} \right\rangle 
  \nonumber \\
  \leq & 
  2 \eta_{\client} \left\langle  \x^{(r,k)}_{m_1} - \x^{(r,k)}_{m_2}  ,  \nabla h (\x^{(r,k)}_{m_1}) - \nabla h(\x^{(r,k)}_{m_2})   \right\rangle
  - 2  \eta_{\client} \left\langle  \x^{(r,k)}_{m_1} - \x^{(r,k)}_{m_2} , \y^{(r,k)}_{m_1} - \y^{(r,k)}_{m_2} \right\rangle 
  \tag{by inequality \cref{eq:quad:2}}
  \\
  = &   - 2  \eta_{\client} \left\langle  \x^{(r,k)}_{m_1} - \x^{(r,k)}_{m_2} , \left(\y^{(r,k)}_{m_1} - \nabla h(\x^{(r,k)}_{m_1}) \right) - \left(\y^{(r,k)}_{m_2} - \nabla h(\x^{(r,k)}_{m_2}) \right)  \right\rangle 
  \label{eq:quad:3}
\end{align}

On the other hand, by definition of $\x^{(r,k)}_m$ we have
\begin{equation*}
  \x^{(r,k)}_{m} = \nabla (h + (rK + k)\eta_{\client} \psi )^* (\y^{(r,k)}_m)
  =
  \argmin_{\x} \left\{ \left\langle -\y^{(r,k)}_m, \x^{\star}  \right\rangle + (rK+k)\eta_{\client} \psi (\x^{\star}) + h(\x^{\star})   \right\}.
\end{equation*}
By subdifferential calculus one has
\begin{equation*}
  \y^{(r,k)}_m - \nabla h(\x^{(r,k)}_m)  \in \partial \left[ \eta_c (rK+k) \psi(\x^{(r,k)}_m) \right].
\end{equation*}
By monotonicity of subgradients one has
\begin{equation}
  \left\langle  \x^{(r,k)}_{m_1} - \x^{(r,k)}_{m_2} , \left(\y^{(r,k)}_{m_1} - \nabla h(\x^{(r,k)}_{m_1}) \right) - \left(\y^{(r,k)}_{m_2} - \nabla h(\x^{(r,k)}_{m_2}) \right)  \right\rangle \geq 0.
  \label{eq:quad:4}
\end{equation}
Combining \cref{eq:quad:3,eq:quad:4} gives
\begin{equation}
  \text{(I) + (II)} \leq 0.
  \label{eq:quad:5}
\end{equation}
Combining \cref{eq:quad:1,eq:quad:5} completes the proof as 
\begin{align*}
  & \expt \left[ \left\| \y^{(r,k)}_{m_1} - \y^{(r,k)}_{m_2} - \eta_{\client} \A \left( \x^{(r,k)}_{m_1} - \x^{(r,k)}_{m_2} \right) - \eta_{\client} \left( \bvarepsilon^{(r,k)}_{m_1} - \bvarepsilon^{(r,k)}_{m_2} \right) \right\|_{\A^{-1}}^2  \middle | \mathcal{F}^{(r,k)} \right]
  \\
  \leq &
  \left\| \y^{(r,k)}_{m_1} - \y^{(r,k)}_{m_2} \right\|_{\A^{-1}}^2 
  + 2 \eta_{\client}^2 \sigma^2  \|\A\|_2^{-1}.
\end{align*}
\end{proof}

\subsubsection{Deferred Proof of  \cref{quad:one_step_stab:claim:3,quad:one_step_stab:claim:4}}
\label{sec:proof:quad:one_step_stab:claim:3}
\begin{proof}[Proof of \cref{quad:one_step_stab:claim:3}]
By triangle inequality and AM-GM inequality,
\begin{align}
  & \expt \left[  \left\|\bdelta^{(r,k)}_{m_1} - \bdelta^{(r,k)}_{m_2} \right\|_{\A^{-1}}^2 | \mathcal{F}^{(r,k)} \right]  
  \nonumber \\
  \leq &
  \expt \left[ \left( \| \bdelta^{(r,k)}_{m_1} \|_{\A^{-1}} + \| \bdelta^{(r,k)}_{m_2} \|_{\A^{-1}} \right)^2  \middle| \mathcal{F}^{(r,k)} \right]\tag{triangle inequality} 
  \\
  \leq & 2 \expt \left[ \| \bdelta^{(r,k)}_{m_1} \|_{\A^{-1}}^2 + \| \bdelta^{(r,k)}_{m_2} \|_{\A^{-1}}^2  \middle| \mathcal{F}^{(r,k)} \right].
  \tag{AM-GM inequality}
\end{align}

By \cref{quad:one_step_stab:claim:4}, 
\begin{equation*}
  \expt \left[  \left\|\bdelta^{(r,k)}_{m_1} - \bdelta^{(r,k)}_{m_2} \right\|_{\A^{-1}}^2 \middle| \mathcal{F}^{(r,k)} \right]  
  \leq
  2 \|\A\|_2^{-1} \expt \left[ \| \bdelta^{(r,k)}_{m_1} \|_*^2 + \| \bdelta^{(r,k)}_{m_2} \|_*^2  \middle| \mathcal{F}^{(r,k)} \right]
  \leq
  4 \|\A\|_2^{-1} \zeta^2,
\end{equation*}
where the last inequality is due to bounded heterogeneity  \cref{a3}(c). This completes the proof of \cref{quad:one_step_stab:claim:3}.
\end{proof}

\begin{proof}[Proof of \cref{quad:one_step_stab:claim:4}]
  Since the primal norm $\|\cdot\|$ is $(\|\A\|_2^{-1} \cdot \A)$-norm by \cref{a3}(b), the dual norm $\|\cdot\|_*$ is $ \left( \|\A\|_2^{-1} \cdot \A \right)^{-1} = \|\A\|_2 \cdot \A^{-1}$-norm.
\end{proof}

\subsection{Details of Step 3: Finishing the Proof of \cref{thm:2}}
\label{sec:feddualavg:step3}
With \cref{quad:stability} at hand, we are ready to prove \cref{thm:2}.
\begin{proof}[Proof of \cref{thm:2}]
  Applying \cref{pia:general,quad:stability} one has
  \begin{align}
    & \expt \left[  \Phi \left(  \frac{1}{KR} \sum_{r=0}^{R-1} \sum_{k=1}^K \widehat{\x^{(r,k)}} \right) - \Phi(\x^{\star}) \right] 
    \nonumber \\
    \leq
    & \frac{1}{\eta_{\client} KR } D_h(\x^{\star}, \x^{(0,0)}) 
    + \frac{\eta_{\client}  \sigma^2}{M}
    + \frac{L}{M KR }   \left[ \sum_{r=0}^{R-1}  \sum_{k=0}^{K-1} \sum_{m=1}^M   \expt\left \| \overline{\y^{(r,k)}} - \y^{(r,k)}_m \right\|_*^2 \right] 
    \tag{by \cref{pia:general}}
    \\
    \leq & 
    \frac{1}{\eta_{\client} KR } D_h(\x^{\star}, \x^{(0,0)}) 
    + \frac{\eta_{\client}  \sigma^2}{M}
    + L \cdot \left(  7 \eta_{\client}^2 K \sigma^2 
    + 14 \eta_{\client}^2 K^2  \zeta^2 \right)
    \tag{by \cref{quad:stability}}
    \\
    = &   \frac{B^2}{\eta_{\client} KR }
    + \frac{\eta_{\client}  \sigma^2}{M}
    + 7 \eta_{\client}^2 L K \sigma^2
    + 14 \eta_{\client}^2 L K^2  \zeta^2,
    \label{eq:proof:thm:2:1}
  \end{align}
  which gives the first inequality in \cref{thm:2}.

  Now set 
  \begin{equation*}
    \eta_{\client} = \min \left\{ \frac{1}{4L}, 
    \frac{M^{\frac{1}{2}} B}{\sigma K^{\frac{1}{2}} R^{\frac{1}{2}}} ,
    \frac{B^{\frac{2}{3}}}{L^{\frac{1}{3}} K^{\frac{2}{3}} R^{\frac{1}{3}} \sigma^{\frac{2}{3}}}, 
    \frac{B^{\frac{2}{3}}}{L^{\frac{1}{3}} K R^{\frac{1}{3}} \zeta^{\frac{2}{3}}}  \right\}.
  \end{equation*}  
  We have
  \begin{equation*}
    \frac{B^2}{\eta_{\client} KR }
    \leq
    \max \left\{  \frac{4L B^2}{KR} 
    ,
    \frac{\sigma B}{M^{\frac{1}{2}} K^{\frac{1}{2}} R^{\frac{1}{2}}}
    ,
    \frac{L^{\frac{1}{3}} B^{\frac{4}{3}} \sigma^{\frac{2}{3}}}{K^{\frac{1}{3}} R^{\frac{2}{3}}}
    ,
    \frac{L^{\frac{1}{3}} B^{\frac{4}{3}} \zeta^{\frac{2}{3}}}{R^{\frac{2}{3}}}\right\},
  \end{equation*}
  and 
  \begin{equation*}
    \frac{\eta_{\client} \sigma^2}{M} \leq  \frac{\sigma B}{M^{\frac{1}{2}} K^{\frac{1}{2}} R^{\frac{1}{2}}}
    ,
    \qquad
    7 \eta_{\client}^2 L K \sigma^2 \leq  \frac{7L^{\frac{1}{3}} B^{\frac{4}{3}} \sigma^{\frac{2}{3}}}{K^{\frac{1}{3}} R^{\frac{2}{3}}}
    ,
    \qquad
    14 \eta_{\client}^2 L K^2  \zeta^2 \leq   \frac{14L^{\frac{1}{3}} B^{\frac{4}{3}} \zeta^{\frac{2}{3}}}{R^{\frac{2}{3}}}.
  \end{equation*}
  Consequently
  \begin{equation*}
    \frac{B^2}{\eta_{\client} KR }
    + \frac{\eta_{\client}  \sigma^2}{M}
    + 7 \eta_{\client}^2 L K \sigma^2
    + 14 \eta_{\client}^2 L K^2  \zeta^2
    \leq
    \frac{4L B^2}{KR} 
    +
    \frac{2\sigma B}{M^{\frac{1}{2}} K^{\frac{1}{2}} R^{\frac{1}{2}}}
    +
    \frac{8L^{\frac{1}{3}} B^{\frac{4}{3}} \sigma^{\frac{2}{3}}}{K^{\frac{1}{3}} R^{\frac{2}{3}}}
    +
    \frac{15L^{\frac{1}{3}} B^{\frac{4}{3}} \zeta^{\frac{2}{3}}}{R^{\frac{2}{3}}},
  \end{equation*}
  completing the proof of \cref{thm:2}.
\end{proof}

% !TEX root = main.tex  
\section{Numerical Experiments}
\label{sec:fco:expr}
In this section, we validate our theory and demonstrate the efficiency of the algorithms via numerical experiments. 

\subsection{General Setup}
\paragraph{Algorithms.} 
In this section we mostly compare \feddualavg (see \cref{alg:feddualavg}) with \fedmid (see \cref{alg:fedmid}) since the latter serves a natural baseline.
We do not present subgradient-\fedavg in this section due to its consistent ineffectiveness, as demonstrated in \cref{fig:haxby:simplified} (marked \fedavg($\partial$)). 

To examine the necessity of client proximal step, we also test two less-principled versions of \fedmid and \feddualavg, in which the proximal steps are only performed on the server-side.
We refer to these two versions as \textsc{FedMiD-OSP} and \textsc{FedDualAvg-OSP}, where ``OSP'' stands for ``only server proximal,''.
%  with pseudo-code provided in \cref{sec:addl_expr_setup}.
% We provide the complete setup details in \cref{sec:additional_expr}, including but not limited to hyper-parameter tuning, dataset processing and evaluation metrics.
% 
% In this paper we mainly test four Federated algorithms, namely \fedmidfull (\fedmid, see \cref{alg:fedmid}), \feddualavgfull (\feddualavg, see \cref{alg:feddualavg}), as well as two less-principled algorithms which skip the client-side proximal operations. 
% We refer to these two algorithms as \fedmid-OSP and \feddualavg-OSP, where ``OSP'' stands for ''only server proximal''. 
We formally state these two OSP algorithms in \cref{alg:fedmid-osp,alg:feddualavg-osp}. 
We study these two OSP algorithms mainly for ablation study purpose, thouse they might be of special interest if the proximal step is computationally intensive. 
For instance, in \fedmid-OSP, the client proximal step is replaced by $\x^{(r,k+1)}_m \gets \nabla h^*(\nabla h (\x^{(r,k)}_m) - \eta_{\client} \g^{(r,k)}_m)$ with no $\psi$ involved (see line 8 of \cref{alg:fedmid-osp}). This step reduces to the ordinary SGD $\x^{(r,k+1)}_m \gets \x^{(r,k)}_m - \eta_{\client} \g^{(r,k)}_m$ if $h(\x) = \frac{1}{2}\| \x \|_2^2$ in which case both $\nabla h$ and $\nabla h^*$ are identity mapping.
% However, we stress that there is \textbf{no} theoretical guarantee on the convergence of either \fedmid-OSP or \feddualavg-OSP.
Theoretically, it is not hard to establish similar rates of \cref{thm:0} for \fedmid-OSP with finite $\psi$. For infinite $\psi$, we need extension of $f$ outside $\textbf{dom}~\psi$ to fix regularity. To keep this thesis focused, we will not establish these results formally.

% !TEX root = main.tex  
\begin{algorithm}
  \caption{\fedmidfull Only Server Proximal (\fedmid-OSP)}
  \begin{algorithmic}[1]
    \label{alg:fedmid-osp}
    \STATE {\textbf{procedure}} \fedmid-OSP ($\x^{(0,0)}, \eta_{\client}, \eta_{\server}$)
      \FOR{$r = 0, \ldots, R-1$}
        \STATE sample a subset of clients $\mathcal{S}^{(r)} \subseteq [M]$
        \FORALL {$m \in \mathcal{S}^{(r)}$ {\bf in parallel}}
          \STATE client initialization $\x^{(r,0)}_m \gets \x^{(r, 0)}$
          \COMMENT{Broadcast primal initialization for round $r$}
          \FOR{$k = 0, \ldots, K-1$}
            \STATE $\g^{(r,k)}_m \gets \nabla f(\x^{(r,k)}_m; \xi^{(r,k)}_m)$
            \COMMENT{Query gradient}
            \STATE $\x^{(r,k+1)}_m \gets \nabla h^*(\nabla h (\x^{(r,k)}_m) - \eta_{\client} \g^{(r,k)}_m$)
            \STATE \COMMENT{Client (primal) update -- proximal operation skipped}
          \ENDFOR
        \ENDFOR
        \STATE $\bDelta^{(r)} = \frac{1}{|\mathcal{S}^{(r)}|} \sum_{m \in \mathcal{S}^{(r)}} (\x^{(r, K)}_m - \x^{(r, 0)}_m)$
        \COMMENT{Compute pseudo-anti-gradient}
        \STATE $\x^{(r+1, 0)} \gets \nabla (h + \eta_{\server}\eta_{\client}K \psi)^*(\nabla h(\x^{(r,0)}) + \eta_{\server} \bDelta^{(r)})$
        \COMMENT{Server (primal) update}
      \ENDFOR
  \end{algorithmic}
\end{algorithm}
% !TEX root = main.tex  
\begin{algorithm}
  \caption{\feddualavgfull Only Server Proximal (\feddualavg-OSP)}
  \begin{algorithmic}[1]
      \label{alg:feddualavg-osp}
    \STATE \textbf{procedure} {\feddualavg-OSP}($\x^{(0,0)}, \eta_{\client}, \eta_{\server}$)
      \STATE server initialization $\y^{(0,0)} \gets \nabla h(\x^{(0,0)})$
      \FOR{$r = 0, \ldots, R-1$}
        \STATE sample a subset of clients $\mathcal{S}^{(r)} \subseteq [M]$
        \FORALL{$m \in \mathcal{S}^{(r)}$ {\bf in parallel}}
          \STATE client initialization $\y^{(r,0)}_m \gets \y^{(r,0)}$ 
           \COMMENT{Broadcast dual initialization for round $r$}
          \FOR{$k = 0, \ldots, K-1$}
            \STATE $\x^{(r,k)}_m \gets \nabla h^*(\y^{(r,k)}_m)$
            \COMMENT{Compute primal point $\x^{(r,k)}_m$ -- proximal operation skipped}
            \STATE $\g^{(r,k)}_m \gets \nabla f (\x^{(r,k)}_m; \xi^{(r,k)}_m) $
            \COMMENT{Query gradient}
            \STATE $\y^{(r,k+1)}_m \gets \y^{(r,k)}_m - \eta_{\client} \g^{(r,k)}_m$
            \COMMENT{Client (dual) update}
          \ENDFOR
        \ENDFOR
        \STATE 
        $\bDelta^{(r)} = \frac{1}{|\mathcal{S}^{(r)}|} \sum_{m \in \mathcal{S}^{(r)}} ( \y^{(r, K)}_m - \y^{(r, 0)}_m)$
        \COMMENT{Compute pseudo-anti-gradient}
        \STATE $\y^{(r+1, 0)} \gets \y^{(r,0)} + \eta_{\server} \bDelta^{(r)}$
        \COMMENT{Server (dual) update}
        \STATE $\x^{(r+1, 0)} \gets \nabla (h + \eta_{\server} \eta_{\client} (r+1) K \psi)^* (\y^{(r+1, 0)})$
        \COMMENT{(Optional) Compute server primal state}
      \ENDFOR
  \end{algorithmic}
\end{algorithm}

\paragraph{Environment.} We simulate the algorithms in the TensorFlow Federated (TFF) framework \cite{Ingerman.Ostrowski-19}. 
The implementation is based on the Federated Research repository available at \url{https://github.com/google-research/federated}.
The source code is available at \url{https://bit.ly/fco-icml21}.

\paragraph{Tasks.} We experiment the following four tasks in this work. 
\begin{enumerate}
    \item Federated Lasso ($\ell_1$-regularized least squares) for sparse feature selection, see \cref{sec:fco:expr:lasso}.
    \item Federated low-rank matrix recovery via nuclear-norm regularization, see \cref{sec:fco:expr:nuclear}.
    \item Federated sparse ($\ell_1$-regularized) logistic regression for fMRI dataset \cite{Haxby-01}, see \cref{sec:fco:expr:fmri}.
    \item Federated constrained optimization for Federated EMNIST dataset \cite{Caldas.Duddu.ea-NeurIPS19}, see \cref{sec:emnist}.
\end{enumerate}
We take the distance-generating function $h$ to be $h(\x) := \frac{1}{2}\|\x\|_2^2$ for all the four tasks.
The detailed setups of each experiment are stated in the corresponding subsections.
  
\subsection{Task 1: Federated LASSO for Sparse Feature Recovery}
  \label{sec:fco:expr:lasso}
  \subsubsection{Setup}
  \label{mainsec:fco:expr:lasso}
  In this subsection, we consider the LASSO ($\ell_1$-regularized least-squares) problem on a synthetic  dataset, motivated by models from biomedical and signal processing literature~(e.g., \cite{Ryali.Supekar.ea-10,Chen.Lin.ea-NIPS12}).   
  \begin{equation}
      \min_{\x \in \reals^d, x^0 \in \reals} ~~ \frac{1}{M} \sum_{m=1}^M \expt_{(\a,b) \sim \mathcal{D}_m} (\a^{\top} \x +  x^0 - b)_2^2 + \lambda \|\x\|_1.
      \label{eq:flasso}
  \end{equation}
  The goal is to recover the sparse signal $\x$ from noisy observations $(\a,b)$.

  \paragraph{Synthetic Dataset Descriptions.}
  To generate the synthetic dataset, we first fix a sparse ground truth $\x_{\mathrm{real}} \in \reals^{d}$ and $x^0_{\mathrm{real}} \in \reals$, and then sample the dataset $(\a, b)$ following $b = \a^{\top} \x_{\mathrm{real}} + x^0_{\mathrm{real}} + \varepsilon$ for some noise $\varepsilon$. 
  We let the distribution of $(\a, b)$ vary over clients to synthesize the heterogeneity. 

  Specifically, we first generate the ground truth $\x_{\mathrm{real}}$ with $d_1$ ones and $d_0$ zeros for some $d_1 + d_0 = d$, namely
  \begin{equation*}
     \x_{\mathrm{real}}  = \begin{bmatrix}
    \textbf{1}_{d_1}
    \\
    \textbf{0}_{d_0}
  \end{bmatrix} \in \reals^{d},
  \end{equation*}
  and ground truth $x^0_{\mathrm{real}} \sim \mathcal{N}(0,1)$. 
  
  The observations $(\a, b)$ are generated as follows to simulate the heterogeneity among clients. 
  Let $(\a_{m}^{(i)}, b_{m}^{(i)})$ denotes the $i$-th observation of the $m$-th client. 
  For each client $m \in [M]$, we first generate and fix the mean $\mu_m \sim \mathcal{N}(0, \I_{d \times d})$. 
  Then we sample $n_m$ pairs of observations following
  \begin{align*}
    \a_m^{(i)} = \boldsymbol{\mu}_m + \boldsymbol{\delta}_m^{(i)}, \quad & \text{where $\delta_m^{(i)} \sim \mathcal{N}(\mathbf{0}_d, \I_{d \times d})$ are \text{i.i.d.}, for $i=1,\ldots,n_m$;}
    \\
    b_m^{(i)} = \x_{\mathrm{real}}^\top \a_m^{(i)} + x^0_{\mathrm{real}} + \varepsilon_m^{(i)}, \quad & \text{where $\varepsilon_m^{(i)} \sim \mathcal{N}(0,1)$ are i.i.d., for $i=1,\ldots,n_m$}.
  \end{align*}
  
  We test four configurations of the above synthetic dataset.
  \begin{enumerate}
    \item [(I)] The ground truth $\x_{\mathrm{real}}$ has $d_1 = 512$ ones and $d_0 = 512$ zeros. 
    We generate $M=64$ training clients where each client possesses $128$ pairs of samples.
    There are 8,192 training samples in total.
    \item [(II)] (sparse ground truth) The ground truth $\x_{\mathrm{real}}$ has $d_1 = 64$ ones and $d_0 = 960$ zeros. The rest of the configurations are the same as dataset (I).
    \item [(III)] (sparser ground truth) The ground truth $\x_{\mathrm{real}}$ has $d_1 = 8$ ones and $d_0 = 1016$ zeros. The rest of the configurations are the same as dataset (I).
    \item [(IV)] (more distributed data) The ground truth is the same as (I). We generate $M=256$ training clients where each client possesses $32$ pairs of samples. The total number of training examples are the same.
  \end{enumerate}
  
  \paragraph{Evaluation Metrics.}
  Since the ground truth $\x_{\mathrm{real}}$ of the synthetic dataset is known, we can evaluate the quality of the sparse features retrieved by comparing it with the ground truth.
  To numerically evaluate the sparsity, we treat all the features in $w$ with absolute values smaller than $10^{-2}$ as zero elements, and non-zero otherwise.
  We evaluate the performance by recording precision, recall, sparsity density, and F1-score. 
  
  \paragraph{Hyperparameters.}
  For all algorithms, we tune the client learning rate $\eta_{\client}$ and server learning rate $\eta_{\server}$ only. 
  We test 49 different combinations of $\eta_{\client}$ and $\eta_{\server}$.  
  $\eta_{\client}$ is selected from $\{0.001, 0.003, 0.01, 0.03, 0.1, 0.3, 1 \}$, and $\eta_{\server}$ is selected from $\{0.01, 0.03, 0.1, 0.3, 1, 3, 10\}$. 
  All methods are tuned to achieve the best averaged recovery error (in F1-score) over the last 100 communication rounds.
  We claim that the best learning rate combination falls in this range for all the algorithms tested.
  We draw 10 clients uniformly at random at each communication round and let the selected clients run local algorithms with batch size 10 for one epoch (of its local dataset) for this round. 
  We run 500 rounds in total, though \feddualavg usually converges to almost perfect solutions in much fewer rounds.
  We select $\lambda$ so that the centralized solver (on gathered data) can successfully recover the sparse pattern. 
  
  \subsubsection{Results on Synthetic Dataset (I)}
  We present the result for the synthetic dataset (I) in \cref{fig:nuclear_row32_rank16_cl64_1x4}.
  The best learning rates configuration is $\eta_{\client} = 0.01, \eta_{\server} = 1$ for  \feddualavg, and  $\eta_{\client} = 0.001, \eta_{\server} = 0.3$ for other algorithms (including \fedmid). This matches our theory that \feddualavg can benefit from larger learning rates.
  \begin{figure}[ht]
    \centering
    \includegraphics[width=\columnwidth]{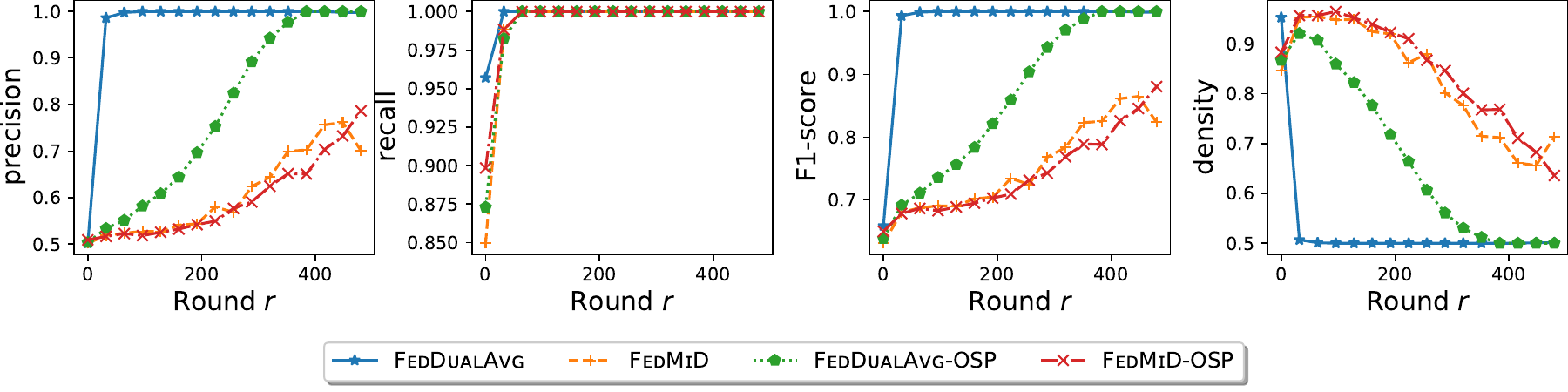}
    \caption{
    \textbf{Sparsity recovery on a synthetic LASSO problem with 50\% sparse ground truth.} 
    Observe that \feddualavg not only identifies most of the sparsity pattern but also is fastest. 
  %   Apart from the expected good performance of \feddualavg, 
    It is also worth noting that the less-principled \textsc{FedDualAvg-OSP} is also very competitive.
    The poor performance of \fedmid can be attributed to the ``curse of primal averaging'', as the server averaging step ``smooths out'' the sparsity pattern, which is corroborated empirically by the least sparse solution obtained by \fedmid.
    }
    \label{fig:lasso_p1024_nnz512_cl64_1x4}
  \end{figure}

  \subsubsection{Results on Synthetic Dataset (II) and (III) with Sparser Ground Truth}
  \label{sec:fco:expr:lasso:2:3}
\begin{figure}[!hbp]
    \centering
     \includegraphics[width=\textwidth]{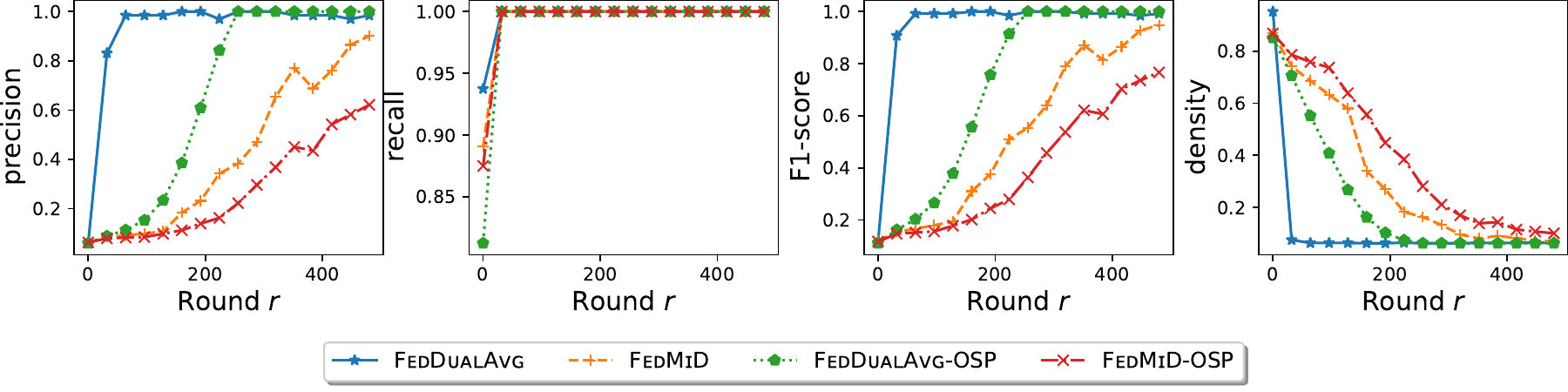}
    \caption{\textbf{Results on Dataset (II): $1/2^{4}$ Ground Truth Density.} See \cref{sec:fco:expr:lasso:2:3} for discussions.}
    \label{fig:lasso_p1024_nnz64_cl64_1x4}
  \end{figure}

\begin{figure}[!hbp]
    \centering
    \includegraphics[width=\textwidth]{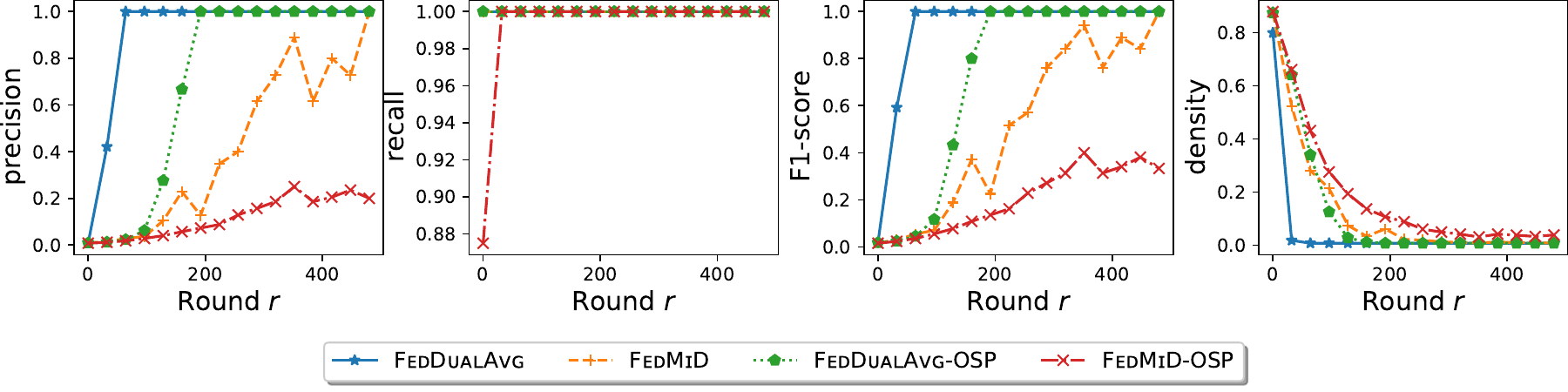}
    \caption{\textbf{Results on Dataset (III): $1/2^{7}$ Ground Truth Density.} See \cref{sec:fco:expr:lasso:2:3} for discussions.}
    \label{fig:lasso_p1024_nnz8_cl64_1x4}
  \end{figure}

  We repeat the experiments on the dataset (II) and (III) with $1/2^{4}$ and $1/{2^7}$ ground truth density, respectively.
  The results are shown in \cref{fig:lasso_p1024_nnz64_cl64_1x4,fig:lasso_p1024_nnz8_cl64_1x4}.
  We observe that \feddualavg converges to the perfect F1-score in less than 100 rounds, which outperforms the other baselines by a margin. 
  The F1-score of \feddualavg-OSP converges faster on these sparser datasets than (I), which makes it comparably more competitive.
  The convergence of \fedmid and \fedmid-OSP remains slow.
  
  \subsubsection{Results on Synthetic Dataset (IV): More Distributed Data (256 clients)}
  \label{sec:fco:expr:lasso:4}
\begin{figure}[!hbp]
    \centering
    \includegraphics[width=\textwidth]{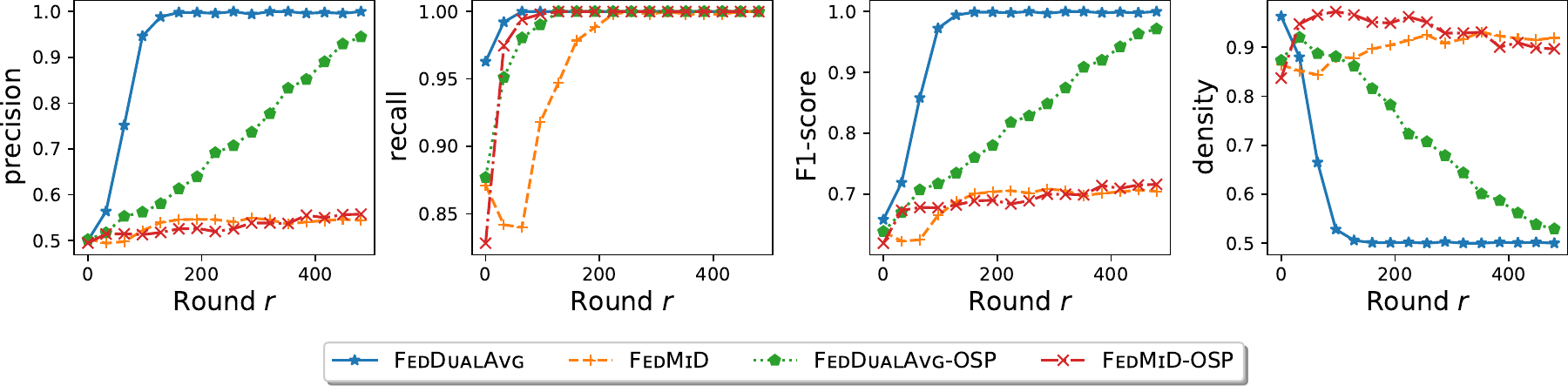}
    \caption{\textbf{Results on Dataset (IV): More Distributed Data.} See \cref{sec:fco:expr:lasso:4} for discussions.}
    \label{fig:lasso_p1024_nnz512_cl256_1x4}
  \end{figure}

  We repeat the experiments on the dataset (IV) with more distributed data (256 clients). 
  The results are shown in \cref{fig:lasso_p1024_nnz512_cl256_1x4}. 
  We observe that all the four algorithms take more rounds to converge in that each client has fewer data than the previous configurations. 
  \feddualavg manages to find perfect F1-score in less than 200 rounds, which outperforms the other algorithms significantly.
  \feddualavg-OSP can recover an almost perfect F1-score after 500 rounds but is much slower than on the less distributed dataset (I).
  \fedmid and \fedmid-OSP have very limited progress within 500 rounds. 
  This is because the server averaging step in \fedmid and \fedmid-OSP fails to aggregate the sparsity patterns properly. 
  Since each client is subject to larger noise due to the limited amount of local data, simply averaging the primal updates will ``smooth out'' the sparsity pattern.
  
\subsection{Task 2: Federated Low-Rank Matrix Estimation via Nuclear-Norm Regularization}
  \label{sec:fco:expr:nuclear}
  \subsubsection{Setup}
  In this subsection, we consider a low-rank matrix estimation problem via the nuclear-norm regularization
  \begin{equation}
    \min_{\X \in \reals^{d_1 \times d_2}, x^0 \in \reals}
     \frac{1}{M} \sum_{m=1}^M \mathbb{E}_{(\A, b) \sim \mathcal{D}_m} \left( \langle \A, \X \rangle + x^0 - b \right)^2 + \lambda \|W\|_{\mathrm{nuc}},
    \label{eq:fed:matrix:estimation}
  \end{equation}
  where $\| \X \|_{\mathrm{nuc}} := \sum_i \sigma_i(\X)$ denotes the nuclear norm (a.k.a. trace norm) defined by the summation of all the singular values. 
  The goal is to recover a low-rank matrix $\X$ from noisy observations $(\A, b)$. 
  This formulation captures a variety of problems such as low-rank matrix completion and recommendation systems  \cite{Candes.Recht-09}.
  Note that the proximal operator with respect to the nuclear-norm regularizer $\|\cdot\|_{\mathrm{nuc}}$ reduces to singular-value thresholding operation \cite{Cai.Candes.ea-SIOPT10}.
  
  \paragraph{Synthetic Dataset Descriptions.}
  We evaluate the algorithms on a synthetic federated dataset with known low-rank ground truth $\X_{\mathrm{real}}$, similar to the above LASSO experiments. Specifically, we generate the following ground truth $\X_{\mathrm{real}} \in \reals^{d \times d}$ of rank $r$ 
  \begin{equation*}
    \X_{\mathrm{real}} = \begin{bmatrix}
      \I_{r \times r} & \zeros_{r \times (d - r)} \\
      \zeros_{(d-r) \times r} & \zeros_{(d-r) \times (d-r)}
    \end{bmatrix},
  \end{equation*}
  and ground truth $x^0_{\mathrm{real}} \sim \mathcal{N}(0,1)$. 
  
  The observations $(\A, b)$ are generated as follows to synthesize the heterogeneity among clients. 
  Let $(\A_{m}^{(i)}, b_{m}^{(i)})$ denotes the $i$-th observation of the $m$-th client. 
  For each client $m$, we first generate and fix the mean $\boldsymbol{\mu}_m \in \reals^{d \times d}$ where all coordinates are i.i.d. standard Gaussian $\mathcal{N}(0,1)$. 
  Then we sample $n_m$ pairs of observations following
  \begin{align*}
    & \A_m^{(i)} = \boldsymbol{\mu}_m + \bdelta_m^{(i)}, \text{where $\bdelta_m^{(i)} \in \reals^{d \times d}$ is a matrix with all coordinates from standard Gaussian;}
    \\
    & b_m^{(i)} = \langle \A_m^{(i)},  \X_{\mathrm{real}} \rangle + x^0_{\mathrm{real}} + \varepsilon_m^{(i)}, \text{where $\varepsilon_m^{(i)} \sim \mathcal{N}(0,1)$ are i.i.d.}
  \end{align*}

  We tested four configurations of the above synthetic dataset.
  \begin{enumerate}
    \item [(I)] The ground truth $\X_{\mathrm{real}}$ is a matrix of dimension $32 \times 32$ with rank $r=16$. 
    We generate $M=64$ training clients where each client possesses $128$ pairs of samples.
    There are 8,192 training samples in total.
    \item [(II)] (rank-4 ground truth) The ground truth $\X_{\mathrm{real}}$ has rank $r=4$. The other configurations are the same as the dataset (I).
    \item [(III)] (rank-1 ground truth) The ground truth $\X_{\mathrm{real}}$ has rank $r=1$. The other configurations are the same as the dataset (I).
    \item [(IV)] (more distributed data) The ground truth is the same as (I). We generate $M=256$ training clients where each client possesses $32$ samples. The total number of training examples remains the same.
  \end{enumerate}

  \paragraph{Evaluation Metrics.}
  We focus on four metrics for this task: the training (regularized) loss, the validation mean-squared-error, the recovered rank, and the recovery error in Frobenius norm $\|\X_{\mathrm{output}} - \X_{\mathrm{real}}\|_{\mathrm{F}}$.
  To numerically evaluate the rank, we count the number of singular values that are greater than $10^{-2}$.
  
  \paragraph{Hyperparameters.}
  For all algorithms, we tune the client learning rate $\eta_{\client}$ and server learning rate $\eta_{\server}$ only. 
  We test 49 different combinations of $\eta_{\client}$ and $\eta_{\server}$.  
  $\eta_{\client}$ is selected from $\{0.001, 0.003, 0.01, 0.03, 0.1, 0.3, 1 \}$, and $\eta_{\server}$ is selected from $\{0.01, 0.03, 0.1, 0.3, 1, 3, 10\}$. 
  All methods are tuned to achieve the best averaged recovery error on the last 100 communication rounds.
  We claim that the best learning rate combination falls in this range for all algorithms tested.
  We draw 10 clients uniformly at random at each communication round and let the selected clients run local algorithms with batch size 10 for one epoch (of its local dataset) for this round. 
  We run 500 rounds in total, though \feddualavg usually converges to perfect F1-score in much fewer rounds.
  We also record the results obtained by the deterministic solver on centralized data, marked as \texttt{optimum}. 

\subsubsection{Results on Synthetic Dataset (I)}
\begin{figure}[ht]
    \centering
    \includegraphics[width=\columnwidth]{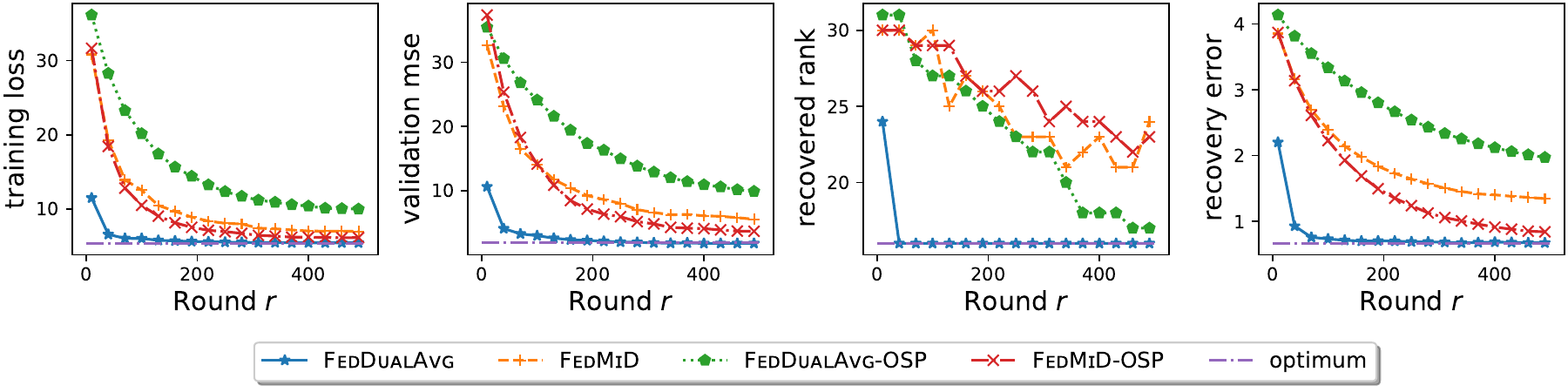}
    \caption{
      \textbf{Low-rank matrix estimation comparison on a synthetic dataset with the ground truth of rank 16.}
  We observe that \feddualavg finds the solution with exact rank in less than 100 communication rounds.
  \fedmid and \fedmid-OSP converge slower in loss and rank.
  The unprincipled \feddualavg-OSP can generate low-rank solutions but is far less accurate. 
    }
    \label{fig:nuclear_row32_rank16_cl64_1x4}
  \end{figure}

  The results for the synthetic dataset (I) are presented in \cref{fig:nuclear_row32_rank16_cl64_1x4}.

\subsubsection{Results on Synthetic Dataset (II) and (III) with Ground Truth of Lower Rank}
\label{sec:fco:expr:nuclear:2:3}
\begin{figure}[ht]
  \centering
  \includegraphics[width=\textwidth]{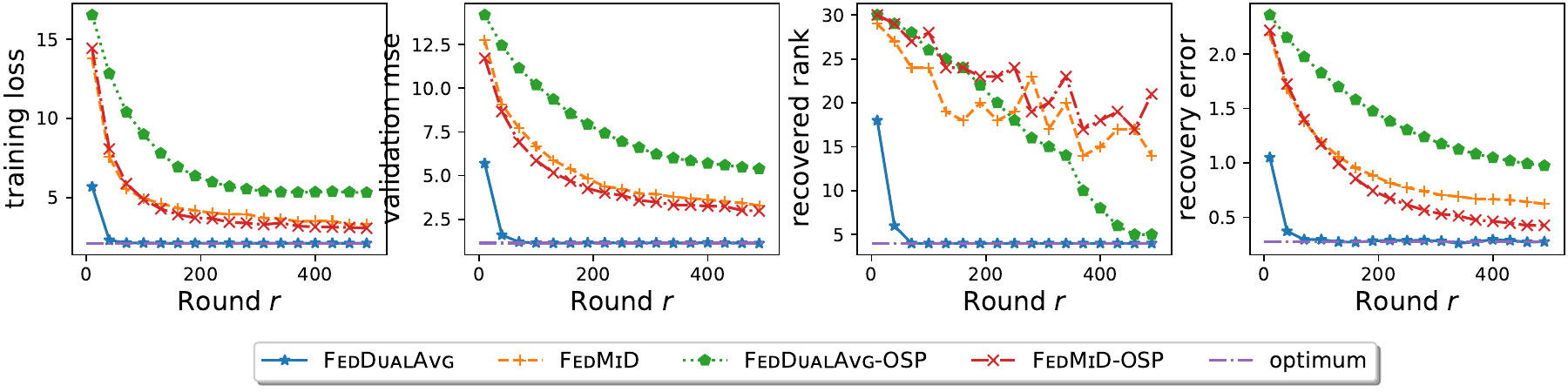}
  \caption{\textbf{Results on Dataset (II): Ground Truth Rank 4.} See \cref{sec:fco:expr:nuclear:2:3} for discussions.}
  \label{fig:nuclear_row32_rank4_cl64_1x4}
\end{figure}

\begin{figure}[ht]
  \centering
  \includegraphics[width=\textwidth]{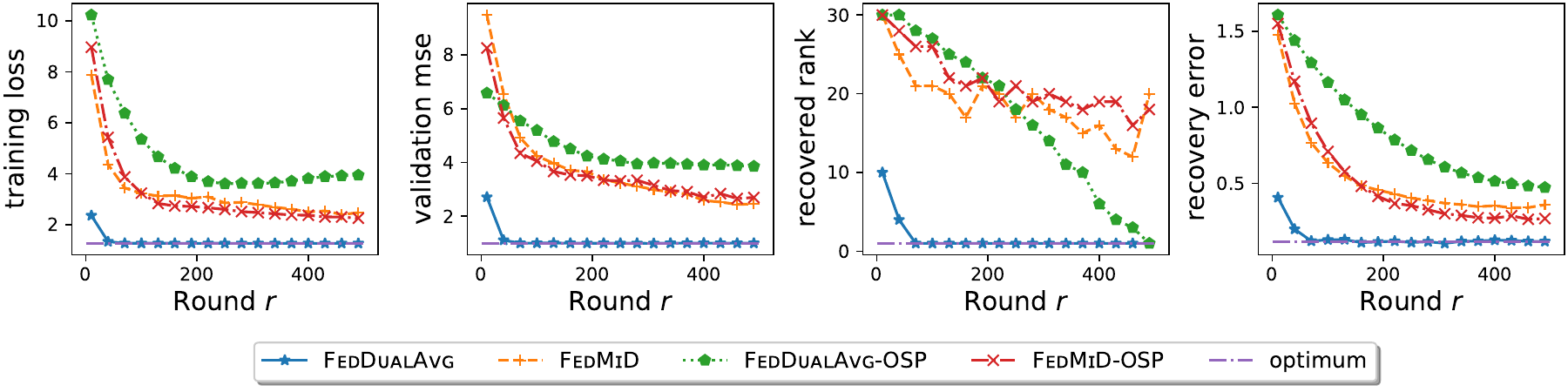}
  \caption{\textbf{Results on Dataset (III): Ground Truth Rank 1.} See \cref{sec:fco:expr:nuclear:2:3} for discussions.}
  \label{fig:nuclear_row32_rank1_cl64_1x4}
\end{figure}

We repeat the experiments on the dataset (II) and (III) with 4 and 1 ground truth rank, respectively.
The results are shown in \cref{fig:nuclear_row32_rank4_cl64_1x4,fig:nuclear_row32_rank1_cl64_1x4}. 
The results are qualitatively reminiscent of the previous experiments on the dataset (I). 
\feddualavg can recover the exact rank in less than 100 rounds, which outperforms the other baselines by a margin. 
\feddualavg-OSP can recover a low-rank solution but is less accurate.
The convergence of \fedmid and \fedmid-OSP remains slow.
\subsubsection{Results on Synthetic Dataset (IV): More Distributed Data (256 clients)}
\label{sec:fco:expr:nuclear:4}
\begin{figure}[ht]
    \centering
    \includegraphics[width=\textwidth]{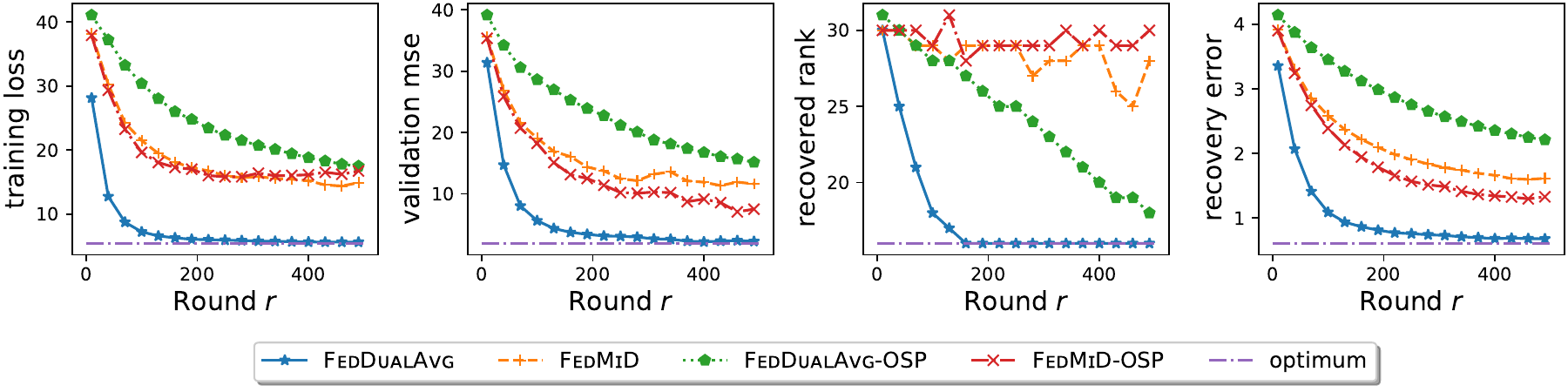}
    \caption{\textbf{Results on Dataset (IV): More Distributed Data.} See \cref{sec:fco:expr:nuclear:4} for discussions.}
    \label{fig:nuclear_row32_rank16_cl256_1x4}
\end{figure}

We repeat the experiments on the dataset (IV) with more distributed data. 
The results are shown in \cref{fig:nuclear_row32_rank16_cl256_1x4}. 
We observe that all four algorithms take more rounds to converge in that each client has fewer data than the previous configurations. 
The other messages are qualitatively similar to the previous experiments -- 
\feddualavg manages to find exact rank in less than 200 rounds, which outperforms the other algorithms significantly.
% This is because the server averaging step in \fedmid and \fedmid-OSP fails to aggregate the sparsity patterns properly. 
% Since each client is subject to larger noise due to the limited amount of local data, simply averaging the primal updates will ``smooth out'' the sparsity pattern.

\subsection{Task 3: Sparse Logistic Regression for fMRI Scan}
  \label{sec:fco:expr:fmri}
  \subsubsection{Setup}
  In this subsection, we consider the cross-silo setup of learning a binary classifier on fMRI scans.
  For this purpose, we use the data collected by \cite{Haxby-01}, to understand the pattern of response in the ventral temporal (vt) area of the brain given a visual stimulus. 
%   There were six subjects doing image recognition in a block-design experiment over 11 to 12 sessions, with a total of 71 sessions. 
%   Each session consists of 18 fMRI scans under the stimuli of a picture of either a house or a face.
%   We use the \texttt{nilearn} package \cite{Abraham.Pedregosa.ea-14} to normalize and transform the four-dimensional raw fMRI scan data into an array with 39,912 volumetric pixels (voxels) using the standard  mask.
  We plan to learn a sparse ($\ell_1$-regularized) binary logistic regression on the voxels to classify the stimuli given the voxels input.
  Enforcing sparsity is crucial for this task as it allows domain experts to understand which part of the brain is differentiating between the stimuli.
%   We select five (out of six) subjects as the training set and the last subject as the held-out validation set. 
%   We treat each session as a client, with a total of 59 training clients and 12 validation clients, where each client possesses the voxel data of 18 scans.
%   As in the previous experiment, we tune the client learning rate $\eta_{\client}$ and server learning rate $\eta_{\server}$ only. 
%   We set the $\ell_1$-regularization strength to be $10^{-3}$.
%   For each setup, we run the federated algorithms for 300 communication rounds.

\paragraph{Dataset Descriptions and Preprocessing.}
  We use data collected by \cite{Haxby-01}. 
  There were 6 subjects doing binary image recognition (from a horse and a face) in a block-design experiment over 11-12 sessions per subject, in which each session consists of 18 scans.
  We use \texttt{nilearn} package \cite{Abraham.Pedregosa.ea-14} to normalize and transform the 4-dimensional raw fMRI scan data into an array with 39,912 volumetric pixels (voxels) using the standard mask.
  We choose the first 5 subjects as training set and the last subject as validation set.
  To simulate the cross-silo federated setup, we treat each session as a client. There are 59 training clients and 12 test clients, where each client possesses the voxel data of 18 scans.
  
\paragraph{Evaluation Metrics.} We focus on three metrics for this task: validation (regularized) loss, validation accuracy, and (sparsity) density.
  To numerically evaluate the density, we treat all weights with absolute values smaller than $10^{-4}$ as zero elements.
  The density is computed as non-zero parameters divided by the total number of parameters.

\paragraph{Hyperparameters.} For all algorithms, we adjust only client learning rate $\eta_{\client}$ and server learning rate $\eta_{\server}$. 
For each federated setup, we tested 49 different combinations of $\eta_{\client}$ and $\eta_{\server}$.  $\eta_{\client}$ is selected from $\{0.001, 0.003, 0.01, 0.03, 0.1, 0.3, 1\}$, and $\eta_{\server}$ is selected from $\{0.01, 0.03, 0.1, 0.3, 1, 3, 10\}$. 
We let each client run its local algorithm with batch-size one for one epoch per round. 
At the beginning of each round, we draw 20 clients uniformly at random.
We run each configuration for 300 rounds and present the configuration with the lowest validation (regularized) loss at the last round. 

\subsubsection{Experimental Results}
\begin{figure}[!hbp]
    \centering
    \includegraphics[width=\columnwidth]{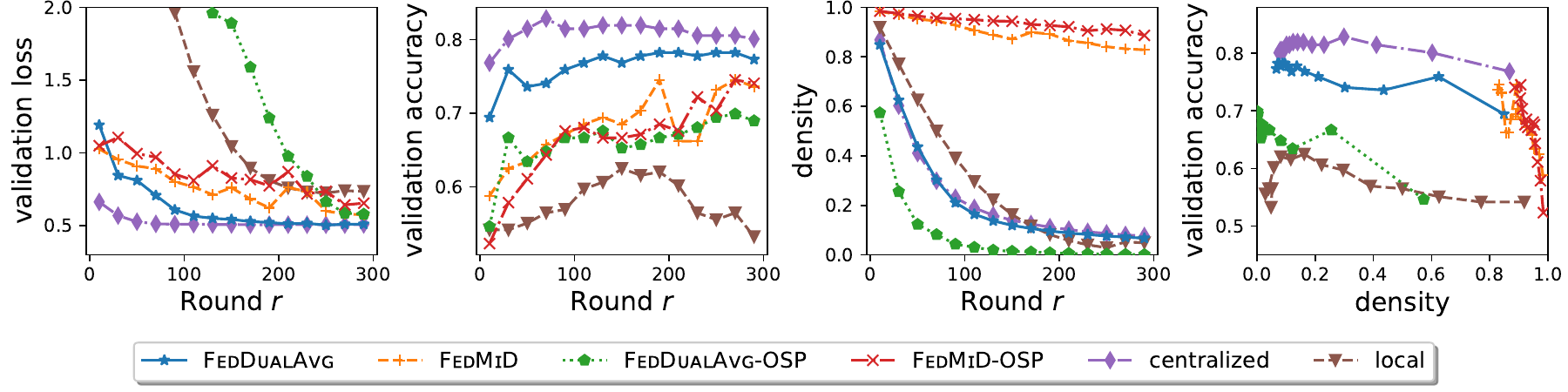}
    \caption{
      \textbf{Results on $\ell_1$-regularized logistic regression for fMRI data from \cite{Haxby-01}}. 
      %  We observe that the result of \feddualavg is comparable with the \texttt{centralized} baseline on gathered data and significantly outperforms the \texttt{local} baseline on isolated data.
      We observe that $\feddualavg$ yields sparse and accurate solutions that are comparable with the centralized baseline. 
      \fedmid and \fedmid-OSP provides denser solutions that are relatively less accurate.
      The unprincipled \feddualavg-OSP can provide sparse solutions but far less accurate.
    }
    \label{fig:haxby59_lambd_1e-03_1x4}
  \end{figure}

We compare the algorithms with two non-federated baselines: 
  1) \texttt{centralized} corresponds to training on the centralized dataset gathered from \textbf{all} the training clients; 2) \texttt{local} corresponds to training on the local data from only \textbf{one} training client without communication. 
  We run proximal gradient descent for these two baselines for 300 epochs.
  The learning rate is tuned from $\{0.0001, 0.0003, 0.001, 0.003, 0.01, 0.03, 0.1, 0.3, 1\}$ to attain the best validation loss at the last epoch.
  The results are shown in \cref{fig:haxby59_lambd_1e-03_1x4}. 

  The results demonstrate that \feddualavg not only recovers sparse and accurate solutions, but also behaves most robust to learning-rate configurations.

  % We also tested two non-federated baselines for comparison, marked as \texttt{centralized} and \texttt{local}. 
% \texttt{centralized} corresponds to training on the centralized dataset gathered from \textbf{all} the 59 training clients.
% \texttt{local} corresponds to training on the local data from only \textbf{one} training client without communication. 
% We run proximal gradient descent for these two baselines for 300 epochs.
% The learning rate is tuned from $\{0.0001, 0.0003, 0.001, 0.003, 0.01, 0.03, 0.1, 0.3, 1\}$ to attain the best validation loss at the last epoch.
% The goal is to understand the pattern of response in ventral temporal area of the brain given a visual stimulus.
% Enforcing sparsity is important as it allows domain experts to understand which part of the brain is differentiating between the stimuli.
% We apply $\ell_1$-regularized logistic regression on the voxels to classify the visual stimuli.

\subsubsection{Progress Visualization across Various Learning Rate Configurations}
\label{sec:lr:config}
\begin{figure}[!hbp]
    \centering
    \includegraphics[width=\textwidth]{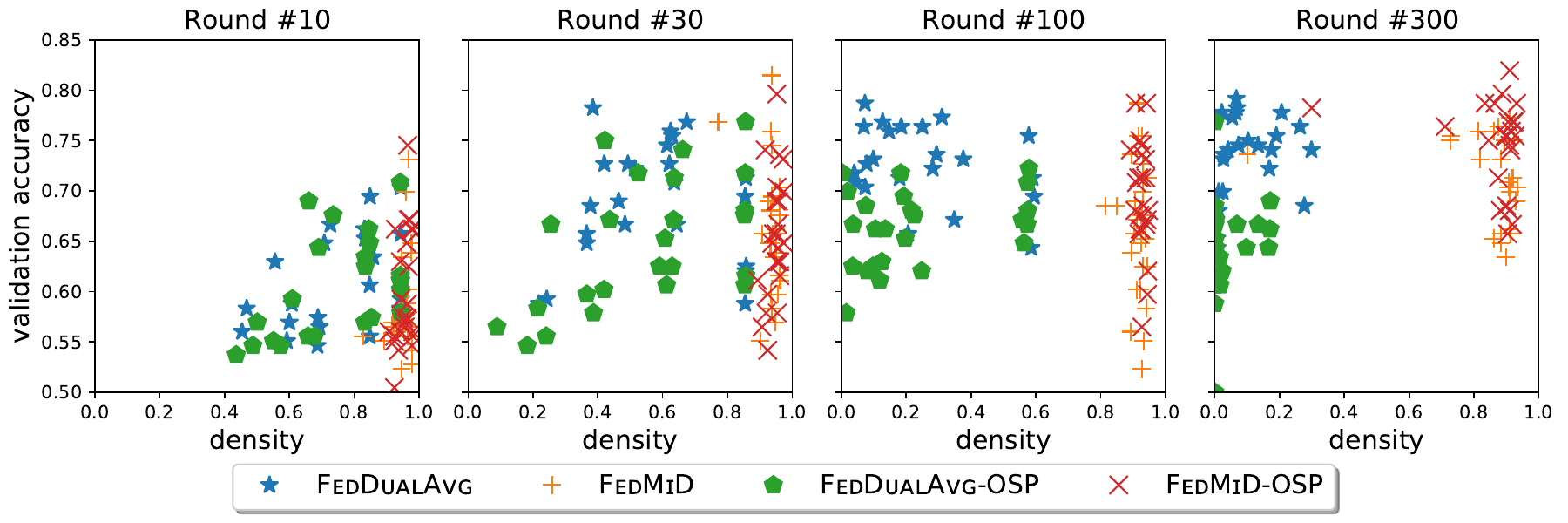}
    \caption{\textbf{Progress of Federated Algorithms Under Various Learning Rate Configurations for fMRI.} Each dot stands for a learning rate configuration (client and server). \feddualavg recovers sparse and accurate solutions, and is robust to learning-rate configurations.
    }
    \label{fig:haxby59_lambd_1e-03_multi_1x4}
\end{figure}

In this subsection, we present an alternative viewpoint to visualize the progress of federated algorithms and understand the robustness to hyper-parameters.
To this end, we run four algorithms for various learning rate configurations (we present all the combinations of learning rates mentioned above such that $\eta_{\client} \eta _{\server} \in [0.003, 0.3]$) and record the validation accuracy and (sparsity) density after 10th, 30th, 100th, and 300th round. 
The results are presented in \cref{fig:haxby59_lambd_1e-03_multi_1x4}. 
Each dot stands for a learning rate configuration (client and server). 
We can observe that most \feddualavg configurations reach the upper-left region of the box, which indicates sparse and accurate solutions. 
\feddualavg-OSP reaches to the mid-left region of the box, which indicates sparse but less accurate solutions.
The majority of \fedmid and \fedmid-OSP lands on the right side region box, which reflects the hardness for \fedmid and \fedmid-OSP to find the sparse solutions.

\subsection{Task 4: Constrained Federated Optimization for Federated EMNIST}
\label{sec:emnist}
\subsubsection{Setup Details}
In this task we test the performance of the algorithms when the composite term $\psi$ is taken to be convex characteristics
$\chi_{\cstr}(\x) := 
  \begin{cases} 
      0 & \text{if $w \in \mathcal{C}$}, \\
      +\infty & \text{if $w \notin \mathcal{C}$}.
  \end{cases}
$
which encodes a hard constraint. 
  
\paragraph{Dataset Descriptions and Models.} 
We tested on the Federated EMNIST (FEMNIST) dataset provided by TensorFlow Federated, which was derived from the Leaf repository \cite{Caldas.Duddu.ea-NeurIPS19}. 
EMNIST is an image classification dataset that extends MNIST dataset by incorporating alphabetical classes. 
The Federated EMNIST dataset groups the examples from EMNIST by writers.

We tested two versions of FEMNIST in this work:
\begin{enumerate}
    \item[(I)] FEMNIST-10: digits-only version of FEMNIST which contains 10 label classes. %There are 3,383 clients, with a total of 40,832 examples.
    We experiment the logistic regression models with $\ell_1$-ball-constraint or $\ell_2$-ball-constraint on this dataset. 
    Note that for this task we only trained on 10\% of the examples in the original FEMNIST-10 dataset because the original FEMNIST-10 has an unnecessarily large number (340k) of examples for the logistic regression model.
    \item[(II)] FEMNIST-62: full version of FEMNIST which contains 62 label classes (including 52 alphabetical classes and 10 digital classes). 
     We test a two-hidden-layer fully connected neural network model where all fully connected layers are simultaneously subject to $\ell_1$-ball-constraint. 
    Note that there is no theoretical guarantee for either of the four algorithms on non-convex objectives. We directly implement the algorithms as if the objectives were convex. 
    We defer the study of \fedmid and \feddualavg for non-convex objectives to the future work.
    % There are 3,400 clients, with a total of 671,585 examples.
\end{enumerate}

\paragraph{Evaluation Metrics.}
We focused on three metrics for this task: training error, training accuracy, and test accuracy. 
Note that the constraints are always satisfied because all the trajectories of all the four algorithms are always in the feasible region.

\paragraph{Hyperparameters.}
For all algorithms, we tune only the client learning rate $\eta_{\client}$ and server learning rate $\eta_{\server}$. For each setup, we tested 25 different combinations of $\eta_{\client}$ and $\eta_{\server}$.  $\eta_{\client}$ is selected from $\{0.001, 0.003, 0.01, 0.03, 0.1\}$, and $\eta_{\server}$ is selected from $\{0.01, 0.03, 0.1, 0.3, 1\}$. 
We draw 10 clients uniformly at random at each communication round and let the selected clients run local algorithms with batch size 10 for 10 epochs (of its local dataset) for this round. 
We run 5,000 communication rounds in total and evaluate the training loss every 100 rounds.
All methods are tuned to achieve the best averaged training loss on the last 10 checkpoints.

\subsubsection{Experimental Results}
\begin{figure}
    \centering
    \includegraphics[width=\textwidth]{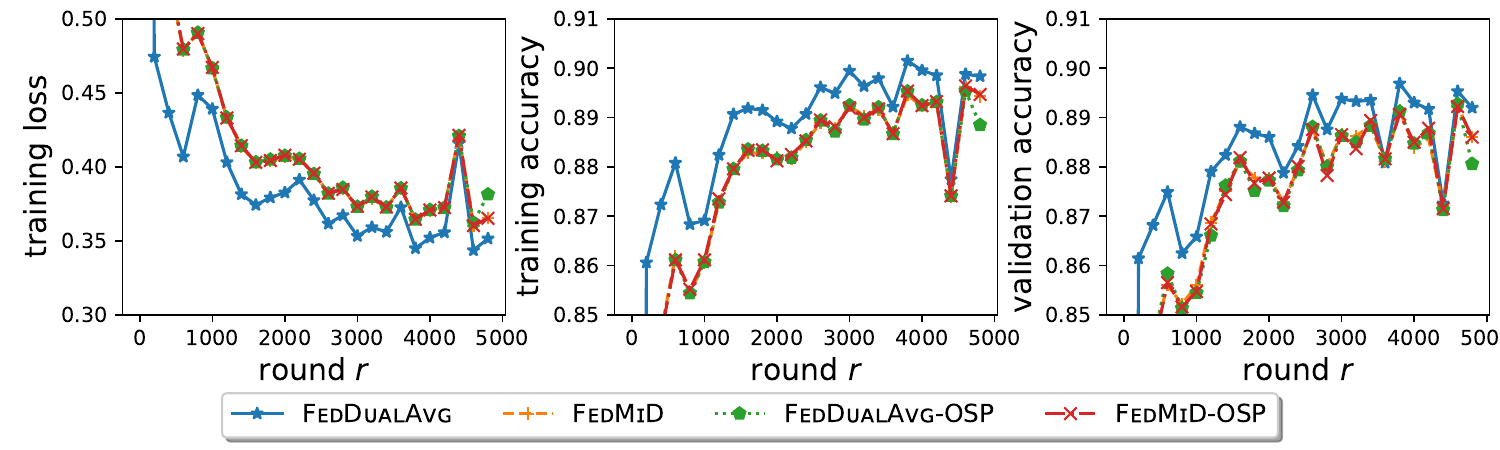}
    \caption{
    \textbf{$\ell_1$-Constrained logistic regression.}
    Dataset: FEMNIST-10.
    Constraint: $\|\x\|_1 \leq 1000$.
    }
    \label{fig:emnist-lr-l1-1000}
\end{figure}

\paragraph{$\ell_1$-Constrained Logistic Regression}
We first test the $\ell_1$-regularized logistic regression. The results are shown in \cref{fig:emnist-lr-l1-1000}. 
We observe that \feddualavg outperforms the other three algorithms by a margin. 
Somewhat surprisingly, we observe that the other three algorithms behave very closely in terms of the three metrics tested. 
This seems to suggest that the client proximal step (in this case projection step) might be saved in \fedmid. 

\begin{figure}
    \centering
    \includegraphics[width=\textwidth]{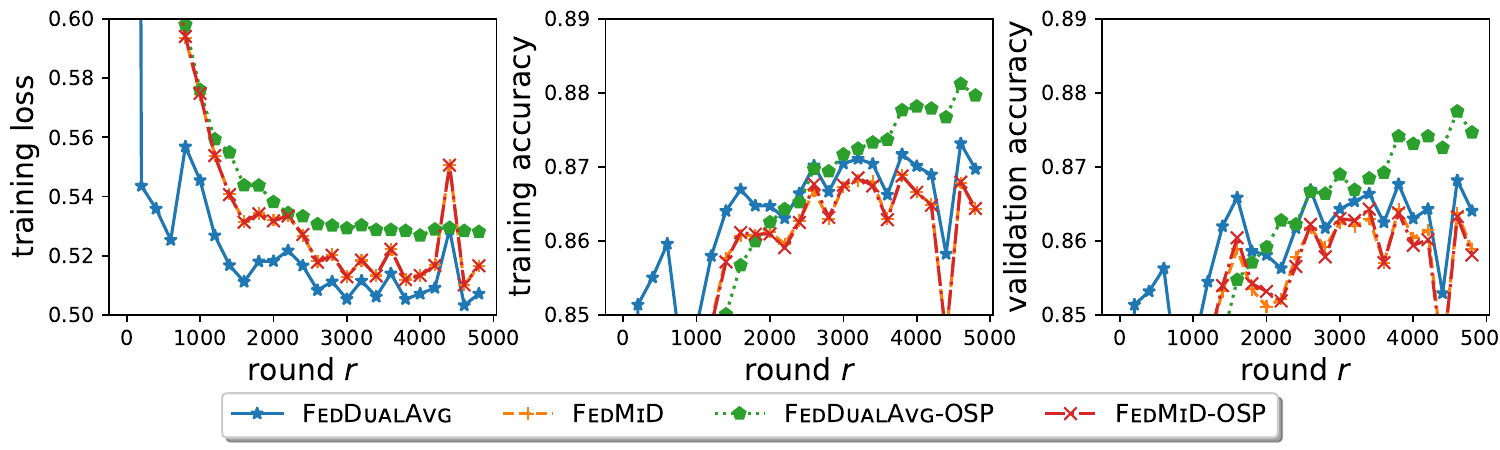}
    \caption{
    \textbf{$\ell_2$-constrained logistic regression.} 
    Dataset: FEMNIST-10.
    Constraint: $\|\x\|_2 \leq 10$.
    }
    \label{fig:emnist-lr-l2-10}
\end{figure}

\paragraph{$\ell_2$-Constrained Logistic Regression}
Next, we test the $\ell_2$-regularized logistic regression. The results are shown in \cref{fig:emnist-lr-l2-10}. 
We observe that \feddualavg outperforms the \fedmid and \fedmid-OSP in all three metrics (note again that \fedmid and \fedmid-OSP share very similar trajectories). 
Interestingly, the \feddualavg-OSP behaves much worse in training loss than the other three algorithms, but the training accuracy and validation accuracy are better. 
We conjecture that this effect might be attributed to the homogeneous property of $\ell_2$-constrained logistic regression which \feddualavg-OSP can benefit from.

\begin{figure}
    \centering
    \includegraphics[width=\textwidth]{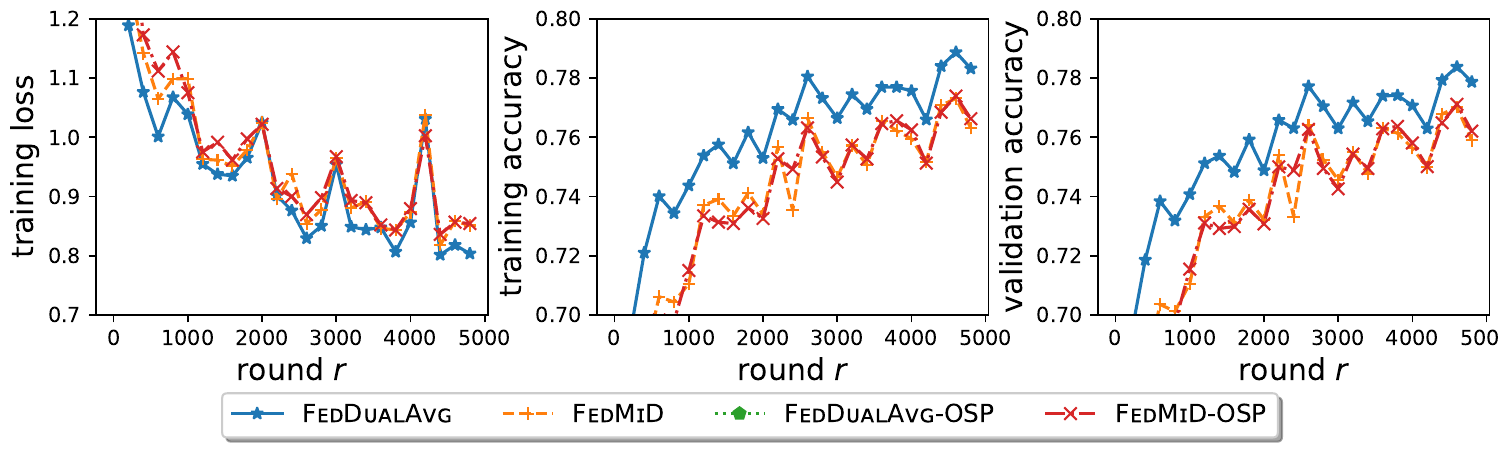}
    \caption{
    \textbf{$\ell_1$-Constrained Two-Hidden-Layer Neural Network.}
  Dataset: FEMNIST-62.
  Constraint: all three dense kernels 
    $\x^{[l]}$ simultaneously satisfy $\|\x^{[l]}\|_1 \leq 1000$.
    }
    \label{fig:emnist-2NN-l1-1000}
\end{figure}

\paragraph{$\ell_1$-Constrained Two-Hidden-Layer Neural Network}
Finally, we test on the two-hidden-layer neural network with $\ell_1$-constraints. 
The results are shown in \cref{fig:emnist-2NN-l1-1000}.
We observe that \feddualavg outperforms \fedmid and \fedmid-OSP in all three metrics (once again, note that \fedmid and \fedmid-OSP share similar trajectories). 
On the other hand, \feddualavg-OSP behaves much worse (which is out of the plotting ranges). 
This is not quite surprising because \feddualavg-OSP does not have any theoretical guarantees.

\appendix
\chapter{Appendix of Chapter 2}
\section{Deferred Proof in \cref{sec:fedavg:2o:ub}}
\label{sec:proof:thm:fedavg:2o:ub}

\subsection{Deferred Proof of \cref{lem:fedavg:2o:ub:1}}
\label{sec:pf:lem:fedavg:2o:ub:1}
We restate the lemma for the readers' convenience. The proof is adapted from \cite{Woodworth.Patel.ea-NeurIPS20} which we include only for completeness.
\LemFedAvgSecondOrderUBFirst*
\begin{proof}[Proof of \cref{lem:fedavg:2o:ub:1}]
    Since $\overline{\x^{(r,k+1)}} = \overline{\x^{(r,k)}} - \eta \cdot \frac{1}{M} \sum_{m=1}^M \nabla f(\x_m^{(r,k)}; \xi_m^{(r,k)})$, by parallelogram law
    \begin{align}
        & \frac{1}{M} \sum_{m=1}^M \left\langle \nabla f (\x_m^{(r,k)}; \xi_m^{(r,k)}), \overline{\x^{(r,k+1)}} - \x^{\star}  \right\rangle 
        \nonumber
        \\
        = &
        \frac{1}{2\eta} \left( \left\|\overline{\x^{(r,k)}} - \x^{\star} \right\|_2^2 
        - \left\| \overline{\x^{(r,k+1)}} -  \overline{\x^{(r,k)}}  \right\|_2^2 
        - \left\|\overline{\x^{(r,k+1)}} - \x^{\star} \right\|_2^2  \right).
        \label{eq:lem:fedavg:2o:ub:1:1}
    \end{align}
    By convexity and $L$-smoothness of $F_m$, one has
    \begin{align}
        & F_m ( \overline{\x^{(r,k+1)}} ) 
        \leq
        F_m ( \x_m^{(r,k)})  + \left\langle \nabla F_m ( \x_m^{(r,k)}), \overline{\x^{(r,k+1)}} - \x_m^{(r,k)} \right\rangle + \frac{L}{2} \left\|  \overline{\x^{(r,k+1)}} -  \x_m^{(r,k)}  \right\|_2^2 
        \tag{$L$-smoothness}
        \\
        \leq &  F_m ( \x^{\star})  + \left\langle \nabla F_m ( \x_m^{(r,k)}), \overline{\x^{(r,k+1)}} - \x^{\star} \right\rangle
        + \frac{L}{2} \left\|  \overline{\x^{(r,k+1)}} -  \x_m^{(r,k)}  \right\|_2^2 
        \tag{convexity}
        \\
        \leq &  F_m ( \x^{\star})  + \left\langle \nabla F_m ( \x_m^{(r,k)}), \overline{\x^{(r,k+1)}} - \x^{\star} \right\rangle
        + L \left\|  \overline{\x^{(r,k+1)}} -  \overline{\x^{(r,k)}} \right\|_2^2 
        + L \left\| \x_m^{(r,k)} -  \overline{\x^{(r,k)}} \right\|_2^2 ,
        \label{eq:lem:fedavg:2o:ub:1:2}
    \end{align}
    where the inequality is by AM-GM. Combining \cref{eq:lem:fedavg:2o:ub:1:1,eq:lem:fedavg:2o:ub:1:2} yields
    \begin{align}
        & F ( \overline{\x^{(r,k+1)}} ) - F(\x^{\star}) = \frac{1}{M} \sum_{m=1}^M \left(  F_m ( \overline{\x^{(r,k+1)}} ) - F(\x^{\star})  \right)
        \nonumber
        \\
    \leq &  \frac{1}{M} \sum_{m=1}^M \left\langle  \nabla F_m ( \x_m^{(r,k)}), \overline{\x^{(r,k+1)}} - \x^{\star} \right\rangle 
    + {L} \left\|  \overline{\x^{(r,k+1)}} -  \overline{\x^{(r,k)}} \right\|_2^2 
    + \frac{L}{M} \sum_{m=1}^M \left\| \x_m^{(r,k)} -  \overline{\x^{(r,k)}} \right\|_2^2 
    \nonumber
    \\
    \leq  &
    \frac{1}{M} \sum_{m=1}^M \left\langle  \nabla F_m ( \x_m^{(r,k)}) - \nabla f (\x_m^{(r,k)}; \xi_m^{(r,k)}), \overline{\x^{(r,k+1)}} - \x^{\star} \right\rangle 
    \nonumber
    \\
    & 
    + {L} \left\|  \overline{\x^{(r,k+1)}} -  \overline{\x^{(r,k)}} \right\|_2^2 
    + \frac{L}{M} \sum_{m=1}^M \left\| \x_m^{(r,k)} -  \overline{\x^{(r,k)}} \right\|_2^2 
    \nonumber
    \\
        & +   \frac{1}{2\eta} \left( \left\|\overline{\x^{(r,k)}} - \x^{\star} \right\|_2^2 
        - \left\|\overline{\x^{(r,k+1)}} - \x^{\star} \right\|_2^2 
        - \left\| \overline{\x^{(r,k+1)}} -  \overline{\x^{(r,k)}}  \right\|_2^2 
        \right).
    \label{eq:lem:fedavg:2o:ub:1:3}
    \end{align}
    Since $\expt \left[ \nabla F_m ( \x_m^{(r,k)}) - \nabla f (\x_m^{(r,k)}; \xi_m^{(r,k)}) \middle| \mathcal{F}^{(r,k)} \right] = 0$ we have
    \begin{align}
        & \expt \left[ \frac{1}{M} \sum_{m=1}^M \left\langle  \nabla F_m ( \x_m^{(r,k)}) - \nabla f (\x_m^{(r,k)}; \xi_m^{(r,k)}), \overline{\x^{(r,k+1)}} - \x^{\star} \right\rangle  \middle| \mathcal{F}^{(r,k)} \right]
        \nonumber
        \\
    =   & \expt \left[ \frac{1}{M} \sum_{m=1}^M \left\langle  \nabla F_m ( \x_m^{(r,k)}) - \nabla f (\x_m^{(r,k)}; \xi_m^{(r,k)}), \overline{\x^{(r,k+1)}} - \overline{\x^{(r,k)}}\right\rangle  \middle| \mathcal{F}^{(r,k)} \right]
    \nonumber
    %     \\
    % \leq & \expt \left[ \left\| \frac{1}{M} \sum_{m=1}^M (\nabla F_m ( \x_m^{(r,k)}) - g_m ( \x_m^{(r,k)})) \right\|_2 \middle| \mathcal{F}^{(r,k)} \right] 
    % \cdot \left\| \overline{\x^{(r,k+1)}} - \overline{\x^{(r,k)}} \right\|_2 \tag{Cauchy-Schwarz inequality}
        \\
    \leq & \eta \cdot \expt \left[ \left\| \frac{1}{M} \sum_{m=1}^M (\nabla F_m ( \x_m^{(r,k)}) - \nabla f(\x_m^{(r,k)}; \xi_m^{(r,k)})) \right\|_2^2 \middle| \mathcal{F}^{(r,k)} \right] 
    \nonumber \\
    & \quad +  \frac{1}{4\eta} \cdot \expt \left[ \left\| \overline{\x^{(r,k+1)}} - \overline{\x^{(r,k)}} \right\|_2^2 \middle| \mathcal{F}^{(r,k)} \right]
    \tag{Young's inequality}
        \\
    \leq & \frac{\eta \sigma^2 }{M} 
    +
    \frac{1}{4\eta} \cdot \expt \left[ \left\| \overline{\x^{(r,k+1)}} - \overline{\x^{(r,k)}} \right\|_2^2 \middle| \mathcal{F}^{(r,k)} \right],
    \label{eq:lem:fedavg:2o:ub:1:4}
    \end{align}
    where the last inequality is by bounded covariance assumptions and independence across clients.
    Plugging \cref{eq:lem:fedavg:2o:ub:1:4} back to the conditional expectation of \cref{eq:lem:fedavg:2o:ub:1:3} yields
    \begin{align}
        & \expt \left[ F( \overline{\x^{(r,k+1)}} ) - F(\x^{\star}) \middle| \mathcal{F}^{(r,k)} \right]
        +   \frac{1}{2\eta} \left( \expt\left[ \left\|\overline{\x^{(r,k+1)}} - \x^{\star} \right\|_2^2 \middle| \mathcal{F}^{(r,k)} \right] 
         - \left\|\overline{\x^{(r,k)}} - \x^{\star} \right\|_2^2  \right)
         \nonumber
        \\
    \leq & \frac{\eta \sigma^2}{M} 
        - \left(\frac{1}{4\eta} - L \right) \expt \left[ \left\| \overline{\x^{(r,k+1)}} - \overline{\x^{(r,k)}} \right\|_2^2 \middle| \mathcal{F}^{(r,k)} \right]
        +
        \frac{L}{M} \sum_{m=1}^M \left\| \x_m^{(r,k)} -  \overline{\x^{(r,k)}} \right\|_2^2 
         \nonumber
        \\
    \leq & \frac{\eta \sigma^2}{M} + \frac{L}{M} \sum_{m=1}^M \left\| \x_m^{(r,k)} -  \overline{\x^{(r,k)}} \right\|_2^2 
    \tag{since $\eta \leq \frac{1}{4L}$}.
    \end{align}
    Telescoping $k$ from $0$ to $K$ gives
    \begin{align*}
        \expt \left[  \frac{1}{K} \sum_{k=1}^{K} F( \overline{\x^{(r,k)}} ) - F(\x^{\star}) \middle| \mathcal{F}^{(r,0)} \right]
        \leq  & \frac{1}{2\eta K} \left( \left\|\overline{\x^{(r,0)}} - \x^{\star} \right\|_2^2 - \expt\left[ \left\|\overline{\x^{(r,K)}} - \x^{\star} \right\|_2^2 \middle| \mathcal{F}^{(r,0)} \right]  \right) \\
        & + \frac{\eta \sigma^2}{M} + \frac{L}{MK} \sum_{m=1}^M \sum_{k=0}^{K-1} \expt \left[ \left\| \x_m^{(r,k)} -  \overline{\x^{(r,k)}} \right\|_2^2 \middle | \mathcal{F}^{(r,0)} \right].
    \end{align*}
    Telescoping $r$ from $0$ to $R$ completes the proof of the lemma.
\end{proof}

\subsection{Deferred Proof of \cref{lem:fedavg:2o:ub:2}}
\label{sec:pf:lem:fedavg:2o:ub:2}
We restate the lemma for the readers' convenience. The proof is adapted from \cite{Woodworth.Patel.ea-NeurIPS20} which we include only for completeness.
\LemFedAvgSecondOrderUBSecond*
\begin{proof}[Proof of \cref{lem:fedavg:2o:ub:2}]
    By bounded gradient variance assumption,
    \begin{align}
          & \expt \left[ \left\| \x_1^{(r,k+1)} - {\x_2^{(r,k+1)}} \right\|_2^2 \middle| \mathcal{F}^{(r,k)} \right]
          \nonumber
          \\
        = & \expt \left[ \left\| \x_1^{(r,k)} - {\x_2^{(r,k)}} - \eta \left( \nabla f(\x_1^{(r,k)}; \xi_1^{(r,k)}) - \nabla f(\x_2^{(r,k)}; \xi_2^{(r,k)})  \right) \right\|_2^2 \middle| \mathcal{F}^{(r,k)} \right]
        \nonumber
        \\
        \leq & \left\| \x_1^{(r,k)} - {\x_2^{(r,k)}} \right\|_2^2 
        - 2 \eta  \left\langle  \nabla F_1(\x_1^{(r,k)}) - \nabla F_2(\x_2^{(r,k)})  , \x_1^{(r,k)} - {\x_2^{(r,k)}}   \right\rangle
        \nonumber
        \\
        & 
        + \eta^2  \left\| \nabla F_1(\x_1^{(r,k)}) - \nabla F_2(\x_2^{(r,k)})  \right\|_2^2 + 2 \eta^2 \sigma^2
        \label{eq:lem:fedavg:2o:ub:2:1}
    \end{align}
    Since $ \max_m \sup_{\x} \|  \nabla F_m(\x) - \nabla F(\x) \| \leq \zeta$, the second term of the RHS of \cref{eq:lem:fedavg:2o:ub:2:1} is bounded as
    \begin{align*}
        & - \left\langle  \nabla F_1(\x_1^{(r,k)}) - \nabla F_2(\x_2^{(r,k)})  , \x_1^{(r,k)} - {\x_2^{(r,k)}}   \right\rangle
        \\
        \leq &   - \left\langle  \nabla F(\x_1^{(r,k)}) - \nabla F(\x_2^{(r,k)})  , \x_1^{(r,k)} - {\x_2^{(r,k)}}   \right\rangle + 2 \zeta \left\| \x_1^{(r,k)} - {\x_2^{(r,k)}} \right\|_2 
        \\
        \leq & - \frac{1}{L} \left\|  \nabla F(\x_1^{(r,k)}) - \nabla F(\x_2^{(r,k)}) \right\|_2^2  + 2 \zeta \left\| \x_1^{(r,k)} - {\x_2^{(r,k)}} \right\|_2 
        \tag{by smoothness and convexity}
        \\
        \leq & - \frac{1}{L} \left\|  \nabla F(\x_1^{(r,k)}) - \nabla F(\x_2^{(r,k)}) \right\|_2^2  + 
        \frac{1}{2 \eta K}\left\| \x_1^{(r,k)} - {\x_2^{(r,k)}} \right\|_2^2 + 2 \eta K \zeta^2 
        \tag{by AM-GM inequality}
    \end{align*}
    Similarly the third term of the RHS of \cref{eq:lem:fedavg:2o:ub:2:1} is bounded as
    \begin{equation*}
        \left\|  \nabla F_1(\x_1^{(r,k)}) - \nabla F_2(\x_2^{(r,k)})  \right\|_2^2 
        \leq
        3 \left\|  \nabla F(\x_1^{(r,k)}) - \nabla F(\x_2^{(r,k)})  \right\|_2^2 + 6 \zeta^2.
    \end{equation*}
    Applying the above two bounds back to \cref{eq:lem:fedavg:2o:ub:2:1} gives (note that $\eta \leq \frac{1}{4L}$)
    \begin{align*}
        \expt \left[ \left\| \x_1^{(r,k+1)} - {\x_2^{(r,k+1)}} \right\|_2^2 \middle| \mathcal{F}^{(r,k)} \right] 
        & \leq
        \left( 1 + \frac{1}{K} \right) \left\| \x_1^{(r,k)} - {\x_2^{(r,k)}} \right\|_2^2+ 4 K \eta^2 \zeta^2 + 6 \eta^2 \zeta^2 + 2 \eta^2 \sigma^2
        \\
        & \leq  \left( 1 + \frac{1}{K} \right) \left\| \x_1^{(r,k)} - {\x_2^{(r,k)}} \right\|_2^2+ 10 K \eta^2 \zeta^2 + 2 \eta^2 \sigma^2.
    \end{align*}
    Telescoping
    \begin{equation*}
        \expt \left[ \left\| \x_1^{(r,k)} - {\x_2^{(r,k)}} \right\|_2^2 \middle| \mathcal{F}^{(r,0)} \right] 
        \leq
        \frac{\left( 1 + \frac{1}{K} \right)^k - 1}{\frac{1}{K}} \cdot \left( 10 K \eta^2 \zeta^2 + 2 \eta^2 \sigma^2 \right)
        \leq
        18 K^2 \eta^2 \zeta^2 + 4 K \eta^2 \sigma^2.
    \end{equation*}
    By convexity, for any $m \in [M]$,
    \begin{equation*}
        \expt \left[ \left\| \x_m^{(r,k)} - \overline{\x^{(r,k)}} \right\|_2^2 \middle| \mathcal{F}^{(r,0)} \right] 
        \leq
        18 K^2 \eta^2 \zeta^2 + 4 K \eta^2 \sigma^2.
    \end{equation*}
\end{proof}

\section{Formal Theorems and Proofs in \cref{sec:fedavg:2o:bias}}
\label{sec:pf:fedavg:2o:bias}
In this section, we state and prove the formal theorems on the lower and upper bounds of iterate bias discussed in \cref{sec:fedavg:2o:bias}.

\subsection{Formal Statement and Proof of \cref{thm:fedavg:2o:bias:ub}}
\begin{theorem}[Upper bound of iterate bias under second-order smoothness, formal version of \cref{thm:fedavg:2o:bias:ub}]
    \label{thm:fedavg:2o:bias:ub:complete}
    Assume $F(\x) := \expt_{\xi} f(\x; \xi)$ satisfies Assumption \ref*{asm:fo:2o}'.
    Let $\{\x_\sgd^{(k)}\}_{k=0}^{\infty}$ be the trajectory of SGD initialized at $\x_{\sgd}^{(0)} = \x$,  and $\{\z_{\gd}^{(k)}\}_{k=0}^{\infty}$ be the trajectory of GD initialized at $\z_{\gd}^{(0)} = \x$, namely
    \begin{equation*}
        \x^{(k+1)}_{\sgd} := \x_{\sgd}^{(k)} - \eta \nabla f(\x_{\sgd}^{(k)}; \xi^{(k)}), \quad 
        \z^{(k+1)}_{\gd} := \z_{\gd}^{(k)} - \eta \nabla F(\z_{\gd}^{(k)}), \quad \text{for } k = 0, 1,\ldots
    \end{equation*}
    Then for any $\eta \leq \frac{1}{L}$, the following inequality holds
    \begin{equation}
        \left \| 
            \expt \x_{\sgd}^{(k)} - \z_{\gd}^{(k)}
        \right\|_2
        \leq
        \min \{4 \eta^2  k^{\frac{3}{2}} L \sigma, \eta k^{\frac{1}{2}} \sigma \}.
        \label{eq:thm:bias:2o:ub}
    \end{equation}
    % Partciularly for $\eta \leq \frac{1}{\eta k L}$, one has $ \left \| \expt \x_{\sgd}^{(k)} - \z_{\gd}^{(k)} \right\|_2 \leq $
\end{theorem}

The proof of \cref{thm:fedavg:2o:bias:ub:complete} is based on the following two lemmas: \cref{lem:bias:2o:ub:1,lem:bias:2o:ub:2}.
\begin{lemma}
    \label{lem:bias:2o:ub:1}
    Under the same settings of \cref{thm:fedavg:2o:bias:ub:complete}, for any $\eta \leq \frac{1}{L}$, the following inequality holds 
    \begin{equation*}
        \left \| 
            \expt \x_{\sgd}^{(k)} - \z_{\gd}^{(k)}
        \right\|_2
        \leq
        \frac{(1 + \eta L)^k - 1}{\eta L} \cdot 2\eta^2 L k^{\frac{1}{2}} \sigma.
    \end{equation*}
\end{lemma}

\begin{proof}[Proof of \cref{lem:bias:2o:ub:1}]
    By definition of $\x_{\sgd}^{(k+1)}$ and $\z_{\gd}^{(k+1)}$ we obtain
    \begin{align*}
     \left\| \expt \x_{\sgd}^{(k+1)} - \z_{\gd}^{(k+1)} \right\|_2
    & = \left\| \left( \expt \x_{\sgd}^{(k)} - \z_{\gd}^{(k)} \right) 
    - 
    \eta \left( \expt \nabla F (\x_{\sgd}^{(k)}) - \nabla F (\z_{\gd}^{(k)}) \right) \right\|_2
    \\
    &   \leq
    \left\| \expt \x_{\sgd}^{(k)} - \z_{\gd}^{(k)}  \right\|_2
    + 
    \eta \left\| \expt \nabla F (\x_{\sgd}^{(k)}) - \nabla F (\z_{\gd}^{(k)}) \right\|_2.
    \end{align*}
    Now we seek an upper bound for $\left\| \expt \nabla F (\x_{\sgd}^{(k)}) - \nabla F (\z_{\gd}^{(k)}) \right\|_2$. Observe that
    \begin{align}
        & \left\| \expt \nabla F (\x_{\sgd}^{(k)}) - \nabla F (\z_{\gd}^{(k)}) \right\|_2 \nonumber
        \\
    \leq & \expt \left\| \nabla F (\x_{\sgd}^{(k)}) - \nabla F (\z_{\gd}^{(k)}) \right\|_2
        \tag{Jensen's inequality}
        \\
    \leq & \eta L  \expt \left\| \x_{\sgd}^{(k)} - \z_{\gd}^{(k)} \right\|_2
        \tag{by $L$-smoothness of $F$}
        \\
    \leq & \eta L  \left( \left\| \expt \x_{\sgd}^{(k)} - \z_{\gd}^{(k)} \right\|_2 + \expt \left\| \x_{\sgd}^{(k)} - \expt \x_{\sgd}^{(k)} \right\|_2 \right)
        \tag{by triangle inequality}
        \\
    \leq & \eta L  \left( \left\| \expt \x_{\sgd}^{(k)} - \z_{\gd}^{(k)} \right\|_2 + \sqrt{ \expt \left\| \x_{\sgd}^{(k)} - \expt \x_{\sgd}^{(k)} \right\|_2^2}\right).
        \tag{by Holder's inequality}
    \end{align}
    % To bound $\expt \left\| \x_{\sgd}^{(k)} - \expt \x_{\sgd}^{(k)} \right\|_2^2$ note that 
    % \begin{align}
    %     & \expt \left\| \x_{\sgd}^{(k+1)} - \expt \x_{\sgd}^{(k+1)} \right\|_2^2
    %     =
    %  \expt \left\| \left(\x_{\sgd}^{(k)} - \expt \x_{\sgd}^{(k)} \right)
    %     -
    %     \eta\left( \nabla f(\x^{(k)}_{\sgd}; \xi^{(k)}) - \expt \nabla F (\x_{\sgd}^{(k)}) \right)
    %     \right\|_2^2
    %     \\
    % \leq &      \expt \left\| \left(\x_{\sgd}^{(k)} - \expt \x_{\sgd}^{(k)} \right)
    %     -
    %     \eta\left( \nabla F(\x^{(k)}_{\sgd}) - \expt \nabla F (\x_{\sgd}^{(k)}) \right)
    %     \right\|_2^2 + \eta^2 \sigma^2
    % \end{align}
    By standard convex stochastic analysis (e.g. \cite{Khaled.Mishchenko.ea-AISTATS20}) one can show that  $\expt \left\| \x_{\sgd}^{(k)} - \expt \x_{\sgd}^{(k)} \right\|_2^2 \leq 2 \eta^2 k \sigma^2$. 
    % \hynote{Potential TODO} 
    Consequently
    \begin{equation}
        \left\| \expt \x_{\sgd}^{(k+1)} - \z_{\gd}^{(k+1)} \right\|_2
        \leq
        (1 + \eta L) \left(  \left\| \expt \x_{\sgd}^{(k)} - \z_{\gd}^{(k)} \right\|_2 \right) + 2 \eta^2 L k^{\frac{1}{2}} \sigma.
        \label{eq:proof:thm:bias:2o:ub:1}
    \end{equation}
    Telescoping \cref{eq:proof:thm:bias:2o:ub:1} completes the proof.
\end{proof}

\begin{lemma}
    \label{lem:bias:2o:ub:2}
    Under the same settings of \cref{thm:fedavg:2o:bias:ub:complete}, or any $\eta \leq \frac{1}{L}$, the following inequality holds
    \begin{equation*}
        \left \| 
            \expt \x_{\sgd}^{(k)} - \z_{\gd}^{(k)}
        \right\|_2
        \leq
        \eta k^{\frac{1}{2}} \sigma.
    \end{equation*}
\end{lemma}

\begin{proof}[Proof of \cref{lem:bias:2o:ub:2}]
     By definition of $\x_{\sgd}^{(k+1)}$ and $\z_{\gd}^{(k+1)}$ we obtain
    \begin{align}
        & \expt \left\| \x_{\sgd}^{(k+1)} - \z_{\gd}^{(k+1)} \right\|_2^2
        =
        \expt \left\| \left(\x_{\sgd}^{(k)} - \z_{\gd}^{(k)} \right)
        - 
        \eta \left( \nabla f(\x_{\sgd}^{(k)}; \xi^{(k)}) - \nabla F(\z_{\gd}^{(k)}) \right)
        \right\|_2^2
        \nonumber
        \\
    \leq & \expt \left\| \left(\x_{\sgd}^{(k)} - \z_{\gd}^{(k)} \right)
        - 
        \eta \left( \nabla F(\x_{\sgd}^{(k)}) - \nabla F(\z_{\gd}^{(k)}) \right)
        \right\|_2^2 + \eta^2 \sigma^2. \tag{by independence and $\sigma^2$-bounded covariance}
    \end{align}
    Note that 
    \begin{align}
        &  \left\| \left(\x_{\sgd}^{(k)} - \z_{\gd}^{(k)} \right)
        - 
        \eta \left( \nabla F(\x_{\sgd}^{(k)}) - \nabla F(\z_{\gd}^{(k)}) \right)
        \right\|_2^2
        \nonumber
        \\
    = & \left\| \x_{\sgd}^{(k)} - \z_{\gd}^{(k)}
        \right\|_2^2
        -
        2 \eta \left \langle \nabla F(\x_{\sgd}^{(k)}) - \nabla F(\z_{\gd}^{(k)}), \x_{\sgd}^{(k)} - \z_{\gd}^{(k)}  \right \rangle
        +
        \eta^2 \left\| \nabla F(\x_{\sgd}^{(k)}) - \nabla F(\z_{\gd}^{(k)})  \right\|_2^2 
        \nonumber
        \\
    \leq & \left\| \x_{\sgd}^{(k)} - \z_{\gd}^{(k)}
        \right\|_2^2
        -
        \left( \frac{2 \eta}{L} - \eta^2 \right) \left\| \nabla F(\x_{\sgd}^{(k)}) - \nabla F(\z_{\gd}^{(k)})  \right\|_2^2 
        \tag{by convexity and $L$-smoothness}
        \\
    \leq & \left\| \x_{\sgd}^{(k)} - \z_{\gd}^{(k)}
        \right\|_2^2 \tag{since $\eta \leq \frac{2}{L}$}.
    \end{align}
    Therefore
    \begin{equation*}
         \expt \left\| \x_{\sgd}^{(k+1)} - \z_{\gd}^{(k+1)} \right\|_2^2
         \leq
         \expt \left\| \x_{\sgd}^{(k)} - \z_{\gd}^{(k)} \right\|_2^2 + \eta^2 \sigma^2.
    \end{equation*}
    Telescoping yields
    \begin{equation*}
        \expt \|\x_{\sgd}^{(k)} - \z_{\gd}^{(k)}\|_2^2 \leq \eta^2 k \sigma^2,
    \end{equation*}
    and thus, by Jensen's inequality and Holder's inequality
    \begin{equation*}
        \left \|\expt  \x_{\sgd}^{(k)} - \z_{\gd}^{(k)} \right\|_2 \leq 
       \expt  \left\| \x_{\sgd}^{(k)} - \z_{\gd}^{(k)} \right\|_2
       \leq 
       \sqrt{ \expt  \left\| \x_{\sgd}^{(k)} - \z_{\gd}^{(k)} \right\|_2^2}
       \leq
       \eta k^{\frac{1}{2}} \sigma.
    \end{equation*}
\end{proof}

With \cref{lem:bias:2o:ub:1,lem:bias:2o:ub:2} at hands we are ready to prove \cref{thm:fedavg:2o:bias:ub:complete}.
\begin{proof}[Proof of \cref{thm:fedavg:2o:bias:ub:complete}]
    We consider the case of $\eta \leq \frac{1}{Lk}$ and $\eta > \frac{1}{Lk}$ separately. In either case we have $ \left \| 
            \expt \x_{\sgd}^{(k)} - \z_{\gd}^{(k)}
        \right\|_2
        \leq
        \eta k^{\frac{1}{2}} \sigma$ by \cref{lem:bias:2o:ub:2}.
    
    If $\eta \leq \frac{1}{Lk}$, by \cref{lem:bias:2o:ub:1}, we have
    \begin{equation*}
        \left \| 
            \expt \x_{\sgd}^{(k)} - \z_{\gd}^{(k)}
        \right\|_2
        \leq
        \frac{(1 + \eta L)^k - 1}{\eta L} 2 \eta^2 L k^{\frac{1}{2}} \sigma
        \leq
        \frac{e^{\eta LK} - 1}{\eta L} 2 \eta^2 L k^{\frac{1}{2}} \sigma
        \leq
        4 \eta^2 Lk^{\frac{3}{2}} \sigma,
    \end{equation*}
    where the last inequality is due to $e^{\eta Lk} - 1 \leq 2 \eta Lk$ since $\eta Lk \leq 1$. Therefore \cref{eq:thm:bias:2o:ub} is satisfied.
    
    If $\eta > \frac{1}{Lk}$, then $\eta k^{\frac{1}{2}} \sigma < \eta^2 Lk^{\frac{3}{2}} \sigma$. Hence \cref{eq:thm:bias:2o:ub} is also satisfied.
\end{proof}

\subsection{Formal Statement and Proof of \cref{thm:fedavg:2o:bias:lb}}
\begin{theorem}[Lower bound of iterate bias under second-order smoothness, complete version of \cref{thm:fedavg:2o:bias:lb}]
  \label{thm:fedavg:2o:bias:lb:complete}
For any $L, \sigma, K$, there exists a function $f(\x; \xi)$ and a distribution $\dist$ satisfying Assumption \ref*{asm:fo:2o}' such that for any $\eta \leq \frac{1}{2L}$, 
for any $k \leq K$ the following iterate bias inequality holds for SGD and GD initialized at the optimum
  \begin{equation*}
      \left\| \expt [\x_{\sgd}^{(k)}] - \z_{\gd}^{(k)} \right\|_2  \geq
      0.002\min\left\{ \eta^2  k^{\frac{3}{2}} L \sigma,  \eta^{\frac{1}{2}}L^{-\frac{1}{2}} \sigma \right\}.
      % \label{eq:2o:bias:lb:complete}
  \end{equation*}
\end{theorem}
\cref{thm:fedavg:2o:bias:lb:complete} is a special case of \cref{lem:fedavg:sgd:2o:lb}, by taking $x^{(0)} = 0$ to be the optimum.

\section{Deferred Proof in \cref{sec:fedavg:2o:lb}}
\label{lem:pf:fedavg:2o:lb}

% When we denote the iterates of SGD run at a single client, we use the notation $x^{(0)}, x^{(1)}, \cdots$ such that the iterates are given by the equation
% \begin{equation}
%     x^{(k + 1)} = x^{(k)} - \eta \left(\nabla \psi(x^{(k)}) + \sigma \xi^{(k)} \right),
% \end{equation}

\subsection{Proof Sketch of \cref{lem:fedavg:lb:f1}}
\label{sec:pf:lem:fedavg:lb:f1}
\begin{figure}
    \centering
    \begin{tabular}{ccc}
             \includegraphics[width=4cm]{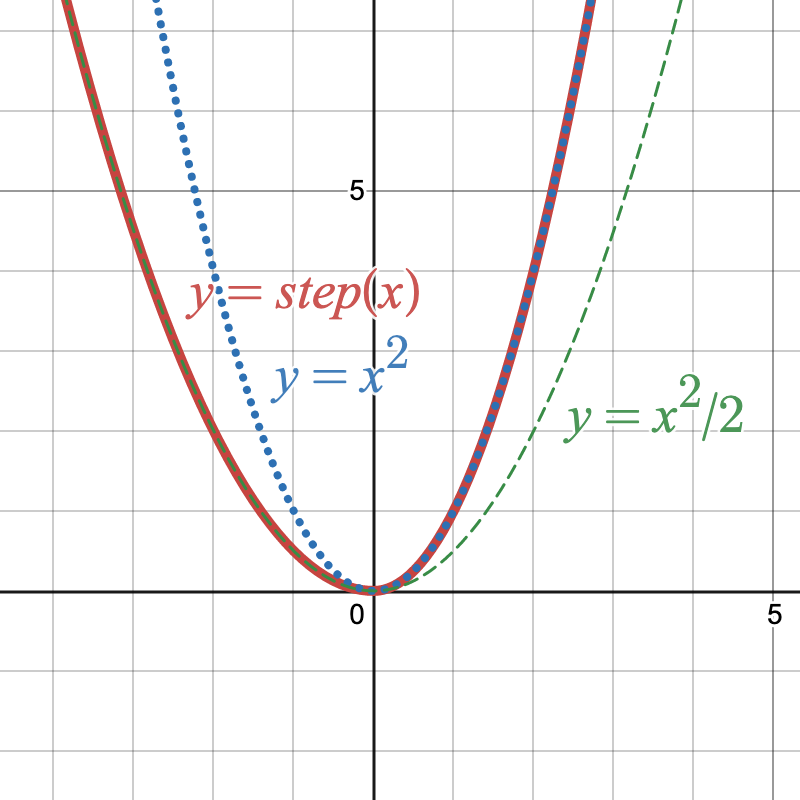} &     \includegraphics[width=4cm]{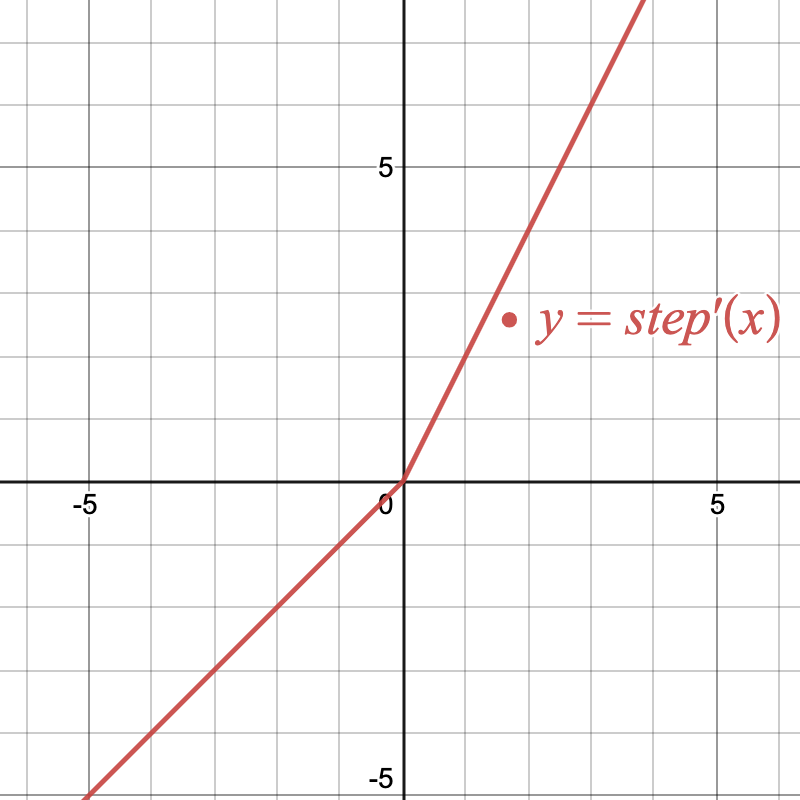} &     \includegraphics[width=4cm]{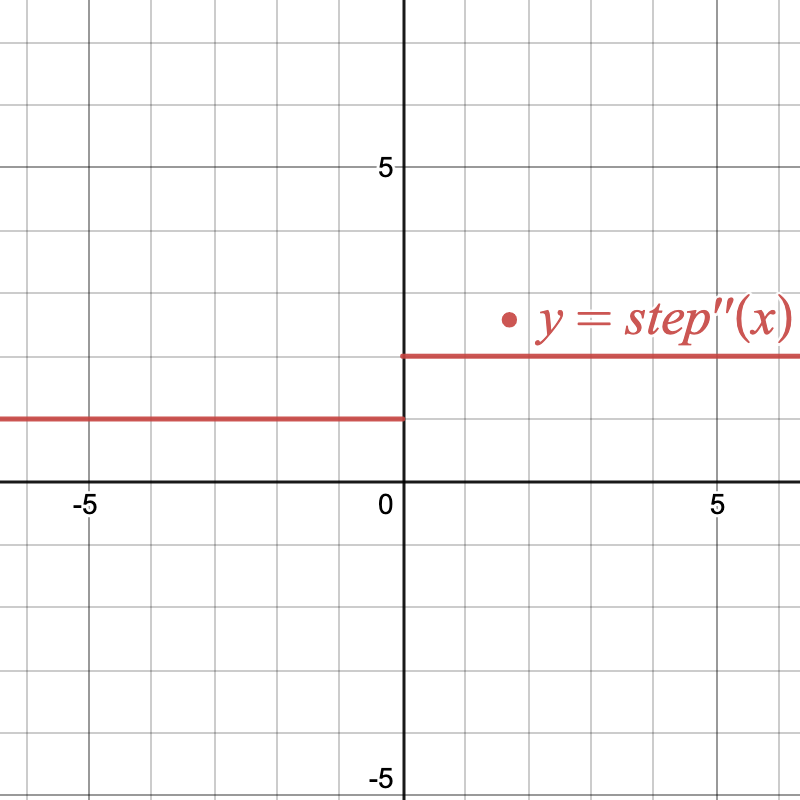} \\
        (a) & (b) & (c)  
    \end{tabular}
    \caption{The piecewise quadratic function and its first two derivatives.}
    \label{fig:step}
\end{figure}

In this subsection, we briefly sketch the proof of \cref{lem:fedavg:lb:f1}. The detailed proof is included in \cite{Glasgow.Yuan.ea-AISTATS22}. We restate the lemma as follows:

\lemsteplb*

To establish the lower bound, we show that when we run \fedavg on the function above, this same iterate bias, $\eta^2k^{\frac{3}{2}} L \sigma$ (recall \cref{thm:fedavg:2o:bias:lb}), persists more generally from any $x$ which is not too far from the optimum $0$. Loosely speaking, we can achieve this same bias whenever a constant fraction of the mass of the iterate $x_{\sgd}^{(k)}$ lies on each side of optimum $0$. Since the variance of $x_{\sgd}^{(k)}$ is on the order of $\eta^2 k \sigma^2$, we can prove that the bias will continue at the rate given in \cref{thm:fedavg:2o:bias:lb} from any $x$ with $|x| \leq \Theta(\eta \sqrt{k} \sigma)$. In fact, we can extend this observation to the case when the initial iterate $x_{\sgd}^{(0)}$ is a random variable, and its expectation is bounded, yielding the following lemma:

We formalize these observations in the following \cref{lem:fedavg:sgd:2o:lb}. Note that this lemma also captures the case when $\eta L k \geq 1$, where $\sigma_y - \sigma_z = \Theta\left(\sigma \eta^{\frac{1}{2}}L^{-\frac{1}{2}}\right)$. The detailed proof of \cref{lem:fedavg:sgd:2o:lb} can be found in Section B.2 of the full paper \cite{Glasgow.Yuan.ea-AISTATS22}.
\begin{lemma}
    \label{lem:fedavg:sgd:2o:lb}
    There exist universal constants $c_1$ and $c_2$ such that the following holds. Suppose we run SGD with step size $\eta$ on the function 
    $f^{(1)}(x; \xi) = \frac{1}{24} L \psi(x) + \xi x$ for $\xi \sim \mathcal{N}(0, \sigma^2)$ with step size $\eta \leq \frac{2}{L}$, starting at a possibly random iterate $x^{(0)}$. If
    $$-\sqrt{c_1}\frac{\sigma_y}{\alpha_y^k} \leq \expt[x^{(0)}] \leq 0,$$ then for any $k$,
    \begin{equation*}
        \expt[x_{\sgd}^{(k)}] \leq \left(1 - \frac{1}{24} \eta L\right)^k\expt[x^{(0)}] - \frac{1}{2} c_2\sigma \eta^{\frac{1}{2}} L^{-\frac{1}{2}}\min(1, \eta L k)^{\frac{3}{2}},
    \end{equation*}
    where $\sigma_y$ and $\alpha_y$ are defined in \cref{eq:fedavg:var_defs}.
\end{lemma}

Using \cref{lem:fedavg:sgd:2o:lb} inductively, we can show that the bias accumulates over many rounds of \fedavg. Loosely speaking, the bias grows linearly with the number of rounds $R$ until the force of the gradient exceeds the drift from the difference $\sigma_y -\sigma_z$.

\subsection{Deferred Proof of \cref{lem:fedavg:lb:f2}}
\label{sec:pf:lem:fedavg:lb:f2}
In this subsection, we prove \cref{lem:fedavg:lb:f2} regarding the second component $F^{(2)}$. We restate the lemma below for the reader's convenience.
\LemFtwo*
\begin{proof}[Proof of \cref{lem:fedavg:lb:f2}]
    Since $F^{(2)}$ is a deterministic function, running \fedavg with $R$ rounds and $K$ local steps per round is equivalent to running gradient descent with $KR$ steps. Consequently
    \begin{equation*}
        x^{(R,0)} = (1-\eta \mu)^{KR} \cdot x^{(0,0)}, \qquad f(x^{(R,0)}) = \frac{1}{2} \mu (1-\eta \mu)^{2KR} \cdot \left( x^{(0,0)} \right)^2.
    \end{equation*}
    Since $\eta \leq \frac{1}{\mu K R}$, $K \geq 2$, $R \geq 1$, we have $(1-\eta \mu)^{2KR} \geq \left( 1 - \frac{1}{KR} \right)^{2KR} \geq \frac{1}{16}$,
    where in the last inequality we applied the inequality $\inf_{z \geq 2} (1- z^{-1})^{z} = \frac{1}{4}$.
    As a result, we obtain $F^{(2)}(x^{(R,0)}) \geq \frac{1}{32} \mu  \left( x^{(0,0)} \right)^2.$
\end{proof}

\subsection{Deferred Proof of \cref{lem:fedavg:lb:f3}}
\label{sec:pf:lem:fedavg:lb:f3}
In this subsection, we prove \cref{lem:fedavg:lb:f3} regarding the third component $F^{(3)}$. We restate the lemma below for the reader's convenience.
\LemFthree*

\begin{proof}[Proof of \cref{lem:fedavg:lb:f3}]
    Since $F^{(3)}$ is a deterministic function, running \fedavg with $R$ rounds and $K$ local steps per round is equivalent to running gradient descent with $KR$ steps. Consequently $x^{(R,0)} = (1- \eta L)^{KR} \cdot x^{(0,0)}.$ Since $\eta \geq \frac{2}{L}$ we have $|1-\eta L| \geq 1$. Therefore $|x^{(R,0)}| \geq x^{(0,0)}$ and thus $F^{(3)}(x^{(R,0)}) \geq F^{(3)}(x^{(0,0)}) = \frac{1}{2} L (x^{(0,0)})^2$.
\end{proof}

\subsection{Deferred Proof of \cref{lem:fedavg:lb:f4}: Lower Bound on Bias of \fedavg with Heterogeneous Distribution}
\label{sec:pf:lem:fedavg:lb:f4}
In this subsection, we prove \cref{lem:fedavg:lb:f4} regarding the fourth component $F^{(4)}$. We restate the lemma below for the reader's convenience.
\LemFedAvgHetero*
\begin{proof}[Proof of \cref{lem:fedavg:lb:f4}]
By definition of $f^{(4)}$, we have $x_m^{(r,k)} = x_1^{(r,k)}$ for all odd $m \in [M]$, and $x_m^{(r,k)} = x_2^{(r,k)}$ for all even $m \in [M]$. Hence it suffices to study the trajectory of $x_1^{(r,k)}$ and $x_2^{(r,k)}$. 

For any $r$ and $0 \leq k < K$, we have
\begin{equation*}
    x_1^{(r, k + 1)} = x_1^{(r, k)} \left(1  - \frac{1}{4}\eta L \right) + \eta\zeta_* = \left(1  - \frac{1}{4}\eta L \right)\left(x_1^{(r, k)} - \frac{4\zeta_*}{L}\right) + \frac{4\zeta_*}{L},
\end{equation*}
and 
\begin{equation*}
    x_2^{(r, k + 1)} = x_2^{(r, k)} \left(1  - \frac{1}{8} \eta L \right) - \eta\zeta_* = \left(1  - \frac{1}{8} \eta L \right)\left(x_2^{(r, k)} + \frac{8 \zeta_*}{L}\right) - \frac{8 \zeta_*}{L}.
\end{equation*}
Recursing for $k$ from $0$ to $K$, we have 
\begin{equation*}
    x_1^{(r, K)} = \left(1  - \frac{1}{4}\eta L \right)^K\left(x_1^{(r, 0)} - \frac{4\zeta_*}{L} \right) + \frac{4\zeta_*}{L},
\quad
    x_2^{(r, K)} = \left(1  - \frac{1}{8} \eta L \right)^K\left(x_2^{(r, 0)} + \frac{8\zeta_*}{L} \right) - \frac{8 \zeta_*}{L}.
\end{equation*}
Since $x_m^{(r+1,0)}= \frac{1}{2}\left(x_1^{(r, K)} + x_2^{(r, K)} \right)$, we have for any $m \in [M]$
\begin{equation}
    x_m^{(r + 1, 0)} = a x_m^{(r, 0)} +  b\zeta_*,
    \label{eq:fedavg:hetero:one:step}
\end{equation}
where $a$ and $b$ are defined by
\begin{equation}
    a = \frac{1}{2}\left((1 - \frac{1}{4} \eta L)^K + (1 - \frac{1}{8} \eta L)^K\right),
\quad
    b = \frac{2}{L} \left(1 - (1 - \frac{1}{4} \eta L)^K  \right) - \frac{4}{L} \left(1 - (1 - \frac{1}{8} \eta L)^K\right).
    \label{eq:fedavg:def:a:b}
\end{equation}
We will show in the following claim that $b$ is upper bounded as follows. 
% We defer the proof of the following claim to the end of the present subsection.
\begin{claim}\label{claim:b}
Under the same condition of \cref{lem:fedavg:lb:f4}, it is the case that
\begin{equation*}
b \leq - \frac{0.001}{L} \min \left\{ 1, (\eta L K)^2 \right\},
\end{equation*}
where $b$ is defined in \cref{eq:fedavg:def:a:b}.
\end{claim}
The proof of \cref{claim:b} is deferred to \cref{sec:proof:claim:b}. We now apply \cref{claim:b} to show \cref{lem:fedavg:lb:f4}.

Recursing \cref{eq:fedavg:hetero:one:step}, since $x^{(0, 0)} \leq 0$, one has 
\begin{equation}
    x^{(R, 0)} = a^R x^{(0, 0)} +  \sum_{j = 0}^{R - 1}a^jb\zeta_* 
    = a^R x^{(0, 0)} + \frac{1 - a^R}{1 - a}b\zeta_* 
    \leq \frac{1 - a^R}{1 - a}b\zeta_*.
    \label{eq:fedavg:hetero:telescope}
\end{equation}
Since $a = \frac{1}{2}\left((1 - \frac{1}{4} \eta L)^K + (1 - \frac{1}{8} \eta L)^K\right) \leq (1 - \frac{1}{8} \eta L)^K$, we have the following lower bound of the numerator in \cref{eq:fedavg:hetero:telescope}: 
\begin{equation}
    1 - a^R \geq 1 - (1 - \frac{1}{8} \eta L)^{KR} \geq 1 - e^{-\frac{1}{8} \eta L KR} \geq \frac{1}{16} \min \left\{1, \eta L K R \right\},
    \label{eq:fedavg:hetero:numerator}
\end{equation}
where in the last inequality we applied $e^{-x} \leq \max \left\{ \frac{1}{2},  1 - \frac{1}{2} x \right\}$ for $x \geq 0$.

Since $a = \frac{1}{2}\left((1 - \frac{1}{4} \eta L)^K + (1 - \frac{1}{8} \eta L)^K\right) \geq (1 - \frac{1}{4} \eta L)^K$, we have the following upper bound of the denominator in \cref{eq:fedavg:hetero:telescope}:
\begin{equation}
    1 - a \leq 1 - \left( 1 - \frac{1}{4} \eta L \right)^K \leq \min \left\{ 1, \eta L K \right\}.
    \label{eq:fedavg:hetero:denominator}
\end{equation}
Taking \cref{eq:fedavg:hetero:telescope,eq:fedavg:hetero:numerator,eq:fedavg:hetero:denominator} together:
\begin{align*}
    x^{(R, 0)}
    \leq &
    - \frac{1}{16000L} \zeta^* \frac{\min \left\{ 1,  \eta L K R \right\}}{\min \left\{ 1, \eta L K \right\}} \min \left\{ 1, (\eta L K)^2 \right\} 
    \\
    = & - \frac{1}{16000L} \zeta^* \min \left\{ 1,  \eta L K R \right\} \min \left\{ 1, \eta L K\right\} 
    \\
    = & - \frac{1}{16000L} \zeta^* \min \left\{ 1, \eta L K, (\eta L K)^2 R\right\}. 
\end{align*}
\end{proof}

\subsubsection{Deferred Proof of \cref{claim:b}}
\label{sec:proof:claim:b}
\begin{proof}[Proof of \cref{claim:b}]
We now finish the proof of \cref{claim:b}.
Since $\eta \leq \frac{2}{L}$ we have the following
\begin{align*}
    b &= \frac{2}{L} \left(1 - (1 - \frac{1}{4} \eta L)^K  \right) - \frac{4}{L} \left(1 - (1 - \frac{1}{8} \eta L)^K\right)\\
    &\leq \frac{2}{L} \left( \frac{1}{4} \eta LK - \frac{1}{16} \binom{K}{2} (\eta L)^2 + \frac{1}{64} \binom{K}{3} (\eta L)^3 \right) 
        -  \frac{4}{L} \left( \frac{1}{8} \eta L K- \frac{1}{64} \binom{K}{2} (\eta L)^2 \right)\\ 
    & = - \frac{1}{16 L} \binom{K}{2} (\eta L)^2 + \frac{1}{32L} \binom{K}{3} (\eta L)^3,
\end{align*}
where in the first inequality we used the fact that for any integer $r \geq 2$ and $0 \leq x \leq 1$, 
\begin{equation*}
    1 - rx + \binom{r}{2}x^2 - \binom{r}{3}x^3 \leq (1 - x)^r \leq 1 - rx + \binom{r}{2}x^2.
\end{equation*}
If $\eta LK \leq 2$, then $\binom{K}{3} (\eta L)^3 \leq \frac{2}{3} \binom{K}{2} (\eta L)^2$, which shows that
\begin{equation*}
    b \leq - \frac{1}{16 L} \binom{K}{2} (\eta L)^2 +  \frac{3}{64 L} \binom{K}{2} (\eta L)^2 = - \frac{1}{64} \binom{K}{2} (\eta L)^2 \leq - \frac{1}{256} \eta^2 K^2 L
\end{equation*}

If $\eta L K > 2$, then consider the following five cases:
\textbf{Case 1}: If $K = 2$, then $\eta L > 1$ and therefore
\begin{equation*}
    b = \frac{2}{L} \left(1 - (1 - \frac{1}{4} \eta L)^2  \right) - \frac{4}{L} \left(1 - (1 - \frac{1}{8} \eta L)^2\right) = - \frac{1}{16} \eta^2 L \leq - \frac{1}{16 L}
\end{equation*}
\textbf{Case 2} If $K = 3$, then $\eta L > \frac{2}{3}$ and therefore
\begin{equation*}
    b = - \frac{3}{16} \eta^2 L + \frac{1}{64} \eta^3 L^2 \leq  - \frac{5}{32} \eta^2 L \leq - \frac{5}{72L}.
\end{equation*}
\textbf{Case 3} If $K = 4$, then $\eta L > \frac{1}{2}$ and therefore
\begin{equation*}
    b = - \frac{3}{16} \eta^2 L + \frac{3}{64} \eta^3 L^2 - \frac{7}{2048} \eta^4 L^3 \leq - \frac{3}{32} \eta^2 L \leq - \frac{3}{128L}
\end{equation*}
\textbf{Case 4}: If $K \geq 5$ and $\eta L \geq 1.04$,
\begin{equation*}
    b = \frac{2}{L} \left( -1 - (1 - \frac{1}{4} \eta L)^K + 2 (1 - \frac{1}{8} \eta L)^K \right) 
    \leq \frac{1}{2L} \left( -1 + 2 \left( 0.87 \right)^5 \right) \leq -\frac{1}{1000L}
\end{equation*}
\textbf{Case 5}: If $K \geq 5$ and $\eta L < 1.04$,
\begin{equation*}
    b = \frac{2}{L} \left( -1 - (1 - \frac{1}{4} \eta L)^K + 2 (1 - \frac{1}{8} \eta L)^K \right) \leq \frac{1}{2L} \left( -1 - e^{-1.16 \eta L K} + 2 e^{-0.5 \eta L K} \right) \leq -\frac{1}{1000L},
\end{equation*}
where in the first inequality we used the fact that $(1-x) \geq e^{-1.16x}$ for $x \leq [0, 0.26]$, and in the second inequality we used the fact that $-1 - e^{-1.16x} + 2e^{-0.5x} \leq 0.002$ for $x \geq 2$.
\end{proof}

\section{Formal Theorems and Proofs in \cref{sec:fedavg:3o:bias}}
\label{sec:pf:fedavg:3o:bias}
In this section, we state and prove the formal theorems on the lower and upper bounds of iterate bias under third-order smoothness \cref{asm:fo:3o} discussed in \cref{sec:fedavg:3o:bias}.

\subsection{Formal Statement and Proof of \cref{thm:fedavg:3o:bias:ub}}
\begin{theorem}[Upper bound of iterate bias under third-order smoothness, complete version of \cref{thm:fedavg:3o:bias:ub}]
    \label{thm:fedavg:3o:bias:ub:complete}
    Assume $(f, \dist)$ satisfies Assumption~\ref*{asm:fo:2o}' and \ref*{asm:fo:3o}'.
    Let $\{\x_\sgd^{(k)}\}_{k=0}^{\infty}$ be the trajectory of SGD initialized at $\x_\sgd^{(0)} = \x$,  and $\{\z_{\mathtt{GD}}^{(k)}\}_{k=0}^{\infty}$ be the trajectory of GD initialized at $\z_{\gd}^{(0)} = \x$, namely
    namely
    \begin{equation*}
        \xi^{(k)} \sim \dist, \quad \x^{(k+1)}_{\sgd} := \x_{\sgd}^{(k)} - \eta \nabla f(\x_{\sgd}^{(k)}; \xi^{(k)}), \quad 
        \z^{(k+1)}_{\gd} := \z_{\gd}^{(k)} - \eta \nabla F(\z_{\gd}^{(k)}), \quad \text{for } k = 0, 1,\ldots
    \end{equation*}
    Then for any $\eta \leq \frac{1}{L}$, the following inequality holds
    \begin{equation*}
        \left \| 
            \expt \x_{\sgd}^{(k)} - \z_{\gd}^{(k)}
        \right\|_2
        \leq
        \min \left\{
        \frac{1}{4} \eta^3 k^2 Q \sigma^2
        , 4 \eta^2 k^{\frac{3}{2}} L \sigma, \eta k^{\frac{1}{2}} \sigma \right\}.
    \end{equation*}
\end{theorem}

The proof of \cref{thm:fedavg:3o:bias:ub:complete} is based on the following lemma.
\begin{lemma}
\label{lem:bias:3o:ub}
Consider the same settings of \cref{thm:fedavg:3o:bias:ub:complete}. For any $k$, define vector-valued function 
\begin{equation*}
    \u^{(k)}(\x) = \expt \left[ \x_{\sgd}^{(k)} \mid \x^{(0)} = \x \right].
\end{equation*}
Then the following results hold.
\begin{enumerate}[(a)]
    \item For any $k$, $\u^{(k+1)}(\x) = \expt_{\xi} \left[ \u^{(k)}(\x - \eta \nabla f(\x; \xi) ) \right]$.
    \item For any $k$,  $\Diff \u^{(k+1)}(\x) = \expt_{\xi} \left[ \Diff  \u^{(k)} (\x - \eta \nabla f(\x; \xi)) \left( \I - \eta \nabla^2 f(\x; \xi) \right) \right]$. Here $\Diff$ denotes the Jacobian operator.
    \item For any $k$,  $\sup_{\x} \|\Diff \u^{(k)}(\x)\| \leq 1$.
    \item For any $k$,  $\sup_{\x} \|\Diff^2 \u^{(k)}(\x)\| \leq \eta k Q$.
    \item For any $k$,  $\left\| \u^{(k+1)}(\x) - \u^{(k)} (\x - \eta \nabla F(\x)) \right\|_2 \leq \frac{1}{2} \eta^3 k Q \sigma^2$.
\end{enumerate}
\end{lemma}
\begin{proof}[Proof of \cref{lem:bias:3o:ub}]
\begin{enumerate}[(a)]
    \item Holds by time-homogeneity of the SGD sequence as 
    \begin{align*}
        \u^{(k+1)}(\x) & = \expt \left[\x_{\sgd}^{(k+1)} \middle| \x^{(0)}_{\sgd} = \x \right]
        =
        \expt_{\xi} \expt \left[\x_{\sgd}^{(k+1)} \middle| \x^{(1)}_{\sgd} = \x - \eta \nabla f(\x; \xi) \right]
        \\
        & =
         \expt_{\xi} \expt \left[\x_{\sgd}^{(k)} \middle| \x^{(0)}_{\sgd} = \x - \eta \nabla f(\x; \xi) \right]
         = 
         \expt_{\xi} \left[ \u^{(k)}(\x - \eta \nabla f(\x; \xi) ) \right].
    \end{align*}
    \item Holds by taking derivative on both sides of (a). Indeed, for any $i \in [d]$, one has
        \begin{equation*}
            \nabla u_i^{(k+1)}(\x)^\top = \expt_{\xi} \left[
            \nabla u_i^{(k)} (\x - \eta \nabla f(\x; \xi))^\top \left( \I - \eta \nabla^2 f(\x; \xi) \right) 
            \right],
        \end{equation*}
        where $u_i^{(k)}$ denotes the $i$-th coordinate of the vector-valued function $\u^{(k)}$.
    \item By (b) one has 
    \begin{equation*}
        \left\|\Diff \u^{(k+1)}(\x)\right\|_2 \leq \expt_{\xi} \left[ 
        \left\|\Diff  \u^{(k)} (\x - \eta \nabla f(\x; \xi)) \right\|_2 
        \left\| \I - \eta \nabla^2 f(\x; \xi) \right\|_2
        \right].
    \end{equation*}
    \sloppy Since $f(\x; \xi)$ is convex and $L$-smooth w.r.t. $\x$, and $\eta \leq \frac{1}{L}$, one has $\sup_{\x, \xi} \left\| \I - \eta \nabla^2 f(\x; \xi) \right\|_2 \leq 1$. Therefore,
    \begin{equation*}
        \sup_{\x} \left\|\Diff \u^{(k+1)}(\x)\right\|_2
        \leq
        \sup_{\x} \left\|\Diff \u^{(k)}(\x)\right\|_2.
    \end{equation*}
    By definition of $\u^{(0)}(\x) = \Diff \u^{(0)}(\x) = \I$. Telescoping the above inequality yields (c).
    \item Taking twice derivatives w.r.t. $\x$ on both sides of (a) gives (for any $i$)
    \begin{align*}
        & \nabla^2 u_i^{(k+1)}(\x) 
        \\
        = &
        \expt_{\xi} \left[ (\I - \eta \nabla^2 f(\x; \xi)) \nabla^2 u_i^{(k)} (\x - \eta \nabla f(\x; \xi))
            (\I - \eta \nabla^2 f(\x; \xi)) \right. \\
        & \qquad \left.
            - \eta \nabla^3 f (\x; \xi) [\nabla u_i^{(k)}(\x - \eta \nabla f(\x; \xi))] \right]
    \end{align*}
    Therefore,
    \begin{align*}
        & \sup_{\x}\| \Diff^2 \u^{(k+1)} (\x)\|_2 \\
        \leq &  \sup_{\x}\| \Diff^2 \u^{(k)}(\x) \|_2 
        \sup_{\x, \xi} \| \I- \eta \nabla^2 f(\x; \xi) \|_2^2 
        + \eta \cdot \left(  \sup_{\x, \xi}\|\nabla^3 f(\x; \xi)\|_2  \right)
        \cdot
        \left( \sup_{\x} \|\Diff \u^{(k)} (\x)\|_2 \right).
    \end{align*}
    Since $f(\x; \xi)$ is convex and $L$-smooth w.r.t. $\x$ and $\eta \leq \frac{1}{L}$, one has $\sup_{\x, \xi} \left\| \I - \eta \nabla^2 f(\x; \xi) \right\|_2 \leq 1$. Also by (c), we arrive at
    \begin{equation*}
        \sup_{\x} \| \Diff^2 \u^{(k+1)} (\x)\|_2
        \leq
        \sup_{\x} \| \Diff^2 \u^{(k)} (\x)\|_2
        +
        \eta Q
    \end{equation*}
    Telescoping from $0$ to $k$ yields (d).
    \item By (a)
    \begin{small}
        \begin{align}
            & \left\| \u^{(k+1)}(\x) - \u^{(k)} (\x - \eta \nabla F(\x)) \right\|_2
            = \left\| \expt_{\xi} \left[ \u^{(k)}(\x - \eta \nabla f(\x; \xi) ) \right] - \u^{(k)} (\x - \eta \nabla F(\x)) \right\|_2 
            \tag{by (a)}
            \\
            = & \left\| \expt_{\xi} \left[ \u^{(k)}(\x - \eta \nabla f(\x; \xi) ) - \u^{(k)} (\x - \eta \nabla F(\x)) 
            - \Diff \u^{(k)} (\x - \eta \nabla F(\x)) \left(\eta \nabla f(\x; \xi) - \eta \nabla F(\x)  \right) \right] 
            \right\|_2 
            \tag{Since $\expt_{\xi} \nabla f(\x; \xi) = \nabla F(\x)$}
            \\
            \leq & \expt_{\xi} \left\| \u^{(k)}(\x - \eta \nabla f(\x; \xi) ) - \u^{(k)} (\x - \eta \nabla F(\x)) 
            - \Diff \u^{(k)} (\x - \eta \nabla F(\x)) \left(\eta \nabla f(\x; \xi) - \eta \nabla F(\x)  \right) 
            \right\|_2 
            \tag{By Jensen's inequality}
            \\
            \leq & \frac{1}{2} \sup_{\x} \|\Diff^2 \u^{(k)}(\x) \|_2 
            \expt_{\xi} \| \eta \nabla F(\x) - \eta \nabla f(\x; \xi) \|_2^2
            \tag{By Taylor's expansion}
            \\
            \leq & \frac{1}{2} \eta k Q \eta^2 \cdot \sigma^2
            =
            \frac{1}{2} \eta^3 k Q \sigma^2.
            \nonumber
        \end{align}
    \end{small}
\end{enumerate}
\end{proof}

We are now ready to finish the proof of \cref{thm:fedavg:3o:bias:ub:complete}.
\begin{proof}[Proof of \cref{thm:fedavg:3o:bias:ub:complete}]
    By \cref{lem:bias:3o:ub}(e), for any $j \in \{0, 1, \ldots, k\}$
    \begin{equation*}
        \left\|\u^{(k-j)}(\z_{\gd}^{(j)}) - \u^{(k-j-1)} (\z_{\gd}^{(j+1)}) \right\|_2 \leq \frac{1}{2} \eta^3 (k-j - 1) Q \sigma^2
    \end{equation*}
    Consequently
    \begin{align*}
        & \left\| \expt \x_{\sgd}^{(k)} - \z_\gd^{(k)} \right\|_2
        =
        \left\| \u^{(k)}(\z_{\gd}^{(0)}) - \u^{(0)} (\z_\gd^{(k)}) \right\|_2
        \\
        \leq &
        \sum_{j=0}^{k-1} \left\|\u^{(k-j)}(\z_{\gd}^{(j)}) - \u^{(k-j-1)} (\z_{\gd}^{(j+1)}) \right\|_2
        \leq
        \frac{1}{4} \eta^3 k^2 Q \sigma^2.
    \end{align*}
    % \begin{equation}
    %     \left\|\u^{(k-1)}(\z_{\gd}^{(1)}) - u^{(k-2)} (\z_{\gd}^{(2)}) \right\|_2 \leq \frac{1}{2} \eta^3 (k-1) Q \sigma^2
    % \end{equation}
\end{proof}

% \begin{equation}
%     \Diff^2 \u^{(k+1)} (\x) = \expt_{\xi} \Diff^2 \u^{(k+1)} (\x)
% \end{equation}

% \begin{equation}
%     \frac{\partial u^{(k+1)}_i (\x)}{\partial x_j}
%     =
%     \expt_{\xi}
%     \left[ 
%         \frac{\partial u^{(k)}_i(\x - \eta \nabla f(\x; \xi))}{\partial x_j}
%     \right]
% \end{equation}

% \begin{equation}
%     \nabla^2 u_{k+1}(\x) = 
%     \expt_{\xi} \left[ (\I - \eta \nabla^2 f(\x; \xi)) \nabla^2 u_k (\x - \eta \nabla f(\x; \xi))
%     (\I - \eta \nabla^2 f(\x; \xi)) 
%     - \eta \nabla^3 f (\x; \xi) \nabla u_k(\x - \eta \nabla f(\x; \xi)) \right]
% \end{equation}

% \begin{lemma}
%     u_{k+1} (\x)
% \end{lemma}

\subsection{Formal Statement and Proof of \cref{thm:fedavg:3o:bias:lb}}
\begin{theorem}[Lower bound of iterate bias under third-order smoothness, complete version of \cref{thm:fedavg:3o:bias:lb}]
    \label{thm:fedavg:3o:bias:lb:complete}
    For any $L, \sigma, K$, for any $Q \leq \frac{L^2}{12 K \sigma}$, there exists a function $f(\x; \xi)$ and a distribution $\dist$ satisfying Assumption~\ref*{asm:fo:2o}' and \ref*{asm:fo:3o}' such that for any $\eta \leq \frac{1}{2L}$, for any $k < K$, the following iterate bias inequality holds for SGD and GD initialized at the optimum
    \begin{equation}
        \left\| \expt [\x_{\sgd}^{(k)}] - \z_{\gd}^{(k)} \right\|_2  \geq
        0.005 \eta^3 \sigma^2 Q  \min \left\{ \frac{k-1}{ \eta L}, k(k-1) \right\}.
        \label{eq:3o:bias:lb:complete}
    \end{equation}
\end{theorem}
Before we state the proof of \cref{thm:fedavg:3o:bias:lb:complete}, let us first describe the following helper function used to construct the lower bound instance. Define
\begin{equation}
    \varphi(x) = \int_0^x \log (\cosh (x)) \diff x.
    \label{eq:3o:phi}
\end{equation}
In the following lemma, we show that this $\varphi(x)$ satisfies the following properties
\begin{lemma}
\label{lem:bias:3o:lb:helper}
The following properties hold for the $\varphi(x)$ defined in  \cref{eq:3o:phi}.
\begin{enumerate}[(a)]
    \item $\varphi'(x) = \log (\cosh (x))$. Therefore, $\varphi'(x) \leq |x|$. In particular $\varphi(0) = 0$.
    \item $\varphi''(x) = \tanh (x)$. In particular $\varphi''(0) = 0$, $\lim_{x \to +\infty} \varphi''(x) = 1$, $\lim_{x \to -\infty} \varphi''(x) = -1$, and $\varphi''(x) \in [-1, 1]$ for any $x \in \reals$.
    \item $\varphi'''(x) = \sech^2 (x)$. In particular $\varphi'''(0) = 1$, $\lim_{x \to +\infty} \varphi'''(x) = 0$, $\lim_{x \to -\infty} \varphi'''(x) = 0$, and $\varphi'''(x) \in [0, 1]$ for any $x \in \reals$. 
    Also $\varphi'''(x) \geq \frac{1}{2}$ for any $x \in [-\frac{1}{2}, +\frac{1}{2}]$
    \item $\varphi''''(x) = -2 \sech^2(x) \tanh(x)$. In particular $\varphi''''(x) \in (-1, 1)$ for any $x \in \reals$.
\end{enumerate}
% \begin{equation}
    
%     \quad
%     \varphi''(x) = \tanh (x),
%     \quad
%     \varphi'''(x) = \sech^2 (x)
% \end{equation}
\end{lemma}
\begin{proof}[Proof of \cref{lem:bias:3o:lb:helper}]
    All results follow by standard trigonometry analysis.
\end{proof}

Next we establish the following lemma
\begin{lemma}
    \label{lem:bias:3o:lb:instance}
    Consider 
    \begin{equation}
        f(x; \xi) = \frac{3}{8}Lx^2 + \frac{L^3}{64Q^2} \varphi \left( \frac{4Q}{L} x \right) +  \xi, \qquad
        F(x) := \expt_{\xi \sim \mathcal{U}[-\sigma, \sigma]} f(x; \xi).
        \label{def:fedavg:bias:3o:lb:instance}
    \end{equation}
    where $\varphi$ is defined in \cref{eq:3o:phi}. 
    Then
    \begin{enumerate}[(a)]
        \item $f''(x; \xi) = F''(x) = \frac{3}{4}L + \frac{1}{4}L \varphi'' \left( \frac{4Q}{L} x \right)$. 
        Therefore, $F''(x)  \in [\frac{1}{2}L, L]$ for any $x \in \reals$.
        \item $f'''(x; \xi) = F'''(x) = Q \varphi'''(\frac{4Q}{L} x )$. Therefore, $F'''(x) \in [0,Q]$ for any $x \in \reals$. In particular $F'''(0) = Q$, and $F'''(x) \geq \frac{1}{2}Q$ for any $x \in [-\frac{L}{8Q}, +\frac{L}{8Q}]$.
        \item $f(x; \xi)$ satisfies \cref{asm:fo:2o,asm:fo:3o}.
    \end{enumerate}
\end{lemma}
\begin{proof}[Proof of \cref{lem:bias:3o:lb:instance}]
    (a,b) follow from \cref{lem:bias:3o:lb:helper}. (c) follows by (a, b) and the fact that the variance of $\mathcal{U}[-\sigma, +\sigma] \leq \sigma^2$.
\end{proof}
% Consider $g(x; \xi) := G(x) + \xi x$ where $\xi \sim \mathcal{U}(-\sigma, +\sigma)$. Then $g'(x; \xi) = G'(x) + \xi$. 

The following lemma studies the SGD trajectory on $f$ defined in \cref{def:fedavg:bias:3o:lb:instance}.
\begin{lemma}
\label{lem:bias:3o:lb:enum}
Let $\{\x_{\sgd}^{(k)}\}_{k=0}^{\infty}$ be the SGD trajectory on the function $f$ defined in \cref{def:fedavg:bias:3o:lb:instance}, with learning rate $\eta$, that is 
\begin{equation*}
    x_{\sgd}^{(k+1)} \gets x_{\sgd}^{(k)} - \eta \cdot f'(x_{\sgd}^{(k)}; \xi^{(k)}), \qquad \xi^{(k)} \sim \mathcal{U}[-\sigma, +\sigma].
\end{equation*}
Define
\begin{equation*}
    u_k(x) := \expt [x_{\sgd}^{(k)} | x_{\sgd}^{(0)} = x].
\end{equation*}
Then the following results hold
\begin{enumerate}[(a)] 
    \item  $u_{k+1}(x) = \expt_{\xi} \left[ u_k (x - \eta f'(x; \xi)) \right]$
    \item $u_{k+1}'(x) = \expt_{\xi} \left[(1 - \eta F''(x)) \cdot u_k'(x - \eta f'(x; \xi))  \right]$.
    \item $u_{k+1}''(x) = \expt_{\xi} \left[ (1 - \eta F''(x))^2 u_k''(x - \eta f'(x; \xi)) - \eta F'''(x) u_k'(x - \eta f'(x;\xi))\right]$. 
    \item For any $k$, $\inf_{x} \{u_k'(x) \} \geq (1- \eta L)^k$  holds.
    \item For any $k$, $\sup_{x} \{u_k''(x)\} \leq 0$.
    \item For any $x \in \reals$ and $k$, it is the case that $u''_{k+1}(x) \leq (1 - \eta L)^2 \expt_{\xi}[u_k''(x - \eta f'(x; \xi))] - \eta (1 - \eta L) F'''(x)$.
\end{enumerate}
\end{lemma}
\begin{proof}[Proof of \cref{lem:bias:3o:lb:enum}]
\begin{enumerate}[(a)]
    \item Proved in \cref{lem:bias:3o:ub}(a).
    \item Proved in \cref{lem:bias:3o:ub}(b).
    \item Holds by taking derivative with respect to $x$ on both sides of (b).
    \item Since $F''(x) \in [\frac{1}{2}L, L]$, by (b), we have
    \begin{equation*}
        \inf_{x} \{u_{k+1}'(x)\} \leq (1 - \eta L) \inf_x \{ u_k'(x)\}.
    \end{equation*}
    By definition of $u_0$ we have $u_0(x) \equiv x$ and thus $u_0'(x) \equiv 1$. Telescoping the above inequality gives (d).
    \item 
    We prove by induction. For $k = 0$ we have $u_0''(x) \equiv 0$ which clearly satisfies (e). Now assume (e) holds for the case of $k$, and we study the case of $k+1$.

    Since $F''(x) \in [\frac{1}{2}L, L]$ and $F'''(x) \geq 0$, by (c) and (d), we have
    \begin{equation*}
        \sup_{x} \{ u_{k+1}''(x) \}
        \leq
        (1 - \eta L)^2 \sup_{x} \{ u_{k}''(x) \}
        -
        \eta  \inf_{x} \{F'''(x)\} (1 - \eta L)^k
        \leq 0,
    \end{equation*}
    completing the induction.
    
    \item Holds by (c-e).
    \end{enumerate}
\end{proof}

% \begin{lemma}
%     \begin{enumerate}[(a)]
%         % \item $u'_{k}(x) \geq (1 - \eta L)^k $ for any $x \in \reals$ and $k$.
%         \item $u''_{k}(x) \leq 0$ for any $x \in \reals$ and $k$.
%         \item 
%     \end{enumerate}
% \end{lemma}

% \begin{lemma}
% % For any non-negative $z$ such that $z \leq \frac{L}{8Q}$, the following inequality holds
% For any $x \in [-\frac{L}{8Q}, +\frac{L}{8Q}]$, the following inequality holds
%     \begin{equation}
%         u_{k+1}''(x) \leq (1 - \eta L)^2 \sup_{z \in [-x-\eta\sigma , x+\eta\sigma]} \{ u_{k}''(z) \} - \eta (1 - \eta L) \frac{Q}{2}.
%     \end{equation}
%     % \begin{equation}
%     %     \sup_{x \in [-z, z]} \{ u_{k+1}''(x) \} \leq (1 - \eta L)^2 \sup_{x \in [-z-\eta\sigma , z+\eta\sigma]} \{ u_{k}''(x) \} - \eta (1 - \eta L) \frac{Q}{2}.
%     % \end{equation}
% \end{lemma}

% \begin{equation}
    % \max_{x \in [-\frac{L}{12Q}, +\frac{L}{12Q}]} \{u_{k}''(x)\}
    % \leq
    % - \sum_{j=0}^{k-1} (1 - \eta L)^{2j}  \cdot \eta (1-\eta L) \frac{Q}{2}
    % \leq 
    % - \m
    % something
% \end{equation}

% \begin{proof}
%     By \cref{lem:bias:3o:lb:enum}(f) and the fact that $\inf_{x \in [-\frac{L}{8Q}, \frac{L}{8Q}]} F'''(x) \geq \frac{Q}{2}$ from \cref{lem:bias:3o:lb:instance}, one has
%     \begin{equation}
%         u''_{k+1}(x) \leq (1 - \eta L)^2 \expt_{\xi}[u_k''(x - \eta f'(x; \xi))] - \eta (1 - \eta L) F'''(x)
%     \end{equation}
%     % DUMP
%     % w.p. $\geq 1 - \eta L$ (need $Q$ bound), one has
%     % $x - \eta g'(x; \xi) \in .....$ OMG \hynote{Lave to shrink ... Need $\eta K \sigma \lesssim \frac{L}{Q}$}
% \end{proof}
We further have the following lemma.
\begin{lemma}
    \label{lem:bias:3o:lb:enum:2}
    Under the same setting of \cref{lem:bias:3o:lb:enum}, the following results hold.
    \begin{enumerate}[(a)]
    \item For any $x \in [-\frac{L}{8Q}, \frac{L}{8Q}]$ and $k$, 
    \begin{equation*}
         u''_{k+1}(x) \leq (1 - \eta L)^2 \sup_{z \in [x - \eta\sigma, x+ \eta \sigma]} \{ u_k''(z) \} - \eta (1 - \eta L) \frac{Q}{2}.
    \end{equation*}
    \item Assuming $Q \leq \frac{L}{24 \eta K \sigma}$, then for any $k < K$, the following inequality holds
        \begin{equation*}
         \sup_{x \in [-\frac{L}{12Q}, +\frac{L}{12Q}]} u''_{k}(x) \leq - \sum_{j=0}^{k-1} (1-\eta L)^{2j
         +1} \cdot \frac{\eta Q}{2}.
        \end{equation*}
    \item Assuming $\eta \leq \frac{1}{2L}$ and $Q \leq \frac{L}{24 \eta K \sigma}$, then for any $k < K$, for any $x \in [-\frac{L}{24Q}, +\frac{L}{24Q}]$, one has
        \begin{equation*}
            u_{k+1}(x) \leq  u_k(x - \eta F'(x)) -  \frac{1}{12} \eta^3 \sigma^2 Q \sum_{j=0}^{k-1} (1 - \eta L)^{2j+1}.
        \end{equation*}
    % \item Assuming $\eta \leq \frac{1}{2L}$ and $Q \leq \frac{L}{24 \eta K \sigma}$, then for any $k < K$, for any $x \in [-\frac{L}{24Q}, +\frac{L}{24Q}]$, one has
    \end{enumerate}
\end{lemma}
\begin{proof}[Proof of \cref{lem:bias:3o:lb:enum:2}]
    \begin{enumerate}[(a)]
        \item  Holds by (f) and the fact that 
        \begin{equation*}
            |f'(x;\xi) - F'(x)| \leq \eta \sigma \quad \text{and} \quad \inf_{x \in [-\frac{L}{8Q}, \frac{L}{8Q}]} F'''(x) \geq \frac{Q}{2}.
        \end{equation*}
        \item Since $ \frac{L}{12Q} + \eta \sigma K \leq \frac{L}{8Q}$ (due to the assumption that $Q \leq \frac{L}{24 \eta K \sigma}$), we can repeatedly apply (a) for $K$ times. Therefore,
        \begin{align*}
                & \sup_{x \in [-\frac{L}{12Q}, +\frac{L}{12Q}]} \{u''_{k}(x) \} 
                \\
            \leq &
            (1 - \eta L)^2 \sup_{x \in [-\frac{L}{12Q} - \eta \sigma, \frac{L}{12Q} + \eta \sigma]}
             \{u''_{k-1}(x) \} - \eta (1-\eta L)\frac{Q}{2}
             \\
             \leq & (1 - \eta L)^{2k} \sup_{x \in [-\frac{L}{12Q} - \eta k \sigma, \frac{L}{12Q} + \eta k \sigma]}
             \{u''_{0}(x) \} - \eta \sum_{j=0}^{k-1} (1-\eta L)^{2j} (1-\eta L)\frac{Q}{2}.
        \end{align*}
        Plugging in $u_0''(x) \equiv 0$ gives (b).
        \item By \cref{lem:bias:3o:lb:enum}(a),
        \begin{align*}
            & u_{k+1}(x) - u_k(x - \eta F'(x)) = \expt_{\xi} \left[ u_k (x - \eta f'(x; \xi)) - u_k(x - \eta F'(x))\right] 
            \\
        \leq & \expt_{\xi} \left[ - \eta \cdot u_k'(x - \eta F'(x)) \cdot (f'(x; \xi) - F'(x))  \right.
        \\
            & \qquad \left. + \frac{1}{2} \sup_{z \in [x - \eta F'(x) - \eta \sigma, x - \eta F'(x) + \eta \sigma]} u''(z) \cdot \eta^2 (f'(x; \xi) - F'(x))^2 \right] 
        \\
        \leq & \frac{1}{6} \eta^2 \sigma^2 \sup_{z \in [x - \eta F'(x) - \eta \sigma, x - \eta F'(x) + \eta \sigma]} u''(z)
        \end{align*}
    \end{enumerate}
    Since $x \in [-\frac{L}{24Q}, \frac{L}{24Q}]$, we know that $x - \eta F'(x) \in [-\frac{L}{24Q}, \frac{L}{24Q}]$ by construction of $F$. Since $Q \leq \frac{L}{24 \eta K \sigma}$ we know that 
    $[x - \eta F'(x) - \eta \sigma, x - \eta F'(x) + \eta \sigma] \subset [-\frac{L}{12Q}, \frac{L}{12Q}]$. Therefore, (b) is applicable, which suggets
    \begin{equation*}
        u_{k+1}(x) - u_k(x - \eta F'(x))  \leq - \frac{1}{12} \eta^3 \sigma^2 Q \sum_{j=0}^{k-1} (1 - \eta L)^{2j+1}.
    \end{equation*}
\end{proof}

We are ready to finish the proof of \cref{thm:fedavg:3o:bias:lb:complete} now.
\begin{proof}[Proof of \cref{thm:fedavg:3o:bias:lb:complete}]
    For $k=1$ the bound trivially holds. From now on assume $k \geq 2$.

    Consider the one-dimensional instance $f$ defined in \cref{def:fedavg:bias:3o:lb:instance}. The optimum of $F = \expt_{\xi} f(x; \xi)$ is clearly 0.
    We will in fact show a stronger result that \cref{eq:3o:bias:lb:complete} holds for any $x \in [-\frac{L}{24Q}, +\frac{L}{24Q}]$, in addition to 0.

    Since $\eta \leq \frac{1}{2L}$, for any $x \in [-\frac{L}{24Q}, +\frac{L}{24Q}]$, one has $x - \eta F'(x) \in  [-\frac{L}{24Q}, +\frac{L}{24Q}]$. Therefore, one can repeatedly apply \cref{lem:bias:3o:lb:enum:2}(c), which yields
    \begin{equation*}
        \expt [x_{\sgd}^{(k)}] - z_{\gd}^{(k)} \leq - \frac{1}{12} \eta^3 \sigma^2 Q \sum_{j=0}^{k-1} \sum_{i=0}^{j-1} (1 - \eta L)^{2i+1}.
    \end{equation*}
    If $k \leq \frac{1}{\eta L}$ then
    \begin{equation*}
        \sum_{j=1}^{k-1} \sum_{i=0}^{j-1} (1 - \eta L)^{2i+1}
        \geq
        k (k-1) (1 - \eta L)^{2k-3}
        \geq
        k (k-1) \left(1 - \frac{1}{k} \right)^{2k-3}
        \geq
        \frac{1}{e^2} k (k-1)
        \geq 
        \frac{k(k-1)}{16}.
    \end{equation*}
    If $k > \frac{1}{\eta L}$ then
    \begin{equation*}
        \sum_{j=1}^{k-1} \sum_{i=0}^{j-1} (1 - \eta L)^{2i+1} 
        =
        \frac{(1 - \eta L) ((1- \eta L)^{2k} + \eta L (2- \eta L) k - 1)}{\eta^2 L^2 (2 - \eta L)^2}
        \geq
        \frac{ \frac{3}{2} \eta L k - 1 }{8 \eta^2 L^2}
        \geq
        \frac{k - 1}{16 \eta L},
    \end{equation*}
    where in the second from the last inequality we used the assumption that $\eta L \leq \frac{1}{2}$.
    In either case we have
    \begin{equation*}
        \sum_{j=1}^{k-1} \sum_{i=0}^{j-1} (1 - \eta L)^{2i+1} 
        \geq 
        \min \left\{ \frac{k - 1}{16 \eta L}, \frac{1}{16} k(k-1) \right\},
    \end{equation*}
    and hence
    \begin{equation*}
        \expt [x_{\sgd}^{(k)}] - z_{\gd}^{(k)} \leq
        - 0.005 \eta^3 \sigma^2 Q (k-1) \min \left\{ \frac{1}{ \eta L}, k \right\}.
    \end{equation*}
\end{proof}
\section{Deferred Proof in \cref{sec:fedavg:ncvx}}
In this section, we prove \cref{thm:fedavg:ncvx:2o,thm:fedavg:ncvx:3o} on the upper bounds of \fedavg in the non-convex settings.

\subsection{Deferred Proof of \cref{thm:fedavg:ncvx:3o}}
\label{sec:pf:fedavg:ncvx:3o}
We first prove \cref{thm:fedavg:ncvx:3o} on the convergence of non-convex \fedavg under the third-order smoothness \cref{asm:fo:3o}. We restate the theorem below for ease of reference.
\ThmFedAvgNcvxTO*
\begin{proof}[Proof of \cref{thm:fedavg:ncvx:3o}]
%In what follows, all expectations are conditional on $\overline{\x}^{(r, 0)}$.
For simplicity of notation let $\g_m^{(r,k)} := \nabla f(\x_m^{(r,k)}; \xi_m^{(r,k)})$, namely the stochastic gradient of the $m$-th client taken at the $k$-th local step of the $r$-th round.
Define the shadow iterate $\overline{\x^{(r, k)}} := \frac{1}{M}\sum_{i = 1}^M\x_m^{(r, k)}$. The following claim bounds the expected difference $F(\overline{\x^{(r, k+1)}}) - F(\overline{\x^{(r, k)}})$. 
By $L$-smoothness, we have
\begin{align*}
    & \expt\left[F(\overline{\x^{(r, k+1)}})\right] = \expt\left[F\left(\overline{\x^{(r, k)}} - 
    \eta\frac{1}{M}\sum_m\g_m^{(r, k)}\right) \right]\\
    \leq & \expt\left[F(\overline{\x^{(r, k)}})\right] - \eta \expt\left[\langle{\nabla F(\overline{\x^{(r, k)}}), \frac{1}{M}\sum_m\g_m^{(r, k))}}\rangle\right] + \frac{L\eta^2}{M^2}\expt\left[\left\|\sum_m\g_m^{(r, k))}\right\|_2^2 \right] \\
    \leq & \expt\left[F(\overline{\x^{(r, k)}})\right] - \eta \expt\left[\langle{\nabla F(\overline{\x^{(r, k)}}), \frac{1}{M}\sum_m\nabla F(\x_m^{(r, k)})}\rangle\right] \\
    & \quad + \frac{L\eta^2}{M^2}\expt\left[\left\|\sum_m\nabla F(\x_m^{(r, k)})\right\|_2^2 \right] + \frac{L\eta^2 \sigma^2}{M}.\\
\end{align*}
Observe that for any real vectors, $\a$ and $\b$, we have $\langle{\a, \b}\rangle \geq \frac{1}{2}\|\a\|_2^2+ \frac{1}{2}\|\b\|_2^2- \|\a - \b\|_2^2 $.
Letting $\a := \nabla F(\overline{\x^{(r, k)}})$, and $\b := \frac{1}{M}\sum_m\nabla F(\x_m^{(r, k)})$, we obtain
\begin{equation}
    \label{eq:fedavg:ncvx:smooth:step}
\begin{split}
    & \expt\left[F(\overline{\x^{(r, k+1)}})\right] \leq \expt\left[F(\overline{\x^{(r, k)}})\right] - \frac{\eta}{2}\expt\left[\left\|\nabla F(\overline{\x^{(r, k)}})\right\|_2^2 \right] - \frac{\eta}{2}\expt\left[\left\|\frac{1}{M}\sum_m\nabla F(\x_m^{(r, k)})\right\|_2^2 \right]\\
    &\quad + \eta\expt\left[\left\|\nabla F(\overline{\x^{(r, k)}}) - \frac{1}{M}\sum_m\nabla F(\x_m^{(r, k)})\right\|_2^2 \right] + \frac{L\eta^2}{M^2}\expt\left[\left\|\sum_m\nabla F(\x_m^{(r, k)})\right\|_2^2 \right] + \frac{L\eta^2 \sigma^2}{M}\\
    &\leq \expt\left[F(\overline{\x^{(r, k)}})\right] - \frac{\eta}{2}\expt\left[\left\|\nabla F(\overline{\x^{(r, k)}})\right\|_2^2 \right] \\
    & \quad + \eta\expt\left[\left\|\nabla F(\overline{\x^{(r, k)}}) - \frac{1}{M}\sum_m\nabla F(\x_m^{(r, k)})\right\|_2^2 \right]  + \frac{L\eta^2 \sigma^2}{M},
\end{split}
\end{equation}
where the last inequality follows because $\eta \leq \frac{1}{L}$.

We will use third-order smoothness to bound $\expt\left[\left\|\nabla F(\overline{\x^{(r, k)}}) - \frac{1}{M}\sum_m\nabla F(\x_m^{(r, k)})\right\|_2^2 \right]$. By helper \cref{helper:3rd:Lip} we have
\begin{equation*}
  \left\| \nabla F(\overline{\x^{(r,k)}}) -  \frac{1}{M} \sum_{m=1}^M \nabla F(\x_m^{(r,k)}) \right\|_2^2 \leq \frac{Q^2}{4M} \sum_{m=1}^M \left\| \x^{(r,k)}_m - \overline{\x^{(r,k)}} \right\|_2^4.
\end{equation*}
% Plugging this into \cref{eq:fedavg:ncvx:smooth:step}, we obtain the claim:
By bounded stochastic gradient assumption we have 
\begin{equation*}
    \|\x_m^{(r,k)} - \overline{\x^{(r,k)}}\|_2^2\leq 4 \eta^2 G^2 K^2.
\end{equation*}
Consequently,
\begin{equation*}
     \expt\left[F(\overline{\x^{(r, k+1)}})\right] \leq \expt\left[F(\overline{\x^{(r, k)}})\right] - \frac{\eta}{2}\expt\left[\left\|\nabla F(\overline{\x^{(r, k)}})\right\|_2^2 \right] 
     + 4 \eta^5 G^4 K^4 Q^2
     + \frac{L\eta^2 \sigma^2}{M}.
\end{equation*}
Telescoping, for any $\eta \leq \frac{1}{L}$, we have
\begin{equation*}
    \frac{1}{KR} \sum_{r = 0}^{R-1} \sum_{k = 0}^{K-1} \expt\left[\left\|\nabla F(\overline{\x^{(r, k)}})\right\|_2^2 \right] \leq \frac{2(F(\x^{(0, 0)}) - F(\x^{\star}))}{\eta K R} + 8Q^2\eta^4 G^4K^4 + \frac{2L\eta \sigma^2}{M}.
\end{equation*}

Choosing 
\begin{equation*}
    \eta = \min\left\{\frac{1}{L}, \frac{\sqrt{M\Delta}}{\sigma\sqrt{LKR}}, \frac{\Delta ^{\frac{1}{5}}}{KR^{\frac{1}{5}}Q^{\frac{2}{5}} G^{\frac{4}{5}}}\right\},
\end{equation*}
we achieve the upper bound in \cref{thm:fedavg:ncvx:3o}.
\end{proof}

\subsection{Deferred Proof of \cref{thm:fedavg:ncvx:2o}}
\label{sec:pf:fedavg:ncvx:2o}
In this subsection we prove \cref{thm:fedavg:ncvx:2o}. The result is adapted from \cite{Yu.Yang.ea-AAAI19} which we include for completeness. We do not claim much novelty here.
\ThmFedAvgNcvxSO*
\begin{proof}[Proof of \cref{thm:fedavg:ncvx:2o}]
The proof is very similar to the $Q$-third order smooth case. Following the proof of \cref{thm:fedavg:ncvx:3o} in the previous section up to \cref{eq:fedavg:ncvx:smooth:step}, we obtain from $L$-smoothness:
\begin{align*}
    & \expt\left[F(\overline{\x^{(r, k+1)}})\right]
    \\
    \leq & \expt\left[F(\overline{\x^{(r, k)}})\right] - \frac{\eta}{2}\expt\left[\left\|\nabla F(\overline{\x^{(r, k)}})\right\|_2^2 \right] + \eta\expt\left[\left\|\nabla F(\overline{\x^{(r, k)}}) - \frac{1}{M}\sum_m\nabla F(\x_m^{(r, k)})\right\|_2^2 \right]  + \frac{L\eta^2 \sigma^2}{M},
\end{align*}
Now we invoke $L$-smoothness:
\begin{equation*}
    \expt\left[\left\|\nabla F(\overline{\x^{(r, k)}}) - \frac{1}{M}\sum_m\nabla F(\x_m^{(r, k)})\right\|_2^2 \right] 
    \leq \frac{L^2}{M}\sum_m\expt\left[\left\|\overline{\x^{(r, k)}} - \x_m^{(r, k)}\right\|_2^2 \right]
\end{equation*}
and by bounded stochastic gradient assumption
\begin{equation*}
    \left\|\overline{\x^{(r, k)}} - \x_m^{(r, k)} \right\|_2^2 \leq 4 \eta^2 G^2 k^2.
\end{equation*}
Consequently
\begin{equation*}
    \expt\left[F(\overline{\x^{(r, k+1)}})\right] \leq \expt\left[F(\overline{\x^{(r, k)}})\right] - \frac{\eta}{2}\expt\left[\left\|\nabla F(\overline{\x^{(r, k)}})\right\|_2^2 \right] + 4\eta^3 L^2 G^2k^2  + \frac{L\eta^2 \sigma^2}{M},
\end{equation*}
Telescoping, we obtain
\begin{equation*}
    \frac{1}{KR} \sum_{r = 0}^{R-1} \sum_{k = 0}^{K-1} \expt\left[\left\|\nabla F(\overline{\x^{(r, k)}})\right\|_2^2 \right] \leq \frac{2(F(\x^{(0, 0)}) - F(\x^{\star}))}{\eta K R} + 8L^2\eta^2 G^2K^2 + \frac{L\eta \sigma^2}{M}.
\end{equation*}
Choosing $\eta = \min\left\{\frac{1}{L}, \frac{\sqrt{\Delta M}}{\sigma\sqrt{LKR}}, \frac{\Delta^{\frac{1}{3}}}{KR^{\frac{1}{3}}L^\frac{2}{3}G^{\frac{2}{3}}}\right\}$, we obtain the upper bound in \cref{thm:fedavg:ncvx:2o}.
\end{proof}

\section{Miscellaneous Helper Lemmas}
In this section we list some miscellaneous technical helper lemmas used throughout the chapter.

\begin{lemma}
  \label{helper:3rd:Lip}
  Let $F$: $\reals^d \to \reals$ be an arbitrary twice-continuous-differentiable function that is $Q$-3rd-order-smooth. Then for any $\x^1, \ldots, \x^M \in \reals^d$, the following inequality holds
  \begin{equation*}
      \left\|  \nabla F (\overline{\x}) - \frac{1}{M} \sum_{m=1}^M \nabla F(\x_m)  \right\|_2^2 
      \leq
      \frac{Q^2}{4M}  \sum_{m=1}^M\left\| \x_m - \overline{\x} \right\|_2^4,
  \end{equation*}
  where $\overline{\x} := \frac{1}{M} \sum_{m=1}^M \x_m$.
\end{lemma}
\begin{proof}[Proof of \cref{helper:3rd:Lip}]
  \begin{align}
           & \left\|  \frac{1}{M} \sum_{m=1}^M \nabla F(\x_m) - \nabla F (\overline{\x})    \right\|_2^2 
           \nonumber
      \\
      =    & \left\|  \frac{1}{M} \sum_{m=1}^M \left( \nabla F(\x_m) - \nabla F (\overline{\x}) - \nabla^2 F (\overline{\x})(\x_m - \overline{\x}) \right)   \right\|_2^2 
      \tag{since $\frac{1}{M} \sum_{m=1}^M \x_m - \overline{\x} = 0$}
      \\
      \leq & \frac{1}{M} \sum_{m=1}^M\left\|   \nabla F(\x_m) - \nabla F (\overline{\x}) - \nabla^2 F (\overline{\x})(\x_m - \overline{\x}) \right\|_2^2 
      \tag{Jensen's inequality}
      \\
      \leq & \frac{Q^2}{4M}  \sum_{m=1}^M\left\| \x_m - \overline{\x} \right\|_2^4.
      \tag{$Q$-3rd-order-smoothness}
  \end{align}
\end{proof}

\begin{lemma}
  \label{helper:diff:4th}
  Let $\x$ and $\y$ be two i.i.d. $\reals^d$-valued random vectors, and assume $\expt \x = 0$, $\expt \|\x\|_2^4 \leq \sigma^4$. Then
  \begin{equation*}
      \expt \|\x+\y\|_2^2\leq 2 \sigma^2, \quad \expt \|\x+\y\|_2^3 \leq 4 \sigma^3, \quad \expt \|\x+\y\|_2^4 \leq 8 \sigma^4.
  \end{equation*}
\end{lemma}
\begin{proof}[Proof of \cref{helper:diff:4th}]
  The first inequality is due to $ \expt \|\x+\y\|_2^2= \expt \|\x\|_2^2+  \expt \|\y\|_2^2= 2\sigma^2$ where $\expt \|\x\|_2^2\leq \sigma^2$ follows by applying H\"older's inequality to the assumption $\expt \|\x\|_2^4 \leq \sigma^4$.

  The \nth{4} moment is bounded as
  \begin{align}
           & \expt \|\x+\y\|_2^4 = \expt \left[ \|\x\|_2^2+ \|\y\|_2^2+ 2 \langle \x,  \y \rangle \right]^2
           \nonumber
      \\
      =    & \expt \left[ \|\x\|_2^4 + \|\y\|_2^4 + 2 \|\x\|_2^2\|\y\|_2^2+ 4 \langle \x,  \y \rangle^2 + 4 \|\x\|_2^2\langle \x,  \y \rangle  + 4\|\y\|_2^2\langle \x,  \y \rangle  \right]
      \nonumber
      \\
      =    & \expt \left[ \|\x\|_2^4 + \|\y\|_2^4 + 2 \|\x\|_2^2\|\y\|_2^2+ 4 \langle \x,  \y \rangle^2 \right] \tag{by independence and mean-zero assumption}
      \\
      \leq & \expt \left[ 4\|\x\|_2^4 + 4\|\y\|_2^4 \right] \tag{Cauchy-Schwarz inequality} \leq 8 \sigma^4.
  \end{align}

  The \nth{3} moment is bounded via Cauchy-Schwarz inequality since
  \begin{equation*}
      \expt \|\x+\y\|_2^3 \leq \sqrt{\expt \|\x+\y\|_2^2\expt \|\x+\y\|_2^4} \leq 4 \sigma^3.
  \end{equation*}
\end{proof}

\chapter{Appendix of Chapter 3}

\label{apx:fedac}
The \cref{apx:fedac} is structured as follows. 
In \cref{sec:fedacii}, we prove the complete version of \cref{fedac:a1,fedacii:a2} on the convergence of \fedacii under \cref{asm:fo:scvx:2o} or \ref{asm:fo:scvx:3o}.
In \cref{sec:fedavg}, we prove \cref{fedavg:a2} on the convergence of \fedavg under \cref{asm:fo:scvx:3o}. In \cref{sec:gcvx}, we  prove the convergence of \fedac (and \fedavg) for general convex objectives. 
% In \cref{sec:instability}, we prove \cref{thm:agd:instability}  on the initial-value instability of standard accelerated gradient descent. 
We include some helper lemmas in \cref{sec:helper}.

% \input{fedac/additional_expr}
% \section{Analysis of \fedaci under \cref{asm:fo:scvx:2o}}
% \label{sec:fedaci}

% \subsection{Main theorem and lemmas: Complete version of \cref{fedac:a1}(a)}

% % \hynote{The proof of \cref{fedaci:full} is based on the following two lemmas regarding convergence and stability respectively.}

\section{Analysis of \fedacii under \cref{asm:fo:scvx:2o} or \ref{asm:fo:scvx:3o}}
\label{sec:fedacii}
In this section we study the convergence of \fedacii.
We provide a complete, non-asymptotic version of \cref{fedacii:a2} on the convergence of \fedacii under \cref{asm:fo:scvx:3o} and provide the detailed proof, which expands the proof sketch in \cref{sec:proof:sketch:2}. 
We also study the convergence of \fedacii under \cref{asm:fo:scvx:2o}, which we defer to the end of this section (see \cref{sec:fedacii:a1}) since the analysis is mostly shared.

Recall that \fedacii is defined as the \fedac algorithm with the following hyperparameter choice:
\begin{equation}
  \eta \in \left(0, \frac{1}{L}\right], \quad  \gamma = \max \left\{ \sqrt{\frac{\eta}{\mu K}}, \eta \right\}, \quad \alpha  = \frac{3}{2 \gamma \mu} - \frac{1}{2}, \quad \beta = \frac{2 \alpha^2 - 1}{\alpha - 1}
\tag{\fedacii}.
\end{equation}
As we discussed in the proof sketch \cref{sec:proof:sketch:2}, for \fedacii, we keep track of the convergence via the ``centralized'' potential $\Phi^{(r,k)}$.
\begin{equation}
  \Phi^{(r,k)} := F( \overline{\x^{(r,k)}_{\mathrm{ag}}})  - F^{\star} + \frac{1}{6}\mu \|\overline{\x^{(r,k)}} - \x^{\star}\|_2^2 .
  \tag{\ref{eq:centralied:potential}}
\end{equation}
Recall $\overline{\x^{(r,k)}}$ is defined as $\frac{1}{M} \sum_{m=1}^M \x^{(r,k)}_m$ and $\overline{\x^{(r,k)}_{\mathrm{ag}}}$ is defined as $\frac{1}{M} \sum_{m=1}^M \x^{(r,k)}_{\mathrm{ag}, m}$. 
Formally, we use $\mathcal{F}^{(r,k)}$ to denote the $\sigma$-algebra generated by $\{\x^{(\rho,\kappa)}_m, \x^{(\rho,\kappa)}_{\mathrm{ag},m}\}$ for $\rho < r$ or $\rho = r$ but $\kappa \leq k$. Since \fedac is Markovian, conditioning on $\mathcal{F}^{(r,k)}$ is equivalent to conditioning on $\{\x_m^{(r,k)}, \x_{\mathrm{ag}, m}^{(r,k)}\}_{m \in [M]}$.

\subsection{Main theorem and lemmas: Complete version of \cref{fedacii:a2}}
Now we introduce the main theorem on the convergence of \fedacii under \cref{asm:fo:scvx:3o}.
\begin{theorem}[Convergence of \fedacii under \cref{asm:fo:scvx:3o}, complete version of \cref{fedacii:a2}]
  \label{fedacii:a2:full}
  Let $F$ be $\mu > 0$ strongly convex, and assume \cref{asm:fo:scvx:3o}, then for 
  \begin{equation*}
    \eta := \min \left\{ \frac{1}{L}, \frac{9}{\mu K R^2} \log^2 \left( \euler
     + \min \left\{ \frac{\mu M K R \Phi^{(0,0)}}{\sigma^2} + \frac{\mu^2 M K R^3 \Phi^{(0,0)}}{L \sigma^2}, \frac{\mu^5 K^2 R^8 \Phi^{(0,0)}}{Q^2 \sigma^4} \right\} \right) \right\},
  \end{equation*}
  \fedacii yields
  \begin{align*}
    \expt [\Phi^{(R,0)}] \leq & \min \left\{ \exp \left( - \frac{\mu K R}{3 L} \right), \exp \left( - \frac{\mu^{\frac{1}{2}} K^{\frac{1}{2}} R}{3 L^{\frac{1}{2}} } \right)\right\} \Phi^{(0,0)} 
    + \frac{4 \sigma^2}{\mu M K R} \log \left( \euler + \frac{\mu M K R \Phi^{(0,0)}}{\sigma^2} \right)
    \\
    & 
    \quad +
    \frac{55 L \sigma^2}{\mu^2 M K R^3} \log^3 \left( \euler + \frac{\mu^2 M K R^3 \Phi^{(0,0)}}{L \sigma^2} \right)
    + \frac{\euler^{18} Q^2 \sigma^4}{\mu^5 K^2 R^8} \log^8\left( \euler
    + \frac{\mu^5 K^2 R^8 \Phi^{(0,0)}}{Q^2 \sigma^4} \right),
  \end{align*}
  where $\Phi^{(r,k)}$ is the ``centralized'' potential defined in \cref{eq:centralied:potential}.
\end{theorem}
\begin{remark}
  The simplified version \cref{fedacii:a2} in main body can be obtained by upper bounding $\Phi^{(0,0)}$ by $LB^2$.
\end{remark}

The proof of \cref{fedacii:a2:full} is based on the following two lemmas regarding convergence and stability respectively.
To clarify the hyperparameter dependency, we state our lemma for general $\gamma \in \left[\eta, \sqrt{ \frac{\eta}{\mu}} \right]$, which has one more degree of freedom than \fedacii where $\gamma = \max \left\{ \sqrt{\frac{\eta}{\mu K}}, \eta\right\}$ is fixed.

\FedAcIIConvMain*
The proof of \cref{fedacii:conv:main} is deferred to \cref{sec:fedacii:conv:main}. Note that \cref{fedacii:conv:main} only requires \cref{asm:fo:scvx:2o} (recall that \cref{asm:fo:scvx:2o} is strictly weaker than \cref{asm:fo:scvx:3o}), which enables us to recycle this Lemma towards the convergence proof of \fedacii under \cref{asm:fo:scvx:2o} (see \cref{sec:fedacii:a1}).

The following lemma studies the discrepancy overhead by \nth{4}-th order stability, which requires \cref{asm:fo:scvx:3o}.

\FedAcIIStabMain*

The proof of \cref{fedacii:stab:main} is deferred to \cref{sec:fedacii:stab:main}.

Now we plug in the choice of $\gamma = \max \left\{ \sqrt{\frac{\eta}{\mu K}}, \eta\right\}$ to \cref{fedacii:conv:main,fedacii:stab:main}, which leads to  the following lemma.
\begin{lemma}[Convergence of \fedacii for general $\eta$]
  \label{fedacii:a2:general:eta}
    Let $F$ be $\mu > 0$-strongly convex, and assume \cref{asm:fo:scvx:3o}, then for any $\eta \in (0, \frac{1}{L}]$, \fedacii yields
    \begin{align}
      \expt[\Phi^{(R,0)}]
      \leq & \exp \left(  - \frac{1}{3} \max \left\{ \eta \mu, \sqrt{\frac{\eta \mu}{K}}\right\} K R \right) \Phi^{(0,0)}
      + \frac{\eta^{\frac{1}{2}} \sigma^2}{\mu^{\frac{1}{2}} M K^{\frac{1}{2}} }
      + \frac{2 \eta^{\frac{3}{2}} L K^{\frac{1}{2}} \sigma^2}{\mu^{\frac{1}{2}} M}
      + \frac{\euler^9 \eta^4 Q^2 K^2 \sigma^4}{\mu},
      \label{eq:fedacii:a2:general:eta}
    \end{align}
    where $\Phi^{(r,k)}$ is the decentralized potential defined in \cref{eq:centralied:potential}.
\end{lemma}
\begin{proof}[Proof of \cref{fedacii:a2:general:eta}]
  It is direct to verify that $\gamma = \max \left\{\eta, \sqrt{\frac{\eta}{\mu K}} \right\} \in \left[\eta, \sqrt{\frac{\eta}{\mu}}\right]$ so both \cref{fedacii:conv:main,fedacii:stab:main} are applicable.
  Applying \cref{fedacii:conv:main} yields
  \begin{align}
    \expt[\Phi^{(R,0)}] \leq & \exp\left( - \frac{1}{3} \max \left\{ \eta \mu, \sqrt{\frac{\eta \mu}{K}}\right\} K R \right) \Phi^{(0,0)}
    +
    \min\left\{ \frac{3\eta L \sigma^2}{2 \mu M}, \frac{3 \eta^{\frac{3}{2}} L K^{\frac{1}{2}} \sigma^2}{2 \mu^{\frac{1}{2}} M} \right\}
    \nonumber \\
                       & + \max\left\{ \frac{\eta \sigma^2}{2M}, \frac{\eta^{\frac{1}{2}} \sigma^2}{2 \mu^{\frac{1}{2}} M K^{\frac{1}{2}} }          \right\}
    + \frac{3}{\mu} \max_{\substack{0 \leq r < R \\ 0 \leq k < K}}  \expt \left[ \left\|  \nabla F(\overline{\x^{(r,k)}_{\mathrm{md}}}) -  \frac{1}{M} \sum_{m=1}^M \nabla F (\x^{(r,k)}_{\mathrm{md}, m})  \right\|_2^2\right].
    \label{eq:fedacii:proof:1}
  \end{align}
  We bound $\min\left\{ \frac{3\eta L \sigma^2}{2 \mu M}, \frac{3 \eta^{\frac{3}{2}} L K^{\frac{1}{2}} \sigma^2}{2 \mu^{\frac{1}{2}} M} \right\}$ with $\frac{3 \eta^{\frac{3}{2}} L K^{\frac{1}{2}} \sigma^2}{2 \mu^{\frac{1}{2}} M}$, and bound $\max\left\{ \frac{\eta \sigma^2}{2M}, \frac{\eta^{\frac{1}{2}} \sigma^2}{2 \mu^{\frac{1}{2}} M K^{\frac{1}{2}} }          \right\}$ with $\frac{\eta \sigma^2}{2M} + \frac{\eta^{\frac{1}{2}} \sigma^2}{2 \mu^{\frac{1}{2}} M K^{\frac{1}{2}} }$.
  By AM-GM inequality and $\mu \leq L$, we have 
  \begin{equation*}
    \frac{\eta \sigma^2}{2M} 
    \leq 
    \frac{ \eta^{\frac{3}{2}} \mu^{\frac{1}{2}} K^{\frac{1}{2}} \sigma^2}{4 M} + \frac{ \eta^{\frac{1}{2}} \sigma^2}{4 \mu^{\frac{1}{2}} M K^{\frac{1}{2}} }
    \leq
    \frac{ \eta^{\frac{3}{2}} L K^{\frac{1}{2}} \sigma^2}{4\mu^{\frac{1}{2}} M} + \frac{ \eta^{\frac{1}{2}} \sigma^2}{4 \mu^{\frac{1}{2}} M K^{\frac{1}{2}} }
  \end{equation*}
  Thus
  \begin{align}
         & \min\left\{ \frac{3\eta L \sigma^2}{2 \mu M}, \frac{3 \eta^{\frac{3}{2}} L K^{\frac{1}{2}} \sigma^2}{2 \mu^{\frac{1}{2}} M} \right\}
    + \max\left\{ \frac{\eta \sigma^2}{2M}, \frac{\eta^{\frac{1}{2}} \sigma^2}{2 \mu^{\frac{1}{2}} M K^{\frac{1}{2}} }          \right\}
    \nonumber \\
    \leq & \frac{3 \eta^{\frac{3}{2}} L K^{\frac{1}{2}} \sigma^2}{2 \mu^{\frac{1}{2}} M} + \frac{\eta \sigma^2}{2M} + \frac{\eta^{\frac{1}{2}} \sigma^2}{2 \mu^{\frac{1}{2}} M K^{\frac{1}{2}} }
    \leq
    \frac{7 \eta^{\frac{3}{2}} L K^{\frac{1}{2}} \sigma^2}{4 \mu^{\frac{1}{2}} M} + \frac{3 \eta^{\frac{1}{2}} \sigma^2}{4 \mu^{\frac{1}{2}} M K^{\frac{1}{2}} },
    \label{eq:fedacii:proof:1:1}
  \end{align}

  Applying \cref{fedacii:stab:main} yields (for all $r, k$)
  \begin{align}
         & \frac{3}{\mu} \expt \left[ \left\|  \nabla F(\overline{\x^{(r,k)}_{\mathrm{md}}}) -  \frac{1}{M} \sum_{m=1}^M \nabla F (\x^{(r,k)}_{\mathrm{md}, m})  \right\|_2^2\right]
    \leq
    \begin{cases}
      \frac{132}{\mu} \eta^4 Q^2 K^2 \sigma^4  \left(1 + \frac{1}{K}\right)^{4K}
       & \text{if~} \gamma = \sqrt{\frac{\eta}{\mu K}}
      \\
      \frac{132}{\mu} \eta^4 Q^2 K^2 \sigma^4,
       &
      \text{if~} \gamma = \eta
    \end{cases}
    \nonumber \\
    \leq & 132 \euler^{4} \mu^{-1} \eta^4 Q^2 K^2 \sigma^4 \leq \euler^9 \mu^{-1} \eta^4 Q^2 K^2 \sigma^4,
    \label{eq:fedacii:proof:2}
  \end{align}
  where in the last inequality we used the estimation that $132 \euler^{4} < \euler^9$.

  Combining \cref{eq:fedacii:proof:1,eq:fedacii:proof:1:1,eq:fedacii:proof:2} yields
  \begin{align*}
    \expt[\Phi^{(R,0)}]
    \leq & \exp \left(  - \frac{1}{3} \max \left\{ \eta \mu, \sqrt{\frac{\eta \mu}{K}}\right\} K R \right) \Phi^{(0,0)}
    + \frac{\eta^{\frac{1}{2}} \sigma^2}{\mu^{\frac{1}{2}} M K^{\frac{1}{2}} }
    + \frac{2 \eta^{\frac{3}{2}} L K^{\frac{1}{2}} \sigma^2}{\mu^{\frac{1}{2}} M}
    + \frac{\euler^9 \eta^4 Q^2 K^2 \sigma^4}{\mu}.
  \end{align*}
\end{proof}

The main \cref{fedacii:a2:full} then follows by plugging the appropriate $\eta$ to \cref{fedacii:a2:general:eta}. 
\begin{proof}[Proof of \cref{fedacii:a2:full}]
  To simplify the notation, we denote the decreasing term in \cref{eq:fedacii:a2:general:eta} in \cref{fedacii:a2:general:eta} as $\varphi_{\downarrow}(\eta)$ and the increasing term as $\varphi_{\uparrow}(\eta)$, namely
  \begin{align*}
    \varphi_{\downarrow}(\eta) & := \exp \left(  - \frac{1}{3} \max \left\{ \eta \mu, \sqrt{\frac{\eta \mu}{K}}\right\} K R \right) \Phi^{(0,0)},
    \\
    \varphi_{\uparrow}(\eta) & := \frac{\eta^{\frac{1}{2}} \sigma^2}{\mu^{\frac{1}{2}} M K^{\frac{1}{2}} }
    + \frac{2 \eta^{\frac{3}{2}} L K^{\frac{1}{2}} \sigma^2}{\mu^{\frac{1}{2}} M}
    + \frac{\euler^9 \eta^4 Q^2 K^2 \sigma^4}{\mu}.
  \end{align*}
  Now let
  \begin{equation*}
    \eta_0 := \frac{9}{\mu K R^2} \log^2 \left( \euler
     + \min \left\{ \frac{\mu M K R \Phi^{(0,0)}}{\sigma^2} + \frac{\mu^2 M K R^3 \Phi^{(0,0)}}{L \sigma^2}, \frac{\mu^5 K^2 R^8 \Phi^{(0,0)}}{Q^2 \sigma^4} \right\} \right)
  \end{equation*}
  then $\eta := \min \left\{ \frac{1}{L}, \eta_0 \right\}$. Therefore, the decreasing term $\varphi_{\downarrow}(\eta)$ is upper bounded by $\varphi_{\downarrow}(\frac{1}{L}) + \varphi_{\downarrow}(\eta_0)$, where
  \begin{equation}
    \varphi_{\downarrow} \left(\frac{1}{L} \right)
    \leq
    \min \left\{ \exp \left( - \frac{\mu K R}{3 L} \right), \exp \left( - \frac{\mu^{\frac{1}{2}} K^{\frac{1}{2}} R}{3 L^{\frac{1}{2}} } \right)\right\} \Phi^{(0,0)},
    \label{eq:fedacii:a2:1}
  \end{equation}
  and
  \begin{align}
    \varphi_{\downarrow}(\eta_0) 
    \leq & \exp \left(  - \frac{1}{3} \sqrt{\eta_0 \mu K R^2} \right) \Phi^{(0,0)}
    \nonumber \\
    = &
    \left( \euler
     + \min \left\{ \frac{\mu M K R \Phi^{(0,0)}}{\sigma^2} + \frac{\mu^2 M K R^3 \Phi^{(0,0)}}{L \sigma^2}, \frac{\mu^5 K^2 R^8 \Phi^{(0,0)}}{Q^2 \sigma^4} \right\} \right)^{-1} \Phi^{(0,0)}
     \nonumber \\
    \leq & \frac{\sigma^2}{\mu M K R} + \frac{L \sigma^2}{\mu^2 M K R^3} + \frac{Q^2 \sigma^4}{\mu^5 K^2 R^8}.
    \label{eq:fedacii:a2:2}
  \end{align}
  On the other hand
  \begin{align}
    \varphi_{\uparrow}(\eta)  \leq \varphi_{\uparrow}(\eta_0) 
    \leq & 
    \frac{3 \sigma^2}{\mu M K R} \log \left( \euler + \frac{\mu M K R \Phi^{(0,0)}}{\sigma^2} \right)
    +
    \frac{54 L \sigma^2}{\mu^2 M K R^3} \log^3 \left( \euler + \frac{\mu^2 M K R^3 \Phi^{(0,0)}}{L \sigma^2} \right)
    \nonumber  \\
    & + \frac{9^4 \euler^9 Q^2 \sigma^4}{\mu^5 K^2 R^8} \log^8\left( \euler
    + \frac{\mu^5 K^2 R^8 \Phi^{(0,0)}}{Q^2 \sigma^4} \right).
    \label{eq:fedacii:a2:3}
  \end{align}
  Combining \cref{fedacii:a2:general:eta,eq:fedacii:a2:1,eq:fedacii:a2:2,eq:fedacii:a2:3} gives
  \begin{align*}
    & \expt[\Phi^{(R,0)}] \leq \varphi_{\downarrow} \left(\frac{1}{L} \right) + \varphi_{\downarrow}(\eta_0) + \varphi_{\uparrow}(\eta_0) 
    \\
    \leq & \min \left\{ \exp \left( - \frac{\mu K R}{3 L} \right), \exp \left( - \frac{\mu^{\frac{1}{2}} K^{\frac{1}{2}} R}{3 L^{\frac{1}{2}} } \right)\right\} \Phi^{(0,0)} 
    + \frac{4 \sigma^2}{\mu M K R} \log \left( \euler + \frac{\mu M K R \Phi^{(0,0)}}{\sigma^2} \right)
    \\
    & 
    +
    \frac{55 L \sigma^2}{\mu^2 M K R^3} \log^3 \left( \euler + \frac{\mu^2 M K R^3 \Phi^{(0,0)}}{L \sigma^2} \right)
    + \frac{\euler^{18} Q^2 \sigma^4}{\mu^5 K^2 R^8} \log^8\left( \euler
    + \frac{\mu^5 K^2 R^8 \Phi^{(0,0)}}{Q^2 \sigma^4} \right),
  \end{align*}
  where in the last inequality we used the estimate $9^4 \euler^9 + 1 < \euler^{18}$.
\end{proof}

\subsection{Perturbed iterate analysis for \fedacii: Proof of \cref{fedacii:conv:main}}
\label{sec:fedacii:conv:main}
In this subsection we will prove \cref{fedacii:conv:main}.
We start by the one-step analysis of the centralized potential defined in \cref{eq:centralied:potential}. 
The following two propositions establish the one-step analysis of the two quantities in $\Phi^{(r,k)}$, namely $\|\overline{\x^{(r,k)}} - \x^{\star}\|_2^2 $ and $F( \overline{\x^{(r,k)}_{\mathrm{ag}}}) - F^{\star}$. 
We only require minimal hyperparameter assumptions, namely $\alpha \geq 1, \beta \geq 1, \eta \leq \frac{1}{L}$ for these two propositions.
We will then show how the choice of $\alpha, \beta$ are determined towards the proof of \cref{fedacii:conv:main} in order to couple the two quantities into potential $\Phi^{(r,k)}$.

\begin{proposition}
  \label{fedacii:conv:1}
  Let $F$ be $\mu>0$-strongly convex, and assume \cref{asm:fo:scvx:2o}, then for \fedac with hyperparameters assumptions $\alpha \geq 1$, $\beta \geq 1$, $\eta \leq \frac{1}{L}$, the following inequality holds
  \begin{align}
         & \expt [ \|\overline{\x^{(r,k+1)}} - \x^{\star}\|_2^2  |\mathcal{F}^{(r,k)}]
    \nonumber \\
    \leq & \left( 1 - \frac{1}{2} \alpha^{-1} \right)\|\overline{\x^{(r,k)}} - \x^{\star}\|_2^2  + \frac{3}{2} \alpha^{-1} \| \overline{\x^{(r,k)}_{\mathrm{md}}} - \x^{\star}\|_2^2  + \frac{3}{2} \gamma^2  \left\| \nabla F (\overline{\x^{(r,k)}_{\mathrm{md}}}) \right\|_2^2 
    \nonumber \\
         & - 2 \gamma  \left( 1 + \frac{1}{2} \alpha^{-1} \right) \left\langle \nabla F (\overline{\x^{(r,k)}_{\mathrm{md}}}) , (1-\alpha^{-1}(1-\beta^{-1})) \overline{\x^{(r,k)}} + \alpha^{-1} (1 - \beta^ {-1}) \overline{\x^{(r,k)}_{\mathrm{ag}}} - \x^{\star}\right\rangle
    \nonumber \\
         &
    + \gamma^2 \left( 1 + 2 \alpha \right)  \left\| \nabla F (\overline{\x^{(r,k)}_{\mathrm{md}}}) - \frac{1}{M} \sum_{m=1}^M \nabla F (\x^{(r,k)}_{\mathrm{md}, m}) \right\|_2^2 
    +  \frac{\gamma^2 \sigma^2}{M}.
    \label{eq:fedacii:conv:1}
  \end{align}
\end{proposition}

\begin{proposition}
  \label{fedacii:conv:2}
  In the same setting of \cref{fedacii:conv:1}, the following inequality holds
  \begin{align}
         & \expt \left[F( \overline{\x^{(r,k+1)}_{\mathrm{ag}}}) - F^{\star}| \mathcal{F}^{(r,k)} \right]
    \nonumber \\
    \leq & \left( 1 - \frac{1}{2} \alpha^{-1} \right) \left(F(\overline{\x^{(r,k)}_{\mathrm{ag}}}) - F^{\star} \right) - \frac{1}{4} \mu \alpha^{-1} \left\|  \overline{\x^{(r,k)}_{\mathrm{md}}} - \x^{\star}  \right\|_2^2- \frac{1}{2} \eta \left\| \nabla F(\overline{\x^{(r,k)}_{\mathrm{md}}}) \right\|_2^2 
    \nonumber \\
         & + {\frac{1}{2} \alpha^{-1} \left\langle \nabla F(\overline{\x^{(r,k)}_{\mathrm{md}}}), 2 \alpha \beta^{-1} \overline{\x^{(r,k)}} + (1 - 2 \alpha \beta^{-1}) \overline{\x^{(r,k)}_{\mathrm{ag}}} -  \x^{\star} \right\rangle}
    \nonumber \\
         & + \frac{1}{2}  \eta \left\|  \nabla F(\overline{\x^{(r,k)}_{\mathrm{md}}}) -  \frac{1}{M} \sum_{m=1}^M \nabla F (\x^{(r,k)}_{\mathrm{md}, m})  \right\|_2^2+  \frac{\eta^2 L \sigma^2 }{2M}.
    \label{eq:fedacii:conv:2}
  \end{align}
\end{proposition}
We defer the proofs of \cref{fedacii:conv:1,fedacii:conv:2} to \cref{sec:proof:fedacii:conv:1,sec:proof:fedacii:conv:2}, respectively.

Now we are ready to prove \cref{fedacii:conv:main}.
\begin{proof}[Proof of \cref{fedacii:conv:main}]
  Since $\gamma \leq \sqrt{\frac{\eta}{\mu}} \leq \sqrt{\frac{1}{\mu L}} \leq \frac{1}{\mu}$, we have $\alpha = \frac{3}{2 \gamma \mu} - \frac{1}{2} \geq 1$, and therefore $\beta = \frac{2 \alpha^2 - 1}{\alpha - 1} \geq 1$. Hence both \cref{fedacii:conv:1,fedacii:conv:2} are applicable.

  Adding \cref{eq:fedacii:conv:2} with $\frac{1}{6} \mu$ times of \cref{eq:fedacii:conv:1} gives (note that the $\|\overline{\x^{(r,k)}_{\mathrm{md}}} - \x^{\star}\|_2^2 $ term is cancelled because $\frac{1}{4}\mu \alpha^{-1} = \frac{1}{6} \mu \cdot \frac{3}{2}\alpha^{-1}$)
  \begin{align}
     & \expt \left[ \Phi^{(r,k+1)} |\mathcal{F}^{(r,k)} \right] 
     \leq 
    \underbrace{\left( 1 - \frac{1}{2} \alpha^{-1} \right) \Phi^{(r,k)}}_{\text{(I)}}
    + 
    \underbrace{\left( \frac{1}{4} \gamma^2 \mu  - \frac{1}{2} \eta  \right) \left\| \nabla F(\overline{\x^{(r,k)}_{\mathrm{md}}}) \right\|_2^2 }_{\text{(II)}}
    \nonumber \\
     & \quad + 
    \underbrace{\frac{1}{2} \alpha^{-1} \left\langle \nabla F(\overline{\x^{(r,k)}_{\mathrm{md}}}), 2 \alpha \beta^{-1} \overline{\x^{(r,k)}} + (1 - 2 \alpha \beta^{-1}) \overline{\x^{(r,k)}_{\mathrm{ag}}} -  \x^{\star} \right\rangle}_{\text{(III)}}
    \nonumber \\
     & \quad - 
    \underbrace{\frac{1}{3} \gamma \mu \left( 1 + \frac{1}{2} \alpha^{-1} \right) \left\langle \nabla F (\overline{\x^{(r,k)}_{\mathrm{md}}}) , (1-\alpha^{-1}(1-\beta^{-1})) \overline{\x^{(r,k)}} + \alpha^{-1} (1 - \beta^ {-1}) \overline{\x^{(r,k)}_{\mathrm{ag}}} - \x^{\star}\right\rangle}_{\text{(IV)}}
    \nonumber \\
     & \quad + 
    \underbrace{\left( \frac{1}{2}  \eta  + \frac{1}{6} \gamma^2 \mu (1 + 2 \alpha)  \right) \left\|  \nabla F(\overline{\x^{(r,k)}_{\mathrm{md}}}) -  \frac{1}{M} \sum_{m=1}^M \nabla F (\x^{(r,k)}_{\mathrm{md}, m})  \right\|_2^2 }_{\text{(V)}}
    +  \frac{\eta^2 L \sigma^2 }{2M} + \frac{\gamma^2 \mu \sigma^2}{6 M}.
    \label{eq:fedacii:conv:main:1}
  \end{align}

  Now we analyze the RHS of \cref{eq:fedacii:conv:main:1} term by term.
  \paragraph{Term (I) of \cref{eq:fedacii:conv:main:1}}
  Note that $\alpha^{-1} = \frac{2 \gamma \mu}{3 - \gamma \mu} \geq \frac{2}{3}\gamma \mu$, we have
  \begin{equation}
    \left( 1 - \frac{1}{2} \alpha^{-1} \right) \Phi^{(r,k)} \leq \left( 1 - \frac{1}{3} \gamma \mu \right) \Phi^{(r,k)}.
    \label{eq:fedacii:conv:main:2}
  \end{equation}

  \paragraph{Term (II) of \cref{eq:fedacii:conv:main:1}}
  Since $\gamma^2 \mu \leq \eta$ we have
  \begin{equation}
    \left( \frac{1}{4} \gamma^2 \mu  - \frac{1}{2} \eta  \right) \left\| \nabla F(\overline{\x^{(r,k)}_{\mathrm{md}}}) \right\|_2^2\leq 0.
    \label{eq:fedacii:conv:main:3}
  \end{equation}

  \paragraph{Term (III) and (IV) of \cref{eq:fedacii:conv:main:1}}
  Since $\beta =\frac{2 \alpha^2 - 1}{\alpha - 1}$, we have $2 \alpha \beta^{-1} = \frac{2 \alpha (\alpha - 1)}{2 \alpha^2 - 1} = (1 - \alpha^{-1} (1 - \beta^{-1}))$, and $1 - 2\alpha\beta^{-1} = \frac{2 \alpha - 1}{2 \alpha^2 - 1} = \alpha^{-1}(1 - \beta^{-1})$. Therefore, the two inner-product terms are cancelled:
  \begin{align}
      & \frac{1}{2} \alpha^{-1} \left\langle \nabla F(\overline{\x^{(r,k)}_{\mathrm{md}}}), 2 \alpha \beta^{-1} \overline{\x^{(r,k)}} + (1 - 2 \alpha \beta^{-1}) \overline{\x^{(r,k)}_{\mathrm{ag}}} -  \x^{\star} \right\rangle
      \nonumber \\
      & \qquad - \frac{1}{3} \gamma \mu \left( 1 + \frac{1}{2} \alpha^{-1} \right) \left\langle \nabla F (\overline{\x^{(r,k)}_{\mathrm{md}}}) , (1-\alpha^{-1}(1-\beta^{-1})) \overline{\x^{(r,k)}} + \alpha^{-1} (1 - \beta^ {-1}) \overline{\x^{(r,k)}_{\mathrm{ag}}} - \x^{\star}\right\rangle
      \nonumber \\
    = & \left( \frac{1}{2} \alpha^{-1} - \frac{1}{3} \gamma \mu  \left( 1 + \frac{1}{2} \alpha^{-1} \right) \right)   \left\langle \nabla F (\overline{\x^{(r,k)}_{\mathrm{md}}}) , \frac{2 \alpha - 1}{2 \alpha^2 - 1} \overline{\x^{(r,k)}_{\mathrm{ag}}} + \left(\frac{2 \alpha^2 - 2 \alpha}{2 \alpha^2 - 1}  \right) \overline{\x^{(r,k)}}  - \x^{\star}\right\rangle
    \nonumber \\
    = & \left( \frac{\gamma \mu}{3 - \gamma \mu} - \frac{1}{3} \gamma \mu  \left( 1 + \frac{\gamma \mu}{3 - \gamma \mu} \right) \right)   \left\langle \nabla F (\overline{\x^{(r,k)}_{\mathrm{md}}}) , \frac{2 \alpha - 1}{2 \alpha^2 - 1} \overline{\x^{(r,k)}_{\mathrm{ag}}} + \left(\frac{2 \alpha^2 - 2 \alpha}{2 \alpha^2 - 1}  \right) \overline{\x^{(r,k)}}  - \x^{\star}\right\rangle
    \tag{since $\alpha^{-1} = \frac{2 \gamma \mu}{3 - \gamma \mu}$}
    \nonumber \\
    = & \zeros.
    \label{eq:fedacii:conv:main:4}
  \end{align}

  \paragraph{Term (V) of \cref{eq:fedacii:conv:main:1}} Since $\alpha = \frac{3 - \gamma \mu}{2 \gamma \mu}$ and $\gamma \geq \eta$ we have
  \begin{equation}
    \left( \frac{1}{2}  \eta  + \frac{1}{6} \gamma^2 \mu (1 + 2 \alpha)  \right)
    = \frac{1}{2} \eta + \frac{1}{6} \gamma^2 \mu \left( \frac{6}{2 \gamma \mu}  \right)
    = \frac{1}{2}(\eta + \gamma) \leq \gamma.
    \label{eq:fedacii:conv:main:5}
  \end{equation}

  Plugging \cref{eq:fedacii:conv:main:2,eq:fedacii:conv:main:3,eq:fedacii:conv:main:4,eq:fedacii:conv:main:5} to \cref{eq:fedacii:conv:main:1} gives
  \begin{align*}
    & \expt \left[\Phi^{(r,k+1)} \middle| \mathcal{F}^{(r,k)} \right]
    \\
    \leq &  \left( 1 - \frac{1}{3}\gamma\mu \right) \Phi^{(r,k)} +  \frac{\eta^2 L \sigma^2 }{2M} + \frac{\gamma^2 \mu \sigma^2}{6 M}
    + \gamma  \left\|  \nabla F(\overline{\x^{(r,k)}_{\mathrm{md}}}) -  \frac{1}{M} \sum_{m=1}^M \nabla F (\x^{(r,k)}_{\mathrm{md}, m})  \right\|_2^2 .
  \end{align*}
  Telescoping the above inequality yields
  \begin{align*}
    \expt \left[\Phi^{(R,0)} \right]
    \leq & \exp \left( - \frac{1}{3} \gamma \mu K R \right) \Phi^{(0,0)} +  \left( \frac{3\eta^2 L \sigma^2 }{2 \gamma \mu M} + \frac{\gamma \sigma^2}{2 M} \right)
    \\
    & \qquad + \frac{3}{\mu}  \cdot \max_{\substack{0 \leq r < R \\ 0 \leq k < K}}   \expt \left[ \left\|  \nabla F(\overline{\x^{(r,k)}_{\mathrm{md}}}) -  \frac{1}{M} \sum_{m=1}^M \nabla F (\x^{(r,k)}_{\mathrm{md}, m})  \right\|_2^2\right].
  \end{align*}
\end{proof}

\subsubsection{Proof of \cref{fedacii:conv:1}}
\label{sec:proof:fedacii:conv:1}
\begin{proof}[Proof of \cref{fedacii:conv:1}]
  By definition of the \fedac procedure (\cref{alg:fedac}),
  \begin{equation*}
    \overline{\x^{(r,k+1)}} - \x^{\star} = (1 - \alpha^{-1}) \overline{\x^{(r,k)}} + \alpha^{-1} \overline{\x^{(r,k)}_{\mathrm{md}}} - \gamma \cdot \frac{1}{M} \sum_{m=1}^M \nabla f(\x^{(r,k)}_{\mathrm{md},m}; \xi^{(r,k)}_m) - \x^{\star}.
  \end{equation*}
  Taking conditional expectation gives
  \begin{align}
    & \expt \left[ \|\overline{\x^{(r,k+1)}} - \x^{\star}\|_2^2  \middle| \mathcal{F}^{(r,k)} \right]
    \nonumber \\
    \leq &
    \left\| (1 - \alpha^{-1}) \overline{\x^{(r,k)}} + \alpha^{-1} \overline{\x^{(r,k)}_{\mathrm{md}}} - \gamma \cdot \frac{1}{M} \sum_{m=1}^M \nabla F (\x^{(r,k)}_{\mathrm{md},m})- \x^{\star} \right\|_2^2 
    + \frac{1}{M}\gamma^2 \sigma^2.
    \label{eq:fedacii:conv:1:0}
  \end{align}
  The squared norm in \cref{eq:fedacii:conv:1:0} is bounded as
  \begin{align}
         & \left\| (1 - \alpha^{-1}) \overline{\x^{(r,k)}} + \alpha^{-1} \overline{\x^{(r,k)}_{\mathrm{md}}} - \gamma \cdot \frac{1}{M} \sum_{m=1}^M \nabla F (\x^{(r,k)}_{\mathrm{md},m})- \x^{\star} \right\|_2^2 
    \nonumber \\
    =    & \left\| (1 - \alpha^{-1}) \overline{\x^{(r,k)}} + \alpha^{-1} \overline{\x^{(r,k)}_{\mathrm{md}}} - \gamma \nabla F (\overline{\x^{(r,k)}_{\mathrm{md}}}) - \x^{\star}
    + \gamma \left(  \nabla F (\overline{\x^{(r,k)}_{\mathrm{md}}}) - \frac{1}{M} \sum_{m=1}^M \nabla F (\x^{(r,k)}_{\mathrm{md}}) \right)\right\|_2^2
    \nonumber \\
    \leq & {\left( 1 + \frac{1}{2} \alpha^{-1} \right)} \left\|  (1 - \alpha^{-1}) \overline{\x^{(r,k)}} + \alpha^{-1} \overline{\x^{(r,k)}_{\mathrm{md}}} - \x^{\star} - \gamma \nabla F (\overline{\x^{(r,k)}_{\mathrm{md}}}) \right\|_2^2 
    \nonumber \\
         & + \gamma^2  {\left( 1 + 2 \alpha \right)} \left\| \nabla F (\overline{\x^{(r,k)}_{\mathrm{md}}}) - \frac{1}{M} \sum_{m=1}^M \nabla F (\x^{(r,k)}_{\mathrm{md}, m}) \right\|_2^2 
    \tag{apply helper \cref{helper:unbalanced:ineq} with $\zeta = \frac{1}{2}\alpha^{-1}$}
     \\
    =    & \underbrace{\left( 1 + \frac{1}{2} \alpha^{-1} \right) \left\| (1 - \alpha^{-1}) \overline{\x^{(r,k)}} + \alpha^{-1} \overline{\x^{(r,k)}_{\mathrm{md}}} - \x^{\star} \right\|_2^2 }_{\text{(I)}}
    + \underbrace{\gamma^2  \left( 1 + \frac{1}{2} \alpha^{-1} \right) \left\| \nabla F (\overline{\x^{(r,k)}_{\mathrm{md}}}) \right\|_2^2 }_{\text{(II)}}
    \nonumber \\
         & \underbrace{- 2 \gamma  \left( 1 + \frac{1}{2} \alpha^{-1} \right) \left\langle \nabla F (\overline{\x^{(r,k)}_{\mathrm{md}}}), (1 - \alpha^{-1}) \overline{\x^{(r,k)}} + \alpha^{-1} \overline{\x^{(r,k)}_{\mathrm{md}}} - \x^{\star}\right\rangle}_{\text{(III)}}
    \nonumber \\
         & + \gamma^2 \left( 1 + 2 \alpha \right)  \left\| \nabla F (\overline{\x^{(r,k)}_{\mathrm{md}}}) - \frac{1}{M} \sum_{m=1}^M \nabla F (\x^{(r,k)}_{\mathrm{md}, m}) \right\|_2^2 .
    \label{eq:fedacii:conv:1:1}
  \end{align}

  The first term (I) of \cref{eq:fedacii:conv:1:1} is bounded via Jensen's inequality as follows:
  \begin{align}
         & \left( 1 + \frac{1}{2} \alpha^{-1} \right) \left\| (1 - \alpha^{-1}) \overline{\x^{(r,k)}} + \alpha^{-1} \overline{\x^{(r,k)}_{\mathrm{md}}} - \x^{\star} \right\|_2^2 
    \nonumber \\
    \leq & \left( 1 + \frac{1}{2} \alpha^{-1} \right) \left( (1 - \alpha^{-1}) \|\overline{\x^{(r,k)}} - \x^{\star}\|_2^2  + \alpha^{-1} \| \overline{\x^{(r,k)}_{\mathrm{md}}} - \x^{\star}\|_2^2  \right) \tag{Jensen's inequality}\
    \nonumber \\
    \leq & \left( 1 - \frac{1}{2} \alpha^{-1} \right)\|\overline{\x^{(r,k)}} - \x^{\star}\|_2^2  + \frac{3}{2} \alpha^{-1} \| \overline{\x^{(r,k)}_{\mathrm{md}}} - \x^{\star}\|_2^2 .
    \label{eq:fedacii:conv:1:2}
  \end{align}
  where in the last inequality of \cref{eq:fedacii:conv:1:2} we used the fact that $( 1 + \frac{1}{2} \alpha^{-1})(1 - \alpha^{-1}) = 1 - \frac{1}{2} \alpha^{-1} - \frac{1}{2} \alpha^{-2} < 1 - \frac{1}{2} \alpha^{-1}$, and $( 1 + \frac{1}{2} \alpha^{-1}) \alpha^{-1} \leq \frac{3}{2}\alpha^{-1}$ as $\alpha \geq 1$.

  The second term (II) of \cref{eq:fedacii:conv:1:1} is bounded as (since $\alpha \geq 1$)
  \begin{equation}
    \gamma^2  \left( 1 + \frac{1}{2} \alpha^{-1} \right) \left\| \nabla F (\overline{\x^{(r,k)}_{\mathrm{md}}}) \right\|_2^2\leq \frac{3}{2} \gamma^2  \left\| \nabla F (\overline{\x^{(r,k)}_{\mathrm{md}}}) \right\|_2^2 .
    \label{eq:fedacii:conv:1:3}
  \end{equation}

  To analyze the third term (III) of \cref{eq:fedacii:conv:1:1}, we note that by definition of $\overline{\x^{(r,k)}_{\mathrm{md}}}$,
  \begin{align}
      & - 2 \gamma  \left( 1 + \frac{1}{2} \alpha^{-1} \right) \left\langle \nabla F (\overline{\x^{(r,k)}_{\mathrm{md}}}), (1 - \alpha^{-1}) \overline{\x^{(r,k)}} + \alpha^{-1} \overline{\x^{(r,k)}_{\mathrm{md}}} - \x^{\star}\right\rangle
    \nonumber \\
    = & - 2 \gamma  \left( 1 + \frac{1}{2} \alpha^{-1} \right) \left\langle \nabla F (\overline{\x^{(r,k)}_{\mathrm{md}}}) , (1-\alpha^{-1}(1-\beta^{-1})) \overline{\x^{(r,k)}} + \alpha^{-1} (1 - \beta^ {-1}) \overline{\x^{(r,k)}_{\mathrm{ag}}} - \x^{\star}\right\rangle.
    \label{eq:fedacii:conv:1:4}
  \end{align}
  Plugging \cref{eq:fedacii:conv:1:1,eq:fedacii:conv:1:2,eq:fedacii:conv:1:3,eq:fedacii:conv:1:4} back to \cref{eq:fedacii:conv:1:0} yields
  \begin{align*}
         & \expt [ \|\overline{\x^{(r,k+1)}} - \x^{\star}\|_2^2  |\mathcal{F}^{(r,k)}]
    \\
    \leq & \left( 1 - \frac{1}{2} \alpha^{-1} \right)\|\overline{\x^{(r,k)}} - \x^{\star}\|_2^2  + \frac{3}{2} \alpha^{-1} \| \overline{\x^{(r,k)}_{\mathrm{md}}} - \x^{\star}\|_2^2  + \frac{3}{2}\gamma^2  \left\| \nabla F (\overline{\x^{(r,k)}_{\mathrm{md}}}) \right\|_2^2 
    \\
         & - 2 \gamma  \left( 1 + \frac{1}{2} \alpha^{-1} \right) \left\langle \nabla F (\overline{\x^{(r,k)}_{\mathrm{md}}}) , (1-\alpha^{-1}(1-\beta^{-1})) \overline{\x^{(r,k)}} + \alpha^{-1} (1 - \beta^ {-1}) \overline{\x^{(r,k)}_{\mathrm{ag}}} - \x^{\star}\right\rangle
    \\
         &
    + \gamma^2 \left( 1 + 2 \alpha \right)  \left\| \nabla F (\overline{\x^{(r,k)}_{\mathrm{md}}}) - \frac{1}{M} \sum_{m=1}^M \nabla F (\x^{(r,k)}_{\mathrm{md}, m}) \right\|_2^2 
    +  \frac{\gamma^2 \sigma^2}{M},
  \end{align*}
  completing the proof of \cref{fedacii:conv:1}.
\end{proof}

\subsubsection{Proof of \cref{fedacii:conv:2}}
\label{sec:proof:fedacii:conv:2}
\begin{proof}[Proof of \cref{fedacii:conv:2}]
  By definition of the \fedac procedure we have
  \begin{equation*}
    \overline{\x^{(r,k+1)}_{\mathrm{ag}}}
    =
    \overline{\x^{(r,k)}_{\mathrm{md}}} - \eta \cdot \frac{1}{M} \sum_{m=1}^M \nabla f(\x^{(r,k)}_{\mathrm{md},m}; \xi^{(r,k)}_m),
  \end{equation*}
  and thus, by $L$-smoothness (\cref{asm:fo:scvx:2o}(b)) we obtain
  \begin{align*}
    & F( \overline{\x^{(r,k+1)}_{\mathrm{ag}}})
    \\
    \leq &
    F(\overline{\x^{(r,k)}_{\mathrm{md}}}) - \eta \left\langle \nabla F(\overline{\x^{(r,k)}_{\mathrm{md}}}), \frac{1}{M} \sum_{m=1}^M \nabla f(\x^{(r,k)}_{\mathrm{md},m}; \xi^{(r,k)}_m) \right\rangle + \frac{\eta^2 L}{2} \left\| \frac{1}{M} \sum_{m=1}^M \nabla f(\x^{(r,k)}_{\mathrm{md},m}; \xi^{(r,k)}_m)  \right\|_2^2 .
  \end{align*}
  Taking conditional expectation, and by bounded variance (\cref{asm:fo:scvx:2o}(c))
  \begin{align}
    & \expt \left[F( \overline{\x^{(r,k+1)}_{\mathrm{ag}}}) |\mathcal{F}^{(r,k)} \right]
    \nonumber \\
    \leq & F(\overline{\x^{(r,k)}_{\mathrm{md}}})
    - \eta \left\langle \nabla F(\overline{\x^{(r,k)}_{\mathrm{md}}}), \frac{1}{M} \sum_{m=1}^M \nabla F (\x^{(r,k)}_{\mathrm{md}, m}) \right\rangle + \frac{\eta^2 L}{2} \left\| \frac{1}{M} \sum_{m=1}^M \nabla F (\x^{(r,k)}_{\mathrm{md}, m})  \right\|_2^2+ \frac{\eta^2 L \sigma^2 }{2M}.
    \label{eq:fedacii:conv:2:0:1}
  \end{align}
  By polarization identity we have
  \begin{align}
    & \left\langle \nabla F(\overline{\x^{(r,k)}_{\mathrm{md}}}), \frac{1}{M} \sum_{m=1}^M \nabla F (\x^{(r,k)}_{\mathrm{md}, m}) \right\rangle
    \nonumber \\
    = &
    \frac{1}{2}
    \left( \left\| \nabla F(\overline{\x^{(r,k)}_{\mathrm{md}}}) \right\|_2^2+ \left\| \frac{1}{M} \sum_{m=1}^M \nabla F (\x^{(r,k)}_{\mathrm{md}, m}) \right\|_2^2- \left\|  \nabla F(\overline{\x^{(r,k)}_{\mathrm{md}}}) -  \frac{1}{M} \sum_{m=1}^M \nabla F (\x^{(r,k)}_{\mathrm{md}, m})  \right\|_2^2\right).
    \label{eq:fedacii:conv:2:0:2}
  \end{align}
  Combining \cref{eq:fedacii:conv:2:0:1,eq:fedacii:conv:2:0:2} gives
  \begin{align}
         & \expt \left[F( \overline{\x^{(r,k+1)}_{\mathrm{ag}}}) |\mathcal{F}^{(r,k)} \right]
    \nonumber \\
    =    & F(\overline{\x^{(r,k)}_{\mathrm{md}}}) - \frac{1}{2} \eta \left\| \nabla F(\overline{\x^{(r,k)}_{\mathrm{md}}}) \right\|_2^2 
    + \frac{1}{2} \eta \left\|  \nabla F(\overline{\x^{(r,k)}_{\mathrm{md}}}) -  \frac{1}{M} \sum_{m=1}^M \nabla F (\x^{(r,k)}_{\mathrm{md}, m})  \right\|_2^2 
    \nonumber \\
         & - \frac{1}{2} \eta (1 - \eta L) \left\| \frac{1}{M} \sum_{m=1}^M \nabla F (\x^{(r,k)}_{\mathrm{md}, m})  \right\|_2^2+  \frac{\eta^2 L \sigma^2 }{2M}
    \nonumber \\
    \leq & F(\overline{\x^{(r,k)}_{\mathrm{md}}}) - \frac{1}{2} \eta \left\| \nabla F(\overline{\x^{(r,k)}_{\mathrm{md}}}) \right\|_2^2\
    + \frac{1}{2}  \eta \left\|  \nabla F(\overline{\x^{(r,k)}_{\mathrm{md}}}) -  \frac{1}{M} \sum_{m=1}^M \nabla F (\x^{(r,k)}_{\mathrm{md}, m})  \right\|_2^2+  \frac{\eta^2 L \sigma^2 }{2M},
    \label{eq:fedacii:conv:2:1}
  \end{align}
  where the last inequality is due to the assumption that $\eta \leq \frac{1}{L}$.

  Now we relate $F( \overline{\x^{(r,k)}_{\mathrm{md}}})$ and $F( \overline{\x^{(r,k)}_{\mathrm{ag}}})$ as follows
  \begin{align}
         & F(\overline{\x^{(r,k)}_{\mathrm{md}}})  - F^{\star}
    \nonumber \\
    =    & \left( 1 - \frac{1}{2} \alpha^{-1} \right) \left(F(\overline{\x^{(r,k)}_{\mathrm{ag}}}) - F^{\star} \right)
    \nonumber \\
    & \quad + \left( 1 - \frac{1}{2} \alpha^{-1} \right) \left(F(\overline{\x^{(r,k)}_{\mathrm{md}}}) - F(\overline{\x^{(r,k)}_{\mathrm{ag}}}) \right)
    + \frac{1}{2} \alpha^{-1} \left(F(\overline{\x^{(r,k)}_{\mathrm{md}}}) - F^{\star} \right)
    \nonumber \\
    \leq & \left( 1 - \frac{1}{2} \alpha^{-1} \right) \left(F(\overline{\x^{(r,k)}_{\mathrm{ag}}}) - F^{\star} \right)
    + \left( 1 - \frac{1}{2} \alpha^{-1} \right) \left\langle \nabla F(\overline{\x^{(r,k)}_{\mathrm{md}}}), \overline{\x^{(r,k)}_{\mathrm{md}}} - \overline{\x^{(r,k)}_{\mathrm{ag}}} \right\rangle
    \nonumber \\
         & + \frac{1}{2} \alpha^{-1}
    \left( \left\langle \nabla F(\overline{\x^{(r,k)}_{\mathrm{md}}}), \overline{\x^{(r,k)}_{\mathrm{md}}} - \x^{\star} \right\rangle  - \frac{\mu}{2}  \left\|  \overline{\x^{(r,k)}_{\mathrm{md}}} - \x^{\star}  \right\|_2^2\right)
    \tag{$\mu$-strong convexity}
    \\
    =    & \left( 1 - \frac{1}{2} \alpha^{-1} \right) \left(F(\overline{\x^{(r,k)}_{\mathrm{ag}}}) - F^{\star} \right) - \frac{1}{4} \mu \alpha^{-1} \left\|  \overline{\x^{(r,k)}_{\mathrm{md}}} - \x^{\star}  \right\|_2^2 
    \nonumber \\
         &
    + {\frac{1}{2} \alpha^{-1} \left\langle \nabla F(\overline{\x^{(r,k)}_{\mathrm{md}}}), 2 \alpha \overline{\x^{(r,k)}_{\mathrm{md}}} - (2 \alpha - 1) \overline{\x^{(r,k)}_{\mathrm{ag}}} -  \x^{\star} \right\rangle}
    \tag{rearranging}
    \\
    =    & \left( 1 - \frac{1}{2} \alpha^{-1} \right) \left(F(\overline{\x^{(r,k)}_{\mathrm{ag}}}) - F^{\star} \right) - \frac{1}{4} \mu \alpha^{-1} \left\|  \overline{\x^{(r,k)}_{\mathrm{md}}} - \x^{\star}  \right\|_2^2 
    \nonumber \\
         &
    + {\frac{1}{2} \alpha^{-1} \left\langle \nabla F(\overline{\x^{(r,k)}_{\mathrm{md}}}), 2 \alpha \beta^{-1} \overline{\x^{(r,k)}} + (1 - 2 \alpha \beta^{-1}) \overline{\x^{(r,k)}_{\mathrm{ag}}} -  \x^{\star} \right\rangle},
    \label{eq:fedacii:conv:2:2}
  \end{align}
  where the last equality is due to the definition of $\overline{\x^{(r,k)}_{\mathrm{md}}}$.

  Plugging \cref{eq:fedacii:conv:2:2} back to \cref{eq:fedacii:conv:2:1} yields
  \begin{align*}
         & \expt \left[F( \overline{\x^{(r,k+1)}_{\mathrm{ag}}})  - F^{\star} \middle|\mathcal{F}^{(r,k)} \right]
    \\
    \leq & \left( 1 - \frac{1}{2} \alpha^{-1} \right) \left(F(\overline{\x^{(r,k)}_{\mathrm{ag}}}) - F^{\star} \right) - \frac{1}{4} \mu \alpha^{-1} \left\|  \overline{\x^{(r,k)}_{\mathrm{md}}} - \x^{\star}  \right\|_2^2- \frac{1}{2} \eta \left\| \nabla F(\overline{\x^{(r,k)}_{\mathrm{md}}}) \right\|_2^2 
    \\
         & + {\frac{1}{2} \alpha^{-1} \left\langle \nabla F(\overline{\x^{(r,k)}_{\mathrm{md}}}), 2 \alpha \beta^{-1} \overline{\x^{(r,k)}} + (1 - 2 \alpha \beta^{-1}) \overline{\x^{(r,k)}_{\mathrm{ag}}} -  \x^{\star} \right\rangle}
    \\
         & + \frac{1}{2}  \eta \left\|  \nabla F(\overline{\x^{(r,k)}_{\mathrm{md}}}) -  \frac{1}{M} \sum_{m=1}^M \nabla F (\x^{(r,k)}_{\mathrm{md}, m})  \right\|_2^2+  \frac{\eta^2 L \sigma^2 }{2M},
  \end{align*}
  completing the proof of \cref{fedacii:conv:2}.
\end{proof}

\subsection{Discrepancy overhead bound for \fedacii: Proof of \cref{fedacii:stab:main}}
\label{sec:fedacii:stab:main}
In this subsection we prove \cref{fedacii:stab:main} regarding the regarding the growth of discrepancy overhead introduced in \cref{fedacii:conv:main}. The core of the proof is the \nth{4}-order stability of \fedacii.
Note that most of the analysis in this subsection follows closely with the analysis on \fedaci (see \cref{sec:fedaci:stab:main}), but the analysis is technically more complicated.

We will reuse a set of notations defined in \cref{sec:fedaci:stab:main}, which we restate here for clearance. Let $m_1, m_2 \in [M]$ be two arbitrary distinct machines. For any timestep $(r, k)$, denote $\bDelta^{(r,k)} := \x^{(r,k)}_{m_1} - \x^{(r,k)}_{m_2}$,  $\bDelta^{(r,k)}_{\mathrm{ag}} := \x^{(r,k)}_{\mathrm{ag}, m_1} - \x^{(r,k)}_{\mathrm{ag}, m2}$ and $\bDelta^{(r,k)}_{\mathrm{md}} := \x^{(r,k)}_{\mathrm{md}, m_1} - \x^{(r,k)}_{\mathrm{md}, m_2}$ be the corresponding vector differences. Let $\bDelta^{(r,k)}_{\varepsilon} = \varepsilon^{(r,k)}_{m_1} - \varepsilon^{(r,k)}_{m_2}$, where $\varepsilon^{(r,k)}_m := \nabla f(\x^{(r,k)}_{\mathrm{md},m}; \xi^{(r,k)}_m) - \nabla F(\x^{(r,k)}_{\mathrm{md}, m})$.

The proof of \cref{fedacii:stab:main} is based on the following propositions.

The following \cref{fedacii:stab:1} studies the growth of $\begin{bmatrix} \bDelta^{(r,k)}_{\mathrm{ag}} \\ \bDelta^{(r,k)} \end{bmatrix}$ at each step. 
\cref{fedacii:stab:1} is analogous to \cref{fedaci:stab:1}, but the $\A$ is different. Note that \cref{fedacii:stab:1} requires only \cref{asm:fo:scvx:2o}.
\begin{proposition}
  \label{fedacii:stab:1}
  Let $F$ be $\mu>0$-strongly convex, assume \cref{asm:fo:scvx:2o} and assume the same hyperparameter choice is taken as in  \cref{fedacii:stab:main} (namely $\alpha = \frac{3}{2 \gamma \mu} - \frac{1}{2}$, $\beta =\frac{2 \alpha^2 - 1}{\alpha - 1}$, $\gamma \in [\eta, \sqrt{ \frac{\eta}{\mu}}]$, $\eta \in (0, \frac{1}{L}]$). Then there exists a matrix $\H^{(r,k)}$ such that $\mu \I \preceq \H^{(r,k)} \preceq L\I$ satisfying
  \begin{equation*}
    \begin{bmatrix}
      \bDelta^{(r,k+1)}_{\mathrm{ag}} \\ \bDelta^{(r,k+1)}
    \end{bmatrix}
    =
    \A (\mu, \gamma, \eta, \H^{(r,k)})
    \begin{bmatrix} \bDelta^{(r,k)}_{\mathrm{ag}} \\ \bDelta^{(r,k)} \end{bmatrix}
    -
    \begin{bmatrix} \eta \I \\ \gamma \I \end{bmatrix}
    \bDelta^{(r,k)}_{\varepsilon},
  \end{equation*}
  where $\A(\mu, \gamma, \eta, \H)$ is a matrix-valued function defined as
  \begin{align}
    & \A(\mu, \gamma, \eta, \H) 
    \nonumber \\
    = & \frac{1}{9 - \gamma \mu (6 + \gamma \mu)}
    \begin{bmatrix}
      (3 - \gamma \mu)(3 - 2\gamma \mu)(\I - \eta \H)
       & 3 \gamma \mu (1 - \gamma \mu) (\I - \eta \H)
      \\
      (3 - 2 \gamma \mu) (2 \gamma \mu \I - (3 - \gamma \mu) \gamma \H)
       &
      3 (1 - \gamma \mu) ((3 - \gamma \mu)\I - \gamma^2 \mu \H)
    \end{bmatrix}.
    \label{eq:fedacii:A:def}
  \end{align}
\end{proposition}
The proof of \cref{fedacii:stab:1} is almost identical with \cref{fedaci:stab:1} except the choice of $\alpha$ and $\beta$ are different. We include this proof in \cref{sec:fedacii:stab:1} for completeness.

The following \cref{fedacii:stab:bound} studies the uniform norm bound of $\A$ under the proposed transformation $\X$. The transformation $\X$ is the same as the one studied in \fedaci, which we restate here for the ease of reference.
The bound is also similar to the corresponding bound for on \fedaci as shown in \cref{fedaci:stab:bound}, though the proof is technically more complicated due to the complexity of $\A$.
We defer the proof of  \cref{fedacii:stab:bound} to \cref{sec:fedacii:stab:bound}.
\begin{proposition}[Uniform norm bound of $\A$ under transformation $\X$]
  \label{fedacii:stab:bound}
  Let $\A(\mu, \gamma, \eta, \H)$ be defined as in \cref{eq:fedacii:A:def}.
  and assume $\mu > 0$, $\gamma \in [\eta, \sqrt{\frac{\eta}{\mu}}]$, $\eta \in (0,\frac{1}{L}]$.
  Then the following uniform norm bound holds
  \begin{equation*}
    \sup_{\mu \I \preceq \H \preceq L \I}
    \left\| \X(\gamma, \eta)^{-1} \A(\mu, \gamma, \eta, \H) \X(\gamma, \eta) \right\|_2 \leq
    \begin{cases}
      1 + \frac{\gamma^2 \mu}{\eta} & \text{if~} \gamma \in \left(\eta, \sqrt{\frac{\eta}{\mu}}\right], \\
      1                              & \text{if~} \gamma =  \eta,
    \end{cases}
  \end{equation*}
  where $\X (\gamma, \eta)$ is a matrix-valued function defined as
  \begin{equation}
    \X(\gamma, \eta) :=
    \begin{bmatrix}
      \frac{\eta}{\gamma} \I & \zeros
      \\
      \I                     & \I
    \end{bmatrix}.
    \label{eq:fedacii:X:def}
  \end{equation}
\end{proposition}

\cref{fedacii:stab:1,fedacii:stab:bound} suggest the one-step growth of $ \left\| \X(\gamma, \eta)^{-1} \begin{bmatrix}
  \bDelta^{(r,k)}_{\mathrm{ag}}
  \\
  \bDelta^{(r,k)}
\end{bmatrix}  \right\|_2^4$ as follows.
\begin{proposition}
  \label{fedacii:stab:2}
  In the same setting of \cref{fedacii:stab:main}, the following inequality holds (for all possible $(r, k)$)
  \begin{align*}
    & \sqrt{
      \expt \left[ \left\| \X(\gamma, \eta)^{-1} \begin{bmatrix}
        \bDelta^{(r,k+1)}_{\mathrm{ag}}
        \\
        \bDelta^{(r,k+1)}
      \end{bmatrix}  \right\|_2^4 \middle| \mathcal{F}^{(r,k)} \right]
    }
    \\
    \leq &
    7 \gamma^2 \sigma^2 +
    \left\| \X(\gamma, \eta)^{-1} \begin{bmatrix}
      \bDelta^{(r,k)}_{\mathrm{ag}}
      \\
      \bDelta^{(r,k)}
    \end{bmatrix}  \right\|_2^2 
    \cdot
    \begin{cases}
      \left(1 + \frac{\gamma^2 \mu}{\eta} \right)^2 & \text{if~} \gamma \in \left(\eta, \sqrt{\frac{\eta}{\mu}}\right], \\
      1                                              & \text{if~} \gamma =  \eta,
    \end{cases}
  \end{align*}
  where $\X$ is the matrix-valued function defined in \cref{eq:fedacii:X:def}.
\end{proposition}
We defer the proof of \cref{fedacii:stab:2} to \cref{sec:fedacii:stab:2}.

The following \cref{fedacii:stab:3} links the discrepancy overhead we wish to bound for \cref{fedacii:stab:main} with the quantity analyzed in \cref{fedacii:stab:2} via \nth{3}-order-smoothness (\cref{asm:fo:scvx:3o}(a)). 
The proof of \cref{fedacii:stab:3} is deferred to \cref{sec:fedacii:stab:3}.
\begin{proposition} 
  \label{fedacii:stab:3}
  In the same setting of \cref{fedacii:stab:main}, the following inequality holds (for all possible $(r, k)$)
  \begin{equation*}
    \left\|  \nabla F(\overline{\x^{(r,k)}_{\mathrm{md}}}) -  \frac{1}{M} \sum_{m=1}^M \nabla F (\x^{(r,k)}_{\mathrm{md}, m})  \right\|_2^2
    \leq
    \frac{289 \eta^4 Q^2}{324 \gamma^4} \left\| \X(\gamma, \eta)^{-1} \begin{bmatrix}
      \bDelta^{(r,k)}_{\mathrm{ag}}
      \\
      \bDelta^{(r,k)}
    \end{bmatrix}  \right\|_2^4,
  \end{equation*}
  where $\X$ is the matrix-valued function defined in \cref{eq:fedacii:X:def}.
\end{proposition}

We are ready to complete the proof of \cref{fedacii:stab:main}.
\begin{proof}[Proof of \cref{fedacii:stab:main}]
  Applying \cref{fedacii:stab:2} gives
  \begin{align*}
    & \sqrt{
      \expt \left[ \left\| \X(\gamma, \eta)^{-1} \begin{bmatrix}
        \bDelta^{(r,k+1)}_{\mathrm{ag}}
        \\
        \bDelta^{(r,k+1)}
      \end{bmatrix}  \right\|_2^4 \middle| \mathcal{F}^{(r,0)} \right]
    }
    \\
    \leq & 7 \gamma^2 \sigma^2 +
    \sqrt{
    \expt \left[
    \left\| \X(\gamma, \eta)^{-1} \begin{bmatrix}
      \bDelta^{(r,k)}_{\mathrm{ag}}
      \\
      \bDelta^{(r,k)}
    \end{bmatrix}  \right\|_2^2\middle| \mathcal{F}^{(r,0)} \right]}
    \cdot
    \begin{cases}
      \left(1 + \frac{\gamma^2 \mu}{\eta} \right)^2 & \text{if~} \gamma \in \left(\eta, \sqrt{\frac{\eta}{\mu}}\right], \\
      1                                              & \text{if~} \gamma =  \eta.
    \end{cases}
  \end{align*}
  Telescoping from $(r,0)$-th step to $(r,k)$-th step gives (note that $\bDelta^{(r,0)}_{\mathrm{ag}} = \bDelta^{(r,0)} = \zeros$)
  \begin{equation*}
    \expt \left[ \left\| \X(\gamma, \eta)^{-1} \begin{bmatrix}
      \bDelta^{(r,k)}_{\mathrm{ag}}
     \\
      \bDelta^{(r,k)}
    \end{bmatrix}  \right\|_2^4 \middle| \mathcal{F}^{(r,0)} \right]
  \leq
  49 \gamma^4 \sigma^4 k^2
  \cdot
  \begin{cases}
    \left( 1 + \frac{\gamma^2 \mu}{\eta} \right)^{4k} & \text{if~} \gamma \in \left(\eta, \sqrt{\frac{\eta}{\mu}} \right], \\
    1                              & \text{if~} \gamma =  \eta.
  \end{cases}
  \end{equation*}
  Consequently, by \cref{fedacii:stab:3} we have
  \begin{align*}
    & 
    \expt \left[  \left\|  \nabla F(\overline{\x^{(r,k)}_{\mathrm{md}}}) -  \frac{1}{M} \sum_{m=1}^M \nabla F (\x^{(r,k)}_{\mathrm{md}, m})  \right\|_2^2\middle| \mathcal{F}^{(r,0)}  \right]
    \\
    \leq & \frac{289 \eta^4 Q^2}{324 \gamma^4} \expt \left[ \left\| \X(\gamma, \eta)^{-1} \begin{bmatrix}
      \bDelta^{(r,k)}_{\mathrm{ag}}
     \\
      \bDelta^{(r,k)}
    \end{bmatrix}  \right\|_2^4 \middle| \mathcal{F}^{(r,0)} \right]
    \\ 
    \leq & 
    \begin{cases}
      44 \eta^4 Q^2 K^2 \sigma^4 \left(1 + \frac{\gamma^2\mu}{\eta}\right)^{4K}
       & \text{if~} \gamma \in \left(\eta, \sqrt{\frac{\eta}{\mu}} \right],
      \\
      44 \eta^4 Q^2 K^2 \sigma^4
       &
      \text{if~} \gamma = \eta,
    \end{cases}
  \end{align*}
  where in the last inequality we used the estimate that $\frac{289}{324} \cdot 49 < 44$. 
\end{proof}

\subsubsection{Proof of \cref{fedacii:stab:1}}

\label{sec:fedacii:stab:1}
\begin{proof}[Proof of \cref{fedacii:stab:1}]
  The proof of  \cref{fedacii:stab:1} follows instantly by plugging $\alpha = \frac{3}{2\gamma \mu} - \frac{1}{2}$, $\beta = \frac{2 \alpha^2 - 1}{\alpha - 1} = \frac{9 - \gamma \mu (6 + \gamma \mu)}{3 \gamma \mu (1 - \gamma\mu)}$ to the general claim on \fedac \cref{fedac:general:stab}:
  \begin{align*}
    & \begin{bmatrix}
      (1 - \beta^{-1}) (\I - \eta \H)
       &
      \beta^{-1} (\I - \eta \H)
      \\
      (1 - \beta^{-1}) (\alpha^{-1} - \gamma \H)
       &
      \beta^{-1} (\alpha^{-1} \I - \gamma \H) + (1 - \alpha^{-1}) \I
    \end{bmatrix}
    \\
    = & \frac{1}{9 - \gamma \mu (6 + \gamma \mu)}
    \begin{bmatrix}
      (3 - \gamma \mu)(3 - 2\gamma \mu)(\I - \eta \H)
       & 3 \gamma \mu (1 - \gamma \mu) (\I - \eta \H)
      \\
      (3 - 2 \gamma \mu) (2 \gamma \mu - (3 - \gamma \mu) \gamma \H)
       &
      3 (1 - \gamma \mu) ((3 - \gamma \mu)\I - \gamma^2 \mu \H)
    \end{bmatrix}.
  \end{align*}
\end{proof}

\subsubsection{Proof of \cref{fedacii:stab:bound}: uniform norm bound}
\label{sec:fedacii:stab:bound}
The proof idea of this proposition is very similar to \cref{fedaci:stab:bound}, though more complicated technically.
\begin{proof}
  Define another matrix-valued function $\tilde{\A}$ as
  \begin{equation*}
    \tilde{\A}(\mu, \gamma, \eta, \H) := \X(\gamma, \eta)^{-1} \A(\mu, \gamma, \eta, \H) \X(\gamma, \eta).
  \end{equation*}
  Since $ \X(\gamma, \eta)^{-1} =
  \begin{bmatrix}
    \frac{\gamma}{\eta} \I  & \zeros
    \\
    -\frac{\gamma}{\eta} \I & \I
  \end{bmatrix}$ 
  we have 
  \begin{align*}
      & \tilde{\A}(\mu, \gamma, \eta, \H) = \frac{1}{(9 - (6 + \gamma \mu)\gamma \mu)\eta} \cdot
    \\
    & \quad 
    \begin{bmatrix}
      \left( 3 \gamma^2 \mu ( 1 - \gamma \mu) + \eta (3 - \gamma \mu)(3 - 2 \gamma \mu) \right) (\I - \eta \H) &
      3\gamma^2 \mu(1 - \gamma \mu)(\I - \eta \H)
      \\
      - (\gamma - \eta)\left( 3 \gamma + 6 \eta - \gamma \mu (3 \gamma + 4\eta) \right) \I                    &
      3 (1 - \gamma \mu)\left( 3 \eta - \gamma \mu (\gamma + \eta)  \right) \I
    \end{bmatrix}.
  \end{align*}
  Define the four blocks of $\tilde{\A}(\mu, \gamma, \eta, \H)$ as $\tilde{\A}_{11}(\mu, \gamma, \eta, \H)$, $\tilde{\A}_{12}(\mu, \gamma, \eta, \H)$, $\tilde{\A}_{21}(\mu, \gamma, \eta)$, $\tilde{\A}_{22}(\mu, \gamma, \eta)$ (note that the lower two blocks do not involve $H$), namely
  \begin{align*}
    \tilde{\A}_{11}(\mu, \gamma, \eta, \H) & = \frac{3 \gamma^2 \mu ( 1 - \gamma \mu) + \eta (3 - \gamma \mu)(3 - 2 \gamma \mu)}{(9 - (6 + \gamma \mu)\gamma \mu) \eta } (\I - \eta \H),
    \\
    \tilde{\A}_{12}(\mu, \gamma, \eta, \H) & = \frac{3\gamma^2 \mu(1 - \gamma \mu)}{(9 - (6 + \gamma \mu)\gamma \mu) \eta }(\I - \eta \H),
    \\
    \tilde{\A}_{21}(\mu, \gamma, \eta)    & = -\frac{(\gamma - \eta) \mu \left( 3 \gamma + 6 \eta - \gamma \mu (3 \gamma + 4\eta) \right)}{(9 - (6 + \gamma \mu)\gamma \mu) \eta } \I,
    \\
    \tilde{\A}_{22}(\mu, \gamma, \eta)    & = \frac{3 (1 - \gamma \mu)\left( 3 \eta - \gamma \mu (\gamma + \eta)  \right)}{(9 - (6 + \gamma \mu)\gamma \mu) \eta} \I.
  \end{align*}
  \paragraph{Case I: $\eta < \gamma \leq \sqrt{\frac{\eta}{\mu}}$.}
  Since $\gamma \mu \leq 1$, we know that the common denominator
  \begin{equation*}
    (9 - (6 + \gamma \mu)\gamma \mu) \eta \geq 2 \eta > 0.
  \end{equation*}
  Now we bound the operator norm of each block as follows.

  \paragraph{Bound for $\|\tilde{\A}_{11}\|_2$.} Since $3 \gamma^2 \mu ( 1 - \gamma \mu) + \eta (3 - \gamma \mu)(3 - 2 \gamma \mu) \geq 0$, we have $\tilde{\A}_{11} \succeq \zeros$, and therefore
  \begin{align}
         & \|\tilde{\A}_{11}(\mu, \gamma, \eta, \H)\|_2
    \nonumber \\
    \leq & \frac{3 \gamma^2 \mu ( 1 - \gamma \mu) + \eta (3 - \gamma \mu)(3 - 2 \gamma \mu)}{(9 - (6 + \gamma \mu)\gamma \mu) \eta } (1 - \eta \mu)
    \nonumber \\
    \leq & \frac{3 \gamma^2 \mu ( 1 - \gamma \mu) + \eta (3 - \gamma \mu)(3 - 2 \gamma \mu)}{(9 - (6 + \gamma \mu)\gamma \mu) \eta }
    \nonumber \\
    =    &  1 + \frac{3 (\gamma - \eta) \gamma \mu  (1 - \gamma \mu)}{(9 - (6 + \gamma \mu)\gamma \mu) \eta }
    \nonumber \\
    \leq & 1 + \frac{3 \gamma^2 \mu }{\eta} \cdot \frac{1 - \gamma \mu}{ 9 - 6 \gamma \mu - \gamma^2\mu^2}
    \tag{since $\gamma - \eta \leq \gamma$}
    \nonumber \\
    \leq & 1 + \frac{\gamma^2 \mu}{3 \eta},
    \label{eq:fedacii:stab:bound:11}
  \end{align}
  where the last inequality is due to $\frac{1 - \gamma \mu}{ 9 - 6 \gamma \mu - \gamma^2\mu^2} \leq \frac{1}{9}$ since $\gamma \mu \leq 1$.

  \paragraph{Bound for $\|\tilde{\A}_{12}\|_2$.} Similarly we have
  \begin{equation}
    \|\tilde{\A}_{12}(\mu, \gamma, \eta, \H)\|_2 \leq \frac{3 \gamma^2 \mu (1 - \gamma\mu)}{(9 - (6 + \gamma \mu)\gamma \mu) \eta}(1-\eta \mu)
    \leq \frac{3 \gamma^2 \mu}{\eta} \cdot \frac{1 - \gamma \mu}{9 - (6 + \gamma \mu)\gamma \mu}
    \leq \frac{\gamma^2 \mu}{3 \eta},
    \label{eq:fedacii:stab:bound:12}
  \end{equation}
  where the last inequality is due to $\frac{1 - \gamma \mu}{ 9 - 6 \gamma \mu - \gamma^2\mu^2} \leq \frac{1}{9}$ since $\gamma \mu \leq 1$.

  \paragraph{Bound for $\|\tilde{\A}_{21}\|_2$.} Since $\gamma \geq \eta$, we have $(\gamma - \eta) \mu \left( 3 \gamma + 6 \eta - \gamma \mu (3 \gamma + 4\eta) \right) \geq 0$.
  Note that
  \begin{align*}
     & (\gamma - \eta) \left( 3 \gamma + 6 \eta - \gamma \mu (3 \gamma + 4\eta) \right)
     \\
     = &  3 \gamma^2 + 3 \gamma \eta - 6 \eta^2 - \gamma \mu (3 \gamma^2 + \gamma \eta - 4 \eta^ 2)
     \\
     = & 4 \gamma^2 - 3\gamma^3 \mu - (\gamma^2 - 3 \gamma \eta + 6 \eta^2 + \gamma^2 \mu \eta - 4\eta^2 \gamma \mu  ),
  \end{align*}
  and
  \begin{align}
        & \gamma^2 - 3 \gamma \eta + 6 \eta^2 + \gamma^2 \mu \eta - 4\eta^2 \gamma \mu 
    \nonumber \\
    \geq & \gamma^2 - 3 \gamma \eta + 6 \eta^2 - 3\eta^2 \gamma \mu  \tag{since $\eta \leq \gamma$}
    \\
    \geq & \gamma^2 - 3 \gamma \eta + 3 \eta^2 \tag{since $\gamma \mu \leq 1$}
    \\
    \geq & 0. \tag{AM-GM inequality}
  \end{align}
  Consequently,
  \begin{equation}
    (\gamma - \eta) \mu \left( 3 \gamma + 6 \eta - \gamma \mu (3 \gamma + 4\eta) \right) \leq 4 \gamma^2 \mu - 3 \gamma^3 \mu^2.
    \label{eq:fedacii:stab:bound:tmp}
  \end{equation}
  It follows that
  \begin{align}
         & \|\tilde{\A}_{21}(\mu, \gamma, \eta)\|_2 =  \frac{\mu (\gamma - \eta)  \left( 3 \gamma + 6 \eta - \gamma \mu (3 \gamma + 4\eta) \right)}{(9 - (6 + \gamma \mu)\gamma \mu) \eta }
    \nonumber \\
    \leq    & \frac{4 \gamma^2 \mu - 3 \gamma^3 \mu^2}{(9 - (6 + \gamma \mu)\gamma \mu) \eta }  
    \tag{by \cref{eq:fedacii:stab:bound:tmp}}
    \\
    =   &  \frac{\gamma^2 \mu}{\eta} \cdot \frac{4 - 3 \gamma \mu}{9 - 6 \gamma \mu - \gamma^2 \mu^2} \leq  \frac{2\gamma^2 \mu}{3\eta}.
    \label{eq:fedacii:stab:bound:21}
  \end{align}
  where the last inequality is due to $\frac{4 - 3 \gamma \mu}{9 - 6 \gamma \mu - \gamma^2 \mu^2} \leq \frac{2}{3}$ since $\gamma \mu \leq 1$.

  \paragraph{Bound for $\tilde{\A}_{22}$.} Since $\gamma > \eta$ and $\gamma^2 \mu \leq \eta$, we have $3\eta - \gamma \mu(\gamma + \eta) \geq 3 \eta - 2 \gamma^2 \mu \geq \eta$. Thus $\tilde{\A}_{22} \succeq \zeros$, which implies
  \begin{equation}
    \|\tilde{\A}_{22}(\mu, \gamma, \eta)\|
    = \frac{3 (1 - \gamma \mu)\left( 3 \eta - \gamma \mu (\gamma + \eta)  \right)}{(9 - (6 + \gamma \mu)\gamma \mu) \eta}
    = 1 + \frac{\gamma \mu \left( -6 \eta - 3 \gamma + \gamma \mu (3 \gamma + 4 \eta ) \right) }{(9 - (6 + \gamma \mu)\gamma \mu) \eta }
    \leq 1.
    \label{eq:fedacii:stab:bound:22}
  \end{equation}

  The operator norm of block matrix $\tilde{\A}$ can be bounded via its blocks via \cref{helper:blocknorm} as
    \begin{align}
         & \tilde{\A}(\mu, \gamma, \eta, \H)
    \nonumber \\
    \leq & \max \left\{\| \tilde{\A}_{11}(\mu, \gamma, \eta, \H) \|_2, \| \tilde{\A}_{22}(\mu, \gamma, \eta) \|)_2 \right\}
    +
    \max \left\{\| \tilde{\A}_{12}(\mu, \gamma, \eta, \H) \|_2, \| \tilde{\A}_{21}(\mu, \gamma, \eta) \|)_2 \right\} 
    \tag{by \cref{helper:blocknorm}}
    \\
    \leq & \max \left\{ 1 + \frac{\gamma^2 \mu}{3\eta}, 1 \right\} + \max \left\{ \frac{\gamma^2 \mu}{3\eta}, \frac{2\gamma^2 \mu}{3\eta} \right\}
    \leq 1 + \frac{\gamma^2 \mu}{\eta}.
    \tag{\cref{eq:fedacii:stab:bound:11,eq:fedacii:stab:bound:12,eq:fedacii:stab:bound:21,eq:fedacii:stab:bound:22}}
  \end{align}
  \paragraph{Case II: $\gamma = \eta$.} In this case we have
  \begin{align*}
    \|\tilde{\A}_{11}(\mu, \gamma, \eta, \H)\|_2 &
    \leq 1 - \eta \mu,
    \\
    \|\tilde{\A}_{12}(\mu, \gamma, \eta, \H)\|_2 &
    \leq \frac{3 \eta \mu - 6 \eta^2 \mu^2 + 3 \eta^3 \mu^3}{9 - 6 \eta \mu - \eta^2 \mu^2},
    \\
    \|\tilde{\A}_{21}(\mu, \gamma, \eta)\|_2    &
    = 0,
    \\
    \|\tilde{\A}_{22}(\mu, \gamma, \eta)\|_2    & = \frac{9- 15 \eta \mu + 6 \eta^2 \mu^2}{9 - 6 \eta \mu - \eta^2 \mu^2} = 1  - \frac{9 \eta \mu - 7 \eta^2 \mu^2}{9 - 6 \eta \mu - \eta^2 \mu^2}.
  \end{align*}

 Similarly, the operator norm of block matrix $\tilde{\A}$ can be bounded via its blocks via \cref{helper:blocknorm} as
   \begin{align}
         & \tilde{\A}(\mu, \gamma, \eta, \H)
    \nonumber \\
    \leq & \max \left\{\| \tilde{\A}_{11}(\mu, \gamma, \eta, \H) \|_2, \| \tilde{\A}_{22}(\mu, \gamma, \eta) \|)_2 \right\}
    +
    \max \left\{\| \tilde{\A}_{12}(\mu, \gamma, \eta, \H) \|_2, \| \tilde{\A}_{21}(\mu, \gamma, \eta) \|)_2 \right\} 
    \tag{\cref{helper:blocknorm}}
    \\
    \leq & \max \left\{ 1 - \eta \mu + \frac{3 \eta \mu - 6 \eta^2 \mu^2 + 3 \eta^3 \mu^3}{9 - 6 \eta \mu - \eta^2 \mu^2}, \frac{9- 15 \eta \mu + 6 \eta^2 \mu^2}{9 - 6 \eta \mu - \eta^2 \mu^2}
    + \frac{3 \eta \mu - 6 \eta^2 \mu^2 + 3 \eta^3 \mu^3}{9 - 6 \eta \mu - \eta^2 \mu^2}\right\}
    \nonumber \\
    \leq & \max \left\{ 1 - \frac{6 \eta \mu - 4 \eta^3 \mu^3}{9 - 6 \eta \mu - \eta^2 \mu^2} , 1 - \frac{6 \eta \mu - \eta^2\mu^2 - 3 \eta^3 \mu^3}{9 - 6 \eta \mu - \eta^2 \mu^2} \right\} \leq 1.
    \nonumber 
  \end{align}
  Summarizing the above two cases completes the proof of \cref{fedacii:stab:bound}.
\end{proof}

\subsubsection{Proof of \cref{fedacii:stab:2}}
\label{sec:fedacii:stab:2}
In this section we apply  \cref{fedacii:stab:1,fedacii:stab:bound} to establish \cref{fedacii:stab:2}.
\begin{proof}[Proof of \cref{fedacii:stab:2}]
  Multiplying $\X(\gamma, \eta)^{-1}$ to the left on both sides of \cref{fedacii:stab:1} gives (we omit the details since the reasoning is the same as in the proof of \cref{fedaci:stab:2}.
  \begin{align}
    & \X(\gamma, \eta)^{-1} \begin{bmatrix}
      \bDelta^{(r,k+1)}_{\mathrm{ag}}
      \\
      \bDelta^{(r,k+1)}
    \end{bmatrix}
    \nonumber \\
    = &
    \X(\gamma, \eta)^{-1} \A(\mu, \gamma, \eta, \H^{(r,k)}) \X(\gamma, \eta)^{-1} \left( \X(\gamma, \eta) \begin{bmatrix}
        \bDelta^{(r,k)}_{\mathrm{ag}}
        \\
        \bDelta^{(r,k)}
      \end{bmatrix}  \right)
    -
    \begin{bmatrix}   \gamma \I \\ \zeros \end{bmatrix} \bDelta^{(r,k)}_{\varepsilon}.
    \label{eq:fedacii:stab:2:1}
  \end{align}
  Before we proceed, we introduce a few more notations to simplify the discussion. 
  Denote the shortcut $\tilde{\A} := \X(\gamma, \eta)^{-1} \A(\mu, \gamma, \eta, \H^{(r,k)}) \X(\gamma, \eta)$,  $\X = \X(\gamma, \eta)$,
  $\tilde{\bDelta} :=  \X^{-1}\begin{bmatrix} \bDelta^{(r,k)}_{\mathrm{ag}} \\  \bDelta^{(r,k)} \end{bmatrix}$, and 
  $\tilde{\bDelta}_{\varepsilon} :=  \begin{bmatrix} \gamma \I \\ \zeros \end{bmatrix} \bDelta^{(r,k)}_{\varepsilon}$.
  Then \cref{eq:fedacii:stab:2:1} becomes $\tilde{\bDelta}^{(r,k+1)} = \tilde{\A} \tilde{\bDelta} - \tilde{\bDelta}_{\varepsilon}$. Thus
  \begin{align}
         & \expt \left[ \left\| \X^{-1}\begin{bmatrix} \bDelta^{(r,k)}_{\mathrm{ag}} \\  \bDelta^{(r,k+1)} \end{bmatrix} \right\|_2^4 |\mathcal{F}^{(r,k)} \right] = \expt \left[ \|  \tilde{\A} \tilde{\bDelta} - \tilde{\bDelta}_{\varepsilon}  \|_2^4 |\mathcal{F}^{(r,k)} \right]
    \tag{by \cref{fedacii:stab:1}}
    \\
    =    & \expt \left[  \left(  \| \tilde{\A} \tilde{\bDelta} \|_2^2 
      +
      \|\tilde{\bDelta}_{\varepsilon} \|_2^2 
      -
      2 \langle \tilde{\A} \tilde{\bDelta} , \tilde{\bDelta}_{\varepsilon}  \rangle \right)^2   \right]
    \nonumber  \\
    =    & \| \tilde{\A} \tilde{\bDelta} \|_2^4
    + \expt \left[ \| \tilde{\bDelta}_{\varepsilon} \|_2^4 |\mathcal{F}^{(r,k)} \right]
    + 4 \expt \left[     \langle \tilde{\A} \tilde{\bDelta} , \tilde{\bDelta}_{\varepsilon}  \rangle^2 |\mathcal{F}^{(r,k)}     \right]
    + 2 \| \tilde{\A} \tilde{\bDelta} \|_2^2  \expt \left[ \| \tilde{\bDelta}_{\varepsilon} \|_2^2  |\mathcal{F}^{(r,k)} \right]
    \nonumber \\
         &
    - 4  \| \tilde{\A} \tilde{\bDelta} \|_2^2   \expt \left[\langle \tilde{\A} \tilde{\bDelta} , \tilde{\bDelta}_{\varepsilon}  \rangle |\mathcal{F}^{(r,k)}     \right]
    - 4 \expt \left[  \| \tilde{\bDelta}_{\varepsilon} \|_2^2    \langle \tilde{\A} \tilde{\bDelta} , \tilde{\bDelta}_{\varepsilon}  \rangle |\mathcal{F}^{(r,k)}     \right]
    \nonumber \\
    =    & \| \tilde{\A} \tilde{\bDelta} \|_2^4 + \expt \left[ \| \tilde{\bDelta}_{\varepsilon} \|_2^4 |\mathcal{F}^{(r,k)} \right]
    + 4 \expt \left[     \langle \tilde{\A} \tilde{\bDelta} , \tilde{\bDelta}_{\varepsilon}  \rangle^2 |\mathcal{F}^{(r,k)}     \right]
    + 2 \| \tilde{\A} \tilde{\bDelta} \|_2^2  \expt \left[ \| \tilde{\bDelta}_{\varepsilon} \|_2^2  |\mathcal{F}^{(r,k)} \right]
    \nonumber \\
         &
    - 4 \expt \left[  \| \tilde{\bDelta}_{\varepsilon} \|_2^2    \langle \tilde{\A} \tilde{\bDelta} , \tilde{\bDelta}_{\varepsilon}  \rangle |\mathcal{F}^{(r,k)}     \right]
    \tag{by independence and $\expt [\tilde{\bDelta}_{\varepsilon} |\mathcal{F}^{(r,k)}] = 0$}
    \\
    \leq & \| \tilde{\A} \tilde{\bDelta} \|_2^4 + \expt \left[ \| \tilde{\bDelta}_{\varepsilon} \|_2^4 |\mathcal{F}^{(r,k)} \right]
    + 6 \| \tilde{\A} \tilde{\bDelta} \|_2^2  \expt \left[ \| \tilde{\bDelta}_{\varepsilon} \|_2^2  |\mathcal{F}^{(r,k)} \right]
    + 4 \| \tilde{\A} \tilde{\bDelta} \| \expt \left[  \| \tilde{\bDelta}_{\varepsilon} \|^3   |\mathcal{F}^{(r,k)}     \right]
    \tag{Cauchy-Schwarz inequality}
    \\
    \leq & \| \tilde{\A} \tilde{\bDelta} \|_2^4 + 5 \expt \left[ \| \tilde{\bDelta}_{\varepsilon} \|_2^4 |\mathcal{F}^{(r,k)} \right]
    + 7 \| \tilde{\A} \tilde{\bDelta} \|_2^2  \expt \left[ \| \tilde{\bDelta}_{\varepsilon} \|_2^2  |\mathcal{F}^{(r,k)} \right]
    \tag{AM-GM inequality}
    \\
    \leq & \| \tilde{\A} \tilde{\bDelta} \|_2^4
    + 40 \gamma^4  \sigma^4
    + 14 \gamma^2 \sigma^2\| \tilde{\A} \tilde{\bDelta} \|_2^2 
    \tag{bounded \nth{4} central moment via \cref{helper:diff:4th}}
    \\
    \leq & \left( \|\tilde{\A} \tilde{\bDelta}\|_2^2  + 7 \gamma^2 \sigma^2 \right)^2 \leq  \left( \|\tilde{\A}\|_2^2  \|\tilde{\bDelta}\|_2^2  + 7 \gamma^2 \sigma^2 \right)^2.
    \nonumber 
  \end{align}
  Applying \cref{fedacii:stab:bound},
  \begin{equation*}
    \sqrt{\expt \left[ \left\| \X^{-1}\begin{bmatrix} \bDelta^{(r,k)}_{\mathrm{ag}} \\  \bDelta^{(r,k)} \end{bmatrix} \right\|_2^4 |\mathcal{F}^{(r,k)} \right]}
    \leq
    7 \gamma^2 \sigma^2 +  \| \tilde{\bDelta}\|_2^2  \cdot  
    \begin{cases}
      \left(1 + \frac{\gamma^2 \mu}{\eta}\right)^2 & \text{if~} \gamma \in \left(\eta, \sqrt{\frac{\eta}{\mu}}\right], \\
      1                              & \text{if~} \gamma =  \eta.
    \end{cases}
  \end{equation*}
  Resetting the notations completes the proof.
\end{proof}

\subsubsection{Proof of \cref{fedacii:stab:3}}
\label{sec:fedacii:stab:3}
In this section we will prove \cref{fedacii:stab:3} in two steps via the following two claims. For both two claims $\X$ stands for the matrix-valued functions defined in \cref{eq:fedacii:X:def}.

\begin{claim} \label{fedacii:stab:3:1}
  In the same setting of \cref{fedacii:stab:main}, the following inequality holds (for all possible $r, k$)
  \begin{align*}
    & \left\|  \nabla F(\overline{\x^{(r,k)}_{\mathrm{md}}}) -  \frac{1}{M} \sum_{m=1}^M \nabla F (\x^{(r,k)}_{\mathrm{md}, m})  \right\|_2^2
    \\
    \leq &
    \frac{Q^2}{4} \left\| \X(\gamma, \eta)^\top
  \begin{bmatrix} \frac{9 - 9 \gamma \mu + 2 \gamma^2 \mu^2}{9 - 6 \gamma \mu - \gamma^2 \mu^2}  \I \\ \frac{3 \gamma \mu - 3\gamma^2 \mu^2}{9 - 6 \gamma \mu - \gamma^2 \mu^2} \I \end{bmatrix}
     \right\|_2^4
     \left\| \X(\gamma, \eta)^{-1} \begin{bmatrix}
      \bDelta^{(r,k)}_{\mathrm{ag}}
      \\
      \bDelta^{(r,k)}
    \end{bmatrix}  \right\|_2^4.
  \end{align*}
\end{claim}
\begin{claim}
  \label{fedacii:stab:left}
  Assume $\mu > 0$, $\gamma \in [\eta,\sqrt{\frac{\eta}{\mu}}]$, then
  \(
    \left\| \X(\gamma, \eta)^\top
  \begin{bmatrix} \frac{9 - 9 \gamma \mu + 2 \gamma^2 \mu^2}{9 - 6 \gamma \mu - \gamma^2 \mu^2}  \I \\ \frac{3 \gamma \mu - 3\gamma^2 \mu^2}{9 - 6 \gamma \mu - \gamma^2 \mu^2} \I \end{bmatrix} \right\|_2 \leq
    \frac{\sqrt{17} \eta }{3\gamma}.
  \)
\end{claim}

\begin{proof}[Proof of \cref{fedacii:stab:3}]
  Follow trivially with \cref{fedaci:stab:3:1,fedacii:stab:left} as
  \begin{align*}
      \left\|  \nabla F(\overline{\x^{(r,k)}_{\mathrm{md}}}) -  \frac{1}{M} \sum_{m=1}^M \nabla F (\x^{(r,k)}_{\mathrm{md}, m})  \right\|_2^2
      & \leq
      \frac{Q^2}{4} \left( \frac{\sqrt{17} \eta}{3 \gamma} \right)^4 \left\| \X(\gamma, \eta)^{-1} \begin{bmatrix}
        \bDelta^{(r,k)}_{\mathrm{ag}}
        \\
        \bDelta^{(r,k)}
      \end{bmatrix}  \right\|_2^4
      \\
      & = 
      \frac{289 \eta^4 Q^2}{324 \gamma^4} \left\| \X(\gamma, \eta)^{-1} \begin{bmatrix}
        \bDelta^{(r,k)}_{\mathrm{ag}}
        \\
        \bDelta^{(r,k)}
      \end{bmatrix}  \right\|_2^4.
  \end{align*}
\end{proof}

Now we finish the proof of these two claims.
\begin{proof}[Proof of \cref{fedacii:stab:3:1}]
  Helper \cref{helper:3rd:Lip} shows that $\left\|  \nabla F(\overline{\x^{(r,k)}_{\mathrm{md}}}) -  \frac{1}{M} \sum_{m=1}^M \nabla F (\x^{(r,k)}_{\mathrm{md}, m})  \right\|_2^2$ can be bounded by $\nth{4}$-moment of difference:
  \begin{align}
    &  \left\|  \nabla F(\overline{\x^{(r,k)}_{\mathrm{md}}}) -  \frac{1}{M} \sum_{m=1}^M \nabla F (\x^{(r,k)}_{\mathrm{md}, m})  \right\|_2^2
    \leq
     \frac{Q^2}{4} \cdot \frac{1}{M} \sum_{m=1}^M  \|\x^{(r,k)}_{\mathrm{md},m} - \overline{\x^{(r,k)}_{\mathrm{md}}}\|_2^4
    \tag{\cref{helper:3rd:Lip}}
\\
\leq & \frac{Q^2}{4} \| \bDelta^{(r,k)}_{\mathrm{md}} \|_2^4 
\tag{convexity of $\|\cdot\|_2^4$}
\\
=  & 
\frac{Q^2}{4}
 \left\|
 \begin{bmatrix} (1 - \beta^{-1}) \I \\ \beta^{-1} \I \end{bmatrix}^\top
 \begin{bmatrix} \bDelta^{(r,k)}_{\mathrm{ag}} \\ \bDelta^{(r,k)} \end{bmatrix}
 \right\|_2^4 
\tag{definition of ``md''}
\\
\leq & \frac{Q^2}{4} \left\| \X(\gamma, \eta)^\top \begin{bmatrix} (1 - \beta^{-1}) \I \\ \beta^{-1} \I \end{bmatrix} \right\|_2^4
\cdot \left\| \X(\gamma, \eta)^{-1}
 \begin{bmatrix} \bDelta^{(r,k)}_{\mathrm{ag}} \\ \bDelta^{(r,k)} \end{bmatrix}
 \right\|_2^4.
\tag{sub-multiplicativity}
\\
= & \frac{Q^2}{4} \left\| \begin{bmatrix} \frac{9 - 9 \gamma \mu + 2 \gamma^2 \mu^2}{9 - 6 \gamma \mu - \gamma^2 \mu^2}  \I \\ \frac{3 \gamma \mu - 3\gamma^2 \mu^2}{9 - 6 \gamma \mu - \gamma^2 \mu^2} \I \end{bmatrix} \right\|_2^4
\cdot \left\| \X(\gamma, \eta)^{-1}
 \begin{bmatrix} \bDelta^{(r,k)}_{\mathrm{ag}} \\ \bDelta^{(r,k)} \end{bmatrix}
 \right\|_2^4.
 \nonumber 
\end{align}
\end{proof}

\begin{proof}[Proof of \cref{fedacii:stab:left}]
  Direct calculation shows that
  \begin{equation*}
    \X(\gamma, \eta)^\top
    \begin{bmatrix} \frac{9 - 9 \gamma \mu + 2 \gamma^2 \mu^2}{9 - 6 \gamma \mu - \gamma^2 \mu^2}  \I \\ \frac{3 \gamma \mu - 3\gamma^2 \mu^2}{9 - 6 \gamma \mu - \gamma^2 \mu^2} \I \end{bmatrix}
    =
    \begin{bmatrix}  \frac{3 \gamma^2 \mu (1-\gamma \mu) + \eta (3 - \gamma \mu)(3 - 2 \gamma \mu)}{\gamma (9 - 6 \gamma \mu - \gamma^2\mu^2)}  \I \\ \frac{3 \gamma^2 \mu ( 1- \gamma \mu)}{\gamma (9 - 6 \gamma \mu - \gamma^2\mu^2)}  \I \end{bmatrix}.
  \end{equation*}
  Since $\gamma^2 \mu \leq \eta$ and $\gamma \mu \leq 1$, we have
  \begin{equation*}
    0 \leq
    \frac{3 \gamma^2 \mu (1-\gamma \mu) + \eta (3 - \gamma \mu)(3 - 2 \gamma \mu)}{\gamma (9 - 6 \gamma \mu - \gamma^2\mu^2)}
    \leq
    \frac{\eta}{\gamma} \cdot \frac{12 - 12 \gamma \mu + 2 \gamma^2 \mu^2}{9 - 6 \gamma \mu - \gamma^2\mu^2}
    \leq
    \frac{4\eta}{3\gamma},
  \end{equation*}
  and
  \begin{equation*}
    0 \leq \frac{3 \gamma^2 \mu (1-\gamma \mu)}{\gamma (9 - 6 \gamma \mu - \gamma^2\mu^2)}
    \leq \frac{\eta}{\gamma} \cdot \frac{3  (1-\gamma \mu)}{9 - 6 \gamma \mu - \gamma^2\mu^2}
    \leq \frac{\eta}{3\gamma}.
  \end{equation*}
  Consequently,
  \begin{equation*}
    \left\|     \begin{bmatrix}  \frac{3 \gamma^2 \mu (1-\gamma \mu) + \eta (3 - \gamma \mu)(3 - 2 \gamma \mu)}{\gamma (9 - 6 \gamma \mu - \gamma^2\mu^2)}  \I \\ \frac{3 \gamma^2 \mu ( 1- \gamma \mu)}{\gamma (9 - 6 \gamma \mu - \gamma^2\mu^2)}  \I \end{bmatrix} \right\|_2
    \leq
    \sqrt{\left( \frac{4 \eta}{3\gamma} \right)^2 + \left( \frac{\eta}{3\gamma} \right)^2}
    \leq
    \frac{\sqrt{17} \eta }{3 \gamma}.
  \end{equation*}
\end{proof}

\subsection{Convergence of \fedacii under \cref{asm:fo:scvx:2o}: Complete version of \cref{fedac:a1}(b)}
\label{sec:fedacii:a1}
\subsubsection{Main theorem and lemma}
In this subsection we establish the convergence of \fedacii under \cref{asm:fo:scvx:2o}. 
We will provide a complete, non-asymptotic version of \cref{fedac:a1}(b) and provide the proof. 
\begin{theorem}[Convergence of \fedacii under \cref{asm:fo:scvx:2o}, complete version of \cref{fedac:a1}(b)]
  \label{fedacii:a1:full}
  Let $F$ be $\mu > 0$ strongly convex, and assume \cref{asm:fo:scvx:2o}, then for 
  \begin{equation*}
    \eta = \min \left\{ \frac{1}{L}, \frac{9}{\mu K R^2} \log^2 \left( \euler
     + \min \left\{ \frac{\mu M K R \Phi^{(0,0)}}{\sigma^2} + \frac{\mu^3 K R^4 \Phi^{(0,0)}}{L^2 \sigma^2}\right\} \right)\right\},
  \end{equation*}
  \fedacii yields
  \begin{align*}
    \expt[\Phi^{(R,0)}] \leq & \min \left\{ \exp \left( - \frac{\mu K R}{3 L} \right), \exp \left( - \frac{\mu^{\frac{1}{2}} K^{\frac{1}{2}} R}{3 L^{\frac{1}{2}} } \right)\right\} \Phi^{(0,0)} 
    \\
    & 
    + \frac{4 \sigma^2}{\mu M K R} \log \left( \euler + \frac{\mu M K R \Phi^{(0,0)}}{\sigma^2} \right)
    + \frac{8101 L^2 \sigma^2}{\mu^3 K R^4} \log^4\left( \euler + \frac{\mu^3 K R^4 \Phi^{(0,0)}}{L^2 \sigma^2} \right),
  \end{align*}
  where $\Phi^{(r,k)}$ is the ``centralized'' potential function defined in \cref{eq:centralied:potential}.
\end{theorem} 
\begin{remark}
  The simplified version \cref{fedac:a1}(b) in the main body can be obtained by upper bounding $\Phi^{(0,0)}$ by $LB^2$.
\end{remark}
Note that most of the results established towards \cref{fedacii:a2:full} can be recycled as long as it does not assume \cref{asm:fo:scvx:3o}. 
In particular, we will reuse the perturbed iterate analysis \cref{fedacii:conv:main}, and provide an alternative version of discrepancy overhead bounds, as shown in \cref{fedacii:a1:stab:main}. 
The only difference is that now we use $L$-smoothness to bound the discrepancy term. 
\begin{lemma}[Discrepancy overhead bounds]
  \label{fedacii:a1:stab:main}
  Let $F$ be $\mu > 0$-strongly convex, and assume \cref{asm:fo:scvx:2o}, then for $\alpha = \frac{3}{2 \gamma \mu} - \frac{1}{2}$, $\beta =\frac{2 \alpha^2 - 1}{\alpha - 1}$, $\gamma \in [\eta, \sqrt{ \frac{\eta}{\mu}}]$, $\eta \in (0, \frac{1}{L}]$, \fedac satisfies (for all $(r, k)$)
  \begin{align*}
     & \expt \left[ \left\|  \nabla F(\overline{\x^{(r,k)}_{\mathrm{md}}}) -  \frac{1}{M} \sum_{m=1}^M \nabla F (\x^{(r,k)}_{\mathrm{md}, m})  \right\|_2^2\right]
    \leq  \begin{cases}
      4 \eta^2 L^2 K \sigma^2 \left(1 + \frac{\gamma^2\mu}{\eta}\right)^{2K}
       & \text{if~} \gamma \in \left(\eta, \sqrt{\frac{\eta}{\mu}} \right],
      \\
      4 \eta^2 L^2 K \sigma^2
       &
      \text{if~} \gamma = \eta.
    \end{cases}
  \end{align*}
\end{lemma}
The proof of \cref{fedacii:a1:stab:main} is deferred to \cref{sec:fedacii:a1:stab:main}.

Now plug in the choice of $\gamma = \max \left\{ \sqrt{\frac{\eta}{\mu K}}, \eta\right\}$ to \cref{fedacii:conv:main,fedacii:a1:stab:main}, which leads to the following lemma.
\begin{lemma}[Convergence of \fedacii for general $\eta$ under \cref{asm:fo:scvx:2o}]
  \label{fedacii:a1:general:eta}
    Let $F$ be $\mu > 0$-strongly convex, and assume \cref{asm:fo:scvx:2o}, then for any $\eta \in (0, \frac{1}{L}]$, \fedacii yields
    \begin{equation}
      \expt[\Phi^{(R,0)}]
      \leq \exp \left(  - \frac{1}{3} \max \left\{ \eta \mu, \sqrt{\frac{\eta \mu}{K}}\right\} K R \right) \Phi^{(0,0)}
      + \frac{\eta^{\frac{1}{2}} \sigma^2}{\mu^{\frac{1}{2}} M K^{\frac{1}{2}} }
      + \frac{100 \eta^2 L^2 K \sigma^2}{\mu}.
      \label{eq:fedacii:a1:general:eta}
    \end{equation}
\end{lemma}
\begin{proof}[Proof of \cref{fedacii:a1:general:eta}]
  Applying \cref{fedacii:conv:main} yields
  \begin{align*}
    \expt[\Phi^{(R,0)}] \leq & \exp\left( - \frac{1}{3} \max \left\{ \eta \mu, \sqrt{\frac{\eta \mu}{K}}\right\} K R \right) \Phi^{(0,0)}
    +
    \min\left\{ \frac{3\eta L \sigma^2}{2 \mu M}, \frac{3 \eta^{\frac{3}{2}} L K^{\frac{1}{2}} \sigma^2}{2 \mu^{\frac{1}{2}} M} \right\}
    \nonumber  \\
                       & + \max\left\{ \frac{\eta \sigma^2}{2M}, \frac{\eta^{\frac{1}{2}} \sigma^2}{2 \mu^{\frac{1}{2}} M K^{\frac{1}{2}} }          \right\}
    + \frac{3}{\mu} \max_{\substack{0 \leq r < R \\ 0 \leq k < K}}  \expt \left[ \left\|  \nabla F(\overline{\x^{(r,k)}_{\mathrm{md}}}) -  \frac{1}{M} \sum_{m=1}^M \nabla F (\x^{(r,k)}_{\mathrm{md}, m})  \right\|_2^2\right].
  \end{align*}
  Applying \cref{fedacii:a1:stab:main} yields (for all $r, k$)
  \begin{align*}
         & \frac{3}{\mu} \expt \left[ \left\|  \nabla F(\overline{\x^{(r,k)}_{\mathrm{md}}}) -  \frac{1}{M} \sum_{m=1}^M \nabla F (\x^{(r,k)}_{\mathrm{md}, m})  \right\|_2^2\right]
    \leq
    \begin{cases}
      12 \mu^{-1} \eta^2 L^2 K \sigma^2  \left(1 + \frac{1}{K}\right)^{2K}
       & \text{if~} \gamma = \sqrt{\frac{\eta}{\mu K}},
      \\
      12 \mu^{-1} \eta^2 L^2 K \sigma^2
       &
      \text{if~} \gamma = \eta
    \end{cases}
    \\
    \leq & 12 \euler^{2} \mu^{-1} \eta^2 L^2 K \sigma^2.
  \end{align*}
  Note that
  \begin{align}
         & \min\left\{ \frac{3\eta L \sigma^2}{2 \mu M}, \frac{3 \eta^{\frac{3}{2}} L K^{\frac{1}{2}} \sigma^2}{2 \mu^{\frac{1}{2}} M} \right\}
    + \max\left\{ \frac{\eta \sigma^2}{2M}, \frac{\eta^{\frac{1}{2}} \sigma^2}{2 \mu^{\frac{1}{2}} M K^{\frac{1}{2}} }          \right\}
    \nonumber \\
    \leq & \frac{3 \eta^{\frac{3}{2}} L K^{\frac{1}{2}} \sigma^2}{2 \mu^{\frac{1}{2}} M} + \frac{\eta \sigma^2}{2M} + \frac{\eta^{\frac{1}{2}} \sigma^2}{2 \mu^{\frac{1}{2}} M K^{\frac{1}{2}} }
    \nonumber \\
    \leq &
    \frac{7 \eta^{\frac{3}{2}} L K^{\frac{1}{2}} \sigma^2}{4 \mu^{\frac{1}{2}} M} + \frac{3 \eta^{\frac{1}{2}} \sigma^2}{4 \mu^{\frac{1}{2}} M K^{\frac{1}{2}} }.
    \tag{by AM-GM inequality, and $\mu \leq L$}
  \end{align}
  By Young's inequality,
  \begin{align}
    \frac{7 \eta^{\frac{3}{2}} L K^{\frac{1}{2}} \sigma^2}{4 \mu^{\frac{1}{2}} M} 
    \leq & \left( \frac{3}{4} \frac{\eta^{\frac{1}{2}} \sigma^2}{\mu^{\frac{1}{2}} M K^{\frac{1}{2}} } \right)^{\frac{1}{3}}
    \left( 3 \cdot \frac{\eta^2 L^\frac{3}{2} K \sigma^2}{\mu^{\frac{1}{2}} M} \right)^{\frac{2}{3}}
    \tag{since $\frac{7}{4} \leq \left( \frac{3}{4} \right)^{\frac{1}{3}} (3)^{\frac{2}{3}} $ }
    \\
    \leq & \frac{1}{4} \cdot \frac{\eta^{\frac{1}{2}} \sigma^2}{\mu^{\frac{1}{2}} M K^{\frac{1}{2}} } + 2 \cdot \frac{\eta^2 L^\frac{3}{2} K \sigma^2}{\mu^{\frac{1}{2}} M}
    \tag{by Young's inequality}
    \\
    \leq & \frac{\eta^{\frac{1}{2}} \sigma^2}{4 \mu^{\frac{1}{2}} M K^{\frac{1}{2}} } + \frac{2\eta^2 L^2 K \sigma^2}{\mu}.
    \tag{since $L \geq \mu$ and $M \geq 1$}
  \end{align}
  Combining the above inequalities gives
  \begin{equation*}
    \expt[\Phi^{(R,0)}]
    \leq \exp \left(  - \frac{1}{3} \max \left\{ \eta \mu, \sqrt{\frac{\eta \mu}{K}}\right\} K R \right) \Phi^{(0,0)}
    + \frac{\eta^{\frac{1}{2}} \sigma^2}{\mu^{\frac{1}{2}} M K^{\frac{1}{2}} }
    + \frac{(12 \euler^2 + 2) \eta^2 L^2 K \sigma^2}{\mu}.
  \end{equation*}
  The proof then follows by the estimate $12 \euler^2 + 2 < 100$.
\end{proof}

\cref{fedacii:a1:full} then follows by plugging in the appropriate $\eta$ to \cref{fedacii:a1:general:eta}.
\begin{proof}[Proof of \cref{fedacii:a1:full}]
  To simplify the notation, we denote the decreasing term in \cref{eq:fedacii:a1:general:eta} in \cref{fedacii:a1:general:eta} as $\varphi_{\downarrow}(\eta)$ and the increasing term as $\varphi_{\uparrow}(\eta)$, namely
  \begin{align*}
    \varphi_{\downarrow}(\eta) := \exp \left(  - \frac{1}{3} \max \left\{ \eta \mu, \sqrt{\frac{\eta \mu}{K}}\right\} K R \right) \Phi^{(0,0)},
    \quad
    \varphi_{\uparrow}(\eta) := \frac{\eta^{\frac{1}{2}} \sigma^2}{\mu^{\frac{1}{2}} M K^{\frac{1}{2}} }
    + \frac{100 \eta^2 L^2 K \sigma^2}{\mu}.
  \end{align*}
  Let
  \begin{equation*}
    \eta_0 := \frac{9}{\mu K R^2} \log^2 \left( \euler
     + \min \left\{ \frac{\mu M K R \Phi^{(0,0)}}{\sigma^2} + \frac{\mu^3 K R^4 \Phi^{(0,0)}}{L^2 \sigma^2}\right\} \right)
     ,
     \quad
     \text{then }
     \eta = \min \left\{ \frac{1}{L}, \eta_0 \right\}.
  \end{equation*}
  Therefore,
  \begin{equation*}
    \varphi_{\downarrow}(\eta)
    \leq
    \min \left\{ \exp \left( - \frac{\mu K R}{3 L} \right), \exp \left( - \frac{\mu^{\frac{1}{2}} K^{\frac{1}{2}} R}{3 L^{\frac{1}{2}} } \right)\right\} \Phi^{(0,0)}
    +
    \frac{\sigma^2}{\mu M K R} + \frac{L^2 \sigma^2}{\mu^3 K R^4}.
  \end{equation*}
  and
  \begin{align*}
    \varphi_{\uparrow}(\eta)  \leq \varphi_{\uparrow}(\eta_0) 
    \leq
    & 
    \frac{3 \sigma^2}{\mu M K R} \log \left( \euler + \frac{\mu M K R \Phi^{(0,0)}}{\sigma^2} \right)
    + \frac{8100 L^2 \sigma^2}{\mu^3 K R^4} \log^4\left( \euler + \frac{\mu^3 K R^4 \Phi^{(0,0)}}{L^2 \sigma^2} \right).
  \end{align*}
  Consequently,
  \begin{align*}
    \expt[\Phi^{(R,0)}] \leq & \varphi_{\downarrow} \left(\frac{1}{L} \right) + \varphi_{\downarrow}(\eta_0) + \varphi_{\uparrow}(\eta_0) \leq \min \left\{ \exp \left( - \frac{\mu K R}{3 L} \right), \exp \left( - \frac{\mu^{\frac{1}{2}} K^{\frac{1}{2}} R}{3 L^{\frac{1}{2}} } \right)\right\} \Phi^{(0,0)} 
    \\
    & 
    + \frac{4 \sigma^2}{\mu M K R} \log \left( \euler + \frac{\mu M K R \Phi^{(0,0)}}{\sigma^2} \right)
    + \frac{8101 L^2 \sigma^2}{\mu^3 K R^4} \log^4\left( \euler + \frac{\mu^3 K R^4 \Phi^{(0,0)}}{L^2 \sigma^2} \right).
  \end{align*}
\end{proof}

\subsubsection{Proof of \cref{fedacii:a1:stab:main}}
\label{sec:fedacii:a1:stab:main}
We first introduce the supporting propositions for \cref{fedacii:a1:stab:main}. We omit most of the proof details since the analysis is largely shared.

The following proposition is parallel to \cref{fedacii:stab:2}, where the difference is that the present proposition analyzes the \nth{2}-order stability instead of \nth{4}-order.
\begin{proposition}
  \label{fedacii:a1:stab:2}
  In the same setting of \cref{fedacii:a1:stab:main}, the following inequality holds (for all possible $(r, k)$)   
    \begin{align*}
        & \expt \left[ \left\| \X(\gamma, \eta)^{-1} \begin{bmatrix}
          \bDelta^{(r,k+1)}_{\mathrm{ag}}
          \\
          \bDelta^{(r,k+1)}
        \end{bmatrix}  \right\|_2^2\middle| \mathcal{F}^{(r,k)} \right]
      \\
      \leq &
      2 \gamma^2 \sigma^2 +
      \left\| \X(\gamma, \eta)^{-1} \begin{bmatrix}
        \bDelta^{(r,k)}_{\mathrm{ag}}
        \\
        \bDelta^{(r,k)}
      \end{bmatrix}  \right\|_2^2 
      \cdot
      \begin{cases}
        \left(1 + \frac{\gamma^2 \mu}{\eta} \right)^2 & \text{if~} \gamma \in \left(\eta, \sqrt{\frac{\eta}{\mu}}\right], \\
        1                                              & \text{if~} \gamma =  \eta,
      \end{cases}
    \end{align*}
    where $\X$ is the matrix-valued function defined in \cref{eq:fedacii:X:def}.
\end{proposition}
\begin{proof}[Proof of \cref{fedacii:a1:stab:2}]
  Apply the uniform norm bound \cref{fedacii:stab:bound}, and the rest of the analysis is the same as \cref{fedaci:stab:2}.
\end{proof}

The following proposition is parallel to \cref{fedacii:stab:3}, where the difference is that the present proposition uses $L$-(\nth{2}-order)-smoothness to bound the LHS quantity.
\begin{proposition} 
  \label{fedacii:a1:stab:3}
  In the same setting of \cref{fedacii:a1:stab:main}, the following inequality holds (for all possible $(r, k)$)
  \begin{equation*}
    \left\|  \nabla F(\overline{\x^{(r,k)}_{\mathrm{md}}}) -  \frac{1}{M} \sum_{m=1}^M \nabla F (\x^{(r,k)}_{\mathrm{md}, m})  \right\|_2^2
    \leq
    \frac{17 \eta^2 L^2}{9 \gamma^2} \left\| \X(\gamma, \eta)^{-1} \begin{bmatrix}
      \bDelta^{(r,k)}_{\mathrm{ag}}
      \\
      \bDelta^{(r,k)}
    \end{bmatrix}  \right\|_2^2 ,
  \end{equation*}
  where $\X$ is the matrix-valued function defined in \cref{eq:fedacii:X:def}.
\end{proposition}
\begin{proof}[Proof of \cref{fedacii:a1:stab:3}]
  By $L$-smoothness (\cref{asm:fo:scvx:2o}(b)), 
  \begin{equation*}
    \left\|  \nabla F(\overline{\x^{(r,k)}_{\mathrm{md}}}) -  \frac{1}{M} \sum_{m=1}^M \nabla F (\x^{(r,k)}_{\mathrm{md}, m})  \right\|_2^2
    \leq
    L^2 \|\bDelta^{(r,k)}_{\mathrm{md}}\|_2^2 .
  \end{equation*}
  By definition of ``md'', sub-multiplicativity, and \cref{fedacii:stab:left},
  \begin{align*}
    \|\bDelta^{(r,k)}_{\mathrm{md}}\|_2^2 
    & =
    \left\|
    \X(\gamma, \eta)^{\top}
    \begin{bmatrix} \frac{9 - 9 \gamma \mu + 2 \gamma^2 \mu^2}{9 - 6 \gamma \mu - \gamma^2 \mu^2}  \I \\ \frac{3 \gamma \mu - 3\gamma^2 \mu^2}{9 - 6 \gamma \mu - \gamma^2 \mu^2} \I \end{bmatrix}
     \right\|_2^2 
     \left\| \X(\gamma, \eta)^{-1} \begin{bmatrix}
      \bDelta^{(r,k)}_{\mathrm{ag}}
      \\
      \bDelta^{(r,k)}
    \end{bmatrix}  \right\|_2^2 
    \\
    & \leq
    \frac{17 \eta^2}{9 \gamma^2} \left\| \X(\gamma, \eta)^{-1} \begin{bmatrix}
      \bDelta^{(r,k)}_{\mathrm{ag}}
      \\
      \bDelta^{(r,k)}
    \end{bmatrix}  \right\|_2^2 .
  \end{align*}
\end{proof}

\cref{fedacii:a1:stab:main} then follows by telescoping \cref{fedacii:a1:stab:2} and plugging in \cref{fedacii:a1:stab:3}.
\begin{proof}[Proof of \cref{fedacii:a1:stab:main}]
  Telescope \cref{fedacii:a1:stab:2} from $(r,0)$-th step to $(r,k)$-th step:
  \begin{equation*}
    \expt \left[ \left\| \X(\gamma, \eta)^{-1}
      \begin{bmatrix} \bDelta^{(r,k)}_{\mathrm{ag}} \\ \bDelta^{(r,k)} \end{bmatrix}
      \right\|_2^2\middle| \mathcal{F}^{(r,0)} \right]
    \leq
    2 \gamma^2 \sigma^2 K \cdot \begin{cases}
      \left(1 + \frac{\gamma^2 \mu}{\eta}\right)^{2K} & \text{if~} \gamma \in \left(\eta, \sqrt{\frac{\eta}{\mu}} \right],
      \\
      1                              & \text{if~} \gamma =  \eta.
    \end{cases}
  \end{equation*}
  Thus, by \cref{fedacii:a1:stab:3},
  \begin{equation*}
    \expt \left[ \left\|  \nabla F(\overline{\x^{(r,k)}_{\mathrm{md}}}) -  \frac{1}{M} \sum_{m=1}^M \nabla F (\x^{(r,k)}_{\mathrm{md}, m})  \right\|_2^2 \middle| \right]
    \leq
    \frac{34}{9} \eta^2 \sigma^2 K \cdot \begin{cases}
      \left(1 + \frac{\gamma^2 \mu}{\eta}\right)^{2K} & \text{if~} \gamma \in \left(\eta, \sqrt{\frac{\eta}{\mu}} \right],
      \\
      1                              & \text{if~} \gamma =  \eta.
    \end{cases}
  \end{equation*}
  The \cref{fedacii:a1:stab:main} then follows by bounding $\frac{34}{9}$ with $4$.
\end{proof}
% !TEX root = ../main.tex
\section{Analysis of \fedavg under \cref{asm:fo:scvx:3o}}
\label{sec:fedavg}
In this section we study the convergence of \fedavg under \cref{asm:fo:scvx:3o}. 
We provide a complete, non-asymptotic version of \cref{fedavg:a2} and provide the proof. 
We formally define \fedavg in \cref{alg:fedavg} for reference. 

Formally, we use $\mathcal{F}^{(r,k)}$ to denote the $\sigma$-algebra generated by $\{\x^{(\rho,\kappa)}_m\}$ for $\rho < r$ or $\rho = r$ but $\kappa \leq k$. Since \fedavg is Markovian, conditioning on $\mathcal{F}^{(r,k)}$ is equivalent to conditioning on $\{\x_m^{(r,k)}\}_{m \in [M]}$.

\subsection{Main theorem and lemma: Complete version of \cref{fedavg:a2}}
\begin{theorem}
  \label{fedavg:a2:full}
  Let $F$ be $\mu > 0$-strongly convex, and assume \cref{asm:fo:scvx:3o}, then for 
  \begin{equation*}
    \eta := \min \left\{ \frac{1}{4L}, \frac{2}{\mu K R} \log \left( \euler + \min \left\{ \frac{\mu^2 M K^2 R^2 B^2}{\sigma^2}, \frac{\mu^6 K^3 R^5 B^2}{Q^2 \sigma^4}\right\} \right)\right\},
  \end{equation*}
  \fedavg yields
  \begin{align*}
    &  \expt \left[ F \left( \sum_{r=0}^{R-1} \sum_{k=0}^{K-1} \rho^{(r,k)} \overline{\x^{(r,k)}} \right)  \right] - F^{\star}
    + \frac{\mu}{2} \expt [\|\overline{\x^{(R,0)}} - \x^{\star}\|_2^2 ]
    \\
    \leq &
    \exp \left( -\frac{\mu K R}{8 L} \right) 4L B^2
    +
    \frac{3 \sigma^2}{ \mu M K R } \log \left( \euler + \frac{\mu^2 M K^2 R^2 B^2}{\sigma^2}  \right)
    \\
    & \quad +
    \frac{3073 Q^2 \sigma^4}{\mu^5 K^2 R^4} \log^4 \left( \euler + \frac{\mu^6 K^3 R^5 B^2}{Q^2 \sigma^4}  \right).
  \end{align*}
  where $\rho^{(r,k)} := \frac{(1 - \frac{1}{2} \eta \mu)^{KR-(rK+k)-1}}{\sum_{t=0}^{KR-1} (1 - \frac{1}{2} \eta \mu)^{KR-t-1}}$, and $B = \|\x^{(0,0)} -\x^{\star}\|_2$.
\end{theorem}
The proof of \cref{fedavg:a2:full} is based on the following two lemmas regarding the convergence and \nth{4}-order stability of \fedavg. The averaging technique applied here is similar to \cite{Stich-arXiv19}.
\begin{lemma}[Perturbed iterate analysis for \fedavg under \cref{asm:fo:scvx:3o}]
  \label{fedavg:a2:conv:main}
  Let $F$ be $\mu > 0$-strongly convex, and assume \cref{asm:fo:scvx:3o}, then for $\eta \in (0, \frac{1}{4L}]$, \fedavg satisfies
  \begin{align*}
         & \expt \left[ F \left(\sum_{r=0}^{R-1} \sum_{k=0}^{K-1} \rho^{(r,k)} \overline{\x^{(r,k)}} \right)  \right] - F^{\star}
    + \frac{\mu}{2} \expt [\|\overline{\x^{(R,0)}} - \x^{\star}\|_2^2 ]
    \\
    \leq & \frac{1}{\eta} \exp \left( - \frac{1}{2} \eta \mu K R \right) B^2
    +
    \frac{1}{M} \eta \sigma^2 + \frac{Q^2}{\mu} \left( \max_{\substack{0 \leq r < R \\ 0 \leq k < K}}  \frac{1}{M} \sum_{m=1}^M \expt \left[\|\overline{\x^{(r,k)}} - \x^{(r,k)}_m\|_2^4 \right] \right).
  \end{align*}
  where $\rho^{(r, k)}$ is defined in the statement of \cref{fedavg:a2:full}.
\end{lemma}
The proof of \cref{fedavg:a2:conv:main} is deferred to \cref{sec:fedavg:a2:conv}.
\begin{proof}[Proof of \cref{fedavg:a2:full}]
  Combining \cref{fedavg:a2:conv:main,fedavg:a2:stab:main} gives
  \begin{align}
      & \expt \left[ F \left( \sum_{r=0}^{R-1} \sum_{k=0}^{K-1} \rho^{(r,k)} \overline{\x^{(r,k)}} \right)  \right] - F^{\star}
  + \frac{\mu}{2} \expt [\|\overline{\x^{(R,0)}} - \x^{\star}\|_2^2 ]
  \nonumber \\
    \leq &
  \frac{1}{\eta} \exp \left( - \frac{1}{2} \eta \mu K R \right) B^2
  +
  \frac{1}{M} \eta \sigma^2 + \frac{192 \eta^4 Q^2 K^2 \sigma^4 }{\mu}.
  \label{eq:fedavg:a2:general:eta}
  \end{align}
  To simplify the notation, denote the terms on the RHS of \cref{eq:fedavg:a2:general:eta} as
  \begin{equation*}
    \varphi_{\downarrow}(\eta) := \frac{1}{\eta} \exp \left( - \frac{1}{2} \eta \mu K R \right) B^2,
    \qquad
    \varphi_{\uparrow}(\eta) := 
    \frac{1}{M} \eta \sigma^2 + \frac{192 \eta^4 Q^2 K^2 \sigma^4 }{\mu}.
  \end{equation*}
  Let
  \begin{equation*}
    \eta_0 := \frac{2}{\mu K R} \log \left( \euler + \min \left\{ \frac{\mu^2 M K^2 R^2 B^2}{\sigma^2}, \frac{\mu^6 K^3 R^5 B^2}{Q^2 \sigma^4}\right\} \right),
    \qquad
    \text{then }
    \eta = \min \left\{ \frac{1}{4L}, \eta_0\right\}.
  \end{equation*}
  Therefore, $\varphi_{\downarrow}(\eta) \leq \varphi_{\downarrow}(\frac{1}{4L}) + \varphi_{\downarrow}(\eta_0)$, where
  \begin{equation}
    \varphi_{\downarrow} \left( \frac{1}{4L} \right) = \exp \left( -\frac{\mu K R}{8 L} \right) 4L B^2,
    \label{eq:fedavg:a2:proof:1}
  \end{equation} 
  and
  \begin{equation}
    \varphi_{\downarrow}(\eta_0) \leq \frac{\mu KR}{2} B^2 \cdot \left( \min \left\{ \frac{\mu^2 M K^2 R^2 B^2}{\sigma^2}, \frac{\mu^6 K^3 R^5 B^2}{Q^2 \sigma^4}\right\}  \right)^{-1}
    \leq
    \frac{\sigma^2}{2 \mu M K R} + \frac{Q^2 \sigma^4}{2 \mu^5 K^2 R^4}.
    \label{eq:fedavg:a2:proof:2}
  \end{equation}
  On the other hand
  \begin{align}
    \varphi_{\uparrow}(\eta) \leq \varphi_{\uparrow}(\eta_0)
    \leq
    \frac{2 \sigma^2}{\mu M K R} \log \left( \euler + \frac{\mu^2 M K^2 R^2 B^2}{\sigma^2}  \right)
    +
    \frac{3072 Q^2 \sigma^4}{\mu^5 K^2 R^4} \log^4 \left( \euler + \frac{\mu^6 K^3 R^5 B^2}{Q^2 \sigma^4}  \right).
    \label{eq:fedavg:a2:proof:3}
  \end{align}
  Combining \cref{eq:fedavg:a2:general:eta,eq:fedavg:a2:proof:1,eq:fedavg:a2:proof:2,eq:fedavg:a2:proof:3} completes the proof.
  % \begin{align*}
  %   &  \expt \left[ F \left( \sum_{r=0}^{R-1} \sum_{k=0}^{K-1} \rho^{(r,k)} \overline{\x^{(r,k)}} \right)  \right] - F^{\star}
  %   + \frac{\mu}{2} \expt [\|\overline{\x^{(R,0)}} - \x^{\star}\|_2^2 ]
  %   \\
  %   \leq &
  %   \exp \left( -\frac{\mu K R}{8 L} \right) 4L B^2
  %   +
  %   \frac{3 \sigma^2}{ \mu M K R } \log \left( \euler + \frac{\mu^2 M K^2 R^2 B^2}{\sigma^2}  \right)
  %   +
  %   \frac{3073 Q^2 \sigma^4}{\mu^5 K^2 R^4} \log^4 \left( \euler + \frac{\mu^6 K^3 R^5 B^2}{Q^2 \sigma^4}  \right).
  % \end{align*}
\end{proof}

\subsection{Perturbed iterative analysis for \fedavg: Proof of \cref{fedavg:a2:conv:main}}
\label{sec:fedavg:a2:conv}
We first state and proof the following proposition on one-step analysis.
\begin{proposition}
  \label{fedavg:a2:conv:onestep}
  Under the same assumption of \cref{fedavg:a2:conv:main}, for all $(r, k)$, the following inequality holds
  \begin{align*}
         & \expt \left[ \|\overline{\x^{(r,k+1)}} - \x^{\star}\|_2^2   |\mathcal{F}^{(r,k)} \right]
    \\
    \leq &
    \left( 1 - \frac{1}{2}\eta \mu  \right)  \|\overline{\x^{(r,k)}} - \x^{\star} \|_2^2  -  \eta (F(\overline{\x^{(r,k)}} ) - F^{\star})
    +  \frac{\eta Q^2}{\mu M}  \sum_{m=1}^M \|\overline{\x^{(r,k)}} - \x^{(r,k)}_m\|_2^4  + \frac{\eta^2\sigma^2}{M}.
  \end{align*}
\end{proposition}

\begin{proof}[Proof of \cref{fedavg:a2:conv:onestep}]
  By definition of the $\fedavg$ procedure (see \cref{alg:fedavg}), for all $m \in [M]$, $\x^{(r,k+1)}_m = \x^{(r,k)}_m - \eta \nabla f(\x^{(r,k)}_m; \xi^{(r,k)}_m)$. Taking average over $m = 1,\ldots, M$ gives
  \begin{equation*}
    \overline{\x^{(r,k+1)}} - \x^{\star} = \overline{\x^{(r,k)}} - \eta \cdot \frac{1}{M} \sum_{m=1}^M \nabla f(\x^{(r,k)}_m; \xi^{(r,k)}_m) - \x^{\star}.
  \end{equation*}
  Taking conditional expectation, by bounded variance \cref{asm:fo:scvx:2o}(c),
  \begin{equation}
    \expt \left[ \|\overline{\x^{(r,k+1)}} - \x^{\star}\|_2^2   |\mathcal{F}^{(r,k)} \right]
    =
    \left\| \overline{\x^{(r,k)}} - \eta \cdot \frac{1}{M} \sum_{m=1}^M \nabla F(\x^{(r,k)}_m) - \x^{\star}  \right\|_2^2+ \frac{1}{M} \eta^2 \sigma^2.
    \label{eq:fedavg:a2:conv:onestep:0}
  \end{equation}
  Now we analyze the $\left\|\overline{\x^{(r,k)}} - \eta \cdot \frac{1}{M} \sum_{m=1}^M \nabla F(\x^{(r,k)}_m) - \x^{\star}  \right\|_2^2 $ term as follows
  \begin{small}
  \begin{align}
    & \left\| \overline{\x^{(r,k)}} - \eta \cdot \frac{1}{M} \sum_{m=1}^M \nabla F(\x^{(r,k)}_m) - \x^{\star}  \right\|_2^2
    \nonumber \\
    =    &
    \left\| \overline{\x^{(r,k)}} - \eta \cdot \nabla F (\overline{\x^{(r,k)}}) - \x^{\star} + \eta \left( \nabla F (\overline{\x^{(r,k)}}) -   \frac{1}{M} \sum_{m=1}^M \nabla F(\x^{(r,k)}_m)  \right) \right\|_2^2
    \nonumber \\
    \leq & \left( 1 + \frac{1}{2} \eta \mu  \right) \left\| \overline{\x^{(r,k)}} - \eta \nabla F (\overline{\x^{(r,k)}}) - \x^{\star} \right\|_2^2 
    + \eta^2 \left( 1 + \frac{2}{\eta \mu} \right) \left\|  \nabla F (\overline{\x^{(r,k)}}) -   \frac{1}{M} \sum_{m=1}^M \nabla F(\x^{(r,k)}_m)   \right\|_2^2
    \tag{apply \cref{helper:unbalanced:ineq} with $a = \frac{1}{2}\eta \mu$}
    \\
    \leq & \left( 1 + \frac{1}{2} \eta \mu  \right) \left\| \overline{\x^{(r,k)}} - \eta \nabla F (\overline{\x^{(r,k)}}) - \x^{\star} \right\|_2^2 
    + \eta^2 \left( 1 + \frac{2}{\eta \mu} \right) \frac{Q^2}{4M} \sum_{m=1}^M \|\overline{\x^{(r,k)}} - \x^{(r,k)}_m\|_2^4 
    \tag{by \cref{helper:3rd:Lip}}
    \\
    \leq & \left( 1 + \frac{1}{2} \eta \mu  \right) \left\| \overline{\x^{(r,k)}} - \eta \nabla F (\overline{\x^{(r,k)}}) - \x^{\star} \right\|_2^2 
    + \frac{\eta Q^2}{\mu M} \sum_{m=1}^M \|\overline{\x^{(r,k)}} - \x^{(r,k)}_m\|_2^4.
    \label{eq:fedavg:a2:conv:onestep:1}
  \end{align}
  \end{small}
  where the last inequality is due to $1 + \frac{2}{\eta \mu} \leq \frac{4}{\eta\mu}$ since $\eta \mu \leq \eta L \leq \frac{1}{4}$.

  The first term of the RHS of \cref{eq:fedavg:a2:conv:onestep:1} is bounded as
  \begin{align}
         & \left\| \overline{\x^{(r,k)}} - \eta \nabla F (\overline{\x^{(r,k)}}) - \x^{\star} \right\|_2^2 
    \nonumber \\
    =    & \left\|  \overline{\x^{(r,k)}} - \x^{\star} \right\|_2^2- 2 \eta \left\langle \nabla F (\overline{\x^{(r,k)}}),   \overline{\x^{(r,k)}} - \x^{\star} \right\rangle + \eta^2 \|\nabla F(\overline{\x^{(r,k)}})\|_2^2 
    \tag{expansion of squared norm}
    \\
    \leq & \left\|  \overline{\x^{(r,k)}} - \x^{\star} \right\|_2^2- \eta \left( \mu \|\overline{\x^{(r,k)}} - \x^{\star} \|_2^2  - 2 (F(\overline{\x^{(r,k)}} ) - F^{\star}) \right) + \eta^2 \cdot (2 L (F(\overline{\x^{(r,k)}} ) - F^{\star}))
    \tag{$\mu$-strongly convexity and $L$-smoothness by \cref{asm:fo:scvx:2o}}
    \\
    =    &
    (1 - \eta \mu )  \|\overline{\x^{(r,k)}} - \x^{\star} \|_2^2  - 2\eta (1 - \eta L) (F(\overline{\x^{(r,k)}} ) - F^{\star})
    \nonumber \\
    \leq & (1 - \eta \mu )  \|\overline{\x^{(r,k)}} - \x^{\star} \|_2^2  - \eta (F(\overline{\x^{(r,k)}} ) - F^{\star}).
    \tag{since $\eta \leq \frac{1}{2L}$}
  \end{align}
  Multiplying $(1 + \frac{1}{2} \eta \mu)$ on both sides gives (note that $(1 + \frac{1}{2} \eta \mu)(1 - \eta \mu) \leq (1 - \frac{1}{2} \eta \mu)$)
  \begin{align}
    & \left( 1 + \frac{1}{2} \eta \mu  \right) \left\| \overline{\x^{(r,k)}} - \eta \nabla F (\overline{\x^{(r,k)}}) - \x^{\star} \right\|_2^2 
    \nonumber \\
    \leq &
    \left( 1 + \frac{1}{2} \eta \mu  \right)\left( 1 - \eta \mu \right)  \|\overline{\x^{(r,k)}} - \x^{\star} \|_2^2  -  \eta \left( 1 + \frac{1}{2} \eta \mu  \right) \left( F (\overline{\x^{(r,k)}}) - F^{\star} \right)
    \nonumber \\
    \leq & \left( 1 - \frac{1}{2} \eta \mu \right) \|\overline{\x^{(r,k)}} - \x^{\star} \|_2^2  -  \eta  \left( F (\overline{\x^{(r,k)}}) - F^{\star} \right).
    \label{eq:fedavg:a2:conv:onestep:2}
  \end{align}
  Combining \cref{eq:fedavg:a2:conv:onestep:0,eq:fedavg:a2:conv:onestep:1,eq:fedavg:a2:conv:onestep:2}
  completes the proof of \cref{fedavg:a2:conv:onestep}.
\end{proof}
With \cref{fedavg:a2:conv:onestep} at hand we are ready to prove \cref{fedavg:a2:conv:main}. The telescoping techniques applied here are similar to \cite{Stich-arXiv19}.
\begin{proof}[Proof of \cref{fedavg:a2:conv:main}]
  Let $S := \sum_{t=0}^{KR-1} (1 - \frac{1}{2} \eta \mu)^{KR-t-1}$. 
  Telescoping \cref{fedavg:a2:conv:onestep} yields
  \begin{align*}
         & \expt \left[\|\overline{\x^{(R,0)}}-\x^{\star}\|_2^2  \right]
    + \eta \sum_{r=0}^{R-1} \sum_{k=0}^{K-1} \left( 1 - \frac{1}{2} \eta \mu \right)^{KR-(rK+k)-1}\left( \expt[F (\overline{\x^{(r,k)}})] - F^{\star} \right)
     \\
    \leq &
    \left( 1 - \frac{1}{2} \eta \mu \right)^{KR} \| \overline{\x^{(0,0)}} - \x^{\star}\|_2^2 
    \\
    & \qquad + \sum_{r=0}^{R-1} \sum_{k=0}^{K-1} \left(1 - \frac{1}{2} \eta \mu \right)^{KR-(rK+k)-1}
    \left(  \frac{1}{M} \eta^2 \sigma^2 + \frac{\eta Q^2}{\mu M} \sum_{m=1}^M \expt \left[ \|\overline{\x^{(r,k)}} - \x^{(r,k)}_m\|_2^4 \right]  \right)
    \\
    \leq &
    \left( 1 - \frac{1}{2} \eta \mu \right)^{KR} \| \overline{\x^{(0,0)}} - \x^{\star}\|_2^2 
    + S \left(  \frac{1}{M} \eta^2 \sigma^2 + \frac{\eta Q^2}{\mu} \max_{\substack{0 \leq r < R \\ 0 \leq k < K}} \frac{1}{M} \sum_{m=1}^M \expt \left[\|\overline{\x^{(r,k)}} - \x^{(r,k)}_m\|_2^4 \right] \right).
  \end{align*}
  Multiplying $\frac{1}{\eta S}$ on both sides and rearranging,
  \begin{align}
         & \sum_{r=0}^{R-1} \sum_{k=0}^{K-1} \rho^{(r,k)} (\expt[F(\overline{\x^{(r,k)}})] - F^{\star})
    + \frac{1}{\eta S} \expt [\|\overline{\x^{(R,0)}} - \x^{\star}\|_2^2 ]
    \nonumber \\
    \leq &
    \frac{(1 - \frac{1}{2}\eta \mu)^{KR}}{\eta S} \|\overline{\x^{(0,0)}} - \x^{\star}\|_2^2 
    +
    \frac{1}{M} \eta \sigma^2 + \frac{Q^2}{\mu} \left( \max_{\substack{0 \leq r < R \\ 0 \leq k < K}}  \frac{1}{M} \sum_{m=1}^M \expt \left[\|\overline{\x^{(r,k)}} - \x^{(r,k)}_m\|_2^4 \right] \right).
    \label{eq:fedavg:a2:conv:main:1}
  \end{align}
  By definition of $S$, we have
  \begin{equation}
    \frac{1}{\eta S} = \frac{\mu}{2\left( 1 - (1 - \frac{1}{2} \eta \mu)^{KR} \right)} \geq \frac{\mu}{2},
    \label{eq:fedavg:a2:conv:main:2}
  \end{equation}
  and
  \begin{equation}
    \frac{(1 - \frac{1}{2}\eta \mu)^{KR}}{\eta S} = \frac{\mu (1 - \frac{1}{2} \eta \mu)^{KR}}{2 \left( 1 - (1 - \frac{1}{2} \eta \mu)^{KR} \right)}
    \leq \frac{\mu(1 - \frac{1}{2} \eta \mu)^{KR}}{\eta \mu}
    \leq \frac{1}{\eta} \exp \left( - \frac{1}{2} \eta \mu K R \right).
    \label{eq:fedavg:a2:conv:main:3}
  \end{equation}
  Also by convexity
  \begin{equation}
    \sum_{r=0}^{R-1} \sum_{k=0}^{K-1} \rho^{(r,k)} (\expt[F(\overline{\x^{(r,k)}})] - F^{\star}) \geq
    \expt \left[ F \left(  \sum_{r=0}^{R-1} \sum_{k=0}^{K-1} \rho^{(r,k)} \overline{\x^{(r,k)}} \right) \right] - F^{\star}.
    \label{eq:fedavg:a2:conv:main:4}
  \end{equation}
  Plugging \cref{eq:fedavg:a2:conv:main:2,eq:fedavg:a2:conv:main:3,eq:fedavg:a2:conv:main:4} to \cref{eq:fedavg:a2:conv:main:1} gives
  \begin{align*}
         & \expt \left[ F \left( \sum_{r=0}^{R-1} \sum_{k=0}^{K-1} \rho^{(r,k)} \overline{\x^{(r,k)}} \right)  \right] - F^{\star}
    + \frac{\mu}{2} \expt [\|\overline{\x^{(R,0)}} - \x^{\star}\|_2^2 ]
    \\
    \leq & \frac{1}{\eta} \exp \left( - \frac{1}{2} \eta \mu K R \right)\|\overline{\x^{(0,0)}} - \x^{\star}\|_2^2 
    +
    \frac{1}{M} \eta \sigma^2 + \frac{Q^2}{\mu} \left( \max_{\substack{0 \leq r < R \\ 0 \leq k < K}}  \frac{1}{M} \sum_{m=1}^M \expt \left[\|\overline{\x^{(r,k)}} - \x^{(r,k)}_m\|_2^4 \right] \right).
  \end{align*}
\end{proof}
% !TEX root = ../main.tex
\section{Analysis of \fedac for general convex objectives}
\label{sec:gcvx}
\subsection{Main theorems}
In this section we study the convergence of \fedac for general convex ($\mu = 0$) objectives. Let $F$ be a general convex function, the main idea is to apply \fedac to the $\ell_2$-augmented $\tilde{F}_{\lambda}(\x)$ defined as
\begin{equation}
    \tilde{F}_{\lambda}(\x) := F(\x) + \frac{1}{2} \lambda \|\x - \x^{(0,0)}\|_2^2 ,
    \label{eq:aug}
\end{equation}
where $\x^{(0,0)}$ is the initial guess. Let $\x_{\lambda}^{\star}$ be the optimum of $\tilde{F}_{\lambda}(\x)$ and define $\tilde{F}_{\lambda}^{\star} := \tilde{F}_{\lambda}(\x_{\lambda}^{\star})$.

One can verify that if $F$ satisfies \cref{asm:fo:scvx:2o} with general convexity ($\mu = 0$) and $L$-smoothness, then $\tilde{F}_{\lambda}$ satisfies \cref{asm:fo:scvx:2o} with smoothness $L+\lambda$ and strong-convexity $\lambda$ (variance does not change). If $F$ satisfies \cref{asm:fo:scvx:3o}, then $\tilde{F}_{\lambda}$ also satisfies \cref{asm:fo:scvx:3o} with the same $Q$-\nth{3}-order-smoothness (\nth{4}-order central moment does not change).
 
Now we state the convergence theorems. Recall $B := \|\x^{(0,0)} - \x^{\star}\|_2$.
\begin{theorem}[Convergence of \fedaci for general convex objective, under \cref{asm:fo:scvx:2o}]
    \label{fedaci:a1:gcvx}
    Assume \cref{asm:fo:scvx:2o} where $F$ is general convex. Then for any $T \geq 24$,\footnote{We assume this constant lower bound for technical simplification.} applying \fedaci to $\tilde{F}_{\lambda}$ \eqref{eq:aug} with 
    \begin{equation*}
        \lambda
        =
        \max \left\{\frac{\sigma}{M^{\frac{1}{2}} K^{\frac{1}{2}} R^{\frac{1}{2}} B},
        \frac{L^{\frac{1}{3}} \sigma^{\frac{2}{3}}}{K^{\frac{1}{3}} R B^{\frac{2}{3}}},  
        \frac{2L}{K R^2} \log^2 \left( \euler^2 + K R^2 \right) \right\},
    \end{equation*}
    and hyperparameter
    \begin{small}
    \begin{equation*}
        \eta = \min \left\{ \frac{1}{L + \lambda},
        \frac{1}{\lambda K R^2} \log^2  \left( \euler + \min \left\{ \frac{\lambda L M K R B^2}{\sigma^2}, 
        \frac{\lambda^2 K R^3 B^2}{\sigma^2} \right\}\right),
        \frac{L^{\frac{1}{3}} B^{\frac{2}{3}}}{\lambda^{\frac{2}{3}} K^{\frac{2}{3}} R \sigma^{\frac{2}{3}}},
        \frac{L^{\frac{1}{4}} B^{\frac{1}{2}}}{\lambda^{\frac{3}{4}} K^{\frac{3}{4}} R \sigma^{\frac{1}{2}}}
        \right\}
    \end{equation*}
    \end{small}
    yields
    \begin{align*}
        \expt \left[F(\overline{\x^{(R,0)}_{\mathrm{ag}}}) - F^{\star} \right] 
       \leq &
       \frac{2L B^2}{K R^2} \log^{2} \left( \euler^2 + K R^2 \right)
       +
       \frac{2\sigma B}{M^{\frac{1}{2}} K^{\frac{1}{2}} R^{\frac{1}{2}}}
       \log^2 \left( \euler^2 + \frac{L M^{\frac{1}{2}} K^{\frac{1}{2}} R^{\frac{1}{2}} B}{\sigma} \right)
       \\
       & \qquad
       +
       \frac{1005 L^{\frac{1}{3}} \sigma^{\frac{2}{3}} B^{\frac{4}{3}}}{K^{\frac{1}{3}} R}
       \log^4 \left( \euler^4 +  \frac{L^{\frac{2}{3}} K^{\frac{1}{3}} R B^{\frac{2}{3}} }{ \sigma^{\frac{2}{3}}} \right).
   \end{align*}
\end{theorem}
The proof of \cref{fedaci:a1:gcvx} is deferred to \cref{sec:fedaci:a1:gcvx}.

\begin{theorem}[Convergence of \fedacii for general convex objective, under \cref{asm:fo:scvx:2o}]
    \label{fedacii:a1:gcvx}
    Assume \cref{asm:fo:scvx:3o} where $F$ is general convex. Then for any $T \geq 10^3$, applying \fedacii to $\tilde{F}_{\lambda}$ \eqref{eq:aug} with 
    \begin{equation*}
        \lambda =
        \max \left\{ \frac{\sigma}{M^{\frac{1}{2}} K^{\frac{1}{2}} R^{\frac{1}{2}} B}, 
        \frac{L^{\frac{1}{2}} \sigma^{\frac{1}{2}} }{K^{\frac{1}{4}} R B^{\frac{1}{2}} },
        \frac{18 L}{K R^2} \log^2 \left( \euler^2 + K R^2 \right)         \right\},
    \end{equation*}
    and hyperparameter
    \begin{equation*}
        \eta = 
        \min \left\{ 
        \frac{1}{L+\lambda}
            ,
        \frac{9}{\lambda K R^2} 
        \log^2 \left( \euler + 
        \min \left\{ \frac{\lambda LM K R B^2}{\sigma^2}, \frac{\lambda^3 K R ^4 B^2}{L \sigma^2} \right\} \right),
        \frac{L^{\frac{1}{3}} B^{\frac{2}{3}} }{\lambda^{\frac{2}{3}} K^{\frac{2}{3}} R^{\frac{2}{3}}  \sigma^{\frac{2}{3}}}
        \right\}
    \end{equation*}
    yields
    \begin{align*}
        \expt \left[F(\overline{\x^{(R,0)}_{\mathrm{ag}}}) - F^{\star} \right] 
        \leq &
        \frac{10L B^2}{K R^2}  \log^2 \left( \euler^2 + K R^2 \right)
        +
        \frac{5 \sigma B}{M^{\frac{1}{2}} K^{\frac{1}{2}} R^{\frac{1}{2}}} \log \left( \euler + \frac{L M^{\frac{1}{2}} K^{\frac{1}{2}} R^{\frac{1}{2}} B}{\sigma} \right)
        \\
        & \qquad +
        \frac{16411 L^{\frac{1}{2}} \sigma^{\frac{1}{2}}  B^{\frac{3}{2}}}{K^{\frac{1}{4}} R}
        \log^4 \left( \euler^4 + 
        \frac{L^{\frac{1}{2}} K^{\frac{1}{4}} R B^{\frac{1}{2}}}{ \sigma^{\frac{1}{2}}} \right).
    \end{align*}
\end{theorem}
The proof of \cref{fedacii:a1:gcvx} is deferred to \cref{sec:fedacii:a1:gcvx}.

\begin{theorem}[Convergence of \fedacii for general convex objective, under \cref{asm:fo:scvx:3o}]
    \label{fedacii:a2:gcvx}
    Assume \cref{asm:fo:scvx:3o} where $F$ is general convex. Then for any $T \geq 10^3$, applying \fedacii to $\tilde{F}_{\lambda}$ \eqref{eq:aug} with 
    \begin{equation*}
        \lambda =
        \max \left\{
        \frac{\sigma}{M^{\frac{1}{2}} K^{\frac{1}{2}} R^{\frac{1}{2}} B},
        \frac{L^{\frac{1}{3}} \sigma^{\frac{2}{3}} }{M^{\frac{1}{3}} K^{\frac{1}{3}} R B^{\frac{2}{3}} },
        \frac{Q^{\frac{1}{3}} \sigma^{\frac{2}{3}}}{ K^{\frac{1}{3}} R^{\frac{4}{3}} B^{\frac{1}{3}}},
        \frac{18 L}{K R^2} \log^2 \left( \euler^2 + K R^2 \right)
        \right\},
    \end{equation*}
    and hyperparameter
    \begin{small}
    \begin{equation*}
        \eta = 
        \min \left\{ 
        \frac{1}{L+\lambda}
            ,
         \frac{9}{\lambda K R^2} 
            \log^2 \left( \euler + 
            \min \left\{ \frac{\lambda LM K R B^2}{\sigma^2},
            \frac{\lambda^2 M K R^3 B^2}{\sigma^2}, 
            \frac{\lambda^5 L K^2 R^8 B^2}{Q^2 \sigma^4} \right\} \right)
        ,
        \frac{ L^{\frac{1}{3}} M^{\frac{1}{3}} B^{\frac{2}{3}}}{\lambda^{\frac{2}{3}} K^{\frac{2}{3}} R \sigma^{\frac{2}{3}}}
        \right\}
    \end{equation*}
    \end{small}
    yields
    \begin{align*}
        & \expt \left[F(\overline{\x^{(R,0)}_{\mathrm{ag}}}) - F^{\star} \right] 
        \leq 
        \frac{10L B^2}{K R^2}  \log^2 \left( \euler^2 + K R^2 \right)
        +
        \frac{5 \sigma B}{M^{\frac{1}{2}} K^{\frac{1}{2}} R^{\frac{1}{2}}} \log \left( \euler + \frac{L M^{\frac{1}{2}} K^{\frac{1}{2}} R^{\frac{1}{2}} B}{\sigma} \right)
        \\
        & +
        \frac{139 L^{\frac{1}{3}} \sigma^{\frac{2}{3}} B^{\frac{4}{3}}}{M^{\frac{1}{3}} K^{\frac{1}{3}} R}
        \log^3 \left( \euler^3 + 
        \frac{L^{\frac{2}{3}} M^{\frac{1}{3}} K^{\frac{1}{3}} R B^{\frac{2}{3}}}{  \sigma^{\frac{2}{3}}} \right)
        +
        \frac{\euler^{19} Q^{\frac{1}{3}} \sigma^{\frac{2}{3}} B^{\frac{5}{3}} }{K^{\frac{1}{3}} R^{\frac{4}{3}}}
        \log^8 \left( \euler^8 +  \frac{L K^{\frac{1}{3}} R^{\frac{4}{3}} B^{\frac{1}{3}}}{Q^{\frac{1}{3}}  \sigma^{\frac{2}{3}} }  \right).
    \end{align*}
\end{theorem}
The proof of \cref{fedacii:a2:gcvx} is deferred to \cref{sec:fedacii:a2:gcvx}.

\subsection{Proof of \cref{fedaci:a1:gcvx} on \fedaci for general-convex objectives under \cref{asm:fo:scvx:2o}}
\label{sec:fedaci:a1:gcvx}
We first introduce the supporting lemmas for \cref{fedaci:a1:gcvx}.
\begin{lemma}
    \label{fedaci:gcvx:1}
    Assume \cref{asm:fo:scvx:2o} where $F$ is general convex, then for any $\lambda > 0$, for any $\eta \leq \frac{1}{L+\lambda}$, applying \fedaci to $\tilde{F}_{\lambda}$ gives
     \begin{align}
        \expt \left[F(\overline{\x^{(R,0)}_{\mathrm{ag}}}) - F^{\star} \right]
        \leq &  
        \frac{1}{2}\lambda  B^2
        +
        \frac{1}{2} LB^2 \exp \left(  -  \sqrt{\eta \lambda K R^2} \right) 
        + 
        \frac{\eta^{\frac{1}{2}} \sigma^2}{2 \lambda^{\frac{1}{2}} M K^{\frac{1}{2}}}
        + 
        \frac{\eta \sigma^2}{2M}
        \nonumber \\
        &  
        + 
        \frac{390 \eta^{\frac{3}{2}} L K^{\frac{1}{2}} \sigma^2} {\lambda^{\frac{1}{2}}}
        +
        7 \eta^2 L K \sigma^2
        + 
        390 \eta^{\frac{3}{2}} \lambda^{\frac{1}{2}} K^{\frac{1}{2}} \sigma^2
        + 
        7 \eta^2 \lambda K \sigma^2.
        \label{eq:fedaci:gcvx:1}
      \end{align}
\end{lemma}
The proof of \cref{fedaci:gcvx:1} is deferred to \cref{sec:fedaci:gcvx:1}. 
Now we plug in $\eta$.
\begin{lemma}
    \label{fedaci:gcvx:2}
    Assume \cref{asm:fo:scvx:2o} where $F$ is general convex, then for any $\lambda > 0$, for
    \begin{small}
    \begin{equation*}
        \eta = \min \left\{ \frac{1}{L + \lambda},
        \frac{1}{\lambda K R^2}
        \log^2  \left( \euler + \min \left\{ \frac{\lambda L M K R B^2}{\sigma^2}, \frac{\lambda^2 K R^3 B^2}{\sigma^2} \right\}\right),
        \frac{L^{\frac{1}{3}} B^{\frac{2}{3}}}{\lambda^{\frac{2}{3}} K^{\frac{2}{3}} R \sigma^{\frac{2}{3}}},
        \frac{L^{\frac{1}{4}} B^{\frac{1}{2}}}{\lambda^{\frac{3}{4}} K^{\frac{3}{4}} R \sigma^{\frac{1}{2}}}
        \right\},
    \end{equation*}
    \end{small}
    applying \fedaci to $\tilde{F}_{\lambda}$ gives
    \begin{align}
        \expt \left[F(\overline{\x^{(R,0)}_{\mathrm{ag}}}) - F^{\star} \right] 
        \leq 
        &
        \frac{1}{2} \lambda B^2
        +
        \frac{3\sigma^2}{2\lambda M K R}  \log^2 \left( \euler^2 + \frac{\lambda LMK R B^2}{\sigma^2} \right)
        \nonumber \\
        & 
        +
        \frac{592 L \sigma^2}{\lambda^2 K R^3} \log^4 \left( \euler^4 + \frac{\lambda^2 K R^3 B^2}{\sigma^2} \right)
        \nonumber \\
        & +  
        \frac{412 L^{\frac{1}{2}} \sigma B }{\lambda^{\frac{1}{2}} K^{\frac{1}{2}} R^{\frac{3}{2}} }
        + 
        \frac{1}{2} L B^2 \exp \left(  -  \sqrt{\frac{KR^2}{1 + L/\lambda}} \right).
        \label{eq:fedaci:gcvx:2}
    \end{align}
\end{lemma}

\begin{proof}[Proof of \cref{fedaci:gcvx:2}]
    To simplify the notation, we name the terms of RHS of \cref{eq:fedaci:gcvx:1} as
    \begin{alignat*}{2}
        & \varphi_0 (\eta) := \frac{1}{2} LB^2 \exp \left(  -  \sqrt{\eta \lambda K R^2} \right),
        &&
        \\
        &
        \varphi_1 (\eta) := \frac{\eta^{\frac{1}{2}} \sigma^2}{2 \lambda^{\frac{1}{2}} M K^{\frac{1}{2}}},
        \quad
        &&
        \varphi_2(\eta) :=\frac{\eta \sigma^2}{2M},
        \\
        & \varphi_3 (\eta) := 
        \frac{390 \eta^{\frac{3}{2}} L K^{\frac{1}{2}} \sigma^2} {\lambda^{\frac{1}{2}}},
        \quad
        &&
        \varphi_4 (\eta) :=
        7 \eta^2 L K \sigma^2,
        \\
        &
        \varphi_5 (\eta) :=
        390 \eta^{\frac{3}{2}} \lambda^{\frac{1}{2}} K^{\frac{1}{2}} \sigma^2,
        \quad
        &&
        \varphi_6 (\eta) :=
        7 \eta^2 \lambda K \sigma^2.
    \end{alignat*}
Define
\begin{align*}
    & \eta_1 := \frac{1}{ \lambda K R^2}
    \log^2  \left( \euler^2 + \min \left\{ \frac{\lambda L M K R B^2}{\sigma^2}, \frac{\lambda^2 K R^3 B^2}{\sigma^2} \right\}\right),
    \\
    \quad
    & \eta_2 := \frac{ L^{\frac{1}{3}} B^{\frac{2}{3}}}{\lambda^{\frac{2}{3}} K^{\frac{2}{3}} R \sigma^{\frac{2}{3}}},
    \quad
    \eta_3 := \frac{ L^{\frac{1}{4}} B^{\frac{1}{2}}}{\lambda^{\frac{3}{4}} K^{\frac{3}{4}} R \sigma^{\frac{1}{2}}}.
\end{align*}
then $\eta = \min \left\{ \eta_1, \eta_2, \eta_3, \frac{1}{L + \lambda}\right\}$. 
Now we bound $\varphi_1(\eta), \ldots, \varphi_6(\eta)$ term by term.
\begin{align*}
    \varphi_1(\eta) 
    & \leq
    \varphi_1(\eta_1)
    \leq
    \frac{\sigma^2}{2\lambda M K R} \log \left( \euler + \frac{\lambda LMK R B^2}{\sigma^2} \right),
    \\
    \varphi_2(\eta) 
    & \leq
    \varphi_2(\eta_1) 
    \leq
    \frac{\sigma^2}{2 \lambda M K R^2} \log^2 \left( \euler + \frac{\lambda LMK R B^2}{\sigma^2}  \right)
    \leq
    \frac{\sigma^2}{2\lambda M K R} \log^2 \left( \euler + \frac{\lambda LMK R B^2}{\sigma^2} \right),
    \\
    \varphi_3(\eta) 
    & \leq
    \varphi_3(\eta_1)
    \leq
    \frac{390L \sigma^2}{\lambda^2 K R^3} \log^3 \left( \euler + \frac{\lambda^2 K R^3 B^2}{\sigma^2} \right),
    \\
    \varphi_4(\eta) 
    & \leq
    \varphi_4(\eta_1)
    \leq
    \frac{7 L\sigma^2}{\lambda^2 K R^4}
    \log^4 \left( \euler + \frac{\lambda^2 K R^3 B^2}{\sigma^2} \right)
    \leq
    \frac{7L \sigma^2}{\lambda^2 K R^3} \log^4 \left( \euler + \frac{\lambda^2 K R^3 B^2}{\sigma^2} \right),
    \\
    \varphi_5(\eta) 
    & \leq
    \varphi_5(\eta_2) 
    =
    \frac{390 L^{\frac{1}{2}} B \sigma}{\lambda^{\frac{1}{2}} K^{\frac{1}{2}} R^{\frac{3}{2}}},
    \\
    \varphi_6(\eta) 
    & \leq
    \varphi_6(\eta_3)
    \leq
    7 \eta_3^2 \lambda K \sigma^2 = \frac{7 L^{\frac{1}{2}} B \sigma}{\lambda^{\frac{1}{2}} K^{\frac{1}{2}} R^2 }
    \leq
    \frac{7 L^{\frac{1}{2}} B \sigma}{\lambda^{\frac{1}{2}} K^{\frac{1}{2}} R^{\frac{3}{2}}}.
\end{align*}
In summary
\begin{equation}
    \sum_{i=1}^6 \varphi_i(\eta) \leq \frac{\sigma^2}{\lambda M K R}\log^2 \left( \euler^2 + \frac{\lambda LMK R B^2}{\sigma^2} \right)
    +
    \frac{397L \sigma^2}{\lambda^2 K R^3} \log^4 \left( \euler^4 + \frac{\lambda^2 K R^3 B^2}{\sigma^2} \right)
    +
    \frac{397 L^{\frac{1}{2}} B \sigma}{\lambda^{\frac{1}{2}} K^{\frac{1}{2}} R^{\frac{3}{2}}}.
    \label{eq:fedaci:gcvx:2:1}
\end{equation}
On the other hand $\varphi_0(\eta) \leq \varphi_0(\eta_1) + \varphi_0(\eta_2) + \varphi_0(\eta_3) + \varphi_0(\frac{1}{L + \lambda})$, where
\begin{align*}
    \varphi_0(\eta_1) 
    & =
    \frac{1}{2} L B^2  \left( \euler^2 + \min \left\{ \frac{\lambda L M K R B^2}{\sigma^2}, \frac{\lambda^2 K R^3 B^2}{\sigma^2} \right\}\right)^{-1}
     \leq  \frac{\sigma^2}{2\lambda M K R} + \frac{195 L \sigma^2}{\lambda^2 K R^3},
    \\
    \varphi_0(\eta_2) 
    & \leq 
    \frac{3!}{2}  L B^2 \left( \sqrt{\eta_2 \lambda K R^2} \right)^{-3}
    =
     \frac{3 L B^2}{\eta_2^{\frac{3}{2}} \lambda^{\frac{3}{2}} K^{\frac{3}{2}} R^3}
     =
     \frac{3 L^{\frac{1}{2}} B \sigma }{\lambda^{\frac{1}{2}} K^{\frac{1}{2}} R^{\frac{3}{2}} },
    \\
    \varphi_0(\eta_3)
    & \leq
    \frac{4!}{2}  L B^2 \left( \sqrt{\eta_3 \lambda K R^2} \right)^{-4}
    =
    \frac{12 L B^2}{\eta_3^{2} \lambda^{2} K^2 R^4}
    =
    \frac{12 L^{\frac{1}{2}} \sigma B}{\lambda^{\frac{1}{2}} K^{\frac{1}{2}} R^{2}}
    \leq
    \frac{12 L^{\frac{1}{2}} B \sigma }{\lambda^{\frac{1}{2}} K^{\frac{1}{2}} R^{\frac{3}{2}} }.
\end{align*}
In summary
\begin{equation}
    \varphi_0(\eta)
    \leq
    \frac{1}{2} L B^2 \exp \left(  -  \sqrt{\frac{\lambda K R^2}{(L + \lambda)}} \right)
    +
    \frac{\sigma^2}{2 \lambda M K R} 
    + \frac{195 L \sigma^2}{\lambda^2 K R^3} 
    + 
    \frac{15 L^{\frac{1}{2}} B \sigma}{\lambda^{\frac{1}{2}} K^{\frac{1}{2}} R^{\frac{3}{2}} }.
    \label{eq:fedaci:gcvx:2:2}
\end{equation}
Combining \cref{fedaci:gcvx:1,eq:fedaci:gcvx:2:1,eq:fedaci:gcvx:2:2} gives
\begin{align*}
    & \expt \left[F(\overline{\x^{(R,0)}_{\mathrm{ag}}}) - F^{\star} \right] 
    \leq \sum_{i=0}^6 \varphi_i(\eta) + \frac{1}{2} \lambda B^2
    \\
    \leq &
    \frac{1}{2} \lambda B^2
    +
    \frac{3\sigma^2}{2\lambda M K R}  \log^2 \left( \euler^2 + \frac{\lambda LMK R B^2}{\sigma^2} \right)
    +
    \frac{592 L \sigma^2}{\lambda^2 K R^3} \log^4 \left( \euler^4 + \frac{\lambda^2 K R^3 B^2}{\sigma^2} \right)
    \\
    & +  
    \frac{412 L^{\frac{1}{2}} \sigma B }{\lambda^{\frac{1}{2}} K^{\frac{1}{2}} R^{\frac{3}{2}} }
    + 
    \frac{1}{2} L B^2 \exp \left(  -  \sqrt{\frac{KR^2}{1 + L/\lambda}} \right).
\end{align*}
\end{proof}
The main \cref{fedaci:a1:gcvx} then follows by plugging in the appropriate $\eta$.
\begin{proof}[Proof of \cref{fedaci:a1:gcvx}]
To simplify the notation, we name the terms on the RHS of \cref{eq:fedaci:gcvx:2} as
\begin{alignat*}{2}
    & 
    \psi_0(\lambda) := \frac{1}{2} \lambda B^2, 
    \quad
    &&
    \psi_1(\lambda) := \frac{3\sigma^2}{2\lambda M K R}  \log^2 \left( \euler^2 + \frac{\lambda LMK R B^2}{\sigma^2} \right),
    \\
    &
    \psi_2(\lambda) :=  \frac{592 L \sigma^2}{\lambda^2 K R^3} \log^4 \left( \euler^4 + \frac{\lambda^2 K R^3 B^2}{\sigma^2} \right),
    \quad
    &&
    \psi_3(\lambda) := \frac{412 L^{\frac{1}{2}} B \sigma}{\lambda^{\frac{1}{2}} K^{\frac{1}{2}} R^{\frac{3}{2}} },
    \\
    &
    \psi_4(\lambda) := \frac{1}{2} L B^2 \exp \left(  -  \sqrt{\frac{KR^2}{1 + L/\lambda}} \right).
    &&
\end{alignat*}
Let
\begin{equation*}
    \lambda_1 := \frac{\sigma}{M^{\frac{1}{2}} K^{\frac{1}{2}} R^{\frac{1}{2}} B},
    \quad
    \lambda_2 := \frac{L^{\frac{1}{3}} \sigma^{\frac{2}{3}}}{K^{\frac{1}{3}} R B^{\frac{2}{3}}},
    \quad
    \lambda_3 := \frac{2L}{K R^2} \log^2 \left( \euler^2 + K R^2 \right),
\end{equation*}
then
\(
    \lambda := \max \left\{ \lambda_1, \lambda_2, \lambda_3 \right\}.
\)
By helper \cref{helper:inv:times:log}, $\psi_1$ and $\psi_2$ are monotonically decreasing w.r.t $\lambda$ for $\lambda > 0$. $\psi_3$ is trivially decreasing. Thus
\begin{align}
    \psi_1(\lambda)
    & \leq
    \psi_1(\lambda_1) 
    \leq
    \frac{3\sigma B}{2M^{\frac{1}{2}} K^{\frac{1}{2}} R^{\frac{1}{2}}}
    \log^2 \left( \euler^2 + \frac{L M^{\frac{1}{2}} K^{\frac{1}{2}} R^{\frac{1}{2}} B}{\sigma} \right),
    \label{eq:fedaci:a1:gcvx:proof:1}
    \\
    \psi_2(\lambda)
    & \leq 
    \psi_2(\lambda_2)
    \leq
    \frac{592 L^{\frac{1}{3}} \sigma^{\frac{2}{3}} B^{\frac{4}{3}}}{K^{\frac{1}{3}} R}
    \log^4 \left( \euler^4 +  \frac{L^{\frac{2}{3}} K^{\frac{1}{3}} R B^{\frac{2}{3}} }{ \sigma^{\frac{2}{3}}} \right),
    \label{eq:fedaci:a1:gcvx:proof:2}
    \\
    \psi_3(\lambda)
    & \leq 
    \psi_3(\lambda_2)
    =
    \frac{412 L^{\frac{1}{3}} \sigma^{\frac{2}{3}} B^{\frac{4}{3}}}{K^{\frac{1}{3}} R}.
    \label{eq:fedaci:a1:gcvx:proof:3}
\end{align}
Now we analyze $\psi_4(\lambda_3)$. Note first that
\(
        \frac{\lambda_3}{L} = \frac{2}{K R^2} \log^{2} \left( \euler^2 + K R^2 \right).
\)
By helper \cref{helper:inv:times:log}, $x^{-1} \log^{2} (\euler^2 + x)$ is monotonically decreasing over $(0, +\infty)$, thus
\begin{equation*}
    \frac{\lambda_3}{L} = \frac{2}{K R^2} \log^{2} \left( \euler^2 + K R^2 \right)
    \leq
    \frac{1}{12} \log^2 (\euler^2 + 24) < 1.
\end{equation*}
Hence
\begin{equation*}
    1 + \frac{L}{\lambda_3}
    \leq
    \frac{2L}{\lambda_3}
    =
    K R^2 \log^{-2}\left( \euler^2 + K R^2 \right).
\end{equation*}
We conclude that
\begin{align}
    \psi_4(\lambda)
    & \leq
    \psi_4(\lambda_3) 
    =
    \frac{1}{2} L B^2 \exp \left(  -  \sqrt{\frac{K R^2}{1 + L/\lambda_3}} \right)
    \leq
    \frac{1}{2} L B^2 \left( \euler^2 + K R^2 \right)^{-1}
    \leq
    \frac{LB^2}{2 K R^2}.
    \label{eq:fedaci:a1:gcvx:proof:4}
\end{align}
Finally note that
\begin{align}
    \psi_0(\lambda) & \leq \frac{1}{2} \lambda_1 B^2 + \frac{1}{2} \lambda_2 B^2 + \frac{1}{2} \lambda_3 B^2 
    = \frac{\sigma B}{2 M^{\frac{1}{2}} K^{\frac{1}{2}} R^{\frac{1}{2}} } + \frac{L^{\frac{1}{3}} \sigma^{\frac{2}{3}} B^{\frac{4}{3}} }{2 K^{\frac{1}{3}} R} + \frac{L B^2}{K R^2} \log^{2} \left( \euler^2 + K R^2 \right).
    \label{eq:fedaci:a1:gcvx:proof:5}
\end{align}
Combining \cref{fedaci:gcvx:2,eq:fedaci:a1:gcvx:proof:1,eq:fedaci:a1:gcvx:proof:2,eq:fedaci:a1:gcvx:proof:3,eq:fedaci:a1:gcvx:proof:4,eq:fedaci:a1:gcvx:proof:5} gives
\begin{align*}
     & \expt \left[F(\overline{\x^{(R,0)}_{\mathrm{ag}}}) - F^{\star} \right] 
    \leq \sum_{i=0}^4 \psi_i(\lambda) 
    \\
    \leq & 
    \frac{2L B^2}{K R^2} \log^{2} \left( \euler^2 + K R^2 \right)
    +
    \frac{2\sigma B}{M^{\frac{1}{2}} K^{\frac{1}{2}} R^{\frac{1}{2}}}
    \log^2 \left( \euler^2 + \frac{L M^{\frac{1}{2}} K^{\frac{1}{2}} R^{\frac{1}{2}} B}{\sigma} \right)
    \\
    & \qquad
    +
    \frac{1005 L^{\frac{1}{3}} \sigma^{\frac{2}{3}} B^{\frac{4}{3}}}{K^{\frac{1}{3}} R}
    \log^4 \left( \euler^4 +  \frac{L^{\frac{2}{3}} K^{\frac{1}{3}} R B^{\frac{2}{3}} }{ \sigma^{\frac{2}{3}}} \right).
\end{align*}
\end{proof}

\subsubsection{Proof of \cref{fedaci:gcvx:1}}
\label{sec:fedaci:gcvx:1}
We first introduce a supporting proposition for \cref{fedaci:gcvx:1}.
\begin{proposition}
    \label{fedaci:gcvx:0}
    Assume $F$ is general convex and $L$-smooth, and let $\Psi^{(r,k)}$ be the decentralized potential \cref{eq:fedaci:potential} for $\tilde{F}_{\lambda}$, namely 
    \begin{equation*}
        \Psi^{(r,k)} := \frac{1}{M} \sum_{m=1}^M \left( \tilde{F}_{\lambda}(\x^{(r,k)}_{\mathrm{ag}, m}) - \tilde{F}_{\lambda}^{\star}  \right)  + \frac{1}{2} \lambda \|\overline{\x^{(r,k)}} - \x_{\lambda}^{\star}\|_2^2 .
    \end{equation*}
    Then
    \begin{equation*}
        \Psi^{(R,0)} \geq F(\overline{\x^{(R,0)}_{\mathrm{ag}}}) - F^{\star} - \frac{1}{2} \lambda B^2
        ,
        \qquad
        \Psi^{(0,0)} \leq \frac{1}{2} L \|\x^{(0,0)} - \x^{\star}\|_2^2 .
    \end{equation*}
\end{proposition}

\begin{proof}[Proof of \cref{fedaci:gcvx:0}]
    Since $\x_{\lambda}^{\star}$ optimizes $\tilde{F}_{\lambda}(\x)$ we have $\tilde{F}_{\lambda}(\x_{\lambda}^{\star}) \leq \tilde{F}_{\lambda}(\x^{\star})$ (recall $\x^{\star}$ is defined as the optimum of the un-augmented objective $F$), and thus
    \begin{equation}
        \tilde{F}_{\lambda}^{\star} = F(\x_{\lambda}^{\star}) + \frac{1}{2}\lambda \|\x_{\lambda}^{\star} - \x^{(0,0)}\|_2^2 
        \leq
        F(\x^{\star}) + \frac{1}{2} \lambda \|\x^{\star} - \x^{(0,0)}\|_2^2 .
        \label{eq:fedaci:gcvx:0:1}
    \end{equation}
    Consequently, $\Psi^{(R,0)}$ is lower bounded as
    \begin{align}
        \Psi^{(R,0)} & = \frac{1}{M} \sum_{m=1}^M \left( \tilde{F}_{\lambda}(\x^{(R,0)}_{\mathrm{ag},m}) - \tilde{F}_{\lambda}^{\star}  \right)
        + \frac{1}{2} \lambda \|\overline{\x^{(R,0)}} - \x_{\lambda}^{\star}\|_2^2 
        \geq
        \frac{1}{M} \sum_{m=1}^M \left( \tilde{F}_{\lambda}(\x^{(R,0)}_{\mathrm{ag},m}) - \tilde{F}_{\lambda}^{\star}  \right)
        \nonumber \\
        & = \frac{1}{M} \sum_{m=1}^M \left[ \left( F(\x^{(R,0)}_{\mathrm{ag},m}) + \frac{1}{2}\lambda  \|\x^{(R,0)}_{\mathrm{ag},m} - \x^{(0,0)}\|_2^2  \right)
        - 
        \tilde{F}_{\lambda}^{\star}
         \right]
         \nonumber \\
        & \geq \frac{1}{M} \sum_{m=1}^M \left[  F(\x^{(R,0)}_{\mathrm{ag},m}) - F^{\star}  + \frac{1}{2}\lambda  \left( \|\x^{(R,0)}_{\mathrm{ag},m}  - \x^{(0,0)}\|_2^2   - \|\x^{\star} - \x^{(0,0)}\|_2^2  \right) \right]
        \tag{by \cref{eq:fedaci:gcvx:0:1}}
        \\
        & \geq \frac{1}{M} \sum_{m=1}^M \left( F(\x^{(R,0)}_{\mathrm{ag},m}) - F^{\star} \right) - \frac{1}{2}\lambda \|\x^{\star} - \x^{(0,0)}\|_2^2 
        \nonumber \\
        & \geq F(\overline{\x^{(R,0)}_{\mathrm{ag}}}) - F^{\star} - \frac{1}{2} \lambda \|\x^{\star} - \x^{(0,0)}\|_2^2 
        \tag{by convexity}
        \\
        &   = F(\overline{\x^{(R,0)}_{\mathrm{ag}}}) - F^{\star} - \frac{1}{2} \lambda B^2.
        \nonumber
    \end{align}
    The initial potential $\Psi^{(0,0)}$ is upper bounded as
    \begin{align}
        \Psi^{(0,0)} & = \tilde{F}_{\lambda}(\x^{(0,0)}) - \tilde{F}_{\lambda}^{\star} 
        + \frac{1}{2} \lambda \|\x_{\lambda}^{\star} - \x^{(0,0)}\|_2^2 
        \nonumber \\
        & =   F(\x^{(0,0)}) - \left( F(\x_{\lambda}^{\star}) + \frac{1}{2}\lambda  \|\x_{\lambda}^{\star} - \x^{(0,0)}\|_2^2  \right) 
        + \frac{1}{2} \lambda \|\x_{\lambda}^{\star} - \x^{(0,0)}\|_2^2 
        \tag{by definition of $\tilde{F}_{\lambda}$ \eqref{eq:aug}}
        \\
        & =  F(\x^{(0,0)}) - F(\x_{\lambda}^{\star}) \leq F(\x^{(0,0)}) - F^{\star} \tag{by optimality $F(\x_{\lambda}^{\star}) \geq F^{\star}$}
        \\
        & \leq \frac{1}{2} L \|\x^{(0,0)} - \x^{\star}\|_2^2  = \frac{1}{2} L B^2.
        \tag{by $L$-smoothness of $F$}
     \end{align}
\end{proof}

\cref{fedaci:gcvx:1} then follows by applying \cref{fedaci:general:eta} and \cref{fedaci:gcvx:0}.
\begin{proof}[Proof of \cref{fedaci:gcvx:1}]
By \cref{fedaci:general:eta} on the convergence of \fedaci, for any $\eta \in (0, \frac{1}{L + \lambda}$),
\begin{equation*}
    \expt \left[ \Psi^{(R,0)} \right]
    \leq
     \exp \left(  -  \sqrt{\eta \lambda K R^2} \right) \Psi^{(0,0)} 
    + \frac{\eta^{\frac{1}{2}} \sigma^2}{2 \lambda^{\frac{1}{2}} M K^{\frac{1}{2}} }
    + \frac{\eta \sigma^2}{2M}
    + \frac{390 \eta^{\frac{3}{2}} (L + \lambda) K^{\frac{1}{2}} \sigma^2} {\lambda^{\frac{1}{2}}}
    + 7 \eta^2 (L + \lambda) K \sigma^2.
\end{equation*}
Applying \cref{fedaci:gcvx:0} gives
    \begin{align*}
    \expt \left[F(\overline{\x^{(R,0)}_{\mathrm{ag}}}) - F^{\star} \right]
    \leq &  
    \frac{1}{2} LB^2 \exp \left(  -  \sqrt{\eta \lambda K R^2} \right) 
    + 
    \frac{1}{2}\lambda  B^2
    + 
    \frac{\eta^{\frac{1}{2}} \sigma^2}{2 \lambda^{\frac{1}{2}} M K^{\frac{1}{2}}}
    + 
    \frac{\eta \sigma^2}{2M}
    \\
    &  
    + 
    \frac{390 \eta^{\frac{3}{2}} L K^{\frac{1}{2}} \sigma^2} {\lambda^{\frac{1}{2}}}
    +
    7 \eta^2 L K \sigma^2
    + 
    390 \eta^{\frac{3}{2}} \lambda^{\frac{1}{2}} K^{\frac{1}{2}} \sigma^2
    + 
    7 \eta^2 \lambda K \sigma^2.
  \end{align*}
\end{proof}

\subsection{Proof of \cref{fedacii:a1:gcvx} on \fedacii for general-convex objectives under \cref{asm:fo:scvx:2o}}
\label{sec:fedacii:a1:gcvx}
We omit some technical details since the proof is similar to \cref{fedaci:a1:gcvx}. 
We first introduce the supporting lemma for \cref{fedacii:a1:gcvx}.
\begin{lemma}
    \label{fedacii:a1:gcvx:1}
    Assume \cref{asm:fo:scvx:2o} where $F$ is general convex, then for any $\lambda > 0$, for any $\eta \leq \frac{1}{L+\lambda}$, applying \fedacii to $\tilde{F}_{\lambda}$ gives
     \begin{equation}
        \expt \left[F(\overline{\x^{(R,0)}_{\mathrm{ag}}}) - F^{\star} \right]
        \leq 
        \frac{1}{2}\lambda  B^2
        +
        \frac{1}{2} LB^2 \exp \left(  -  \sqrt{\frac{\eta \lambda K R^2}{9}} \right) 
        + 
        \frac{\eta^{\frac{1}{2}} \sigma^2}{\lambda^{\frac{1}{2}} M K^{\frac{1}{2}}}
        +
        \frac{200 \eta^2 L^2 K \sigma^2}{\lambda}
        +
        200 \eta^2 \lambda K \sigma^2.
        \label{eq:fedacii:a1:gcvx:1}
      \end{equation}
\end{lemma}
The proof of \cref{fedacii:a1:gcvx:1} is deferred to \cref{sec:fedacii:a1:gcvx:1}.
\begin{lemma}
    \label{fedacii:a1:gcvx:2}
    Assume \cref{asm:fo:scvx:2o} where $F$ is general convex, then for any $\lambda > 0$, for
    \begin{equation*}
        \eta = 
        \min \left\{ 
        \frac{1}{L+\lambda}
            ,
        \frac{9}{\lambda K R^2} 
        \log^2 \left( \euler + 
        \min \left\{ \frac{\lambda LM K R B^2}{\sigma^2}, \frac{\lambda^3 K R ^4 B^2}{L \sigma^2} \right\} \right),
        \frac{L^{\frac{1}{3}} B^{\frac{2}{3}} }{\lambda^{\frac{2}{3}} K^{\frac{2}{3}} R^{\frac{2}{3}}  \sigma^{\frac{2}{3}}},
        \right\}
    \end{equation*}
    applying \fedacii to $\tilde{F}_{\lambda}$ gives
    \begin{align}
        \expt \left[F(\overline{\x^{(R,0)}_{\mathrm{ag}}}) - F^{\star} \right]
        \leq &
        \frac{1}{2}\lambda  B^2
        +
        \frac{1}{2} LB^2 \exp \left(  -  \sqrt{\frac{K R^2}{9(1 + L/\lambda)}} \right) 
        +
        \frac{209 L^{\frac{2}{3}} B^{\frac{4}{3}} \sigma^{\frac{2}{3}}}{\lambda^{\frac{1}{3}} K^{\frac{1}{3}} R^{\frac{4}{3}}}
        \nonumber \\
        & + 
        \frac{4 \sigma^2}{\lambda M K R} \log \left( \euler + 
         \frac{\lambda LM K R B^2}{\sigma^2} \right)
        + \frac{16201 L^2 \sigma^2}{\lambda^3 K R^4}
        \log^4 \left( \euler^4 + 
       \frac{\lambda^3 K R ^4 B^2}{L \sigma^2} \right).
       \label{eq:fedacii:a1:gcvx:2}
    \end{align}
\end{lemma}
\begin{proof}[Proof of \cref{fedacii:a1:gcvx:2}]
    To simplify the notation, define the terms on the RHS of \cref{eq:fedacii:a1:gcvx:1} as
    \begin{alignat*}{2}
        & \varphi_0(\eta) := 
        \frac{1}{2} LB^2 \exp \left(  -  \sqrt{\frac{\eta \lambda K R^2}{9}} \right),
        \quad
        &&
        \varphi_1(\eta) :=
        \frac{\eta^{\frac{1}{2}} \sigma^2}{\lambda^{\frac{1}{2}} M K^{\frac{1}{2}}},
        \\
        &
        \varphi_2(\eta) :=
        \frac{200 \eta^2 L^2 K \sigma^2}{\lambda},
        \quad
        &&
        \varphi_3(\eta) :=
        200 \eta^2 \lambda K \sigma^2.
    \end{alignat*}
    Define
    \begin{equation*}
        \eta_1 :=  \frac{9}{\lambda K R^2} 
        \log^2 \left( \euler + 
        \min \left\{ \frac{\lambda LM K R B^2}{\sigma^2}, \frac{\lambda^3 K R ^4 B^2}{L \sigma^2} \right\} \right),
        \qquad
        \eta_2 := 
        \frac{L^{\frac{1}{3}} B^{\frac{2}{3}} }{\lambda^{\frac{2}{3}} K^{\frac{2}{3}} R^{\frac{2}{3}}  \sigma^{\frac{2}{3}}},
    \end{equation*}
    Then $\eta = \min \left\{ \eta_1, \eta_2\right\}$.
    Since $\varphi_1, \varphi_2, \varphi_3$ are increasing we have
    \begin{align*}
        \varphi_1(\eta) \leq \varphi_1 (\eta_1)
        & \leq
        \frac{3 \sigma^2}{\lambda M K R} \log \left( \euler + 
         \frac{\lambda LM K R B^2}{\sigma^2} \right),
        \\
        \varphi_2(\eta) \leq \varphi_2 (\eta_1)
        & \leq
        \frac{16200 L^2 \sigma^2}{\lambda^3 K R^4}
        \log^4 \left( \euler + 
       \frac{\lambda^3 K R ^4 B^2}{L \sigma^2} \right),
       \\
       \varphi_3(\eta) \leq \varphi_3 (\eta_2)
       & \leq
       \frac{200 L^{\frac{2}{3}} B^{\frac{4}{3}} \sigma^{\frac{2}{3}}}{\lambda^{\frac{1}{3}} K^{\frac{1}{3}} R^{\frac{4}{3}}}.
    \end{align*}
    On the other hand, since $\varphi_0$ is decreasing we have $\varphi_0(\eta) \leq \varphi_0(\eta_1) + \varphi_0(\eta_2) + \varphi_0(\frac{1}{L + \lambda})$, where
    \begin{align*}
        \varphi_0(\eta_1) 
        & \leq
        \frac{\sigma^2}{2 \lambda M K R} + \frac{L^2 \sigma^2}{2 \lambda^3 K R^4},
        \\
        \varphi_0(\eta_2)
        & \leq
        \frac{2!}{2} LB^2 \left( \sqrt{\frac{\eta_2 \lambda K R^2}{9}}  \right)^{-2}
        =
        \frac{9L B^2}{\eta_2 \lambda K R^2}
        =
       \frac{9 L^{\frac{2}{3}} B^{\frac{4}{3}} \sigma^{\frac{2}{3}}}{\lambda^{\frac{1}{3}} K^{\frac{1}{3}} R^{\frac{4}{3}}}.
    \end{align*}
    Combining the above bounds completes the proof.
\end{proof}
\cref{fedacii:a1:gcvx} then follows by plugging in an appropriate $\lambda$.
\begin{proof}[Proof of \cref{fedacii:a1:gcvx}]
    To simplify the notation, define the terms on the RHS of \cref{eq:fedacii:a1:gcvx:2} as
    \begin{alignat*}{2}
        & \psi_0(\lambda):=
        \frac{1}{2}\lambda  B^2,
        \quad
        &&
        \psi_1(\lambda):=
        \frac{1}{2} LB^2 \exp \left(  -  \sqrt{\frac{K R^2}{9(1 + L/\lambda)}} \right),
        \\
        &
        \psi_2(\lambda):=
        \frac{209 L^{\frac{2}{3}} B^{\frac{4}{3}} \sigma^{\frac{2}{3}}}{\lambda^{\frac{1}{3}} K^{\frac{1}{3}} R^{\frac{4}{3}}},
        \quad
        && \psi_3(\lambda):=
        \frac{4 \sigma^2}{\lambda M K R} \log \left( \euler + 
         \frac{\lambda LM K R B^2}{\sigma^2} \right),
        \\
        &
        \psi_4(\lambda):=
       \frac{16201 L^2 \sigma^2}{\lambda^3 K R^4}
        \log^4 \left( \euler^4 + 
       \frac{\lambda^3 K R ^4 B^2}{L \sigma^2} \right).
    \end{alignat*}
    Define
    \begin{align*}
        \lambda_1 := \frac{\sigma}{M^{\frac{1}{2}} K^{\frac{1}{2}} R^{\frac{1}{2}} B},
        \quad
        \lambda_2 := \frac{L^{\frac{1}{2}} \sigma^{\frac{1}{2}} }{B^{\frac{1}{2}} K^{\frac{1}{4}} R},
        \quad
        \lambda_3 := \frac{18 L}{K R^2} \log^2 \left( \euler^2 + K R^2 \right).
    \end{align*}
    Then $\lambda = \max \left\{ \lambda_1, \lambda_2, \lambda_3 \right\}$.
    By helper \cref{helper:inv:times:log} $\psi_3$, $\psi_4$ are decreasing; $\psi_2$ is trivially decreasing, thus
    \begin{align*}
        \psi_2(\lambda)
        &
        \leq \psi_2(\lambda_2) = \frac{209 L^{\frac{1}{2}} B^{\frac{3}{2}} \sigma^{\frac{1}{2}}}{K^{\frac{1}{4}} R},
        \\
        \psi_3(\lambda)
        &
        \leq \psi_3(\lambda_1) 
        =
        \frac{4 \sigma B}{M^{\frac{1}{2}} K^{\frac{1}{2}} R^{\frac{1}{2}}} \log \left( \euler + \frac{L M^{\frac{1}{2}} K^{\frac{1}{2}} R^{\frac{1}{2}} B}{\sigma} \right),
        \\
        \psi_4 (\lambda)
        & \leq
        \psi_4(\lambda_2)
        =
        \frac{16201 L^{\frac{1}{2}} B^{\frac{3}{2}} \sigma^{\frac{1}{2}}}{K^{\frac{1}{4}} R}
        \log^4 \left( \euler^4 + 
        \frac{L^{\frac{1}{2}} K^{\frac{1}{4}} R B^{\frac{1}{2}}}{ \sigma^{\frac{1}{2}}} \right).
    \end{align*}
    For $\psi_1(\lambda)$ since $T \geq 1000$ we have $K R^2 \geq 1000$, thus
    \begin{equation*}
        \frac{\lambda_3}{L} = \frac{18 }{K R^2} \log^2 \left( \euler^2 + K R^2 \right) 
        \leq
        \frac{18}{1000}\log^2 \left( \euler^2 + 1000 \right) < 1.
    \end{equation*}
    Thus $1 + \frac{L}{\lambda_3} \leq \frac{2L}{\lambda_3}$, and therefore
    \begin{equation*}
        \psi_1(\lambda) \leq \psi_1(\lambda_3) 
        =
        \frac{1}{2} L B^2 \left( \euler^2 + K R^2 \right)^{-1} \leq \frac{L B^2}{2 K R^2}.
    \end{equation*}
    Finally
    \begin{align*}
        \psi_0(\lambda)  \leq \sum_{i=1}^3 \psi_0(\lambda_i)
        \leq \frac{\sigma B}{2 M^{\frac{1}{2}} K^{\frac{1}{2}} R^{\frac{1}{2}}} 
        +
        \frac{L^{\frac{1}{2}} B^{\frac{3}{2}} \sigma^{\frac{1}{2}}}{2 K^{\frac{1}{4}} R}
        +
        \frac{9L B^2}{K R^2}  \log^2 \left( \euler^2 + K R^2 \right).
    \end{align*}
    Consequently,
    \begin{align*}
        \sum_{i=0}^4 \psi(\lambda)
        \leq &
        \frac{10L B^2}{K R^2}  \log^2 \left( \euler^2 + K R^2 \right)
        +
        \frac{5 \sigma B}{M^{\frac{1}{2}} K^{\frac{1}{2}} R^{\frac{1}{2}}} \log \left( \euler + \frac{L M^{\frac{1}{2}} K^{\frac{1}{2}} R^{\frac{1}{2}} B}{\sigma} \right)
        \\
        & \qquad +
        \frac{16411 L^{\frac{1}{2}} B^{\frac{3}{2}} \sigma^{\frac{1}{2}}}{K^{\frac{1}{4}} R}
        \log^4 \left( \euler^4 + 
        \frac{L^{\frac{1}{2}} K^{\frac{1}{4}} R B^{\frac{1}{2}}}{ \sigma^{\frac{1}{2}}} \right),
    \end{align*}
    completing the proof.
\end{proof}

\subsubsection{Proof of \cref{fedacii:a1:gcvx:1}}
\label{sec:fedacii:a1:gcvx:1}
\cref{fedacii:a1:gcvx:1} is parallel to \cref{fedaci:gcvx:1} where the main difference is the following supporting proposition.
\begin{proposition}
    \label{fedacii:gcvx:0}
    Assume $F$ is general convex and $L$-smooth, and let $\Phi^{(r,k)}$ be the centralized potential \cref{eq:centralied:potential} for $\tilde{F}_{\lambda}$ (with strong convexity estimate $\mu = \lambda$), namely 
    \begin{equation*}
        \Phi^{(r,k)} := \left( \tilde{F}_{\lambda}(\overline{\x^{(r,k)}_{\mathrm{ag}}}) - \tilde{F}_{\lambda}^{\star}  \right)  + \frac{1}{6} \lambda \|\overline{\x^{(R,0)}} - \x_{\lambda}^{\star}\|_2^2 .
    \end{equation*}
    Then
    \begin{equation*}
        \Phi^{(R,0)} \geq F(\overline{\x^{(R,0)}_{\mathrm{ag}}}) - F^{\star} - \frac{1}{2} \lambda B^2
        ,
        \qquad
        \Phi^{(0,0)} \leq \frac{1}{2} L \|\x^{(0,0)} - \x^{\star}\|_2^2 .
    \end{equation*}
\end{proposition}
\begin{proof}[Proof of \cref{fedacii:gcvx:0}]
    The proof is almost identical to \cref{fedaci:gcvx:0}.
\end{proof}
\begin{proof}[Proof of \cref{fedacii:a1:gcvx:1}]
    Follows by applying \cref{fedacii:a1:general:eta} and plugging in the bound of \cref{fedacii:gcvx:0}. The rest of proof is the same as \cref{fedaci:gcvx:1} which we omit the details.
\end{proof}

\subsection{Proof of \cref{fedacii:a2:gcvx} on \fedacii for general-convex objectives under \cref{asm:fo:scvx:3o}}
\label{sec:fedacii:a2:gcvx}
We omit some of the proof details since the proof is similar to \cref{fedaci:a1:gcvx}. 
We first introduce the supporting lemma for \cref{fedacii:a2:gcvx}.
\begin{lemma}
    \label{fedacii:a2:gcvx:1}
    Assume \cref{asm:fo:scvx:3o} where $F$ is general convex, then for any $\lambda > 0$, for any $\eta \leq \frac{1}{L+\lambda}$, applying \fedacii to $\tilde{F}_{\lambda}$ gives
    \begin{align}
        \expt \left[F(\overline{\x^{(R,0)}_{\mathrm{ag}}}) - F^{\star} \right]
        & \leq 
        \frac{1}{2}\lambda  B^2
        +
        \frac{1}{2} LB^2 \exp \left(  -  \sqrt{\frac{\eta \lambda K R^2}{9}} \right) 
        \nonumber \\
        & 
        + 
        \frac{\eta^{\frac{1}{2}} \sigma^2}{\lambda^{\frac{1}{2}} M K^{\frac{1}{2}}}
        +
        \frac{2 \eta^{\frac{3}{2}} L K^{\frac{1}{2}} \sigma^2}{\lambda^{\frac{1}{2}}M }
        +
        \frac{2 \eta^{\frac{3}{2}} \lambda^{\frac{1}{2}} K^{\frac{1}{2}} \sigma^2}{M}
        +
         \frac{\euler^9 \eta^4 Q^2 K^2 \sigma^4}{\lambda}.
        \label{eq:fedacii:a2:gcvx:1}
    \end{align}
\end{lemma}
\begin{proof}[Proof of \cref{fedacii:a2:gcvx:1}]
    Follows by \cref{fedacii:a2:general:eta,fedacii:gcvx:0}. The proof is similar to \cref{fedaci:gcvx:1} so we omit the details.
\end{proof}
\begin{lemma}
    \label{fedacii:a2:gcvx:2}
    Assume \cref{asm:fo:scvx:3o} where $F$ is general convex, then for any $\lambda > 0$, for
    \begin{equation*}
        \eta = 
        \min \left\{ 
        \frac{1}{L+\lambda}
            ,
         \frac{9}{\lambda K R^2} 
            \log^2 \left( \euler + 
            \min \left\{ \frac{\lambda LM K R B^2}{\sigma^2},
            \frac{\lambda^2 M K R^3 B^2}{\sigma^2}, 
            \frac{\lambda^5 L K^2 R^8 B^2}{Q^2 \sigma^4} \right\} \right)
        ,
        \frac{ L^{\frac{1}{3}} M^{\frac{1}{3}} B^{\frac{2}{3}}}{\lambda^{\frac{2}{3}} K^{\frac{2}{3}} R \sigma^{\frac{2}{3}}}
        \right\},
    \end{equation*}
    applying \fedacii to $\tilde{F}_{\lambda}$ gives
    \begin{align}
        & \expt \left[F(\overline{\x^{(R,0)}_{\mathrm{ag}}}) - F^{\star} \right]
        \leq
        \frac{1}{2}\lambda  B^2
        +
        \frac{1}{2} LB^2 \exp \left(  -  \sqrt{\frac{K R^2}{9(1 + L/\lambda)}} \right) 
        +
        \frac{4 \sigma^2}{\lambda M K R} \log \left( \euler + 
         \frac{\lambda LM K R B^2}{\sigma^2} \right)
         \nonumber \\
        & 
         + 
         \frac{55 L \sigma^2}{\lambda^2 M K R^3}
         \log^3 \left( \euler^3 + 
         \frac{\lambda^2 M K R^3 B^2}{\sigma^2} \right)
         +
         \frac{83 L^{\frac{1}{2}} B \sigma}{\lambda^{\frac{1}{2}} M^{\frac{1}{2}} K^{\frac{1}{2}} R^{\frac{3}{2}}}
         +
         \frac{\euler^{18} Q^2 \sigma^4}{\lambda^5 K^2 R^8}  \log^8 \left( \euler^8 +  \frac{\lambda^5 L K^2 R^8 B^2}{Q^2 \sigma^4} \right).
         \label{eq:fedacii:a2:gcvx:2}
    \end{align}
\end{lemma}
\begin{proof}[Proof of \cref{fedacii:a2:gcvx:2}]
    To simplify the notation, define the terms on the RHS of \cref{eq:fedacii:a2:gcvx:1} as
    \begin{alignat*}{3}
        & \varphi_0(\eta) := 
        \frac{1}{2} LB^2 \exp \left(  -  \sqrt{\frac{\eta \lambda K R^2}{9}} \right),
        \quad
        &&
        \varphi_1(\eta) :=
        \frac{\eta^{\frac{1}{2}} \sigma^2}{\lambda^{\frac{1}{2}} M K^{\frac{1}{2}}},
        \quad
        &&
        \varphi_2(\eta) :=
        \frac{2 \eta^{\frac{3}{2}} L K^{\frac{1}{2}} \sigma^2}{\lambda^{\frac{1}{2}}M },
        \\
        & 
        \varphi_3(\eta) := 
        \frac{2 \eta^{\frac{3}{2}} \lambda^{\frac{1}{2}} K^{\frac{1}{2}} \sigma^2}{M},
        \quad
        &&
        \varphi_4(\eta) :=
         \frac{\euler^9 \eta^4 Q^2 K^2 \sigma^4}{\lambda}.
        &&
    \end{alignat*}
    Define
    \begin{equation*}
        \eta_1 :=  \frac{9}{\lambda K R^2} 
        \log^2 \left( \euler + 
        \min \left\{ \frac{\lambda LM K R B^2}{\sigma^2},
        \frac{\lambda^2 M K R^3 B^2}{\sigma^2}, 
        \frac{\lambda^5 L K^2 R^8 B^2}{Q^2 \sigma^4} \right\} \right),
        \quad
        \eta_2 := 
        \frac{ L^{\frac{1}{3}} M^{\frac{1}{3}} B^{\frac{2}{3}}}{\lambda^{\frac{2}{3}} K^{\frac{2}{3}} R \sigma^{\frac{2}{3}}}.
    \end{equation*}
    Then $\eta = \min \left\{ \eta_1, \eta_2\right\}$.
    Since $\varphi_1, \ldots, \varphi_4$ are increasing we have
    \begin{align*}
        \varphi_1(\eta) \leq \varphi_1 (\eta_1)
        & \leq
        \frac{3 \sigma^2}{\lambda M K R} \log \left( \euler + 
         \frac{\lambda LM K R B^2}{\sigma^2} \right),
        \\
        \varphi_2(\eta) \leq \varphi_2 (\eta_1)
        & \leq
        \frac{54 L \sigma^2}{\lambda^2 M K R^3}
        \log^3 \left( \euler + 
        \frac{\lambda^2 M K R^3 B^2}{\sigma^2} \right),
       \\
       \varphi_3(\eta) \leq \varphi_3 (\eta_2)
       & =
       \frac{2 L^{\frac{1}{2}} B \sigma}{\lambda^{\frac{1}{2}} M^{\frac{1}{2}} K^{\frac{1}{2}} R^{\frac{3}{2}}},
       \\
       \varphi_4(\eta) \leq \varphi_4 (\eta_1)
       & \leq
       \frac{9^4 \euler^9 Q^2 \sigma^4}{\lambda^5 K^2 R^8}  \log^8 \left( \euler +  \frac{\lambda^5 L K^2 R^8 B^2}{Q^2 \sigma^4} \right).
    \end{align*}
    On the other hand $\varphi_0(\eta) \leq \varphi_0(\eta_1) + \varphi_0(\eta_2) + \varphi_0(\frac{1}{L + \lambda})$, where
    \begin{align*}
        \varphi_0(\eta_1) 
        & \leq
        \frac{\sigma^2}{2 \lambda M K R} + \frac{L \sigma^2}{2 \lambda^2 M K R^3} + \frac{Q^2 \sigma^4}{2 \lambda^5 K^2 R^8},
        \\
        \varphi_0(\eta_2)
        & \leq
        \frac{3!}{2} LB^2 \left( \sqrt{\frac{\eta_2 \lambda K R^2}{9}}  \right)^{-3}
        =
        \frac{81 L B^2}{\eta_2^{\frac{3}{2}} \lambda^{\frac{3}{2}} K^{\frac{3}{2}} R^3}
        =
        \frac{81 L^{\frac{1}{2}} B \sigma}{\lambda^{\frac{1}{2}} M^{\frac{1}{2}} K^{\frac{1}{2}} R^{\frac{3}{2}}}.
    \end{align*}
    Combining the above bounds completes the proof.
\end{proof}

\cref{fedacii:a2:gcvx} then follows by plugging in an appropriate $\lambda$.
\begin{proof}[Proof of \cref{fedacii:a2:gcvx}]
    To simplify the notation, define the terms on the RHS of \cref{eq:fedacii:a2:gcvx:2} as
    \begin{alignat*}{2}
        & \psi_0(\lambda):=
        \frac{1}{2}\lambda  B^2
        ,
        &&
        \psi_1(\lambda):=
        \frac{1}{2} LB^2 \exp \left(  -  \sqrt{\frac{K R^2}{9(1 + L/\lambda)}} \right),
        \\
        & \psi_2(\lambda):=
        \frac{4 \sigma^2}{\lambda M K R} \log \left( \euler + 
         \frac{\lambda LM K R B^2}{\sigma^2} \right),
        &&
        \psi_3(\lambda):=
        \frac{55 L \sigma^2}{\lambda^2 M K R^3}
         \log^3 \left( \euler^3 + 
         \frac{\lambda^2 M K R^3 B^2}{\sigma^2} \right),
        \\
        & \psi_4(\lambda):= \frac{83 L^{\frac{1}{2}} B \sigma}{\lambda^{\frac{1}{2}} M^{\frac{1}{2}} K^{\frac{1}{2}} R^{\frac{3}{2}}},
        &&
        \psi_5(\lambda):=
        \frac{\euler^{18} Q^2 \sigma^4}{\lambda^5 K^2 R^8}  \log^8 \left( \euler^8 +  \frac{\lambda^5 L K^2 R^8 B^2}{Q^2 \sigma^4} \right).
    \end{alignat*}
    Define
    \begin{align*}
        \lambda_1 := \frac{\sigma}{M^{\frac{1}{2}} K^{\frac{1}{2}} R^{\frac{1}{2}} B},
        \quad
        \lambda_2 := \frac{L^{\frac{1}{3}} \sigma^{\frac{2}{3}} }{M^{\frac{1}{3}} K^{\frac{1}{3}} R B^{\frac{2}{3}} },
        \quad
        \lambda_3 := \frac{Q^{\frac{1}{3}} \sigma^{\frac{2}{3}}}{B^{\frac{1}{3}} K^{\frac{1}{3}} R^{\frac{4}{3}}},
        \quad
        \lambda_4 := \frac{18 L}{K R^2} \log^2 \left( \euler^2 + K R^2 \right).
    \end{align*}
    Then $\lambda = \max \left\{ \lambda_1, \lambda_2, \lambda_3 \right\}$.
    By \cref{helper:inv:times:log}, $\psi_2$, $\psi_3$, $\psi_5$ are increasing. $\psi_4$ is trivially decreasing, thus
    \begin{align*}
        \psi_2(\lambda)
        &
        \leq \psi_2(\lambda_1) 
        =
        \frac{4 \sigma B}{M^{\frac{1}{2}} K^{\frac{1}{2}} R^{\frac{1}{2}}} \log \left( \euler + \frac{L M^{\frac{1}{2}} K^{\frac{1}{2}} R^{\frac{1}{2}} B}{\sigma} \right),
        \\
        \psi_3 (\lambda)
        & \leq
        \psi_3(\lambda_2)
        =
        \frac{55 L^{\frac{1}{3}} B^{\frac{4}{3}} \sigma^{\frac{2}{3}}}{M^{\frac{1}{3}} K^{\frac{1}{3}} R}
        \log^3 \left( \euler^3 + 
        \frac{L^{\frac{2}{3}} M^{\frac{1}{3}} K^{\frac{1}{3}} R B^{\frac{2}{3}}}{  \sigma^{\frac{2}{3}}} \right),
        \\
        \psi_4(\lambda) 
        & \leq
        \psi_4(\lambda_2)
        =
        \frac{83 L^{\frac{1}{3}} B^{\frac{4}{3}} \sigma^{\frac{2}{3}}}{M^{\frac{1}{3}} K^{\frac{1}{3}} R},
        \\
        \psi_5(\lambda)
        & \leq 
        \psi_5(\lambda_3)
        = 
        \frac{\euler^{18} Q^{\frac{1}{3}} B^{\frac{5}{3}} \sigma^{\frac{2}{3}} }{K^{\frac{1}{3}} R^{\frac{4}{3}}}
        \log^8 \left( \euler^8 +  \frac{L K^{\frac{1}{3}} R^{\frac{4}{3}} B^{\frac{1}{3}}}{Q^{\frac{1}{3}}  \sigma^{\frac{2}{3}} }  \right).
    \end{align*}    
    For $\psi_1(\lambda)$ since $T \geq 1000$ we have $K R^2 \geq 1000$, thus
    \begin{equation*}
        \frac{\lambda_3}{L} = \frac{18 }{K R^2} \log^2 \left( \euler^2 + K R^2 \right) 
        \leq
        \frac{18}{1000}\log^2 \left( \euler^2 + 1000 \right) < 1.
    \end{equation*}
    Thus $1 + \frac{L}{\lambda_3} \leq \frac{2L}{\lambda_3}$, and therefore
    \begin{equation*}
        \psi_1(\lambda) \leq \psi_1(\lambda_3) 
        =
        \frac{1}{2} L B^2 \left( \euler^2 + K R^2 \right)^{-1} \leq \frac{L B^2}{2 K R^2}.
    \end{equation*}
    Finally
    \begin{align*}
        \psi_0(\lambda)  \leq \sum_{i=1}^4 \psi_0(\lambda_i)
        \leq \frac{\sigma B}{2 M^{\frac{1}{2}} K^{\frac{1}{2}} R^{\frac{1}{2}}} 
        +
        \frac{L^{\frac{1}{3}} B^{\frac{4}{3}} \sigma^{\frac{2}{3}}}{2 M^{\frac{1}{3}} K^{\frac{1}{3}} R}
        +
        \frac{Q^{\frac{1}{3}} B^{\frac{5}{3}} \sigma^{\frac{2}{3}} }{2 K^{\frac{1}{3}} R^{\frac{4}{3}}}
        +
        \frac{9L B^2}{K R^2}  \log^2 \left( \euler^2 + K R^2 \right).
    \end{align*}
    Consequently,
    \begin{align*}
        & \sum_{i=0}^4 \psi(\lambda)
        \leq
        \frac{10L B^2}{K R^2}  \log^2 \left( \euler^2 + K R^2 \right)
        +
        \frac{5 \sigma B}{M^{\frac{1}{2}} K^{\frac{1}{2}} R^{\frac{1}{2}}} \log \left( \euler + \frac{L M^{\frac{1}{2}} K^{\frac{1}{2}} R^{\frac{1}{2}} B}{\sigma} \right)
        \\
        & +
        \frac{139 L^{\frac{1}{3}} \sigma^{\frac{2}{3}} B^{\frac{4}{3}}}{M^{\frac{1}{3}} K^{\frac{1}{3}} R}
        \log^3 \left( \euler^3 + 
        \frac{L^{\frac{2}{3}} M^{\frac{1}{3}} K^{\frac{1}{3}} R B^{\frac{2}{3}}}{  \sigma^{\frac{2}{3}}} \right)
        +
        \frac{\euler^{19} Q^{\frac{1}{3}} \sigma^{\frac{2}{3}} B^{\frac{5}{3}} }{K^{\frac{1}{3}} R^{\frac{4}{3}}}
        \log^8 \left( \euler^8 +  \frac{L K^{\frac{1}{3}} R^{\frac{4}{3}} B^{\frac{1}{3}}}{Q^{\frac{1}{3}}  \sigma^{\frac{2}{3}} }  \right).
    \end{align*}
\end{proof}
\section{Miscellaneous Helper Lemmas}
\label{sec:helper}
In this section we include some generic helper lemmas. Most of the results are standard and we provide the proof for completeness.
\begin{lemma}
    \label{helper:blocknorm}
    Let $\A = \begin{bmatrix}
            \A_{11} & \A_{12}
            \\
            \A_{21} & \A_{22}
        \end{bmatrix}$ be an arbitrary $2d \times 2d$ block matrix, where $\A_{11}, \A_{12}, \A_{21}, \A_{22}$ are $d \times d$ matrix blocks. Then the operator norm of $\A$ is bounded by
    \begin{equation*}
        \| \A\|_2 \leq \max \left\{ \|\A_{11}\|_2, \|\A_{22}\|_2 \right\} + \left\{ \|\A_{12}\|_2, \|\A_{21}\|_2 \right\}.
    \end{equation*}
\end{lemma}
\begin{proof}[Proof of \cref{helper:blocknorm}]
    Let $\A_{ij} = \U_{ij} \bSigma_{ij} \V_{ij}^{KR}$ be the SVD decomposition of matrix $\A_{ij}$, for $i=1,2$, and $j=1,2$. Then
    \begin{equation*}
        \begin{bmatrix}
            \A_{11} & \\ & \A_{22}
        \end{bmatrix}
        =
        \begin{bmatrix}
            \U_{11} \bSigma_{11} \V_{11}^\top & \\ & \U_{22} \bSigma_{22} \V_{22}^\top
        \end{bmatrix}
        =
        \begin{bmatrix}
            \U_{11} & \\ & \U_{22}
        \end{bmatrix}
        \begin{bmatrix}
            \bSigma_{11} & \\ & \bSigma_{22}
        \end{bmatrix}
        \begin{bmatrix}
            \V_{11} & \\ & \V_{22}
        \end{bmatrix}^\top,
    \end{equation*}
    thus
    \begin{equation*}
        \left\| \begin{bmatrix}
            \A_{11} & \\ & \A_{22}
        \end{bmatrix} \right\|_2
        =
        \left\| \begin{bmatrix}
            \bSigma_{11} & \\ & \bSigma_{22}
        \end{bmatrix} \right\|_2
        =
        \max\left\{ \|\bSigma_{11}\|_2, \|\bSigma_{22}\|_2\right\}
        =
        \max\left\{ \|\A_{11}\|_2, \|\A_{22}\|_2\right\}.
    \end{equation*}
    Similarly
    \begin{equation*}
        \begin{bmatrix}
             & \A_{12} \\ \A_{21} &
        \end{bmatrix}
        =
        \begin{bmatrix}
             & \U_{12} \bSigma_{12} \V_{12}^\top \\ \U_{21} \bSigma_{21} \V_{21}^\top &
        \end{bmatrix}
        =
        \begin{bmatrix}
             & \U_{12} \\ \U_{21} &
        \end{bmatrix}
        \begin{bmatrix}
            \bSigma_{21} & \\  & \bSigma_{12}
        \end{bmatrix}
        \begin{bmatrix}
            \V_{21} & \\ & \V_{12}
        \end{bmatrix}^\top,
    \end{equation*}
    thus
    \begin{equation*}
        \left\| \begin{bmatrix}
             & \A_{12} \\ \A_{21 }&
        \end{bmatrix} \right\|_2
        =
        \left\| \begin{bmatrix}
            \bSigma_{21} & \\ & \bSigma_{12}
        \end{bmatrix} \right\|_2
        =
        \max\left\{ \|\bSigma_{12}\|_2, \|\bSigma_{21}\|_2\right\}
        =
        \max\left\{ \|\A_{12}\|_2, \|\A_{21}\|_2\right\}.
    \end{equation*}
    Consequently, by the subadditivity of the operator norm,
    \begin{equation*}
        \|\A\| \leq  \left\| \begin{bmatrix}
            \A_{11} & \\ & \A_{22}
        \end{bmatrix} \right\|_2 + \left\| \begin{bmatrix}
             & \A_{12} \\ \A_{21 }&
        \end{bmatrix} \right\|_2
        \leq
        \max\left\{ \|\A_{11}\|_2, \|\A_{22}\|_2\right\} + \max\left\{ \|\A_{12}\|_2, \|\A_{21}\|_2\right\}.
    \end{equation*}
\end{proof}

\begin{lemma}
    \label{helper:unbalanced:ineq}
    Let $\x, \y \in \reals^d$, then for any $a > 0$, the following inequality holds
    \begin{equation*}
        \|\x + \y\|_2^2  \leq (1 + a) \|\x\|_2^2  + (1 + a^{-1}) \|\y\|_2^2 .
    \end{equation*}
\end{lemma}
\begin{proof}[Proof of \cref{helper:unbalanced:ineq}]
    First note that $\|\x + \y\|_2^2  = \|\x\|_2^2  + \|\y\|_2^2  + 2 \langle \x, \y \rangle$, then the proof follows by $2\langle \x, \y \rangle \leq \zeta \|\x\|_2^2  + \zeta^{-1} \|\y\|_2^2 $ due to Cauchy-Schwartz inequality.
\end{proof}

\begin{lemma}
    \label{helper:inv:times:log}
    Let $\varphi(x) := \frac{1}{x^q} \log^p (a + b x)$, where $a, p, q \geq 1$, $b > 0$ are constants. Then suppose $a \geq \exp (p/q)$, it is the case that $\varphi(x)$ is monotonically decreasing over $(0, +\infty)$.
\end{lemma}
\begin{proof}[Proof of \cref{helper:inv:times:log}]
    Without loss of generality assume $b = 1$, otherwise we put $\psi(x) = \varphi(x/b)$ then $\psi$ has the same form (up to constants) with $b = 1$. Taking derivative for $\varphi(x) = x^{-q} \log^p (a + x)$ gives
    \begin{align*}
        \varphi'(x)
        & =
        \frac{px^{-q} \log^{p -1} (a+x) }{a + x} - q x^{-q-1} \log^p (a+x)
        \\
        & =
        \frac{x^{-q-1}\log^{p-1}(a+x)}{a+ x} \left( p x - q(a + x) \log(a+x) \right).
    \end{align*}
    Since $a \geq 1$ and $x > 0$ we always have $ \frac{x^{-q-1}\log^{p-1}(a+x)}{a+ x} \geq 0$. Suppose $a \geq \exp(p/q)$ then
    \begin{equation*}
        px - q(a+x) \log (a+x)
        <
        px - qx \log (a)
        \leq
        px - qx \cdot \frac{p}{q}
        \leq
        0.
    \end{equation*}
    Hence $\varphi'(x) < 0$ and thus $\varphi(x)$ is monotonically decreasing.
\end{proof}

\chapter{Appendix of Chapter 4}
\section{Theoretical Background and Technicalities}
\label{sec:background}
In this section, we introduce some definitions and propositions that are necessary for the proof of our theoretical results. Most of the definitions and results are standard and can be found in the classic convex analysis literature (e.g., \cite{Rockafellar-70,Hiriart-Urruty.Lemarechal-01}), unless otherwise noted.

The following definition of the \emph{effective domain} extends the notion of \emph{domain} (of a finite-valued function) to an extended-valued convex function $\reals^d \to \reals \cup \{+ \infty\}$. 
\begin{definition}[Effective domain]
  Let $g: \reals^d \to \reals \cup \{+ \infty\}$ be an extended-valued convex function. The \textbf{effective domain} of $g$, denoted by $\dom g$, is defined by
  \begin{equation*}
    \dom g := \{\x \in \reals^d: g(\x) < +\infty   \}.
  \end{equation*}
\end{definition}
In this work we assume all extended-valued convex functions discussed are \textbf{proper}, namely the effective domain is nonempty.

Next, we formally define the concept of \emph{strict} and \emph{strong convexity}. Note that the strong convexity is parametrized by some parameter $\mu > 0$ and therefore implies strict convexity. 
\begin{definition}[Strict and Strong convexity {\cite[Definition B.1.1.1]{Hiriart-Urruty.Lemarechal-01}}]
  A convex function $g: \reals^d \to \reals \cup \{+ \infty\}$ is \textbf{strictly convex} if for any $\x_1, \x_2 \in \dom g$, for any $\alpha \in (0,1)$, it is the case that 
  \begin{equation*}
    g(\alpha \x_1 + (1- \alpha) \x_2) < \alpha g (\x_1) + (1-\alpha) g(\x_2).
  \end{equation*}
  Moreover, $g$ is $\mu$-\textbf{strongly convex} with respect to $\|\cdot\|$ norm if for any $\x_1, \x_2 \in \dom g$, for any $\alpha \in (0,1)$, it is the case that 
  \begin{equation*}
    g(\alpha \x_1 + (1- \alpha) \x_2) \leq \alpha g(\x_1) + (1- \alpha) g(\x_2) - \frac{1}{2} \mu \alpha (1-\alpha) \|\x_2 - \x_1\|^2.
  \end{equation*}
\end{definition}

The notion of \emph{convex conjugate} (a.k.a. \emph{Legendre-Fenchel transformation}) is defined as follows. The outcome of convex conjugate is always convex and closed.
\begin{definition}[Convex conjugate]
  Let $g: \reals^d \to \reals \cup \{+\infty\}$ be a convex function. The convex conjugate is defined as
  \begin{equation*}
    g^*(\y) := \sup_{\x \in \reals^d} \left\{ \left \langle \y, \x \right \rangle - g(\x) \right\}.
  \end{equation*}
\end{definition}

The following result shows that the differentiability of the conjugate function and the strict convexity of the original function is linked.
\begin{proposition}[Differentiability of the conjugate of strictly convex function {\cite[Theorem E.4.1.1]{Hiriart-Urruty.Lemarechal-01}}]
  Let $g$: $\reals^d \to \reals \cup \{+\infty \}$ be a closed, strictly convex function. 
  Then we have $\interior \dom g^* \neq \emptyset$ and $g^*$ is continuously differentiable on $\interior \dom g^*$ (where $\interior$ stands for interior). 
  
  Moreover, for $z \in \interior \dom g^*$, it is the case that
  \begin{equation*}
    \nabla g^*(\y) = \argmin_{\x} \left\{ \left\langle -\y, \x \right\rangle + g(\x)\right\}.
  \end{equation*}
  \label{conjugate:strictly:convex}
\end{proposition}

The differentiability in \cref{conjugate:strictly:convex} can be strengthened to smoothness if we further assume the strong convexity of the original function $g$.
\begin{proposition}[Smoothness of the conjugate of strongly convex function {\cite[Theorem E.4.2.1]{Hiriart-Urruty.Lemarechal-01}}]
  \label{strongly:convex:conjugate}
  Let $g: \reals^d \to \reals \cup \{+\infty \}$ be a closed, $\mu$-strongly convex function. 
  Then $g^*$ is continuously differentiable on $\reals^d$, and $g^*$ is $\frac{1}{\mu}$-smooth on $\reals^d$, namely
  $\|\nabla g^*(\y_1) - \nabla g^*(\y_2) \|_* \leq \frac{1}{\mu} \|\y_1-\y_2\|$.
\end{proposition}

Next we define the \emph{Legendre function class}. 
\begin{definition}[Legendre function {\cite[§26]{Rockafellar-70}}]  
  \label{def:legendre}
  A proper, convex, closed function $h: \reals^d \to \reals \cup \{+\infty\}$ is \textbf{of Legendre type} if 
  \begin{enumerate}
    \item [(a)] $h$ is \textbf{strictly convex}.
    \item [(b)] $h$ is \textbf{essentially smooth}, namely $h$ is differentiable on $\interior \dom h$, and $\| \nabla h(\x^{(k)} \| \to \infty$ for every sequence $\{\x^{(k)}\}_{k=0}^{\infty} \subset \interior \dom h$ converging to a boundary point of $\dom h$ as $k \to +\infty$.
  \end{enumerate}
\end{definition}

An important property of the Legendre function is the following proposition \cite{Bauschke.Borwein.ea-JCA97}.
\begin{proposition}[{\cite[Theorem 26.5]{Rockafellar-70}}]
  \label{prop:legendre}
  A convex function $g$ is of Legendre type if and only if its conjugate $g^*$ is. 
  In this case, the gradient mapping $\nabla g$ is a toplogical isomorphism with inverse mapping, namely $(\nabla g)^{-1} = \nabla g^*$.
\end{proposition}

Next, recall the definition of Bregman divergence:
\begin{definition}[Bregman divergence {\cite{Bregman-67}}]
  \label{def:bregman}
  Let $g: \reals^d \to \reals \cup \{+\infty\}$ be a closed, strictly convex function that is differentiable in $\interior \dom g$. The \textbf{Bregman divergence} $D_g(\x, \y)$ for $\x \in \dom g$, $\w \in \interior \dom g$ is defined by 
  \begin{equation*}
    D_g(\x, \w) = g(\x) - g(\w) - \left \langle \nabla g(\w), \x - \w \right \rangle.
  \end{equation*}
\end{definition}

Note the definition of Bregman divergence requires the differentiability of the base function $g$.
To extend the concept of Bregman divergence to non-differentiable function $g$, we consider the following generalized Bregman divergence (slightly modified from \cite{Flammarion.Bach-COLT17}). 
The generalized Bregman divergence plays an important role in the analysis of \feddualavg.
\begin{definition}[Generalized Bregman divergence {\cite[slightly modified from][Section B.2]{Flammarion.Bach-COLT17}}]
  \label{def:generalized_bregman}
  Let $g: \reals^d \to \reals \cup \{+\infty\}$ be a closed strictly convex function (which may not be differentiable).
  The \textbf{Generalized Bregman divergence} $\tilde{D}_{g}(\x, \y)$ for $\x \in \dom g$, $\y \in \interior \dom g^*$ is defined by 
  \begin{equation*}
    \tilde{D}_g(\x, \y) = g(\x) - g(\nabla g^*(\y)) - \left \langle \y, \x - \nabla g^*(\y) \right \rangle.
  \end{equation*}
  Note that $\nabla g^*$ is well-defined because $g^*$ is differentiable in $\interior \dom g^*$ according to \cref{conjugate:strictly:convex}.
\end{definition}

The generalized Bregman divergence is lower bounded by the ordinary Bregman divergence in the following sense. 
\begin{proposition}[{\cite[Lemma 6]{Flammarion.Bach-COLT17}}]
  \label{generalized:bregman}
  Let $h: \reals^d \to \reals \cup \{ +\infty\}$ be a Legendre function. Let $\psi: \reals^d \to \reals$ be a convex function (which may not be differentiable). Then for any $\x \in \dom h$, for any $\y \in \interior \dom (h + \psi)^*$, the following inequality holds
  \begin{equation*}
    \tilde{D}_{h + \psi}(\x, \y) \geq D_h (\x, \nabla (h + \psi)^*(\y)).
  \end{equation*}
\end{proposition}
\begin{proof}[Proof of \cref{generalized:bregman}] 
  The proof is very similar to Lemma 6 of \cite{Flammarion.Bach-COLT17}, and we include for completeness.
  By definition of the generalized Bregman divergence (\cref{def:generalized_bregman}),
  \begin{equation*}
    \tilde{D}_{h + \psi}(\x, \y) = (h + \psi)(\x) - (h + \psi)(\nabla (h+\psi)^*(\y)) - \left\langle \y, \x - \nabla (h + \psi)^*(\y) \right\rangle.
  \end{equation*}
  By definition of the (ordinary) Bregman divergence (\cref{def:bregman}),
  \begin{equation*}
    D_{h}(\x, \nabla (h + \psi)^*(\y)) = h(\x) - h( \nabla (h+\psi)^*(\y)) - \left\langle \nabla h \left( \nabla (h+ \psi)^*(\y) \right), \x - \nabla (h + \psi)^*(\y) \right\rangle.
  \end{equation*}
  Taking difference,
  \begin{align}
    & \tilde{D}_{h + \psi}(\x, \y) - D_{h}(\x, \nabla (h + \psi)^*(\y))
    \nonumber \\
    = &
    \psi(\x) - \psi \left( \nabla (h+\psi)^*(\y) \right) - \left\langle \y - \nabla h \left( \nabla (h+ \psi)^*(\y) \right), \x - \nabla (h + \psi)^*(\y) \right\rangle.
    \label{eq:generalized:bregman:1}
  \end{align}
  By \cref{conjugate:strictly:convex}, one has $\y \in \partial (h + \psi) (\nabla (h + \psi)^*(\y))$. 
  Since $h$ is differentiable in $\interior \dom h$, we have (by subgradient calculus)
  \begin{equation*}
    \y - \nabla h (\nabla (h + \psi)^*(\y)) \in \partial \psi (\nabla (h+\psi)^*(\y)).
  \end{equation*}
  Therefore, by the property of subgradient as the supporting hyperplane,
  \begin{equation}
    \psi(\x) \geq \psi ( \nabla (h + \psi)^*(\y)) + \left\langle \y - \nabla h \left( \nabla (h + \psi)^*(\y) \right) , \x - \nabla \left( h + \psi \right)^*(\y) \right\rangle
    \label{eq:generalized:bregman:2}
  \end{equation}
  Combining \cref{eq:generalized:bregman:1} and \cref{eq:generalized:bregman:2} yields
  \begin{equation*}
    \tilde{D}_{h + \psi}(\x, \y) - D_{h}(\x, \nabla (h + \psi)^*(\y)) \geq 0,
  \end{equation*}
  completing the proof.
\end{proof}

% !TEX root = main.tex  
\section{Proof of \cref{thm:1:simplified}}
\label{sec:proof:thm:1}
In this section, we provide a complete, non-asymptotic version of \cref{thm:1:simplified} with detailed proof.

We now formally state the assumptions of \cref{thm:1:simplified} for ease of reference.
\begin{assumption}[Bounded gradient]
  \;
  \label{a2} In addition to \cref{a1}, assume 
  that the gradient is $G$-uniformly-bounded, namely 
  \begin{equation*}
    \sup_{\x \in \dom \psi} \|\nabla f(\x; \xi)\|_* \leq G
  \end{equation*}
\end{assumption}
This is a standard assumption in analyzing classic distributed composite optimization \cite{Duchi.Agarwal.ea-TACON12}.

% In this section, we study the convergence of \feddualavg under \cref{a2} with unit server learning rate $\eta_{\server} = 1$.
Before we start, we introduce a few more notations to simplify the exposition and analysis throughout this section. 
Let $h_{r,k}(\x) = h(\x) +  (r K + k) \eta_{\client} \psi(\x)$.
Let $\overline{\y^{(r,k)}} := \frac{1}{M} \sum_{m=1}^M \y^{(r,k)}_m$ denote the average over clients, and $\widehat{\x^{(r,k)}} := \nabla h_{r,k}^*(\overline{\y^{(r,k)}})$ denote the primal image of $\overline{\y^{(r,k)}}$.
Formally, we use $\mathcal{F}^{(r,k)}$ to denote the $\sigma$-algebra generated by $\{\y^{(\rho,\kappa)}_{m}: \rho < r \text{ or } (\rho = r \text{ and } \kappa \leq k), m \in [M]\}$. 

\subsection{Main Theorem and Lemmas}
Now we introduce the full version of \cref{thm:1:simplified} regarding the convergence of \feddualavg with unit server learning rate $\eta_{\server} = 1$ under bounded gradient assumptions. 
\begin{theorem}[Detailed version of \cref{thm:1:simplified}]
  \label{thm:1}
  Assume \cref{a2}, then for any initialization $\x^{(0,0)} \in \dom \psi$, for unit server learning rate $\eta_{\server} = 1$ and any client learning rate $\eta_{\client} \leq \frac{1}{4L}$, \feddualavg yields
  \begin{equation}
    \expt \left[  \Phi \left(  \frac{1}{KR} \sum_{r=0}^{R-1} \sum_{k=1}^K \widehat{\x^{(r,k)}} \right) - \Phi(\x^{\star}) \right] 
    \leq
    \frac{B^2}{\eta_{\client} KR } 
    + \frac{\eta_{\client}  \sigma^2}{M}
    + 4 \eta_{\client}^2 L (K-1)^2 G^2,
    \label{eq:thm:1:1}
  \end{equation}
  where $B := \sqrt{D_h(\x^{\star}, \x^{(0,0)})}$ is the Bregman divergence between the optimal $\x^*$ and the initial $\x^{(0,0)}$.

  Particularly for 
  \begin{equation*}
      \eta_{\client} = \min \left\{ \frac{1}{4L}, \frac{M^{\frac{1}{2}} B}{\sigma K^{\frac{1}{2}} R^{\frac{1}{2}}} , \frac{B^{\frac{2}{3}}}{L^{\frac{1}{3}} K R^{\frac{1}{3}} G^{\frac{2}{3}}}  \right\},
  \end{equation*}
  one has
  \begin{equation*}
    \expt \left[  \Phi \left(  \frac{1}{KR} \sum_{r=0}^{R-1} \sum_{k=1}^K \widehat{\x^{(r,k)}} \right) - \Phi(\x^{\star}) \right]
    \leq
      \frac{4 L B^2}{KR} 
      +
      \frac{2\sigma B}{M^{\frac{1}{2}} K^{\frac{1}{2}} R^{\frac{1}{2}}}
      +
      \frac{5 L^{\frac{1}{3}} B^{\frac{4}{3}}  G^{\frac{2}{3}}}{R^{\frac{2}{3}}}.
  \end{equation*}
\end{theorem}
The proof of \cref{thm:1} is based on the two lemmas regarding perturbed convergence and stability respectively. The first lemma is \cref{pia:general} which we restate below for readers' convenience.

\PiaGeneral*

The following \cref{bounded_gradient}  bounds the stability term under the additional bounded gradient assumptions.
\begin{lemma}[Stability of \feddualavg under bounded gradient assumption]
  \label{bounded_gradient}
  In the same settings of \cref{thm:1}, it is the case that
  \begin{equation*}
    \frac{1}{M}\sum_{m=1}^M \expt \left\| \y^{(r,k)}_m - \overline{\y^{(r,k)}} \right\|_*^2 \leq 4 \eta_{\client}^2 (K-1)^2 G^2.
  \end{equation*}
\end{lemma}
We defer the proof of \cref{bounded_gradient} to \cref{sec:bounded_gradient}.
With \cref{pia:general,bounded_gradient} at hands the proof of \cref{thm:1} is immediate.
\begin{proof}[Proof of \cref{thm:1}]
  \cref{eq:thm:1:1} follows immediately from \cref{pia:general,bounded_gradient} by putting $\x = \x^{\star}$ in \cref{pia:general}. 

  Now put 
  \begin{equation*}
    \eta_{\client} =  \min \left\{ \frac{1}{4L}, 
    \frac{M^{\frac{1}{2}} B}{\sigma K^{\frac{1}{2}} R^{\frac{1}{2}}}, 
    \frac{B^{\frac{2}{3}}}{L^{\frac{1}{3}} K R^{\frac{1}{3}} G^{\frac{2}{3}}}  \right\},
  \end{equation*}
  which yields
  \begin{equation*}
    \frac{B^2}{\eta_{\client} KR} = \max \left\{ \frac{4L B^2}{KR},  
    \frac{\sigma B}{M^{\frac{1}{2}} K^{\frac{1}{2}} R^{\frac{1}{2}}},  
    \frac{L^{\frac{1}{3}} B^{\frac{4}{3}} G^{\frac{2}{3}}}{R^{\frac{2}{3}}}
       \right\}
    \leq
    \frac{4L B^2}{KR} 
    + \frac{\sigma B}{M^{\frac{1}{2}} K^{\frac{1}{2}} R^{\frac{1}{2}}}
    +   \frac{L^{\frac{1}{3}} B^{\frac{4}{3}} G^{\frac{2}{3}}}{R^{\frac{2}{3}}},
  \end{equation*}
  and
  \begin{equation*}
    \frac{\eta_{\client} \sigma^2}{2M} \leq {\frac{M^{\frac{1}{2}} B}{\sigma T^{\frac{1}{2}}}} \cdot \frac{\sigma^2}{2M} 
    =
    \frac{\sigma B}{2 M^{\frac{1}{2}} K^{\frac{1}{2}} R^{\frac{1}{2}} } , 
    \quad
     4 \eta_{\client}^2 L K^2 G^2
    \leq
    4 \left( \frac{B^{\frac{2}{3}}}{L^{\frac{1}{3}} K R^{\frac{1}{3}} G^{\frac{2}{3}}}   \right)^2 LK^2 G^2
    =
    \frac{4 L^{\frac{1}{3}} B^{\frac{4}{3}} G^{\frac{2}{3}}}{R^{\frac{2}{3}}}.
  \end{equation*}
  Summarizing the above three inequalities completes the proof of \cref{thm:1}.
\end{proof}

\subsection{Stability of \feddualavg Under Bounded Gradient Assumptions: Proof of \cref{bounded_gradient}}
\label{sec:bounded_gradient}
The proof of \cref{bounded_gradient} is straightforward given the assumption of bounded gradient and the fact that $\y^{(r,0)}_{m_1} = \y^{(r,0)}_{m_2}$ for all $m_1, m_2 \in [M]$.
\begin{proof}[Proof of \cref{bounded_gradient}]
  Let $m_1, m_2 \in [M]$ be two arbitrary clients, then
\begin{align}
  & \expt \left[  \|\y^{(r,k)}_{m_1} - \y^{(r,k)}_{m_2}\|_*^2  \middle| \mathcal{F}^{(r,0)} \right]
  \nonumber \\
  = &
  \eta_{\client}^2 \expt \left[ \left\| \sum_{\kappa=0}^{k-1} \left( \nabla f(\x^{(r,\kappa)}_{m_1}; \xi^{(r,\kappa)}_{m_1}) - \nabla f(\x^{(r,\kappa)}_{m_2}; \xi^{(r,\kappa)}_{m_2})  \right) \right\|_*^2 \middle| \mathcal{F}^{(r,0)} \right]
  \tag{since $\y^{(r,0)}_{m_1} = \y^{(r,0)}_{m_2}$}
  \\
  \leq &
  \eta_{\client}^2 \expt \left[ \left( \sum_{\kappa={0}}^{k-1}  \left\|  \nabla f(\x^{(r,\kappa)}_{m_1}; \xi^{(r,\kappa)}_{m_1})\right\|_* + \sum_{\kappa={0}}^{k-1}  \left\| \nabla f(\x^{(r,\kappa)}_{m_2}; \xi^{(r,\kappa)}_{m_2}) \right\|_* \right)^2 
  \middle| \mathcal{F}^{(r,0)} \right]
  \tag{triangle inequality of $\|\cdot\|_*$} 
  \\
  \leq &
  \eta_{\client}^2 (2 (k-1) G)^2 = 4\eta_{\client}^2 (K-1)^2 G^2.
  \nonumber
\end{align}
By convexity of $\|\cdot\|_*$,
\begin{equation*}
  \frac{1}{M}\sum_{m=1}^M \expt \left\| \y^{(r,k)}_m - \overline{\y^{(r,k)}} \right\|_*^2 
  \leq \expt \left\| \y^{(r,k)}_{m_1} - \y^{(r,k)}_{m_2} \right\|_*^2
  \leq 4 \eta_{\client}^2 (K-1)^2 G^2,
\end{equation*}
completing the proof of \cref{bounded_gradient}.
\end{proof}

% % !TEX root = main.tex  
% \section{Proof of Theorem \ref{thm:2}: Convergence of \feddualavg Under Bounded Heterogeneity and Quadratic Assumptions}
% \label{sec:proof:thm:2}
% % In this section, we study the convergence of \feddualavg under \cref{a3} (quadraticness) with unit server learning rate $\eta_{\server} = 1$.
% % We provide a complete, non-asymptotic version of \cref{thm:2} with detailed proof, which expands the proof sketch in \cref{sec:proof_sketch}. 

% % \subsection{Main Theorem and Lemmas}
% % Now we state the full version of \cref{thm:2} on \feddualavg with unit server learning rate $\eta_{\server} = 1$ under quadratic assumptions.

% % The proof of \cref{thm:2} relies on the perturbed iterate analysis \cref{pia:general} of \feddualavg and a stability bound for quadratic objectives, as stated below in \cref{quad:stability}. 
% % Note that \cref{pia:general} only assumes \cref{a1} and therefore applicable to \cref{thm:2}.
% % \hynote{former \cref{quad:stability}}

% !TEX root = main.tex  
\section{Proof of \cref{thm:0}}
\label{sec:small_lr}
In this section, we state and prove \cref{thm:0} on the convergence of \feddualavg for small client learning rate $\eta_{\client}$. 
The intuition is that for sufficiently small client learning rate, \feddualavg is almost as good as stochastic mini-batch with $R$ iterations and batch-size $MK$.
The proof technique is very similar to the above sections and \cite{Karimireddy.Kale.ea-ICML20} so we skip a substantial amount of the proof details. 
We present the proof for \feddualavg only since the analysis of \fedmid is very similar.

To facilitate the analysis we re-parameterize the hyperparameters by letting $\eta := \eta_\server \eta_{\client}$, and we treat $(\eta, \eta_{\client})$ as independent hyperparameters (rather than $(\eta_{\client}, \eta_{\server})$).
We use the notation $h_{r,k} := h + \tilde{\eta}^{(r,k)} \cdot \psi  = h + (\eta r K + \eta_{\client} k) \psi$, $\overline{\y^{(r,k)}} := \frac{1}{M} \sum_{m=1}^M \y^{(r,k)}_m$, and $\widehat{\x^{(r,k)}} := \nabla h_{r,k}^* (\overline{\y^{(r,k)}})$. 
Note that $\widehat{\x^{(r,0)}} = \x^{(r,0)}_m$ for all $m \in [M]$ by definition.
\subsection{Main Theorem and Lemmas}
Now we state the full version of \cref{thm:0} on \feddualavg with small client learning rate $\eta_{\client}$.
\begin{theorem}[Detailed version of \cref{thm:0}]
  \label{small_lr}
  Assuming \cref{a1}, then for any $\eta \in (0, \frac{1}{4KL}]$, for any initialization $\x^{(0,0)} \in \dom \psi$,
  there exists an $\eta_{\client}^{\max} > 0$ (which may depend on $\eta$ and $\x^{(0,0)}$) such that for any $\eta_{\client} \in (0, \eta_{\client}^{\max}]$, \feddualavg yields
  \begin{equation*} 
    \expt \left[ \Phi\left( \frac{1}{R}  \sum_{r=1}^{R}  \widehat{\x^{(r,0)}} \right) - \Phi(\x^{\star}) \right]
    \leq
    \frac{B^2}{\eta KR}
    +
    \frac{3 \eta \sigma^2}{M},
  \end{equation*}
  where $B := \sqrt{D_h(\x^{\star}, \x^{(0,0)})}$ is the Bregman divergence between the optimal $\x^{\star}$ and the initialization $\x^{(0,0)}$.

  In particular for 
  \begin{equation*}
    \eta = \min \left\{ \frac{1}{4KL}, \frac{B M^{\frac{1}{2}}}{K^{\frac{1}{2}} R^{\frac{1}{2}} \sigma} \right\},
  \end{equation*}
  one has
  \begin{equation*}
    \expt \left[ \Phi\left( \frac{1}{R}  \sum_{r=1}^{R}  \widehat{\x^{(r,0)}} \right) - \Phi(\x^{\star}) \right]
     \leq
    \frac{4 L B^2}{R}
    +
    \frac{4 \sigma B}{ M^{\frac{1}{2}} K^{\frac{1}{2}} R^{\frac{1}{2}}}.
  \end{equation*}
\end{theorem}

The proof of \cref{small_lr} relies on the following lemmas. 

The first \cref{small_lr:1} analyzes $\tilde{D}_{h_{r+1,0}} (\x, \overline{\y^{(r+1,0)}})$. The proof of \cref{small_lr:1} is deferred to \cref{sec:proof:small_lr:1}.
\begin{lemma}
  Under the same settings of \cref{small_lr}, the following inequality holds.
  \label{small_lr:1}
\begin{align*}
  & \tilde{D}_{h_{r+1,0}} (\x, \overline{\y^{(r+1,0)}}) -  \tilde{D}_{h_{r,0}} (\x, \overline{\y^{(r,0)}}) 
  \\
  \leq & - \tilde{D}_{h_{r,0}} ( \widehat{\x^{(r+1,0)}}, \overline{\y^{(r,0)}})  - \eta K \left( \Phi( \widehat{\x^{(r+1,0)}}) - \Phi(\x)  \right) + \frac{L}{2} \eta K  \|\widehat{\x^{(r+1,0)}}-\widehat{\x^{(r,0)}}\|^2 
  \\
  & + \eta  K 
  \left\langle  \nabla F (\widehat{\x^{(r,0)}}) - \frac{1}{MK} \sum_{m=1}^M \sum_{k=0}^{K-1} \nabla f(\x^{(r,k)}_m; \xi^{(r,k)}_m), \widehat{\x^{(r+1,0)}}- \x  \right\rangle
\end{align*}
\end{lemma}

The second lemma analyzes $\tilde{D}_{h_{r+1,0}} (\x, \overline{\y^{(r+1,0)}})$ under conditional expectation. The proof of \cref{small_lr:2} is deferred to \cref{sec:proof:small_lr:2}.
\begin{lemma}
  \label{small_lr:2}
  Under the same settings of \cref{small_lr},   there exists an $\eta_{\client}^{\max} > 0$ (which may depend on $\eta$ and $\x^{(0,0)}$) such that for any $\eta_{\client} \in (0, \eta_{\client}^{\max}]$, \feddualavg yields
  \begin{align*}
    & \expt \left[ \tilde{D}_{h_{r+1,0}} (\x, \overline{\y^{(r+1,0)}}) \middle | \mathcal{F}^{(r,0)} \right] -  \tilde{D}_{h_{r,0}} (\x, \overline{\y^{(r,0)}}) 
    \\
    \leq & 
    - \eta K \expt \left[ \left( \Phi( \widehat{\x^{(r+1,0)}}) - \Phi(\x)  \right) \middle| \mathcal{F}^{(r,0)} \right]
    + \frac{3 \eta^2 K \sigma^2}{M}.
  \end{align*}
\end{lemma}

With \cref{small_lr:1,small_lr:2} at hand we are ready to prove \cref{small_lr}.
\begin{proof}[Proof of \cref{small_lr}]
  Telescoping \cref{small_lr:2} and dropping the negative terms gives
\begin{equation*} 
  \frac{1}{R} \sum_{r=1}^{R} \expt \left[ \Phi( \widehat{\x^{(r,0)}}) - \Phi(\x) \right]
  \leq
  \frac{1}{\eta KR}\tilde{D}_{h_{r,0}} (\x, \overline{\y^{(r,0)}})
  +
  \frac{3 \eta \sigma^2}{M}
  =
  \frac{B^2}{\eta KR}
  +
  \frac{3 \eta \sigma^2}{M}.
\end{equation*}
The second inequality of \cref{small_lr} follows immediately once we plug in the specified $\eta$.
\end{proof}

\subsection{Deferred Proof of \cref{small_lr:1}}
\label{sec:proof:small_lr:1}
\begin{proof}[Proof of \cref{small_lr:1}]
The proof of this lemma is very similar to \cref{one:step:analysis:claim:1,one:step:analysis:claim:2} so we skip most of the details.

  We start by analyzing $\tilde{D}_{h_{r+1,0}}(\x, \overline{\y^{(r+1,0)}})$. 
  \begin{align}
    & \tilde{D}_{h_{r+1,0}}(\x, \overline{\y^{(r+1,0)}})
    \nonumber \\
    = & h_{r+1,0}(\x) - h_{r+1,0} \left( \nabla h_{r+1,0}^*(\overline{\y^{(r+1,0)}}) \right)
    - 
    \left\langle \overline{\y^{(r+1,0)}}, \x - \nabla h_{r+1,0}^* (\overline{\y^{(r+1,0)}})  \right\rangle 
    \tag{By definition of generalized Bregman divergence $\tilde{D}$}
    \\
    = & h_{r+1,0}(\x) - h_{r+1,0}( \widehat{\x^{(r+1,0)}}) - \left\langle \overline{\y^{(r+1,0)}}, \x - \widehat{\x^{(r+1,0)}}\right\rangle 
    \tag{By definition of $\widehat{\x^{(r+1,0)}}$}
    \\
    = & h_{r+1,0}(\x) - h_{r+1,0}( \widehat{\x^{(r+1,0)}}) - \left\langle \overline{\y^{(r,0)}} - \eta K \cdot \frac{1}{MK} \sum_{m=1}^M \sum_{k=0}^{K-1} \nabla f(\x^{(r,k)}_m; \xi^{(r,k)}_m) , \x - \widehat{\x^{(r+1,0)}} \right\rangle 
    \tag{By \feddualavg procedure}
    \\
    = & \left( h_{r,0}(\x) + \eta K \psi (\x) \right)
    - \left(h_{r,0}( \widehat{\x^{(r+1,0)}}) + \eta K \psi (\widehat{\x^{(r+1,0)}})  \right)
    \nonumber \\
    & - \left\langle \overline{\y^{(r,0)}} - \eta K \cdot \frac{1}{MK} \sum_{m=1}^M \sum_{k=0}^{K-1} \nabla f(\x^{(r,k)}_m; \xi^{(r,k)}_m) , \x - \widehat{\x^{(r+1,0)}} \right\rangle 
    \tag{By definition of $h_{r+1, 0}$}
    \nonumber \\
    = & 
    \left( h_{r,0}(\x) - h_{r,0}( \widehat{\x^{(r,0)}}) - \left\langle \overline{\y^{(r,0)}}, \x - \widehat{\x^{(r,0)}}  \right\rangle \right)
    \nonumber \\
    & - 
    \left( h_{r,0}(\widehat{\x^{(r+1,0)}}) - h_{r,0}( \widehat{\x^{(r,0)}}) - \left\langle \overline{\y^{(r,0)}},  \widehat{\x^{(r+1,0)}}  - \widehat{\x^{(r,0)}}  \right\rangle \right)
    \nonumber \\
    & - \eta K \left( \psi ( \widehat{\x^{(r+1,0)}}) -\psi(\x) \right) 
    - \eta K \left\langle \frac{1}{MK} \sum_{m=1}^M \sum_{k=0}^{K-1} \nabla f(\x^{(r,k)}_m; \xi^{(r,k)}_m) , \widehat{\x^{(r+1,0)}} - w\right\rangle 
    \tag{Rearranging}
    \\
    = & \tilde{D}_{h_{r,0}} (\x, \overline{\y^{(r,0)}}) - \tilde{D}_{h_{r,0}} ( \widehat{\x^{(r+1,0)}}, \overline{\y^{(r,0)}}) 
    - \eta K \left( \psi ( \widehat{\x^{(r+1,0)}}) -\psi(\x) \right) 
    \nonumber \\
    & - \eta K \left\langle \frac{1}{MK} \sum_{m=1}^M \sum_{k=0}^{K-1} \nabla f(\x^{(r,k)}_m; \xi^{(r,k)}_m) , \widehat{\x^{(r+1,0)}} - w\right\rangle 
    \tag{By definition of $\tilde{D}$}
  \end{align}
  By smoothness and convexity of $F$ we have
  \begin{align}
    F(\widehat{\x^{(r+1,0)}})
    \leq &
    F(\widehat{\x^{(r,0)}}) 
    + \left\langle \nabla F (\widehat{\x^{(r,0)}}), \widehat{\x^{(r+1,0)}}-\widehat{\x^{(r,0)}}  \right\rangle + \frac{L}{2} \|\widehat{\x^{(r+1,0)}}-\widehat{\x^{(r,0)}}\|^2
    \tag{by $L$-smoothness of $F$}
    \\
    \leq & 
    F(\x) 
    + \left\langle \nabla F (\widehat{\x^{(r,0)}}), \widehat{\x^{(r+1,0)}}- \x  \right\rangle + \frac{L}{2} \|\widehat{\x^{(r+1,0)}}-\widehat{\x^{(r,0)}}\|^2
    \tag{by convexity of $F$}
  \end{align}
  Combining the above two (in)equalities gives
  \begin{align*}
    & \tilde{D}_{h_{r+1,0}} (\x, \overline{\y^{(r+1,0)}}) -  \tilde{D}_{h_{r,0}} (\x, \overline{\y^{(r,0)}}) 
    \\
    \leq & - \tilde{D}_{h_{r,0}} ( \widehat{\x^{(r+1,0)}}, \overline{\y^{(r,0)}})  - \eta K \left( \Phi( \widehat{\x^{(r+1,0)}}) - \Phi(\x)  \right) + \frac{L}{2} \eta K  \|\widehat{\x^{(r+1,0)}}-\widehat{\x^{(r,0)}}\|^2 
    \\
    & + \eta K 
    \left\langle  \nabla F (\widehat{\x^{(r,0)}}) - \frac{1}{MK} \sum_{m=1}^M \sum_{k=0}^{K-1} \nabla f(\x^{(r,k)}_m; \xi^{(r,k)}_m), \widehat{\x^{(r+1,0)}}- \x  \right\rangle.
  \end{align*}
  \end{proof}

\subsection{Deferred Proof of \cref{small_lr:2}}
  \label{sec:proof:small_lr:2}
\begin{proof}[Proof of \cref{small_lr:2}]
We start by splitting the inner product term in the inequality of \cref{small_lr:1}:
\begin{align*}
  & \left\langle  \nabla F (\widehat{\x^{(r,0)}}) - \frac{1}{MK} \sum_{m=1}^M \sum_{k=0}^{K-1} \nabla f(\x^{(r,k)}_m; \xi^{(r,k)}_m), \widehat{\x^{(r+1,0)}}- \x  \right\rangle
  \\
  = &
  \underbrace{
  \left\langle  \nabla F (\widehat{\x^{(r,0)}}) - \frac{1}{MK} \sum_{m=1}^M \sum_{k=0}^{K-1} \nabla f(\widehat{\x^{(r,0)}}; \xi^{(r,k)}_m), \widehat{\x^{(r,0)}}- \x  \right\rangle}_{\text{(I)}}
  \\ &
   + 
  \underbrace{
    \left\langle  \nabla F (\widehat{\x^{(r,0)}}) - \frac{1}{MK} \sum_{m=1}^M \sum_{k=0}^{K-1} \nabla f(\widehat{\x^{(r,0)}}; \xi^{(r,k)}_m), \widehat{\x^{(r+1,0)}}-\widehat{\x^{(r,0)}}  \right\rangle}_{\text{(II)}}
  \\ &
  +
  \underbrace{\frac{1}{MK} \sum_{m=1}^M \sum_{k=0}^{K-1} 
  \left\langle  \nabla f(\widehat{\x^{(r,0)}}; \xi^{(r,k)}_m) - \nabla f(\x^{(r,k)}_m; \xi^{(r,k)}_m), \widehat{\x^{(r+1,0)}}- \x  \right\rangle}_{\text{(III)}}.
\end{align*}
Now we investigate the terms (I)-(III). 
By conditional independence we know $\expt[\text{(I)} | \mathcal{F}^{(r,0)}] = 0$. For (II), we know that
\begin{align*}
 & \expt \left[ \text{(II)} \middle| \mathcal{F}^{(r,0)} \right] 
 \\
 \leq &
 \expt \left[ \left\|\nabla F (\widehat{\x^{(r,0)}}) - \frac{1}{MK} \sum_{m=1}^M \sum_{k=0}^{K-1} \nabla f(\widehat{\x^{(r,0)}}; \xi^{(r,k)}_m) \right\|_*\middle | \mathcal{F}^{(r,0)} \right]
 \expt \left[  \left\|  \widehat{\x^{(r+1,0)}}-\widehat{\x^{(r,0)}}   \right\|\middle| \mathcal{F}^{(r,0)} \right]
 \\
 \leq & 
 \frac{\sigma}{\sqrt{MK}} \cdot \expt \left[  \left\|  \widehat{\x^{(r+1,0)}}-\widehat{\x^{(r,0)}}   \right\|\middle| \mathcal{F}^{(r,0)} \right]
\end{align*}
For (III) we observe that (by smoothness assumption)
\begin{align*}
  \text{(III)} \leq &  \frac{1}{MK} \sum_{m=1}^M \sum_{k=0}^{K-1}  \left\| \nabla f(\widehat{\x^{(r,0)}}; \xi^{(r,k)}_m) - \nabla f(\x^{(r,k)}_m; \xi^{(r,k)}_m) \right\|_*  \|\widehat{\x^{(r+1,0)}} - \x\| 
  \\
  \leq & \frac{L}{MK} \sum_{m=1}^M \sum_{k=0}^{K-1}  \left\| \widehat{\x^{(r,0)}} - \x^{(r,k)}_m \right\|  \|\widehat{\x^{(r+1,0)}} - \x\|.
\end{align*}
Taking conditional expectation,
\begin{align*}
  & \expt \left[ \text{(III)} \middle| \mathcal{F}^{(r,0)} \right] 
  \\
  \leq &  
  \frac{1}{MK} \sum_{m=1}^M \sum_{k=0}^{K-1}  \expt \left[ \left\| \nabla f(\widehat{\x^{(r,0)}}; \xi^{(r,k)}_m) - \nabla f(\x^{(r,k)}_m; \xi^{(r,k)}_m) \right\|_* \middle| \mathcal{F}^{(r,0)} \right]
  \expt \left[ \|\widehat{\x^{(r+1,0)}} - \x\| \middle| \mathcal{F}^{(r,0)} \right]
  \\ 
  \leq & \frac{L}{MK} \left(  \sum_{m=1}^M \sum_{k=0}^{K-1} \expt \left[ \|\widehat{\x^{(r,0)}} - \x^{(r,k)}_m \| \middle| \mathcal{F}^{(r,0)} \right] \right) 
  \expt \left[ \|\widehat{\x^{(r+1,0)}} - \x\| \middle| \mathcal{F}^{(r,0)} \right]
  \\
\end{align*}
Combining the above inequalities with \cref{small_lr:1} gives
\begin{align*}
   & \expt \left[ \tilde{D}_{h_{r+1,0}} (\x, \overline{\y^{(r+1,0)}}) \middle | \mathcal{F}^{(r,0)} \right] -  \tilde{D}_{h_{r,0}} (\x, \overline{\y^{(r,0)}}) 
   \\
   \leq & 
   - \eta K \expt \left[ \left( \Phi( \widehat{\x^{(r+1,0)}}) - \Phi(\x)  \right) \middle| \mathcal{F}^{(r,0)} \right]
  - \left( \frac{1}{2} - \frac{L}{2} \eta K  \right) \expt \left[  \|\widehat{\x^{(r+1,0)}}-\widehat{\x^{(r,0)}}\|^2  \middle| \mathcal{F}^{(r,0)} \right]
   \\ & + \frac{\eta \sigma \sqrt{K}}{\sqrt{M}}  \cdot \expt \left[  \left\|  \widehat{\x^{(r+1,0)}}-\widehat{\x^{(r,0)}}   \right\|\middle| \mathcal{F}^{(r,0)} \right] 
   \\ & +  \frac{\eta L}{M} \left(  \sum_{m=1}^M \sum_{k=0}^{K-1} \expt \left[ \|\widehat{\x^{(r,0)}} - \x^{(r,k)}_m \| \middle| \mathcal{F}^{(r,0)} \right] \right) 
   \expt \left[ \|\widehat{\x^{(r+1,0)}} - \x\| \middle| \mathcal{F}^{(r,0)} \right]
 \end{align*}
 Note that 
 \begin{align}
  & - \left( \frac{1}{2} - \frac{L}{2} \eta K  \right) \expt \left[  \|\widehat{\x^{(r+1,0)}}-\widehat{\x^{(r,0)}}\|^2  \middle| \mathcal{F}^{(r,0)} \right]
  + \frac{\eta \sigma \sqrt{K}}{\sqrt{M}} \cdot \expt \left[  \left\|  \widehat{\x^{(r+1,0)}}-\widehat{\x^{(r,0)}}   \right\|\middle| \mathcal{F}^{(r,0)} \right] 
  \nonumber \\
  \leq & - \frac{3}{8} \expt \left[  \|\widehat{\x^{(r+1,0)}}-\widehat{\x^{(r,0)}}\|^2  \middle| \mathcal{F}^{(r,0)} \right]
  + \frac{\eta \sigma \sqrt{K}}{\sqrt{M}} \cdot \expt \left[  \left\|  \widehat{\x^{(r+1,0)}}-\widehat{\x^{(r,0)}}   \right\|\middle| \mathcal{F}^{(r,0)} \right]    \tag{since $\eta \leq \frac{1}{4KL}$}
  \\
  \leq & - \frac{1}{4} \expt \left[  \|\widehat{\x^{(r+1,0)}}-\widehat{\x^{(r,0)}}\|^2  \middle| \mathcal{F}^{(r,0)} \right] + \frac{2\eta^2 K \sigma^2}{M}.
  \tag{by quadratic optimum}
 \end{align}
 Therefore
 \begin{align*}
  & \expt \left[ \tilde{D}_{h_{r+1,0}} (\x, \overline{\y^{(r+1,0)}}) \middle | \mathcal{F}^{(r,0)} \right] -  \tilde{D}_{h_{r,0}} (\x, \overline{\y^{(r,0)}}) 
  \\
  \leq & 
  - \eta K \expt \left[ \left( \Phi( \widehat{\x^{(r+1,0)}}) - \Phi(\x)  \right) \middle| \mathcal{F}^{(r,0)} \right]
  - \frac{1}{4} \expt \left[  \|\widehat{\x^{(r+1,0)}}-\widehat{\x^{(r,0)}}\|^2  \middle| \mathcal{F}^{(r,0)} \right] 
    + \frac{2 \eta^2 K \sigma^2}{M} 
  \\
  & +  \frac{\eta L}{M} \left(  \sum_{m=1}^M \sum_{k=0}^{K-1} \expt \left[ \|\widehat{\x^{(r,0)}} - \x^{(r,k)}_m \| \middle| \mathcal{F}^{(r,0)} \right] \right) 
  \expt \left[ \|\widehat{\x^{(r+1,0)}} - \x\| \middle| \mathcal{F}^{(r,0)} \right].
\end{align*}
Since $\x^{(r,k)}_m$ is generated by running local dual averaging with learning rate $\eta_{\client}$, one has
\begin{equation*}
  \lim_{\eta_{\client} \downarrow 0} \left[ \left(  \sum_{m=1}^M \sum_{k=0}^{K-1} \expt \left[ \|\widehat{\x^{(r,0)}} - \x^{(r,k)}_m \| \middle| \mathcal{F}^{(r,0)} \right] \right) 
  \expt \left[ \|\widehat{\x^{(r+1,0)}} - \x\| \middle| \mathcal{F}^{(r,0)} \right] \right] = 0.
\end{equation*}
There exists an upper bound $\eta_{\client}^{\max}$ such that for any $\eta_{\client} \in (0, \eta_{\client}^{\max}]$, it is the case that 
\begin{equation*}
   \left(  \sum_{m=1}^M \sum_{k=0}^{K-1} \expt \left[ \|\widehat{\x^{(r,0)}} - \x^{(r,k)}_m \| \middle| \mathcal{F}^{(r,0)} \right] \right) 
  \expt \left[ \|\widehat{\x^{(r+1,0)}} - \x\| \middle| \mathcal{F}^{(r,0)} \right] 
  \leq
  \frac{\eta K \sigma^2}{L}.
\end{equation*}
Therefore, for any $\eta_{\client} \in (0, \eta_{\client}^{\max}]$,
\begin{align*}
   & \expt \left[ \tilde{D}_{h_{r+1,0}} (\x, \overline{\y^{(r+1,0)}}) \middle | \mathcal{F}^{(r,0)} \right] -  \tilde{D}_{h_{r,0}} (\x, \overline{\y^{(r,0)}}) 
   \\
  \leq &
  - \eta K \expt \left[ \Phi( \widehat{\x^{(r+1,0)}}) - \Phi(\x)  \middle| \mathcal{F}^{(r,0)} \right]
  + \frac{3 \eta^2 K \sigma^2}{M}.
\end{align*}
\end{proof}

\bibliography{thesis}

\end{document}